\documentclass{book}

\newif\ifdigital
\digitaltrue

\ifdigital
  \usepackage[a4paper]{geometry}
\else
  \usepackage{geometry}
\fi

\usepackage[T1]{fontenc}
\usepackage[utf8]{inputenc}
\usepackage[english]{babel}

\usepackage{newtxtext}

\usepackage{mathtools} 
\usepackage{amsthm}

\usepackage[nonewtxmathopt]{newtxmath} 
\usepackage[cal=cm,bb=ams,frak=pxtx]{mathalpha}

\PassOptionsToPackage{hyphens}{url}
\usepackage[hyphens]{url}
\usepackage[hidelinks]{hyperref}

\usepackage{graphicx}
\usepackage{caption}
\usepackage{tikz}
\usetikzlibrary{hobby,decorations.markings}
\usepackage{pgfplots}
\pgfplotsset{compat=1.17}

\usepackage{booktabs}
\usepackage{microtype}
\usepackage{xcolor}
\usepackage{enumitem}
\setitemize{itemsep=0pt,parsep=0pt,topsep=0pt}
\setenumerate{itemsep=0pt,parsep=0pt,topsep=0pt}
\usepackage[ruled,vlined]{algorithm2e}

\usepackage{xfrac}

\usepackage[sectionbib,authoryear,round]{natbib}
\usepackage{appendix}
\usepackage{psl-cover}

\DeclareMathOperator{\PP}{\mathbb{P}}
\newcommand{\Prob}[1]{\Delta_{#1}}

\newcommand{\card}[1]{\left\vert #1 \right\vert}
\newcommand{\diff}{\mathop{}\!\mathrm{d}}
\newcommand{\ind}[1]{\mathbf{1}_{#1}}
\newcommand{\prob}[1]{\Delta_{#1}}
\DeclareMathOperator{\E}{\mathbb{E}}
\DeclareMathOperator{\Pbb}{\mathbb{P}}
\DeclareMathOperator*{\argmax}{arg\,max}
\DeclareMathOperator*{\argmin}{arg\,min}

\newcommand{\uniform}[1]{{\cal U}\paren{#1}}
\DeclareMathOperator{\Div}{div}
\DeclareMathOperator{\Span}{Span}

\DeclareMathOperator{\hull}{Conv}
\DeclareMathOperator{\ima}{im}
\DeclareMathOperator{\op}{op}
\DeclareMathOperator{\sign}{sign}
\DeclareMathOperator{\spec}{spec}
\DeclareMathOperator*{\supp}{supp}
\DeclareMathOperator{\trace}{Tr}

\newcommand{\C}{\mathbb{C}}
\newcommand{\N}{\mathbb{N}}
\newcommand{\Q}{\mathbb{Q}}
\newcommand{\R}{\mathbb{R}}
\newcommand{\Z}{\mathbb{Z}}

\newcommand{\bbracket}[1]{\left\llbracket #1 \right\rrbracket}
\renewcommand{\brace}[1]{\left\{ #1 \right\}}
\newcommand{\bracket}[1]{\left[ #1 \right]}
\newcommand{\floor}[1]{\left\lfloor #1 \right\rfloor}
\newcommand{\ceil}[1]{\left\lceil #1 \right\rceil}
\newcommand{\paren}[1]{\left( #1 \right)}
\newcommand{\midvert}{\,\middle\vert\,}
\newcommand{\midvertvert}{\,\middle\vert\vert\,}

\newcommand{\abs}[1]{\left| #1 \right|}
\newcommand{\module}[1]{\left| #1 \right|}
\newcommand{\norm}[1]{\left\| #1 \right\|}
\newcommand{\scap}[2]{\left\langle #1, #2 \right\rangle}

\newcommand{\Sfrak}{\mathfrak{S}}
\newcommand{\X}{\mathcal{X}}
\newcommand{\Y}{\mathcal{Y}}

\renewcommand{\epsilon}{\varepsilon}
\renewcommand{\phi}{\varphi}

\theoremstyle{plain}
\newtheorem{theorem}{Theorem}

\newtheorem{lemma}{Lemma}
\newtheorem{proposition}[lemma]{Proposition}
\newtheorem{definition}[lemma]{Definition}
\newtheorem{assumption}{Assumption}
\newtheorem{remark}[lemma]{Remark}
\newtheorem{example}{Example}

\newcommand{\minimize}[2]{
\begin{array}{rl}
	\text{minimize } & #1 \\
	\text{subject to } & #2 
\end{array}
}

\newcommand{\myfunction}[5]{
\begin{array}{cccc}
	#1 : & #2 & \rightarrow & #3 \\
	 & #4 & \rightarrow & #5
\end{array}
}

\title{From Weakly Supervised Learning to Active Labeling}
\author{Vivien Cabannes}
\date{18/07/2022}

\jurymember{1}{Marc Lelarge}{INRIA / DI ENS / PSL}{Président}
\jurymember{2}{Eyke H\"ullermeier}{Paderborn University}{Rapporteur}
\jurymember{3}{Guillaume Lecu\'e}{ENSAE}{Rapporteur}
\jurymember{4}{Joan Bruna}{NYU}{Examinateur}
\jurymember{5}{Francis Bach}{INRIA / DI ENS / PSL}{Directeur de thèse}
\jurymember{6}{Alessandro Rudi}{INRIA / DI ENS / PSL}{Directeur de thèse}

\frabstract{
	Les mathématiques appliquées et le calcul nourrissent beaucoup d'espoirs à la suite des succès récents de l'apprentissage supervisé.
	Dans l'industrie, beaucoup d'ingénieurs cherchent à remplacer leurs anciens paradigmes de pensée par l'apprentissage machine.
	Étonnamment, ces ingénieurs passent plus de temps à collecter, annoter et nettoyer des données qu'à raffiner des modèles.
	Ce phénomène motive la problématique de cette thèse: peut-on définir un cadre théorique plus général que l'apprentissage supervisé pour apprendre grâce à des données hétérogènes?
	Cette question est abordée via le concept de supervision faible, faisant l'hypothèse que le problème que posent les données est leur annotation.
	On modélise la supervision faible comme l'accès, pour une entrée donnée, non pas d'une sortie claire, mais d'un ensemble de sorties potentielles.
  On plaide pour l'adoption d'une perspective \guillemotleft~optimiste~\guillemotright\ et l'apprentissage d'une fonction qui vérifie la plupart des observations.
	Cette perspective nous permet de définir un principe pour lever l'ambiguïté des informations faibles.
	On discute également de l'importance d'incorporer des techniques sans supervision d'appréhension des données d'entrée dans notre théorie, en particulier de compréhension de la variété sous-jacente via des techniques de diffusion, pour lesquelles on propose un algorithme réaliste afin d'éviter le fléau de la dimension, à l'inverse de ce qui existait jusqu'alors.
	Enfin, nous nous attaquons à la question de collecte active d'informations faibles, définissant le problème de \guillemotleft~catalogage en ligne~\guillemotright, où un intendant doit acquérir une maximum d'informations fiables sur ses données sous une contrainte de budget.
	Entre autres, nous tirons parti du fait que pour obtenir un gradient stochastique et effectuer une descente de gradient, il n'y a pas besoin de supervision totale.
}

\enabstract{
  Applied mathematics and machine computations have raised a lot of hope since the recent success of supervised learning.
  Many practitioners in industries have been trying to switch from their old paradigms to machine learning.
  Interestingly, those data scientists spend more time scrapping, annotating and cleaning data than fine-tuning models.
  This thesis is motivated by the following question: can we derive a more generic framework than the one of supervised learning in order to learn from clutter data?
  This question is approached through the lens of weakly supervised learning, assuming that the bottleneck of data collection lies in annotation.
  We model weak supervision as giving, rather than a unique target, a set of target candidates.
  We argue that one should look for an ``optimistic'' function that matches most of the observations.
  This allows us to derive a principle to disambiguate partial labels.
  We also discuss the advantage to incorporate unsupervised learning techniques into our framework, in particular manifold regularization approached through diffusion techniques, for which we derived a new algorithm that scales better with input dimension then the baseline method.
  Finally, we switch from passive to active weakly supervised learning, introducing the ``active labeling'' framework, in which a practitioner can query weak information about chosen data.
  Among others, we leverage the fact that one does not need full information to access stochastic gradients and perform stochastic gradient descent.
}

\frkeywords{Apprentissage statistique. Donn\'ees faiblement supervis\'ees. Acquisition active d'informations partielles.}
\enkeywords{Statistical learning. Weakly supervised learning; partial supervision. Active labeling; weak queries.}

\begin{document}
\ifdigital
	\maketitle
\fi

\thispagestyle{empty}
\begin{flushright}
	\vspace*{\fill}
	\Large\it\`A Kwama,
	\vspace*{\fill}
\end{flushright}

\clearpage

\thispagestyle{empty}
\vspace*{\fill}
\begin{verbatim}
@PhDThesis{Cabannes2022,
  author    = {Vivien Cabannes},
  title     = {From Weakly Supervised Learning to Active Labeling},
  year      = 2022,
  school    = {\'Ecole Normale Sup\'erieure},
  type      = {PhD Thesis}
}
\end{verbatim}
\clearpage
\tableofcontents

\chapter{Forewords}

\section{Acknowledgement}

This thesis would not have been the same without the brilliant minds of Francis Bach and Alessandro Rudi, who have constantly impressed me by the speed at which they formed deep thoughts on any subject I would bring up in meetings.
I consider myself really fortunate to have worked under your direction.
Beside your sagacity, you have been a great source of inspiration as a person: always wise and caring, even when I was wondering about life in uncalled-for fashions, I was abusing academic freedom, or I was claiming false statements.
My second thoughts go for my colleagues at INRIA Paris.
I am particularly grateful to Alex Nowak for our many conversations around structured prediction, and to Yann Labb\'e for his continuous support.
My last year as a PhD student has been highly enjoyable, which correlates with the arrival of Quentin Le Lidec in my office.
Finally, I would like to warmly thank Guillaume Lecu\'e and Eyke H\"ullermeier for having accepted to review this manuscript, and especially considering how valuable your research hence your time is.
In particular, Eyke, you were the first person to give me feedback on my work, which I have experienced as an accolade that marked my entry in the research world.

Une thèse de doctorat représente un accomplissement académique certain.
Ma reconnaissance s'exprime également envers toutes les entités et les personnes extérieures qui m'ont permis d'y accéder.
Tout d'abord tous les professeurs qui ont su me transmettre leur enthousiasme, en particulier Xavier Lacroix et ses merveilleuses digressions.
Rétrospectivement, je m'étonne du dévouement et de la pédagogie remarquable d'un grand nombre d'entre eux, l'exemple le plus frappant étant celui d'Alain Camanès qui a su m'enseigner la rigueur nécessaire pour entrer dans le monde des sciences du supérieur.
Enfin, mon parcours n'aurait pas été le même sans la bienveillance d'un certain nombre de personnes qui ont cru en moi plus que moi-même n'y croyait, me laissant nombre de vifs souvenirs.
Remercions également les institutions qui font de leur mieux pour permettre ce genre d'entreprises, notamment en les finançant.
En particulier, cette thèse a été financée par le gouvernement français via le programme \guillemotleft~Investissements d'avenir~\guillemotright\ de l'Agence Nationale de la Recherche (référence ANR-19-P3IA-0001).
Francis et Alessandro sont également soutenus financièrement par l'European Research Council (bourses SEQUOIA 724063 et REAL 94790).

Cet accomplissement s'inscrit également dans une histoire plus intime.
Celle-ci s'écrit depuis bien avant ma naissance, et si je peux ici ressusciter les morts pour les remercier, je le ferai pour l'importance qu'ils ont porté à l'éveil, l'instruction et la curiosité, des valeurs qu'à leur suite, mon père et ma grand-mère maternelle ont tout particulièrement souhaite m'inculquer.
Je dois également à ma famille de précieux sentiments, nourris par le souvenir, qui me fournissent souvent les aspirations nécessaires pour repartir quand la difficulté se fait trop présente en moi.
J'espère qu'en retour, je participe un peu au bonheur et au développement de chacun.
Si ce court texte ne permet pas de remercier personnellement toutes les personnes qui m'importent, je n'oublierai ni mon papy qui m'a hébergée pendant ces trois dernières années, ni ma mère qui s'est occupée de moi si longtemps.
À la famille s'ajoutent mes amis.
J'aimerai les remercier tous, qui sont des éléments constitutifs de ma personne.
En particulier, je citerai Thomas Kerdreux pour sa magnanimité et Charles Arnal pour ses pertinents conseils.
Tous deux ont une vigueur intellectuelle et physique qui ne cesse de m'impressionner.

\subsection{Résumé}
	Les mathématiques appliquées et le calcul nourrissent beaucoup d'espoirs à la suite des succès récents de l'apprentissage supervisé.
	Dans l'industrie, beaucoup d'ingénieurs cherchent à remplacer leurs anciens paradigmes de pensée par l'apprentissage machine.
	Étonnamment, ces ingénieurs passent plus de temps à collecter, annoter et nettoyer des données qu'à raffiner des modèles.
	Ce phénomène motive la problématique de cette thèse: peut-on définir un cadre théorique plus général que l'apprentissage supervisé pour apprendre grâce à des données hétérogènes?
	Cette question est abordée via le concept de supervision faible, faisant l'hypothèse que le problème que posent les données est leur annotation.
	On modélise la supervision faible comme l'accès, pour une entrée donnée, non pas d'une sortie claire, mais d'un ensemble de sorties potentielles.
  On plaide pour l'adoption d'une perspective \guillemotleft~optimiste~\guillemotright\ et l'apprentissage d'une fonction qui vérifie la plupart des observations.
	Cette perspective nous permet de définir un principe pour lever l'ambiguïté des informations faibles.
	On discute également de l'importance d'incorporer des techniques sans supervision d'appréhension des données d'entrée dans notre théorie, en particulier de compréhension de la variété sous-jacente via des techniques de diffusion, pour lesquelles on propose un algorithme réaliste afin d'éviter le fléau de la dimension, à l'inverse de ce qui existait jusqu'alors.
	Enfin, nous nous attaquons à la question de collecte active d'informations faibles, définissant le problème de \guillemotleft~catalogage en ligne~\guillemotright, où un intendant doit acquérir une maximum d'informations fiables sur ses données sous une contrainte de budget.
	Entre autres, nous tirons parti du fait que pour obtenir un gradient stochastique et effectuer une descente de gradient, il n'y a pas besoin de supervision totale.

\subsection{Abstract}
  Applied mathematics and machine computations have raised a lot of hope since the recent success of supervised learning.
  Many practitioners in industries have been trying to switch from their old paradigms to machine learning.
  Interestingly, those data scientists spend more time scrapping, annotating and cleaning data than fine-tuning models.
  This thesis is motivated by the following question: can we derive a more generic framework than the one of supervised learning in order to learn from clutter data?
  This question is approached through the lens of weakly supervised learning, assuming that the bottleneck of data collection lies in annotation.
  We model weak supervision as giving, rather than a unique target, a set of target candidates.
  We argue that one should look for an ``optimistic'' function that matches most of the observations.
  This allows us to derive a principle to disambiguate partial labels.
  We also discuss the advantage to incorporate unsupervised learning techniques into our framework, in particular manifold regularization approached through diffusion techniques, for which we derived a new algorithm that scales better with input dimension then the baseline method.
  Finally, we switch from passive to active weakly supervised learning, introducing the ``active labeling'' framework, in which a practitioner can query weak information about chosen data.
	Among others, we leverage the fact that one does not need full information to access stochastic gradients and perform stochastic gradient descent.

\section{Summary of contributions}

\begin{enumerate}
    \item The main contribution of this thesis is to provide a generic framework to deal with partial supervision.
        Along this framework, we provide a consistent algorithm for any structured prediction problem \citep{Cabannes2020}; a principled algorithm to disambiguate weak information into full supervision which go beyond the partial supervision setting \citep{Cabannes2021}, as well as a statistically and computationally efficient way to incorporate Laplacian regularization \citep{Cabannes2021c}.
    \item We provide a method to derive fast rates for any discrete output problem \citep{Cabannes2021b} based on least-squares surrogate. Although geared toward least-squares surrogate with Tikhonov regularization, those derivations can easily be extended to self-concordant losses based on \cite{MarteauFerey2019} and to any spectral filtering techniques based on \cite{Lin2020}.
    We hope that this could be useful for statistical learning people, and we wish to see in the future similar results for other losses, as well as a better theory to compare surrogate problems for classification.
        As a first step in this direction, we provide some derivations for the hinge loss \citep{Cabannes2022b}.
    \item We provide a statistically and computationally efficient method to approach Laplacian regularization \citep{Cabannes2021c}.
    Our method could be useful for a vast number of applications. Our low-rank approximation could help scale up exciting methods for sampling based on Langevin dynamics \citep{PillaudVivien2020}, Gaussian processes with derivatives \citep{Eriksson2018}, sparse models based on regularization with derivatives \citep{Rosasco2013}.
    The fact that we ``kernelized'' Laplacian regularization, making it statistically superior to existing local averaging methods, is exciting when considering the impact of ``local'' methods such as diffusion maps \citep{Coifman2006}.
    \item We introduce the active labeling problem, which unify several problems of searching for information through imprecise query, and might be useful to deal with privacy issues.
			We provide a stochastic gradient descent method that does not require full supervision to tackle this active weakly supervised learning problem that we formalized in \cite{Cabannes2022}.
\end{enumerate}

\subsection{List of publications}
On supervised learning, reproduced in Part \ref{part:discrete}.
\begin{itemize}
	\item \citep{Cabannes2021b} Vivien Cabannes, Alessandro Rudi, and Francis Bach. Fast rates in structured prediction. In {\em Conference on Learning Theory}, 2021.
	\item \citep{Cabannes2022b} Vivien Cabannes and Stefano Vigogna. A case of exponential convergence rates for SVM. {\em In preparation} 2022.
\end{itemize}

\noindent
On partial supervision, reproduced in Parts \ref{part:weakly} and \ref{part:collection}.
\begin{itemize}
	\item \citep{Cabannes2020} Vivien Cabannes, Alessandro Rudi, and Francis Bach. Structured prediction with partial labeling through the infimum loss. In {\em International Conference on Machine Learning}, 2020.
	\item \citep{Cabannes2021c} Vivien Cabannes, Alessandro Rudi, and Francis Bach. Disambiguation of weak supervision with exponential convergence rates. In {\em International Conference on Machine Learning}, 2021.
	\item \citep{Cabannes2021c} Vivien Cabannes, Loucas Pillaud-Vivien, Francis Bach, and Alessandro Rudi. Overcoming the curse of dimensionality with Laplacian regularization in semi-supervised learning. In {\em Neural Information Processing Systems}, 2021.
	\item \citep{Cabannes2022} Vivien Cabannes, Francis Bach, Vianney Perchet, and Alessandro Rudi. Active labeling: streaming stochastic gradients. {\em In preparation} 2022.
\end{itemize}

\noindent
Some artistic reflections formatted for workshops (not reproduced).
\begin{itemize}
  \item \citep{Cabannes2019} Vivien Cabannes, Thomas Kerdreux, Louis Thiry, and Tina \& Charly. Dialog on a canvas with a machine. In {\em NeurIPS workshop on Creativity}, 2019.
	\item \citep{Cabannes2020b} Vivien Cabannes, Thomas Kerdreux, Louis Thiry. Diptychs of human and machine perception. {\em In NeurIPS workshop on Creativity}, 2020.
\end{itemize}

\noindent
Side works done for French reviews (not reproduced).
\begin{itemize}
	\item Vivien Cabannes. Perquisition de données sur le cloud : les Etats-Unis en avance, l’Europe à la traîne. In {\em le Grand Continent}, 2019.
	\item \citep{Cabannes2022c} Vivien Cabannes. Le futur du numérique sera-t-il incarné ? {\em Esprit}, 487:117-125, 2022.
\end{itemize}

\section{Research chronology}

This section traces back some of my research history, before summarizing the contributions of this thesis.
I wish it to be useful in order to understand the thought process behind this thesis and to ease its reading.

The thesis originated from the motivations of my supervisors to confront the interest of Francis for weak supervision with the least-squares surrogate method of Alessandro.
My interest was sparked by the fact that I wondered how to better incorporate fuzzy human knowledge into learning processes.
Many abstractions can be learned in a hierarchical fashion, {\em e.g.} first recognizing edges on input images, then aggregating edges into tires, doors, handles, and those last components into cars, fridges, bridges, before understanding the semantic of an image and outputting a wanted label.
Often, we have a good understanding of this hierarchical structure, but it is hard to formulate it clearly in order to guide the learning process.
In this thesis, we focus on weak supervision, which is understood as specifying crucial elements in the last layer of abstraction before labels, such as specifying some animal attributes that allow us to discriminate the animals that our algorithm should learn to recognize.
Other priors on the problem to solve are assumed to be incorporated into the model or into regularizers.
Note that priors are distinct from supervision in the sense that they are given once and for all, while supervision can be refined ({\em e.g.} samples $(X_i, Y_i)_{i\leq n}$ are a function of $n\in\N$ that can be made larger).

From a theoretical point of view, successes of machine learning are understood as successes of supervised learning, which are successes of statistics when given enough clean data.
Yet, in many problems, accessing tidy data is too costly, so many heuristics have been developed to deal with settings that do not completely fit in the supervised learning setting.
The goal of this thesis is to ground those heuristics into principles.
To make those principles clear and rigorous, this work finds its roots in the statistical learning theory.
In particular, we assume that there is a unique metric to quantify the quality of a learning algorithm, and that this metric is specified by a loss function.

The first task was to define the setup to work on.
Many directions could have been taken.
For example, supervision can come as experts specifying what is their guess for the label, and what is their levels of certitude.
To avoid dealing with fuzziness, we decided to assume that our supervision provides sure information that allows us to locate the label into a set of label candidates, assuming that errors on the supervision could be captured by the randomness of the label conditionally to the input.
This generic yet interesting framework turns out to have already been studied in the literature under the name of partial labeling as well as superset learning.
In a first paper \citep{Cabannes2020}, we described this framework, advocate for combining weak information in a parsimonious way and came up with some theoretical results to strengthen our point.
We put on top the framework of Alessandro to exhibit a consistent algorithm with decent convergence rates.
While we initially wanted to compare this principle with several heuristics and associated principles, we soon realized that this task was endless, and we reduced the experimental comparison to ``comparable'' principles.

Several indications point us toward a second paper \citep{Cabannes2021}.
First of all, the algorithm we came up with was dealing with a really big related surrogate problem, hard to understand, hiding in ugly constants in our convergence rates.
Fundamentally, we were dealing with functions from input to set of labels, and were paying the price of the combinatorial structure of power sets.
To stay in the space of functions from input to labels, we decided to explicitly retrieve labels from weak supervision before learning with a fully supervised dataset.
To come up with a clear empirical objective, we thought in terms of distributions rather than functions, leveraging the kernel mean embedding structure beyond Alessandro framework.
However, theoretical guarantees were much harder to obtain, missing generic tools to understand deviation between empirical principle and population principle when these principles are expressed on distributions.
This forces us to incorporate assumptions to ease the theoretical understanding.
Surprisingly, those assumptions were not that strong and led to much better convergence rates, and this was not only the case for our problem but for the whole framework of Alessandro.
We published this interesting result in a different paper \citep{Cabannes2021b}, and left open research questions on how far we can push those derivations.
We came back to those questions recently, extending the proof mechanism to support vector machines \citep{Cabannes2022b}.

In the first two papers, we built on combining weak information conditionally to a given input.
While relation between inputs were captured by the model of functions we learn, the model we used was not smartly leveraging the input distribution.
As such, we wanted to strengthen our principle to deal with settings akin to semi-supervised learning.
In particular, we wanted to incorporate Laplacian regularization into our framework.
Surprisingly, there were no good ``kernelized'' methods for Laplacian regularization but only ``local averaging'' methods.
Benefiting from the expertise of my supervisors on kernel methods, as well as exchanging with Loucas Pillaud-Vivien, who already realized this fact, we came up with a paper on the matter \citep{Cabannes2021c}.
Arguably, this gave a satisfying final piece to the theoretical framework built in this thesis in order to learn from partial supervision.

From there were several natural continuations.
Deepen the study of partial supervision by understanding and incorporating more heuristics into our frameworks, such as collaborative filtering.
Refine our algorithms with refinement of the structured prediction framework of Alessandro, for example, what can we say about conditional random fields and partial supervision? could we simplify the link between the structure of the supervision and the structure in ``structured prediction'', as we have implicitly done in the first two papers applications and algorithms?
Widen our scope to all weak supervision frameworks, or even all unclutter data problems ({\em e.g.} missing input data).
Ultimately, we decided to go for dataset annotation.

The ultimate goal of this work is to be useful to the practitioner hustling with data.
Indeed, our research was not motivated by the fact that many datasets come with weak supervision, but by the idea that weak supervision is easier to collect for the practitioner.
As a consequence, if we suppose that the practitioner is annotating data, we might help him to efficiently provide the most useful supervision.
This problem has deep links with active learning, sequential design, non i.i.d. statistics and search games; links that were attractive to me.
Building on our previous framework, we came up with a simple model of annotation cost and looked at techniques to minimize this cost for a given level of probably approximately correct bound.
In dimension one, our atomic questions are exactly the gradient of the $L^1$ loss; this pushed us to realize that one does not need full supervision to acquire unbiased stochastic gradients \citep{Cabannes2022}.
However, when dealing with highly structured prediction, naive stochastic gradient descent on continuous surrogates suffer from suboptimality issues.
Hence, in order to have a finer control on the samples we collected, we are currently switching to a bandit setting, for which, with the help of Vianney Perchet, we hope to come up with creative solutions.

\subsection*{Ethical considerations}
This work aims at advancing our understanding of weakly supervised learning.
Weakly supervised learning enrolls in the quest of an automated artificial intelligence, free from the need of human supervision.
Automation, which is at the basis of computer science \citep{Turing1950}, is known to increase productivity at a reduced human labor cost, and is associated with several political/societal issues.
\part{Introduction}
\chapter{The Machine Learning Paradigm}

In 2021, the International Data Corporation estimates that by 2025, the yearly global spending on artificial intelligence (AI) will overrun \$204 billions \citep{IDC2021}.
To give an order of magnitude, this corresponds to the gross domestic product of countries such as Greece, New Zealand or Peru.\footnote{According to the World Bank Open Data.}
It means that if this business was concentrated in Peru, its 33 millions inhabitants would only produce goods and services related to AI.
Nowadays, the core engine behind AI is machine learning, that is the learning of rules and algorithms by computers.
In this chapter, we provide a subjective exposition of what machine learning is about.
It is written to be accessible for a large public beyond the machine learning community.

\section{From algorithmics to machine learning}

In this section, we present machine learning as the evolution of algorithmics.

In a prosaic fashion, an algorithm can be thought of as a cooking recipe.
It is a sequence of instructions that, given some inputs, produces an output.
For example, the inputs could be some apples, lard, eggs, salt and flour, the sequence of instructions be the one of an apple pie recipe, and the output be a delicious pie.
This can be formalized mathematically, an algorithm being understood as a sequence of instruction $I$, leading to a mapping $f:\X\to\Y$, which, given an input $x$ belonging to an input space $\X$, outputs $y = f(x)$ belonging to an output space $\Y$.

\begin{example}[Cooking recipe]
  \label{ex:cooking}
  $\X$ is a set of potential ingredients, $\Y$ is a set of potential pies.
  For some ingredients $x$, {\em e.g.} $x = ($5 apples, 100g of lard, 200g of flour, 1 egg, 1 pinch of salt$)$, $y = f(x)$ describes the pie obtained by following the instructions $I$ of a recipe.
\end{example}

\begin{example}[Sorting algorithm]
  \label{ex:sorting}
  $\X = \R^n$ is the set of lists of size $n$, $\Y = \brace{(y_1, \cdots, y_n) \in \R^n \midvert y_1 \leq \cdots \leq y_n}$ is the set of sorted lists of size $n$.
  Any sorting algorithm corresponds to $f:\X\to\Y$ that maps a list $x = (x_i)_{i\leq n}$ to its sorted version $y = (x_{\sigma(i)})_{i\leq n} \in \Y$ with $\sigma \in \Sfrak_n$ a permutation.
\end{example}

\begin{example}[House pricing]
  \label{ex:housing}
  $\X = \R^2_+$ represents real estate properties with two features: the living space of the house, the outdoor space of the garden, both in meter square.
  $\Y = \R_+$ represents the price of the house in US dollars.
  The mapping $f$ is a model for house prices.
\end{example}

\begin{example}[Classifying cats and dogs]
  \label{ex:image_classification}
  $\X = [255]^{d_1\times d_2\times 3}$ is a set of images of cats and dogs. Those images are represented by $d_1 \times d_2$ pixels, with the three RGB channels.
  $\Y = \brace{-1, 1}$ with $f(x)=1$ if $x$ is an image of cats, and $f(x)=-1$ if $x$ is an image of dogs.
\end{example}

The problematics arising in our examples are all slightly different and will be helpful to illustrate differences between algorithmics, statistics and machine learning.

\paragraph{Algorithmics.}
In classical computer science, algorithms are implemented on $m$ bits of memories. As such, the sequence of instructions can be written as $I = (I_t)_{0\leq t\leq T+1}$, with $I_0:\X\to\brace{0, 1}^m$ an encoding of $\X$ on $m$ bits, $I_{T+1}:\R^m \to \Y$ a decoding of $m$ bits to $\Y$, and, for $t\in [T]$, $I_t:\brace{0,1}^m\to\brace{0,1}^m$ a basic bitwise operation.
In problems such as sorting lists, we know exactly which $f$ we want to achieve, but we do not know how this mapping can be decomposed into a set of basic operations which could be implemented practically on a computer.
In those algorithmic problems, an algorithm is thought of as the sequence of instructions $I$,and its quality is discussed under the light of two notions.
\begin{itemize}
  \item Correctness: does this sequence of instructions correspond to the desired mapping $f$? In other terms, can we guarantee the equality $f = I_{T+1}\circ I_{T} \circ \cdots \circ I_0$?
  \item Complexity: how much basic operations $T$ and computer memory $m$ does it take to perform the full sequence of instruction $I$?
\end{itemize}
In the cooking example \ref{ex:cooking}, the first question can be rephrased as: does our recipe, given a specification of ingredients, always lead to the same pie?
The second question is related to the equipment and the time required by the recipe.
When it comes to sorting lists, example \ref{ex:sorting}, deriving and implementing a sequence of basic operations to solve this task is a reasonable idea.
For example, think of how you sort a deck of cards, write rules about it, and implement those rules in your favorite programming language that will convert it into a sequence of bitwise instructions for the computer.
When it comes to understanding images, deriving and implementing a sequence of basic operations to differentiate cats and dogs have not been proven a successful way to proceed.
Indeed, even for language translation, efforts, made by linguists to describe translation as a set of simple syntactic rules, have not led to efficient translation algorithms.

\paragraph{Learning optimal sequence of instructions.}
Rather than designing an algorithm from a set of humanly-thought rules, machine learning suggests designing an algorithm by quantifying a notion of optimality, or equivalently a notion of error, cost or risk ${\cal R}$, and looking among all potential sequences of instructions $I \in {\cal I}$, in a class ${\cal I}$ of potential sequences of instruction, for the one that minimize the risk ${\cal R}(I)$.
For example, in the cooking example, we could consider ${\cal I}$ as all potential combinations of basic instructions such as ``breaking eggs'', ``mixing some ingredients'', ``frying/baking a batter'' and so on.
Optimality, and the risk ${\cal R}$, could be designed by some experts testing the pie, or through molecular gastronomy considerations.
To find the best recipe, that is the best sequence of basic instructions, a naive approach consists in trying all potential combinations of basic instructions $I$, and quantifying the result quality through ${\cal R}(I)$, and considering the sequence $I$ that achieves the minimal risk.
Of course, there are zillions of ways to combine basic instructions, and baking zillions of pie, and having them tested by experts is not a viable option.
But on a computer that can perform millions of operations by seconds, the picture is quite different.
Applied to the list sorting problem, such a procedure leads to a completely different approach from the classic algorithmic one.
Roughly speaking, the machine learning paradigm is to try all potential algorithms and take the best one, where ``the best'' is quantified through the risk ${\cal R}$.

\section{The emergence of statistics}

In this section, we get more specific on what machine learning is today, that is, statistics in the era of powerful computers.

While machine learning could be seen as the art of automatically designing algorithms, nowadays, it is rather a new take on statistics in the era of powerful computers.
As such, it is less focused on having a ``good'' correct sequence of instruction $I$ to perform a task, but rather in finding a mapping $f$ that is optimal for a task, such as pricing houses or distinguishing cats and dogs.
In such a setting, the risk ${\cal R}:\Y^\X\to \R_+$ associates a score ${\cal R}(f)$, quantifying a notion of risk, cost or error, to a function $f:\X\to\Y$.
Ideally, we would like to retrieve an optimal mapping $f^*:\X\to\Y$ such that
\begin{equation}
  {\cal R}(f^*) = {\cal R}^*, \qquad\text{where}\qquad
  {\cal R}^* = \inf_{f:\X\to\Y} {\cal R}(f).
\end{equation}
The problematics of machine learning are slightly different from the ones of algorithmics.
The quality of a procedure, that allows obtaining automatically a mapping $f:\X\to\Y$ to solve a specified task, is discussed under the light of three notions.
\begin{itemize}
  \item Optimality: is the risk ${\cal R}(f)$ of the obtained mapping $f$ close to ${\cal R}^*$?
  \item Inference complexity: given an input $x\in\X$, is it easy to compute $f(x)$?
  \item Training complexity: is it computationally easy to obtain the mapping $f$?
\end{itemize}
The first two issues are the direct translation of the algorithmic issues of correctness and complexity, while the third one is specific to machine learning, and has strong links with the field of optimization.

\begin{remark}[Correctness and Optimality]
  In classical algorithms, there is the idea that an algorithm is perfect/correct, that we know exactly what it is doing.
  In machine learning, the idea of $f$ being correct is dropped by the notion of $f$ being optimal or quite optimal.
  For critical applications such as flying planes, replacing the notion of correctness brought in by classical algorithms, by a notion of quasi-optimality is not innocuous.
  Yet, once a mapping $f$ is retrieved by an artificial intelligence system, it can be tested for correctness through formal proof management systems, or tested for statistical quality through an extensive number of testings.
  While replacing ``being correct'' by ``being statistically good'' can offend purists, note that, in medicine, drugs are more often approved based on statistical studies, rather than on a precise description of their mechanisms and a complete understanding of their interactions with the body.
\end{remark}

If we consider the cooking example \ref{ex:cooking}, we could translate those three issues into prosaic terms. Suppose that you have a way to design new recipes.
Optimality translates into ``are the new dishes you make good?''.
Inference complexity translates into ``are the recipes easy to reproduce?''.
Optimization complexity translates into ``is your procedure for designing new recipes difficult, is it time and money consuming?''.

\subsection{Supervised learning}
A simple idea to learn a mapping $f$ between inputs $x$ and outputs $y$ is to consider a set of $n$ examples $(x_i, y_i)_{i\leq n} \in (\X\times\Y)^n$, where $f(x_i)$ should be $y_i$, and try to infer from those examples a general law between $x$ and $y$.
That is, for house pricing, example \ref{ex:housing}, consider $n$ houses, report their characteristics $(x_i)_{i\leq n}$ and their prices $(y_i)_{i\leq n}$, and try to find a law that links house characteristics $x$ to their prices $y$.
In particular, a real estate agent can try to fit a linear model $y = w^\top x$, for $w \in \R^d$ a weight vector, the house characteristics $x$ assumed to belong to $\X\subset\R^d$, and $y$ the price of the house.
It is usual to fit $w$ by minimizing the least-squares error
\[
  \hat w \in \argmin_{w\in\R^d} \frac{1}{n} \sum_{i=1}^n \norm{w^\top x_i -
    y_i}^2 = (\sum_{i=1}^n x_i x_i^\top )^{\dagger} \sum_{i=1}^n y_i x_i,
\]
where, for a vector $x\in\R^d = \R^{d\times 1}$, $x^\top \in \R^{1\times d}$ designs its transpose, and, for a matrix $A$, $A^\dagger$ its pseudo-inverse.
When needing to price a new house, the real estate agent could report its characteristics $x$ and price it at $y = \hat{w}^\top x$.

Such a model raises many questions.
Which features $x$ govern the price of a house?
Can the price of the house be deduced as a linear combination of those features?
How many houses $n$ should I consider as training examples $(x_i, y_i)_{i\leq n}$ to retrieve a $\hat w$ such that the value $\hat w^\top x$ does reflect well the market price of a new house characterized by $x$?
How much should we expect the transaction price of a house to deviate from our model of its market price?
Those questions are addressed by statisticians and machine learning practitioners.
They are respectively linked with features engineering, model selection, data collection, and generalization guarantees.

Perhaps the most well known result of statistics is the large law number.
In essence, it tells you that if you repeat the same experiment, what you see as the result of this experiment, reflects what you are likely to see if you repeat this experiment one more time.
For example, if you have rolled a die zillion times, and you have gotten six forty-five percent of the time, the next time you roll this die, you can expect to get a six with forty-five percent of chance.
In the past two centuries, statisticians have developed many concentration inequalities that quantify, given the number of experiments you have run so far, how much the statistics collected from those experiments are likely to reflect the outcome of future experiments.
This provides tools to give generalization guarantees on models learned from data.
But beyond those results are some axioms on the experiments you run.
In particular, statistics are grounded in probability theory, assuming that the experiment's outcomes are governed by some probabilistic process that is stable over time, allowing prediction of future outcomes from prior ones.
Typically, when talking about dice, we assume that you roll the same dice, in the same fashion, in the same environment.

Learning theory, and in particular supervised learning, builds on this idea of understanding observable phenomena by assuming an underlying probabilistic model.
This thesis is set in this framework. It assumes the existence of a joint probability distribution $\rho \in \prob{\X\times\Y}$ over $\X$ and $\Y$, that have generated the dataset ${\cal D}_n = (X_i, Y_i)_{i\leq n} \sim \rho^{\otimes n}$ in an {\em independent identically distributed} fashion.
Moreover, it assumes that the measure of error results from the integration of a pointwise measure of error $\ell(f(X), Y)$, between a prediction $f(X)$ made at the point $X$ and its observed label $Y$, for $\ell \in \Y\times\Y\to\R$ a loss function, in the sense that for $f:\X\to\Y$,
\begin{equation}
  {\cal R}(f) = \E_{(X, Y)\sim\rho}\bracket{\ell(f(X), Y)}.
\end{equation}
As one has no access to the distribution $\rho$ but only to samples $(X_i, Y_i)_{i\leq n}$, it is natural to substitute the risk with its empirical counterpart ${\cal R}_{{\cal D}_n}:\Y^\X\to \R$, defined as
\begin{equation}
  {\cal R}_{{\cal D}_n}(f) = \frac{1}{n}\sum_{i=1}^n\bracket{\ell(f(X_i), Y_i)}.
\end{equation}
A popular approach to learn a mapping $f:\X\to\Y$ is to consider some model of functions ${\cal F} \subset \Y^\X$ and to perform empirical risk minimization over this class of functions, leading to the estimate
\[
  f_n \in \argmin_{f\in{\cal F}} {\cal R}_{{\cal D}_n}(f).
\]
Yet, how optimal is this mapping $f_n$? In other terms, does it exhibit a low risk ${\cal R}(f)$?
In the machine learning community, this question refers to generalization properties of $f_n$: does what we have learned based on $n$ examples generalize to new situations characterized by new inputs?
Statistical learning theory provides some answers to this question.

\begin{remark}[Why probability?]
  Arguably, we live in a deterministic world. Hence, if everything has a reason, why introduce randomness in our model?
  Think about rolling dice, the face that pops up is a pure deterministic function of the landscape of the support the dice are rolled on, as well as the initial moment, impulsion and position of the dice, which themselves depend on the mood of the person rolling the dice.
  But as we cannot model those factors, we suppose them to be random, leading to the random distribution on the rolling dice, at the end of the causal chain from mood to initial physical values to dice output.
  The same applies to house price, many idiosyncratic factors come into play when a buying offer is made on a house.
  When tracking those explanation variables is too difficult, statisticians deal with this fact by adding randomness into the model.
\end{remark}

\subsection{Generalization bounds}
In this subsection, we mimic typical derivations provided by statistical learning theory which are used in this thesis.
Recall the house pricing example \ref{ex:housing}.
Supervised learning consists in assuming the existence of some abstract distribution $\rho\in\X\times\Y$, and that pairs of house characteristics and their prices $(x_i, y_i)_{i\leq n}$ are actually sampled from this distribution $(x_i, y_i) = (X_i, Y_i) \sim \rho^{\otimes n}$ -- using capital letters to denote the randomness in the sampling.
Trying to predict house prices from a linear combination of house characteristics leads to the model of functions ${\cal F} = \brace{x\to w^\top x\midvert w\in\R^d}$, assuming $x\in\X\subset \R^d$.
The best linear model, based on the mean-squares error, is defined by $f^*_{\text{lin}}:x\to x^\top w^*$, with the weighting scheme
\[
  w^* \in \argmin_{w\in\R^d} \E_{(X, Y)\sim\rho}\bracket{\norm{X^\top w - Y}^2}
  = (\E_{X\sim\rho_\X}\bracket{XX^\top})^{\dagger} \E_{(X, Y)\sim\rho}\bracket{Y X}.
\]
Here $\rho_\X$ designs the marginal of $\rho$ over $\X$, that is, the distribution of $X$ according to $\rho$.
In practice, we do not have access to $\rho$ but to the samples $(X_i, Y_i)$.
As a consequence, we can replace $\rho$ by $\hat\rho = \frac{1}{n} \sum_{i=1}^n \delta_{(X_i, Y_i)}$, which leads to the empirical risk minimizer $f_n: x\to x^\top \hat{w}$ with
\[
  \hat{w} \in \argmin_{w\in\R^d} \frac{1}{n}\sum_{i=1}^n\norm{X_i^\top w - Y_i}^2
  = \paren{\frac{1}{n}\sum_{i=1}^n X_iX_i^\top}^{\dagger} \paren{\frac{1}{n}
    \sum_{i=1}^n Y_i X_i}.
\]
As the quality of a function $f$ is measured through the risk ${\cal R}$, it is natural to measure the quality of $f_n$ in terms of the excess of risk ${\cal R}(f_n) - {\cal R}^*$, with respect to the minimum achievable risk ${\cal R}^*$.
This error can be split in the following way
\begin{equation}
  \label{eq:risk-dec}
  {\cal R}(f_n) - {\cal R}^* =
  \underbrace{{\cal R}(f_n) - {\cal R}(f_{\text{lin}}^*)}_{\text{estimation error}}
  \ + \underbrace{{\cal R}(f_{\text{lin}}^*) - {\cal R}^*}_{\text{approximation error}}.
\end{equation}
This split separates an error due to the need for more data so that $\hat{w}$ estimates $w^*$ correctly and a part due to the fact that the best linear predictor $f^*_{\text{lin}}$ might not be the best pricing model.
The decomposition with estimation and approximation errors, is sometimes called bias-variance decomposition.
The variance, or estimation error, is due to the randomness of the samples, while the bias, or approximation error, is due to the gap between the average (or best) estimate we expect and the solution.

The approximation error is due to the fact that our model might be inaccurate.
This error can be controlled by assuming some sort of density of ${\cal F}$ in $\X\to\Y$ regarding the topology inherited from ${\cal R}$, or by assuming that the best predictor $f^*$ does belong to our class of functions ${\cal F}$ (or is well approximated by it with respect to the topology associated with ${\cal R}$).

The estimation error is due to the fact that we have estimated a quantity defined through the distribution $\rho$ thanks to sample $(X_i, Y_i)_{i\leq n}$.
Similarly to rolling dice or tossing coins, when knowing $\rho$, statistics and probability can quantify, for a given number of training examples $n$, what is the distribution of ${\cal R}(f_n) - {\cal R}(f_{\text{lin}}^*)$.
This distribution should be seen as the pushforward of $\rho^{\otimes n}$ regarding the process that leads to the estimate $f_n$ from samples $(X_i, Y_i)_{i\leq n}$.
When not knowing $\rho$, this pushforward cannot be estimated, but with few hypotheses on $\rho$, such as control of extreme behaviors, or equivalently control of high moments, it is possible to control deviation of ${\cal R}(f_n)$.
This is exactly what provide concentration inequalities, such as Hoeffding or Bernstein inequalities, that allows upper-bounding the deviations between the empirical means $\frac{1}{n}\sum_{i=1}^n X_i X_i^\top$ and $\frac{1}{n}\sum_{i=1}^n Y_i X_i$ and their population counterpart $\E[XX^\top]$ and $\E[YX]$, and therefore the deviation between $\hat{w}$ and $w^*$.

With assumptions to control the approximation and estimation errors, it is possible to get convergence results of the type,
\[
  \forall\, t > 0,\qquad \Pbb_{{\cal D}_n}({\cal R}(f_n) - {\cal R}^* > t)
  \leq \exp(-\sigma^{-2} nt^2).
\]
This specific result tells you that if you have trained your model with $n$ independent identically distributed samples $(X_i, Y_i)_{i\leq n}$, the risk of the learned mapping $f_n$ is comparable to a random number that follows a Gaussian distribution ${\cal N}(0, \sigma / \sqrt{n})$, the randomness being due to the randomness of the samples.
Such a bound is called a generalization bound as it tells us what risk or error we can expect on generic examples, which is a measure of how well our predictor generalizes to new situations.
When an algorithm benefits from such a bound, we call it probably approximately correct \citep{Valient2013}, in the sense that with high probability (the randomness being due to the sample population), it approximately minimizes the risk.
It is worth noting that, in practice, we only have one realization of ${\cal D}_n$, and that this result does not tell us that the risk ${\cal R}(f_n)$ is small, it only tells us that we would have been really unlucky if it was not the case.

\subsection{Unsupervised learning}
In the preceding sections, we have presented machine learning as concerned with the learning of a mapping from an input space $\X$ to an output space $\Y$.
As such, it differs from statistics that are rather concerned with inferring properties of a distribution $\rho_\X$ from samples $(X_i)_{i\leq n}$.
Indeed, many papers published as machine learning papers are concerned with this setting, usually referred to as unsupervised learning.
There are two major motivations for unsupervised learning.

First, statistical problems such as density estimation, eventually done through maximum likelihood estimation, are instances of problems where we would like to infer properties of a distribution $\rho_\X$ through samples $(X_i)_{i\leq n}$.
The usage of a computer, and of optimization routines, is crucial to deal with a large number of data in this setting.
Indeed, many of the computational and statistical problematics arising from those problems are shared with standard supervised learning, justifying results on those problems to be published in machine learning conferences, and justifying the machine learning community to welcome statisticians.

Second, learning a mapping from inputs to outputs, highly benefits from features engineering, which might be done without access to the supervision $(Y_i)_{i\leq n}$.
From a machine learning perspective, the main goal of unsupervised learning is to discover structure beyond input data, assuming that this structure will help to find laws that correlate those inputs to potential outputs.
For example, when described as an array of pixels, images are understood as points in $\R^d$ for $d$ of order $10^9$.
Yet, we expect natural images to have a strong structure that forbid them to be any points in $\R^d$.
Indeed, we can expect natural images to be concentrated close to a low-dimensional manifold, potentially parametrized in $\R^m$ for $m$ much smaller than $d$, describing the many parameters of freedom of a natural image.
Finding such a small representation would ease ``downstream'' tasks, such as learning to recognize the content of an image.
Having been approached with many perspectives, unsupervised learning has been divided into many sub-problems, such as clustering, manifold learning, sparse dictionary learning, or self-supervised learning.

\section{A tour of problems}
In this section, we review diverse practical problems that are arguably understood as machine learning problems.
We review machine learning to create artificial intelligence with human-alike capability.
We showcase business motivations and driving forces beyond machine learning.
We explain how data science can fit into intelligent systems, and discuss how machine learning can help science and how its scope is growing behind statistical learning.
We do not aim for completeness, but for simple examples.

\subsection{Seeing, hearing and talking}
For non-experts of machine learning, probably the most important result of this last decade was the result achieved by deep learning trained on GPU on the ImageNet challenge of 2012 \citep{Krizhevsky2012}. In essence, for the first time, a computer was able to showcase human-like performance when recognizing simple objects in images.
Similarly, we can dictate to our electronic devices and have those devices writing down our thoughts as if they were secretaries.
Moreover, those devices can emit sounds articulated into words and sentences to answer a question we have eventually asked.
All as if machines were to see, hear and speak.
This echoes a long fantasy of machines bearing signs of human intelligence.\footnote{Indeed, some researchers use concepts of human psychology to describe their algorithms such as attention and saliency maps \citep{Cabannes2020b}.}
Interestingly, this ``artificial intelligence'' replicates tasks that many animals do automatically and instantaneously, concerned with processing simple information.
As such it is more concerned with intelligence as collecting pieces of information (such as ``military intelligence''), rather than the more deep thought and contemplative process captured by the French notion of {\em  intelligence}.
Those recent successes are somewhat based on supervised learning, providing strong evidence of the usefulness of this theoretical framework, especially when compared to previous attempts to build computer vision, audio and natural language processing from so-called ``expert systems''.
Solving those problems has required machine learning to scale with millions of training examples $n$, as well as millions of input dimensions $d$, which has strongly pushed to advance calculations engineering \citep[see, \emph{e.g.},][]{Chowdhery2022}.

\subsection{Playing games}
Another highly publicized result of machine learning was the beating of all humans at Go, an abstract strategy board game known for the richness of all potential strategies \citep{Silver2018}.
Games such as Go or chess do not fit exactly the framework of supervised learning.
Those games are characterized by a position, or state, which mainly corresponds to the disposition of pieces on the game board, that is changed by moves, or action, made iteratively by two players.
The game is over when a terminal state is reached.
Each terminal state is associated with a game output, could it be a draw, a win of player one or a win of player two.
What needs to be learned is a strategy, or policy, that can be seen as a function that maps a game state to an action or move.
The goal is to come up with a winning strategy whatsoever the opponent strategy.
These problems can be approached with reinforcement learning \citep{Sutton2018}, whose goal is to navigate an environment over time in order to maximize some reward functions.
The wording ``reinforcement'' came from behavioral psychology and describes the fact that the strategy is learned by trials and errors.
In the case of games, the computer can simulate many games against itself before finding a winning strategy.

\subsection{Recommender systems}
\label{int:sec:coll_filt}
Far away from the fantasy of creating superhuman creatures, a consequent amount of machine learning development is the fruit of businesses trying to better identify potential customers and prospects of bestseller products.
A particularly hot topic is recommender systems, whose goal is to propose to a given customer characterized by some features $x$, the most adequate content $y$.
Recommender systems are now everywhere to suggest personalized content, could it be for browsing music and movies, or for targeted advertising.

In 2009, Netflix ran a competition to predict how many stars over five a user might rank a movie. Their goal was clear : have a system that finds the most relevant movies to recommend to the user. They offered \$1,000,000 to anyone coming up with the best performance on this prediction problem \citep{Bennett2007,Lohr2009}.
A specificity of this problem was the absence of features to describe movies or users. Among suggestions of many participants, the best solution was provided by "collaborative filtering". This technique consists in trying to factorize the note given to a movie by a user as a product $u^\top m$, where $u$ characterizes the user, and $m$ characterizes the movie with the smallest (in terms of number of coordinates) vectors $u$ and $m$.
The goal is to find few characteristics for each user that respond to the same number of few characteristics for each movie in order to explain given scores and to predict future scores -- think of $u = (\text{how much I like romantic movie}, \text{how much I like action movie})$ and of
$a = (\text{how much is it a romantic movie}, \text{how much is it an action movie})$.
Formally this technique is formalized by trying to factorize the table $S=(\text{score}(\text{user},\text{movie}))_{\text{user}, \text{movie}}$ as $S = U M$ on observed scores for $U$ and $M$ of smallest rank.
While the rank is not a continuous function, making this problem hard to solve, it is possible to relax it into a concave function thanks to the nuclear norm and get nicely behaving solutions.

\subsection{Intelligent infrastructure}
The machine learning of today is the science of data. The competency to manage massive amounts of data opens the way for massive systems of information.
Moreover, advances in sensors and measurement tools as well as robots allowing vast automation of tasks create optimistic perspectives in fields such as farming or medicine.
The big dreamer might fall for the internet of things governing hydrometry sensors, cameras recognizing aphids, automatic release of pesticides, analysis of the best companion plants and so on.
But without going that far, simply putting in place numerical tools to easily perform large scale cross-sectional and longitudinal studies would highly accelerate medicine -- recall debates on Hydroxychloroquine as a COVID treatment based on small studies such as \cite{Raoult2020}.
Arguably, the revolution promised by machine learning is not the revolution of machine learning per se, nor the creation of supra-humanoid, but the future deployment of massive active monitoring systems \citep{Jordan2019}.

To give a personal touch to this discussion, in 2019, with two friends, we created an ``artificial intelligent system'' to interact with artists during their creation process \citep{Cabannes2019}. In this project, much of our work was engineering to make sure that the process was innocuous for the artists. The machine learning algorithm running in the middle was only a part of the complete picture.

\subsection{Making sense of scientific data}
While we have described the application of machine learning for non-science purposes, in particular for marketing and agriculture, there is traction for machine learning as science for science.
Of course, supervised learning techniques can be applied to scientific problems, such as the protein folding problem \citep{Jumper2021}.
But machine learning techniques are also handy to make sense of data coming from massive scientific experiments such as the quest for the Higgs boson at the CERN \citep{Larkoski2020}, or the analysis of gravitational waves \citep{Gebhard2019}.
In those last two examples, the goal is to retrieve a signal that has caused some observations given a potential high-level of noise.

\subsection{Signal recovery and inverse problems}

Signal recovery from noisy observations is an active field of research with many appreciable results.
The goal of compression is to, given a signal $x$ that is heavy to describe, compress it into a signal $y=f(x)$ much lighter to describe, store, transmit, {\em etc.} such that given the knowledge of $f$ and a prior on $x$, one can retrieve the signal $x$ from the observation $y$.
Historically, this has been studied for telecommunications with several results coming from the Bell Labs, such as the Nyquist-Shannon theorem \citep{Shannon1949,Nyquist1928}: assuming $x$ to be a function with bounded frequencies, it can be reconstructed by interpolation \citep{Whittaker1915}.
More recently, \cite{Candes2006} and \cite{Donoho2006} looked at the transmission of a vector $x\in\R^n$ known to be sparse, {\em i.e.} having only $s<<n$ non-zero components, under the form $y = A x\in\R^m$ for a chosen $A\in\R^{m\times n}$.
They showed that by taking the row of $A$ sufficiently ``incoherent'' it is possible to recover $x$ with $m\approx 2s$.
Moreover, when the coefficients of $A$ are random, normally distributed, the ``incoherence'' property holds with high probability.
This allows to drastically reduce compression, as well as acquisition, of sparse signal, and had successful applications in medical imaging \citep{Lustig2008,Sidky2008}.
Think of the impact of compressed sensing when it permits reducing X-ray exposure during tomography for the same amount of reconstruction fidelity.

Signal recovery is not the primary focus of machine learning, but as the field is gaining traction, its tools and scope of applications are growing, and similarly to compressed sensing papers that find their place in machine learning journals and conferences, other imaging techniques might as well.
To conclude on those inverse problems beyond machine learning, let us mention the work of \cite{Fink2002}, which leverages wave reversibility and information preservation to design time-reversal mirrors in order to locate a wave source and recreate the emitted wave.
In the same vein, techniques such as sonar or medical ultrasonography are based on propagating waves into a medium and deduce a topography of the medium from the waves echo.
Among others, they have been used to locate oil in the ground, after creating a wave with dynamite.
Quite remarkably, recent research has shown those explosions to be a waste of resources as it is possible to leverage ambient noise of the Earth's crust in order to image it \citep{Garnier2016}.

\section{A tour of models}

In this section, we focus on the supervised learning settings, and we discuss classical models to learn functions.
Our classification into local averaging, reproducing kernel and deep learning methods is aligned with the class taught by Francis, on learning from first principles \citep{Bach2023}.

\subsection{Local averaging methods}
If you are a real estate agent pricing houses, and you encounter a new house and need to price it, a natural idea is to look for similar houses that have been sold recently, and infer a price from those similar examples.
This is exactly what captures local averaging methods.
Suppose that you have collected examples $(X_i, Y_i)_{i\leq n}$ with features $X_i$ characterizing a house indexed by $i \in [n]$ and its price $Y_i$.
Given a new house with characteristics $x$, you can look at the $k$-th nearest neighbors ${\cal N}_k(x)$ that contains $k$ elements in $[n]$ and such that for $i\in{\cal N}_k(x)$ and $j\in[n]\setminus{\cal N}_k(x)$, $d(x, X_i) \leq d(x, X_j)$ according to a distance $d:\X\times\X\to\R$.
This defines the nearest neighbors predictor $f_n:\X\to\Y$ as
\(
f_n(x) = \frac{1}{k}\sum_{i\in{\cal N}_k(x)} Y_i
\).
Eventually, you might want to give more importance to houses that are really similar to the one you are trying to price, and less to the $k$-th nearest neighbor.
To do so, you can introduce a system of weights $w_i=w_i^{(n)}:\X\to\R$ that are positive and sum to one, and refine $f_n$ as
\[
  f_n(x) = \sum_{i=1}^n w_i(x) Y_i =
  \sum_{i=1}^n w_i(x) f^*(X_i) + \sum_{i=1}^n w_i(x) \epsilon_i
\]
with $\epsilon_i = Y_i - f^*(X_i)$ linked with the noise of having observed $Y_i$ when we would have preferred to observe $f^*(X_i)$.

\paragraph{Universal consistency.}
While this pricing method sounds really sensible, the job of statisticians is to make formal statements to prove its soundness.
The most classical statement is to prove that when the number of examples goes to infinity, we retrieve the optimal pricing rule.
This property is known as {\em consistency}.
Consistency is usually defined formally through convergence in probability (although some prefer to define it with convergence almost surely) of the risk ${\cal R}(f_n)$, which is seen a random variable depending on the sampling of the dataset ${\cal D}_n = (X_i, Y_i)_{i\leq n}\sim \rho^{\otimes n}$, toward the infimum risk ${\cal R}^*$.
Methods that are consistent without assumptions on the solution $f^*$ are known as {\em universally consistent}.

Consistency of local averaging methods has first been derived by \cite{Stone1977}.
In essence, it consists in assuming that the noises $(\epsilon_i)_{i\leq n}$ are well-behaved, such that $\sum_{i=1}^n w_i(x) \epsilon_i$ cancels out as the number of samples $n$ goes to infinity, and assuming the weighting scheme concentrates locally around $x$ such that, for $\rho_\X$-almost all $x$, informally,
\[
  \lim_{n\to+\infty}f_n(x) =
  \lim_{n\to+\infty} \sum_{i=1}^n w_i(x)f^*(X_i)
  = \lim_{r\downarrow 0} \frac{1}{\rho_\X(B(x, r))} \int_{B(x, r)}
  f^*(t)\rho_\X(\diff t) = f^*(x).
\]
The last inequality being due to Lebesgue differentiation theorem (the derivative of the integral of a function equals the original function), which is true when $f^*$ is bounded.

\paragraph{Curse of dimensionality.}
Once universal consistency has been derived, statisticians focus on asymptotic rates of convergence, as well as non-asymptotic generalization bounds.
Those results are more informative than universal consistency, similarly to the fact that concentration inequalities are arguably stronger than the central limit theorem which is stronger than the law of large numbers.
When providing non-asymptotic generalization bounds, we should specify a loss function.
When we output continuous values, $\Y=\R$, for algebraic reasons, people tend to use the mean-squares loss, defined as $\ell(z, y) = \norm{z-y}^2$.
In this setting, $f^*(x) = \E_\rho\bracket{Y\midvert X=x}$, and under the condition that $f^*$ is Lipschitz plus some mild condition on $\rho$ and classical assumptions on the weighting scheme \citep{Gyorfi2002}, it is possible to get convergence rates of the type
\begin{equation}
  \label{par:eq:nn_rates}
  {\cal R}(f_n) - {\cal R}(f^*) \leq c n^{-\frac{2}{d+2}},
\end{equation}
for $c$ a constant depending on the hardness of the problem, and $d$ the dimension of the input space $\X$.
The dependence in $d$ is explained by the fact that to cover the space $[0,1]^d$ with precision $\epsilon$ in order to have meaningful neighbors to predict the value of a Lipschitz function, one need $\epsilon^{-d}$ points.
Meaning that as the dimension increases, one will need exponentially more points in order to guarantee a given risk level, a phenomenon which is referred to as the {\em curse of dimensionality}.
The curse of dimensionality is troublesome for machine learning problems where the input dimension can be really large such as on images with millions of pixels.

\paragraph{Leveraging smoothness and higher-order information.}
It is possible to break the curse of dimensionality in~\eqref{par:eq:nn_rates} by leveraging additional assumed structure of $f^*$.
In particular, if $f^*$ admits smooth derivatives, one can use Taylor formula, and try to infer derivatives values by finite differences.
This will lead to a much more precise estimate $f_n$ of $f^*$.
In particular, if $f^*$ is $s+1$ times differentiable in every direction, when a zeroth order Taylor expansion makes a local error of order $\epsilon$, an $s$-th order expansion will make an error of order $\epsilon^s$.
As such, it is natural to hope for an estimator $f_n$ that achieve the following generalization bound,
\begin{equation}
  \label{par:eq:smooth_nn_rates}
  {\cal R}(f_n) - {\cal R}(f^*) \leq c n^{-\frac{2s}{d+2s}},
\end{equation}
This is exactly the rates that local polynomial methods are achieving.
We refer to \cite{Tsybakov2009} for additional details on the matter.
Note that higher-order derivatives need more points in the neighborhood of $x$ in order for finite difference methods to provide a good estimate of those derivatives.
As such, we might expect a transitory regime, before the number of samples $n$ gets sufficiently large, where there are not enough points in the neighborhood of $x$ to benefit from high-order smoothness.

\subsection{Reproducing kernel methods}
\label{int:krr}
Suppose that you are the same real agent, and that you encounter a new house that is really different from the one you have in your database, so that local averaging prediction is meaningless.
For example, suppose that the house is huge, but it is not close to any good schools, and while you have already sold big houses, or houses without good nearby schools, you have no records for big houses without good nearby schools.
Then a natural idea is to proceed with factor analysis, trying to figure how the price of a house is driven by its size, and by good nearby schools.
Simple factor analysis model supposes that characteristics $x \in \R^d$ combine linearly to form the output $y\in\R$.
This leads to the search of an estimate $f_n$ as a linear model $f_n(x) = w^\top x$, for $w\in\R^d$ to be determined in order to ensure that $f_n(X_i) \approx Y_i$ on your training dataset $(X_i, Y_i)_{i\leq n}$.

\begin{figure}
  \centering
  \includegraphics[width=.3\textwidth]{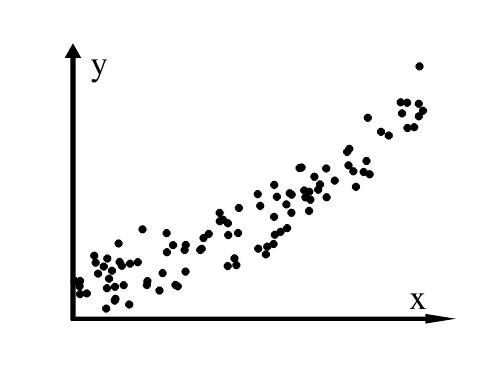}
  \includegraphics[width=.3\textwidth]{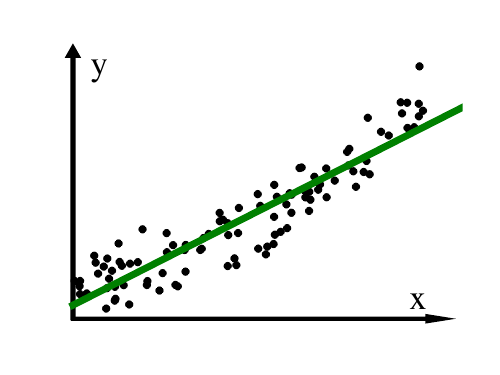}
  \includegraphics[width=.3\textwidth]{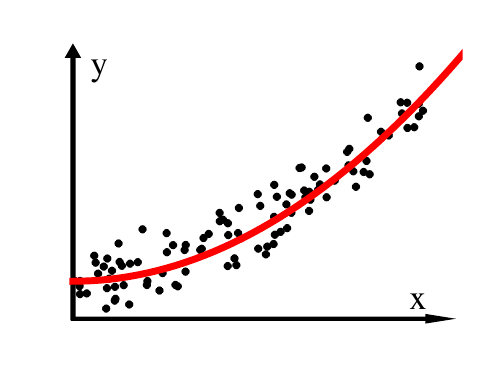}
  \caption{Plotting $(X_i, Y_i)_{i\leq n}$ (left) indicates that a quadratic regression (right) is better suited to derive $y$ from $x$ than a linear regression (middle). 
  However a quadratic regression is nothing but a linear regression with features $(1, x, x^2)$.}
  \label{par:fig:lin_reg}
\end{figure}

\paragraph{Packing features.}
Consider the case where $\X=\R$ and when plotting $(X_i, Y_i)_{i\leq n}$ seems to indicate that a quadratic relationship links $x$ to $y$ rather than a linear one, as illustrated on Figure \ref{par:fig:lin_reg}.
In this case, one can enrich the input $x$ with the features $\phi(x) = (1, x, x^2) \in \R^3$, and perform a linear regression with the data $(\phi(X_i), Y_i)_{i\leq n}$.
When it is hard to visualize the relationship between $x$ and $y$, one might be tempted to concatenate a lot of features in the vector $\phi(x)$ in order to make sure to be able to predict $y$ from $\phi(x)$.
However, such a technique is prone to learn spurious correlation between features and outputs of the training data, and not to generalize well to unseen I/O pairs.
This phenomenon is called overfitting.

\paragraph{Avoiding overfitting.}
If we directly find the predictor $f_n(x) = w^\top\phi(x)$ by minimizing $\sum_i \ell(w^\top \phi(X_i), Y_i)$, our method will likely overfit the training data.
Overfitting is often the fruit of a linear combination of features $\omega^\top\phi(X) \approx 0$ for $\omega \in \R^d$ (assuming $\phi(X)\in\R^d$) that is noise spuriously correlated with $Y$ on the training data $(X_i, Y_i)_{i\leq n}$, leading to $w = C\cdot\omega$ for a big $C$.
As a consequence, a way to avoid overfitting is to constraint $w$ to have a small norm, or to look for the minimizer of the regularized objective $\frac{1}{n} \sum_i \ell(w^\top \phi(X_i), Y_i) + \lambda \Omega(w)$, for $\Omega:\R^d\to\R$ a regularizing function and $\lambda\in\R_+$ a regularizing parameter.
To ease this minimization, it is possible to use $\Omega(w) = \norm{w}^2_2$.
An interesting alternative is to use the $\ell_1$-norm $\Omega(w) = \norm{w}_1$ as it pushes $w$ to be sparse \citep{Tibshirani1996}, which makes the predictor $f_n(x) = w^\top\phi(x)$ relatively simple to interpret.
Sparsity is a rich concept \citep{Bach2012}, yet it was not a focus of this thesis.
In the following, we will consider $\Omega(w) = \norm{w}^2_2$.

\paragraph{Kernel methods.}
When minimizing $\sum_i \ell(w^\top\phi(X_i), Y_i) + n\lambda \norm{w}^2_2$, it is clear that $w^* \in \Span(\phi(X_i))_{i\leq n}$ as any part of $w$ supported on the orthogonal of the span of the $(\phi(X_i))_{i\leq n}$ will not change the value of any $w^\top \phi(X_i)$ but will higher $\norm{w}_2$, a fact that is referred as the representer theorem \citep{Scholkopf2001}.
Looking for $w\in\R^d$ under the form $w = \sum_{i} \alpha_i \phi(X_i)$ with $\alpha \in \R^n$ leads to the new problem $\sum_i \ell([K\alpha]_i, Y_i) + \lambda \alpha^\top K \alpha$, with $K = (\scap{\phi(X_i)}{\phi(X_j)})_{i, j\leq n} \in \R^{n\times n}$, $[K\alpha]_i$ the $i$-th entry of the vector $K\alpha \in\R^n$ and $f_n(x) = \sum_i \alpha_i \scap{\phi(X_i)}{\phi(x)}$.
As one can notice, everything can be expressed through the scalar product $k:\X\times\X\to\R_+; (x, x') = \scap{\phi(x)}{\phi(x')}$, forgetting about the features map $\phi:\X\to\R^d$.
As such, rather than making explicit the features, one can make explicit the kernel $k$, under the sole condition that for any $m\in\N$ and $(x_i)_{i\leq m}\in\X^m$, $k(x_i, x_j)_{i,j\leq m}$ is symmetric semidefinite positive.
In particular, when $\X$ is a collection of structured objects, it can be easier to describe how similar are two $x$ through $k$ rather than deriving features $\phi$.

When $\ell(z, y) = \norm{z - y}^2$, the empirical risk minimization admits a closed form solution
\[
  f_n(x) = K_x^\top (K + n\lambda I)^{-1} \mathbf{Y},
\]
with $\mathbf{Y} = (Y_i)_{i\leq n} \in \R^n$ and $K_x = (k(X_i, x))_{i\leq n} \in \R^n$.
As $n$, the number of data, goes to infinity, this converges to
\(
f_\lambda(x) = \bar{K}_x (\bar{K} + \lambda I)^{-1} f^*
\)
where $\bar{K}$ operates on functions as $\bar{K}(f) = x\to \int k(x, x')f(x')\diff\rho_\X(x')$ and $\bar{K}_x(f) = \bar{K}(f)(x)$.
And as $\lambda$ goes to zero, under mild reachability assumption, $f_\lambda$ converges to $f^*$ \citep{Caponnetto2007}.
An advantage of kernel methods is that it reduces the search of $w\in\R^d$ to the search of $\alpha\in\R^n$ which is valuable when $d >> n$.
This is known as the ``kernel trick''.

\paragraph{Reproducing kernel Hilbert space (RKHS).}
Kernel methods are backed-up by a rich mathematical concept, linked with well-behaved classes of functions.
It can be introduced in a completely different fashion, as it has been historically by \cite{Aronszajn1950}.
Consider a scalar product of functions mapping $\X$ to $\Y=\R$ and the class of functions ${\cal F} = \brace{f:\X\to\Y\midvert \norm{f} = \sqrt{\scap{f}{f}}<+\infty}$.
Suppose also that the evaluation maps $L_x:{\cal F}\to\R; f\to f(x)$ are bounded linear operators.
This implies the existence of representer $k_x\in{\cal F}$ such that the ``reproducing property'' $L_x(f) = \scap{k_x}{f}$ holds for any $x\in\X$ and $f\in{\cal F}$.
It can be shown that ${\cal F}$ is the completion of the linear combination of $(k_x)_{x\in\X}$, ${\cal F} = \overline{\Span(k_x)_{x\in\X}}^{\norm{\cdot}}$.
One inclusion is due to the fact that $k_x\in{\cal F}$ and that ${\cal F}$ is close by linear combination and completion; the other due to the fact that if $f\in{\cal F}\cap (k_x)_{x\in\X}^\perp$ than $f(x) = \scap{k_x}{f} = 0$.
Hence, there is a unique association between the kernel $k:\X\times\X\to\R; (x, x') \to \scap{k_x}{k_{x'}}$ and the so-called reproducing kernel Hilbert space ${\cal F}$ \citep{Mercer1909}.

\paragraph{Smoothness adaptability.}
Under the knowledge that the target function $f^*$ belongs to a RKHS ${\cal F}$, using the associated kernel $k$ generates good learning rates, akin to linear regression.
Moreover, when using the mean-squares error, one can guarantee universal consistency of the method described above, under the assumption that ${\cal F}$ is dense in $L^2(\rho_\X)$, a property that is verified for many usual kernels \citep{Micchelli2006}.
But what about the rates in this case, do they deteriorate badly when $f^*\notin{\cal F}$?
How does kernel methods adapt to the regularity of the function $f^*$?
Indeed the operator $\bar{K}$ introduced above provides the answer.
In fact, it is possible to show that $L^2(\rho_\X) = \ima \bar{K}^0$ and ${\cal F} = \ima \bar{K}^{1/2}$ (we provide many details on this operator in sections \ref{rates:app:rkhs} and \ref{lap:app:operators}), and that by only adapting the regularization parameter $\lambda$ (which is usually done through cross-validation) the excess of risk for the least-squares error of the estimator introduced above is ${\cal O}(n^{-2q/(2q+1)})$ for the biggest $q\in(0,1)$ such that $f^*\in\ima \bar{K}^q$ \citep{Caponnetto2007}.

\paragraph{Transitory regimes.}
While research papers tend to showcase rates that are best in terms of exponents raising the number of samples, those rates are often not observed in practice before accessing a high number of samples.
As such, practitioners might prefer to write
\begin{equation}
  \label{par:eq:kkr_rates}
  {\cal R}(f_n) - {\cal R}(f^*) \leq \inf_{0\leq t\leq q} c_t n^{-\frac{2t}{2t+1}},
\end{equation}
for the biggest $q\in(0,1)$ such that $f^*\in\ima\bar{K}^q$, and some constants $(c_t)_{0\leq t\leq q}$.
The transitory regime corresponds to the case where the best bound is not achieved for $t=q$, and which is when, if the kernel $k$ is leveraging smoothness, points are too far apart to benefit from higher-order information.

\subsection{Neural networks and deep learning}

Most of the state-of-the-art algorithms derived through machine learning are currently based on deep learning.
Deep learning consists in long artificial neural networks.
Artificial neural networks are a cascade of linear operators and non-linearities that are optimized through gradient descent on empirical risks.
This model appeared in the middle of the 20\textsuperscript{th} century \citep{McCulloch1943,Rosenblatt1958}, and were advocated by a few researchers in the 90s \citep{Lecun1995}.
It rose to dominance when implemented on graphical processing units, achieving astonishing results on image recognition \citep{Krizhevsky2012}.

\begin{figure}
  \centering
  \begin{tikzpicture}
  \draw[thick] (0, 0) circle (.25);
  \draw[thick] (0, 1) circle (.25);
  \draw[thick] (0, 2) circle (.25);
  
  \draw[thick] (2, -1) circle (.25);
  \draw[thick] (2, 0) circle (.25);
  \draw[thick] (2, 1) circle (.25);
  \draw[thick] (2, 2) circle (.25);
  \draw[thick] (2, 3) circle (.25);

  \draw[thick] (4, -1) circle (.25);
  \draw[thick] (4, 0) circle (.25);
  \draw[thick] (4, 1) circle (.25);
  \draw[thick] (4, 2) circle (.25);
  \draw[thick] (4, 3) circle (.25);

  \draw[thick] (6, -.5) circle (.25);
  \draw[thick] (6, .5) circle (.25);
  \draw[thick] (6, 1.5) circle (.25);
  \draw[thick] (6, 2.5) circle (.25);

  \draw[thick] (8, .5) circle (.25);
  \draw[thick] (8, 1.5) circle (.25);

  \draw[thick,-latex] (.25, 0) -- (1.75, -1);
  \draw[thick,-latex] (.25, 0) -- (1.75, 0);
  \draw[thick,-latex] (.25, 0) -- (1.75, 1);
  \draw[thick,-latex] (.25, 0) -- (1.75, 2);
  \draw[thick,-latex] (.25, 0) -- (1.75, 3);

  \draw[thick,-latex] (.25, 1) -- (1.75, -1);
  \draw[thick,-latex] (.25, 1) -- (1.75, 0);
  \draw[thick,-latex] (.25, 1) -- (1.75, 1);
  \draw[thick,-latex] (.25, 1) -- (1.75, 2);
  \draw[thick,-latex] (.25, 1) -- (1.75, 3);

  \draw[thick,-latex] (.25, 2) -- (1.75, -1);
  \draw[thick,-latex] (.25, 2) -- (1.75, 0);
  \draw[thick,-latex] (.25, 2) -- (1.75, 1);
  \draw[thick,-latex] (.25, 2) -- (1.75, 2);
  \draw[thick,-latex] (.25, 2) -- (1.75, 3);

  \draw[thick,-latex] (2.25, -1) -- (3.75, -1);
  \draw[thick,-latex] (2.25, -1) -- (3.75, 0);
  \draw[thick,-latex] (2.25, -1) -- (3.75, 1);
  \draw[thick,-latex] (2.25, -1) -- (3.75, 2);
  \draw[thick,-latex] (2.25, -1) -- (3.75, 3);

  \draw[thick,-latex] (2.25, 0) -- (3.75, -1);
  \draw[thick,-latex] (2.25, 0) -- (3.75, 0);
  \draw[thick,-latex] (2.25, 0) -- (3.75, 1);
  \draw[thick,-latex] (2.25, 0) -- (3.75, 2);
  \draw[thick,-latex] (2.25, 0) -- (3.75, 3);

  \draw[thick,-latex] (2.25, 1) -- (3.75, -1);
  \draw[thick,-latex] (2.25, 1) -- (3.75, 0);
  \draw[thick,-latex] (2.25, 1) -- (3.75, 1);
  \draw[thick,-latex] (2.25, 1) -- (3.75, 2);
  \draw[thick,-latex] (2.25, 1) -- (3.75, 3);

  \draw[thick,-latex] (2.25, 2) -- (3.75, -1);
  \draw[thick,-latex] (2.25, 2) -- (3.75, 0);
  \draw[thick,-latex] (2.25, 2) -- (3.75, 1);
  \draw[thick,-latex] (2.25, 2) -- (3.75, 2);
  \draw[thick,-latex] (2.25, 2) -- (3.75, 3);

  \draw[thick,-latex] (2.25, 3) -- (3.75, -1);
  \draw[thick,-latex] (2.25, 3) -- (3.75, 0);
  \draw[thick,-latex] (2.25, 3) -- (3.75, 1);
  \draw[thick,-latex] (2.25, 3) -- (3.75, 2);
  \draw[thick,-latex] (2.25, 3) -- (3.75, 3);

  \draw[thick,-latex] (4.25, 3) -- (5.75, -.5);
  \draw[thick,-latex] (4.25, 3) -- (5.75, .5);
  \draw[thick,-latex] (4.25, 3) -- (5.75, 1.5);
  \draw[thick,-latex] (4.25, 3) -- (5.75, 2.5);

  \draw[thick,-latex] (4.25, 2) -- (5.75, -.5);
  \draw[thick,-latex] (4.25, 2) -- (5.75, .5);
  \draw[thick,-latex] (4.25, 2) -- (5.75, 1.5);
  \draw[thick,-latex] (4.25, 2) -- (5.75, 2.5);

  \draw[thick,-latex] (4.25, 1) -- (5.75, -.5);
  \draw[thick,-latex] (4.25, 1) -- (5.75, .5);
  \draw[thick,-latex] (4.25, 1) -- (5.75, 1.5);
  \draw[thick,-latex] (4.25, 1) -- (5.75, 2.5);

  \draw[thick,-latex] (4.25, 0) -- (5.75, -.5);
  \draw[thick,-latex] (4.25, 0) -- (5.75, .5);
  \draw[thick,-latex] (4.25, 0) -- (5.75, 1.5);
  \draw[thick,-latex] (4.25, 0) -- (5.75, 2.5);

  \draw[thick,-latex] (4.25, -1) -- (5.75, -.5);
  \draw[thick,-latex] (4.25, -1) -- (5.75, .5);
  \draw[thick,-latex] (4.25, -1) -- (5.75, 1.5);
  \draw[thick,-latex] (4.25, -1) -- (5.75, 2.5);

  \draw[thick,-latex] (6.25, 2.5) -- (7.75, .5);
  \draw[thick,-latex] (6.25, 2.5) -- (7.75, 1.5);

  \draw[thick,-latex] (6.25, 1.5) -- (7.75, .5);
  \draw[thick,-latex] (6.25, 1.5) -- (7.75, 1.5);

  \draw[thick,-latex] (6.25, .5) -- (7.75, .5);
  \draw[thick,-latex] (6.25, .5) -- (7.75, 1.5);

  \draw[thick,-latex] (6.25, -.5) -- (7.75, .5);
  \draw[thick,-latex] (6.25, -.5) -- (7.75, 1.5);

  \draw (0, -1.75) node[font=\footnotesize] {$x$};
  \draw (2, -1.75) node[font=\footnotesize] {$x_1 = \sigma(W_1 x)$};
  \draw (4, -1.75) node[font=\footnotesize] {$x_2 = \sigma(W_2 x_1)$};
  \draw (6, -1.75) node[font=\footnotesize] {$x_3 = \sigma(W_3 x_2)$};
  \draw (8, -1.75) node[font=\footnotesize] {$y = W_4 x_3$};
\end{tikzpicture}
  \caption{Representation of a neural network with three hidden layers and parameters $(d, h_1, h_2, h_3, m) = (3, 5, 5, 4, 2)$, as a weighted directed acyclic graph.
    The input $x = (x_{(i)}) \in \R^3$ is passed through the first set of edges, leading to the first layer with value $x_1 = (x_{1, (i)}) \in \R^5$ specified by $x_{1,(i)} = \sigma(\sum_{1\leq j\leq 3} W_{1, (i,j)} x_{(j)})$ based on the set of weights $W_1 = (W_{1, (i, j)}) \in \R^{5\times 3}$ and the non-linearity $\sigma:\R\to\R$, and so on until reaching the output $y \in \R^2$.
    The wording ``deep learning'' refers to learning with neural networks that
    have a really large number of hidden layers, which can lead to hundreds of
    billions of parameters \citep[see {\em e.g.}][]{Brown2020}.
  }
  \label{par:fig:nn_archi}
\end{figure}
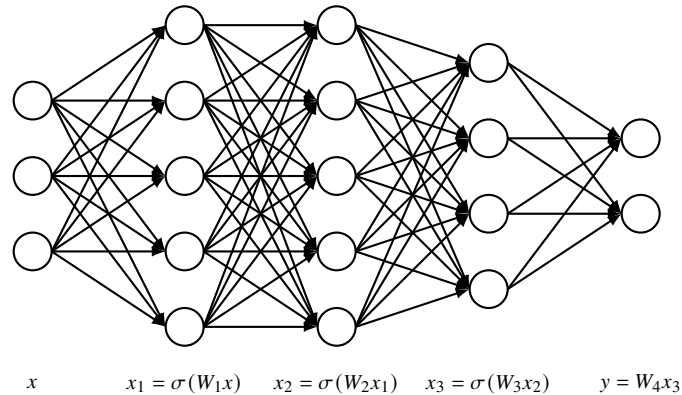

\paragraph{Cascade of linear operation.}
A neural network $f_n:\R^d\to\R^m$ with three hidden layers is parametrized with four matrices $W_1 \in \R^{h_1\times d}$, $W_2\in \R^{h_2\times h_1}$, $W_3\in \R^{h_3\times h_2}$, $W_4\in \R^{m\times h_3}$ and reads
\[
  f_n(x) = W_4\sigma (W_3\sigma (W_2 \sigma(W_1x))),
\]
with $h_1, h_2, h_3 \in \N$ the number of neurons in the first, second and third hidden layers, $\sigma:\R\to\R$ a non-linearity, {\em e.g.} $\sigma(x) = \max(x, 0)$, and the convention $\sigma((x_1, \cdots, x_n)^\top) = (\sigma(x_1), \cdots, \sigma(x_n))^\top$.
We illustrate it in Figure~\ref{par:fig:nn_archi}.
Given a loss $\ell$, some data ${\cal D}_n = (X_i, Y_i)_{i\leq n}$, a regularizer term $\lambda\Omega(W_1, W_2, W_3, W_4)$ and the consequent regularized empirical risk ${\cal R}_{{\cal D}_n}$, the parameters $W_1, W_2, W_3, W_4$ can be optimized in order to minimize ${\cal R}_{{\cal D}_n}(f_n)$.
This minimization is usually done with gradient descent, or stochastic gradient descent \citep{Robbins1951}, which is better suited for large scale learning problems \citep{Bottou2007}.
Gradients are obtained thanks to the chain rule, and while the computation of $f(x)$ can be seen as forward propagation from $x$ to $W_1x$ to $W_2\sigma(W_1x)$ and so on, the chain rule naturally leads to {\em backward propagation} or {\em backpropagation}, which is the name used by researchers to refer to the derivation of gradients with neural networks.
Astonishingly, while such a procedure is supposed to stall into undesirable local minima, it has not constrained the recent successes of deep learning.

\paragraph{Hierarchy and convolution.}
In the last paragraph, we have described the most basic architecture of a neural network, only made of fully connected layers.
There are many architectures that refine this basic neural network, the most well known are recurrent neural networks \citep{Williams1986} that allow dealing with time series of different length and convolutional neural networks that were proven successful to understand images \citep{LeCun2015}.
We refer the interested reader to this last paper to get a more precise idea on what convolutional neural networks are.
In substance, a one dimensional convolutional filter replaces the linear mapping $\R^{h_i}\to\R^{h_o} x\to Wx$ with $W \in \R^{h_0\times h_i}$ by the linear mapping $\R^{h_i}\to\R^{h_i-p}; x \to w*x$ with $w\in \R^p$ and $(w*x)_i = \sum_{j=1}^p w_i x_{i+j}$.
This is useful when $x$ has a translation-invariant structure \citep{Fukushima1980}.
Rather than learning to recognize the same pattern at different places by tuning similarly different columns of $W$, it allows for parameter sharing.
As one filter allows for the recognition of one pattern, a convolutional layer is made of several filters, whose responses are concatenated along a new dimension in order to form the layer output.
After passing through a filter bank, the response is usually pruned by keeping only local maxima, an operation known as max-pooling.
This subsampling operation builds a hierarchical structure, allowing for further filters to learn broader patterns \citep{Behnke2003}.
Translation invariance and hierarchy are two important concepts \citep{Simon1962,Mallat1989} that have been used to design others ``hand-made'' methods that were used in the past to analyze images \citep{Lowe1999,Dalal2005}.

\paragraph{Plethora of architectures.}
While convolutional neural networks are the model of choice to work with images, they do not exhaust what deep learning is about.
Recently, transformer models \citep{Vaswani2017} have become the model of choice to deal with time series in natural language processing, with amazing realizations in language understanding \citep{Devlin2019} and text generation \citep{Brown2020}, and generative adversarial networks \citep{Goodfellow2014} introduced earlier have also led to impressive applications such as ``deepfakes'' \citep{Karras2019}, which are synthetic images that look real and might be used for malicious purposes.
The rapid spread of deep learning is partly due to its ease to implement through differential programming libraries that implement automatic differentiation and backpropagation when forward propagation is specified \citep{Abadi2016,Paszke2019}.
This allows practitioners to easily develop and test new ideas and architectures, even though training neural networks is known to be quite unstable and greedy in terms of time and energy \citep[see {\em e.g.}][]{Garcia2019}.

\paragraph{Statistical properties.}
While neural networks work better than local averaging and kernel methods, statisticians have not arrived at a consensus in order to describe theoretically why neural networks work so well.
In particular, it is currently hard to state generalization bounds akin~\eqref{par:eq:nn_rates}, \eqref{par:eq:smooth_nn_rates} and \eqref{par:eq:kkr_rates}.
Because of their wide-spread use, many are trying to derive a better understanding of those models.
Some are trying to come up with similar models based on concepts that are better understood such as wavelets \citep{Bruna2013}, kernels \citep{Mairal2014} or sparse coding \citep{Papyan2017}.
Others are looking at the properties of neural networks once the training has taken place \citep{Mahendran2015,Selvaraju2017}, notably discovering the possibility to fool them with imperceptible noise \citep{Szegedy2014,Ilyas2019}.
Finally, statisticians are trying to explain some of their characteristics with different tools, could it be some measure of complexity \citep{Bartlett2017,Zhang2021}, group theory and harmonic analysis \citep{Mallat2016}, statistical physics \citep{Spigler2019} or asymptotic limit \citep{Jacot2018,Chizat2019,Mei2019}.

\section{The practice}
In this section, we discuss the daily job of data scientists before confronting this reality with the prior theory in order to discuss the limitations of the supervised learning framework.

\subsection{The job of data scientists}
There are arguably four steps in a project of a data scientist when it comes to machine learning.
While we present those tasks in a downstream fashion, in practice, it is important to get a big picture and a rough sketch of each step first in order to ease the transitions between steps and avoid going back too often to prior steps.
All along the process, intuition can be found by reflecting on how the work would be accomplished manually, that is without learning perspectives.

\paragraph{1. Define the problem.}
First of all, data scientists should be clear about the task to solve, look at the current solution, and see how it can be improved with machine learning.
In order to prove the usefulness of the new method, they should define some clear measures of success, and benchmark simple baselines.
At this point comes the definition of the output space $\Y$ and of the loss $\ell:\Y\times\Y\to\R$ that should be defined in coherence with the precedent measures of success.
Once the target space $\Y$ is well-defined, the next step is to find a set of relevant features that defines the input space $\X$ and allows prediction of targets.
Of course, the features that can be accessed, hence the definition of $\X$, depend on the data one can get.

\paragraph{2. Get clean data.}
Once $\X$ and $\Y$ are well-defined, it is time to find a set of I/O examples $(X_i, Y_i)_{i\leq n}$.
This is often the most time-consuming task.
How many samples are needed regarding the statistical hardness of the task at hand? Where to scrap data? How to easily associate outputs to inputs?
Once enough samples have been collected, it is usual to set aside a part of those to form a test dataset, that would only be looked at before deployment to validate the working status of the built system.
The remaining data form the train set.
At this point, data is often very cluttered, implying a consequent amount of cleaning before feeding a machine learning model.
In particular, when collected from different sources, there might be some coordinates missing for various input samples.
For example when working with countries, it will be hard to get an indication of the number of people below poverty line in North Korea.
A decision should be made with those missing data, could it be filling with the feature average, or dropping completely the corresponding sample.
Similarly, the practitioners might have to deal with outliers or data sources that got stalled on wrong values.

\paragraph{3. Train a model.}
Once clean data have been collected, machine learning models are ready to be trained.
In order to help models, some features engineering might be useful, such as centering and normalizing the variance of the data, or applying more advanced transformations to get samples that are well-behaved, {\em e.g.} following a unit Gaussian distribution.
It is usual to combine different models together, when those models capture different relationships between inputs and outputs, in order to boost the overall performance, a procedure known as boosting \citep{Schapire1990}.
Hyperparameters are generally tuned with cross-validation.
Cross-validation consists in splitting the samples into a given number of folds, using all the folds but one to train the models and getting a measure of success on the last fold before averaging this measure over the different folds that can be retained for testing.
This associates a score to a method with a given set of hyperparameters.
Best hyperparameters are chosen as the ones leading to the highest score.
Practitioners should be careful not to try too many models or hyperparameters as the more they try, the more likely will be learned spurious correlations that only explain the training samples I/O relationship.

\paragraph{4. Deploy the model.}
Once a model has been learned to predict output from input, one should excavate the test dataset, and see if the method performs better than baselines.
If so, it is time to benefit from this new method and put it into production.
At this point, it is important to monitor the quality of new input data, to make sure that it is not corrupted, and that its distribution is similar to the distribution of training data.
There might also be an emphasis on engineering problems, such as managing databases and computing architectures.

For the curious reader, there are many good references to prepare future data scientists for their job \cite[see {\em e.g.}][]{Geron2017}.

\subsection{Framework limitations}

\paragraph{Loss design.}
In the formalization of supervised learning that we gave earlier, we assumed that the loss $\ell:\Y\times\Y\to\R$ is clearly defined and that the risk ${\cal R}$ to minimize is nothing but the average over samples of the error made by a predictor measured with this loss.
In some critical applications, such as medical one, one might prefer to build ${\cal R}_{{\cal D}_n}$ as the worst case of $(\ell(f(X_i), Y_i))$ or build ${\cal R}$ as some quantile of the pushforward distribution $\ell(f(\cdot), \cdot)_\# \rho$ rather than building them as their respective means.
Moreover, in practice, it might be hard to define clearly the loss $\ell$.
For example, it is the case when it comes to style transfer, that is taking two images and outputting an image with the content of the first one and the style of the second, or to super-resolution, that is enhancing the resolution of an image \citep{Johnson2016}.
The same is true with our cooking example, how to access the quality of a recipe derived by a computer? Should it be the most healthy, the most tasty or somewhat healthy and tasty? Can we derive such losses based on chemical considerations? Or should we only measure success based on people reviews, meaning high costs to evaluate the risk ${\cal R}$?

Sometimes, losses can be partially or totally designed by leveraging the structure of the problem.
For example, on problems where there is a desired invariance defined by a set $G$ of functions $g:\X\to\X$ such that we should have $f\circ g = f$, if this invariance is not built into the model, it can be built into the loss by replacing $\ell(f(X), Y)$ with $\sum_{g \in G}\ell(f(g(X)), Y)$, which is notably the idea of ``data augmentation''.
Another example is when we have a parametric model of the relation between $X$ and $Y$, $Y = g(X, \theta, \epsilon)$ for $g$ a known function, $\theta$ an unknown parameter to optimize and $\epsilon$ some random noise.
The parameter $\theta$ can be fitted by maximizing the likelihood of having observed ${\cal D}_n = (X_i, Y_i)_{i\leq n}$ under the model.
Assuming the samples are independent, this probability factorizes as the product of the probability of observing each $(X_i, Y_i)$.
As a consequence, maximizing the likelihood, or equivalently the log-likelihood, is similar to minimizing the risk ${\cal R}$ defined through the loss $\ell(X, Y, \theta) =-\log \Pbb\paren{Y\midvert X; \theta}$ with the last term designing the probability of observing $Y$ given the observation of $X$ under the model parametrized by $\theta$.
When the parametric model is an exponential family, this procedure is to be linked with generalized linear models \citep{Nelder1972}.

\paragraph{Independent identically distributed variables?}
Suppose you want to predict annual growth from indicators, such as the gross domestic product, regarding a country.
You might collect a dataset $(X_{c,t}, Y_{c,t})_{c\in C, t\in T}$ with many countries, indexed by a set $C$, and many years, indexed by a set $T$, features $X$ and corresponding growth $Y$,  but this dataset will violate the independent identically distributed assumption of statistical learning.
For example, an oil crisis happening in year $t$ will have a repercussion on each $(Y_{c,t})_{c\in C}$, and if one try to add oil prices in the features $X$ to enforce the independence of the conditional variables $\paren{Y_{c,t}\midvert X_{c,t}}_{c\in C}$, than the $(X_{c,t})_{c\in C}$ will not be independent as this factor will be constant over countries.
Similarly, for a given country, indicators are arguably forming a non-stationary process that depends on politics in place, hence $(X_{c,t})_{t\in T}$ is not a family of independent random variables.
However, many data scientists might be tempted to still use the many resources provided by supervised learning, without trying to leverage the underlying processes.

More importantly, suppose that one trains a model on western countries, benefiting from higher quality data, before deploying the model to help some decision-making in developing countries.
There might be some important cofactors such as quantitative easing policies, or trust of populations in money, that change completely the picture when it comes to predicting growth from country indicators.
As such, the testing distribution is not the same as the training distribution, a phenomenon known as {\em covariate shift} \citep{Heckman1979,Shimodaira2000}.
Recently, this problem has come back to denounce the risk of automatic systems that reproduce human-biases \citep{Caliskan2017}, which has found an echo in the civil society \citep{Benjamin2019}.

Interestingly, while data scientists often try to artificially squeeze their problems into the framework of supervised learning, many researchers are trying to leverage specific settings, such as the fact that data might be coming from different datasets \citep{Arjovsky2019}.

\paragraph{Practice and theory regarding generalization.}
In theory, generalization error might be given by derivations allowing to get results such as~\eqref{par:eq:nn_rates}, \eqref{par:eq:smooth_nn_rates} and \eqref{par:eq:kkr_rates}.
In practice, those bounds might be hard to compute because of unknown constants, yet they provide precious insights on the number of samples to consider in order to achieve statistical significance.
For example, bounds reading $d/n$ suggests that the number of samples $n$ should be bigger than the number of dimensions $d$ of the input space $\X$, while bounds in $n^{-1/d}$ suggests that the number of samples should be exponentially bigger than the number of dimensions.
Anyhow, data scientists mostly compute generalization scores by computing the empirical risk of a predictor $f_n$ on a fresh test set of data.
According to this practice, researchers might study the benefits of the test set validation procedure.
For example, they may leverage the distribution of the empirical test error in order to transform a predictor $f_n:\X\to\Y$ into a confidence interval predictor $f_n:\X\to 2^\Y$ with a given probability to be valid \citep{Vovk2008}.
\chapter{Learning Rates with Discrete Outputs}

Understanding the reliability of machine learning is crucial to deploy learned models in the wild.
It was also an important point of this thesis that focuses on algorithms for weakly supervised learning with appreciable theoretical guarantees.
This chapter elaborates on such guarantees and discusses some of our work \citep{Cabannes2021b,Cabannes2022b}, which is reproduced entirely in part \ref{part:discrete}.
For simplicity, it is set in the supervised learning formalization made in the precedent chapter.
In particular, we consider $\X\subset\R^d$ an input space, $\Y$ a discrete output space, $\rho$ a joint distribution on $\X\times\Y$, and $\ell$ a loss function.
Our goal is to minimize the risk ${\cal R}:\Y^\X\to\R$ defined as
\begin{equation}
  {\cal R}(f) = \E_{(X, Y)\sim\rho}\bracket{\ell(f(X), Y)}.
\end{equation}

\section{Statistical learning theory}

In this section, we discuss results offered by statistical learning theory to get insights on what can be learned from data.
In particular, we get more precise about classical derivations of generalization bounds.

\subsection{Insights from information theory}
Artificial intelligence designs a putative system created by humans that would display signs of intelligence, whatsoever this means.
Machine learning consists in implementing such a system with a machine that learns, {\em i.e.} finds in a sort of autonomous fashion, how to behave in order to produce a desired output given some input parameters.
Statistical learning consists in presenting the machine with a set of training examples that show desired outputs for different inputs in order for the machine to infer a rule to produce outputs from inputs.
Statistical learning theory borrows many tools from information theory.

Information theory is concerned with transmitting information.
Think that we want to describe a function $f:\X\to\Y$ to a friend, with only a finite number of information.
If we know that $f$ belongs to a finite set of functions ${\cal F}$, based on dichotomy, we only need $\log_2(\card{\cal F})$ bits to encode this function among all functions in ${\cal F}$.
Now, if ${\cal F}$ is continuous, we cannot encode all functions on a finite number of bits.
Yet, if we only want to unravel a signal $f\in{\cal F}$ up to a precision $\epsilon$, given a certain error metric\footnote{Here, the word ``metric'' is used in a prosaic fashion since $L$ might not be a distance.} $L$ (typically $L(\hat{f}, f) = \E\bracket{\ell(\hat{f}(X), Y) - \ell(f(X), Y)}$), we could do so with an encoding on less than $\ceil{\log_2({\cal N}(\epsilon))}$, with ${\cal N}(\epsilon)$ the minimum number of balls of size $\epsilon$ (with the metric $L$) to cover ${\cal F}$.
To quantify the complexity of the set ${\cal F}$, it is natural to look at the behavior of the number of bits we need to encode any function in ${\cal F}$ up to an error $\epsilon$ as $\epsilon$ get to zero.
As a consequence, it is useful to define a notion of size of ${\cal F}$ (with respect to the error metric $L$) as the superior limit
\[
  C_{\cal F} = \limsup_{\epsilon\to 0} - \log_2({\cal N}(\epsilon)) / \log_2(\epsilon).
\]
Note the analogy with statistical mechanics, where signals are microscopic arrangements of unit elements, and the set ${\cal F}$ is the corresponding macroscopic system.
Assuming that all arrangements have the same probability to appear, the Boltzmann entropy is exactly, up to the Boltzmann constant, the logarithm in base two of the cardinality of ${\cal F}$.
This explains the wording Kolmogorov entropy for the logarithm in base two of the covering number ${\cal N}$.

To make the preceding paragraph more concrete, suppose that we want to transmit a function $f:[0,1]^d \to \R$ to our friend.
Typically, we would give to our friend $m$ bits of discrete information, plus some information on the function class ({\em e.g.} $f$ is a polynomial of order $d$, or, thinking in terms of Fourier transform, it has a low energy on high harmonics).
For example, assume that $f$ belongs to the Sobolev space $H^m(\diff x)$, {\em i.e.} it is $m$ times differentiable, with derivatives up to order $m$ square integrable against the Lebesgue measure.
Suppose that our friend knows that $\norm{f}_{H^m} \leq 1$, and suppose that we want to minimize the $L^2$ error. Then, we can give $n$ binary information to localize this function in a covering of ${\cal F} = \brace{f\midvert \norm{f}_{H^m} \leq 1}$ with $2^n$ $L^2$-balls.\footnote{Note that such a covering of ${\cal F}$ can be taken as it is compact with respect to the $L^2$-topology - even though it is not compact with the $H^m$-topology, $H^m$ being infinite dimensional.}
At best, our friend will be able to localize the signal in an $L^2$-ball of radius $\epsilon(2^n)$, thus making at most an error $\epsilon(2^n)$ on the reconstructed signal, with
\(
\epsilon(n) = \inf \bracket{\epsilon \midvert {\cal N}(\epsilon) \leq n} \approx n^{-C_{\cal F}},
\)
and $C_{\cal F}$ the size of ${\cal F}$ with respect to the $L^2$-metric.
We refer the interested reader to the seminal paper of \cite{Kolmogorov1959} for precise quantification of space sizes.

\subsection{Vapnik-Chervonenkis theory}

Machine learning is concerned with inferring information.
In this setting, we want to learn a task $f:\X\to\Y$, and we do so by collecting examples of the solved instance of this task $(x_i, y_i = f(x_i))$.
In the current machine learning paradigm, we do not choose the bits of information to transmit, but we are given some points $(x_i, y_i)_{i\leq n}$ that are assumed independently sampled according to the distribution that matters to us.
Let us review the picture offered by the classical statistical learning theory for the sake of completeness.
Recall that we want to study the excess of risk. We have, with $f_{\cal F}^*$ and $f_n$ the respective minimizers of the population and empirical risks over ${\cal F}$,
\[
  {\cal R}(f_n) - {\cal R}(f_{\cal F}^*) \leq
  {\cal R}(f_n) - {\cal R}_{{\cal D}_n}(f_n) +
  {\cal R}_{{\cal D}_n}(f^*_{\cal F}) - {\cal R}(f^*_{\cal F}).
\]
For any function $f\in{\cal F}$, we can use concentration inequality (such as Bernstein inequality) to control the difference between the empirical risk ${\cal R}_{{\cal D}_n} (f)$, which inherits its randomness from ${\cal D}_n$, and its average ${\cal R}(f)$.
This works well for the second part of the equation, due to $f^*_{\cal F}$.
Sadly, as $f_n$ depends on ${\cal D}_n$, we can not apply concentration inequality directly to control the deviation between its empirical and population risk.
The classical way to proceed is to find a uniform concentration bound over ${\cal F}$, which we can do by controlling the following random quantity
\[
  \sup_{f\in{\cal F}} {\cal R}(f) - {\cal R}_{{\cal D}_n}(f).
\]
When ${\cal F}$ is finite, we can control this supremum with a union bound, joining together all individual concentration inequality.
Similarly, when ${\cal F}$ is continuous, we could do similar things with a well-specified $\epsilon$-cover of ${\cal F}$, which can be refined based on chaining techniques, to avoid redundancy of events when performing union bound for supremum.
In the statistical learning literature, it is classical to rather use a symmetrization trick to relate ${\cal R}(f) - {\cal R}_{{\cal D}_n}(f)$ to ${\cal R}_{{\cal D}'_n}(f) - {\cal R}_{{\cal D}_n}(f)$ for ${\cal D}'_n$ another dataset, and study the Rademacher complexity defined as $\E\sup_{f\in{\cal F}} n^{-1} \sum_{i=1}^n \epsilon_i f(X_i)$ for $\E$ the expectation taken over the $\epsilon_i$ defined as variables taking value one or minus one with probability one half.

In the case of binary classification with the 0-1 loss, {\em i.e.} $\Y = \brace{-1, 1}$, $\ell(y, y') = \ind{y\neq y'}$, Vapnik and Chervonenkis proposed slightly different derivations leading to the following bound.
\begin{equation}
  \label{eq:vcbound}
  \E_{{\cal D}_n}{\cal R}(f_n) - {\cal R}(f_{\cal F}^*) \leq
  c\sqrt{\frac{V_{\cal F}}{n}},
\end{equation}
for $c$ a universal constant, and $V_{\cal F}$ the so-called VC dimension of ${\cal F}$, that relates to the average number of points that the class ${\cal F}$ is able to shatter \citep{Vapnik1995}.

A natural question with generalization bounds is how well they quantify the behavior of the excess of risk ${\cal R}(f_n) - {\cal R}^*$.
Indeed, the estimation of the excess of risk given by~\eqref{eq:vcbound} is known to be optimal, in the sense that, for any class of functions ${\cal F}$, it is possible to find a specific distribution $\rho$ such that this bound is a lower bound up to a multiplicative constant.
This behavior is referred to as {\em minimax optimality}, meaning that it is not possible to find a better bound, given the class ${\cal F}$ of estimators considered.

\section{Surrogate methods}

In this section, we discuss the difficulty of learning discrete-valued functions, and the possibility to tackle the learning problem through surrogate problems that consist of learning continuous-valued functions.

\subsection{Practical limitations of VC theory}
While VC theory was a keystone in machine learning, it does not exhaust practitioner issues, as this theory does suffer from some limitations.

\paragraph{Approximation/estimation trade-off.}
First of all, VC theory controls the estimation error in~\eqref{eq:risk-dec}, that is the error in terms of population risk ${\cal R}$ between $f_n$ the minimizer of the empirical risk in a hypothesis class ${\cal F}$ and $f_{\cal F}^*$ the minimizer of the population risk in the class ${\cal F}$.
In particular, equation \eqref{eq:vcbound} does not tell anything about the approximation error, that is the gap between ${\cal R}(f_{\cal F}^*)$ and ${\cal R}^*$.
To control the approximation error, we need to make assumptions on how good our model ${\cal F}$ is to minimize the risk ${\cal R}$.
There is a trade-off between choosing a big model ${\cal F}$, so that the optimal risk ${\cal R}^*$ can be achieved by ${\cal R}(f_{\cal F}^*)$ inside the hypothesis class, and choosing a small model, so that its capacity, captured by $C_{\cal F}$ or ${\cal V}_{\cal F}$, is small.

\paragraph{Optimization issues.}
Finally, if we were to apply VC theory to learn from a continuous input space $\X \subset \R^d$, to a discrete output space $\Y$, we could consider a model ${\cal F}$ of functions from $\X$ to $\Y$.
Those functions could be characterized through their decision regions and decisions boundaries, defined for $y, z \in \Y$ as
\[
  f^{-1}(y) = \brace{x\in\X \midvert f(x) = y} \subset \X,\qquad
  \Delta_{y, z}(f) = \overline{f^{-1}(y)} \cap \overline{f^{-1}(z)} \subset \X,
\]
where the bar stands for the closure of the set.
As such, a model ${\cal F}$ could be defined as a collection of putative decision regions.
The region of disagreement between two functions $f$ and $g$ for a prediction $y\in \Y$ can be expressed through the symmetric difference ${f^{-1}(y) \triangle g^{-1}(y)}$.
When using the 0-1 loss, the excess of risk between $f_n$ and $f^*_{\cal F}$ is controlled by the measure of the union of disagreement regions, that is $\rho_\X(\cup_{y\in\Y} (f_n)^{-1}(y)\triangle (f_{\cal F}^*)^{-1}(y))$.
It is then possible to retake the VC theory, using the covering number of ${\cal F}$ with respect to this pseudo-distance \citep{Mammen1999}.
Sadly, even for a simple model ${\cal F}$, such as binary classification with half-plane decision regions, minimizing the empirical risk is NP-hard \citep{Arora1997}.
This echoes optimization issues in machine learning. While VC theory asks to consider the empirical risk minimizer $f_n$, it might be really hard to find it in practice, and one might have to settle with an estimate $\hat{f}$ of $f_n$.
This adds a third term to the risk decomposition that is an optimization error term
\[
  {\cal R}(\hat{f}) - {\cal R}^* =
  \underbrace{{\cal R}(\hat{f}) - {\cal R}(f_n)}_{\text{optimization error}}
  \ + \underbrace{{\cal R}(f_n) - {\cal R}(f_{\text{lin}}^*)}_{\text{estimation error}}
  \ + \underbrace{{\cal R}(f_{\text{lin}}^*) - {\cal R}^*}_{\text{approximation error}}.
\]

\subsection{Related continuous surrogate problems}
When learning with discrete outputs, direct risk minimization leads to many combinatorial difficulties that translate into computational intractability.
This is related to the difficulties of studying and optimizing discrete-valued functions.
One technique to overcome this issue consists in learning a discrete-valued function, by learning a surrogate continuous-valued function and thresholding its output in order to make it discrete.
Consider binary classification, that is $\Y = \brace{-1, 1}$ and $\ell(y, z) = \ind{y\neq z}$.
In this setting, the optimal predictor $f^*:\X\to\Y$ is defined as
\[
  f^*(x) = \sign(g^*(x)), \qquad\text{where}\qquad
  g^*(x) = \E_{(X, Y)\sim\rho}\bracket{Y \midvert X=x}.
\]
This suggests learning the discrete-valued function $f^*:\X\to\Y$ by learning the continuous-valued function $g^*:\X\to\R$.
To an estimate $g$ of $g^*$, we associate the estimate $f = \sign(g)$ of $f^*$.
One method to learn the conditional expectation $g^*$ is through its characterization with the least-squares error
\[
  g^* \in \argmin_{g:\X\to\R} {\cal R}_S(g) := \E_{(X, Y)\sim\rho}\bracket{\abs{g(X) - Y}^2}.
\]
Given data $(X_i, Y_i)$, one can then consider a model of real-valued functions ${\cal G} \subset \R^\X$ and the empirical risk minimization
\[
  \hat{g} \in \argmin_{g\in{\cal G}} \frac{1}{n}\sum_{i=1}^n \abs{g(X_i) - Y_i}^2.
\]
This estimate is cast as an estimate of $f^*$ through the decoding $\hat{f} = \sign\hat{g}$.
It is natural to wonder how a good estimate of $g^*$ translate into a good estimate of $f^*$.
This question is answered by calibration inequalities that relate the excess of risk on the surrogate problem with the excess on risk of the original problem \citep{Bartlett2006}.
For example, for the least-squares surrogate considered here, we have
\[
  {\cal R}(\sign g) - {\cal R}(f^*) \leq \sqrt{{\cal R}_S(g) - {\cal R}_S(g^*)}.
\]
While we only present the least-squares surrogate for binary classification, other surrogates such as the hinge loss, leading to support vector machine, the logistic loss, leading to softmax regression, can be considered.
Similarly, surrogate methods can be extended beyond binary classification, for example, in multiclass problem with the $0-1$ loss, that is $\Y = \brace{1, \cdots, m}$ and $\ell(y, z) = \ind{y\neq z}$, we can consider $g^*:\X\to\R^\Y$, defined as
\[
  g^*(x) = \paren{\Pbb\paren{Y=y\midvert X=x}}_{y\in\Y}
  = \E\bracket{(\ind{Y=y})_{y\in\Y} \midvert X=x},
\]
and the decoding
\[
  f^*(x) = \argmax_{y\in\Y} g^*_y(x).
\]
In such a setting, $g^*$ is often referred to as a {\em score} function.

\begin{remark}[The pros of predicting scores]
  From a formal perspective, if we are only interested in the optimal mapping $f:\X\to\Y$, learning $g$ can be seen as a waste of resources.
  In essence, this waste of resources is similar to the one when learning the full probability function $(p(y))_{y\in\Y}$ for some $p \in\prob{\Y}$ while we only care about the argmax $y^* \in \argmax_{y\in\Y} p(y)$.
  In practice, however, it might be of interest to get an estimate of $g^*(x) = \Pbb\paren{Y=y\midvert X=x}$, as it might provide important information about how specified is $f^*(x)$ and how we could confidently discard other potential labels $y$ for the input $x$.
\end{remark}

We refer the interested reader to \cite{Nowak2021} for deeper reflections about learning with surrogate methods.

\section{Fast rates derivation}
In this section, we sketch roughly the underlying machinery beyond \citet{Cabannes2021b}.
While writing this section, we gave a fresh look at it to avoid redundancy with results already published and reproduced in Chapter~\ref{chap:fast}.
Confronting this machinery with SVM has led us to~\cite{Cabannes2022b} which is reproduced in Chapter~\ref{chap:svm}.

\paragraph{Regularized risk minimization.}
For simplicity, consider a surrogate problem consisting of learning a real-valued function; with a linear parametric model of functions  ${\cal G} =\brace{g_\theta: x\to \scap{\theta}{\phi(x)}\midvert \theta\in\Theta}\subset \R^\X$, parametrized by some Hilbert space $\Theta$, and some features $\phi:\X\to\Theta$.
Consider the regularized empirical risk minimization
\[
  \theta_n \in \argmin_{\theta} {\cal R}_{S, {\cal D}_n}(g_\theta) + \lambda_n \norm{\theta}^2.
\]
Here, $\lambda_n$ is a regularization parameter that goes to zeros as the number of samples $n$ goes to infinity and the risk of overfitting vanishes, and ${\cal R}_{S, {\cal D}_n}$ is the empirical surrogate risk computed from the dataset ${\cal D}_n$.
Eventually, we could also consider $\hat{\theta}$ an approximation of $\theta_n$ related to some optimization techniques.
Finally, it is useful to introduce the bias estimate
$
  \theta_\lambda \in \argmin_{\theta} {\cal R}_S(g_\theta) + \lambda \norm{\theta}^2.
$

\paragraph{Classical versus new convergence study.}
Classically, to prove that $\hat{f}$, the decoding of $g_{\hat{\theta}}$, will minimize the original risk ${\cal R}$ as $n$ goes to infinity and to give rates of convergence, one can use capacity/estimation assumptions to state that $\hat{\theta}$ concentrate around $\theta_\lambda$, as well as source/approximation assumptions to state that ${\cal R}_S(g_{\theta_\lambda})$ will convergence to ${\cal R}_S(g^*)$ as $\lambda$ goes to zero.
Finally, using calibration inequalities, one can relate concentration in $\theta$ to concentration in ${\cal R}_S$, and then to concentration in ${\cal R}$.
The problem with this technique is that cascading calibration inequalities can lead to quite suboptimal rates, as shown by the work of \citet{Audibert2007}, which bypassed this procedure, and, under a simple well-thought margin condition, derived dramatically faster convergence rates than classical ones.
In our view, this work can be generalized by deriving directly calibration inequality that relates the original excess of risk to concentration in the parameter space.
The specificity of those new calibration inequalities is that they are not universal, but depends on some approximation hypothesis that quantifies how hard or easy it is to solve the original problem based on the parametrized surrogate problem.
The most well-known such hypothesis is the Tsybakov margin condition, but others were proposed \citep[for example by][]{Steinwart2007}.

\begin{figure}
  \centering
  \begin{tikzpicture} 
  
  \filldraw[yellow] (1.9, 0) circle (1);

  \draw[dashed] (0, 0) ellipse (.5 and .5);
  \draw[dashed] (.2, 0) ellipse (.9 and .75);
  \draw[dashed] (.4, 0) ellipse (1.3 and 1);
  \draw[dashed] (.6, 0) ellipse (1.7 and 1.25);
  \draw[densely dotted,red,thick] (.8, 0) ellipse (2.1 and 1.5);
  \draw[dashed] (1, 0) ellipse (2.5 and 1.75);

  \begin{scope}[decoration={markings, 
      mark= at position .15 with {\arrow{latex}},
      mark= at position .35 with {\arrow{latex}},
      mark= at position .55 with {\arrow{latex}},
      mark= at position .75 with {\arrow{latex}},
      mark= at position .95 with {\arrow{latex}}}
    ]
    \draw [postaction={decorate},gray,thick] (4,0) -- (0,0);
  \end{scope}

  \filldraw (0, 0) circle (2pt) node[anchor=north west] {$\theta^*$};
  \filldraw (1.9, 0) circle (2pt) node[anchor=north west] {$\theta_\lambda$};

  \draw[dashed] (4, 2) -- (5, 2) node[anchor=west] {level lines of ${\cal R}(\sign(g_n))$};
  \begin{scope}[decoration={markings, 
      mark= at position .5with {\arrow{latex}}}
    ]
    \draw[gray,thick,postaction={decorate}] (5, 1.5) node[anchor=west,black] {path $\brace{\theta_\lambda; \lambda\in[0, 4]}$} -- (4, 1.5);
  \end{scope}
  \filldraw[yellow] (4.5, 1) circle (.25);
  \draw (5, 1) node[anchor=west,black]{region $\brace{\theta; \|\theta - \theta_\lambda \| < 1}$};
 
  \draw[densely dotted,red,thick] (4, .5) -- (5, .5) node[anchor=west,black] {Certified value of ${\cal R}(g_n)$ when $\|\theta_n - \theta_\lambda\| < 1$};

\end{tikzpicture}
  \caption{Our new convergence analysis consists in relating natural concentration given by the surrogate method to the original excess of risk without passing by the surrogate excess of risk.
    As the drawing shows, concentration in parameter space $\Theta$ can be cast as deviation on the original excess of risk.
    Yet, this casting relation depends on the geometry of this picture, which itself depends on which surrogate is used, what is the function to learn, how the bias estimator approaches it, and how our empirical estimate concentrates around the bias estimator.
  }
  \label{fig:discrete_analysis}
\end{figure}
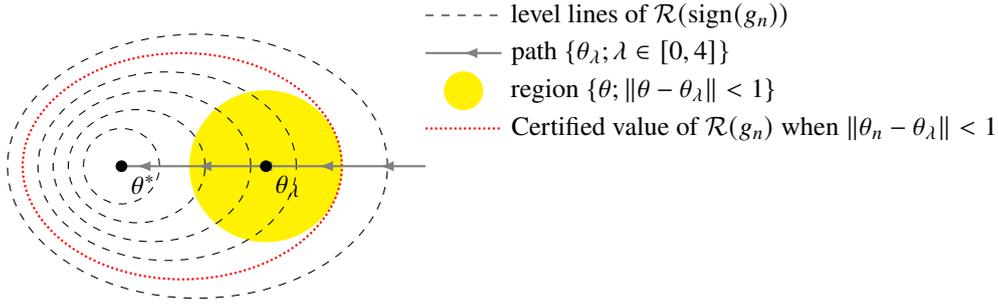

\paragraph{Special case of exponential convergence rates.}
To give more light on the previous paragraph, consider the binary classification problem. Suppose that there exists $\lambda$ such that $\sign(g_{\theta_\lambda}) = \sign(g^*)$.
Suppose also that there exists $\epsilon > 0$ such that $\norm{\theta - \theta_\lambda} \leq \epsilon$ implies that $\sign(g_{\theta}) = \sign(g_{\theta_\lambda})$.
With those two approximation hypothesis, we have the following calibration inequality
\[
  {\cal R}(\sign(g_\theta)) - {\cal R}(f^*) \leq
  \ind{\norm{\theta - \theta_\lambda} > \epsilon}.
\]
As a consequence,
\(
\E_{{\cal D}_n}\bracket{{\cal R}(\sign(g_{\theta_{{\cal D}_n}}))} - {\cal R}(f^*) \leq
\Pbb_{{\cal D}_n}\paren{\norm{\theta_{{\cal D}_n} - \theta_\lambda} > \epsilon}.
\)
Assuming that this last $\theta_{{\cal D}_n}$ concentrated around $\theta_\lambda$ with sub-Gaussian tails, that is
\(
\Pbb_{{\cal D}_n}\paren{\norm{\theta_{{\cal D}_n} - \theta_\lambda} >
  \epsilon}
\leq \exp(-cn\epsilon^2),
\)
for a constant $c$, we get exponential convergence rates.

\section{Discussion}

In this section, we discuss the practical usefulness of generalization bounds, and the different paradigms to define and quantify learnability.

\paragraph{Practical bounds on expected error.}
In theory, generalization bounds provide guarantees on how much error we might expect when deploying a model learned from data.
In practice, those bounds are rather taken as indications that the learning methods are sound, but rates are not reported in order to get confidence on the learned models.
This might have to do with the fact that bounds often depend on parameters or constants that are hard to know in practice. As such, practitioners often prefer to derive error indications from test samples.
In this line, research on {\em conformal prediction} is trying to leverage test samples to provide useful confidence information on learned models.

\paragraph{Assumptions to quantify learnability.}
Given data, it is highly valuable to get an idea of what can be learned from it.
Of course, learnability depends on regularity assumptions of the problem at hand.
Under such assumptions, lower bounds given by minimax rates are supposed to give optimal baselines for learning rates, thus to quantify learnability.
In practice, those lower bounds are derived by considering the most degenerate function respecting those assumptions.
Sometimes a simple well-thought additional assumption can lead to much better bounds.
As such, is there a natural paradigm of assumptions on the function to learn to discuss its learnability?
For example, assumptions on the surrogate solution $g^*$ are not intrinsic to the original problem but depends on the surrogate problem that is considered.
On the contrary, for a fixed decoding $d$, one can quantify the regularity of $f$ through the most regular function $g$ such that $f = d(g)$ (think of the binary classification case with $d = \sign$), this will correspond to the regularity of the decision frontier, rather than the one of the conditional expectation $x\to\E\bracket{Y\midvert X=x}$.
We refer to the paragraph ``Beyond least-squares'' in Section \ref{lap:sec:structured_prediction} for a practical example of such considerations.
This echoes the difference between realizable-consistency and Fisher-consistency of surrogate methods \citep{Long2013}.

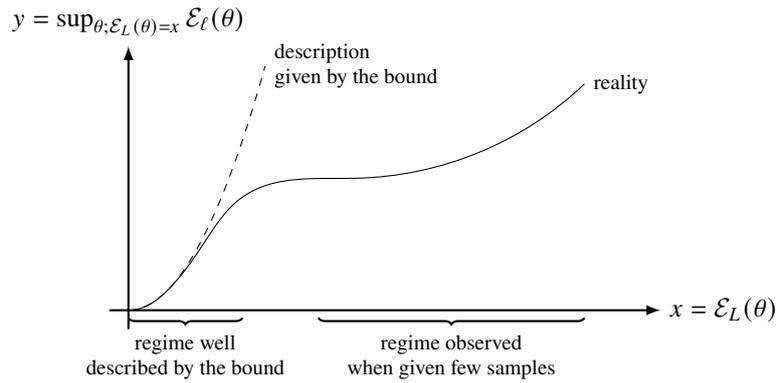
\begin{figure}[t]
  \centering
  \begin{tikzpicture} 
  \draw[-latex, thick] (0, -.25) -- (0, 3.5) node[anchor=south] {$y = \sup_{\theta; {\cal E}_L(\theta) = x} {\cal E}_\ell(\theta)$};
  \draw[-latex, thick] (-.25, 0) -- (7, 0) node[anchor=west] {$x = {\cal E}_L(\theta)$};
  \draw[dashed] plot[smooth,domain=0:1.8] (\x, {\x^2}) node[anchor=west,align=left,font=\footnotesize] {description\\ given by the bound};
  
  \draw (0, 0) to [curve through = {(.1, .01) .. (.2, .04) .. (.3, .09) .. (.4, .16) .. (.5, .25) .. (1, .9) .. (1.5, 1.5) .. (3, 1.75)}] (6, 3);
  \draw (6, 3) node[anchor=west,font=\footnotesize] {reality};

  \draw [decoration={brace,mirror}, decorate, thick] (0, -.1) -- (1.5, -.1);
  \draw [decoration={brace,mirror}, decorate, thick] (2.5, -.1) -- (6, -.1);
  \draw (.75, -.2) node[anchor=north,align=center,font=\footnotesize] {regime well\\ described by the bound};
  \draw (4.25, -.2) node[anchor=north,align=center,font=\footnotesize] {regime observed\\ when given few samples};
\end{tikzpicture}
  \caption{
    The behavior described by generalization bound might be relevant only when accessing an indecently large number of samples, thus being of little use in practice.
  }
  \label{fig:sur_bounds}
\end{figure}

\paragraph{Non-asymptoticness of finite-sample bounds?}
There are two points to mention about probably approximately correct bounds.
First they are derived by considering worst-cases, and as we said in the previous paragraph, simple assumptions often allow refining those bounds by removing pathological cases.
Second they tend to describe the behavior of how risks decrease with the number of samples when accessing numerous samples. As such, while those bounds are non-asymptotic in the sense that they are valid for any number of samples $n$, the described regime might not take place without accessing an indecent number of data, thus being of little use in practice.

To get more concrete, consider a loss $\ell$ and a surrogate loss $L$.
Suppose that we want to bound the excess of risk ${\cal E}_\ell(\theta)$ on the original problem with the excess of risk ${\cal E}_L(\theta)$ on the surrogate problem, where $\theta$ is a learned parameter.
The goal of calibration theory is to find the best function $\xi$ such that for any $\theta$, ${\cal E}_\ell(\theta) \leq \xi({\cal E}_L(\theta))$.
In practice, people try to find $\xi$ concave (because the inequality can be derived for a single $x$ and propagated to the whole space thanks to Jensen inequality), and will look for it under the form $\xi(x) = cx^\alpha$ with $c > 0$ and $\alpha \in (0, 1]$, trying to get the highest $\alpha$ -- that is trying to optimize the behavior of $\xi$ close to zero.
This will lead to somewhat tight bounds only in the regime where ${\cal E}_L(\theta)$ is small, a regime which might not be reached without an indecent number of samples.
We illustrate the regime where constants matter more than exponents on Figure \ref{fig:sur_bounds}.

In practice, it would be more useful to take into consideration the number of samples first, before looking for a function $\xi$ that verifies ${\cal E}_\ell(\theta) \leq \xi({\cal E}_L(\theta))$ for any $\theta$, and minimizes $\max\xi([a, b])$ for $[a,b]$ the range of value that we expect the surrogate risk to take given our fixed number of samples.

\chapter{Partially Supervised Learning}

Many data scientists spend more time scrapping, annotating and cleaning data than fine-tuning models.
This motivates the following question: can we derive a more generic framework than the one of supervised learning in order to learn from cluttered data?
In this thesis, this question is approached through the lens of weakly supervised learning, assuming that the bottleneck of data collection lies in data annotation.
This chapter summarizes our main contributions to weakly supervised learning.
It is based on our published work \cite{Cabannes2020,Cabannes2021,Cabannes2021c}, which is reproduced in part \ref{part:weakly}.

\section{Navigating frameworks of weakly supervised learning}

In this section, we give a brief introduction to weakly supervised learning.

Weakly supervised learning is concerned with the setting where it is easy to get $n$ inputs $(X_i)_{i\leq n}$, but it is hard to get the corresponding outputs $(Y_i)_{i\leq n}$, a setting which is relevant for many applications.
For example, when dealing with images, one can easily scrap zillions of images on the web, which provides many input data, but getting the corresponding outputs demands laborious annotations of data.
The situation is similar when it comes to medical applications, such as understanding drug interactions \citep{CTDPfizer2013} or recognizing cancers with radiography.
While one can access databases from hospitals to get many X-ray images, recognizing cancers on those requires the expertise of a radiologist, which is expensive to access.
Indeed, annotation costs can critically add up when one plans to use powerful models that require millions of data points to be trained.
As a counter example, notice that weakly supervised learning does not help when input data are features coming from heterogeneous sources leading to missing data.
We shall discern three types of weak supervision, the last two being not completely disjoint.

\paragraph{1. Global statistics on groups of inputs.}
This setting consists in accessing global information on bags of samples -- {\em e.g.} knowing that half of the $(Y_i)_{i\in I}$ are zeros for a given $I\subset[n]$.
Examples of global statistics supervision include multiple-instance learning \citep{Dietterich1997} and learning from label proportion \citep{Quadrianto2009}.
We refer to those articles for motivations, descriptions and applications of those two problems.

\paragraph{2. Weak classifiers.}
A second approach consists in assuming the access to many weak classifiers $\phi_j:\X\to\Y$ for $j\in[p]$ that weakly correlate with $f^*$.
Those classifiers might model labelers from a crowdsourcing platform, experts or noisy measurements.
More generally, they might be provided by side information, coming from {\em distant supervision} \citep{Mintz2009,Craven1999} or {\em transferred} from algorithms that have been designed for a related task \citep{Pan2010}.
Recently, \cite{Ratner2020} have implemented a software interface based on this approach.

\paragraph{3. Incomplete annotation.}
Finally, weak supervision might be understood as the access to partial knowledge on what $Y_i$ is for each $i\in[n]$.
In some instances, partial observation on $Y_i$ can be cast as a set of potential labels $S_i\subset\Y$ that are compatible with this partial observation, which is the setting of partial supervision \citep{Cour2011}.
Partial supervision is a generalization of semi-supervised learning, which has been the classical approach to overcome the bottleneck of data annotation \citep{Chapelle2006} and consists in learning from a set of labeled data, $S_i=\brace{Y_i}$ for $i \in L\subset[n]$, and of unlabeled data, $S_i = \Y$ for $i\in [n]\setminus L$.

Beyond those three settings, limitations that motivate weakly supervised learning might be tackled by leveraging human knowledge under the form of priors \citep{Mann2010} or of function architectures \citep[reviving old approaches of artificial intelligence such as][]{Muggleton1994}.

\section{The importance to create consensus}

In this section, we review our first approach to the problem of learning from partial supervision. We first formalize this setting before introducing a variational objective to minimize and providing guarantee regarding this approach.

\subsection{Partial supervision setting}
In this thesis, hence in the following, we focus on partial supervision, which is also known as {\em superset learning}.
In other terms, we assume the access to weak supervision for each label $y$ under the form of a set $s\subset\Y$ that contains the true labels $y \in s$.
To formalize this problem, we introduce the set ${\cal S}\subset 2^\Y$ of closed subsets of $\Y$ and the distribution $\tau\in\prob{\X\times{\cal S}}$ generating samples $(X_i, S_i)_{i\leq n} \sim \tau^{\otimes n}$.
The goal is still to minimize the risk
\begin{equation}
  \label{int:eq:risk}
  {\cal R}(f) = \E_{(X, Y)\sim\rho}\bracket{\ell(f(X), Y)},
\end{equation}
for a known loss function $\ell:\Y\times\Y\to\R$ and a distribution on I/O pairs $\rho\in\prob{\X\times\Y}$.
In contrast with supervised learning, $\rho$ will never be accessed, nor any samples $(X_i, Y_i) \sim \rho$.

In order to motivate our theoretical setting, we now discuss simple instances of partial supervision.

\begin{example}[Classification with attributes]
  Suppose that you want to learn fine-grained classes on images, {\em e.g.} to precisely distinguish ``caracals'' and ``domestic cats'', as well as ``hedgehog mushrooms'' and ``shiitakes''.
  It will probably be hard for commoners hired on crowdsourcing platforms to label images with precise classes.
  Yet, this commoner might easily label attributes, such as ``tufted ears'' characterizing ``caracals'' among ``felines'', or ``spines rather gills underside of the cap'' and ``tan irregular cap'' characterizing ``hedgehogs'' among ``mushrooms''.
  Those partial labels can easily be cast as sets of labels: ``feline'' would  be $\brace{\text{``lion'', ``panther'', \dots}}$ and ``tufted ears''  be $\brace{\text{``great horned owl'', ``Araucana chicken'', \dots}}$.
  The key idea here is that it is much cheaper to get a set of many partial labels that is informative enough to learn $f^*$ than a set of few complete labels enabling the same learning.
\end{example}

\begin{example}[Ranking with partial ordering]
  Consider a ranking problem where, given a user characterized by some features $x$, the goal is to learn their preference over $m$ flowers, defining the output $y\in\Sfrak_m$ as an ordering of $m$ elements.
  Once again, getting the full $y$ is a laborious task, as one might have a hard time to rank all $m$ flowers. In contrast, one might easily provide partial ordering information, such as ``I prefer roses to lilies'', or ``tulips are my favorite'', which can be cast as the set of total orderings that verify those partial orderings.
\end{example}

\begin{example}[Regression with censored data]
  For many regression problems, it is not possible to get exact labels but only possible to access bins or censored labels $[a,b] \ni y$.
  This could be due to the use of a measuring scale, or, if retaking example \ref{ex:housing}, due to some uncertainty regarding a price at which a house has been sold.
\end{example}

\subsection{Solution definition through the infimum loss}

Currently, the problem $(\ell, \tau)$ is ill-defined since we cannot define clearly the goal \eqref{int:eq:risk} from those two objects only -- $\ell$ being a loss function on pairs of outputs and $\tau$ a distribution on $(X, S)$.
One way to redefine a solution in this context is to keep the precedent variational point of view and define a loss $L:\Y\times 2^{\Y}\to\R$ that given a prediction $z\in \Y$ and an observation $S\subset \Y$, provides a compatibility or performance score.
Hence the new objective
\[
  {\cal R}_S(f) = \E_{(X, S)\sim\tau}\bracket{L(f(X), S)}.
\]
Yet, how to derive the loss $L(z, S)$ for $z\in\Y$ and $S\subset\Y$ when the original task was specified by the loss $\ell(z, y)$ with $y\in S$?
Arguably there are three possibilities.
\begin{enumerate}
  \item Bounding the original excess risk from above with the supremum loss $L(z, S) = \sup_{y\in S}\ell(z, y)$.
        Suprema have been used in statistics for robustness purposes \citep{Wald1945}.
        The idea here is that if we prevent ourselves against the worst possible candidates (in terms of error given a prediction $z$), we will prevent ourselves against whatever is the ground truth label in $S$.
        Yet, as we will discuss later, this approach is too conservative in our setting.
  \item Matching a bit of all the elements in the set $S$ with the average loss $L(z, S) = \frac{1}{\abs{S}} \sum_{y\in S} \ell(z, S)$.
        This loss has the benefit of being agnostic on what should be the true $y\in S$ regardless of what the prediction $z$ is.
        Under symmetry assumptions of the original loss, for example when $\ell$ is the 0-1 loss in classification, averaging candidates is a reasonable solution.
        When the loss is not symmetric, it can insidiously bias the solution as we describe on Figure \ref{il:fig:inconsistency}.
        In many cases, it is possible to correct for this asymmetry, which techniques implicitly implied by Proposition \ref{df:prop:init} and illustrated with the filling with zero techniques in section \ref{df:sec:ranking}.
  \item Making sure that the prediction $z$ matches at least one element with the infimum loss $L(z, S) = \inf_{y\in S} \ell(z, S)$.
        When $\ell$ is seen as a distance, $L$ is its natural extension to sets.
        \cite{Hullermeier2014} has referred to it as the optimistic loss, since given a prediction $z$, it somehow disambiguates $y\in S$ by considering the best label in order to minimize $\ell(z, y)$ under the constraints $y\in S$.
\end{enumerate}

\begin{figure}
  \centering
  \begin{tikzpicture} 


  \filldraw[gray!30, rotate=45] (0, 0) ellipse (2 and 1.3);
  \filldraw[gray!50] (3, .7) circle (.7);

  \filldraw[gray] (0, 0) circle (.5pt) node[anchor=south] {$y_1^{a}$};
  \filldraw[gray] (1.7, .7) circle (.5pt) node[anchor=south] {$y_1^{i}$};

  \filldraw[gray] (3, .7) circle (.5pt) node[anchor=south] {$y_2^{a}$};
  \filldraw[gray] (2.3, .7) circle (.5pt) node[anchor=south] {$y_2^{i}$};

  \filldraw (1.5, .35) circle (1pt) node[anchor=north] {$z^{a}$};
  \filldraw (2, .7) circle (1pt) node[anchor=north] {$z^{i}$};
\end{tikzpicture}
  \hspace{2cm}
  \begin{tikzpicture} 


  \filldraw[gray!30] (0, 0) ellipse (2 and 1.3);
  \filldraw[gray!50] (1.5, 0) circle (.4);

  \filldraw[gray] (0, 0) circle (.3pt) node[anchor=south] {$y_1^{a}$};
  \filldraw[gray] (1.5, 0) circle (.3pt) node[anchor=south] {$y_2^{a}$};

  \filldraw (.75, 0) circle (1pt) node[anchor=north] {$z^{a}$};
  \filldraw (1.5, 0) circle (1pt) node[anchor=north] {$z^{i}$};
\end{tikzpicture}
  \caption{Two different figures of partial supervision in $\Y=\R^2$ without input (or $\X$ being a singleton).
    The loss is given by the mean square error, {\em i.e.} the usual Euclidean geometry.
    On both figures, we represent $\Y$ with two sets, $S_1$ in light gray and $S_2$ in gray, and the estimates $z = \argmin_{z\in\Y} \brace{L(z, S_1) + L(z, S_2)}$ provided by the infimum loss ($z^i$) and the average loss ($z^a$).
    The point $y_1^a$ represents the disambiguation of $S_1$ by the average loss.
    On the left figure, although $S_1$ intersects $S_2$, only the infimum loss verifies $z^i \in S_1\cap S_2$.
  }
  \label{int:fig:il_ac}
\end{figure}
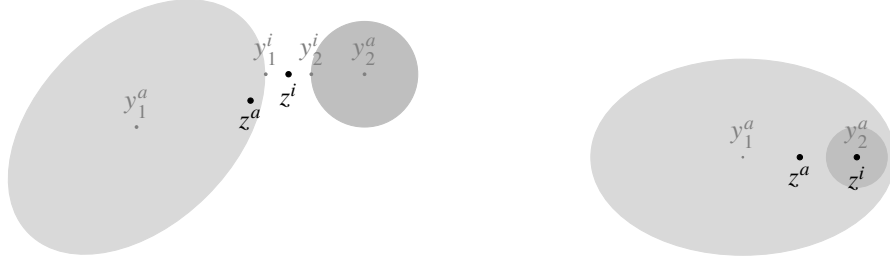

Let us take a step back and recall the commoner providing the set ``feline'' that is $S_1 = \brace{\text{``cat'', ``lion'', \dots}}$ and the set ``tufted ears'', that is $S_2 = \brace{\text{``owl'', ``Araucana chicken''}, \dots}$, when annotating the picture of a ``caracal''.
Naturally, we would like to output a prediction $z\in S_1\cap S_2 = \brace{\text{``caracal''}}$ of a feline with tufted ears.
In other terms, we want to create consensus between observations, which is what the infimum loss provides.
The supremum loss is outcast from the good solutions, since it might disambiguate $y_1 \in S_1$ as ``leopard'' and $y_2 \in S_2$ as ``owl'', as well as the average loss as we illustrated on Figure \ref{int:fig:il_ac}.

\subsection{Theoretical guarantees}

With the infimum loss, we switch to the original supervised learning problem \eqref{int:eq:risk} to the new problem defined through the risk
\begin{equation}
  \label{int:eq:inf_risk}
  {\cal R}_S(f) = \E_{(X, S)\sim\tau}\bracket{\inf_{y\in S} \ell(f(X), y)}.
\end{equation}
How are those two problems related?
In particular, we would like to make sure that solving the partially supervised problem does help to tackle the fully supervised one.
Following the formalism of the previous chapter, we can try to derive a calibration inequality that relates the excess of the original risk \eqref{int:eq:risk} to the excess of the ``surrogate'' risk \eqref{int:eq:inf_risk}.
To derive such a generic inequality, we need strong assumptions, which are usually given by ensuring non-ambiguity of the partial supervision (Definition \ref{il:def:non-ambiguity}), that is for each $x$, the sets in the support of the conditional distribution $(S\,\vert\,X=x)$ intersect into a singleton $\brace{y_x} = \brace{f^*(x)}$.
Such an inequality was first provided by \cite{Cour2011} in the case of classification with the 0-1 loss, and is generalized to generic problems by Proposition \ref{il:thm:calibration}.
This technique allows us to derive rates on a structured prediction algorithm learning from partially supervised data in Theorem \ref{il:thm:learning-rates}.
In practice, given samples $(X_i, S_i)$, the practitioner can use their favorite algorithm to translate this population principle into an empirical objective to be optimized in order to get an estimate $f_n$ of the risk minimizer.

\begin{remark}[A false case against partial supervision]
  Partially supervised learning and the infimum loss are often criticized because of the strength of the non-ambiguity assumption.
  This should rather be seen as a limitation of theoretical analysis, rather than a limitation of the framework.
  Actually, a more careful analysis based on the idea provided in the previous chapter shows that the non-ambiguity assumption is akin to the Massart noise condition, and can be used to derive much faster rates without modifying algorithms as shown by Theorem \ref{df:thm:convergence}.
  This suggests that there are many interesting behaviors to characterize beyond the non-ambiguity assumption.
\end{remark}
\section{Label disambiguation to complete supervision}

In this section, we characterize the solution provided by the infimum loss as a disambiguation strategy, that consists in retrieving full supervision first, before learning from it.
This leads to a different perspective in terms of learning and practical implementations.

\subsection{Expected testing distribution}

Let us roll back a little and think back on which solution to look for when in presence of partial supervision.
The ultimate goal is to find a mapping that minimizes the generalization error on the testing distribution $\rho\in\prob{\X\times\Y}$.
Yet which distribution $\rho$ should we expect given the loss $\ell$ and the distribution $\tau \in \prob{\X\times 2^\Y}$?
The only clearly defined constraints is that the distribution should be compatible with the weak information we have seen on the data.
The notion of compatibility can be formalized through the following definition, which we introduced in \cite{Cabannes2020}.

\begin{definition}[Compatibility]
  \label{int:def:comp}
  Two distributions $\rho\in\prob{\X\times\Y}$ and $\tau\in\prob{\X\times{\cal S}}$ are said to be {\em compatible} if there exists an {\em underlying distribution} of probability $\pi\in\prob{\X\times\Y\times{\cal S}}$ such that $\rho$ and $\tau$ are the respective marginals of $\pi$ according to $\X\times\Y$ and $\X\times{\cal S}$ and such that for any $(x, y, s) \in \supp\pi$, we have $y\in s$.
  This compatibility property will be denoted by $\rho\vdash\tau$.
\end{definition}

To discriminate the expected $\rho$ among all the distributions compatible with $\tau$, we need a principle.
It is natural to look for a distribution that satisfies some properties.
As mentioned in the last section, we would like to go for a distribution that is ``consensual'', {\em i.e.} we would like to reduce the noise of the conditional $\paren{Y\midvert X}$.
In other terms, we would like those conditional distributions to be as deterministic as possible.
As a consequence, we define the expected testing distribution as
\[
  \rho^\star \in \argmin_{\rho\vdash\tau}{\cal E}(\rho),
\]
for ${\cal E}:\prob{\X\times\Y}\to\R$ a measure of determinism.
A well-known measure of determinism is the entropy ${\cal E}(\rho):=\E[-\log(\rho(X, Y))]$.
This measure is independent of the loss $\ell$, which might be a desirable property if one would like to reuse the same supervised distribution for different tasks linked with different losses.
Yet, when there is a structure on $\Y$ such that its explicit dimension is much bigger than the intrinsic dimension of this space seen through the loss $\ell$, the entropy will scale with the explicit dimension.
For example, in ranking problems where the goal is to output an ordering over $m$ items, which is similar to a permutation in $\Sfrak_m$, the output space is of cardinality $m!$, while the intrinsic dimension is $m$.
For those problems, it is wise to incorporate the loss in the objective, in particular, we suggest the following loss-based variance ${\cal E}(\rho) := \inf_{f:\X\to\Y} \E_\rho[\ell(f(X), Y)]$.
Once the distribution $\rho$ is recovered, a function $f$ can be learned in a supervised learning fashion
\begin{equation}
  \label{eq:int:df}
  f^* \in \argmin_{f:\X\to\Y} \E_{(X, Y)\sim\rho^\star}[\ell(f(X), Y)], \qquad\text{with}\qquad \rho^\star = \argmin_{\rho\vdash\tau}  \inf_{f:\X\to\Y} \E_\rho[\ell(f(X), Y)].
\end{equation}
Remarkably, this second solution to the partially supervised learning problem is exactly the same as the one provided by the infimum loss.
This is based on the fact that, under really mild definition assumptions
\[
  \inf_{\rho\vdash\tau} \inf_{f:\X\to\Y} \E_{\rho}[\ell(f(X), Y)] = \inf_{f:\X\to\Y}\inf_{\rho\vdash\tau} \E_{\rho}[\ell(f(X), Y)] = \inf_{f:\X\to\Y} \E_{\tau}[\inf_{y\in S}\ell(f(X), y)].
\]
A nice property of this new characterization of the solution $f^*$ is that it goes beyond the partial labeling problem.
As a matter of fact, \eqref{eq:int:df} can be applied more generally to any weakly supervised learning paradigm where the weak information acquired on the problem can be translated into hard constraints, {\em e.g.} in the case of label proportion.

\subsection{Optimization issues}

Before going any further, let us point out that any objective ${\cal E}(\rho)$ that is minimized for deterministic distributions is intrinsically not-convex.
Consider the case where there is no input space (or $\X$ is a singleton), $\Y$ is discrete, and ${\cal E}$ is a function taking as argument elements of the simplex $\prob{\Y}$ and outputting a real-valued score.
If ${\cal E}$ is minimized for deterministic distributions, it is minimized on extremities of the simplex.
In other terms, ${\cal E}$ should really be thought of as a concave function, making its minimization a non-convex optimization problem.
We picture level lines of such an objective on Figure \ref{df:fig:objective}.

Regarding the constraints, it should be noted that $\brace{\rho\in\prob{\X\times\Y}\midvert \rho\vdash\tau}$ is a convex set.
So we are trying to minimize a concave function over a convex set.
In \cite{Cabannes2021}, we suggested two different heuristics to approach this minimization procedure.
The first one consists in starting with a good ``center point'' and finding an extremity of the definition domain after gradient descent.
Our notion of center takes into account the loss to avoid insidious bias (similarly to what we discussed the average loss).
The second one is based on the Diffrac algorithm \citep{Bach2007}, and consists in convexifying the objective by adding a quadratic term that is constant on extremities, perform gradient descent, and get a solution that is a convex combination of extreme points.
We refer to sections \ref{df:sec:optim} and \ref{df:app:diffrac} for further discussions on those techniques.

\subsection{Empirical objective}

Given data $(X_i, S_i)_{i\leq n}\sim \tau^{\otimes n}$, how to translate the disambiguation principle \eqref{eq:int:df} into an empirical objective allowing to disambiguate the sets $(S_i)_{i\leq n}$ into labels $(\hat{Y}_i)_{i\leq n}$?
To answer this question, we need to translate an objective on distribution into an objective on samples.
The usual translation of risk minimization into empirical risk minimization can be thought as the approximation $\rho \approx \hat{\rho} = \frac{1}{n} \sum_{i\leq n} \delta_{X_i} \otimes \delta_{Y_i}$ plus some constraints $f\in{\cal F}$ on the functions that we minimize.
As such, we could go for
\[
  \min_{(y_i \in S_i)_{i\leq n}} \inf_{f\in{\cal F}} \sum_{i=1}^n \ell(f(X_i), y_i).
\]
This will be really hard to solve in practice.
The solution we suggested is based on kernel mean embedding \citep{Muandet2017}, which is a natural way to understand the surrogate approach of \cite{Ciliberto2016}.
It consists in incorporating directly the hypothesis on the solution of the problem in the estimated distribution with $\hat{\rho} \approx \frac{1}{n} \sum_{i,j\leq n} \delta_{X_i} \otimes \alpha_j(X_i)\delta_{Y_j}$, where $\alpha:\X\to\R^n$ is a weighting scheme that states how to diffuse information $(Y_i)_{i\leq n}$ seen at $(X_i)_{i\leq n}$ to any point $x\in\X$.
In particular, those weights might be used to estimate the conditional distribution $\paren{Y\vert X=x}$ as $\sum_{j\leq n} \alpha_j(x)\delta_{Y_j}$.
This leads to the following empirical objective
\begin{equation}
  \label{eq:int:df_emp}
  (\hat{y}_i)_{i\leq n} \in \argmin_{(y_i \in S_i)_{i\leq n}} \inf_{(z_i\in \Y)_{i\leq n}} \sum_{i,j=1}^n \alpha_i(X_j)\ell(z_i, y_j),
\end{equation}
In the case of kernel ridge regression, those weights are given by
$\alpha(x) = (K + \lambda n)^{-1}K_x$ with the notations of section \ref{int:krr}.
But they could also be given by similarity metrics, or derived with more advanced unsupervised techniques such as the one we present in \cite{Cabannes2021c} as we detail in section \ref{lap:sec:extension}.

\subsection{Theoretical guarantees}

The convergence of an estimate $f_n$ learned on the dataset $(X_i, \hat{y}_i)_i$ with $(\hat{y}_i)_{i\leq n}$ recovered through \eqref{eq:int:df_emp} cannot easily be performed with the tools presented so far.
Indeed, it is hard to understand how the different disambiguation $(y_i \in S_i)$ interacts when studying \eqref{eq:int:df_emp}.
This relates to the non-concavity of this objective: a small change in the optimal $(z_i)_{i\leq n}$ can lead to completely different optimal $(y_i)_{i\leq n}$.
While there are probably good tools to think in terms of distribution and derive a consistent empirical estimate of \eqref{eq:int:df}, in \cite{Cabannes2021}, we used the fact that usual learnability assumptions for the partial labeling problem are actually quite strong, which allows us to prove impressively fast rates.

To derive consistency of algorithm learning with partially supervised data, it is classical to assume non-ambiguity of the distribution $\paren{S\midvert X=x}$ as well as some regularity of the function $\X\to\prob{\Y}; x\to\paren{Y\midvert X=x}$.
Those assumptions actually imply that the data are clustered by groups where labels are all equal.
When this class separation hypothesis holds true, one can leverage this structure by considering a weighting scheme based on nearest neighbors.
As we discussed in more details in section \ref{df:sec:stat}, this endows the estimate $f_n$ with exponential convergence rates, meaning that the risk ${\cal R}(f_n)$ converges toward the minimum risk exponentially fast as the number of samples $n$ goes to infinity.
This is the statement of Theorem \ref{df:thm:convergence}.

\subsection{Can collaborative filtering help us to deal with weak supervision?}

There is a natural link between completion of weak information and the collaborative filtering problem described in section \ref{int:sec:coll_filt}.
To make this link we will need to introduce more concepts.

Assume that the loss $\ell:\Y\to\Y\to\R$ can be written as $\ell(z, y) = -\scap{\psi(z)}{\phi(y)}$ with $\psi,\phi:\Y\to{\cal H}$ two embeddings into a Hilbert space ${\cal H}$, encoding the structure of the loss.
For example, the Kendall loss in ranking corresponds to the correlation measure $\phi = -\psi = (\ind{\sigma(i) > \sigma(j)})_{i,j\leq m}$ for $\sigma \in \Y = \Sfrak_m$ a permutation.
We refer to \cite{Nowak2019} for more examples.
Assume also that the weak supervision correspond to observing $A_i\phi(Y_i)$ instead of $\phi(Y_i)$ for a known masking operator $A_i:{\cal H}\to{\cal H}$.
For example, observing a partial ordering that $\sigma$ should verify is equivalent to observing some coordinates of the vector $\phi(\sigma)$.
The disambiguation problem can be reduced to the matrix completion problem of retrieving $(\phi(y_i))_{i\leq n}$ from the observations $(A_i\phi(Y_i))_{i\leq n}$.
Our algorithm suggests solving the minimization problem
\[
  \minimize{\norm{
      \paren{\begin{array}{ccc}
          \vert     &        & \vert     \\
          \psi(z_1) & \cdots & \psi(z_n) \\
          \vert     &        & \vert     \\
        \end{array}}
      \paren{\begin{array}{ccc}
           &               & \\
           & \alpha_i(X_j) & \\
           &               & \\
        \end{array} }
      \paren{\begin{array}{ccc}
          \vert     &        & \vert     \\
          \phi(y_1) & \cdots & \phi(y_n) \\
          \vert     &        & \vert     \\
        \end{array}}^\top
    }_F^2}{A_i\phi(y_i) = A_i\phi(Y_i).}
\]

In the case where $\psi = -\phi$ and there is no context variables, hence $\alpha_i(X_j) = 1$, the slightly different measure of determinism exposed in section \ref{df:app:diffrac} leads to the problem
\[
  \minimize{-\norm{ \sum_{i=1}^n \phi(y_i)}^2}
  {A_i\phi(y_i) = A_i\phi(Y_i).}
\]
For correlation losses, it is usual for $\norm{\phi(\cdot)}$ to be constant, so to minimize the last objective, one should try to align all $\phi(y_i)$ in one direction.
This differs from the collaborative filtering solution which completes the $(\phi(y_i))$ by minimizing the cardinality of the set $\brace{\phi(y_i)\midvert i\in[N]}$.
Formally, collaborative aims at solving the following problem
\[
  \minimize{\operatorname{rank}
    \paren{\begin{array}{ccc}
        \vert     &        & \vert     \\
        \phi(y_1) & \cdots & \phi(y_n) \\
        \vert     &        & \vert     \\
      \end{array}}
  }{A_i\phi(y_i) = A_i\phi(Y_i),}
\]
which is practically implemented with the nuclear norm instead of the rank.
In essence, the infimum loss builds an understanding of $\paren{Y\midvert X}$ by collapsing all observations onto a single vector in $\phi(\prob{\Y})$, while collaborative filtering builds such an understanding by minimizing the number of vectors in $\phi(\Y)$ to complete observations without discrepancy.
Bining observations into several groups makes sense when context variables do not allow for inputs to characterize outputs univocally.

Further investigations of the link between our work and collaborative filtering is a research direction that we left open.
We conjecture that this link might prove useful to incorporate context variables into collaborative filtering techniques.

\section{Leveraging input structure}
In this section, we discuss unsupervised learning techniques that could be incorporated in the frameworks described previously in order to leverage unlabeled input data.

\subsection{The case of semi-supervised learning}
The infimum loss and the disambiguation framework described previously are based on a pointwise principle that discriminates a distribution $\paren{Y\midvert X}$ from a weakly supervised distribution $\paren{S\midvert X}$.
Such a principle fails to provide any guideline for semi-supervised learning, which is a specific instance of partial supervision where $S$ is either a singleton or the full set $\Y$.
In particular for unsupervised parts of the input space, which correspond to points in the support of $\paren{X\midvert S=\Y}$ that do not belong to the support of $\paren{X\midvert \card S = 1}$, this principle does not provide any information on what $f^*(x)$ should be.

Eventually, the solution on those points could be inferred from the solution on the other points by looking for the simplest function completion with respect to some criterion quantifying how ``simple'' a function is.
For regression problems, if one searches to minimize the empirical risk in a Banach space of function, the norm of this Banach space provides a simple criterion for ``simplicity''.
More generally, in metric space of function, this norm can be replaced more generally as the distance between a neutral function ({\em e.g.} the null function in a vector space) and the function itself.

\subsection{Laplacian regularization}
The classical approach to solve semi-supervised learning regression problem is to choose the criterion of ``simplicity'' as the Dirichlet energy $f\to \E_{\rho_\X}[\norm{\nabla f(X)}^2]$.
This regularization is similar to the square norm of the Hilbert space of functions $W^{1,2}(\rho_\X)$, which is the weighted Sobolev space of functions with weak derivative endowed with the probability measure $\rho_\X$.
Considering a square of the norm rather than the norm itself is classical in machine learning, making analysis as well as computations easier.
Measuring regularity through the probability measure $\rho_\X$ rather than the Lebesgue measure is crucial in order to leverage the unsupervised data.

This regularization has been introduced in the seminal paper of \cite{Zhu2003} which estimates the Dirichlet energy through finite difference methods akin to the Nadayara-Watson estimator.
This regularization formalized many intuitions.
For regression problems, it is natural to assume that the target function does not vary too much on densely populated input regions, or said otherwise, that its variations concentrate on regions without too much data.
For classification problems, this echoes the low-density separation hypothesis, assuming that decision frontiers, that is frontier between classes in the input space, lies on sparsely populated regions.
In \cite{Cabannes2021c}, we proposed a ``kernelized'' version of this estimation in order to bypass the curse of dimensionality under regularity assumptions, together with some clever low-rank factorization in order to reduce computation time to a decent amount.
It should be noted that Laplacian regularization has also a natural interpretation in terms of diffusion, which is captured by the Langevin dynamic, and explains the wording ``label propagation'' found in semi-supervised literature.

\subsection{Complete framework}
With Laplacian regularization, one can deepen the {\em lex parsimoniae} of the precedent sections when this rule does not discriminate a unique distribution by looking for the distribution that will minimize the Dirichlet energy of its (or a continuous surrogate) risk minimizer.
In particular, this approach is interesting as it could remove the usual non-ambiguity assumption usually made to define and study solutions of the partial supervision problems \citep{Cour2011,Luo2010,Liu2014,Cabannes2020,Cabannes2021}.
Yet, removing such an assumption will come at the price of adding assumptions on the surrogate problem we are solving that might be harder to verify in practice.

Finally, it should be noted that Laplacian regularization can be introduced as soon as during the preprocessing/feature engineering part of a practical machine learning pipeline.
In particular, it provides principled guidelines to retrieve and parametrize a small manifold on which the input data might lie, and might prove itself useful to strengthen self-supervision techniques.
At this point, it is less a theory than experiments that could confirm those intuitions.

\section{Active labeling}

Weakly supervised learning is concerned with retrieving a target function under weak observations.
It is often motivated by the bottleneck of annotating data.
Yet it fails to answer the most simple question practitioners might ask themselves: how to collect the most discriminative dataset to learn this function under some cost constraints.
Hence, the last part of this thesis is devoted to the ``active'' collection of weak supervision.
Active refers to the fact that we will iteratively and adaptively search for annotations, in contrast with passive settings in which annotations are given once and for all by an exterior mechanism.

The cost of dataset annotation depends on different task specifications.
Hence, a model of annotation cost can only describe a limited number of situations.
As well as we split our understanding of weakly supervised learning in three parts (group statistics, weak classifiers, and labels corruption), we could split dataset annotation with the same three categories.
In practice, it is frequent to annotate dataset by grouping input data according to a classifier output before identifying the most present class in each group and filtering mistakes, which allows annotating large chunks of data at once.
In a similar fashion, ImageNet was collected by searching for different categories on image search engines, before spotting outliers in a batch of images resulting from a single query \citep{ImageNet}.

The final part of this thesis touches upon the active variant of partially supervised learning \citep{Cabannes2022}.
In particular, we assume that given data $(X_i)_{i\leq n}$, one can query any information of the type $\ind{Y_{i_t}\in S_t}$ for $i_t \in [n]$ an index to choose and $S_t\in {\cal S}\subset 2^\Y$ a set to choose among certain subsets of labels.
We refer the reader to Part \ref{part:collection} for additional considerations on the matter.

\part{Considerations on Learning Theory}
\label{part:discrete}
\chapter{Fast Rates for Structured Prediction}
\label{chap:fast}

The following is a reproduction of \cite{Cabannes2021b}.

Discrete supervised learning problems such as classification are often tackled by introducing a continuous surrogate problem akin to regression.
Bounding the original error, between estimate and solution, by the surrogate error endows discrete problems with convergence rates already shown for continuous instances.
Yet, current approaches do not leverage the fact that discrete problems are essentially predicting a discrete output when continuous problems are predicting a continuous value.
In this paper, we tackle this issue for general structured prediction problems, opening the way to ``superfast'' rates, that is, convergence rates for the excess risk faster than $n^{-1}$, where $n$ is the number of observations, with even exponential rates with the strongest assumptions.
We first illustrate it for predictors based on nearest neighbors, generalizing rates known for binary classification to any discrete problem within the framework of structured prediction.
We then consider kernel ridge regression where we improve known rates in $n^{-1/4}$ to arbitrarily fast rates, depending on a parameter characterizing the hardness of the problem, thus allowing, under smoothness assumptions, to bypass the curse of dimensionality.
\section{Introduction}

Machine learning is raising high hopes to tackle a wide variety of prediction problems, such as language translation, fraud detection, traffic routing, speech recognition, self-driving cars, DNA-binding proteins, {\em etc}.
Its framework is appreciated as it removes humans from the burden to come up with a set of precise rules to accomplish a complex task, such as recognizing a cat on an array of pixels.
Yet, it comes at a price, which is of forgetting about algorithm correctness, meaning that machine learning algorithms can make mistakes, {\em i.e.}, wrong predictions, which can have dramatic implications, {\em e.g.}, in medical applications.
This motivates work on generalization error bounds, quantifying how often one should expect errors.

Many of the problems discussed above are of discrete nature, in the sense that the number of potential outputs is finite, or infinite countable.
To learn such problems, a classical technique consists in defining a continuous surrogate problem, which is easier to solve, and such that:
\begin{itemize}
  \item[(1)] an algorithm on the surrogate problem translates into an algorithm on the original problem;
  \item[(2)] errors on the original problem are bounded by errors on the surrogate problem.
\end{itemize}
The first point refers to the concept of plug-in algorithms, while the second point to the notion of calibration inequalities.
For example, binary classification can be approached through regression by estimating the conditional expectation of the output $Y$ given an input $X$ \citep{Bartlett2006}.

On the one hand, continuous surrogates for discrete problems are interesting, as they benefit from functional analysis knowledge, when discrete problems are more combinatorial in nature.
On the other hand, continuous surrogates can be deceptive, as they are asking to solve for more than needed.
Considering the example of binary classification, where $Y\in\brace{-1, 1}$, one only has to predict the sign of the conditional expectation, rather than its precise value.
Interestingly, without modifying the continuous surrogate approach, this last remark can be leveraged in order to tighten generalization bounds derived through calibration inequalities \citep{Audibert2007}.
In this work, we extend those considerations, known in binary classification \citep[\emph{e.g.},][]{Koltchinskii2005, Chaudhuri2014}, to generic discrete supervised learning problems, and show how it can be applied to the kernel ridge regression algorithm introduced by \citet{Ciliberto2016}.

\subsection{Contributions}
Our contributions are organized in the following order.
\begin{itemize}
  \item In Section \ref{rates:sec:structured}, we consider the general structured prediction from \citet{Ciliberto2020} based on Lipschitz-continuous losses and derive refined calibration inequalities to leverage the fact that learning a mapping into a discrete output space is easier than learning a mapping into a continuous space.
  \item In Section \ref{rates:sec:margin}, we show how to exploit exponential concentration inequalities to turn them into fast rates under a condition generalizing the Tsybakov margin condition.
  \item In Section \ref{rates:sec:nn}, we apply Section \ref{rates:sec:margin} to local averaging methods with the particular example of nearest neighbors.
        This leads to extending the rates known for regression and classification to a wide variety of structured prediction problems, with rates that match minimax rates known in binary classification.
  \item In Section \ref{rates:sec:rkhs}, we show how Section \ref{rates:sec:margin} can be applied to kernel ridge regression.
        This allows us to improve rates known in $n^{-1/4}$ to arbitrarily fast rates depending on the hardness of the associated discrete problem.
\end{itemize}

\subsection{Related work}

\paragraph{Surrogate framework.}
The surrogate problem we will consider to tackle structured prediction finds its roots in the approximate Bayes rule proposed by \citet{Stone1977}, analyzed through the prism of mean estimation as suggested by \citet{Friedman1994} for classification, and analyzed by \citet{Ciliberto2020} in the wide context of structured prediction.
In particular, we will specify results on two classes of surrogate estimators: local averaging methods, or kernel ridge regression.

\paragraph{Local averaging methods.}
Neighborhood methods were first studied by \citet{Fix1951} for statistical testing through density estimation.
Similarly, Parzen–Rosenblatt window methods \citep{Parzen1962,Rosenblatt1956} were developed.
Those methods were cast in the context of regression as nearest neighbors \citep{Cover1957} and Nadayara-Watson estimators \citep{Watson1964,Nadaraya1964}.
\citet{Stone1977} was the first to derive consistency results for a large class of localized methods, among which are nearest neighbors and some window estimators \citep{Spiegelman1980,Devroye1980}.
Rates were then derived, with minimax optimality \citep{Stone1980,Yang1999}.
Several reviews can be found in the literature, such as \citet{Gyorfi2002,Tsybakov2009,Biau2015,Chen2018}.

\paragraph{Reproducing kernel ridge regression.}
The theory of real-valued reproducing kernel Hilbert spaces was formalized by \citet{Aronszajn1950}, before finding applications in machine learning \citep[\emph{e.g.},][]{Scholkopf2001}.
Minimax rates for kernel ridge regression were achieved by casting the empirical solution estimate as a result of integral operator approximation \citep{Smale2007,Caponnetto2007}, allowing to control convergence through concentration inequalities in Hilbert spaces \citep{Yurinskii1970,Pinelis1986} and on self-adjoint operator on Hilbert spaces \citep{Minsker2017}.
First derived in $L^{2}$-norm, rates were cast in $L^{\infty}$-norm through interpolation inequalities \citep[\emph{e.g.},][]{Fischer2020,Lin2020}.

\paragraph{Tsybakov margin condition.}
Learning a mapping into a discrete output space is indeed easier than learning a continuous mapping, as, for binary classification for example, one typically only has to predict the sign of $\E[Y\vert X]$ rather than its precise value.
As such, calibration inequalities that relate the error on a discrete structured prediction problem to an error on a smooth surrogate problem are often suboptimal.
This phenomenon was exploited for density discrimination, a problem consisting of testing if samples were drawn from one or the other of two potential distributions, by \citet{Mammen1999}, and for binary classification by \citet{Audibert2007}.
Those works introduce a parameter $\alpha\in[0, \infty)$ characterizing the hardness of the discrete problem, and leverage concentration inequalities to accelerate rates known for regression by a power $\alpha+1$ \citep{Audibert2007}, while rates plugged-in directly through calibration inequalities only present an acceleration by a power $\sfrac{2(\alpha+1)}{(\alpha + 2)}$ \citep[see, \emph{e.g.}][]{Boucheron2005,Bartlett2006,Bartlett2006b,Ervan2015,Nowak2019}.

\section{Structured prediction with surrogate control}
\label{rates:sec:structured}

In this section, we introduce the classical supervised learning problem, and a surrogate problem that consists of conditional mean estimation.
We recall a calibration inequality relating the original problem to the surrogate one.
We mention how empirical estimations of the conditional means usually deviate from the real means following a sub-exponential tail bound, similarly to bounds obtained through Bernstein inequality.
We end this section by providing refined surrogate control, that is the key toward ``superfast'' rates, that is, rates faster than $1/n$.

\subsection{Surrogate mean estimation}
Consider a classic supervised learning problem, where given an input space $\X$, an observation space~$\Y$, a prediction space $\cal Z$, a joint distribution $\rho \in \prob{\X\times\Y}$ and a loss function $\ell:{\cal Z}\times\Y\to\R_{+}$, one would like to retrieve $f^{*}:\X\to{\cal Z}$ minimizing the risk $\cal R$.
\[
  f^{*} \in \argmin_{f:\X\to{\cal Z}} {\cal R}(f)
  \qquad\text{with}\qquad
  {\cal R}(f) = \E_{(X, Y)\sim\rho}\bracket{\ell(f(X), Y)}.
\]
In practice, $\X$, $\Y$, ${\cal Z}$ and $\ell$ are givens of the problem, while $\rho$ is unknown, yet partially observed thanks to a dataset ${\cal D}_{n} = (X_{i}, Y_{i})_{i\leq n} \sim \rho^{\otimes n}$, with data $(X_{i}, Y_{i})$ sampled independently from $\rho$.
Note that in fully supervised learning, the observation space is the same as the prediction space $\Y = {\cal Z}$, yet we distinguish the two for our results to stand in more generic settings, such as instances of weak supervision \citep{Cabannes2020}.
In the following, we consider ${\cal Z}$ finite.
In several cases, solving the supervised learning problem can be done through solving a surrogate problem that is easier to handle.
\cite{Ciliberto2016} provides a setup that reduces a wide variety of structured prediction problems $(\ell, \rho)$ to a problem of mean estimation.
It works under the following assumption.
\begin{assumption}[Bilinear loss decomposition]\label{rates:ass:loss}
  There exists a Hilbert space ${\cal H}$ and two mappings
  $\psi:{\cal Z}\to{\cal H}$, $\phi:\Y\to{\cal H}$ such that
  \[
    \ell(z, y) = \scap{\psi(z)}{\phi(y)}.
  \]
  We will also assume that $\psi$ is bounded (in norm) by a constant $c_\psi$.
\end{assumption}
This assumption is not really restrictive \citep{Ciliberto2020}.
Among others, it works for any losses on finite spaces, usually with spaces $\cal H$ whose dimensionality is only polylogarithmic with respect to the cardinality of $\cal Z$ \citep{Nowak2019}. 
Under Assumption \ref{rates:ass:loss}, solving the supervised learning problem can be done through estimating the surrogate conditional mean $g^{*}:\supp\rho_{\X}\to{\cal H}$, defined as
\begin{equation}
  \label{rates:eq:gast}
  g^{*}(x) = \E_{Y\sim\rho\vert_{x}}\bracket{\phi(Y)},
\end{equation}
where we denote $\rho\vert_{x}$ the conditional law of $\paren{Y\midvert X}$ under $(X, Y) \sim \rho$.

\begin{lemma}[\citet{Ciliberto2016}]\label{rates:lem:cal}
  Given an estimate $g_{n}$ of $g^{*}$ in \eqref{rates:eq:gast}, consider the estimate $f_{n}:\X\to{\cal Z}$ of $f^{*}$, which is obtained from ``decoding'' $g_{n}$ as
  \begin{equation}\label{rates:eq:decoding}
    f_{n}(x) = \argmin_{z\in{\cal Z}} \scap{\psi(z)}{g_{n}(x)}.
  \end{equation}
  Then the excess risk is controlled through the surrogate error as
  \begin{equation}
    \label{rates:eq:L1}
    {\cal R}(f_{n}) - {\cal R}(f^{*}) \leq 2c_\psi
    \norm{g_{n} - g^{*}}_{L^{1}(\X, {\cal H}, \rho)}.
  \end{equation}
\end{lemma}

Inequalities relating the original excess risk ${\cal R}(f_{n}) - {\cal R}(f^{*})$ with a measure of error on a surrogate problem are called \emph{calibration inequalities}.
They are useful when the measure of error between $g_{n}$ and $g^{*}$ is easier to control than the one between $f_{n}$ and $f^{*}$.

\begin{example}[Binary classification]\label{rates:ex:binary}
  Binary classification corresponds to $\Y={\cal Z} = \brace{-1, 1}$ and $\ell(z, y) = \ind{z\neq y}$ (or equivalently $\ell(z, y) = 2\ind{z\neq y}-1$).
  The classical surrogate consists of taking $\cal H = \R$, with $\phi=\textit{id}$ and $\psi=-\textit{id}$.
  In this setting, we have $g^{*}(x) = \E_{\rho}[Y\vert X=x]$, and the decoding $f_{n}(x) := \sign g_{n}(x)$, for any $g_{n}(x) \in\cal H$.
  In this case
  \(
  {\cal R}(f_{n}) - {\cal R}(f^{*})
  = \E_{X}\bracket{\ind{f_{n}(X)\neq f^{*}(X)} \abs{g^{*}(X)}}
  \leq 2\norm{g_{n} - g^{*}}_{L^{1}}
  \leq 2\norm{g_{n} - g^{*}}_{L^{2}}.
  \)
  Note that in regression the excess risk reads as the square of the $L^{2}$ norm, explaining a loss of a power one half in convergence rates, when going from regression to classification \citep[\emph{e.g.}][]{Chen2018}.
\end{example}

Differences between an empirical estimate and its population version are generally handled through concentration inequalities.
In this work, we will leverage concentration on $\norm{g_{n}(x) - g(x)}$ that is uniform for $x \in \supp \rho_{\X}$, motivating the introduction of Assumption \ref{rates:ass:concentration}.
\begin{assumption}[Exponential concentration inequality]\label{rates:ass:concentration}
  Suppose that for $n\in\N$, there exists two reals $L_n$ and $M_n$, such that the tails of $\norm{g_{n}(x) - g(x)}$ can be controlled for any $t > 0$ as
  \begin{equation}
    \label{rates:eq:concentration}
    \sup_{x\in\supp\rho_{\X}} \Pbb_{{\cal D}_{n}}\paren{\norm{g_{n}(x) - g(x)} > t} \leq
    \exp\paren{-\frac{L_n t^{2}}{1 + M_n t}}.
  \end{equation}
\end{assumption}
Note that to satisfy Assumption \ref{rates:ass:concentration}, it is sufficient, yet \emph{not necessary}, to have a uniform control on $g_{n} - g^{*}$, {\emph i.e.}, a control on the tail of $\norm{g_{n} - g^{*}}_{L^{\infty}}$, since
\(
\sup_{x} \Pbb\paren{{A_{x} > t}}
\leq \Pbb\paren{\cup_{x} \brace{A_{x} > t}} = \Pbb\paren{\sup_{x} A_{x} > t},
\)
with $(A_{x})$ a family of random variables indexed by $x\in\X$.

Usually, in bounds like \eqref{rates:eq:concentration}, $M_n$ is a constant of the problem, while $L_n$ depends on the number of samples, therefore, we would like to give rates depending on $L_n$.
Typically, in Bernstein inequalities (see Theorem \ref{rates:thm:bernstein-vector} in Appendix), $L_n = n\sigma^{-2}$ with $\sigma^{2}$ a variance parameter and $M_n = c\sigma^{-2}$ with $c$ a constant of the problem that does not depend on $n$.

\subsection{Refined calibration}
While it is sufficient to control the excess risk through an $L^{1}$-norm control on $g$ from~\eqref{rates:eq:L1}, it is not always necessary.
In other words, the calibration bound in Lemma \ref{rates:lem:cal} is not always tight.
Indeed, we do not predict optimally, that is, $\brace{f_{n}(x) \neq f^{*}(x)}$ only if $g_{n}(x)$ and $g^{*}(x)$ do not lead to the same decoding $f_{n}(x)$ and $f^{*}(x)$.
When $\cal Z$ is finite, this is characterized by $g_{n}(x)$ and $g^{*}(x)$ not falling in the same region $R_z$ of ${\cal H}$, where
\[
  R_z = \big\{\xi \in {\cal H} \big| z \in \argmin_{z' \in{\cal Z}}
  \scap{\psi(z')}{\xi}\big\}.
\]
To ensure that $g_{n}(x)$ and $g^{*}(x)$ fall in the same region, one can ensure that $g_{n}(x)$ is closer to $g^{*}(x)$ than $g^{*}(x)$ is on the frontier of those regions.
Those frontiers are defined by points leading to at least two minimizers in~\eqref{rates:eq:decoding}:
\begin{align*}
  F = \brace{\xi\in{\cal H} \midvert
    \big|\argmin_{z\in{\cal Z}}\scap{\psi(z)}{\xi}\big| > 1}.
\end{align*}
The introduction of $F$ is motivated by the following geometric results.
\begin{lemma}[Refined surrogate control]\label{rates:lem:calibration}
  When $\cal Z$ is finite, for any $x\in\supp\rho_{\X}$,
  \[
    \norm{g_{n}(x) - g^{*}(x)} < d(g^{*}(x), F) \qquad\Rightarrow\qquad f_{n}(x) = f^{*}(x),
  \]
  with $d$ the extension of the norm distance to sets as $d(g^{*}(x), F) = \inf_{\xi\in F} \norm{g^{*}(x) - \xi}$.
  This result allows refining the calibration control from Lemma \ref{rates:lem:cal} as
  \begin{equation}
    \label{rates:eq:calibration}
    {\cal R}(f_{n}) - {\cal R}(f^{*})
    \leq 2c_{\psi} \E_{X} \bracket{ \ind{\norm{g_{n}(X) - g^{*}(X)} \geq d(g^{*}(X), F)}\norm{g_{n}(X) - g^{*}(X)}}.
  \end{equation}
\end{lemma}

\begin{example}[Binary classification]\label{rates:ex:binary-2}
  In binary classification (\emph{cf.} Example \ref{rates:ex:binary}), $F = \brace{0}$, and, for any $x\in\supp\rho_{\X}$, $d(g^{*}(x), F) = \abs{g^{*}(x)}$.
  Lemma \ref{rates:lem:calibration} is based on the fact that $f^{*}(x) \neq f_{n}(x)$ implies that $\sign g^{*}(x) \neq \sign g_{n}(x)$ which itself implies that $\abs{g^{*}(x) - g_{n}(x)} = \abs{g^{*}(x)} + \abs{g_{n}(x)} \geq \abs{g^{*}(x)}$.
\end{example}

To leverage~\eqref{rates:eq:calibration}, we need to control $d(g^{*}(x), F)$ below and $\norm{g_{n}(x)- g^{*}(x)}$ above.
While upper bounds on $\norm{g_{n}(x)- g^{*}(x)}$ are assumed to have been derived through concentration inequalities, lower bounds on $d(g^{*}(x), F)$ will be assumed as a given parameter of the problem, see~\eqref{rates:eq:no-density}~and~\eqref{rates:eq:low-density}.

\begin{remark}[Scope of our work]
  While we derived the refined calibration inequality~\eqref{rates:eq:calibration} for the surrogate conditional mean $g^*$ and the associated pointwise metric $\norm{\cdot}_{\cal H}$, similar inequality could be obtained for other types of surrogate methods.
  This suggests that our work could be extended to any smooth surrogate such as the ones considered by \citet{Nowak2019b}, as well as Fenchel-Young losses \citep{Blondel2020}.
\end{remark}

\subsection{Geometric understanding}

\begin{figure}[ht]
  \centering
  \begin{tikzpicture}[scale=2.5]
  \coordinate(a) at (0, 0);
  \coordinate(b) at ({1/2}, {sin(60)});
  \coordinate(c) at (1, 0);
  \coordinate(ha) at ({3/4}, {sin(60)/2});
  \coordinate(hb) at ({1/2}, 0);
  \coordinate(hc) at ({1/4}, {sin(60)/2});

  \fill[fill=red!20] (a) -- (hb) -- (ha) -- (hc) -- cycle;
  \fill[fill=green!20] (b) -- (ha) -- (hc) -- cycle;
  \fill[fill=blue!20] (c) -- (ha) -- (hb) -- cycle;

  \draw (a) node[anchor=north east]{a} -- (b) node[anchor=south]{b} --
  (c) node[anchor=north west]{c} -- cycle;
  \draw (hb) -- (ha) -- (hc);

  \draw[dashed] (hc) -- (0, {sin(60)/2}) node[left] {$F$};
  \node at ({3/8}, {sin(60)/4}) {$R_a$};
  \node at ({1/2}, {3*sin(60)/4 - 1/16}) {$R_b$};
  \node at ({3/4}, {sin(60)/4 - 1/16}) {$R_c$};
\end{tikzpicture}
\hspace{1cm}
\begin{tikzpicture}[scale=2.5]
  \coordinate(a) at (0, 0);
  \coordinate(b) at ({1/2}, {sin(60)});
  \coordinate(c) at (1, 0);
  \coordinate(ha) at ({3/4}, {sin(60)/2});
  \coordinate(hb) at ({1/2}, 0);
  \coordinate(hc) at ({1/4}, {sin(60)/2});

  \fill[fill=red!20] (a) -- (hb) -- (ha) -- (hc) -- cycle;
  \fill[fill=green!20] (b) -- (ha) -- (hc) -- cycle;
  \fill[fill=blue!20] (c) -- (ha) -- (hb) -- cycle;

  \draw (a) node[anchor=north east]{a} -- (b) node[anchor=south]{b} --
  (c) node[anchor=north west]{c} -- cycle;
  \draw (hb) -- (ha) -- (hc);
  \fill[fill=black!60] (.4, {sin(60)/3}) circle (.05em);
  \fill[fill=black!60] (.5, {sin(60)/3 -.05}) circle (.05em);
  \draw (.4, {sin(60)/3}) circle ({sin(60)/6});
  \draw[dashed] (.4, {sin(60)/3}) -- (0, {sin(60)/3}) node[left] {$\mu^\star$};
  \draw[dashed] (.5, {sin(60)/3-.05}) -- (1, {sin(60)/3-.05}) node[right] {$\mu_n$};
\end{tikzpicture}
  \caption{Illustration of Lemma \ref{rates:lem:calibration}.
  Simplex $\prob{\Y}$, for $\Y = {\cal Z} = \brace{a, b, c}$ and $\ell$ a symmetric loss defined as $\ell(a, b) = \ell(a, c) = 1$ and $\ell(b, c) = 2$, while $\ell(z, z) = 0$.
  This leads to the decision regions $R_z$ represented in colors.
  Given $x\in\X$, if $g^*(x)$ corresponds to a distribution $\mu^* := \rho\vert_{x}$ falling in $R_a$, and if $g_n(x)$ represented by $\hat\mu$ falls closer to $\mu^*$ than the distance between $\mu^*$ and the decision frontier $F$ (represented by a circle on the right figure), then $\hat\mu$ is also in $R_a$, and therefore $f^*(x)=f_n(x)=a$.
  }
  \label{rates:fig:separation}
  \vspace*{-.5cm}
\end{figure}
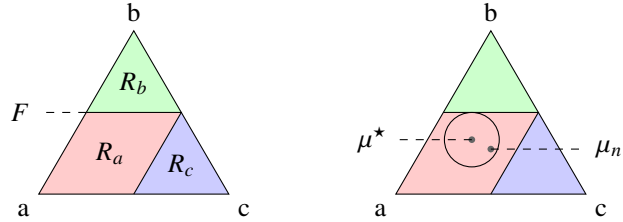

In this subsection, we detail how to understand geometrically Lemma \ref{rates:lem:calibration}.
While the introduction of $\phi$ and $\psi$ could seem arbitrary, it can be thought in a more intrinsic manner by considering the embedding $\phi(y) = \delta_y$ belonging to the Banach space ${\cal H}$ of signed measured, $g^{*}(x) = \rho\vert_{x}$, with the bracket operator, for $\mu\in{\cal H}$ and $z\in{\cal Z}$, $\scap{z}{\mu} = \int_{\Y} \ell(z, y)\mu(\diff y)$, and the distance between signed measures being $d(\mu_{1}, \mu_{2}) = \sup_{z\in{\cal Z}} \scap{z}{\mu_{1} - \mu_{2}}$.
Note that Lemma \ref{rates:lem:calibration} is a pointwise result, holding for any $x\in\X$, that is integrated over $\X$ afterwards.
Therefore, it is enough to consider $\X = \brace{x}$ and remove the dependency in $\X$ to understand it.
The simplex $\prob{\Y}$ naturally splits into the decision region $R_z$ for $z\in{\cal Z}$ as illustrated on Figure~\ref{rates:fig:separation}.
The main idea of Lemma \ref{rates:lem:calibration} is that one does not have to precisely estimate $g^*(x) = \rho\vert_x$ but only has to make sure that $g_n(x)$ falls in the same region on Figure~\ref{rates:fig:separation}.

\section{Rate acceleration under margin condition}
\label{rates:sec:margin}

In this section, we introduce a condition that $g^{*}$ is not too often close to the decision frontier~$F$.
It generalizes the so-called ``Tsybakov margin condition'' known for classification.
Under this condition, we prove rates that generalize the results of \cite{Audibert2007} from binary classification to generic structured prediction problems, which opens the way to ``superfast'' rates in structured prediction.

\subsection{No density separation}
To get fast convergence rates, one has to make assumptions on the problem.
A classical assumption is that $g^{*}$ is smooth enough in order to get concentration bounds similar to Assumption \ref{rates:ass:concentration} when considering a specific class of estimates $g_{n}$.
In our decoding setting (Lemma \ref{rates:lem:cal}), learning is made easy when it is easy to estimate in which region $R_z$ the optimal $g^{*}$ will fall in.
This is in particular the case, when there is a margin $t_{0}>0$, for which, for no point $x \in \supp\rho_{\X}$, $g^{*}(x)$ falls at distance $t_{0}$ of the decision frontier $F$, motivating the following definition.

\begin{assumption}[No-density separation]\label{rates:ass:no-density}
  A surrogate solution $g^{*}$ will be said to satisfy the \emph{no-density separation}, if there exists a $t_{0} > 0$, such that
  \begin{equation}
    \label{rates:eq:no-density}
    \Pbb_{X}(d(g^{*}(X), F) < t_{0}) = 0.
  \end{equation}
  This condition is alternatively called the \emph{hard margin condition}, or sometimes ``Massart's noise condition'' for binary classification~\citep{Massart2006}.
\end{assumption}

\begin{figure}[ht]
  \vspace*{-.5cm}
  \centering
  \begin{tikzpicture}[xscale=2]
  \def\a{.02}
  \draw[->] (-1.2, 0) -- (1.2, 0) node[right] {$x$};
  \draw[->] (-1, -1.2) -- (-1, 1.2) node[above] {$\E[Y\vert X=x]$};
  \draw[-] (-.95, -1) -- (-1.05, -1) node[left] {$-1$};
  \draw[-] (-.95, 1) -- (-1.05, 1) node[left] {$1$};
  \draw[-] (0, 0)--(.125, 1-\a)--(.25, \a)--(.375, 1-\a)--(.5, \a)--(.625, 1-\a)--(.75, \a)--(.875, 1-\a)--(1, .5) node[right] {$g^*(x)$};
  \draw[-] (-1, -\a)--(-.875, -1+\a)--(-.75, -\a)--(-.625, -1+\a)--(-.5, -\a)--(-.375, -1+\a)--(-.25, -\a)--(-.125, -1+\a)--(0, 0);
  \draw[-,red] (-1, -1) -- (0, -1) -- (0, 1) -- (1, 1) node[right] {$f^*(x)$};
\end{tikzpicture}
\hspace{1cm}
\begin{tikzpicture}[xscale=2]
  \draw[->] (-1.2, 0) -- (1.2, 0) node[right] {$x$};
  \draw[->] (-1, -1.2) -- (-1, 1.2) node[above] {$\E[Y\vert X=x]$};
  \draw[-] (-.95, -1) -- (-1.05, -1) node[left] {$-1$};
  \draw[-] (-.95, 1) -- (-1.05, 1) node[left] {$1$};
  \draw[-,red] (-1,1)--(-.75, 1)--(-.75, -1)--(-.5, -1)--(-.5, 1)--(-.25, 1)--(-.25, -1)--(0, -1)--
  (0,1)--(.25, 1)--(.25, -1)--(.5, -1)--(.5, 1)--(.75, 1)--(.75, -1)--(1, -1) node[right] {$f^*(x)$};
  \draw[-] (-1,.8)--(-.75,1);
  \draw[-] (-.75,-1)--(-.5,-.75);
  \draw[-] (-.5,.3)--(-.25,.4);
  \draw[-] (-.25,-.6)--(0,-.5);
  \draw[-] (0,.6)--(.25,.5);
  \draw[-] (.25,-.6)--(.5,-.6);
  \draw[-] (.5,.8)--(.75,.9);
  \draw[-] (.75,-.5)--(1,-.5) node[right] {$g^*(x)$};
  \draw[dashed] (-1,.3)--(1,.3);
  \draw[dashed] (-1,-.3)--(1,-.3);
  \draw[-] (-.95,.3) -- (-1.05,.3) node[left] {$t_0$};
\end{tikzpicture}
  \caption{Illustration of Remark \ref{rates:rmk:separation}.
    We represent two instances of binary classification (see Examples \ref{rates:ex:binary} and \ref{rates:ex:binary-2}).
    On the left example, when $\rho_{\X}$ is such that there is no mass where the sign of $g^{*}$ changes, classes are separated in $\X$, yet the no-density separation is not verified.
    On the right, classes are not separated in $\X$, but the problem satisfies the no-density separation as there is no $x$ such that $d(g^{*}(x), F) = \abs{g^{*}(x)} < t_{0}$.
    Note that when $\rho_{\X}$ is uniform, the left problem satisfies a milder separation condition, introduced thereafter and called the $1$-low-density separation.}
  \label{rates:fig:no-density}
\end{figure}
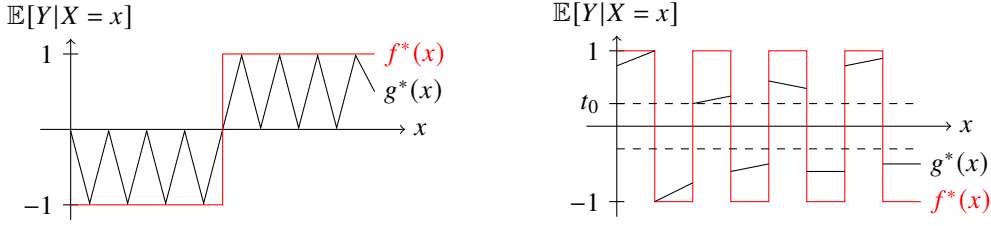

\begin{remark}[Separation in $\Y$ and separation in $\X$]\label{rates:rmk:separation}
  It is important to realize that~\eqref{rates:eq:no-density} is a condition of separation in $\prob{\Y}$ that should hold for all $x\in\X$, but it does not state any separation between classes in $\X$ for $f^{*}:\X\to{\cal Z}$.
  To visualize it, consider the classification problem where $\X = [-1,1]$, $\Y = {\cal Z} = \brace{-1, 1}$ and $\ell(z, y) = \ind{z\neq y}$.
  \begin{itemize}
    \item A situation where $\rho_{\X}$ is uniform on $\X$ and $\E[Y\vert X=x] = 2\cdot\ind{x\in p\N + \brace{a\midvert \abs{a} < p/4}} - 1$, for $p = 1/50$, satisfies separation in $\prob{\Y}$ \eqref{rates:eq:no-density}, but classes are not separated in $\X$.
    \item A situation where $\rho$ is uniform on $[-1, -.5] \cup [.5, 1]$, with $\E[Y\vert X=x] = \sign(x) (1 - \abs{x})^{p}$, for $p>0$, satisfies a separation of classes in $\X$ but does not satisfy \eqref{rates:eq:no-density}.
  \end{itemize}
  Note that continuity of $g^{*}$ and the no-density separation in~\eqref{rates:eq:no-density} imply separation of classes in $\X$.
  Note also that to get concentration inequality such as~\eqref{rates:eq:concentration}, one usually supposes that $g^{*}$ is smooth.
  We refer the curious reader to Section 2.4 in \citet{Steinwart2007} for separation in $\X$.
\end{remark}

The introduction of Assumption \ref{rates:ass:no-density} is motivated by the following result.

\begin{theorem}[Rates under no-density separation]\label{rates:thm:no-density}
  When $\ell$ is bounded by $\ell_{\infty}$ ({\em i.e.}, $\ell(z, y) \leq \ell_{\infty}$ for any $(z, y) \in {\cal Z} \times \Y$) and satisfies Assumption \ref{rates:ass:loss}, and $\mathcal{Z}$ is finite, under the no-density separation Assumption \ref{rates:ass:no-density}, and the concentration Assumption \ref{rates:ass:concentration}, the excess risk is controlled
  \[
    \E_{{\cal D}_{n}}{\cal R}(f_{n}) - {\cal R}(f^{*}) \leq \ell_{\infty} \exp\paren{-\frac{L_n t_0^2}{1 + M_n t_0}}.
  \]
\end{theorem}
\begin{proof}
  Because we make a mistake only when $d(g^{*}(x), F) \geq \norm{g_{n}(x) -	g^{*}(x)}$, we make no mistake when $\norm{g_{n}(x) - g^{*}(x)} < t_{0}$; otherwise we can consider the worse error we are going to pay, that is~$\ell_{\infty}$, leading to
  \[
    {\cal R}(f_{n}) - {\cal R}(f^{*}) \leq \ell_{\infty} \Pbb_{X}(\norm{g_{n}(x) - g^{*}(x)} > t_{0}).
  \]
  Taking the expectation with respect to ${\cal D}_{n}$ and using the fact that \(\E_A\Pbb_B(Z) = \E_A\E_B[\ind{Z}] = \E_B\E_A[\ind{Z}] = \E_B\Pbb_A(Z)\), and plug-in the concentration inequality~\eqref{rates:eq:concentration}, we get the result.
\end{proof}

\begin{example}[Image classification]
  In image classification, one can arguably assume that the class of an image is a deterministic function of this image.
  With the 0-1 loss, it implies that the image classification problem verifies the no-density separation.
  The same holds for any discrete problem where the label is a deterministic function of the input.
  Based on Theorem \ref{rates:thm:no-density} and~\eqref{rates:eq:concentration} in which $M$ is generally a constant when $L$ is proportional to the number of data, it is reasonable to ask for exponential convergences rates on such problems.
\end{example}

\subsection{Low density separation}

While we presented the no-density separation first for readability, it is a strong assumption.
Recall our example, Remark \ref{rates:rmk:separation}, with $\E[Y\vert X=x] = \sign(x)(1 - \abs{x})^{p}$, only around $x=1$ and $x=-1$ is $d(g^{*}(x), F)$ not bounded away from zero.
While the neighborhood of those points should be studied carefully, the error on all other points $x \in [-1+t, 1-t]$ can be controlled with exponential rates.
The low-density separation, also known as the Tsybakov margin condition in binary classification, will allow a refined control to get fast rates in such a setting.
\begin{assumption}[Low-density separation]\label{rates:ass:low-density}
  A surrogate solution $g^{*}$ is said to satisfy the \emph{low-density separation}, if there exists $c_{\alpha} > 0$, and $\alpha > 0$, such that for any $t > 0$
  \begin{equation}
    \label{rates:eq:low-density}
    \Pbb_{X}(d(g^{*}(X), F) < t) \leq c_{\alpha} t^{\alpha}.
  \end{equation}
  This condition is alternatively called the \emph{margin condition}.
\end{assumption}

The low-density separation spans all situations from the hard margin condition, that can be seen as $\alpha = +\infty$, to situations without any margin assumption corresponding to $\alpha = 0$.
The coefficient~$\alpha$ is an intrinsic measure of the easiness of finding $f^{*}$ in the problem $(\ell, \rho)$.
For example, the setting described in the last paragraph corresponds to the case $\alpha = 1/p$.
We discuss the equivalence of Assumption \ref{rates:ass:low-density} to definitions appearing in the literature in Remark \ref{rates:rmk:low-density}.

\begin{theorem}[Optimal rates under low density separation]\label{rates:thm:low-density}
  Under refined calibration in~\eqref{rates:eq:calibration}, concentration (Assumption \ref{rates:ass:concentration}), and low-density separation (Assumption \ref{rates:ass:low-density}), the risk is controlled as
  \[
    \E_{{\cal D}_{n}} {\cal R}(f_{n}) - {\cal R}(f^{*}) \leq
    2c_{\psi} c_{\alpha} c \paren{M_n^{\alpha + 1}L_n^{-(\alpha + 1)} + L_n^{-\frac{\alpha + 1}{2}}},
  \]
  for $c$ a constant that only depends on $\alpha$, that can be expressed through the Gamma function evaluated in quantity depending on $\alpha$, meaning that when $\alpha$ is big, $c$ behaves like $\alpha^{\alpha}$.
  Note that it is not possible to derive a better bound only given~\eqref{rates:eq:concentration}, \eqref{rates:eq:calibration} and \eqref{rates:eq:low-density}.
\end{theorem}

\begin{proof}[Sketch for Theorem \ref{rates:thm:low-density}, details in Appendix \ref{rates:proof:low-density}]
  Based on the refined calibration inequality in~\eqref{rates:eq:calibration}, and using that $\E[X] = \int_{0}^{\infty} \Pbb(X > t)\diff t$, it is possible to show that the expectation of the excess risk behave like
  \[
    \int_{0}^{\infty} \Pbb_{X}(d(g^{*}(x), F) < t) \sup_{x} \Pbb_{{\cal D}_{n}}(\norm{g_{n}(x)
      - g^{*}(x)} > t)\diff t.
  \]
  Based on Assumptions \ref{rates:ass:concentration} and \ref{rates:ass:low-density}, the integrand behaves like $t^{\alpha} \exp(- L_n t^{2} /(1+M_n t))$.
  A change of variable and the study of the Gamma function leads to the result.
  We provide all the details in Appendix \ref{rates:proof:low-density}.
  Note that while we stated Theorem \ref{rates:thm:low-density} under an exponential inequality of Bernstein type (Assumption \ref{rates:ass:concentration}), similar theorems can be derived for any type of exponential concentration inequality, as stated in Lemma \ref{rates:lem:ref-low-density} in Appendix \ref{rates:app:ref-low-density}.
\end{proof}

Theorem \ref{rates:thm:low-density} is to put in perspective with the work of \citet{Nowak2019} which considers the same setup as ours, yet only succeeds to derive acceleration by a power $2(\alpha+1) / (\alpha + 2)$, while we got an acceleration by a power $(\alpha + 1) / 2$ as already mentioned in the related work section.
This gain will appear more clearly in Theorem \ref{rates:thm:krr-low-density}.

\begin{remark}[Independence to the decomposition of $\ell$]\label{rates:rmk:low-density}
  While we have stated results based on the quantity $d(g^{*}(x), F)$, generalization of the Tsybakov margin condition has also been expressed through the quantity $\inf_{z\neq z^{*}} \E_{Y\sim \rho\vert_{x}}\ell(z, Y) - \E_{Y\sim \rho\vert_{x}}\ell(z^{*}, Y)$ instead of $d(g^{*}(x), F)$ \citep{Nowak2019}.
  We show in Appendix \ref{rates:proof:def-low-density} that the two definitions of the margin condition are equivalent.
\end{remark}

\begin{remark}[Scope of our work]\label{rates:rmk:l2-pointwise}
  Our work relies on pointwise exponential concentration inequalities (Assumption \ref{rates:ass:concentration}) which are specially designed to work well with the Tsybakov margin condition.
  It is natural for localized averaging methods such as nearest neighbors, or for surrogate methods leading to $L^\infty$ concentration.
  For surrogate methods leading to concentration of other quantities, it is possible to use similar tricks under different ``margin'' conditions ({\em e.g.} \citet{Steinwart2007} for a margin condition designed for the Hinge loss).

  Note that $L^2$ concentration on $g_n$ toward $g^*$ (such as the one derived by \citet{MarteauFerey2019} for logistic regression) could also be turned into fast convergence of $f_n$ toward $f^*$, since, in essence, for points $x\in\X$ where $\rho(\diff x)$ is high, the quantity $g^*(x) - g_n(x)$ will have a non-negligible contribution to $\norm{g^* - g_n}_{L^2}$ -- allowing to cast concentration in $L^2$ to concentration pointwise in $x$ -- and for points $x\in\X$ where $\rho(\diff x)$ is negligible, it is acceptable to pay the worst error, since it will have a small contribution to the excess of risk.

  Finally, note that it is also possible to let the right hand-side term in~\eqref{rates:eq:concentration} depends on $x$, and to modify Theorem \ref{rates:thm:no-density} with $L = \E[L(X)]$.
\end{remark}

\subsection{The importance of constants}

In this subsection, we discuss the importance of constants when providing learning rates.
Assumption \ref{rates:ass:no-density} corresponds to asking for $g^*(x)$ never to enter a neighborhood of $F$ defined through~$t_0$.
Similarly, when $\X$ is parametrized such that $\rho_{\X}$ is uniform, the parameter $\alpha$ in Assumption~\ref{rates:ass:low-density} corresponds to the speed at which $g^*(x)$ ``get through'' the decision frontier $F$.
In order to have a higher $\alpha$ and optimize the dependency in $n$ in the bound of Theorem \ref{rates:thm:low-density}, it is natural to think of infinitesimal perturbations of $g^*$ to make it cross the boundary orthogonally (or even jump over it and satisfy the no-density separation).
To give a precise example, in binary classification, let us artificially add smoothness to the function $g^*(x) = x^q$ when approaching zero.
Consider $g^*:[0, 1]\to[-1,1], x\to c^{q-p}x^p\ind{x<c} + x^q\ind{x\geq c}$, and $x$ uniform, and $p < q$.
In this setting, $\alpha$ can be taken anywhere in $[0, p^{-1})$.
Naively, we could ask for the biggest possible $\alpha$ in order to have the best dependency in $n$ in the learning rates given by Theorem \ref{rates:thm:low-density}.
While this approach will higher $\alpha$, it will also higher $c_\alpha$, compensating the gain one could expect from such a strategy.
Indeed, for $\alpha \in [0, p^{-1}]$, at best, we can take $c_\alpha = \ind{\alpha < q^{-1}} + c^{1-q\alpha}\ind{\alpha \geq q^{-1}}$.
This shows the importance of optimizing both $\alpha$ and $ c_\alpha c$ to minimize the lower bound appearing in Theorem \ref{rates:thm:low-density} when given a fixed number of sample $n$.

In a word, while we only give results that are optimized in $n$, when $n$ is fixed, better bounds could be given by optimizing parameters and constants simultaneously.
For example, when $\X = \R^d$ and $g^*$ belongs to the Sobolev space $H^m$ for all $m \in [0, m_*]$, and satisfies Assumption \ref{rates:ass:low-density} for all $\alpha \in [0, \alpha_*]$, we expect the best bound, that could be derived from our proof technique, to be of form
\[
  \E_{{\cal D}_n}{\cal R}(f_n) - {\cal R}(f^*) \leq
  \min_{m \leq m_*, \alpha < \alpha_*} \alpha^\alpha c_\alpha c_\psi\norm{g^*}_{H^m}
  n^{-\frac{m(\alpha + 1)}{2 d}}.
\]
Yet, for simplicity, we will express those bounds as $b n^{-\frac{m_*(\alpha_* + 1)}{2d}}$, for $b$ a big constant.

\section{Application to nearest neighbors}
\label{rates:sec:nn}
In this section, we consider the Bayes approximate risk estimator proposed by \citet{Stone1977}, with weights given by nearest neighbors \citep{Cover1957}.
We prove, under regularity assumptions, concentration inequalities similar to~\eqref{rates:eq:concentration}, which allow us to derive exponential and polynomial rates.
Given samples $(X_{i},Y_{i}) \sim \rho^{\otimes n}$, $k \in \N$ and a metric $d$ on $\X$, the estimator is
\begin{equation}\label{rates:eq:nn}
  \!\!\! g_{n}(x) = \sum_{i=1}^{n} \alpha_{i}(x) \phi(Y_{i}),\ \
  \text{with}\ \
  \alpha_{i}(x) = \left\{
  \begin{array}{cl}
    k^{-1}    & \text{if}\quad \sum_{j=1}^{n} \ind{d(x, X_{j}) \leq d(x, X_{i})} < k   \\
    0         & \text{if}\quad \sum_{j=1}^{n} \ind{d(x, X_{j}) < d(x, X_{i})} \geq k   \\
    (pk)^{-1} & \text{else, with } p = \sum_{j=1}^{n} \ind{d(x, X_{j}) = d(x, X_{i})}.
  \end{array}\right.
\end{equation}
To study the convergence of $g_{n}$, we introduce the noise free estimator $g_{n}^{*} = \sum_{i=1}^{n} \alpha_{i}(x)g^{*}(X_{i})$.
This allows separating the error due to the randomness of the labels $Y_{i} \sim \rho\vert_{X_{i}}$, and the error due to the difference between $g^{*}(x)$ and the averaging of $g^{*}$ on the neighbors of $x$ defining $g_{n}$.
To control the fist error, we need a bounded moment condition on $\phi(Y)$.
We reuse an assumption from \citet{Bernstein1924}, that is classic in machine learning \citep[\emph{e.g.},][]{Caponnetto2007,Lin2020}.

\begin{assumption}[Sub-exponential moment of $\rho\vert_{x}$]\label{rates:ass:moment}
  Suppose that there exists $\sigma^{2}, M > 0$ such that for any $x\in\supp\rho_{\X}$, for any $m \geq 2$, we have
  \[
    \E_{Y\sim\rho\vert_{x}}\bracket{\norm{\phi(Y) - g^{*}(x)}^{m}}
    \leq \frac{1}{2} m! \sigma^{2} M^{m-2}.
  \]
\end{assumption}

\begin{example}[Moment bound on $\phi(Y)$]
  Assumption \ref{rates:ass:moment} is a classical assumption that is notably satisfied when $\phi(Y)$ is bounded by $M$, with $\sigma^2$ its variance, or when $\paren{\phi(Y)\midvert X}$ is Gaussian with covariance bounded by a constant independent of $X$ \citep[see a proof of this standard result by][]{Fischer2020}.
\end{example}

To control the second error, we notice, for $x\in\supp\rho_{\X}$, that the quantity $\norm{g^{*}(x) - g_{n}^{*}(x)}$ behaves like $\sup_{x'\in{\cal B}(x, r)} \norm{g^{*}(x) - g^{*}(x')}$, with $r$ such that $\rho_{\X}({\cal B}(x, r)) \approx k/n$, such an $r$ modeling the distance between $x$ and its $k$-th neighbor.
This motivates the following assumption.

\begin{assumption}[Modified Lipschitz condition \citep{Chaudhuri2014}]\label{rates:ass:lipschitz}
  $g^{*}$ is said to verify the $\beta$-Modified Lipschitz condition if there exists $c_{\beta} > 0$ such that for any $x, x' \in \supp\rho_{\X}$
  \[
    \norm{g^{*}(x) - g^{*}(x')} \leq c_{\beta} \rho_{\X}({\cal B}(x, d(x,x')))^{\beta},
  \]
  where $d$ is the distance on $\X$, and ${\cal B}(x, t)\subset \X$ the ball of center $x$ and radius $t$.
\end{assumption}

Typically, the $\beta$ that appears in Assumption \ref{rates:ass:lipschitz} is linked with the dimension of a subset of $\X$ containing most of the mass of $\rho_{\X}$ (see below).
This will slow the rates according to this dimension parameter, a property referred to as the curse of dimensionality.

\begin{example}[Classical assumptions]\label{rates:ex:nn}
  When $\X = \R^{d}$, if $g$ is $\beta'$-H\"older continuous, and $\rho_{\X}$ is regular in the sense that, there exists a constant $c$ and $t^{*} > 0$ such that for $x\in\supp\rho_{\X}$ and any $t \in [0, t^{*}]$,
  \(
  \rho_{\X}({\cal B}(x, t)) \geq c \lambda({\cal B}(x, t)),
  \)
  with $\lambda$ the Lebesgue measure on $\X$, then $g$ satisfies the modified Lipschitz condition with $\beta = \beta' / d$.
  The condition on $\rho_{\X}$ is usually split in a condition of minimal mass of $\rho_{\X}$, and a condition of regular boundaries of $\supp\rho_{\X}$ \citep[\emph{e.g.},][]{Audibert2007}.
  We provide more details in Appendix \ref{rates:app:nn-assumptions}.
\end{example}

We now state convergence results, respectively proven in Appendices \ref{rates:proof:nn-concentration}, \ref{rates:proof:nn-no-density} and \ref{rates:proof:nn-low-density}, in which the constant values $b_1$ to $b_6$ appear explicitly.
Note that results provided by Lemma \ref{rates:lem:nn-concentration} are already known in the literature \citep{Gyorfi2002}, while Theorems \ref{rates:thm:nn-no-density} and \ref{rates:thm:nn-low-density} were only known in binary classification, but we generalize them to any discrete structured prediction problem.
It should be noted that rates in Theorem \ref{rates:thm:nn-low-density} match the minimax rates derived by \cite{Audibert2007} in the case of binary classification.

\begin{lemma}[Nearest neighbors concentration]\label{rates:lem:nn-concentration}
  Under Assumptions \ref{rates:ass:moment} and \ref{rates:ass:lipschitz}, there exist constants $b_{1}, b_{2}, b_{3} > 0$, such that for any $x\in\supp\rho_{\X}$ and any $t > 0$,
  \[
    \Pbb_{{\cal D}_{n}}\paren{\norm{g_{n}(x) - g_{n}^{*}(x)} > t} \leq
    2\exp\paren{-\frac{b_{1}kt^{2}}{1 + b_{2}t}}.
  \]
  And for $t > \paren{k/2n}^{\beta}$, when $\rho_{\X}$ is continuous\footnote{Note that this topological assumption ease derivations but	is not fundamental for such non-asymptotic results.}
  \[
    \Pbb_{{\cal D}_{n}}\paren{\norm{g_{n}^{*}(x) - g^{*}(x)} > t} \leq
    \exp\paren{-b_{3} nt^{\frac{1}{\beta}}}.
  \]
\end{lemma}
\begin{theorem}[Nearest neighbors fast rates under no-density assumption]
  \label{rates:thm:nn-no-density}
  When $\ell$ is bounded by $\ell_{\infty}$, satisfies Assumption \ref{rates:ass:loss}, and $\mathcal{Z}$ is finite, under the no-density separation, Assumption \ref{rates:ass:no-density}, and Assumptions \ref{rates:ass:moment} and \ref{rates:ass:lipschitz}, there exist two constants $b_{4}, b_{5} > 0$ that do not depend on $n$ or $k$ such that for any $n \in \N^{*}$ and any $k$ such that $\paren{k/2n}^{\beta} < t_{0}$, we have
  \begin{equation}
    \E_{{\cal D}_{n}}{\cal R}(f_{n}) - {\cal R}(f^{*}) \leq
    2\ell_{\infty}\exp(-b_{4}k) + \ell_{\infty}\exp(-b_{5}n).
  \end{equation}
\end{theorem}

\begin{theorem}[Nearest neighbors fast rates under low-density assumption]
  \label{rates:thm:nn-low-density}
  When $\ell$ satisfies Assumption \ref{rates:ass:loss}, and ${\cal Z}$ is finite, under the low-density separation, Assumption \ref{rates:ass:low-density}, and Assumptions \ref{rates:ass:moment} and \ref{rates:ass:lipschitz}, considering the scheme $k_{n} = \floor{k_{0}n^{\frac{2\beta}{2\beta + 1}}}$, for any $k_{0} > 0$, there exists a constant $b_{6} > 0$ that does not depend on $n$ such that for any $n \in \N^{*}$,
  \begin{equation}\label{rates:eq:nn_rates}
    \E_{{\cal D}_{n}}{\cal R}(f_{n}) - {\cal R}(f^{*}) \leq
    b_{6} n^{-\frac{\beta(\alpha + 1)}{2\beta + 1}}.
  \end{equation}
\end{theorem}

\begin{remark}[Scope of our work]
  The same type of argument works for other local averaging methods, such as Nadaraya-Watson \citep{Nadaraya1964,Watson1964}, local polynomials \citep{Cleveland1979,Audibert2007} or decision trees \citep{Breiman1984}.
\end{remark}
\begin{figure}[ht]
  \vspace*{-.5cm}
  \centering
  \includegraphics{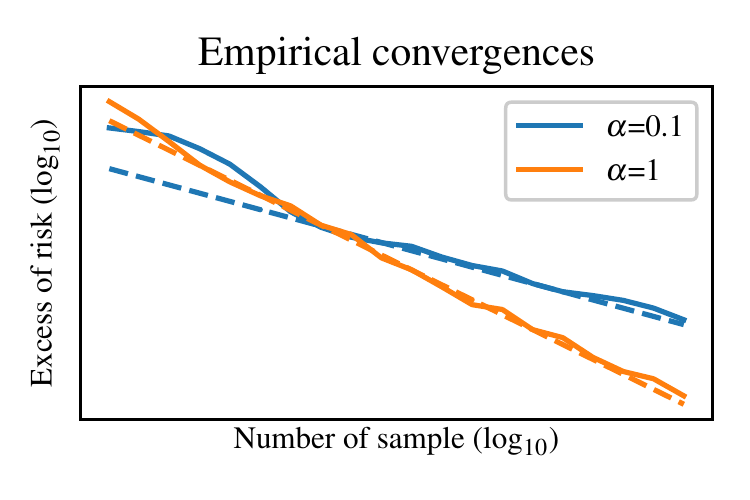}
  \caption{Empirical convergence rates.
  We consider binary classification, with $\X = [-1, 1]$, $g^{*}(x) = \sign(x)*\abs{x}^{\frac{1}{\alpha}}$, for $\alpha \in \brace{.1, 1}$ and $\rho_{\X}$ uniform.
  We plot in solid the logarithm of the excess risk averaged over 100 trials against the logarithm of the number of samples for $n \in [10, 10^6]$, and plot in dashed the expected slope of those curves due to Theorem \ref{rates:thm:nn-low-density} (\emph{i.e.}, we fit the constant $C$ in the rate $Cn^{-\gamma}$ with $\gamma$ obtained from the bound in~\eqref{rates:eq:nn_rates}).}
  \label{rates:fig:rates}
\end{figure}

\section{Application to reproducing kernel ridge regression}
\label{rates:sec:rkhs}

In this section, we consider the kernel ridge regression estimate $g_{n}$ of $g^{*}$ first proposed by \cite{Ciliberto2016}, and we prove, under regularity assumptions, uniform concentration inequalities similar to~\eqref{rates:eq:concentration}, which allow us to derive superfast rates at the end of the section.
Given a symmetric, positive semi-definite kernel $k:\X\times\X\to\R_+$, the kernel ridge regression estimation $g_{n}$ of $g^{*}$ is defined similarly to~\eqref{rates:eq:nn} yet with weights $\alpha(x) \in \R^{n}$ defined as
\[
  \alpha(x) = (\hat K + \lambda)^{-1} \hat K_{x},\quad
  \hat K = \paren{\frac{1}{n} k(X_{i}, X_{j})}_{i,j\leq n} \in \R^{n\times n},\
  \hat K_{x} = \paren{\frac{1}{n} k(x, X_{i})}_{i\leq n} \in \R^{n}.
\]
To state regularity assumptions, we introduce a minimal setup linked to the reproducing kernel $k$.
To keep the exposition clear, we relegate technicalities in Appendix \ref{rates:app:rkhs}.
We define the operator $K$ operating on functions $f\in L^{2}(\X, {\cal H}, \rho_{\X})$ and $K_{\X}$ operating on $f\in L^{2}(\X, \R, \rho_{\X})$, both defined as
\[
  (Kf)(x') = \int_{\X} k(x', x)f(x)\diff\rho_{\X}(x).
\]
Inheriting from the symmetry and positive semi-definiteness of $k$, $K$ is self-adjoint with the spectrum in $\R_+$.
To study the convergence of $g_{n}$ to $g^{*}$, it is useful to introduce the approximate orthogonal projection on $\ima K^{\frac{1}{2}}$, defined for $\lambda > 0$ as
\[
  g_\lambda = K (K+\lambda)^{-1} g^{*}.
\]
We introduce three assumptions linked with the regularity of the problem, referred to as the capacity condition, interpolation inequality and source condition.
Those are classical assumptions to prove uniform rates of the kernel ridge regression estimates.
They could be found, in particular, by \citet{Fischer2020} under the respective names of (EVD), (EMB) and (SRC), but also by \citet{PillaudVivien2018,Lin2020}.
Our assumptions differ in that they are expressed for vector-valued functions, which usually generate compactness issues \citep{Caponnetto2007}.
However, when ${\cal Z}$ is finite, ${\cal H}$ is finite dimensional, and $K$ can be shown to be a compact operator, which allows considering fractional power without definition issues.

\begin{assumption}[Capacity condition]\label{rates:ass:capacity}
  Suppose $\trace\paren{K_{\X}^{\sigma}} < +\infty$ for
  $\sigma \in [0, 1]$.
\end{assumption}

\begin{assumption}[Interpolation inequality]\label{rates:ass:interpolation}
  Assume the existence of $p \in [0,\frac{1}{2}]$, $c_p > 0$ such that
  \[
    \forall\ g\in \paren{\ker K}^{\perp}, \qquad
    \norm{K^{p}g}_{L^{\infty}} \leq c_p\norm{g}_{L^{2}}.
  \]
\end{assumption}

\begin{assumption}[Source condition]\label{rates:ass:source} Suppose
  $g^{*} \in \ima K^{q}$ for $q \in (p, 1]$.
\end{assumption}

When $q=\sfrac{1}{2}$, the source condition is often expressed as $g^{*}$ belonging to the reproducing kernel Hilbert space associated to the kernel $k$.
Note that when $k$ is bounded, Assumptions~\ref{rates:ass:capacity} and~\ref{rates:ass:interpolation} hold with $\sigma=1$ and $p=\sfrac{1}{2}$.
In those assumptions, for $p$ and $\sigma$ the smaller, and for $q$ the bigger, the faster the convergence rates will be.

\begin{example}[Classical assumptions]
  For Assumption \ref{rates:ass:interpolation} to hold, minimal mass and regular support of $\rho$, similarly to Example \ref{rates:ex:nn}, are often assumed, as well as regularity of functions in $\ima K^{p}$, in coherence with Remark \ref{rates:rmk:l2-pointwise}.
  For Assumption \ref{rates:ass:source} to hold, it is classical to assume regularity of $g^{*}$, matching the regularity of function spaces derived from the kernel $k$.
  The value of $\sigma$ in Assumption \ref{rates:ass:capacity} often comes as a bonus of regularity assumptions on $\rho$ and specificity of the RKHS implied by $k$.
  See Example 2 by \citet{PillaudVivien2018} and Section 4 by \citet{Fischer2020} as well as references therein for concrete examples.
\end{example}

We now state convergence results respectively proven in Appendices \ref{rates:proof:krr-1} and \ref{rates:proof:krr-2}, \ref{rates:proof:krr-no-density}, and \ref{rates:proof:krr-low-density}.
Lemma \ref{rates:lem:rkhs} is a generalization to vector-valued functions of kernel ridge regression uniform convergence rates known for real-valued function \citep[see][]{Fischer2020}.
Note that a similar result to Theorem \ref{rates:thm:krr-no-density} was provided for binary classification by \citet{Koltchinskii2005}, but we generalize exponential rates with kernel ridge regression to any discrete structured prediction problem.
Theorem \ref{rates:thm:krr-low-density} is new, even in the context of binary classification.
It states that, while, up to now, only rates in $n^{-1/4}$ were known for $f_{n}$ \citep{Ciliberto2020}, one can indeed hope for arbitrarily fast rates, depending on the hardness of the problem, read in the value of $\alpha\in[0, \infty)$.

\begin{lemma}[Reproducing kernel concentration]\label{rates:lem:rkhs}
  Under Assumptions \ref{rates:ass:capacity}, \ref{rates:ass:interpolation} and \ref{rates:ass:source}, for any $\lambda > 0$,
  \[
    \norm{g_\lambda - g^{*}}_{L^{\infty}} \leq b_{1} \lambda^{q-p}.
  \]
  With $b_{1} = c_p\norm{K^{-q}g^{*}}_{L^{2}}$.
  Moreover, when the kernel $k$ is bounded and under Assumption \ref{rates:ass:moment}, there exists three constants $b_{2}, b_{3}, b_{4}, b_{5} > 0$ that does not depend nor on $\lambda$ nor on $n$ such that
  \[
    \Pbb\paren{\norm{g_{n} - g_\lambda}_{\infty} > t} \leq
    b_{2} \lambda^{-\sigma}\exp\paren{-b_{3}n\lambda^{2p}}
    + 4\exp\paren{-\frac{n\lambda^{2p + \sigma}t^{2}}
    {b_{4}+b_{5}t}}.
  \]
  As long as $b_{3}n \geq \lambda^{-p}$, and $\lambda \leq \min\paren{\norm{K}_{\op}, 1}$.
\end{lemma}

\begin{theorem}[Kernel ridge regression fast rates under no-density assumption]
  \label{rates:thm:krr-no-density}
  When the loss $\ell$ is bounded, satisfies Assumption \ref{rates:ass:loss} and ${\cal Z}$ is finite, under the $t_{0}$-no-density separation condition, and Assumptions \ref{rates:ass:moment}, when $k$ is bounded, if $\lambda_{n} = \lambda$, for any $\lambda > 0$ such that $\norm{g^{*} - g_\lambda}_{L^{\infty}} < t_{0}$, then there exist two constants $b_{6}, b_{7} > 0$ such that, for any $n\in\N^{*}$,
  \begin{equation}
    \E_{{\cal D}_{n}} {\cal R}(f_{n}) - {\cal R}(f^{*}) \leq b_{6} \exp(-b_{7} n),
  \end{equation}
  with $f_{n}$ given by the kernel ridge regression surrogate estimate.
\end{theorem}

\begin{theorem}[Kernel ridge regression fast rates under low-density assumption]
  \label{rates:thm:krr-low-density}
  When $\ell$ satisfies Assumption \ref{rates:ass:loss}, is bounded and ${\cal Z}$ is finite, under the $\alpha$-low-density separation condition, and Assumptions \ref{rates:ass:moment}, \ref{rates:ass:capacity}, \ref{rates:ass:interpolation} and \ref{rates:ass:source}, if $\lambda_{n} = \lambda_{0} n^{-\frac{1}{2q + \sigma}}$, for any $\lambda_{0} > 0$, there exists $b_{8} > 0$, such that for any $n \in \N^{*}$,
  \begin{equation}
    \E_{{\cal D}_{n}} {\cal R}(f_{n}) - {\cal R}(f^{*}) \leq b_{8}	n^{-\frac{(q-p)(1+\alpha)}{2q + \sigma}}.
  \end{equation}
\end{theorem}

\section{Conclusion}

In this paper, we have shown how, for discrete problems, to leverage exponential concentration inequalities derived on continuous surrogate problems, in order to derive faster rates than rates directly obtained through calibration inequalities.
Those rates are arbitrarily fast, depending on a parameter characterizing the hardness of the discrete problem.
We have shown how this method directly applies to local averaging methods and to kernel ridge regression, which allows deriving ``superfast'' rates for any discrete structured prediction problem.

This opens the way to several follow-up, such as
\begin{itemize}
  \item Application follow-up, consisting of tackling concrete problem instances, such as predicting properties of DNA-sequence \citep{Jaakkola2000}, \emph{e.g.}, gene mutations responsible for diseases, with well-designed kernels on DNA in order to higher the exponent appearing in Theorem \ref{rates:thm:krr-low-density}.
  \item Computational follow-up, pushing our analysis further to understand how to design better algorithms on discrete problems.
        For example, by adding a regularization pushing $g_{n}$ away from the decision frontier $F$, and adding a term in $\ind{\norm{g_{n}(x) - g^{*}(x)} > d(g_{n}(x), F)}$ in~\eqref{rates:eq:calibration} for the analysis.
  \item Theoretical follow-up, to widen our analysis to other types of smooth surrogates, and to parametric methods, such as deep learning models, assuming that functions are parametrized by a parameter $\theta$, that some analysis gives concentration on $\theta_{n} - \theta^{*}$ similar to~\eqref{rates:eq:concentration} and that calibration inequalities relate the error on $\theta$ with the error between $f_{n}=f_{\theta_{n}}$ and $f^{*} = f_{\theta^{*}}$.
\end{itemize}

\begin{subappendices}
  \chapter*{Appendix}
  \addcontentsline{toc}{chapter}{Appendix}
  \section{Fast rates}
\label{rates:sec:proof}

In the following, we consider $\X$ and $\Y$ to be Polish spaces, {\em i.e.}, separable completely metrizable topological spaces, in order to define the distribution $\rho$.
We also consider $\cal Z$ endowed with a topology that makes it compact, and that makes $z\to\E_{Y\sim\mu}\ell(z, Y)$ continuous for any $\mu\in\prob{\Y}$, in order to have minimizer well-defined.
For a Polish space ${\cal A}$, we denote by $\prob{\cal A}$ the simplex formed by the set of Borel probability measures on this space.
For $\rho\in\prob{\X\times\Y}$, we denote by $\rho\vert_{x}$ the conditional distribution of $Y$ given $x$, and by $\rho_{\X}$ the marginal distribution over $\X$.
We suppose ${\cal H}$ separable Hilbert and that the mapping $\phi$ is measurable in order to define the pushforward measure $\phi_*\rho\vert_{x}$.
We assume that, for $\rho_{\X}$-almost every $x$, $(\phi(Y)\vert X=x)$ has a second moment, in order to consider the conditional mean $g^*(x)$ as the solution of the well-defined problem consisting of minimizing $\norm{\xi - \phi(Y)}^2$ for $\xi \in {\cal H}$.
We consider $\psi$ to be continuous, in order to have the decoding problem well posed.

\subsection{Proof of Lemma \ref{rates:lem:cal}}
\label{rates:proof:cal}

With the notation of Lemma \ref{rates:lem:cal}, for $x\in\supp\rho_{\X}$
\begin{align*}
  \E_{Y\sim\rho\vert_{x}} & \bracket{\ell(f_n(x), Y) - \ell(f^*(x), Y)}
  = \scap{\psi(f_n(x)) - \psi(f^*(x))}{g^*(x)}_{\cal H}
  \\& = \scap{\psi(f_n(x))}{g_n(x)}
  + \scap{\psi(f_n(x))}{g^*(x) - g_n(x)}
  - \scap{\psi(f^*(x))}{g^*(x)}
  \\& \leq \scap{\psi(f^*(x))}{g_n(x)}
  + \scap{\psi(f_n(x))}{g^*(x) - g_n(x)}
  - \scap{\psi(f^*(x))}{g^*(x)}
  \\& = \scap{\psi(f_n(x)) - \psi(f^*(x))}{g^*(x) - g_n(x)}
  \\& \leq \norm{\psi(f_n(x)) - \psi(f^*(x))}_{\cal H}\norm{g^*(x) - g_n(x)}_{\cal H}
  \\& \leq 2c_{\psi}\norm{g^*(x) - g_n(x)}_{\cal H},
\end{align*}
where the inequality $\scap{\psi(f_n(x))}{g_n(x)} \leq \scap{\psi(f^*(x))}{g_n(x)}$ is due to the fact that $f_n(x)$ minimizes the functional $z\to \scap{\psi(z)}{g_n(x)}$.
Integrating over $\X$ leads to the results in Lemma \ref{rates:lem:cal}.

\subsection{Proof of Lemma \ref{rates:lem:calibration}}
\label{rates:proof:calibration}

The first part of the lemma is a geometrical result stating that to go from two elements $\xi_1$ and $\xi_2$ in $\prob{\phi(\Y)}$, leading to two different decoding, one has to pass by a point $\xi_{1/2} \in F$, where there is at least two possible decodings.
Let us make it clearer.
Consider $x\in\supp\rho_{\X}$ and suppose that $f_n(x) \neq f^*(x)$, define the path
\[
  \myfunction{\zeta}{[0,1]}{\prob{\phi(\Y)}}{\lambda}{\lambda g_n(x) +
    (1-\lambda)g^*(x).}
\]
Consider $d:\prob{\phi(\Y)}\to{\cal Z}$ the decoding function used to retrieve $f^*$ and $f_n$, from $g^*$ and $g_n$, satisfying $d(\xi) \in \argmin_{z\in{\cal Z}} \scap{\psi(z)}{\xi}$.
Consider the path $d\circ\zeta:[0, 1] \to {\cal Z}$, it goes from $\zeta(0) = f^*(x)$ to $\zeta(1) = f_n(x)$.
Consider $\lambda_{\infty}$ the supremum of $(d\circ\zeta)^{-1}(f^*(x))$.
We will show that $\zeta(\lambda_{\infty}) \in F$, this will lead to
\[
  \norm{g_n(x) - g^*(x)} = \norm{g_n(x) - \zeta(\lambda_{\infty})}
  + \norm{\zeta(\lambda_{\infty}) - g^*(x)}
  \geq \norm{\zeta(\lambda_{\infty}) - g^*(x)}
  \geq d(g^*(x), F),
\]
and to Lemma \ref{rates:lem:calibration} by contraposition.

To show that $\zeta(\lambda_{\infty}) \in F$, we will show that \( f^*(x) \in \argmin_{z} \scap{\psi(z)}{\zeta(\lambda_{\infty})} \not\subset\brace{f^*(x)}.\)
By definition of the supremum, there exists a sequence $(\lambda_p)_{p\in\N}$ converging to $\lambda_{\infty}$ such that
\[
  f^*(x) \in
  \argmin_{z} \scap{\psi(z)}{\lambda_p g_n(x) + (1-\lambda_p)g^*(x)},
\]
meaning that for all $z\neq f^*(x)$
\[
  \scap{\psi(f^*(x))}{\lambda_p g_n(x) + (1-\lambda_p)g^*(x)}
  \leq \scap{\psi(z)}{\lambda_p g_n(x) + (1-\lambda_p)g^*(x)}.
\]
By continuity of the scalar product, it holds for $p=\infty$, which means $f^*(x) \in \argmin_z \scap{\psi(z)}{\zeta(\lambda_{\infty})}$.
Now, suppose that $\argmin_z\scap{\psi(z)}{\zeta(\lambda_{\infty})} = \brace{f^*(x)}$, this means that for all $z\neq f^*(x)$,
\[
  \scap{\psi(f^*(x))}{\lambda_{\infty} g_n(x) + (1-\lambda_{\infty})g^*(x)}
  < \scap{\psi(z)}{\lambda_{\infty} g_n(x) + (1-\lambda_{\infty})g^*(x)}.
\]
By continuity of this function according to $\lambda$, this means that this still holds for $\lambda_{\infty} + \epsilon_z$ for $\epsilon_z > 0$.
Taking $\epsilon = \inf_{z\in{\cal Z}} \epsilon_z$, it means that $\lambda_{\infty} + \epsilon \in (d\circ\zeta)^{-1}(f^*(x))$.
When ${\cal Z}$ is finite, $\epsilon > 0$, which contradicts the definition of $\lambda_{\infty}$.
Therefore, $\zeta(\lambda_{\infty}) \in F$.

The second part of Lemma \ref{rates:lem:calibration} follows from derivations in Appendix \ref{rates:proof:cal}.

\begin{remark}[Extension to discrete cases]\label{rates:rmk:extension-1}
  Note that the same argument can be generalized to discrete problems -- which could be defined as ${\cal Z}$ endowed with a topology that makes $z\to\E_{Y\sim\mu}[\ell(z, Y)]$ continuous with respect to $z$, and ${\cal Z}\setminus\brace{z}$ locally compact for any $z\in\cal Z$ -- that are not degenerate, in the sense that $\rho_{\X}$ almost all $x\in\X$, there exists $t > 0$ such that the cardinality of the set defined as $\brace{z\midvert\E_{Y\sim\rho\vert_{x}}[\ell(z, Y)] - \inf_{z'\in\Y}\E_{Y\sim\rho\vert_{x}}[\ell(z', Y)] < t}$ if finite.
  This holds for classification with infinite countable classes, but it does not for regression on the set of rational numbers.
\end{remark}

\begin{remark}[Extension to general cases]\label{rates:rmk:extension-2}
  To remove the condition ${\cal Z}$ finite, one can change the definition of $d(g^*(x), F)$ to
  \(
  \inf_{\xi\in{\cal H}; \brace{f^*(x)} \neq \argmin\scap{\psi(z)}{\xi}} \norm{\xi - g^*(x)},
  \)
  in order to make Lemma \ref{rates:lem:calibration} hold for any ${\cal Z}$.
\end{remark}

\subsection{Equivalence between generalizations of the Tsybakov margin condition}
\label{rates:proof:def-low-density}

While we state the margin condition with $d(g^*(x), F)$, it could also be stated with $d(g^*(x), F\cap\hull(\phi(\Y)))$ or with, which is the quantity considered by \citep{Nowak2019},
\[
  \gamma(x) = \inf_{z\neq z^*} \E_{Y\sim \rho\vert_{x}}\ell(z, Y) -
  \E_{Y\sim \rho\vert_{x}}\ell(z^*, Y)
  = \inf_{z\neq z^*}\scap{\psi(z) - \psi(z^*)}{g^*(x)}.
\]
Indeed, when $\cal Z$ is finite and $\ell$ is proper in the sense that $\ell(\cdot, y) = \ell(\cdot, z)$ implies $z=y$, and that there is no $z$ that minimizes a linear combination of $(\ell(\cdot, y))_{y\in\Y}$ without minimizing a convex combination of the same family, we have the existence of two constants such that
\[
  c\gamma(x) \leq d(g^*(x), F\cap\hull(\phi(\Y))) \leq d(g^*(x), F) \leq c'\gamma(x).
\]

\subsubsection{Mildness of our condition}
Let $z'$ be the argmin defining $\gamma$, geometric properties of the scalar product imply the existence of a $\xi \in (\psi(z') - \psi(z^*))^\perp$ such that
\[
  \scap{\psi(z') - \psi(z^*)}{g^*(x)}
  = \norm{\psi(z') - \psi(z^*)} \norm{g^*(x) - \xi}.
\]
Therefore,
\[
  \scap{\psi(z') - \psi(z^*)}{g^*(x)}
  \geq \min_{y, y'} \norm{\psi(y) - \psi(y')}
  \norm{g^*(x) - \xi}.
\]
Note that, by definition of $\xi$, $\scap{\xi}{\psi(z')} = \scap{\xi}{\psi(z^*)}$.
If $\xi \in R_{z^*}$ then $\xi \in F$, otherwise $\xi \notin R_{z^*}$ and then, there exists a point between $\xi$ and $g^*(x)$ that belongs to the decision frontier (see Appendix \ref{rates:proof:calibration} for a proof - for which we need some regularity assumption such as $\cal Z$ finite).
In every case,
\[
  \norm{g^*(x) - \xi} \geq d(g^*(x), F).
\]
This implies the existence of $c'$.

\subsubsection{Strength of our condition}
For any $g_n$ such that $f_n(x) = z$, we have
\begin{align*}
  \scap{\psi(z) - \psi(z^*)}{g^*(x)}
   & = \scap{\psi(z)}{g^*(x) - g_n(x)}
  + \scap{\psi(z)}{g_n(x)} - \scap{\psi(z^*)}{g^*(x)}
  \\&\leq \scap{\psi(z)}{g^*(x) - g_n(x)}
  + \scap{\psi(z^*)}{g_n(x)} - \scap{\psi(z^*)}{g^*(x)}
  \\&\leq 2 c_\psi\norm{g^*(x) - g_n(x)}.
\end{align*}
If we take the infimum on both sides we have
\[
  d(g^*(x), F) = \inf_{g_n(x)\notin R_{f^*(x)}} \norm{g_n(x) - g^*(x)}
  \geq \frac{1}{2c_\psi}\inf_{z\neq z^*} \scap{\psi(z) - \psi(z^*)}{g^*(x)},
\]
where the left equality is provided, when ${\cal Z}$ is finite, by a similar reasoning to the one in Appendix \ref{rates:proof:calibration}.
This implies the existence of $c$.
Note also that if the loss is proper in the sense that if $z$ minimizes $\scap{\psi(z)}{\xi}$ for a $\xi\in{\cal H}$, there exists a $\xi \in \hull{\phi(\Y)}$ such that $z$ minimizes $\scap{\psi(z)}{\xi}$, we can consider $g_n(x) \in \hull(\phi(\Y))$, and therefore restrict $F$ to $F\cap\hull{\phi(\Y)}$.
Finally, we have shown that, when $\cal Z$ finite and $\ell$ proper
\[
  c\gamma(x) \leq d(g^*(x), F\cap\hull(\phi(\Y))) \leq d(g^*(x), F) \leq c'\gamma(x).
\]
This explains why we would consider $\gamma(x)$, $d(g^*(x), F\cap\hull(\phi(\Y)))$ or $d(g^*(x), F)$ to define the margin condition, it will only change the value of constants in Assumptions \ref{rates:ass:no-density} and \ref{rates:ass:low-density}.

\subsection{Refinement of Theorem \ref{rates:thm:no-density}}
\label{rates:proof:no-density}

It is possible to refine Theorem \ref{rates:thm:no-density} to remove the condition that the loss $\ell$ is bounded.
In the following, we omit the dependency of $L_n$ and $M_n$ to $n$.

\begin{lemma}[Refinement of Theorem \ref{rates:thm:no-density}]
  \label{rates:lem:ref-no-density}
  Under refined calibration \eqref{rates:eq:calibration}, concentration, Assumption \ref{rates:ass:concentration}, and no-density separation, Assumption \ref{rates:ass:no-density}, the risk is controlled as
  \[
    \E_{{\cal D}_n} {\cal R}(f_n) - {\cal R}(f^*) \leq
    4c_\psi L^{-1/2} \exp\paren{-\frac{t_0^2 L}{2}}^{1/2}
    + 4c_\psi ML^{-1} \exp\paren{- \frac{t_0 L}{2M}}.
  \]
  Note that it is not possible to derive a better bound only given~\eqref{rates:eq:concentration}, \eqref{rates:eq:calibration} and \eqref{rates:eq:no-density}.
  Yet when $\ell$ is bounded by $\ell_{\infty}$, we have
  \[
    \E_{{\cal D}_n} {\cal R}(f_n) - {\cal R}(f^*) \leq \ell_{\infty}
    \exp\paren{-\frac{Lt_0^2}{1 + Mt_0}}.
  \]
\end{lemma}
\begin{proof}
  Using the calibration inequality along with the no-density separation one, we get
  \begin{align*}
    {\cal R}(f_n) - {\cal R}(f^*)
     & \leq 2c_\psi \E_{X}\bracket{\ind{\norm{g_n(X) - g^*(X)} \geq t_0} \norm{g_n(X) - g^*(X)}} \\
     & = 2c_\psi \int_{t_0}^\infty \Pbb_{X}\paren{\norm{g_n(X) - g^*(X)} \geq t} \diff t.
  \end{align*}
  Taking the expectation over ${\cal D}_n$ and using concentration inequality we have
  \begin{align*}
    \E_{{\cal D}_n}{\cal R}(f_n) - {\cal R}(f^*)
     & \leq 2c_\psi \int_{t_0}^\infty \Pbb_{X, {\cal D}_n}\paren{\norm{g_n(X) - g^*(X)} \geq t} \diff t \\
     & \leq 2c_\psi \int_{t_0}^\infty \exp\paren{-\frac{Lt^2}{1+Mt}} \diff t.
  \end{align*}
  We only need to study the integral
  \(
  \int_{t_0}^\infty \exp\paren{-\frac{Lt^2}{1+Mt}} \diff t.
  \)
  We first clean the dependency on $t$ inside the exponential using that
  \[
    \frac{1}{2}\paren{\exp(-Lt^2) + \exp\paren{-\frac{Lt}{M}}}
    \leq \exp\paren{-\frac{Lt^2}{1 + Mt}}
    \leq \exp\paren{-\frac{Lt^2}{2}} + \exp\paren{-\frac{Lt}{2M}}.
  \]
  We are left with the study of
  \(
  \int_{t_0}^\infty \exp(-At^p)\diff t,
  \)
  for $p \in \brace{1, 2}$ and $A > 0$.
  The case $p=1$, directly leads to
  \(
  A^{-1}\exp(-At_0),
  \)
  explaining the part in $L / M$.
  The case $p=2$ is similar to the Gaussian integral, and can be handled with the following tricks
  \begin{align*}
    \int_{t_0}^\infty \exp(-At^2)\diff t
     & = \frac{1}{2} \int_{(-\infty, -t_0]\cup[t_0, \infty)} \exp(-At^2)\diff t
    \\&= \frac{1}{2}\paren{\int_{((-\infty, -t_0]\cup[t_0, \infty))^2} \exp(-A\norm{x}^2))
    \diff x}^{1/2}.
  \end{align*}
  This last integral corresponds to integrate the function $x\to\exp(-A\norm{x}^2)$ for $x\in\R^2$ on the domain $((-\infty, -t_0]\cup[t_0, \infty))^2$.
  This function being positive and the domain being included in the domain $\brace{\norm{x} \geq t_0}$ and containing the domain $\brace{\norm{x} \geq \sqrt{2}t_0}$, we get
  \begin{align*}
    \int_{\brace{\norm{x}\geq \sqrt{2}t_0}}^\infty \exp(-A\norm{x}^2)\diff x
    \leq \paren{2\int_{t_0}^\infty \exp(-At^2)\diff t}^2
    \leq \int_{\brace{\norm{x}\geq t_0}}^\infty \exp(-A\norm{x}^2)\diff x.
  \end{align*}
  Using polar coordinate we get
  \[
    \int_{\brace{\norm{x}\geq t_0}}^\infty \exp(-A\norm{x}^2)\diff x
    = 2\pi\int_{t_0}^\infty r\exp(-Ar^2)\diff r
    = \pi A^{-1} \exp(-At_0^2).
  \]
  Therefore,
  \[
    2^{-1}\sqrt{\pi} A^{-1/2} \exp(-A 2 t_0^2)^{1/2}
    \leq \int_{t_0}^\infty \exp(-At^2)\diff t
    \leq 2^{-1}\sqrt{\pi} A^{-1/2} \exp(-A t_0^2)^{1/2}.
  \]
  This explains the rates in $L$.
\end{proof}

\subsection{Proof of Theorem \ref{rates:thm:low-density}}
\label{rates:proof:low-density}

Using the calibration and Bernstein inequalities we get, omitting the dependency of $L_n$ and $M_n$ to $n$,
\begin{align*}
  \E_{{\cal D}_n}{\cal R}(f_n) - {\cal R}(f^*)
   & \leq 2c_{\psi}\E_{{\cal D}_n, X}\bracket{\ind{d(g_n(X), g^*(X)) \geq d(g^*(X), F)}\norm{g_n(X) - g^*(X)}}
  \\& = 2c_{\psi}\int_{0}^\infty \Pbb_{{\cal D}_n, X}\paren{\ind{d(g_n(X), g^*(X)) \geq d(g^*(X), F)}\norm{g_n(X) - g^*(X)} \geq t} \diff t
  \\& = 2c_{\psi}\int_{0}^\infty \E_{X}\Pbb_{{\cal D}_n}\paren{\norm{g_n(X) - g^*(X)} \geq \max\brace{t, d(g^*(X), F)}} \diff t
  \\& \leq 2c_{\psi}\int_{0}^\infty \E_{X}\exp\paren{-\frac{L\max\brace{t, d(g^*(X), F)}^2}{1 + M \max\brace{t, d(g^*(X), F)}^2}} \diff t
  \\& = 2c_{\psi}\int_{0}^\infty \E_{X}\bracket{\ind{d(g^*(X), F) < t} \exp\paren{-\frac{Lt^2}{1 + Mt}}}\diff t
  \\&\qquad + 2c_{\psi} \int_0^\infty \E_{X}\bracket{\ind{d(g^*(X), F) \geq t}\exp\paren{-L\frac{d(g^*(X), F)^2}{1 + Md(g^*(X), F)}}}\diff t
  \\& = 2c_{\psi}\int_{0}^\infty \Pbb_{X}\paren{d(g^*(X), F) < t} \exp\paren{-\frac{Lt^2}{1 + Mt}}\diff t
  \\& \qquad + 2c_{\psi}\E_{X}\bracket{d(g^*(X), F) \exp\paren{-\frac{Ld(g^*(X), F)^2}{1 + Md(g^*(X), F)}}}.
\end{align*}
Let us begin by working on the first term.
We have, using the low-density separation hypothesis
\begin{align*}
  \int_{0}^\infty \Pbb_{X}\paren{d(g^*(X), F) < t} \exp\paren{-\frac{Lt^2}{1 + Mt}}\diff t
   & \leq c_{\alpha} \int_{0}^\infty t^\alpha \exp(-\frac{Lt^2}{1+Mt})\diff t.
\end{align*}
Recall the expression of the Gamma integral
\[
  \int_0^\infty t^\alpha \exp(-Lt) \diff t = \frac{\Gamma(\alpha+1)}{L^{\alpha+1}}
  \qquad\text{and}\qquad
  \int_0^\infty t^\alpha \exp(-Lt^2) \diff t = \frac{\Gamma\paren{\frac{\alpha+1}{2}}}{2 L^{\frac{\alpha+1}{2}}}.
\]
Let us briefly talk about optimality.
Up to now, we have only used three inequalities: calibration, exponential concentration and low-density separation.
Therefore, when those inequalities hold as equalities, we get a lower bound of order on the excess of risk as
\begin{align*}
  \E_{{\cal D}_n}{\cal R}(f_n) - {\cal R}(f^*)
   & \geq 2c_{\psi}c_{\alpha} \int_{0}^\infty t^\alpha \exp(-\frac{Lt^2}{1+Mt})\diff t
  \\&\geq 2c_{\psi}c_{\alpha} \int_{0}^\infty \frac{1}{2} t^\alpha \paren{\exp(-Lt^2) + \exp\paren{-\frac{Lt}{M}}}\diff t
  \\&= 2c_{\psi} c_{\alpha}\paren{\frac{\Gamma\paren{\frac{\alpha+1}{2}}}{4} L^{-\frac{\alpha+1}{2}}
  + \frac{\Gamma(\alpha+1)}{2} M^{\alpha+1}L^{-(\alpha + 1)}}.
\end{align*}
For the upper bound, using that $\exp(-a/1+b) \leq \exp(-a/2) + \exp(-a/2b)$, we get
\begin{align*}
  \int_{0}^\infty t^\alpha \exp(-\frac{Lt^2}{1+Mt})\diff t
   & \leq \int_{0}^\infty t^\alpha \exp(-\frac{Lt^2}{2})\diff t
  + \int_{0}^\infty t^\alpha \exp(-\frac{Lt}{2M})\diff t
  \\&= 2^{\frac{\alpha - 1}{2}} \Gamma\paren{\frac{\alpha+1}{2}} L^{-\frac{\alpha + 1}{2}}
  + 2^{\alpha+1} \Gamma(\alpha + 1) M^{\alpha + 1} L^{-(\alpha + 1)} .
\end{align*}

Let study the second term in the excess of risk inequality.
To enhance readability, write $\eta(X) = d(g^*(X), F)$.
We will first dissociate the two parts in the exponential with
\begin{align*}
  \E_{X}\bracket{\eta(X) \exp\paren{-\frac{L\eta(X)^2}{1 + M\eta(X)}}}
  \leq \E_{X}\bracket{\eta(X) \paren{\exp\paren{-\frac{L\eta(X)^2}{2}} + \exp\paren{-\frac{L\eta(X)}{2M}}}.}
\end{align*}
We are left with studying $\E[\eta(X) \exp(-A\eta(X)^p)]$, for $A > 0$ and $p\in \brace{1, 2}$.
The function $t\to~t\exp(-At^p)$ achieves its maximum in $t_0 = (pA)^{-1/p}$, increasing before and decreasing after.
Notice that the quantity
\begin{align*}
  \Pbb(\eta(X) < t_0)\E_{X}\bracket{\eta(X) \exp(-A\eta(X)^p)\midvert \eta(X) < t_0}
   & \leq c_{\alpha} t_0^{\alpha + 1} \exp(-A t_0^p)
  \\&= c_{\alpha} p^{-\frac{\alpha + 1}{p}}\exp(-p^{-1/p}) A^{-\frac{\alpha + 1}{p}},
\end{align*}
is exactly of the same order as the control we had on the first term in the excess of risk decomposition.
This suggests considering the following decomposition
\begin{align*}
   & \E_{X}\bracket{\eta(X)\exp(-A\eta(X)^p)}=
  \Pbb(\eta(X) < t_0)\E_{X}\bracket{\eta(X) \exp(-A\eta(X)^p)\midvert \eta(X) < t_0}
  \\&\qquad +\sum_{i=0}^\infty \Pbb(2^i t_0\leq \eta(X) < 2^{i+1}t_0)\E_{X}\bracket{\eta(X) \exp(-A\eta(X)^p)\midvert 2^it_0 \leq \eta(X) < 2^{i+1}t_0}
  \\&\qquad \leq c_{\alpha} t_o^{\alpha+1} \exp(-At_0^p)
  + \sum_{i=0}^\infty c_{\alpha} 2^\alpha (2^i t_0)^{\alpha + 1} \exp(-At_0^p (2^i)^p)
  \\&\qquad = c_{\alpha} t_o^{\alpha+1} \paren{\exp(-p^{-1/p})
    + \sum_{i=0}^\infty 2^\alpha 2^{i(\alpha + 1)} \exp(-p^{-1/p} 2^{ip})}.
\end{align*}
The convergence of the last series, ensures the existence of a constant $c$ such that
\begin{align*}
  \E_{X}\bracket{\eta(X) \exp\paren{-\frac{L\eta(X)^2}{1 + M\eta(X)}}}
   & \leq c \paren{\paren{\frac{L}{2M}}^{-(\alpha + 1)} + \paren{\frac{L}{2}}^{-\frac{\alpha + 1}{2}}}.
\end{align*}
Adding everything together, we get the existence of two constants $c', c''$, such that
\[
  \E_{{\cal D}_n}{\cal R}(f_n) - {\cal R}(f^*) \leq 2c_{\psi}c_{\alpha}\paren{c' M^{\alpha+1}
    L^{-(\alpha+ 1)} + c'' L^{-\frac{\alpha + 1}{2}}}.
\]
This ends the proof by considering $c = \max\paren{c', c''}$.

\subsection{Refinement of Theorem \ref{rates:thm:low-density}}
\label{rates:app:ref-low-density}

Some convergence analyses lead to exponential inequalities that are not of Bernstein type, indeed, our result still holds in those settings, as mentioned by the following lemma.
In the following, we omit the dependency of $L_n$ and $M_n$ to $n$.

\begin{lemma}[Refinement of Theorem \ref{rates:thm:low-density}]
  \label{rates:lem:ref-low-density}
  Under the assumptions of Theorem \ref{rates:thm:low-density}, if the concentration is not given by Assumption \ref{rates:ass:concentration} but given, for some positive constants $(a_{i}, b_{i}, p_{i})_{i\leq m}$, by, for all $x\in\supp\rho_{\X}$ and $t > 0$,
  \[
    \Pbb_{{\cal D}_n}(\norm{g_n(x) - g^*(x)} > t) \leq \sum_{i=1}^n a_{i} \exp(-b_{i}t^{p_{i}}).
  \]
  Then the excess of risk is controlled by
  \[
    \E_{{\cal D}_n}{\cal R}(f_n) - {\cal R}(f^*) \leq c\sum_{i=1}^n a_{i} b_{i}^{-\frac{\alpha + 1}{p_{i}}},
  \]
  for a constant $c$ that does not depend on $(a_{i}, b_{i})_{i\leq m}$.
\end{lemma}
\begin{proof}
  First of all, remark that the proof of Theorem \ref{rates:thm:low-density} is linear in $\Pbb_{{\cal D}_n}(\norm{g_n(x) - g^*(x)} > t)$, therefore we only need to prove this lemma for $(a, b, p)$, for which we proceed as in Theorem \ref{rates:thm:low-density}
  \begin{align*}
    \E_{{\cal D}_n}{\cal R}(f_n) - {\cal R}(f^*)
     & \leq 2c_{\psi} \E_{{\cal D}_n, X}\bracket{\ind{\norm{g_n(X) - g(X)} \geq d(g(X), F)}\norm{g_n(X) - g(X)}}
    \\& = 2c_{\psi}\int_{0}^\infty \Pbb_{{\cal D}_n, X}\paren{\ind{\norm{g_n(X) - g(X)} \geq d(g(X), F)}\norm{g_n(X) - g(X)} \geq t} \diff t
    \\& = 2c_{\psi}\int_{0}^\infty \E_{X}\Pbb_{{\cal D}_n}\paren{\norm{g_n(X) - g(X)} \geq \max\brace{t, d(g(X), F)}} \diff t
    \\& \leq 2c_{\psi}a \int_{0}^\infty \E_{X}\exp\paren{- b\max\brace{t, d(g(X), F)}^p} \diff t
    \\& = 2c_{\psi}a\int_{0}^\infty \E_{X}\bracket{\ind{d(g(X), F) < t} \exp\paren{-bt^p}}\diff t
    \\&\qquad\qquad + 2c_{\psi}a \int_0^\infty \E_{X}\bracket{\ind{d(g(X), F) \geq t}\exp\paren{-bd(g(X), F)^p}}\diff t
    \\& = 2c_{\psi}a\int_{0}^\infty \Pbb_{X}\paren{d(g(X), F) < t} \exp\paren{-bt^p}\diff t
    \\&\qquad\qquad+ 2c_{\psi}a\E_{X}\bracket{d(g(X), F) \exp\paren{-bd(g(X), F)^p}}.
  \end{align*}
  Let us begin by working on the first term.
  We have, using the low-density separation hypothesis
  \begin{align*}
     & \int_{0}^\infty \Pbb_{X}\paren{d(g(X), F) < t} \exp\paren{-bt^2}\diff t
    \leq c_{\alpha} \int_{0}^\infty t^\beta \exp(-bt^p)\diff t.
    \\ &\qquad\qquad= b^{-\frac{1+\beta}{p}} c_{\alpha} \int_{0}^\infty (b^{1/p}t)^\beta \exp(-(b^{1/p}t)^p)\diff (b^{1/p}t).
    \\ &\qquad\qquad= b^{-\frac{1+\beta}{p}} c_{\alpha} \int_{0}^\infty t^\beta \exp(-t^p)\diff t = c_{\alpha} \Gamma(\beta, p) b^{-\frac{1+\beta}{p}}.
  \end{align*}
  Let study the second term in the excess of risk inequality.
  To enhance readability, write $\eta(X) = d(g(X), F)$.
  We are left with studying $\E[\eta(X) \exp(-b\eta(X)^p)]$.
  The function $t\to~t\exp(-bt^p)$ achieves its maximum in $t_0 = (pb)^{-1/p}$, increasing before and decreasing after.
  Notice that the quantity
  \begin{align*}
    \Pbb(\eta(X) < t_0)\E_{X}\bracket{\eta(X) \exp(-b\eta(X)^p)\midvert \eta(X) < t_0}
     & \leq c_{\alpha} t_0^{\beta + 1} \exp(-b t_0^p)
    \\&= c_{\alpha} p^{-\frac{\beta + 1}{p}}\exp(-p^{-1/p}) b^{-\frac{\beta + 1}{p}},
  \end{align*}
  is exactly of the same order as the control we had on the first term in the excess of risk decomposition.
  This suggests considering the following decomposition
  \begin{align*}
     & \E_{X}\bracket{\eta(X)\exp(-b\eta(X)^p)}
    =
    \Pbb(\eta(X) < t_0)\E_{X}\bracket{\eta(X) \exp(-b\eta(X)^p)\midvert \eta(X) < t_0}
    \\&\quad\qquad +\sum_{i=0}^\infty \Pbb(2^i t_0\leq \eta(X) < 2^{i+1}t_0)\E_{X}\bracket{\eta(X) \exp(-b\eta(X)^p)\midvert 2^it_0 \leq \eta(X) < 2^{i+1}t_0}
    \\&\qquad \leq c_{\alpha} t_o^{\beta+1} \exp(-bt_0^p)
    + \sum_{i=0}^\infty c_{\alpha} 2^\beta (2^i t_0)^{\beta + 1} \exp(-bt_0^p (2^i)^p)
    \\&\qquad = c_{\alpha} t_o^{\beta+1} \paren{\exp(-p^{-1/p})
      + \sum_{i=0}^\infty 2^\beta 2^{i(\beta + 1)} \exp(-p^{-1/p} 2^{ip})}.
  \end{align*}
  The convergence of the last series ensures the existence of a constant $c'$ such that
  \begin{align*}
    \E_{X}\bracket{\eta(X) \exp(-b\eta(X)^p)}
    \leq c' b^{-\frac{\beta+1}{p}}.
  \end{align*}
  Adding everything together ends the proof of this lemma.
  Note that we have the same type of optimality as the one stated in Theorem \ref{rates:thm:low-density}.
\end{proof}

Because we use concentration inequalities for terms that are not necessarily centered, we usually get that~\eqref{rates:eq:concentration} only holds for $t > \epsilon_0$ where, typically $\epsilon_0 = \norm{\E_{{\cal D}_n}g_n(x) - g^*(x)}$, we can bypass this problem by adding in $\ind{t<\epsilon_0}$ in the probability, motivating the study leading to the following lemma.

\begin{lemma}[Handling bias in concentration inequality]
  Under the assumptions of Theorem \ref{rates:thm:low-density}, if the concentration is not given by Assumption \ref{rates:ass:concentration} but given, for an $\epsilon_0 > 0$, by, for all $x\in\supp\rho_{\X}$ and $t > 0$,
  \[
    \Pbb_{{\cal D}_n}(\norm{g_n(x) - g^*(x)} > t) \leq \ind{t<\epsilon_0}.
  \]
  Then the excess of risk is controlled by
  \[
    \E_{{\cal D}_n}{\cal R}(f_n) - {\cal R}(f^*) \leq 2c_{\psi}c_{\alpha} \epsilon_0^{\alpha + 1}.
  \]
\end{lemma}
\begin{proof}
  We retake the beginning of the proof of Theorem \ref{rates:thm:low-density}, and change its ending with
  \begin{align*}
    \E_{{\cal D}_n}{\cal R}(f_n) - {\cal R}(f^*)
     & \leq 2c_{\psi} \E_{{\cal D}_n, X}\bracket{\ind{\norm{g_n(X) - g(X)} \geq d(g(X), F)}\norm{g_n(X) - g(X)}}
    \\& = 2c_{\psi}\int_{0}^\infty \Pbb_{{\cal D}_n, X}\paren{\ind{\norm{g_n(X) - g(X)} \geq d(g(X), F)}\norm{g_n(X) - g(X)} \geq t} \diff t
    \\& = 2c_{\psi}\int_{0}^\infty \E_{X}\Pbb_{{\cal D}_n}\paren{\norm{g_n(X) - g(X)} \geq \max\brace{t, d(g(X), F)}} \diff t
    \\& \leq 2c_{\psi} \int_{0}^\infty \E_{X} \ind{t<\epsilon_0}\ind{d(g(X), F) < \epsilon_0} \diff t
    = 2c_{\psi} \epsilon_0 \Pbb_{X}\paren{d(g(X), F) <\epsilon_0} \diff t.
  \end{align*}
  This leads to the result after applying the $\alpha$-margin condition.
\end{proof}
  \section{Nearest neighbors}
\subsection{Usual assumptions to derive nearest neighbors convergence rates}
\label{rates:app:nn-assumptions}

Assumption \ref{rates:ass:lipschitz} can be seen as the backbone to control $\norm{g_n^*(x) - g^*(x)}$ in a uniform manner.
This assumption that relates the regularity of $g^*$ with the density of $\rho_{\X}$ has been historically approached in the following manner.
Assume that $g^*$ is $\beta'$-H\"older, that is, for any $x, x' \in \supp\rho_{\X}$
\[
  \norm{g^*(x) - g^*(x')} \leq a_1 d(x, x')^{\beta'}.
\]
Suppose that $\X = \R^d$, that $\rho_{\X}$ is continuous against $\lambda$, the Lebesgue measure, with minimal mass in the sense that there exists a $p_{\min} > 0$ such that $\frac{\diff\rho_{\X}}{\diff\lambda}(\X)$ does not intersect $(0, p_{\min})$, and that $\supp\rho_{\X}$ has regular boundaries in the sense that there exist $a_2, t_0 > 0$ such that for any $x\in\supp\rho_{\X}$ and $t \in (0, t_0)$
\[
  \lambda\paren{{\cal B}(x, t)\cap\supp\rho_{\X}} \geq a_2\lambda\paren{{\cal B}(x, t)}.
\]
For example an orthant satisfies this property with $a_2 = 2^{-d}$ and $t_0 = \infty$, and ${\cal B}(0, 1)$ satisfies this property with $a_2 = \lambda\paren{{\cal B}(0, 1) \cap {\cal B}(1, 1)} / \lambda\paren{{\cal B}(0, 1)}$ and $t_0 = 1$.
In such a setting, we get
\[
  \norm{g^*(x) - g^*(x')} \leq a_1 d(x, x')^{\beta'}
  = a_1 \paren{\frac{\lambda({\cal B}(x, d(x, x')))}{\lambda({\cal B}(0, 1))}}^{\frac{\beta'}{d}}.
\]
When $d(x, x') < t_0$, we have
\[
  \lambda({\cal B}(x, d(x,x'))) \leq a_2^{-1} \lambda\paren{{\cal B}(x, d(x,
    x'))\cap \supp\rho_{\X}}
  \leq a_2^{-1}p_{\min}^{-1} \rho_{\X}({\cal B}(x, d(x,x')))^\beta.
\]
This means that for any $x \in \supp\rho_{\X}$ and $x' \in {\cal B}(x, t_0)$ we have, with $\beta = \frac{\beta'}{d}$ and the constant $a_3= a_1 a_2^{-\beta} p_{\min}^{-\beta}\lambda({\cal B}(0,1))^{-\beta}$
\[
  \norm{g^*(x) - g^*(x')} \leq a_3 \rho_{\X}({\cal B}(x, d(x, x')))^{\beta}.
\]
While, we actually do not need the bound to hold for $d(x, x') > t_0$ in the following proof, to check the veracity of our remark on Assumption \ref{rates:ass:lipschitz}, one can verify that under our assumptions on $\rho_{\X}$, $\supp\rho_{\X}$ is bounded, and therefore $g^*$ is too.
And if $g^*$ is bounded by $c_\phi$, by considering $a_3' = \max\paren{2c_\phi a_2^{-\beta}p_{\min}^{-\beta} t_0^{-\beta'}, a_3}$, this bound holds for any $x, x' \in \supp\rho_{\X}$.

\subsection{Proof of Lemma \ref{rates:lem:nn-concentration}}
\label{rates:proof:nn-concentration}

\paragraph{Control of the variance term.}
For $x\in\rho_{\X}$, the variance term can be written
\[
  \norm{g_n(x) - g_n^*(x)} = \norm{\frac{1}{k}\sum_{i=1}^k \phi(Y_{(i)}) -
    \E\bracket{Y_{(i)}\midvert X_{(i)}}}.
\]
Where the index $(i)$ is such that $X_{(i)}$ is the $i$-th nearest neighbor of $x$ in $(X_i)_{i\leq n}$.
Since, given $(X_i)_{i\leq n}$, the $(Y_i)_{i\neq n}$ are independent, distributed according to $\otimes_{i\leq n}\rho\vert_{X_i}$, we can use a concentration inequality to control it.
We recall Bernstein concentration inequality in such spaces, derived by \citet{Yurinskii1970}, we will use the formulation of Corollary 1 from \citet{Pinelis1986}.

\begin{theorem}[Concentration in Hilbert space \citep{Pinelis1986}]
  \label{rates:thm:bernstein-vector-full}
  Let denote by ${\cal A}$ a Hilbert space and by $(\xi_i)$ a sequence of independent random vectors on ${\cal A}$ such that $\E[\xi_i] = 0$, and that there exists $M, \sigma^2 > 0$ such that for any $m \geq 2$
  \[
    \sum_{i=1}^n \E\bracket{\norm{\xi_i}^m} \leq \frac{1}{2} m!\sigma^2 M^{m-2}.
  \]
  Then for any $t>0$
  \[
    \Pbb(\norm{\sum_{i=1}^n \xi_i} \geq t) \leq
    2\exp\paren{-\frac{t^2}{2\sigma^2 + 2tM}}.
  \]
\end{theorem}
This explains Assumption \ref{rates:ass:moment}, allowing, because there is only $k$ $\xi_i$ active in $\sum_{i=1}^n \alpha_i(x)\xi_i$, to get
\[
  \Pbb_{{\cal D}_n}\paren{\norm{g_n(x) - g_n^*(x)} > t}
  \leq 2\exp\paren{-\frac{kt^2}{2\sigma^2 + 2M t}}.
\]

\paragraph{Control of the bias term.}
Under the Modified Lipschitz condition, Assumption \ref{rates:ass:lipschitz},
\begin{align*}
  \norm{g_n^*(x) - g_n(x)} & = \norm{\sum_{i=1}^n \alpha_i(x) \paren{g_n(x) - g^*(X_i)}}
  \leq \sum_{i=1}^n \alpha_i(x) \norm{g_n(x) - g^*(X_i)}
  \\&\leq c_{\beta} \sum_{i=1}^n \alpha_i(x) \rho_{\X}\paren{{\cal B}(x, d(x, X_i))}^\beta
  \leq c_{\beta} \rho_{\X}\paren{{\cal B}(x, d(x, X_{k}(x)))}^\beta.
\end{align*}
When $\rho_{\X}$ is continuous, it follows from the probability integral transform (also known as universality of the uniform) that $\rho_{\X}\paren{{\cal B}(x, d(x, X_{k}(x)))}$ behaves like the $k$-th order statistics of a sample $(U_i)_{i\leq n}$ of $n$ uniform distributions on $[0,1]$.
Therefore, for any $s\in[0,1]$
\begin{align*}
  \Pbb_{{\cal D}_n}\paren{\rho_{\X}\paren{{\cal B}(x, d(x, X_{k}(x)))} > s}
   & = \Pbb\paren{\sum_{i=1}^n \ind{U_i < s} \leq k}.
\end{align*}
Recall the multiplicative Chernoff bound, stating that for $(Z_i)_{i\leq n}$ $n$ independent random variables in $\brace{0,1}$, if $Z = \sum_{i=1}^n Z_i$, and $\mu = \E[Z]$, for any $\delta > 0$
\[
  \Pbb\paren{Z \leq (1-\delta)\mu} \leq \exp\paren{-\frac{\delta^2\mu}{2}}.
\]
Since, for $s \in [0, 1]$, $\E[\ind{U_i < s}] = \Pbb(U_i < s) = s$, we get, when $k \leq ns / 2$
\[
  \Pbb\paren{\sum_{i=1}^n \ind{U_i < s} \leq k} \leq
  \exp\paren{-\frac{\paren{ns - k}^2}{2 ns}}
  \leq \exp\paren{-\frac{ns}{8}}.
\]
With $s = c_{\beta}^{-1}t^{\frac{1}{\beta}}$, we get
\[
  \Pbb_{{\cal D}_n}\paren{\norm{g_n^*(x) - g_n(x)} > t}
  \leq \exp\paren{-\frac{nt^{\frac{1}{\beta}}}{8c_{\beta}}}.
\]
Remark that when $g^*$ is $\beta'$ H\"older, we get the same result with $\rho_{\X}({\cal B}(x, t))$ instead of $t^{\frac{1}{\beta}}$ by considering $\ind{X_i \in {\cal B}(x, t)}$ instead of $\ind{U_i \leq t}$.
Note that there is a way to bound a Binomial distribution with a Gaussian for $t$ smaller than the mean of the binomial distribution, which would lead to a bound that holds for any $t>0$ \citep{Slud1977}.

\subsection{Proof of Theorem \ref{rates:thm:nn-no-density}}
\label{rates:proof:nn-no-density}
Using the proof of Theorem \ref{rates:thm:no-density}, we get
\[
  \E_{{\cal D}_n}{\cal R}(f_n) - {\cal R}(f^*)
  \leq \ell_\infty\Pbb_{{\cal D}_n}\paren{\norm{g(x) - g_n^*(x)} > t_0}.
\]
Because $\norm{g_n(x) - g^*(x)} > t_0$ implies that either $\norm{g_n(x) - g_n^*(x)} > t_0/2$ or $\norm{g_n^*(x) - g^*(x)} > t_0/2$, we get using Lemma \ref{rates:lem:nn-concentration}
\[
  \Pbb_{{\cal D}_n}\paren{\norm{g(x) - g_n^*(x)} > t_0}
  \leq
  2\exp\paren{-\frac{b_1kt_0^2}{4 + 2b_2t_0}}
  + \exp\paren{-2^{-\frac{1}{\beta}}b_3 nt_0^{\frac{1}{\beta}}}
  +\ind{t_0 < \paren{k/2n}^\beta}.
\]
This explains the result of Theorem \ref{rates:thm:nn-no-density}.

\subsection{Proof of Theorem \ref{rates:thm:nn-low-density}}
\label{rates:proof:nn-low-density}

First of all for $t > 0$, and $x \in \supp\rho_{\X}$, because $\norm{g_n(x) - g^*(x)} >t$ implies that either $\norm{g_n(x) - g_n^*(x)} > t / 2$ or $\norm{g_n^*(x) - g^*(x)} > t / 2$, we have the inclusion of events:
\[
  \brace{{\cal D}_n \midvert \norm{g_n(x) - g^*(x)} > t}
  \subset \brace{{\cal D}_n \midvert \norm{g_n(x) - g_n^*(x)} > t/2}
  \cup \brace{{\cal D}_n \midvert \norm{g_n(x) - g^*(x)} > t/2},
\]
which translates in terms of probability as
\begin{align*}
  \Pbb_{{\cal D}_n}\paren{\norm{g_n(x) - g^*(x)} > 2t}
   & \leq \Pbb_{{\cal D}_n}\paren{\norm{g_n(x) - g_n^*(x)} > t}
  +\Pbb_{{\cal D}_n}\paren{ \norm{g_n^*(x) - g^*(x)} > t}.
  \\&\leq 2\exp\paren{-\frac{b_1kt^2}{1 + b_2t}}
  + \exp\paren{-b_3 nt^{\frac{1}{\beta}}}
  + \ind{t < \paren{\frac{k}{2n}}^\beta}.
\end{align*}
Using the refinements of Theorem \ref{rates:thm:low-density} exposed in Appendix \ref{rates:app:ref-low-density}, we get that there exists a constant $c > 0$ that does not depend on $k$ or $n$ such that
\[
  \E_{{\cal D}_n}{\cal R}{f_n} - {\cal R}(f^*)
  \leq
  c \paren{k^{-\frac{\alpha + 1}{2}} + n^{-\beta(\alpha + 1)} +
    (nk^{-1})^{\beta(\alpha + 1)}}.
\]
We optimize this last quantity with respect to $k$ by taking $k = n^{\gamma}$, and choosing $\gamma$ such that $\gamma = 2(1-\gamma)\beta$ leading to $\gamma = 2\beta / (2\beta + 1)$ and to rates in $n$ to the power minus $\beta(\alpha + 1) / (2\beta + 1)$.

\subsection{Numerical experiments}

\begin{figure}[ht]
  \centering
  \includegraphics{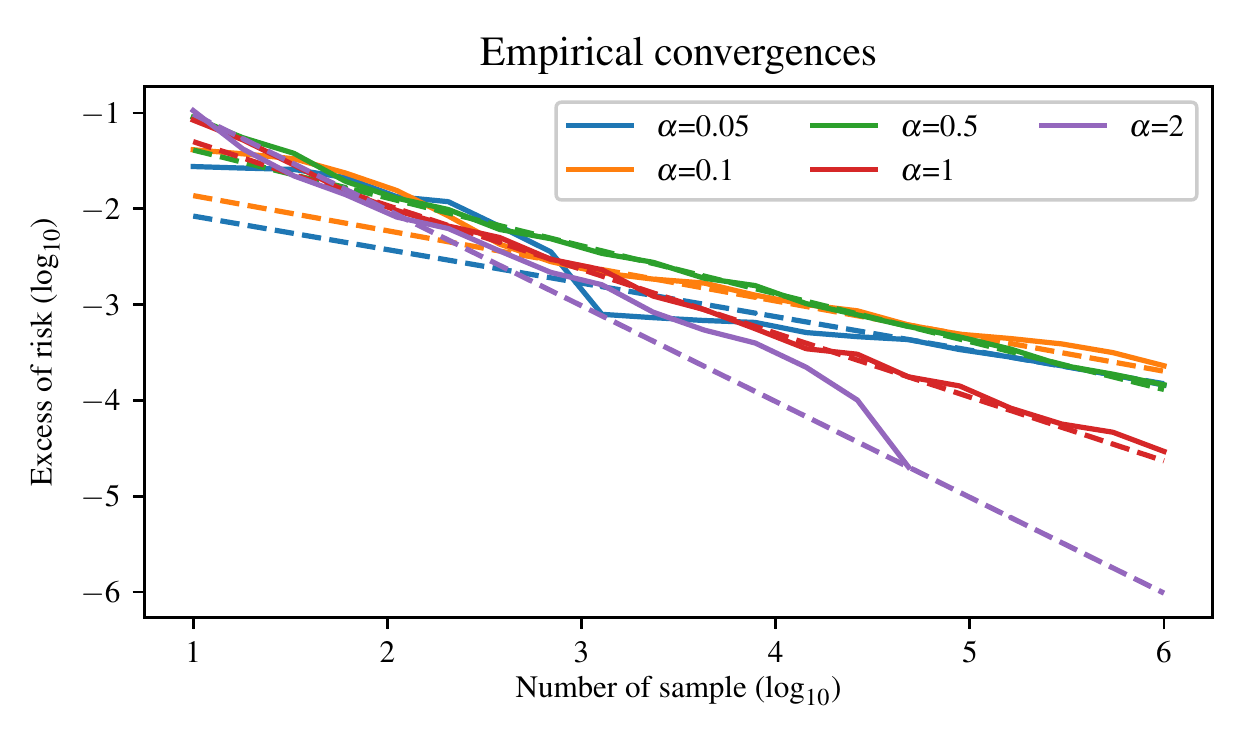}
  \caption{Supplement to Figure \ref{rates:fig:rates}.
  We specify the error is evaluated on 100 points forming a regular partition of $\X = [-1, 1]$, and the expectation $\E_{{\cal D}_n}$ is approximated by considering 100 datasets.
  The violet curve is cropped at $n\approx 10^5$, because the error was null afterwards with our evaluation parameters (only 100 points to evaluate the error), forbidding us to consider the logarithm of the excess of risk.}
  \label{rates:fig:more_rates}
\end{figure}

Interestingly, on numerical simulations such as the one presented on Figure \ref{rates:fig:rates}, we observed two regimes.
A first regime where bound is meaningless because of constants being too big, and where the error decreases independently of the exponent expected through Theorem \ref{rates:thm:nn-low-density}, and a final regime where rates correspond to the bond given by the theorem.
Note that when $\alpha >> 1$, with our computation parameter, we do not get to really illustrate convergence rates, as this final regime get place for bigger $n$ than what we have considered ($n_{\max} = 10^6$), this being partly due to the constant $c_\beta$ in Assumption \ref{rates:ass:lipschitz} being big for the $g^*$ we considered.
Furthermore, note that, for example, if a problem satisfied Assumption \ref{rates:ass:no-density} with a tiny $t_0$, we expect that exponential convergence rates are only going to be observed for $n > N$, with $N$ huge, and for which the excess of risk is already minuscule.
  \section{Kernel proofs}
\label{rates:app:rkhs}

In this section, we study $L^\infty$ convergence rates of the kernel ridge regression estimate.
We use the $L^2$-proof scheme of \citet{Caponnetto2007} with the remark of \citet{Ciliberto2016} to factorize the action of $K$ on $L^2(\X, {\cal H}, \rho_{\X})$ through its action on $L^2(\X, \R, \rho_{\X})$.
We retake the work of \citet{PillaudVivien2018} to relax the source condition, and use \citet{Fischer2020} to cast in $L^\infty$ thanks to interpolation inequality.
While those results, leading to Lemma \ref{rates:lem:rkhs}, are not new, we present them entirely to provide the reader with self-contained materials.

\subsection{Construction of reproducing kernel Hilbert space (RKHS)}
\label{rates:app:rkhs-construction}

In the following, we suppose that $k$ is bounded by $\kappa^2$.

\paragraph{Vector-valued RKHS.}
To study the estimator $g_{n}$, it is useful to introduce the reproducing kernel Hilbert space ${\cal G}$ associated with $k$ and ${\cal H}$ \citep{Aronszajn1950}.
To define ${\cal G}$, define the atoms $k_{x}:{\cal H}\to{\cal G}$ and the scalar product, for $x, x' \in \X$ and $\xi, \xi' \in {\cal H}$ as
\[
  \scap{k_{x}\xi}{k_{x'}\xi'}_{\cal G} = \scap{\xi}{\Gamma(x, x')\xi'} =
  k(x, x')\scap{\xi}{\xi'}_{\cal H}.
\]
Where $\Gamma$ is the vector valued kernel inherited from $k$ as $\Gamma(x, x') = k(x, x') I_{\cal H}$ \citep{Schwartz1964}.
${\cal G}$ is defined as the closure, under the metric induced by this scalar product, of the span of the atoms $k_{x} \xi$ for $x\in\X$ and $\xi\in{\cal H}$.
Note that $k_{x}$ is linear, and continuous of norm $\norm{k_{x}}_{\op} = \sqrt{k(x, x)}$.
When $k(\cdot, x)$ is square integrable for all $x\in\supp\rho_{\X}$, ${\cal G}$ is homomorphic to a functional space in $L^2(\X, {\cal H}, \rho_{\X})$ through the linear mapping $S$ that associates the atom $k_{x}\xi$ in ${\cal G}$ to the function $k(\cdot, x)\xi$ in $L^2$, defined formally as
\[
  \myfunction{S}{\cal G}{L^2}{\gamma}{x\to k_{x}^{\star} \gamma.}
\]
While intrinsically similar, it is useful to distinguish between $\cal G$ and $\ima S \subset L^2$.
Note that $S$ is continuous, since on atom $k_x \xi$, $\norm{Sk_x\xi}_{L^2} \leq \norm{k_x(\cdot)}_{L^2} \norm{\xi}_{\cal H} \leq k(x,x)\norm{\xi}_{\cal H} = \norm{k_x\xi}_{\cal G}$.
The fact that $S$ is a bounded operator justifies the introduction of the following operators.

\paragraph{Central operators.}
In the following, we will make an extensive use of $S^{\star}:L^2(\X, {\cal H}, \rho_{\X})\to{\cal G}$ the adjoint of $S$, defined as $S^{\star} g = \E_{\rho_{\X}}[k_{X}g(X)]$; the covariance operator $\Sigma:{\cal G}\to{\cal G}$, defined as $\Sigma := S^{\star} S = \E_{\rho_{\X}}[k_{X} k_{X}^{\star}]$; and its action on $L^2$, $K:L^2(\X, {\cal H}, \rho_{\X}) \to L^2(\X, {\cal H}, \rho_{\X})$, defined as $Kg := SS^{\star} g = \E_{X}[k(\cdot, X)g(X)]$.
Finally, we have defined the four central operators
\begin{equation}
  \begin{array}{ll}
    S\gamma = k_{(\cdot)}^{\star} \gamma,                              & S^{\star} g = \E_{\rho_{\X}}[k_{X}g(X)]       \\
    \Sigma := S^{\star} S = \E_{\rho_{\X}}[k_{X} k_{X}^{\star}],\qquad & Kg := SS^{\star} g = \E_{X}[k(\cdot, X)g(X)].
  \end{array}
\end{equation}
It should be noted that this construction is usually avoided since, based on the fact that the Frobenius norm of $K$ behave like $\dim({\cal H})$, meaning that when $\cal H$ is infinite dimensional, $K$ is not a compact operator.
However, since we consider $\cal Z$ finite, we can always consider ${\cal H} = \R^{\card{\cal Z}}$ with $\phi(y) = (\ell(z, y)_{z\in\cal Z}$ and $\psi(z) = (\ind{z=z'})_{z'\in{\cal Z}}$, and moreover, we will see that a way can be worked out, even when $\cal H$ is infinite dimensional, which was already shown by \citet{Ciliberto2016}.

\subsubsection*{Relation between real-valued versus vector-valued RKHS.}
Usually convergence in RKHS is studied for real-valued functions.
We need convergence results for vector-valued functions.
As mentioned above, we only need the results for Euclidean space, however, we will do it for functions that are maps going into potentially infinite dimensional Hilbert space.
Indeed, this does not lead to major complications.
We provide here one way to get around this issue.
An alternative formal way to proceed can be found \citep{Ciliberto2016}.

\paragraph{Real-valued RKHS.}
We build the real-valued RKHS ${\cal G}_{\X}$ as the closure of the span of the atoms $\bar{k}_{x}$ for $x\in\X$, under the metric induced by the scalar product $\scap{\bar{k}_{x}}{\bar{k}_{x'}} = k(x, x')$.
Similarly, we build $\bar{S}$, $\bar{S}^{\star}$, $\bar\Sigma$ and $\bar{K}$.
We shall see that the action of $\Sigma$ on ${\cal G}$ can be factorized through its actions $\bar\Sigma$ on ${\cal G}_{\X}$.

\paragraph{Algebraic equivalences.}
Based on the fact that $\norm{\bar{k}_{x}}_{{\cal G}_{\X}} = \norm{k_{x}}_{\op} = \sqrt{k(x, x)}$, it is possible to build an isometry that match $\bar{k}_{x}$ in ${\cal G}_{\X}$ to $k_{x}$ in the space of continuous linear operator from ${\cal H}$ to ${\cal G}$.
With $(\bar{e}_{i})_{i\in \N}$ an orthogonal basis of ${\cal G}_{\X}$, and $(f_{j})_{j\in\N}$ an orthogonal basis of ${\cal H}$, we get an orthogonal basis $(e_{i}f_{j})_{i,j\in\N}$ of ${\cal G}$.
This is exactly the construction ${\cal G} = {\cal G}_{\X} \otimes {\cal H}$ of \citep{Ciliberto2016}.

Note that for $\mu_1, \mu_2$ two measures on $\X$, we can check that
\begin{align*}
  \norm{\E_{X\sim\mu_1}[k_{X}k_{X}^{\star}]\E_{X_0\sim\mu_2}[k_{X_0}]}_{\op}^2
   & = \norm{\E_{X\sim\mu_1}[\bar{k}_{X}\bar{k}_{X}^{\star}]\E_{X_0\sim\mu_2}[\bar{k}_{X_0}]}_{{\cal G}_{\X}}^2
  \\&=\E_{X, X'\sim\mu_1; X, X'\sim\mu_2} [k(X_0, X)k(X, X')k(X', X'_0)].
\end{align*}
This explains that we will allow ourselves to write derivations of the type
\[
  \norm{(\Sigma+\lambda)^{-\frac{1}{2}} k_{x} g_{n}(x)}_{\cal G} \leq
  \norm{(\bar\Sigma+\lambda)^{-\frac{1}{2}} \bar{k}_{x}}_{{\cal G}_{\X}}
  \norm{g_{n}(x)}_{\cal H}.
\]

Note also that for $g := \sum_{ij} c_{ij} e_{i}f_{j} \in {\cal G}$, with $\sum_{ij}c_{ij}^2 = 1$, $\bar{c}_{i} := (c_{ij})_{j\in\N} \in \ell^2$, $\bar{A}$ a self-adjoint operator on ${\cal G}_{\X}$ and $A$ its version on ${\cal G}$, we have
\begin{align*}
  \norm{Ag}^2_{{\cal G}}
   & = \sum_{ijk} c_{ij} c_{kj} \scap{\bar{A}\bar{e}_{i}}{\bar{A}\bar{e}_k}_{\cal G}
  = \sum_{ij} \scap{\bar{c}_{i}}{\bar{c}_{j}}_{\ell^2} \scap{\bar{A}\bar{e}_{i}}{\bar{A}\bar{e}_k}_{{\cal G}_{\X}}
  \\& \leq \sum_{ij}\norm{\bar{c}_{i}}_{\ell^2}\norm{\bar{c}_{j}}_{\ell^2} \scap{\bar{A}\bar{e}_{i}}{\bar{A}\bar{e}_k}_{{\cal G}_{\X}}
  = \norm{\bar{A} \sum_{i} \norm{\bar{c}_{i}}_{\ell^2} \bar{e}_{i}}^2_{{\cal G}_{\X}}
  \leq \norm{\bar{A}}_{\op}^2,
\end{align*}
which explains why we will consider derivations of the type
\[
  \norm{(\Sigma+\lambda)^{\frac{1}{2}}(\hat\Sigma+\lambda)^{-1}(\Sigma+\lambda)^{\frac{1}{2}}}_{\op}
  \leq \norm{(\bar\Sigma+\lambda)^{\frac{1}{2}}(\hat{\bar\Sigma}+\lambda)^{-1}(\bar\Sigma+\lambda)^{\frac{1}{2}}}_{\op}.
\]
Finally, notice that because of the same consideration, if $(\bar{u}_i)_{i\in\N} \in {\cal G}_\X^\N$ diagonalize $\bar{A}$, $(u_if_j)_{i,j\leq \N}\in{\cal G}^{\N\times\N}\simeq{\cal G}^\N$ diagonalize $A$ in ${\cal G}$.
This justifies the consideration of fractional operators $A^p$ for $p \in \R_+$, such as in Assumptions \ref{rates:ass:interpolation} and \ref{rates:ass:source}.
Based on this equivalence, we will forget the bar notations, we incite the careful and attentive reader to recover them.

\subsection{Estimate \texorpdfstring{$g_{n}$}{} as an empirical approximate projection on RKHS}
To obtain bounds like~\eqref{rates:eq:concentration}, it is sufficient to control the convergence of $g_{n}$ to $g^{*}$ in $L^\infty$.
Assumption \ref{rates:ass:interpolation} allow us to cast in $L^2$ the study of the convergence in $L^\infty$.
The convergence of $g_{n}$ toward $g^{*}$ can be split in two terms, a term expressing the convergence of $g_\lambda$ toward $g^{*}$ that is based on geometrical properties and a term expressing the convergence of $g_{n}$ toward $g_\lambda$, that is based on concentration inequalities in ${\cal G}$, such as the ones given by \citet{Pinelis1986,Minsker2017}.
For this last term, we need to characterize $g_{n}$ and $g_\lambda$ with the following lemma.

\begin{lemma}[Approximation of integral operators]
  $g_{n}$ can be understood as the empirical approximation of $g_\lambda$ since
  \[
    g_{n} = S(\E_{\hat\rho}[k_{X} k_{X}^{\star}] + \lambda)^{-1}
    \E_{\hat\rho}[k_{X}\phi(Y)],
    \qquad
    g_\lambda = S(\E_{\rho}[k_{X} k_{X}^{\star}] + \lambda)^{-1}
    \E_{\rho}[k_{X}\phi(Y)],
  \]
  with $\hat\rho = \frac{1}{n} \sum_{i=1}^{n} \delta_{X_{i}} \otimes \delta_{Y_{i}}$,
\end{lemma}
\begin{proof}
  Indeed, the expression of $g_{n}$ and its convergences toward $g^{*}$ will be understood thanks to the operator $S$ and its derivatives.
  When $\ima S$ is closed in $L^2$, on can be defined the orthogonal projection of $g^{*}$ to $\ima S$, with the $L^2$ metric as $\pi_{\ima S}(g^{*}) = S(S^{\star} S)^\dagger S^{\star} g^{*}$.
  When $\ima S$ is not closed, or equivalently when $\Sigma$ has positive eigenvalues converging to zero, one can define approximate orthogonal projection, through eigenvalue thresholding or Tikhonov regularization.
  This last choice leads to the estimate
  \[
    g_\lambda = S(\Sigma + \lambda)^{-1}S^{\star} g^{*} = S (S^{\star} S + \lambda)^{-1} S^{\star} g^{*}
    = SS^{\star} (SS^{\star} +\lambda)^{-1}g^{*} = K (K + \lambda)^{-1}g^{*}.
  \]
  Note that, because of the Bayes optimum characterization of $g^{*}$, $S^{\star} g^{*} = \E_{\rho}[k_{X} \phi(Y)]$.
  This explains the characterization of $g_\lambda$.

  Interestingly, the approximation of $\rho$ by $\hat\rho$ can be thought with the approximation of $L^2(\X, {\cal H}, \rho_{\X})$ by $\ell^2({\cal H}^{n}) \simeq L^2(\X, {\cal H}, \hat\rho_{\X})$ where for $\Xi = (\xi_{i}), Z = (\zeta_{i}) \in {\cal H}^{n}$,
  \[
    \scap{\Xi}{Z}_{\ell^2} = \frac{1}{n}\sum_{i=1}^{n} \scap{\xi_{i}}{\zeta_{i}}_{\cal H},
  \]
  and with the empirical probability measure $\hat\rho = \frac{1}{n}\sum_{i=1}^{n} \delta_{x_{i}}\otimes \delta_{y_{i}}$.
  We redefine the natural homomorphism of $\cal G$ into $\ell^2$ with
  \[
    \myfunction{\hat{S}}{\cal G}{\ell^2}{\gamma}{\paren{k_{x_{i}}^{\star} \gamma}_{i\leq n}.}
  \]
  We check that its adjoint is, for $\Xi\in{\cal H}^{n}$ and $\gamma\in{\cal G}$
  \[
    \scap{\hat{S}^{\star} \Xi}{\gamma}_{\cal G} = \scap{\Xi}{\hat S \gamma}_{\ell^2}
    = \frac{1}{n} \sum_{i=1}^{n}\scap{\xi_{i}}{k_{x_{i}}^{\star}\gamma}_{\cal H}
    = \scap{\frac{1}{n} \sum_{i=1}^{n} k_{x_{i}}\xi_{i}}{\gamma}_{\cal G}.
  \]
  Similarly, we define $\hat K:{\cal H}^{n}\to{\cal H}^{n}$ and $\hat\Sigma:{\cal G}\to{\cal G}$, with
  \[
    \hat K\Xi = \hat S \hat S^{\star} \Xi = \paren{\frac{1}{n} \sum_{i=1}^{n} k(x_{j}, x_{i})
      \xi_{i}}_{j\leq n},\qquad
    \hat\Sigma = \frac{1}{n}\sum_{i=1}^{n} k_{x_{i}} \otimes k_{x_{i}} = \E_{\hat\rho_{\X}}[k_{X}k_{X}^{\star}].
  \]
  Finally, we define $\hat{\Phi} = (\phi(y)_{i})_{i\leq n} \in {\cal H}^{n}$, so that
  \[
    \widehat{S^{\star} g^{*}} := \E_{\hat\rho}[\phi(Y)\cdot k_{X}] = \hat S^{\star} \hat\Phi.
  \]
  Finally, we can express $g_{n}$ as
  \[
    g_{n} = S(\hat\Sigma + \lambda)^{-1}\hat S^{\star} \hat\Phi
    = S(\hat S^{\star} \hat S + \lambda)^{-1}\hat S^{\star} \hat\Phi
    = S\hat S^{\star} (\hat S\hat S^{\star} + \lambda)^{-1} \hat\Phi
    = S\hat S^{\star} (\hat K + \lambda)^{-1} \hat\Phi.
  \]
  This explains the equivalence between $g_{n}$ defined at the beginning of Section \ref{rates:sec:rkhs} and the $g_{n}$ expressed in the lemma, that will be used for derivations of theorems.
\end{proof}

\subsection{Linear algebra and equivalent assumptions to Assumptions \ref{rates:ass:capacity},
  \ref{rates:ass:interpolation}}

To proceed with the study of the convergence of $g_{n}$ toward $g_\lambda$ in $L^2$, it is helpful to pass by ${\cal G}$.
To do so, we need to express Assumptions \ref{rates:ass:capacity} and \ref{rates:ass:interpolation} in ${\cal G}$, which we can do using the following linear algebra property.

\begin{lemma}[Linear algebra on compact operators]
  There exist $(u_{i})_{i\in\N}$ an orthogonal basis of ${\cal G}_{\X}$, $(v_{i})_{i\in\N}$ an orthogonal basis of $L^2(\X, \R, \rho_{\X})$, and $(\lambda_{i})_{i\in \N}$ a decreasing sequence of positive real number such that
  \begin{align}
    S = \sum_{i\in\N} \lambda_{i}^{1/2}u_{i}v_{i}^{\star},\qquad
    S^{\star} = \sum_{i\in\N} \lambda_{i}^{1/2}v_{i}u_{i}^{\star},\qquad
    \Sigma = \sum_{i\in\N} \lambda_{i} u_{i}u_{i}^{\star},\qquad
    K = \sum_{i\in\N} \lambda_{i}v_{i}v_{i}^{\star},
  \end{align}
  where the convergence of series has to be understood with the operator norms.
  Moreover, we have that, if the kernel $k$ is bounded by $\kappa^2$,
  \[
    \sum_{i\in\N} \lambda_{i} \leq \kappa^2 < +\infty.
  \]
  Therefore, $K$ and $\Sigma$ are trace-class, and $S$ and $S^{\star}$ are Hilbert-Schmidt.
\end{lemma}
\begin{proof}
  Notice that $\Sigma = \E_{X}[k_{X}\otimes k_{X}]$ and that $\norm{k_{x}\otimes k_{x}}_{\op({\cal G}_{\X})} = \norm{k_{x}}_{{\cal G}_{\X}} = k(x, x) \leq \kappa^2$.
  Therefore, $\Sigma$ is a nuclear operator, so it is trace class and so it is compact.

  The first point results from diagonalization of kernel operators, known as Mercer's Theorem \citep{Mercer1909,Steinwart2012}.
  $\Sigma$ is a compact operator, therefore, the Spectral Theorem gives the existence of a sequence $(\lambda_{i}) \in \R^{\N}$ and an orthonormal basis $(u_{i}) \in {\cal G}_{\X}^{\N}$ of ${\cal G}_{\X}$ such that
  \[
    \Sigma = \sum_{i\in\N} \lambda_{i} u_{i} u_{i}^{\star},
  \]
  where the convergence has to be understood with the operator norm.
  Because $\Sigma$ is of the form $S^{\star} S$, one can consider $(\lambda_{i})$ a decreasing sequence of positive eigenvalues.
  Then, by defining, for all $i\in\N$ with $\lambda_{i} > 0$,
  \[
    v_{i} = \lambda_{i}^{-1/2} Su_{i}
  \]
  we check that $(v_{i})$ are orthonormal, and we complete them to form an orthonormal basis of $(L^2(\X, \R, \rho_{\X}))$.
  Finally, we check that
  \[
    S = \sum_{i\in\N} \lambda_{i}^{1/2} v_{i}u_{i}^{\star},
  \]
  and that the other equalities hold too.

  To check the second assertion, we use that $k_{x}k_{x}^{\star}$ is rank one when operating on ${\cal G}_{\X}$ and therefore
  \begin{align*}
    \trace{\Sigma} & = \trace\paren{\E_{X}[k_{X}k_{X}^{\star}]} = \E_{X}\bracket{\trace\paren{k_{X}k_{X}^{\star}}}
    = \E_{X}\bracket{\norm{k_{X}k_{X}^{\star}}_{\op({\cal G}_{\X})}}
    \\&= \E_{X}\bracket{\norm{k_{X}}_{{\cal G}_{\X}}} = \E_{X}[k(x, x)] \leq \kappa^2.
  \end{align*}
  This shows that $S$ and $S^{\star}$ are Hilbert-Schmidt operators and that $K$ is also trace class.
\end{proof}

This allows us to cast in ${\cal G}_{\X}$ the assumptions expressed in $L^2$.

\begin{lemma}[Equivalence of capacity condition]
  For $\sigma \in (0, 1]$, it is equivalent to suppose that
  \begin{itemize}
    \item $\trace_{L^2(\X, {\cal H}, \rho_{\X})}(K^\sigma) < +\infty$.
    \item $\trace_{{\cal G}_{\X}}(\Sigma^\sigma) < +\infty$.
    \item $\sum_{i\in\N} \lambda_{i}^\sigma < +\infty$.
  \end{itemize}
\end{lemma}

In Assumption \ref{rates:ass:capacity}, the smaller $\sigma$, the faster the $\lambda_{i}$ decreases, the easier it will be to approximate $\Sigma$ based on approximation of $\rho$.
This appears explicitly in Theorem \ref{rates:thm:matrix}.
Indeed, for $\sigma=0$, the condition should be defined as $\Sigma$ of finite rank.
Note that when $k$ is bounded, we know that $\Sigma$ is trace class, and therefore, Assumption \ref{rates:ass:capacity} holds with $\sigma = 1$.

\begin{lemma}[Interpolation inequality in RKHS]
  \label{rates:lem:interpolation}
  Assumption \ref{rates:ass:interpolation} implies that
  \begin{equation}
    \forall\ \gamma\in{\cal G},\qquad
    \norm{S\gamma}_{L^\infty} \leq c_p \norm{\Sigma^{\frac{1}{2}-p}\gamma}_{\cal G}.
  \end{equation}
\end{lemma}
\begin{proof}
  We begin by showing the property for $\gamma \in {\cal G}_{\X}$.
  When $\gamma = \sum_{i\in\N} c_{i} v_{i}$ with $\sum_{i\in\N} c_{i}^2 < +\infty$, denote $g = \sum_{i\in\N} \lambda_{i}^{\frac{1}{2}-p}c_{i}u_{i}$, we have $g\in L^2$, therefore, using Assumption \ref{rates:ass:interpolation},
  \begin{align*}
    \norm{S\gamma}_{L^\infty}
    = \norm{K^pg}_{L^\infty}
    \leq c_p\norm{g}_{L^2}
    = c_p\norm{\Sigma^{\frac{1}{2}-p}\gamma}_{{\cal G}_{\X}}.
  \end{align*}
  This ends the proof for ${\cal G}_{\X}$.
  Note also that when the result of the Lemma holds, then Assumption \ref{rates:ass:interpolation} holds for any $g\in\ima_{L^2(\X, \R,\rho_{\X})} K^{\frac{1}{2}-p}$.

  Let us switch to ${\cal G}$ now.
  Let $\gamma \in {\cal G}$, and denote $g = S\gamma$.
  Suppose that $g$ achieve it maximum in $x_\infty$, define the direction $\xi = \sfrac{g(x_\infty)}{\norm{g(x_\infty)}_{\cal H}}$, and define $g_{\xi}: x\to \scap{g(x)}{\xi}_{\cal H} \in L^2(\X, \R, \rho_{\X})$, and $\gamma_{\xi} = \sum_{j\in\N} \scap{g_{\xi}}{v_{i}}_{L^2} u_{i} \in {\cal G}_{\X}$.
  We have
  \[
    \norm{S\gamma}_{L^\infty} = \norm{S\gamma_{\xi}}_{L^\infty} \leq
    c_p\norm{\Sigma^{\frac{1}{2}-p}\gamma_{\xi}}_{{\cal G}_{\X}}
    \leq c_p\norm{\Sigma^{\frac{1}{2}-p}\gamma}_{{\cal G}}.
  \]
  When $g$ does not achieve its maximum, one can do a similar reasoning by considering a basis $(f_{i})_{i\in\N}$ of ${\cal H}$ and decomposition $\gamma$ on the basis $(u_{i}f_{j})_{i,j\in\N}$, before summing the directions.
\end{proof}

In Assumption \ref{rates:ass:interpolation}, the bigger $\sfrac{1}{2}-p$ the more we are able to control our problem in ${\cal G}$, the better.
Note that this reformulation of the interpolation inequality allows generalizing it for $p$ smaller than zero.
Note that when $k$ is bounded,
\(
\norm{(S\gamma)(x)}_{\cal H} = \norm{k_{X}^{\star} \gamma}_{\cal H}
\leq \norm{k_{X}}_{\op} \norm{\gamma}_{\cal G} = \sqrt{k(x, x)}\norm{\gamma}_{\cal G},
\)
hence Assumption \ref{rates:ass:interpolation} holds with $p=\sfrac{1}{2}$.

\subsection{Linear algebra with atoms \texorpdfstring{$k_{x}$}{} and useful inequalities}

From the study of the convergence of $g_{n}$ to $g_\lambda$ will emerge two quantities linked to eigenvalues of $\Sigma$ and the position of $k_{x}$ regarding eigenspaces, that are
\begin{equation}\label{rates:eq:eigen-quantity}
  {\cal N}(\lambda) = \trace\paren{(\Sigma+\lambda)^{-1}\Sigma},\qquad
  {\cal N}_\infty(\lambda) = \sup_{x\in\supp\rho_{\X}}\norm{(\Sigma+\lambda)^{-\frac{1}{2}}k_{x}}_{\op}.
\end{equation}
While those quantities could be bounded with brute force consideration, Assumptions \ref{rates:ass:capacity} and \ref{rates:ass:interpolation} will help to control them more subtly.

\begin{proposition}[Characterization of capacity condition]
  The property $\sum_{i\in\N}\lambda_{i}^\sigma < +\infty$, can be rephrased in terms of eigenvalues of $\Sigma$ as the existence of $a_1 > 0$ such that, for all $i > 0$,
  \begin{equation}
    \lambda_{i} \leq a_1 (i+1)^{-\frac{1}{\sigma}}.
  \end{equation}
\end{proposition}
\begin{proof}
  Denote by $u_{i}$ and $S_{n}$ the respective quantities $\lambda_{i}^{\sigma}$ and $\sum_{i=1}^{n} u_{i}$.
  Because $S_{n}$ converge, it is a Cauchy sequence, so there exists $N$ such that for any $p > q > N$, \( S_p - S_{q} = \sum_{i=q+1}^p u_{i} \leq 1.\)
  In particular, considering $p=2q$, and because $(\lambda_{i})$ is decreasing, we have \( q u_{2q} \leq \sum_{i=q+1}^{2q} u_{i} \leq 1.\)
  Therefore, we have that for all $i > 2N$, $u_{i} \leq 3(i+1)^{-1}$, considering $(a_1)^{\sigma} = 3 + \max_{i\leq 2N}\brace{(i+1)u_{i}}$, we get the desired result.
\end{proof}

\begin{proposition}[Characterization of ${\cal N}$]
  When $\trace\paren{K^{\sigma}} < +\infty$, with $a_2 = \int_0^\infty \frac{a_1}{a_1 + t^{\frac{1}{\sigma}}}\diff t$,
  \begin{equation}
    \forall\ \lambda > 0, \qquad {\cal N}(\lambda, r) \leq a_2 \lambda^{-\sigma}.
  \end{equation}
\end{proposition}
\begin{proof}
  Expressed with eigenvalues, we have
  \[
    {\cal N}(\lambda) = \trace\paren{(\Sigma + \lambda)^{-1}\Sigma} =
    \sum_{i\in\N} \frac{\lambda_{i}}{\lambda_{i} + \lambda}.
  \]
  Using that $\lambda_{i} \leq a_1(i+1)^{-\frac{1}{\sigma}}$, that $x\to\frac{x}{x+a}$ is increasing with respect to $x$ for any $a> 0$ and the series-integral comparison, we get for $\sigma \in (0, 1]$
  \begin{align*}
    {\cal N}(\lambda) & \leq \sum_{i\in\N} \frac{a_1(i+1)^{-\frac{1}{\sigma}}}{a_1(i+1)^{-\sigma} + \lambda}
    \leq \int_0^\infty \frac{a_1t^{-\frac{1}{\sigma}}}{a_1t^{-\frac{1}{\sigma}} + \lambda}\diff t
    = \int_0^\infty \frac{a_1}{a_1 + \lambda t^{\frac{1}{\sigma}}}\diff t
    \\&= \lambda^{-\sigma}\int_0^\infty \frac{a_1}{a_1 + (\lambda^\sigma t)^{\frac{1}{\sigma}}}\diff (\lambda^\sigma t)
    = a_2\lambda^{-\sigma},
  \end{align*}
  where we check the convergence of the integral.
\end{proof}

Indeed, Assumption \ref{rates:ass:interpolation} has a profound linear algebra meaning, it is a condition on $\rho_{\X}$-almost all the vector $k_{x} \in {\cal G}_{\X}$ not to be excessively supported on the eigenvector corresponding to small eigenvalue of $\Sigma$.

\begin{proposition}[Characterization of interpolation condition]
  The interpolation Assumption \ref{rates:ass:interpolation} implies that, for all $i\in\N$
  \begin{equation}
    \sup_{x\in\rho_{\X}} \abs{\scap{k_{x}}{u_{i}}_{{\cal G}_{\X}}} \leq c_p\lambda_{i}^{\frac{1}{2}-p}.
  \end{equation}
\end{proposition}
\begin{proof}
  Consider the decomposition of $k_{x} \in {\cal G}_{\X}$ according to the eigenvectors of $\Sigma$, with $a_{i}(x) = \scap{k_{x}}{u_{i}}$.
  The interpolation condition Assumption \ref{rates:ass:interpolation}, expressed in ${\cal G}_{\X}$ with Lemma \ref{rates:lem:interpolation}, leads to for any $\gamma_{\X} \in {\cal G}_{\X}$, and $S\gamma_{\X} :\X\to\R$,
  \begin{align*}
    \abs{(S\gamma_{\X})(x)}
    = \abs{\scap{k_{x}}{\gamma_{\X}}_{{\cal G}_{\X}}}
    \leq \norm{S\gamma_{\X}}_{L^\infty}
    \leq c_p \norm{\Sigma^{\frac{1}{2}-p} \gamma_{\X}}_{{\cal G}_{\X}}
  \end{align*}
  This implies that
  \[
    \scap{k_{x}}{\gamma_{\X}}_{{\cal G}_{\X}}^2 = \paren{\sum_{i\in\N}
      \scap{k_{x}}{u_{i}}\scap{\gamma_{\X}}{u_{i}}}^2 \leq c_p^2
    \norm{\Sigma^{\frac{1}{2}-p} \gamma_{\X}}_{{\cal G}_{\X}}^2 = c_p^2
    \sum_{i\in\N} \lambda_{i}^{1-2p} \scap{\gamma_{\X}}{u_{i}}^2.
  \]
  Taking $\gamma_{\X} = u_{i}$, we get that
  \[
    \abs{\scap{k_{x}}{u_{i}}} \leq c_p\lambda_{i}^{\frac{1}{2}-p}.
  \]
  This result relates the interpolation condition to the fact that $k_{x}$ is not excessively supported on the eigenvectors corresponding to vanishing eigenvalues of $\Sigma$.
\end{proof}

\begin{proposition}[Characterization of ${\cal N}_{\infty}(\lambda, r)$]
  Under the interpolation condition, Assumption \ref{rates:ass:interpolation}, we have with $a_3 = c_p (2p)^{-p} (1-2p)^{\frac{1}{2}-p}$, or $a_3=c_p$ when $p=\sfrac{1}{2}$,
  \begin{equation}
    {\cal N}_{\infty}(\lambda) \leq a_3\lambda^{-p}.
  \end{equation}
\end{proposition}
\begin{proof}
  First of all, notice that
  \begin{align*}
    \norm{(\Sigma + \lambda)^{-\frac{1}{2}}k_{x}}_{{\cal G}_{\X}}
     & = \sup_{\norm{\gamma_{\X}}_{\cal \X} = 1} \scap{\gamma_{\X}}{(\Sigma + \lambda)^{-\frac{1}{2}}k_{x}}_{{\cal G}_{\X}}
    = \sup_{c; \sum_{i\in\N} c_{i}^2 = 1} \sum_{i\in\N} \frac{c_{i} \scap{k_{x}}{u_{i}}}{(\lambda + \lambda_{i})^{\frac{1}{2}}}
    \\& \leq c_p \sup_{c;\sum_{i\in\N c_{i}^2=1}} \sum_{i\in\N} \frac{c_{i}\lambda_{i}^{\frac{1}{2}-p}}{(\lambda + \lambda_{i})^{\frac{1}{2}}}
    \leq \sup_{t\in\R_+} c_p \frac{t^{\frac{1}{2}-p}}{(\lambda + t)^{\frac{1}{2}}}.
  \end{align*}
  When $p\in(0, \sfrac{1}{2})$, this last function is zero in zero and in infinity, therefore its maximum $t_0$ verifies, taking the derivative of its logarithm,
  \[
    \frac{\sfrac{1}{2}-p}{t_0} = \frac{1}{2(t_0+\lambda)} \quad\Rightarrow\quad
    t_0 = \frac{(1-2p)\lambda}{2p} \quad\Rightarrow\quad
    \sup_{t\in\R_+}\frac{t^{\frac{1}{2}-p}}{(\lambda + t)^{\frac{1}{2}}} =
    (2p)^{-p} (1-2p)^{\frac{1}{2}-p}\lambda^{-p}.
  \]
  The cases $p\in\brace{0, 1}$ are easy to treat.
\end{proof}

In the previous analysis, one fact does not appear, it is that $\Sigma$ and $k_{x}$ are linked to one another, since $\Sigma = \E_{X}[k_{X}k_{X}^{\star}]$.
The following remark builds on it to relate $\cal N$ and ${\cal N}_\infty$.

\begin{remark}[Relation between interpolation and capacity condition]
  The capacity and interpolation condition are related by the fact that it is unreasonable not to consider that
  \(
  p \leq \sfrac{\sigma}{2}.
  \)
\end{remark}
\begin{proof}
  Because $k_{x}k_{x}^{\star}$ is of rank one in ${\cal G}_{\X}$, we have
  \begin{align*}
    {\cal N}(\lambda) & = \trace\paren{(\Sigma + \lambda)^{-1}\Sigma}
    = \E_{X}\bracket{\trace\paren{(\Sigma + \lambda)^{-1}k_{X}k_{X}^{\star}}}
    = \E_{X}\bracket{\trace\paren{k_{X}^{\star}(\Sigma + \lambda)^{-1}k_{X}}}
    \\&= \E_{X}\bracket{\norm{k_{X}^{\star}(\Sigma + \lambda)^{-1}k_{X}}_{\op}}
    = \E_{X}\bracket{\norm{(\Sigma + \lambda)^{-\frac{1}{2}}k_{X}}^2_{{\cal G}_{\X}}}.
  \end{align*}
  So indeed, ${\cal N}(\lambda)$ is the expectation of the square $\norm{(\Sigma + \lambda)^{-\frac{1}{2}}k_{X}}_{{\cal G}_{\X}}$, when ${\cal N}_{\infty}(\lambda)$ is the supremum of this last quantity.
  Therefore,
  \[
    {\cal N}(\lambda) \leq {\cal N}_{\infty}(\lambda)^2
  \]
  Supposing that the dependencies in $\lambda$ proved above are tight, we should have \( \sigma \geq 2p, \) which is the statement of this remark.
  We refer the reader to Lemma 6.2 of \citet{Fischer2020} for more consideration to relates $\sigma$ and $p$ (reading $p$ and $\sfrac{\alpha}{2}$ with their notations)
\end{proof}

\subsection{Geometrical control of the residual \texorpdfstring{$\norm{g_\lambda - g^{*}}_{L^\infty}$}{}}
\label{rates:proof:krr-1}

The proof of the first assertion in Lemma \ref{rates:lem:rkhs} follows from, using Assumption \ref{rates:ass:source}, with $g_0 \in K^{-q}g^{*}$,
\begin{align*}
  g_\lambda - g^{*} & = (K(K + \lambda)^{-1} - I)g^{*} = -\lambda (K+\lambda)^{-1} g^{*} = -\lambda (K + \lambda)^{-1}K^{q}g_0
  \\&= -\lambda K^{p}(K + \lambda)^{-1}K^{q-p}g_0.
\end{align*}
Then using Assumption \ref{rates:ass:interpolation},
\begin{align*}
  \norm{g^{*} - g_\lambda}_{\infty} & \leq c_p \lambda\norm{K^{q-p}(K + \lambda)^{-1}}_{\op}\norm{g_0}_{L^2}
  \\&\leq c_p\lambda \norm{K(K + \lambda)^{-1}}^{q-p}_{\op} \norm{(K + \lambda)^{-1}}^{1+p-q}_{\op}\norm{g_0}_{L^2}
  \\&\leq c_p\lambda 1^{q-p} \lambda^{-(1+p-q)}\norm{g_0}_{L^2}
  = b_1\lambda^{q-p},
\end{align*}
where we have used that $\norm{K(K + \lambda)^{-1}}_{\op} = \sfrac{\norm{K}_{\op}}{(\norm{K}_{\op} + \lambda)} \leq 1$ and that $\norm{(K + \lambda)^{-1}} \leq \lambda^{-1}$.

\subsection{Convergence of \texorpdfstring{$\norm{g_{n} - g_\lambda}$}{} through concentration inequality}
\label{rates:proof:krr-2}

For the proof of the second assertion in Lemma \ref{rates:lem:rkhs}, we will put ourselves in ${\cal G}$.
For this, we define in ${\cal G}$
\begin{equation}
  \gamma = \E_\rho[k_{X}\phi(Y)],\qquad
  \gamma_\lambda = (\Sigma + \lambda)^{-1}\gamma,\qquad
  \hat\gamma = \E_{\hat\rho}[k_{X}\phi(Y)],
\end{equation}
so that $g_\lambda = S\gamma_\lambda$, and $g_{n} = S(\hat\Sigma + \lambda)^{-1}\hat\gamma$.

\subsubsection{Decomposition into a matrix and a vector term}
We begin by expressing $g_{n} - g_\lambda$ in ${\cal G}$ with
\begin{align*}
  g_{n} - g_\lambda
   & = S\paren{(\hat\Sigma + \lambda)^{-1} \hat\gamma - (\Sigma + \lambda)^{-1} \gamma}
  \\&= S\paren{(\hat\Sigma + \lambda)^{-1} (\hat\gamma - \gamma) + ((\hat\Sigma + \lambda)^{-1} - (\Sigma + \lambda)^{-1}) \gamma)}
  \\&= S\paren{(\hat\Sigma + \lambda)^{-1} (\hat\gamma - \gamma) + (\hat\Sigma + \lambda)^{-1}(\Sigma - \hat\Sigma)(\Sigma + \lambda)^{-1} \gamma)}
  \\&= S\paren{(\hat\Sigma + \lambda)^{-1} ((\hat\gamma - \hat\Sigma \gamma_\lambda) - (\gamma - \Sigma \gamma_\lambda))},
\end{align*}
where we have used that $A^{-1} - B^{-1} = A^{-1}(B - A)B^{-1}$.
Therefore, using the expression, Lemma \ref{rates:lem:interpolation}, of Assumption \ref{rates:ass:interpolation} in ${\cal G}$, we get
\begin{align*}
  \norm{g_{n} - g_\lambda}_{L^\infty}
   & \leq c_p\norm{\Sigma^{\frac{1}{2}-p} (\hat\Sigma + \lambda)^{-1} (\Sigma + \lambda)^{\frac{1}{2}+p}}_{\op} \times \cdots                         \\
   & \qquad\qquad \norm{(\Sigma + \lambda)^{-(\frac{1}{2}+p)}((\hat\gamma - \hat\Sigma \gamma_\lambda) - (\gamma - \Sigma \gamma_\lambda))}_{\cal G}.
\end{align*}
On the one hand, we have concentration of matrix term toward $\Sigma^{\frac{1}{2}-p} (\Sigma + \lambda)^{-\paren{\frac{1}{2}-p}} \preceq I$.
On the other hand, we have concentration of the vector $\hat\gamma - \hat\Sigma \gamma_\lambda$ toward $\gamma - \Sigma \gamma_\lambda$.
Indeed, the concentration of the matrix term is hard to prove (it is only a conjecture), therefore we will go for another decomposition, that will result in similar rates when $p \geq 0$, that is
\begin{equation}
  \begin{split}
    &\norm{g_{n} - g_\lambda}_{L^\infty}
    \leq c_p\norm{\Sigma^{\frac{1}{2}-p} (\Sigma + \lambda)^{-\frac{1}{2}}}_{\op} {\cal A}(\lambda){\cal B}(\lambda)\\
    &{\cal A}(\lambda) = \norm{(\Sigma + \lambda)^{\frac{1}{2}}(\hat\Sigma + \lambda)^{-1} (\Sigma + \lambda)^{\frac{1}{2}}}_{\op}, \\
    &{\cal B}(\lambda) = \norm{(\Sigma + \lambda)^{-\frac{1}{2}}((\hat\gamma - \hat\Sigma \gamma_\lambda) - (\gamma - \Sigma \gamma_\lambda))}_{\cal G}.
  \end{split}
\end{equation}
Recall the definition of the following important quantity that are going to pop up from the analysis
\begin{equation}\tag{\ref{rates:eq:eigen-quantity}}
  {\cal N}(\lambda) = \trace\paren{(\Sigma+\lambda)^{-1}\Sigma},\qquad
  {\cal N}_\infty(\lambda) = \sup_{x\in\supp\rho_{\X}}\norm{(\Sigma+\lambda)^{-\frac{1}{2}}k_{x}}_{\op}.
\end{equation}

\subsubsection{Extra matrix term}
We control the extra matrix term with
\[
  \norm{\Sigma^{\frac{1}{2}-p} (\Sigma + \lambda)^{-\frac{1}{2}}}_{\op}
  = \norm{\Sigma^{\frac{1}{2}-p} (\Sigma +
  \lambda)^{-\paren{\frac{1}{2}-p}}}_{\op} \norm{(\Sigma + \lambda)^{-p}}_{\op}
  \leq \lambda^{-p}.
\]
Using that $\norm{(\Sigma + \lambda)^{-1}}_{\op} \leq \lambda^{-1}$ and that $\norm{(\Sigma + \lambda)^{-1}\Sigma}_{\op} \leq \sfrac{\norm{\Sigma}_{\op}}{(\norm{\Sigma}_{\op} + \lambda)} \leq 1$.

\subsubsection{Matrix concentration}
Let us make explicit the concentration in the matrix term with
\begin{align*}
  (\Sigma + \lambda)^{\frac{1}{2}} (\hat\Sigma + \lambda)^{-1} (\Sigma + \lambda)^{\frac{1}{2}}
   & = I +
  (\Sigma + \lambda)^{\frac{1}{2}} \paren{(\hat\Sigma + \lambda)^{-1} - (\Sigma + \lambda)^{-1}} (\Sigma + \lambda)^{\frac{1}{2}}
  \\&= I +
  (\Sigma + \lambda)^{\frac{1}{2}} (\hat\Sigma + \lambda)^{-1} \paren{\Sigma - \hat\Sigma} (\Sigma + \lambda)^{-1} (\Sigma + \lambda)^{\frac{1}{2}}.
\end{align*}
From here, notice the following implications (that are actually equivalence)
\begin{align*}
  \Sigma - \hat\Sigma \preceq t (\Sigma + \lambda)
   & \quad\Rightarrow\quad \hat\Sigma + \lambda \succeq (1 - t)(\Sigma + \lambda)
  \\&\quad\Rightarrow\quad (\hat\Sigma + \lambda)^{-1} \preceq (1 - t)^{-1}(\Sigma + \lambda)^{-1}.
  \\&\quad\Rightarrow\quad (\hat\Sigma + \lambda)^{-1} - (\Sigma + \lambda)^{-1} \preceq t(1 - t)^{-1}(\Sigma + \lambda)^{-1}.
  \\&\quad\Rightarrow\quad (\Sigma + \lambda)^{\frac{1}{2}}
  \paren{(\hat\Sigma + \lambda)^{-1} - (\Sigma + \lambda)^{-1}} (\Sigma + \lambda)^{\frac{1}{2}}\preceq t(1-t)^{-1}
  \\&\quad\Rightarrow\quad (\Sigma + \lambda)^{\frac{1}{2}}
  (\hat\Sigma + \lambda)^{-1} (\Sigma + \lambda)^{\frac{1}{2}} \preceq (1-t)^{-1}.
\end{align*}
The probability of the event $\Sigma - \hat\Sigma \preceq t (\Sigma + \lambda)$, can be studied through the probability of the event $(\Sigma + \lambda)^{-\frac{1}{2}}(\Sigma - \hat\Sigma)(\Sigma + \lambda)^{-\frac{1}{2}} \preceq t$, which can be studied through concentration of self adjoint operators.
Finally, we have shown that
\begin{equation}
  \norm{(\Sigma + \lambda)^{-\frac{1}{2}}(\Sigma - \hat\Sigma)(\Sigma + \lambda)^{-\frac{1}{2}}}_{\op} \leq t
  \quad\Rightarrow\quad {\cal A}(\lambda)
  \leq \frac{1}{1-t}.
\end{equation}
The best result that we are aware of, for covariance matrix inequality, is the extension to self-adjoint Hilbert-Schmidt operators provided by \citet{Minsker2017} in Section 3.2 of its concentration inequality on random matrices Theorem 3.1.
It can be formulated as the following.

\begin{theorem}[Concentration of self-adjoint operators \citep{Minsker2017}]
  \label{rates:thm:matrix}
  Let denote by $(\xi_{i})_{i\leq n}$ a sequence of independent self-adjoint operator acting on a separable Hilbert space ${\cal A}$, such that $\ker(\E[\xi_{i}]) = {\cal A}$, that are bounded by a constant $M \in \R$, in the sense $\norm{\xi_{i}}_{\op} \leq M$, with finite variance $\sigma^2 = \norm{\E\sum_{i=1}^{n} \xi_{i}^2}_{\op}$.
  For any $t>0$ such that $6 t^2 \geq (\sigma^2 + \sfrac{Mt}{3})$,
  \[
    \Pbb\paren{\norm{\sum_{i=1}^{n}\xi_{i}}_{\op} > t} \leq 14\
    r\paren{\sum_{i=1}^{n} \E\xi_{i}^2}
    \exp\paren{-\frac{t^2}{2\sigma^2 + 2t M / 3}},
  \]
  with $r(\xi) = \sfrac{\trace{\xi}}{\norm{\xi}_{\op}}$.
\end{theorem}

Let us define $\xi$ that goes from $\X$ to the space of self-adjoint operator action on ${\cal G}_{\X}$ as
\begin{equation}
  \xi(x) = (\Sigma + \lambda)^{-\frac{1}{2}} k_{x}k_{x}^{\star} (\Sigma + \lambda)^{-\frac{1}{2}}.
\end{equation}
We have that $(\Sigma + \lambda)^{-\frac{1}{2}}(\Sigma - \hat\Sigma)(\Sigma + \lambda)^{-\frac{1}{2}} = \E_\rho[\xi(X)] - \frac{1}{n}\sum_{i=1}^{n} \xi(x_{i})$.
To apply operator concentration, we need to bound $\xi$ and its variance.

\paragraph{Bound on $\xi$.}
To bound $\xi$ we proceed with, because $k_{x}k_{x}^{\star}$ is of rank one,
\begin{align*}
  \norm{\xi(x)}_{\op} & = \norm{ (\Sigma + \lambda)^{-\frac{1}{2}} k_{x}k_{x}^{\star} (\Sigma + \lambda)^{-\frac{1}{2}}}_{\op}
  = \trace\paren{ (\Sigma + \lambda)^{-\frac{1}{2}} k_{x}k_{x}^{\star} (\Sigma + \lambda)^{-\frac{1}{2}}}
  \\&= \trace\paren{ k_{x}^{\star} (\Sigma + \lambda)^{-1} k_{x}}
  = \norm{(\Sigma + \lambda)^{-\frac{1}{2}}k_{x}}_{{\cal G}_{\X}}^2
  \leq {\cal N}_{\infty}(\lambda)^2.
\end{align*}

\paragraph{Variance of $\xi$.}
For the variance of $\xi$ we proceed by noticing that
\begin{align*}
  \E\xi(X) & = \E_{X}(\Sigma + \lambda)^{-\frac{1}{2}} k_{X}k_{X}^{\star} (\Sigma + \lambda)^{-\frac{1}{2}}
  = (\Sigma + \lambda)^{-\frac{1}{2}} \E_{X}\bracket{k_{X}k_{X}^{\star}} (\Sigma + \lambda)^{-\frac{1}{2}}
  \\&= (\Sigma + \lambda)^{-\frac{1}{2}} \Sigma (\Sigma + \lambda)^{-\frac{1}{2}}
  = (\Sigma + \lambda)^{-1} \Sigma.
\end{align*}
Hence,
\begin{align*}
  \E\xi(X)^2 \preceq \sup_{x\in\X} \norm{\xi(x)}_{\op} \E[\xi(X)] \preceq {\cal N}_\infty(\lambda)^2(\Sigma + \lambda)^{-1} \Sigma.
\end{align*}
And as a consequence
\begin{equation*}
  \norm{\E\xi(x)^2} \leq {\cal N}_\infty(\lambda)^2,
\end{equation*}
where we have used that $\norm{(\Sigma + \lambda)^{-1} \Sigma}_{\op} = \sfrac{\norm{\Sigma}_{\op}}{(\norm{\Sigma}_{\op} + \lambda)} \leq 1$.

\paragraph{Concentration bound on $\xi$.}
Using the self-adjoint concentration theorem, we get for any $t>0$, such that $6nt^2 \geq {\cal N}_{\infty}(\lambda)^2 (1 + \sfrac{t}{3})$,
\begin{align*}
  \Pbb_{{\cal D}_{n}}\paren{\norm{\E_{\hat\rho}[\xi] - \E_\rho[\xi]}_{\op} > t}
  \leq 14\, \frac{\norm{\Sigma}_{\op} + \lambda}{\norm{\Sigma}_{\op}}
  {\cal N}(\lambda)
  \exp\paren{-\frac{nt^2}{2{\cal N}_{\infty}(\lambda)^2 (1 + t/3)}}.
\end{align*}
Therefore, using the contraposition of the prior implication, we get
\begin{equation}
  \Pbb_{{\cal D}_{n}}\paren{{\cal A}(\lambda) > \frac{1}{1-t}} \leq
  14\, \frac{\norm{\Sigma}_{\op} + \lambda}{\norm{\Sigma}_{\op}}
  {\cal N}(\lambda)
  \exp\paren{-\frac{nt^2}{2{\cal N}_{\infty}(\lambda)^2 (1 + t/3)}}.
\end{equation}

\subsubsection{Decomposition of vector term in a variance and a bias term}
Let switch to the vector term, consider $\xi:\X\times\Y\to{\cal G}$, defined as
\begin{equation*}
  \xi = (\Sigma + \lambda)^{-\frac{1}{2}}k_{x}(\phi(y) - k_{x}^{\star} \gamma_\lambda).
\end{equation*}
It allows expressing in simple form the vector term as
\[
  {\cal B}(\lambda) =
  \norm{\frac{1}{n}\sum_{i=1}^{n} \xi(X_{i}, Y_{i}) - \E_{(X, Y)\sim\rho}[\xi(X, Y)]}.
\]
We can study this term through concentration inequality in ${\cal G}$.
To proceed we will dissociate the variability due to $Y$ to the one due to $X$, recalling that $g_\lambda(x) = k_{x}^{\star}\gamma_\lambda$ and going for the following decomposition
\begin{equation}
  \begin{split}
    &\xi(x, y) = \xi_v(x, y) + \xi_b(x)\\
    &\xi_v(x, y) = (\Sigma + \lambda)^{-\frac{1}{2}}k_{x}(\phi(y) - g^{*}(x)),\\
    &\xi_b(x) = (\Sigma + \lambda)^{-\frac{1}{2}}k_{x}(g^{*}(x) - g_\lambda(x)),
  \end{split}
\end{equation}
which corresponds to the decomposition
\begin{equation}
  \begin{split}
    &{\cal B}(\lambda) \leq {\cal B}_{v}(\lambda) + {\cal B}_{b}(\lambda)\\
    &{\cal B}_v(\lambda) = \norm{\E_{\hat\rho}[\xi_v(X, Y)]-\E_{\rho}[\xi_v(X, Y)]}\\
    &{\cal B}_b(\lambda) = \norm{\E_{\hat\rho}[\xi_b(X, Y)]-\E_{\rho}[\xi_b(X, Y)]}.
  \end{split}
\end{equation}
The first term is due to the error because of having observed $\phi(y)$ rather than $g^{*}(x)$, often called ``variance'', and the second term is due to the aiming for $g_\lambda$ instead of $g^{*}$ often called ``bias''.

\subsubsection{Control of the variance}
To control the variance term, we will use the Bernstein inequality stated Theorem \ref{rates:thm:bernstein-vector-full}.

\paragraph{Bound on the moment of $\xi_v$.}
First of all notice that
\[
  \norm{\xi_v(x, y)}_{\cal G} \leq \norm{(\Sigma+\lambda)^{-\frac{1}{2}}k_{x}}_{\op} \norm{\phi(y) -
    g^{*}(x)}_{\cal H}.
\]
Therefore, under Assumption \ref{rates:ass:moment}, for $m \geq 2$:
\begin{align*}
  \E_{(X, Y)\sim \rho}\bracket{\norm{\xi_v(X, Y)}^m}
   & \leq
  \E_{X\sim\rho_{\X}}\bracket{\norm{(\Sigma+\lambda)^{-\frac{1}{2}}k_{x}}_{\op}^m
  \E_{Y\sim\rho\vert_{X}}\bracket{\norm{\phi(y) -
      g^{*}(x)}_{\cal H}^m}}
  \\& \leq \frac{1}{2} m! \sigma^2 M^{m-2}\E_{X\sim\rho_{\X}}\bracket{\norm{(\Sigma+\lambda)^{-\frac{1}{2}}k_{x}}_{\op}^m}.
\end{align*}
We bound the last term with
\begin{align*}
  \E_{X\sim\rho_{\X}}\bracket{\norm{(\Sigma+\lambda)^{-\frac{1}{2}}k_{x}}_{\op}^m}
   & \leq \sup_{x\in\supp\rho_{\X}} \norm{(\Sigma+\lambda)^{-\frac{1}{2}}k_{x}}_{\op}^{m-2}
  \E_{X\sim\rho_{\X}}\bracket{\norm{(\Sigma+\lambda)^{-\frac{1}{2}}k_{x}}_{\op}^2}
  \\&= {\cal N}_{\infty}(\lambda)^{(m-2)} {\cal N}(\lambda).
\end{align*}

\paragraph{Concentration on $\xi_v$.}
Applying Theorem \ref{rates:thm:bernstein-vector-full}, we get, for any $t > 0$, that
\begin{equation}
  \Pbb\paren{{\cal B}_v(\lambda) > t}
  \leq 2\exp\paren{-\frac{nt^2}{2\sigma^2{\cal N}(\lambda) + 2 M{\cal N}_\infty(\lambda)t}}.
\end{equation}

\subsubsection{Control of the bias}

To control the bias, we recall a simpler version of Bernstein concentration inequality, that is a corollary of Theorem \ref{rates:thm:bernstein-vector-full}.

\begin{theorem}[Concentration in Hilbert space \citep{Pinelis1986}]
  \label{rates:thm:bernstein-vector}
  Let denote by ${\cal A}$ a Hilbert space and by $(\xi_{i})$ a sequence of independent random vectors on ${\cal A}$ such that $\E[\xi_{i}] = 0$, that are bounded by a constant $M$, with finite variance $\sigma^2 = \E[\sum_{i=1}^{n}\norm{\xi_{i}}^2]$.
  For any $t>0$,
  \[
    \Pbb(\norm{\sum_{i=1}^{n} \xi_{i}} \geq t) \leq 2\exp\paren{-\frac{t^2}{2\sigma^2 +
        2tM / 3}}.
  \]
\end{theorem}

\paragraph{Bound on $\xi_b$.}
We have
\[
  \norm{\xi_b(x)}_{\cal G} \leq \sup_{x\in\supp\rho_{\X}} \norm{(\Sigma+\lambda)^{-\frac{1}{2}}k_{x}}_{\op}
  \norm{g_\lambda(x) - g^{*}(x)}_{\cal H}
  \leq {\cal N}_{\infty}(\lambda)\norm{g_\lambda - g^{*}}_{\infty}.
\]
Therefore, with Appendix \ref{rates:proof:krr-1}, we get
\[
  \norm{\xi_b(x)}_{\cal G} \leq b_1 \lambda^{q-p} {\cal N}_{\infty}(\lambda).
\]

\paragraph{Variance of $\xi_b$.}
For the variance we proceed with
\[
  \norm{\xi_b(x)}_{\cal G}^2 \leq {\cal N}_{\infty}(\lambda)^2
  \norm{g_\lambda(x) - g^{*}(x)}_{\cal H}^2.
\]
Therefore,
\[
  \E[\norm{\xi_b(X)}^2] \leq {\cal N}_{\infty}(\lambda)^2 \norm{g_\lambda -
    g^{*}}_{L^2}^2.
\]
Using the derivations made in Appendix \ref{rates:proof:krr-1}, we have, using that $q \leq 1$,
\begin{align*}
  \norm{g_\lambda - g^{*}}_{L^2} & = \lambda \norm{(K+\lambda)^{-1}K^q g_0}_{L^2}
  \leq \lambda \norm{(K+\lambda)^{-(1-q)}}_{\op}
  \norm{(K+\lambda)^{-q}K^q}_{\op}\norm{g_0}_{L^2}
  \\&\leq \lambda^q \norm{g_0}_{L^2}.
\end{align*}

\paragraph{Concentration on $\xi_b$.}
Adding everything together, we get
\begin{equation}
  \Pbb\paren{{\cal B}_b(\lambda) > t}
  \leq 2\exp\paren{-\frac{nt^2}{2
  \paren{\lambda^{2q} {\cal N}_{\infty}(\lambda)^{2}\norm{g_0}^2_{L^2} + b_1\lambda^{q-p}{\cal N}_{\infty}(\lambda)t / 3}}}.
\end{equation}
Note that based on the bound on the variance, we would like ${\cal N}_{\infty}(\lambda)^2\lambda^{2q} \approx\lambda^{2(q-p)}$ to be smaller than ${\cal N}(\lambda) \approx \lambda^{-\sigma}$.
It is the case since $q > p$.

\subsubsection{Union bound}
To control $\norm{g_{n} - g_\lambda}_{L^\infty} \leq c_p\lambda^{-p} {\cal A}(\lambda) ({\cal B}_v(\lambda) + {\cal B}_b(\lambda))$, we need to perform a union bound on the control of ${\cal A}$ and the control of ${\cal B} := {\cal B}_v + {\cal B}_b$, we use that for any $t > 0$ and $0<s<1$, $c_p \lambda^{-p} {\cal A}{\cal B} > t$ implies ${\cal A} > \sfrac{1}{(1-s)}$ or ${\cal B} > \sfrac{(1-s)t \lambda^p}{c_p}$.
Similarly, ${\cal B}_v + {\cal B}_b > t$, implies that either ${\cal B}_v > \sfrac{t}{2}$, either ${\cal B}_b > \sfrac{t}{2}$.
Therefore, we have, the following inclusion of events (with respect to ${\cal D}_{n}$)
\[
  \brace{\norm{g_{n} - g_\lambda}_{L^\infty} > t} \subset \brace{{\cal A} >
    \frac{1}{1-s}}
  \cup \brace{{\cal B}_v > \frac{(1-s)t\lambda^{p}}{2c_p}}
  \cup \brace{{\cal B}_b > \frac{(1-s)t\lambda^{p}}{2c_p}}.
\]
In terms of probability this leads to
\begin{equation}
  \Pbb_{{\cal D}_{n}}\paren{\norm{g_{n} - g_\lambda}_{L^\infty} > t} \leq
  \Pbb_{{\cal D}_{n}}\paren{{\cal A} > \frac{1}{1-s}}
  + \Pbb_{{\cal D}_{n}}\paren{{\cal B} > \frac{(1-s)t\lambda^p}{c_p}}.
\end{equation}
Looking closer it is the term in ${\cal B}$ that will be the more problematic, therefore we would like $s$ to be small.
If we take $s$ to be a constant with respect to $t$, we will get something that behaves like $\Pbb(B > t\lambda^p)$, which is the best we can hope for (this also explain why we divide ${\cal B} > t$ in ${\cal B}_v > \sfrac{t}{2}$ or ${\cal B}_b > \sfrac{t}{2}$).
We will consider $s=\sfrac{1}{2}$.
We express concentration based on the expression of ${\cal N}$ and ${\cal N}_{\infty}$, assuming $\lambda \leq \norm{\Sigma}_{\op}$, and $n > a_3^2\lambda^{-2p}$
\[
  \Pbb_{{\cal D}_{n}}({\cal A} > 2) \leq 28 a_2 \lambda^{-\sigma}\exp(-
  \frac{n\lambda^{2p}}{10 a_3^2}).
\]
Similarly, we get, when $\lambda \leq 1$, using that $\lambda^{-\sigma} \geq 1$
\[
  \Pbb_{{\cal D}_{n}}({\cal B}_v > \sfrac{t}{4}) \leq
  2 \exp\paren{-\frac{n\lambda^\sigma t^2}{32\sigma^2 a_2 + 8Ma_3\lambda^{-p}t}}.
\]
For the bias term, we can proceed at a brutal bounding, based on the fact that for $\lambda \leq 1$, $\lambda^{q-p} \leq 1 \leq \lambda^{-\sigma}$, to get
\[
  \Pbb_{{\cal D}_{n}}({\cal B}_b > \sfrac{t}{4}) \leq
  2 \exp\paren{-\frac{n\lambda^\sigma t^2}{32 a_3^2\norm{g_0}_{L^2} +
      8b_1a_3\lambda^{-p}t/3}}.
\]
With $b_4 = \max(32\sigma^2 a_2, 32 a_3^2\norm{g_0}_{L^2})$ and $b_5 = \max(8Ma_3, 8b_1a_3/3)$, we get the following union bound
\[
  \Pbb_{{\cal D}_{n}}\paren{{\cal B} > \frac{t\lambda^p}{2}}
  \leq 4 \exp\paren{-\frac{n\lambda^{2p+\sigma}t^2}{b_4 + b_5t}}.
\]
We proceed with the union bound on $\norm{g_{n} - g_\lambda}_{L^\infty}$ as
\[
  \Pbb_{{\cal D}_{n}}(\norm{g_{n} - g_\lambda}_{L^\infty} > t)
  \leq b_2 \lambda^{-\sigma}\exp(-b_3n\lambda^{2p})
  + 4 \exp\paren{-\frac{n\lambda^{2p+\sigma}t^2}{b_4 + b_5t}},
\]
with $b_2 = 28a_2$ and $b_3^{-1} = 10a_3^2$, as long as $b_3n > \lambda^{-2p}$, and $\lambda \leq \max(1, \norm{K}_{\op})$.

\subsubsection{Refinement of Lemma \ref{rates:lem:rkhs}}

Remark that the uniform control in Lemma \ref{rates:lem:rkhs} is more than we need, we only need control for each $x$ as described in Assumption \ref{rates:ass:concentration}.
Indeed, if $p(x)$ is such that there exists a constant $\tilde{c_p}$ (that does not depend on $x$ or $\lambda$), such that for any $i\in\N$
\[
  \scap{k_{x}}{u_{i}}_{{\cal G}_{\X}} \leq \tilde{c_p}\lambda_{i}^{p(x)},
\]
then considering that
\[
  g_{n}(x) - g_\lambda(x) = k_{x}^{\star}(\gamma_{n} - \gamma_\lambda)
  = k_{x}^{\star} \paren{\Sigma+\lambda}^{-\frac{1}{2}} \paren{\Sigma+\lambda}^{\frac{1}{2}}\paren{\gamma_{n} - \gamma_\lambda},
\]
we can improve the results of Lemma \ref{rates:lem:rkhs} by replacing $p$ by $p(x)$.
While we considered $p = {\sup_{x\in\rho_{\X}} p(x)}$ as a consequence of our proof scheme, one can expect to end up with the $\E_{X}[\lambda^{p(X)}]$ instead of $\lambda^p$ when deriving the proof of Theorems \ref{rates:thm:krr-no-density} and \ref{rates:thm:krr-low-density} (for which one has to refine Theorem \ref{rates:thm:low-density} in order to integrate dependency of $L$ to $x$, similarly to what is done in Lemma \ref{rates:lem:ref-no-density}), which will lead to better rates.
Yet, because of the complexity of expressing a quantity of the type $\E_{X}[\phi(p(X))]$, for some function $\phi$, we decided not to present this improved version in the paper.

\subsection{Proof of Theorem \ref{rates:thm:krr-no-density}}
\label{rates:proof:krr-no-density}

Based on the proof of Theorem \ref{rates:thm:no-density}, we know that
\[
  \E_{{\cal D}_{n}} {\cal R}(f_{n}) - {\cal R}(f^{*}) \leq \ell_\infty
  \Pbb_{{\cal D}_{n}}\paren{\norm{g_{n} - g^{*}}_{\infty} > t_0}.
\]
Now we use that
\[
  \Pbb_{{\cal D}_{n}}\paren{\norm{g_{n} - g^{*}}_{\infty} > t_0}
  \leq \Pbb_{{\cal D}_{n}}\paren{\norm{g_{n} - g_\lambda}_{\infty} > t_0 - \norm{g_\lambda - g^{*}}_{\infty}}.
\]
The result follows from derivations in Appendix \ref{rates:proof:krr-2}, where we used that when $k$ is bounded, Assumptions \ref{rates:ass:capacity} and \ref{rates:ass:interpolation} are verified with $\sigma=1$ and $p=\sfrac{1}{2}$.
Note that we do not need the source assumption, since we can bound directly $\norm{g_\lambda - g^{*}}_{L^2} \leq \norm{g_\lambda - g^{*}}_{L^\infty} < t_0$ while retaking the proof in Appendix \ref{rates:proof:krr-2}.
Moreover, the results of this last proof holds under the condition $n \lambda b_3 > 1$, but, since $\E_{{\cal D}_{n}} {\cal R}(f_{n}) - {\cal R}(f^{*}) \leq \ell_\infty$, we can augment the constant $b_6$ so that the result in Theorem \ref{rates:thm:krr-no-density} still holds for any $n \in \N^{*}$.

\subsection{Proof of Theorem \ref{rates:thm:krr-low-density}}
\label{rates:proof:krr-low-density}

We can rephrase Lemma \ref{rates:lem:rkhs}, using a union bound
\begin{align*}
  \Pbb_{{\cal D}_{n}}(\norm{g_{n} - g^{*}} > t)
   & \leq \Pbb_{{\cal D}_{n}}(\norm{g_{n} - g_\lambda} > \sfrac{t}{2})
  + \Pbb_{{\cal D}_{n}}(\norm{g_\lambda - g^{*}} > \sfrac{t}{2})
  \\& \leq b_2\lambda^{-\sigma}\exp\paren{-b_3n\lambda^{2p}}
  + 4\exp\paren{-\frac{n\lambda^{2p+\sigma}t^2}{4b_4 + 2b_5t}}
  + \ind{t\leq 2\lambda^{q-p}}.
\end{align*}
Using variant of Theorem \ref{rates:thm:low-density} presented in Appendix \ref{rates:app:ref-low-density}, we get
\begin{align*}
  {\cal R}(f_{n}) - {\cal R}(f^{*})
   & \leq \ell_\infty b_2 \lambda^{-\sigma} \exp\paren{-b_3n\lambda^{2p}}
  + 2c_{\psi}c_{\alpha} 2^{\alpha + 1} \lambda^{(q-p)(\alpha+1)}
  \\&\qquad\qquad+ 2c_{\psi}c_{\alpha} c\paren{b_4^{\frac{\alpha+1}{2}} (n\lambda^{2p+\sigma})^{-\frac{\alpha+1}{2}} + b_5^{\alpha+1}(n\lambda^{2p+\sigma})^{-(\alpha + 1)}}.
\end{align*}
As long as $\lambda \leq \max(\norm{K}_{\op}, 1)$ and $n \geq (b_3\lambda^{2p})^{-1}$.
We optimize those rates with $\lambda = \lambda_0 n^{-\gamma}$, and $\gamma$ satisfying
\[
  2\gamma(q-p) = 1 - \gamma(2p+\sigma) \qquad\Rightarrow\qquad
  \gamma = (2q + \sigma)^{-1}.
\]
This leads to, for $n$ after a certain $N\in\N^{*}$
\begin{align*}
  {\cal R}(f_{n}) - {\cal R}(f^{*})
   & \leq \ell_\infty b_2 \lambda_0^{-\sigma} n^{\frac{\sigma}{2q+\sigma}} \exp\paren{-b_3n\lambda_0^{2p} n^{\frac{2(q-p) + \sigma}{2q+\sigma}}}
  \\&\qquad + 2c_{\psi}c_{\alpha} 2^{\alpha + 1} \lambda_0^{(q-p)(\alpha+1)} n^{-\frac{(q-p)(\alpha+1)}{2q+\sigma}}
  \\&\qquad + 2c_{\psi}c_{\alpha} c\paren{b_4^{\frac{\alpha+1}{2}} \lambda_0^{\frac{(2p+\sigma)\alpha+1}{2}} n^{-\frac{(q-p)(\alpha+1)}{2q+\sigma}} + b_5^{\alpha+1} \lambda_0^{(2p+\sigma)\alpha+1} n^{-\frac{2(q-p)(\alpha+1)}{2q+\sigma}}}
  \\&\leq b_8 n^{-\frac{2(q-p)(\alpha+1)}{2q+\sigma}}.
\end{align*}
Since $\ell$ is bounded, ${\cal R}(f_{n}) - {\cal R}(f^{*}) \leq \ell_\infty$, and we can always higher $b_8$, in order to have the inequality for any $n\in\N^{*}$.
\end{subappendices}

\chapter{Exponential Convergence Rates for SVM}
\label{chap:svm}

The following is a reproduction of \cite{Cabannes2022b}.

Classification is often the first problem described in introductory machine learning classes.
Generalization guarantees of classification have historically been offered by Vapnik-Chervonenkis theory.
Yet those guarantees are based on intractable algorithms, which has led to the theory of surrogate methods in classification.
Guarantees offered by surrogate methods are based on calibration inequalities, which have been shown to be highly suboptimal under some margin conditions, failing short to capture exponential convergence phenomena.
Those ``super fast'' rates are becoming well understood for smooth surrogates, but the picture remains blurry for non-smooth losses such as the hinge loss, associated with the renowned support vector machines.
In this paper, we present a simple mechanism to obtain fast convergence rates, and we investigate its usage for SVM.
In particular, we show that SVM can exhibit exponential convergence rates even without assuming the hard Tsybakov margin condition.

\section{Introduction}
To solve a problem with computer calculations, classical computer science consists in handcrafting a set of rules.
In contrast, machine learning is based on the collection of a vast amount of solved instances of this problem, and on the automatic tuning of an algorithm that maps inputs defining the problem to the desired outputs.
Denote by $x$ in a space $\X$ the inputs, by $y\in\Y$ the outputs, and by $f:\X\to\Y$ the input/output mappings.
To learn a mapping $f^*$, it is customary to introduce an explicit metric of error, and search for the function that minimizes it.
Define this metric through a loss $\ell:\Y\times\Y\to\R$ that quantifies how bad a prediction $f(x)$ is when we observe $y$.
If we assume the existence of a distribution over I/O pairs, $\rho\in\prob{\X\times\Y}$, that generates the instances of the problem we mean to solve, we aim to minimize the average loss value
\begin{equation}
	\label{svm:eq:obj}
	{\cal R}(f) = \E_{(X, Y)\sim\rho}\bracket{\ell(f(X), Y)}.
\end{equation}
In practice, this ``risk'' ${\cal R}$ can be evaluated approximately with samples ${\cal D}_n = (X_i, Y_i)_{i\leq n}$, collected by the machine learning scientist and assumed to have been drawn independently accordingly to $\rho$.

We shall focus on the binary classification problem where $\Y = \brace{-1, 1}$, and $\ell$ is the zero-one loss $\ell(y, z) = \ind{y\neq z}$.
In this setting, the risk ${\cal R}(f)$ captures the probability of mistakes of a classifier $f$, and its minimizer is characterized by
\begin{equation}
	f^* = \argmin_{f:\X\to\Y} {\cal R}(f) = \sign \eta,
	\qquad\text{where}\qquad
	\eta(x) = \E\bracket{Y\midvert X=x}.
\end{equation}
Ideally, leveraging the dataset ${\cal D}_n$, we would like to find a mapping $f_{{\cal D}_n}:\X\to\Y$ that is close to be optimal, in the sense that the excess of risk
\(
{\cal E}(f_{{\cal D}_n}) = {\cal R}(f_{{\cal D}_n}) - {\cal R}(f^*)
\)
is as small as it could be.
Since this quantity is actually random, inheriting from the randomness of the samples, we will focus on controlling its average.\footnote{Note that one could also be interested in controlling its tail, but this is out-of-score of the current literature on ``super fast'' rates.}
In particular, we will show that, when our model is well-specified, as the number of samples grows, this average decays actually much faster than what usual statistical learning theory suggests.
We give a brief historical review of related literature before précising our contributions.

\subsection{Statistical learning theory}
The classical approach to minimize \eqref{svm:eq:obj} without the knowledge of $\rho$ but with the sole access to samples ${\cal D}_n \sim \rho^{\otimes n}$ is to restrict the search over functions in a class ${\cal F}\subset\Y^\X$, and look for an empirical risk minimizer
\begin{equation}
	\label{svm:eq:erm}
	f^*_{{\cal D}_n} \in \argmin_{f\in{\cal F}} {\cal R}_{{\cal D}_n},
	\qquad\text{where}\qquad
	{\cal R}_{{\cal D}_n}(f) = \frac{1}{n} \sum_{i=1}^n \ell(f(X_i), Y_i).
\end{equation}
If we denote by $f^*_{\cal F}$ the minimizer of ${\cal R}$ in ${\cal F}$, using the fact that ${\cal R}_{{\cal D}_n}(f^*_{\cal F}) \geq {\cal R}_{{\cal D}_n}(f^*_{{\cal D}_n})$, the excess of risk can be bounded as
\begin{equation}
	\label{svm:eq:app_est}
	{\cal R}(f_{{\cal D}_n}^*) - {\cal R}(f^*)\leq
	\underbrace{2\sup_{f\in{\cal F}} \abs{{\cal R}(f) - {\cal R}_{{\cal D}_n}(f)}}_{\text{estimation error}}
	+ \underbrace{{\cal R}(f^*_{\cal F}) - {\cal R}(f^*)}_{\text{approximation error}}.
\end{equation}
This bound can be seen as highly suboptimal because it bounds the deviation of a random function with the worst deviation in the function class.
However, for any class~${\cal F}$, there exists an ``adversarial'' distribution $\rho$ for which convergence rates (of the excess of risk toward zero as a function of the number of samples $n\in\N$) derived through this bound can not be improved beside lowering some multiplicative constants \citep{Vapnik1995}.
On the one hand, the estimation error can be controlled with general tools to bound the supremum of a random process \citep[\emph{e.g.},][]{Dudley1967}, and will decrease with a decrease in the size of the class ${\cal F}$.
On the other hand, the approximation error depends on assumptions of the problem, and the bigger the size of the class ${\cal F}$, the less restrictive it will be to assume that $f^*$ is not too different from $f_{\cal F}^*$.
Hence, there is a clear trade-off between controlling both errors, which should be balanced in order to optimize a bound on the full excess of risk.

\subsection{Surrogate methods}
In practice, due to the combinatorial nature of discrete-valued functions, finding the empirical risk minimizer \eqref{svm:eq:erm} is often an intractable problem \citep[\emph{e.g.},][]{Hoffgen1992,Arora1997}.
Therefore, people have approached the original problem with other perspectives.
A straightforward approach is given by \emph{plug-in classifiers}, \emph{i.e.}, classifiers of the form $\sign \hat\eta$, for $\hat\eta$ some estimator of $\eta$.
For example, such an estimator can be constructed as
\(
\hat\eta(x) = \sum_{i=1}^n \alpha_i(x) Y_i,
\)
for $\alpha_i(x)$ some weights that specify how much the observation $Y_i$ made at the point $X_i$ should diffuse to the point $x$ \citep[see][for an example]{Friedman1994}.
Another popular approach to solve classification problems is provided by \emph{support vector machines} (SVM), which were introduced from geometric considerations to maximize the margin between the classes $\brace{x\in\X\midvert f^*(x) = y}$ for $y \in\brace{-1, 1}$ \citep{Cortes1995}.

These two approaches can be conjointly understood as introducing a surrogate loss $L:\R\times\Y\to\R$ and looking for a continuous-valued function $g:\X\to\R$ that solves the surrogate problem
\begin{equation}
	\label{svm:eq:decoding}
	f = \sign g, \qquad
	g^*\in \argmin_{g:\X\to\R} {\cal R}_S(g),
	\qquad\text{where}\qquad {\cal R}_S(g) = \E_{(X,Y)\sim\rho}\bracket{L(g(X), Y)},
\end{equation}
where the notation $S$ stands for ``surrogate''.
To an estimate $g:\X\to\R$ of $g^*$ we associate an estimate $f:\X\to\Y$ of $f^*$ through the decoding step $f=\sign g$.
In particular, using the variational characterization of the mean, $\eta$ can be estimated through $L(z, y) = \abs{z - y}^2$.
Regarding SVM, they are related to the \emph{hinge loss} \cite[see, \emph{e.g.},][]{Steinwart2008}
\begin{equation}
	\label{svm:eq:Hinge}
	L(z, y) = \max\paren{0, 1 - zy},
\end{equation}
Surrogate methods benefit from their relative easiness to optimize and the quality of their practical results.
Arguably, they define the current state of the art in classification, softmax regression being particularly popular to train neural networks on classification tasks.

Surrogate methods were studied in depth by \cite{Bartlett2006}, who proposed a generic framework to relate the excess of the original risk to the excess of surrogate risk through an inequality of the type
\begin{equation}
	\label{svm:eq:calibration}
	{\cal R}(f) - {\cal R}(f^*) \leq \psi\paren{{\cal R}_S(g) - {\cal R}_S(g^*)},
\end{equation}
where $f = \sign g$
and $\psi$ is a concave function, uniquely defined from $L$ and verifying $\psi(0) = 0$.
The use of a concave function is motivated by Jensen inequality, allowing to integrate an inequality derived pointwise (conditionally on an input $x$).

\subsection{Exponential convergence rates}
On the one hand, calibration inequalities \eqref{svm:eq:calibration} are appealing, as they allow casting directly rates derived on the surrogate problem to rates on the original problem.
On the other hand, because~$\psi$ has to be concave, rates in $O(n^{-r})$ on the surrogate problem can not be cast as better rates on the original problem, corresponding to the optimal inequalities where $\psi(x) = cx$ for some $c > 0$,
Yet, one can find cases where the sign of $\eta$ can be estimated much faster than $\eta$ itself, even when this sign is estimated with surrogate methods.
In particular, \cite{Mammen1999} \citep[see also][]{Massart2006} introduced the following condition.

\begin{assumption}[Hard margin condition]
	\label{svm:ass:margin}
	The binary classification problem defined through the distribution $\rho$ is said to verify the (Tsybakov) hard margin condition if the conditional mean $\eta$ is bounded away from zero, i.e.,
	\begin{equation}
		\label{svm:eq:margin}
		\exists\,\eta_0 > 0; \qquad \abs{\eta(X)} > \eta_0\qquad\textit{a.s.},
	\end{equation}
	where the notation a.s. stands for almost surely.
	Equivalently, $\abs{\eta}^{-1} \in L^\infty(\rho_\X)$.
\end{assumption}

Indeed, as shown in Appendix \ref{svm:app:proofs}, under Assumption \ref{svm:ass:margin}, leveraging sign equality, we get for any estimate $g_{{\cal D}_n}:\X\to\R$ computed from the dataset ${\cal D}_n$,
\[
	\E_{{\cal D}_n}[{\cal R}(\sign g_{{\cal D}_n})] - {\cal R}(f^*)
	\leq \Pbb_{{\cal D}_n}\paren{\norm{g_{{\cal D}_n} - \eta}_{L^\infty} > \eta_0}.
\]
As a consequence, an exponential concentration inequality on the $L^\infty$ distance between $g_{{\cal D}_n}$ and $\eta$ directly translates to exponential convergence rates on the average excess of risk.
In particular, estimation methods for~$\eta$ based on H\"older classes of functions, such as local polynomials, are known to be well-behaved with respect to the $L^\infty$ norm \citep[see, \emph{e.g.}, the construction of covering number by][]{Kolmogorov1959}.
This was leveraged by \cite{Audibert2007} in a seminal paper that shows how better rates can be achieved on the classification problem under Assumption \ref{svm:ass:margin} and a variety of weaker conditions (described later in Assumption \ref{svm:ass:weak_margin}).

Surprisingly, such an approach has remained somehow less popular than approaches based on calibration inequalities, and we are missing a framework to fully apprehend fast rates phenomena.
Some results were achieved by \cite{Koltchinskii2005}.
Recently, \cite{Cabannes2021b} showed that this result generalizes to any discrete output learning problem, and that approaches that naturally lead to concentration in $L^2$ could be turned into fast rates based on interpolation inequalities that relate the $L^2$ norm with the $L^\infty$ one (notably reusing the work of \cite{Fischer2020} on interpolation spaces).
Exploiting the work of \cite{MarteauFerey2019}, this can be generalized to any self-concordant loss (using self-concordance to reduce the problem to a least-squares problem); and, through the work of \cite{Lin2020}, to any spectral filtering technique (beyond Tikhonov regularization), such as stochastic gradient descent, which was actually shown earlier for binary classification by \cite{PillaudVivien2018b} and \cite{Nitanda2019}.
In the same stream of research, \cite{Vigogna2022} proposed a general framework to study exponential rates for smooth losses in multiclass classification beyond least-squares.

\subsection{Contribution}
The proofs of exponential convergence in the works quoted above are all based on the basic mechanism outlined in \cite{Audibert2007}.
Unfortunately, such a mechanism does not easily extend to the hinge loss.
Does this mean that support vector machines do not exhibit super fast rates, and thus they are inferior to other surrogate methods?
The practice seems to answer negatively.
In this paper, we give a firm theoretical answer to this question.
In particular, we show not only that support vector machines do achieve exponential rates, but also that they can do so even without assuming the hard margin condition.
Our main contribution is to introduce a general framework to prove exponential convergence rates, and show how this framework can be applied to the hinge loss while only considering classical assumptions.

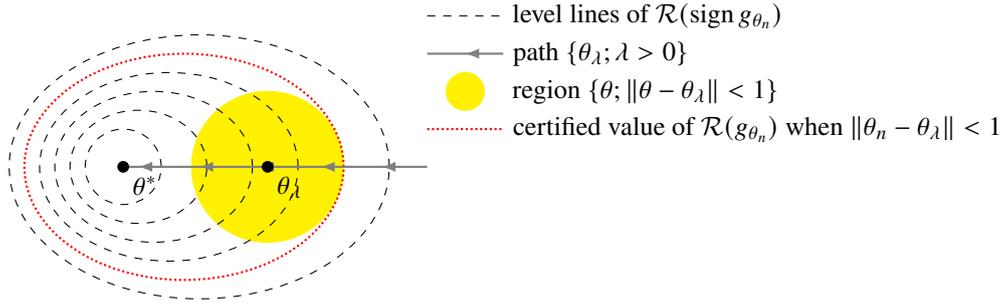
\begin{figure}
	\centering
	\begin{tikzpicture} 
  
  \filldraw[yellow] (1.9, 0) circle (1);

  \draw[dashed] (0, 0) ellipse (.5 and .5);
  \draw[dashed] (.2, 0) ellipse (.9 and .75);
  \draw[dashed] (.4, 0) ellipse (1.3 and 1);
  \draw[dashed] (.6, 0) ellipse (1.7 and 1.25);
  \draw[densely dotted,red,thick] (.8, 0) ellipse (2.1 and 1.5);
  \draw[dashed] (1, 0) ellipse (2.5 and 1.75);

  \begin{scope}[decoration={markings, 
      mark= at position .15 with {\arrow{latex}},
      mark= at position .35 with {\arrow{latex}},
      mark= at position .55 with {\arrow{latex}},
      mark= at position .75 with {\arrow{latex}},
      mark= at position .95 with {\arrow{latex}}}
    ]
    \draw [postaction={decorate},gray,thick] (4,0) -- (0,0);
  \end{scope}

  \filldraw (0, 0) circle (2pt) node[anchor=north west] {$\theta^*$};
  \filldraw (1.9, 0) circle (2pt) node[anchor=north west] {$\theta_\lambda$};

  \draw[dashed] (4, 2) -- (5, 2) node[anchor=west] {level lines of ${\cal R}(\sign g_{\theta_n})$};
  \begin{scope}[decoration={markings, 
      mark= at position .5with {\arrow{latex}}}
    ]
    \draw[gray,thick,postaction={decorate}] (5, 1.5) node[anchor=west,black] {path $\brace{\theta_\lambda; \lambda>0}$} -- (4, 1.5);
  \end{scope}
  \filldraw[yellow] (4.5, 1) circle (.25);
  \draw (5, 1) node[anchor=west,black]{region $\brace{\theta; \|\theta - \theta_\lambda \| < 1}$};
 
  \draw[densely dotted,red,thick] (4, .5) -- (5, .5) node[anchor=west,black] {certified value of ${\cal R}(g_{\theta_n})$ when $\|\theta_n - \theta_\lambda\| < 1$};

\end{tikzpicture}
	\caption{Our convergence analysis consists in relating natural concentration given by surrogate methods to the original excess of risk without passing by the surrogate excess of risk.
		As the drawing shows, concentration in parameter space $\Theta$ can be cast as deviation on the original excess of risk.
		Yet, such a casting relation depends on the geometry of this picture, which itself depends on what surrogate is used, what is the function to learn, how a regularized estimator approached it, and how our empirical estimate concentrates around the regularized estimator.
		Note that this figure illustrates an abstract mechanism that generalizes the simpler mechanism we use to derive exponential convergence rates.
	}
	\label{svm:fig:discrete_analysis}
\end{figure}

\paragraph{Outline.}
Our general strategy is illustrated on Figure \ref{svm:fig:discrete_analysis} and consists in first finding a relation
\[
	{\cal R}_S(g_{\theta}) - {\cal R}_S(g_{\theta^*}) \geq \norm{\theta - \theta^*},
\]
for some natural parameter $\theta$ in a Banach space $\Theta$ parametrizing a class of functions $g_\theta\in{\cal F}$, and then show that $\sign g_\theta = \sign g_{\theta^*}$ when $\norm{\theta - \theta^*}$ is small enough, that is,
\[
	\exists\, \epsilon > 0; \qquad
	\norm{\theta - \theta^*} \leq \epsilon \quad\Rightarrow\quad
	\sign g_\theta = \sign g_{\theta^*}.
\]
Assuming that $\sign g_{\theta^*} = f^*$, we deduce that, as shown in Appendix \ref{svm:app:proofs},
\[
	\E_{{\cal D}_n}[{\cal R}(\sign g_{\theta_n})] - {\cal R}(f^*)
	\leq \Pbb_{{\cal D}_n}\paren{{\cal R}_S(g_{\theta_n}) - {\cal R}_S(g_{\theta^*}) \geq \epsilon},
\]
where $g_{\theta_n}$ is an estimate of $g_\theta$ based on the samples ${\cal D}_n$.
Finally, we conclude with an exponential concentration inequality that controls the deviation of the excess of risk based on classical statistical learning theory.

\section{Exponential convergence of SVM}
This section is devoted to the proof of exponential convergence rates for the hinge loss.
We shall fix the notation ${\cal R}_S$ as the surrogate risk associated with \eqref{svm:eq:Hinge}.
All the proofs are collected in Appendix \ref{svm:app:proofs}.

\subsection{Refined calibration for the hinge loss}

We start by introducing the classical weak margin condition \citep{Mammen1999}.

\begin{assumption}[Weak margin condition]
	\label{svm:ass:weak_margin}
	The binary classification problem defined through the distribution $\rho$ is said to verify the (Tsybakov) $p$-margin condition, with $p \in (0, \infty)$, if there exists a constant $c > 0$ such that
	\begin{equation}
		\Pbb_{\rho_\X}(0 < \abs{\eta(X)} < t) \leq c t^p,
	\end{equation}
	where the notation $\rho_\X$ denotes the marginal of $\rho$ over $\X$.
\end{assumption}

Assumption \ref{svm:ass:weak_margin} is equivalent to asking for the inverse of the conditional mean~$\abs{\eta}^{-1}$ (with the convention $0^{-1} = 0$) to belong to the Lorentz space $L^{p,\infty}(\rho_\X)$ (also known as weak-$L^p$ space), which is the Banach space endowed with the norm (quasi-norm and quasi-Banach if $p < 1$)
\begin{equation}
	\norm{f}_{p, \infty} = \sup_{t > 0} t\Pbb_{\rho_\X}(f(X) > t)^{\frac{1}{p}},
\end{equation}
where the $\rho_\X$ denotes the marginal of $\rho$ with respect to $\X$.
This definition can be extended to the case $p=\infty$ by setting $ L^{p,\infty}(\rho_\X) = L^\infty(\rho_\X) $, which characterizes the hard margin condition in Assumption \ref{svm:ass:margin}.
We will also use $\norm{\cdot}_p$, for $p\in[1,\infty]$, to denote the $L^p$-norm on $\X$ endowed with $\rho_\X$.

We now relate the excess of risk on the hinge loss to the deviation in these spaces.

\begin{lemma}[Weak-$L^q$ concentration due to the hinge loss]
	\label{svm:lem:l1}
	For any functions $g_1, g_2:\X\to[-1,1]$,
	\begin{equation} \label{svm:lem1a}
		{\cal R}_S(g_2) - {\cal R}_S(g_1) = \E_{\rho_\X}[ -\eta(X) (g_2(X) - g_1(X))].
	\end{equation}
	In particular, under Assumption \ref{svm:ass:margin}, for any $g:\X\to\R$,
	\begin{equation} \label{svm:lem1b}
		{\cal R}_S(g) - {\cal R}_S(g^*) \geq \norm{\abs{\eta}^{-1}}_{\infty}^{-1} \norm{\pi(g) - g^*}_{1},
	\end{equation}
	where $g^* = \sign \eta$ is a minimizer of ${\cal R}_S$ and $\pi$ is the projection of $\R$ on $[-1, 1]$, defined as mapping $t\in\R$ to $\pi(t) = \sign(t) \min\{|t|,1\}$.
	Similarly, under Assumption \ref{svm:ass:weak_margin}, with $q = \sfrac{p}{p+1}$,
	\begin{equation} \label{svm:lem1c}
		{\cal R}_S(g) - {\cal R}_S(g^*)
		\geq 2^{-1} \norm{\abs{\eta}^{-1}}_{p,\infty}^{-1} \norm{\pi(g) - g^*}_{q,\infty}.
	\end{equation}
\end{lemma}

Lemma \ref{svm:lem:l1} shows that we can set the minimizer $g^* = f^* \in \brace{-1, 1}^\X$.
This is a useful fact as it implies that the excess of the original risk is zero as soon as $\norm{g - g^*}_\infty < 1$.
In essence, the only piece missing in order to prove fast convergence rates is an interpolation inequality between $L^{q,\infty}$ and $L^\infty$.
In the following, we will leverage Lemma \ref{svm:lem:l1} more subtly by considering a class of functions ${\cal G}$ and assumptions on the distribution $\rho_\X$ such that, if an estimate $g\in{\cal G}$ has not the same sign almost everywhere as the estimand $g^*$, then $\norm{g-g^*}_{q, \infty}$ is bounded away from zero.
By contraposition, if $g\in{\cal G}$ presents a small excess of surrogate risk, then $\sign g = \sign g^*$.
When $\X$ is a metric space, one way to proceed is to assume that $g$ is Lipschitz-continuous, together with some minimal mass assumptions.
Let us begin with the minimal mass assumption.
We first need the following definition.

\begin{definition}[Well-behaved sets]
	A set $U\subset\X$ is said to be well-behaved with respect to $\rho$ if there exist constants $c, r, d > 0$ such that, for any $x\in U$,
	\begin{equation}
		\label{svm:eq:mass}
		\forall\,\epsilon \in [0, r]; \qquad \rho_\X(U\cap{\cal B}(x, \epsilon)) \geq c\epsilon^d,
	\end{equation}
	and ${\cal B}(x, \epsilon)$ the ball in $\X$ of center $x$ and radius $\epsilon$.
\end{definition}

The following examples show that the coefficient~$d$ that appears in \eqref{svm:eq:mass} results from the dimension of the ambient space, the regularity of singularities of the border of the set, and the decay of the density when approaching the frontier of the set.

\begin{example}
	The set $[0, 1]^p$ is well-behaved with coefficients $r=1$, $d=p$ and $c=2^{-d}\operatorname{vol}(\mathbb{S}^{d-1})$ with respect to the Lebesgue measure in $\R^d$.
\end{example}
\begin{example}
	The set $\brace{(x, y) \in \R^2\midvert x\in[0,1], y\in[0, x^{n-1}]}$ is well-behaved with coefficient $r=1$, $d=n$ and $c=n^{-1}$ with respect to the Lebesgue measure.
	Reciprocally, the set $[0,1]$ is well-behaved with coefficient $r=1$, $d=n$ and $c=n^{-1}$ with respect to the measure whose density equals $p(x) = x^{n-1}$.
\end{example}

\begin{assumption}[Minimal mass assumption]
	\label{svm:ass:mass}
	The decision regions $\X_y = \brace{x\in\supp(\rho_\X)\midvert f^*(x) = y}$ for $y \in \brace{-1, 1}$ are well-behaved.
\end{assumption}

Assumption \ref{svm:ass:mass} is a weakening of an assumption that is commonly found in the statistical learning literature.
More precisely, it is often assumed that $\rho$ is absolutely continuous according to the Lebesgue measure $\lambda$ on $\X$ (assumed to be a Euclidean space), that its density is bounded away from zero on its support, and that its support has smooth boundary, so that $\lambda(\supp\rho_\X\cap {\cal B}(x, \epsilon)) > c'\lambda({\cal B}(x, \epsilon))$ \citep[see the strong density assumption in][]{Audibert2007}.

The minimal mass requirement allows relating misclassification events to $L^{q,\infty}$ deviation.

\begin{lemma} \label{svm:lem2}
	Under Assumption \ref{svm:ass:mass}, there exists a constant $c_0$ such that if $g$ is $G$-Lipschitz-continuous for $G > r^{-1}$, for any $q \in (0, 1]$
	\begin{equation}
		\exists\, x\in\supp\rho_\X; \abs{g(x) - g^*(x)} \geq 1 \quad\Rightarrow\quad \norm{g - g^*}_{q, \infty} \geq c_0 G^{-\frac{d}{q}}.
	\end{equation}
\end{lemma}

Putting together Lemmas~\ref{svm:lem:l1} and~\ref{svm:lem2}, we obtain the following refined calibration.

\begin{proposition}
	\label{svm:prop:mid}
	Under Assumptions \ref{svm:ass:weak_margin} and \ref{svm:ass:mass}, if $g$ is $G$-Lipschitz-continuous with $G>r^{-1}$, we have
	\begin{equation}
		\label{svm:eq:mid}
		{\cal R}_S(g) - {\cal R}_S(g^*) \leq 2^{-1}\norm{\abs{\eta}^{-1}}_{p,\infty}^{-1} c_0 G^{-\frac{d(p+1)}{p}} \qquad\Rightarrow\qquad {\cal R}(\sign g) = {\cal R}(f^*).
	\end{equation}
\end{proposition}

\subsection{Trade-off between estimation and approximation errors}
We are now left with the research of $g_{{\cal D}_n}$ inside a class of Lipschitz-continuous functions such that ${\cal R}_S(g_{{\cal D}_n}) - {\cal R}_S(g^*)$ is sub-Gaussian (its randomness being inherited from the dataset ${\cal D}_n$ from which $g_{{\cal D}_n}$ is built).
To do so, let us consider a linear class of functions
\begin{equation}
	\label{svm:eq:f_k}
	{\cal G}_{M, \sigma} = \brace{x\mapsto\scap{\theta}{\phi(\sigma^{-1} x)}\midvert \theta \in {\cal H}, \norm{\theta}_{\cal H} \leq M} ,
\end{equation}
where ${\cal H}$ is a separable Hilbert space, $\phi:\X\to{\cal H}$ is a $G_\phi$-Lipschitz-continuous mapping, and $\sigma > 0$ is a scaling (or bandwidth) parameter.
Such a class of functions can be entirely described from the kernel $k(x, x') = \scap{\phi(x)}{\phi(x')}$ \citep[see][for a primer on kernel methods]{Scholkopf2001}.
An example for ${\cal G}$ is given by the Gaussian kernel, a.k.a. radial basis function, $k(x, x') = \exp(-\norm{x - x'}^2)$.
Using Cauchy-Schwarz, it is easy to show that any function in ${\cal G}_{M,\sigma}$ is $MG_\phi \sigma^{-1}$-Lipschitz-continuous.

In order to find a function $g_{{\cal D}_n}$ that is likely to minimize ${\cal R}_S$ without accessing the distribution $\rho$, but only i.i.d. samples ${\cal D}_n = (X_i, Y_i)_{i\leq n} \sim \rho^{\otimes n}$, it is classical to consider the empirical risk minimizer
\begin{equation}
	\label{svm:eq:sur_erm}
	g_{{\cal D}_n} \in \argmin_{g\in{\cal G}_{M, \sigma}} \frac{1}{n} \sum_{i=1}^n L(g(X_i), Y_i).
\end{equation}
This problem is convex with respect to $\theta$ parametrizing $g \in {\cal G}_{M,\sigma}$, and is easily optimized with duality.
We refer the curious reader to the extensive literature on SVM \citep[see][for books on the matter]{Cristianini2000,Scholkopf2001,Steinwart2008}.

In order to show that ${\cal R}_S(g_{{\cal D}_n})$ is close to ${\cal R}_S(g^*)$, one can apply classical results from statistical learning theory, and in particular \eqref{svm:eq:app_est}.
The estimation error can be bounded using the extensive literature on Rademacher complexity for linear classes of functions on Lipschitz-continuous losses \citep{Bartlett2002}.
To bound the approximation error, one needs to make additional assumptions on the problem.
We refer to \cite{Steinwart2007,Blaschzyk2018} for advanced considerations on the matter.
In view of our calibration result \eqref{svm:eq:mid}, the following additional assumption suffices to prove exponential convergence of SVM.

\begin{assumption}[Source condition]
	\label{svm:ass:source}
	The classification problem verifies the $p$-margin condition \ref{svm:ass:weak_margin}, and there exist $M, \sigma$ and a function $g\in{\cal G}_{M, \sigma}$ such that ${\cal R}_S(g) - {\cal R}_S(g^*) \leq 4^{-1}\|\abs{\eta}^{-1}\|_{p, \infty}^{-1} c_0 M^{-r} G_\phi^{-r} \sigma^r$ with $r = d(p+1) / p$.
\end{assumption}

It should be noted that Assumption \ref{svm:ass:source}, together with Assumption \ref{svm:ass:mass}, implies that the decision frontier $\overline{\X_{-1}} \cap \overline{\X_1}$ (the bar notation corresponding to space closure), inherits from the regularity of $g$, since it is included in the set $\brace{x\in\X\midvert g(x) = 0}$.
In particular, if ${\cal G}_{M,\sigma}$ is included in ${\cal C}^m$, this frontier would be in ${\cal C}^m$.
Hence, for Assumptions \ref{svm:ass:source} and \ref{svm:ass:mass} to hold, the boundary frontier should match the regularity implicitly defined by ${\cal G}_{M, \sigma}$.

We are finally ready to state our main result, establishing exponential convergence rates for SVM.

\begin{theorem}[Exponential convergence rates for SVM]
	\label{svm:thm}
	Under Assumptions \ref{svm:ass:weak_margin}, \ref{svm:ass:mass} and \ref{svm:ass:source}, there exists a constant $c > 0$ such that the empirical minimizer $g_{{\cal D}_n}$ defined by \eqref{svm:eq:sur_erm} verifies
	\begin{equation}
		\E_{{\cal D}_n}{\cal R}(\sign g_{{\cal D}_n}) - {\cal R}(f^*)
		\leq \exp(-cn).
	\end{equation}
\end{theorem}

\begin{figure}[t]
	\centering
	\includegraphics{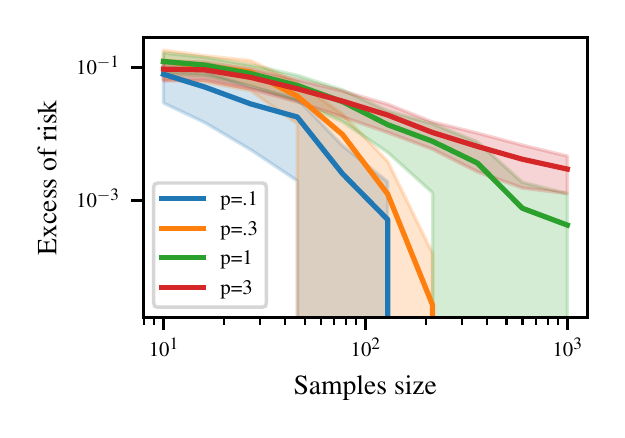}
	\includegraphics{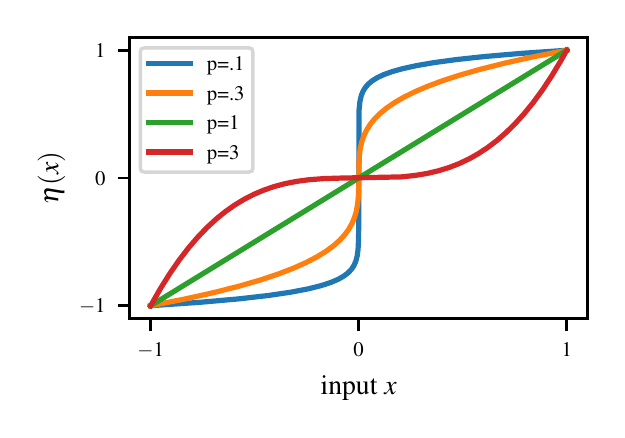}
	\caption{
		SVM generalization error as a function of the number of samples (left) for a problem where $X$ is uniform on $[-1, .-1] \cup [.1, 1]$ and $\eta(x) = \sign(x) \abs{x}^p$ (right). We observe exponential convergence rates.
	}
	\label{svm:fig:fig1}
\end{figure}

\subsection{Relaxing assumptions}

Exponential convergence rates rely on strong assumptions in order to set the approximation error to zero.
In particular, it is customary to assume that the surrogate function to learn lies in the model we have chosen, that is, in our notation, $g^* \in {\cal G}_{M, \sigma}$.
In our case this would be a strong assumption, since $g^*$ is piecewise linear while ${\cal G}_{M, \sigma}$ is a smooth space of functions.
It turns out that the assumption $g^*\in{\cal G}_{M, \sigma}$ is not necessary, and what we actually need is a sufficiently small risk.
How small is enough is quantified by the statement of Proposition \ref{svm:prop:mid}. 
From a qualitative point of view, the requirement of Assumption \ref{svm:ass:source} is natural (surrogate risk must be smaller than $\epsilon$), with the technical part being just a quantification of the needed behavior (bound on such $\epsilon$).

To deepen the study of the approximation error, one could leverage the following geometrical characterization of the risk of misclassification.
For $f:\X\to\brace{-1, 1}$, we have
\begin{equation}
	\label{svm:eq:pre_source}
	{\cal R}(f) - {\cal R}(f^*) = \E[\abs{\eta(X)} \ind{f(x) \neq f^*(X)}] \leq
	\Pbb(f(X)\neq f^*(X)) = \rho_\X\paren{f^{-1}(\{1\})\,\triangle\, \X_1},
\end{equation}
where $\triangle$ denotes the symmetric difference of sets, {\em i.e.} $A \triangle B = (A \cup B) \setminus (A \cap B)$.
In particular, if the function class ${\cal G}_{M,\sigma}$ is rich enough, and the classes $\X_1$ and $\X_{-1}$ are separated, in the sense that the distance between any two points in each set is bounded away from zero,\footnote{This property is sometimes referred to as the \emph{cluster assumption} \citep{Rigollet2007} under which even weakly supervised learning techniques may exhibit exponential convergence rates \citep[\emph{e.g.}][]{Cabannes2021}. In terms of practical applications, this assumption says that no one can continuously modify an input to go from a region of the space linked with one class to a region linked with another class without going through inputs that will never exist.
This is typically true for well-curated image datasets such as CIFAR10: one can not continuously transform an image of a truck into an image of a horse without going through images that will never appear in the CIFAR10 dataset \citep{Krizhevsky2009}.} then Assumption \ref{svm:ass:source} holds, and the minimizer $g_{{\cal G}_{M,\sigma}}$ of the surrogate risk in ${\cal G}_{M,\sigma}$ verifies
\begin{equation}
	\label{svm:eq:source}
	\rho_\X\paren{(\sign g_{{\cal G}_{M,\sigma}})^{-1}(\{1\})\,\triangle\, \X_1} \leq \psi(M, \sigma),
\end{equation}
for $\psi$ a function that vanishes for sufficiently large $M$ and small $\sigma$.

On the one hand, one could control the approximation error by assuming or deriving inequalities akin to \eqref{svm:eq:pre_source} and \eqref{svm:eq:source}, with different profiles of $\psi$.
We conjecture that this can be done by assuming margin conditions that are well adapted to the geometric nature of SVM, such as the one proposed by \cite{Steinwart2007} \citep[see also][]{Gentile1999,Cristianini2000}.
On the other hand, the estimation error can be controlled by extending the ideas presented in this paper to study the worst value of the estimation error ${\cal R}(\sign g_{{\cal D}_n}) - {\cal R}(\sign g_{{\cal G}_{M,\sigma}})$ under the knowledge of ${\cal R}_S(g_{{\cal D}_n}) - {\cal R}_S(g_{{\cal G}_{M,\sigma}})$.
Fitting $M$ and $\sigma$ to trade estimation and approximation error, such derivations would open the way to fast polynomial rates under less restrictive assumptions.

\begin{figure}[t]
	\centering
	\includegraphics{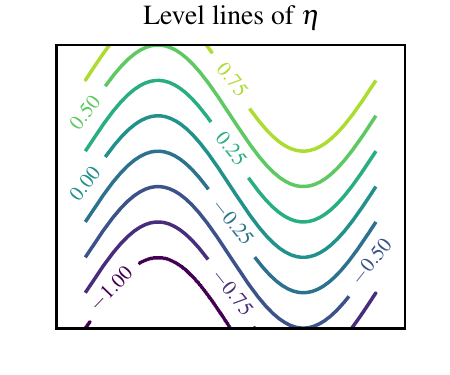}
	\includegraphics{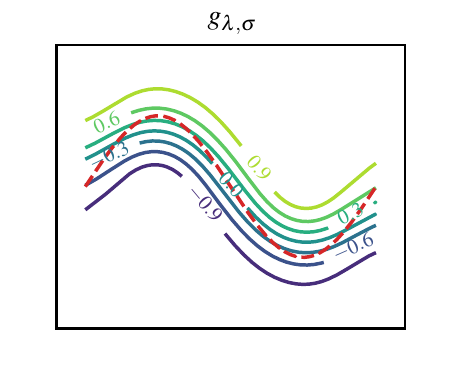}
	\includegraphics{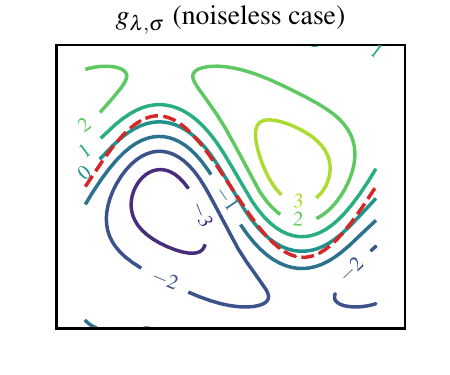}
	\includegraphics{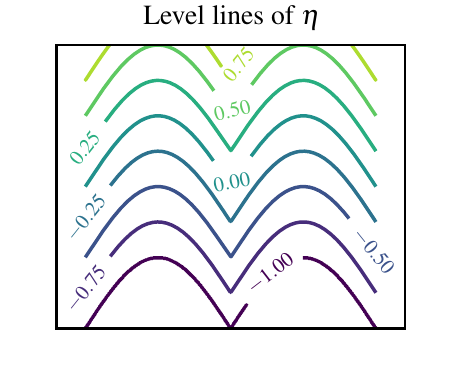}
	\includegraphics{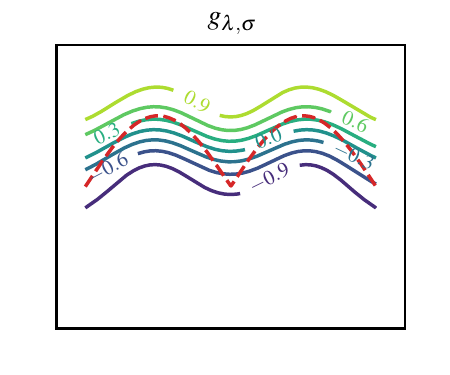}
	\includegraphics{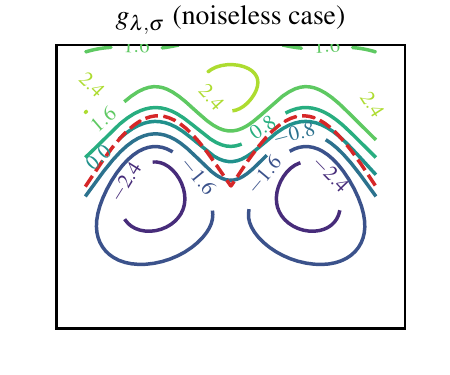}
	\caption{
	Study of the level lines of $g_{\lambda, \sigma}$ when $\eta^{-1}(0) \in {\cal C}^\infty$ (top) and $\eta^{-1}(0)\in{\cal C}^0\setminus {\cal C}^1$ (bottom).
	The function $g^*$ takes values $-1$ below the optimal decision frontier plotted in red and $+1$ above, independently of the noise.
	We observe that the bias error ${\cal R}(\sign g_{\lambda,\sigma}) - {\cal R}(f^*)$, which is bounded by the volume between the level lines $\brace{x\in\X\midvert g_{\lambda,\sigma}(x) = 0}$ and $\brace{x\in\X\midvert \eta(x) = 0}$ (plotted in red), depends on both the regularity of the latter, and on the noise level.
	Here, $\sigma$ is taken to be of the order of 15\% of the diameter of the domain, which explains the regularity of the observed level lines.
	The noiseless cases on the right correspond to the situations where $\E[Y\vert X] = \sign \eta(X)$ for $\eta$ plotted on the left.
	}
	\label{svm:fig:fig2}
\end{figure}

\section{Numerical analysis}

In this section, we provide experiments to illustrate and validate our theoretical findings.
In order to be inline with the current practice of machine learning, instead of considering the hard constraint $\norm{\theta} \leq M$ when minimizing a risk functional, we add a penalty $\lambda \norm{\theta}^2$ to the risk to be minimized.
Going from a constrained to a penalized framework does not change the nature of the statistical analysis, and one might loosely think of $\lambda$ as $1 / M$ \citep[see, for example,][]{Bach2023}.
All experiments are made with the Gaussian kernel.
Precise details of the different settings are provided in Appendix \ref{svm:app:experiments}.

First, we observe that the regime described in this paper kicks in when the error is already pretty small.
On many real-world problems, we do not expect the generalization error as a function of the number of data used for training to exhibit a clear exponential behavior until an unusually big number of samples is used.
This fact is illustrated on Figure~\ref{svm:fig:fig1}, where for hard problems, the exponential behaviors still do not kick in after a thousand of samples.

Second, this paper shows that, in order to get exponential convergence rates for SVM, one needs the minimizer $g_{M,\sigma}$ of the surrogate risk over the selected class of functions to be a perfect classifier, {\em i.e.} its sign equals the sign of $g^*$.
While this is not constraining under the cluster assumption, we inspect divergences from this condition on Figure~\ref{svm:fig:fig2}.
We observe that, even if $g^*$ does not depend on the noise, $g_{M,\sigma}$ does.
We also observe that the regularity of the decision boundary $\brace{x\in \X\midvert \eta(x) = 0}$ should match the regularity defined implicitly by the kernel $k$ and the scale parameter $\sigma$.

\begin{figure}[t]
	\centering
	\includegraphics{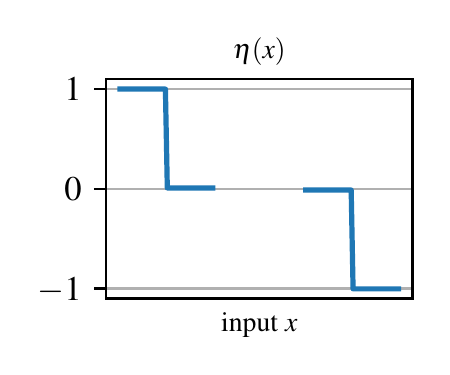}
	\includegraphics{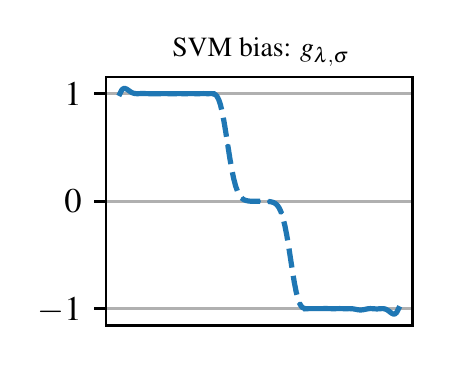}
	\includegraphics{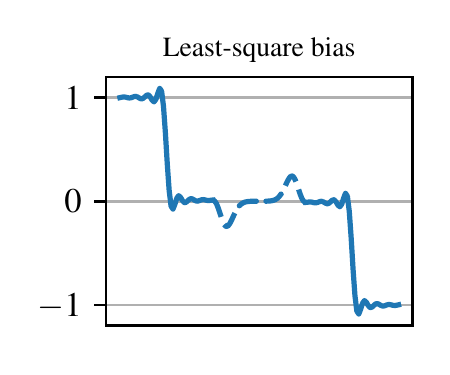}
	\caption{
		Comparison of the regularized risk ({\em i.e.} ${\cal R}_S(g_\theta) + \lambda\norm{\theta}^2$) minimizer for the hinge loss surrogate (middle) and the least-squares surrogate (right), when $\eta$ is not regular (left).
		In this setting, the hinge loss is minimized for $g = \sign(\eta)$, which can be chosen regular, while the least-squares loss is minimized for $g = \eta$, which can not be chosen regular.
		The reconstruction is made with $\sigma$ about 3\% of the domain diameter, and $\lambda$ relatively small.
		We assume no density in the middle of the domain, explaining the absence of definition of $\eta$ and the dashed lines of the right figures.
		The oscillation on the later figures is related to the Gibbs phenomenon \citep{Wilbraham1848}.
		This phenomenon prevents the regularized least-squares solution from being a perfect classifier.
	}
	\label{svm:fig:fig3}
\end{figure}

Experimental comparisons of different classification approaches have been done by many people, and our goal is not to showcase the superiority of the SVM over least-squares, which might be considered as general wisdom that led to the golden age of SVM in the pre-deep-learning area \citep[see][for example]{Joachims1998}.
In comparison with previous works based on calibration inequalities \citep{Rosasco2004,Steinwart2007b}, our analysis proves the robustness of SVM to noise far away from the decision boundary, in the sense that one does not need $\eta$ to be bounded away from zero.
This is a distinctive aspect of SVM compared to smooth surrogate methods \citep{Nowak2019b}, such as softmax regression, that implicitly estimate conditional probabilities and whose performance depends on the regularity of $\eta$.
We illustrate this fact graphically on Figure~\ref{svm:fig:fig3}.

\section{Limitations}

\paragraph{Are surrogate methods only a proxy for classification?}
From a theoretical perspective, if we are only interested in the optimal mapping $f:\X\to\Y$, learning surrogate quantities can be seen as a waste of resources.
In essence, this waste of resources is similar to the one occurring when we learn the full probability function $(p(y))_{y\in\Y}$ for some probability distribution $p$ on $\Y$, while we only care about its mode.
Yet, in practice, what we call a ``surrogate'' problem might actually be a problem of prime interest when we do not only want to predict $f^*(x)$, but we would also like to know how much we can confidently discard other potential outputs for an input $x$.
Furthermore, assuming that a problem is exactly defined through an ``original'' loss that defines a clear and unique measure of error can be questioned when some practitioners evaluate methods with several metrics of performance \cite[\emph{e.g.}][]{Chowdhery2022}.

\paragraph{Do PAC-bounds provide confidence levels?}
Since the parameters in Assumptions \ref{svm:ass:weak_margin} and \ref{svm:ass:mass} are hard to estimate in practice, it would be difficult to directly plug our bounds into a practical problem to derive confidence levels on how much error one might expect when deploying a model in production.
Less ambitiously, we see theorems akin to Theorem \ref{svm:thm} as providing theoretical indications that a learning method or a set of hyperparameters is sound.
This is a generic downfall of probably approximately correct (PAC) generalization bounds \citep{Valient2013}, which might explain why practitioners often prefer to derive error indications from test samples \citep[see, \emph{e.g.},][]{Geron2017}.
Along this line, research on conformal prediction provides interesting considerations to obtain useful confidence information from test samples \citep{Vovk2008}.
Finally, all these statistical methods to get confidence intervals assume representative (if not i.i.d.) data, an assumption sometimes hard to meet in practice, which is a problem that has found echoes in the civil society \citep[\emph{e.g.}][]{Benjamin2019}.

\section{Conclusion}

In this work, we were keen to illustrate a simple mechanism to get exponential convergence rates for a loss that is quite popular and whose understanding can not be easily reduced to existing work.
In particular, we show that the hard margin condition is not crucial in order to derive exponential convergence rates for the SVM.

This provides a crucial step to better understand convergence rates on classification problems.
An extension to generic discrete output problems could be made by considering polyhedral losses, and deriving variants of Lemma~\ref{svm:lem:l1} \citep[see][for calibration inequalities for such losses]{Frongillo2021}.
An important follow-up would be to provide a more global picture of fast polynomial rates for SVM under relaxations of Assumptions \ref{svm:ass:mass} and \ref{svm:ass:source}.

Finally, \cite{Chizat2020} have made a link between two-layer wide neural networks in the interpolation regime (which implies Assumption \ref{svm:ass:margin} with $\eta_0 = 1$) and max-margin classifiers over specific linear classes of functions.
As a consequence, we could directly plug in our analysis to prove exponential convergence rates for those small neural networks in this noiseless setting.
Studying rates, constants and hyperparameter tuning in this setting would be of particular interest if it was to provide practical guidelines to deep learning practitioners in the spirit of \cite{Yang2021}.

\begin{subappendices}
	\chapter*{Appendix}
	\addcontentsline{toc}{chapter}{Appendix}
	\section{Proofs}
\label{svm:app:proofs}

\paragraph{Notation.}
This paper makes use of the following standard notations.
We used the simplex notation $\prob{A}$ to denote the space of probability measure over a Polish space $A$. We also used the notation $\Y^\X$ to denote functions from $\X$ to $\Y$ (which can be seen as a sequence of elements in $\Y$ indexed by $\X$).

\paragraph{Equations.}
Some standard derivations in the fast rates literature were omitted in the main paper. 
In particular, under Assumption \ref{svm:ass:margin}, we know that when $\abs{g(x) - \eta(x)} < \abs{\eta(x)}$ then $\sign g(x) = \sign \eta(x) = f^*(x)$, hence ${\cal R}(\sign g) - {\cal R}(f^*) \leq \Pbb_X(\sign g(X) \neq f^*(X)) \leq 1_{\norm{g - \eta}_{\infty} \geq \eta_0}$.
As a consequence,
$\E_{{\cal D}_n}[{\cal R}(\sign g_n)] - {\cal R}(f^*) \leq \Pbb_{{\cal D}_n}(\|g - \eta\|_{\infty} \geq \eta_0)$.

Similarly, the outline exposition follows from the fact that
\begin{align*}
\E_{{\cal D}_n}[{\cal R}(\sign g_{\theta_n})] - {\cal R}(f^*) 
&= \E_{{\cal D}_n}[{\cal R}(\sign g_{\theta_n}) - {\cal R}(\sign g_{\theta^*})]
\leq \E_{{\cal D}_n}[1_{\sign g_{\theta_n} \neq (\sign g_{\theta^*})}]
\\&\leq \E_{{\cal D}_n}[1_{\norm{\theta_n - \theta^*}\geq \epsilon}]
\leq \Pbb_{{\cal D}_n}(\norm{\theta_n - \theta^*}\geq \epsilon)
\\&\leq \Pbb_{{\cal D}_n}({\cal R}_S(g_{\theta_n}) - {\cal R}(g_{\theta^*})\geq \epsilon).
\end{align*}

\subsection{Proof of Lemma \ref{svm:lem:l1}}

The first part follows by integration of a pointwise result.
Consider the function $h_p: \R\to\R; q \mapsto p(1-q)_+ + (1-p)(1+q)_+$, where $p\in(0, 1)$ represents $\Pbb(Y=1\vert X)$ and $q$ represents $g(x)$.
The function $h_p$ has a slope equal to $-p$ for $q < -1$, then slope $1-2p$ for $q\in(-1, 1)$, and $1-p$ for $q > 1$. Therefore, when $q_1, q_2 \in (-1, 1)$, we have
\[
	h_p(q_2) - h_p(q_1) = (1-2p) (q_2 - q_1).
\]
Taking $p = \Pbb(Y=1\vert X)$, $q_2 = g_2(X)$ and $q_1 = g_1(X)$, we get $1-2p = -\E[Y\vert X] = -\eta(X)$. By integration, we obtain the claim.
From the previous slope considerations, it also follows that $h_p$ is minimized by $q=\sign(2p-1)$, meaning that one can take $g^*(X) = \sign\eta(X)$.

The second part follows from the fact that projecting on $[-1, 1]$ can only reduce the value of the hinge loss, that $\eta(x)(\pi(g)(x) - g^*(x))$ is always negative, and the reverse H\"older inequality:
\[
	{\cal R}_S(g) - {\cal R}(g^*) \geq {\cal R}_S(\pi(g)) - {\cal R}(g^*) = \norm{\eta (\pi(g) - g^*)}_{1}
	\geq \norm{\pi(g) - g^*}_{q} \norm{\abs{\eta}^{-1}}_{p}^{-1}.
\]
A H\"older inequality also holds for weak Lebesgue spaces \citep[see][Theorem 5.23]{Castillo2016}, whence
\[
	{\cal R}_S(g) - {\cal R}(g^*) \geq \norm{\eta (\pi(g) - g^*)}_{1}
	\geq \norm{\eta (\pi(g) - g^*)}_{1,\infty}
	\geq \frac{1}{2} \norm{ |\eta|^{-1} }_{p,\infty}^{-1} \norm{ \pi(g) - g^* }_{\frac{p}{p+1},\infty}.
\]
This completes the proof.

\subsection{Proof of Lemma \ref{svm:lem2}}
Assume without restrictions that there exists $x \in \X_1$ such that $|g^*(x) - g(x)| \geq 1 $.
For any event $A = A(X)$, by the law of total probability we have
\[
	\Pbb(A) =
	\rho_\X(\X_1) \Pbb\paren{A\midvert X\in \X_1}
	+ \rho_\X(\X_{-1}) \Pbb\paren{A\midvert X\in \X_{-1}}
	\geq \rho_\X(\X_1) \Pbb\paren{A\midvert X\in \X_1}.
\]
Hence, since $g^*(\X_1) = \brace{1}$,
\begin{align*}
	\norm{g - g^*}_{q, \infty}^q
	= \sup_{t > 0} t^q \Pbb(\abs{g(X) - g^*(X)} > t)
	\geq \sup_{t > 0} t^q\Pbb\paren{\abs{g(X) - 1} > t\midvert X\in\X_1}\rho_\X(\X_1).
\end{align*}
Using the triangular inequality, the $G$-Lipschitz continuity of $g$, and the definition of $x$, we have that, for any $x'\in\X$,
\[
	|g(x') - 1| \geq |g(x) - 1| - |g(x') - g(x)| \geq 1 - Gd(x, x').
\]
As a consequence,
\[
	\Pbb\paren{|g(X) - 1| > t\midvert X\in\X_1}
	\geq \Pbb\paren{X\in{\cal B}\paren{x, \frac{1-t}{G}} \mid X\in\X_1}
	= \frac{\rho_\X\paren{\X_1\cap {\cal B}\paren{x, \frac{1-t}{G}}}}{\rho_\X\paren{\X_1}}.
\]
Combined with the previous facts, we get
\[
	\norm{g - g^*}_{q, \infty}^q
	\geq \sup_{t > 0} t^q \rho_\X\paren{\X_1\cap {\cal B}\paren{x, \frac{1-t}{G}}}.
\]
Thanks to Assumption \ref{svm:ass:mass}, there exists $(c, r, d)$ such that \eqref{svm:eq:mass} holds for $\X_1$. Hence, when $G^{-1} < r$, we get the following lower bound:
\[
	\norm{g - g^*}_{q, \infty}^q
	\geq c G^{-d} \sup_{t \in [0, 1]} t^q (1 - t)^d
	= c G^{-d} \frac{q^q d^d}{(d+q)^{d+q}}.
\]
This proves the statement in the lemma.

\subsection{Proof of Proposition \ref{svm:prop:mid}}
Suppose $ {\cal R}(\sign g) > {\cal R}(f^*) $.
Then, observing that $ \sign(\pi(t)) = \sign(t) $ for all $ t \in \R $, and taking $ g^* = f^* $, we know there must be $ x \in \supp \rho_\X $ such that $ \abs{ \pi(g(x)) - g^*(x) } \geq 1 $.
Hence, by Lemma \ref{svm:lem2}, we get $ \norm{ \pi(g) - g^* }_{q,\infty} \geq c_0 G^{-\frac{d}{q}} $, and therefore, by Lemma \ref{svm:lem:l1},
$
	{\cal R}_S(g) - {\cal R}_S(g^*)
	\geq 2^{-1} c_0 G^{-\frac{d}{q}} \norm{\abs{\eta}^{-1}}_{p,\infty}^{-1} .
$
Thus, the proposition is proved.

\subsection{Proof of Theorem \ref{svm:thm}}
From Proposition \ref{svm:prop:mid} and Assumption \ref{svm:ass:source}, we get, with $\tilde{L} = \max\brace{M G_\phi \sigma^{-1}, r^{-1}}$ and $q = \sfrac{p}{p+1}$,
\begin{align*}
	\E_{{\cal D}_n}[{\cal R}(\sign g_{{\cal D}_n})] - {\cal R}(f^*)
	&\leq \Pbb_{{\cal D}_n} \paren{{\cal R}_S(\pi\circ g_{{\cal D}_n}) - {\cal R}_S(g^*) \geq 2^{-1}\norm{\abs{\eta}^{-1}}_{p,\infty}^{-1} c_0 \tilde{L}^{-\frac{d}{q}}}
	\\&\leq \Pbb_{{\cal D}_n} \paren{{\cal R}_S(\pi\circ g_{{\cal D}_n}) - {\cal R}_S(g_{M,\sigma}) \geq 4^{-1}\norm{\abs{\eta}^{-1}}_{p,\infty}^{-1} c_0 \tilde{L}^{-\frac{d}{q}}}.
\end{align*}
To deal with this last quantity, we proceed by usign the fact that
\[
    {\cal R}_{S,{\cal D}_n}(\pi\circ g_{{\cal D}_n}) \leq  {\cal R}_{S,{\cal D}_n}(g_{{\cal D}_n}) \leq  {\cal R}_{S,{\cal D}_n}(g_{M,\sigma}),
\]
where ${\cal R}_{S,{\cal D}_n}$ denotes the empirical surrogate risk,
to deduce that
\[
   {\cal R}_S(\pi\circ g_{{\cal D}_n}) - {\cal R}_S(g_{M,\sigma})
   \leq {\cal R}_S(\pi\circ g_{{\cal D}_n}) - {\cal R}_{S,{\cal D}_n}(\pi\circ g_{{\cal D}_n}) + {\cal R}_{S,{\cal D}_n}(g_{M,\sigma}) + {\cal R}_S(g_{M,\sigma}).
\]
Hence, we get the following union bound
\begin{align*}
	\E_{{\cal D}_n}[{\cal R}(\sign g_{{\cal D}_n})] - {\cal R}(f^*)
    &\leq \Pbb_{{\cal D}_n} \paren{{\cal R}_S(\pi\circ g_{{\cal D}_n}) - {\cal R}_{S, {\cal D}_n}(g_{{\cal D}_n}) \geq 8^{-1}\norm{\abs{\eta}^{-1}}_{p,\infty}^{-1} c_0 \tilde{L}^{-\frac{d}{q}}}
	\\&\qquad+ \Pbb_{{\cal D}_n} \paren{{\cal R}_S(\pi\circ g_{M,\sigma}) - {\cal R}_{S,{\cal D}_n}(g_{M,\sigma}) \geq 8^{-1}\norm{\abs{\eta}^{-1}}_{p,\infty}^{-1} c_0 \tilde{L}^{-\frac{d}{q}}}.
\end{align*}
Regarding the first term, we can reuse the literature on Rademacher complexity for linear models on convex risks \citep{Bartlett2002}, which ensures that
\[
	\E_{{\cal D}_n}\bracket{\sup_{g\in{\cal G}_{M,\sigma}}\abs{{\cal R}_S(\pi\circ g) - {\cal R}_{S, {\cal D}_n}(\pi\circ g)}} \leq M \norm{\phi}_{\infty} n^{-1/2}.
\]
Note that Assumption \ref{svm:ass:mass} implies that $\supp\rho_\X$ is compact, hence, if $\phi$ is Lipschitz-continuous, it is bounded on $\supp\rho_\X$.
This allows us to use McDiarmid inequality to get the same type of bound on the deviation of ${\cal R}_S(g_{{\cal D}_n})$ around its mean.
Let $H({\cal D}_n) = \sup_{g\in{\cal G}_{M,\sigma}} {\cal R}_S(\pi\circ g) - {\cal R}_{{\cal D}_n}(\pi\circ g)$. 
Let us decompose ${\cal D}_n = ((x_1, y_1), \cdots, (x_n, y_n))$. We would like to show that if ${\cal D}_n'$ is equal to ${\cal D}_n$ for each datapoint but for $(x_i, y_i)$ that becomes $(x_i', y_i')$ then $H({\cal D}_n) - H({\cal D}_n')$ is bounded.
We have
\begin{align*}
  H({\cal D}_n) - H({\cal D}_n') 
  &= \sup_{g\in{\cal G}_{M,\sigma}} {\cal R}_S(\pi\circ g) - {\cal R}_{S,{\cal D}_n}(\pi\circ g) - \sup_{g'\in{\cal G}_{M,\sigma}} {\cal R}_S(\pi\circ g') - {\cal R}_{S,{\cal D}_n'}(\pi\circ g')
  \\&\leq \sup_{g\in{\cal G}_{M,\sigma}} {\cal R}_{S, {\cal D}_n}(\pi\circ g) - {\cal R}_{S,{\cal D}_n'}(\pi\circ g)
  \\& = n^{-1}\sup_{g\in{\cal G}_{M, \sigma}} L(\pi\circ g(x_i'), y_i') - L(\pi\circ g(x_i), y_i)
  \leq n^{-1}
\end{align*}
Using McDiarmid's inequality, we get
\[
    \Pbb(H({\cal D}_n) - \E[H({\cal D}_n)] \geq t) \leq \exp(-2nt^2).
\]
In other terms, when adding the control we have on the expectation, we get
\begin{equation}
	\Pbb_{{\cal D}_n}\paren{\sup_{g\in{\cal G}_{M,\sigma}} {\cal R}_S(\pi\circ g) - {\cal R}_{S, {\cal D}_n}(\pi\circ g) > t + M\norm{\phi}_{\infty} n^{-1/2}} \leq \exp\paren{-2nt^2}.
\end{equation}
When $8^{-1}\norm{\abs{\eta}^{-1}}_{p,\infty}^{-1} c_0 \tilde{L}^{-\frac{d}{q}} \geq \norm{\phi}_{\infty} M n^{-1/2}$, this leads to
\begin{align*}
	&\Pbb_{{\cal D}_n} \paren{{\cal R}_S(\pi\circ g_{{\cal D}_n}) - {\cal R}_{S, {\cal D}_n}(g_{{\cal D}_n}) \geq 8^{-1}\norm{\abs{\eta}^{-1}}_{p,\infty}^{-1} c_0 \tilde{L}^{-\frac{d}{q}}}
	\\&\leq \exp\paren{-\frac{n}{8} \paren{8^{-1}\norm{\abs{\eta}^{-1}}_{p,\infty}^{-1} c_0 \tilde{L}^{-\frac{d}{q}} - M\norm{\phi}_{\infty} n^{-1/2}}^2}
	\\&\leq \exp\paren{-\frac{c_0^2 \sigma^{\frac{2d(p+1)}{p}}}{512\norm{\abs{\eta}^{-1}}_{p,\infty}^2(MG_\phi)^{\frac{2d(p+1)}{p}}}\cdot n + \frac{c_0\norm{\phi}_{\infty}}{32\norm{\abs{\eta}^{-1}}_{p,\infty}M^{\frac{d(p+1)}{p}-1}G_\phi^{\frac{d(p+1)}{p}}} \cdot n^{-1/2} - \frac{M^2\norm{\phi}_{\infty}^2}{8}}.
\end{align*}

Regarding the second term, we can use classical concentration of ${\cal R}_{{\cal D}_n}(g_{M,\sigma})$ around its mean.
For example, using the fact that Assumption \ref{svm:ass:mass} implies that $\rho_\X$ is compact, and using the fact that $L$ and $g_{\sigma, M}$ are Lipschitz, we deduce that $L(g_{\sigma, M}, Y)$ is bounded, hence one can applies Hoeffding's inequality to get the same type of exponential control on this term.

The result follows form those concentration inequality and the fact that ${\cal R}$ is bounded by one and that any $\min(1, a \exp(-bn))$ for $a, b > 0$ and $n>1$ can be bounded by $\exp(-cn)$ for a $c>0$).

\section{Experimental details}
\label{svm:app:experiments}

In our experiments, we used the SVM implementation of \cite{Chang2011} through its {\em Scikit-learn} wrapper \citep{Pedregosa2011} in {\em Python}.
We used {\em Numpy} \citep{Harris2020} to reduce our work to high-level array instructions, and {\em Matplotlib} for visualization \citep{Hunter2007}.
Randomness in experiments was controlled with the random seed provided by {\em Numpy}, which we initialized at zero.

\begin{figure}[ht]
	\centering
	\includegraphics{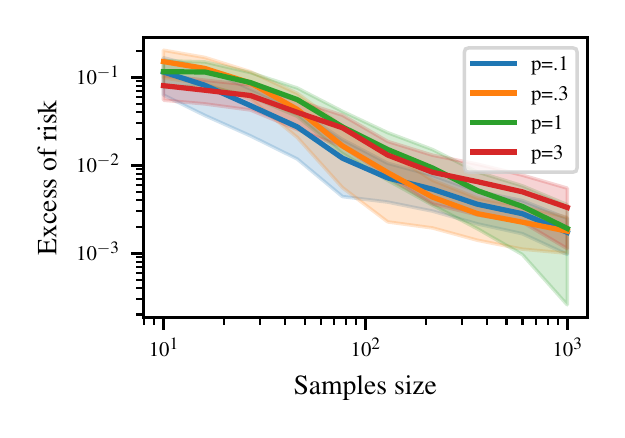}
	\includegraphics{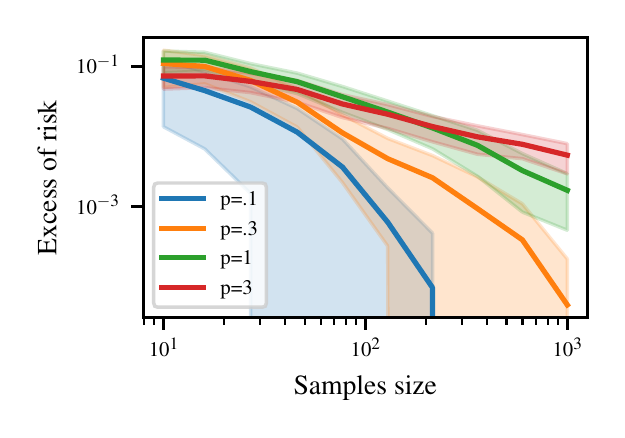}
	\caption{
		(Left) Similar setting as Figure \ref{svm:fig:fig1} but with $X$ uniform on $[-1, 1]$.
		The behavior of the excess of risk is quite different without the separation in $\X$: no exponential convergence rate is kicking in after a thousand of samples.
		(Right) Similar setting as Figure \ref{svm:fig:fig1}, using kernel ridge regression with the least-squares surrogate.
		Exponential convergence rates are observed with a slight delay compared to the hinge loss, and are explained by the hard-margin condition \ref{svm:ass:margin}.
	}
	\label{svm:fig:fig1_app}
\end{figure}

Figures \ref{svm:fig:fig1} and \ref{svm:fig:fig1_app} are derived by averaging 100 trials of the following procedure.
We draw uniformly at random $n$ independent samples uniformly distributed on $\X \in \brace{[-1, 1], [-1, -.1]\cup [.1, 1]}$.
We draw randomly one output $y_i$ for each input $x_i$, according to $\eta(x_i)$.
We consider the Gaussian kernel $k(x, x') = \exp(-\norm{x-x'}^2 / 2\sigma^2)$ for $\sigma = .2$, and solve the empirical risk minimization associated to the hinge loss with the penalization $\lambda \norm{\theta}^2$ (rather than the hard constraint $\norm{\theta} < M$) for $\lambda = 10^{-4}$.
The generalization error is measured through the formula $\E[\norm{\eta(x)} \ind{f(x) \neq f^*(x)}]$, with an empirical approximation of this sum with the points $(x_i)_{i\leq n}$ chosen such that $\rho_{\X}([x_i, x_{i+1}]) = 1/n$ and $\rho_{\X}([x_n, +\infty)) = 1/n$, with $n = 10^4$ (which makes sure that the exponential behavior observed is not due to the lack of testing samples).
For each $x$, the height of each dark part corresponds to one standard deviation of the generalization error computed from the 100 trials, and the solid line corresponds to the empirical average.
The fact that the dark parts are not centered around the averages is due to the fact that we have drawn $\log$-plots but centered the interval for linearly-scaled plots.

Figure~\ref{svm:fig:fig2} is obtained by considering $\X = [0, 1]^2$ with uniform input distribution, the Gaussian kernel with $\sigma = .2$, and the penalty parameter $\lambda = 10^{-3}$ (instead of a hard constraint leading to a parameter $M$ as in the main text derivations).
We take $n=10^4 = 100^2$ points uniformly spread out on $\X$ (on the regular lattice $\frac{1}{\sqrt{n}}\cdot \Z^2 \cap \X$) to approximate $g_{\lambda, \sigma}$ with empirical risk minimization on this curated dataset.
We consider $\eta(x) = \pi_{[-1, 1]}(2 x_2 - .5\sin(2\pi x_1) - 1)$, and assign to each $x$ in the dataset a sample $(x, 1)$ weighted by $\Pbb\paren{Y=1\midvert X=x} = (\eta(x) - 1) / 2$, and a sample $(x, -1)$, weighted by $\Pbb\paren{Y=-1\midvert X=x}$.
The ``noiseless'' setting denotes the setting where $\paren{Y\midvert X}$ is deterministic, but with the same decision frontier between the classes $\X_1$ and $\X_{-1}$ characterized by $\brace{(x, .5 + .25\sin(2\pi x))\midvert x\in [0,1]}$.
Once we fit the support vector machine with this dataset, we test it with $n=2.5\cdot 10^5 = 500^2$ data points uniformly spread out on $\X$, and use {\em Matplotlib} to automatically draw level lines.

\begin{figure}[ht]
	\centering
	\includegraphics{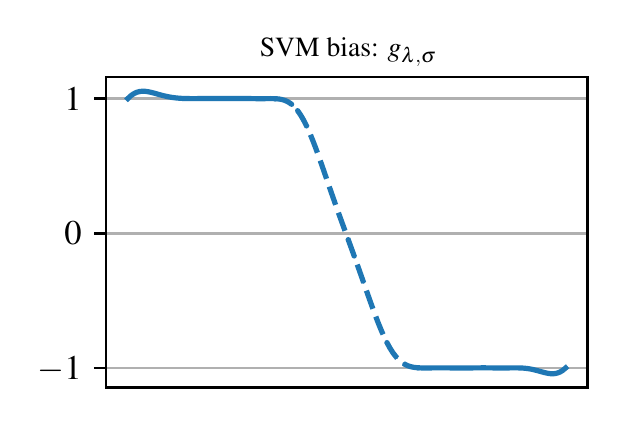}
	\includegraphics{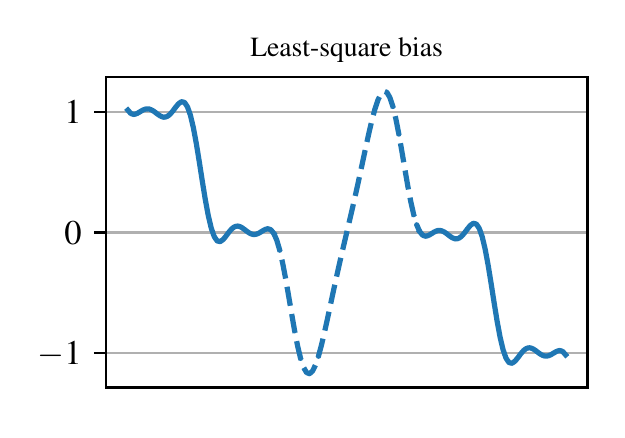}
	\includegraphics{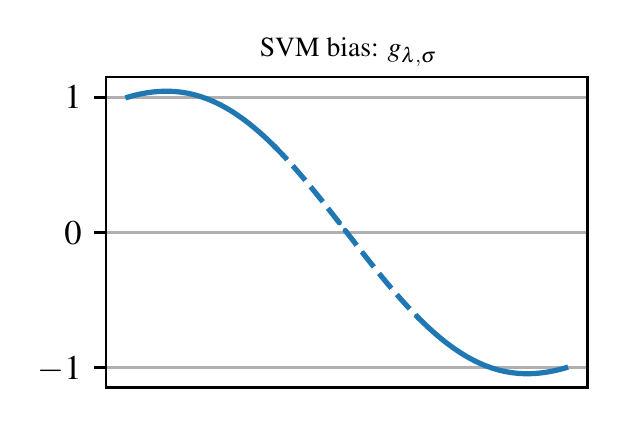}
	\includegraphics{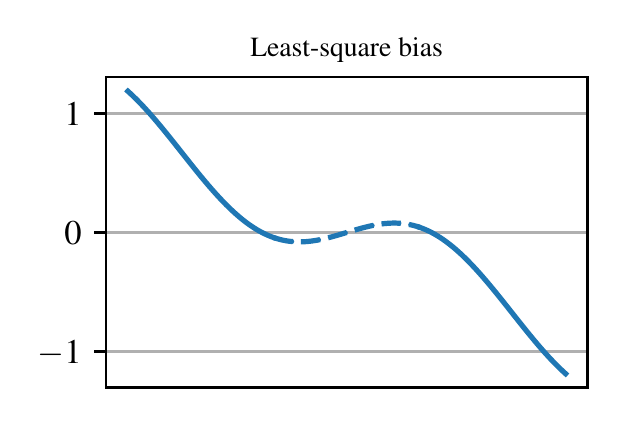}
	\caption{
		Same setting as Figure \ref{svm:fig:fig3}, with $\sigma = .2$ and $\lambda = 10^{-6}$ (top), and with $\sigma = 1$ and $\lambda = 10^{-3}$ (bottom).
	}
	\label{svm:fig:fig3_app}
\end{figure}

Figures~\ref{svm:fig:fig3} and \ref{svm:fig:fig3_app} correspond to $\X = [0, 3]$ with the input distribution uniform on $[0,1]\cup[2,3]$.
Figure~\ref{svm:fig:fig3} is obtained with $\sigma=.1$ and $\lambda=10^{-6}$.
We derive it by considering $n=100$ points uniformly spread out on the domain of $\eta$, solving the equivalent curated empirical risk minimization, that approximates both
\begin{align}
	g_{\lambda, \sigma} & = \argmin_{g:\X\to\R} \E_{\rho}[(0, 1- Y\scap{\theta}{\phi\paren{\frac{x}{\sigma}}})_+] + \lambda \norm{\theta}^2 ,
	\\g_{(\text{LS})} &= \argmin_{g:\X\to\R} \E_{\rho}[\norm{\scap{\theta}{\phi\paren{\frac{x}{\sigma}}} - Y}^2] + \lambda \norm{\theta}^2.
\end{align}

The robustness of SVM might be understood from its geometrical definition: when trying to find the maximum separating margin, infinitesimal modifications that change the regularity properties of $\eta$ do not really matter.
The picture is different for the least-squares surrogate with kernel methods, where from few point evaluations, the system reconstructs a function by assuming regularity and inferring information on high-order derivatives.
This is similar to the Runge phenomenon with Hermite interpolation.
More precisely, the Gaussian kernel is linked to a space of functions with rapidly decreasing Fourier coefficients \citep[see, for example,][for a more precise link]{Bach2023}.
The function $\eta$ that needs to be approximated on Figure~\ref{svm:fig:fig3} is similar to the Heaviside function, whose Fourier coefficients are of the form $(\frac{1}{i\pi k})_{k\in\N^*}$ and do not decrease fast enough to be all reconstructed.
This leads to some high-frequency oscillations missing in the reconstruction as it appears on Figure~\ref{svm:fig:fig3}.
\end{subappendices}

\part{Learning with Partial Supervision}
\label{part:weakly}
\chapter{Infimum Loss}

The following is a reproduction of \cite{Cabannes2020}.

Annotating datasets is one of the main costs nowadays in supervised learning.
The goal of weak supervision is to enable models to learn while using only forms of labeling which are cheaper to collect, as partial labeling.
This is a type of incomplete annotation where, for each data point, supervision is cast as a set of labels containing the real one.
The problem of supervised learning with partial labeling has been studied for specific instances such as classification, multi-label, ranking or segmentation, but a general framework is still missing.
This paper provides a unified framework based on structured prediction and on the concept of {\em infimum loss} to deal with partial labeling over a wide family of learning problems and loss functions.
The framework leads naturally to explicit algorithms that can be easily implemented and for which proved statistical consistency and learning rates.
Experiments confirm the superiority of the proposed approach over commonly used baselines.
\section{Introduction}

Fully supervised learning demands tight supervision of large amounts of data, a supervision that can be quite costly to acquire and constrains the scope of applications.
To overcome this bottleneck, the machine learning community is seeking to incorporate weaker sources of information in the learning framework.
In this paper, we address those limitations through partial labeling: {\em e.g.}, giving only partial ordering when learning user preferences over items, or providing the label ``flower'' for a picture of Arum Lilies, instead of spending a consequent amount of time to find the exact taxonomy.

Partial labeling has been studied in the context of classification~\citep{Cour2011,Nguyen2008}, multilabeling~\citep{Yu2014}, ranking~\citep{Hullermeier2008,Korba2018}, as well as segmentation~\citep{Verbeek2007,Papandreou2015}, however a generic framework is still missing.
Such a framework is a crucial step toward understanding how to learn from weaker sources of information, and widening the spectrum of machine learning beyond rigid applications of supervised learning.
Some interesting directions are provided by~\citet{CidSueiro2014,vanRooyen2018}, to recover the information lost in a corrupt acquisition of labels.
Yet, they assume that the corruption process is known, which is a strong requirement that we want to relax.

In this paper, we make the following contributions:
\begin{itemize}
  \item We provide a principled framework to solve the problem of learning with partial labeling, via {\em structured prediction}.
        This approach naturally leads to a variational framework built on the {\em infimum loss}.
  \item We prove that the proposed framework is able to recover the original solution of the supervised learning problem under identifiability assumptions on the labeling process.
  \item We derive an explicit algorithm which is easy to train and with strong theoretical guarantees.
        In particular, we prove that it is consistent, and we provide generalization error rates.
  \item Finally, we test our method against some simple baselines, on synthetic and real examples.
        We show that for certain partial labeling scenarios with symmetries, our infimum loss performs similarly to a simple baseline.
        However, in scenarios where the acquisition process of the labels is more adversarial in nature, the proposed algorithm performs consistently better.
\end{itemize}

\section{Partial labeling with infimum loss}\label{il:sec:partial-labeling} In this section, we introduce a statistical framework for partial labeling, and we show that it is characterized naturally in terms of risk minimization with the infimum loss.
First, let's recall some elements of fully supervised and weakly supervised learning.

  {\em Fully supervised learning} consists in learning a function~${f\in\Y^\X}$ between an input space~$\X$ and an output space~$\Y$, given a joint distribution ${\rho \in \Prob{\X \times \Y}}$ on~${\X \times \Y}$, and a loss function ${\ell \in \R^{\Y\times\Y}}$, that minimizes the risk
\begin{equation}\label{il:eq:risk}
  {\cal R}(f; \rho) = \E_{(X, Y)\sim\rho}\bracket{\ell(f(X), Y)},
\end{equation}
given observations ${(x_i, y_i)_{i\leq n} \sim \rho^{\otimes n}}$.
We will assume that the loss~$\ell$ is proper, {\em i.e.} it is continuous non-negative and is zero on, and only on, the diagonal of~${\Y \times \Y}$, and strictly positive outside.
We will also assume that $\Y$ is compact.

In \emph{weakly supervised learning}, given~$(x_i)_{i\leq n}$, one does not have direct observations of~$(y_i)_{i\leq n}$ but weaker information.
The goal is still to recover the solution~${f \in \Y^\X}$ of the fully supervised problem~\eqref{il:eq:risk}.
In \emph{partial labeling}, also known as \emph{superset learning} or as \emph{learning with ambiguous labels}, which is an instance of weak supervision, information is cast as closed sets $(S_i)_{i\leq n}$ in ${\cal S}$, where ${\cal S}\subset 2^\Y$ is the space of closed subsets of $\Y$, containing the true labels ${(y_i \in S_i)}$.
In this paper, we model this scenario by considering a data distribution ${\tau \in \Prob{\X \times {\cal S}}}$, that generates the samples $(x_i, S_i)$.
We will denote $\tau$ as {\em weak distribution} to distinguish it from $\rho$.
Capturing the dependence on the original problem, $\tau$~must be compatible with $\rho$, a matching property that we formalize with the concept of eligibility.
\begin{definition}[Eligibility]\label{il:def:eligibility}
  Given a probability measure $\tau$ on $\X \times {\cal S}$, a probability measure $\rho$ on $\X \times \Y$ is said to be eligible for $\tau$ (denoted by $\rho \vdash \tau$), if there exists a probability measure $\pi$ over $\X \times \Y \times {\cal S}$ such that $\rho$ is the marginal of $\pi$ over $\X \times \Y$, $\tau$ is the marginal of $\pi$ over $\X \times {\cal S}$, and, for $y\in\Y$ and $S\in{\cal S}$
  \[
    y\notin S \qquad \Rightarrow \qquad \PP_{\pi}\paren{S\midvert
      Y=y} = 0.
  \]
  We will alternatively say that $\tau$ is a {\em weakening} of~$\rho$, or that $\rho$ and $\tau$ are {\em compatible}.
\end{definition}
\subsection{Disambiguation principle}\label{il:sec:disambiguation-principle}

According to the setting described above, the problem of partial labeling is completely defined by a loss and a weak distribution~$(\ell, \tau)$.
The goal is to recover the solution of the original supervised learning problem in~\eqref{il:eq:risk} assuming that the original distribution verifies $\rho\vdash\tau$.
Since more than one $\rho$ may be eligible for $\tau$, we would like to introduce a guiding principle to identify a $\rho^\star$ among them.
With this goal we define the concept of {\em non-ambiguity} for $\tau$, a setting in which a natural choice for $\rho^\star$ appears.
\begin{definition}[Non-ambiguity]\label{il:def:non-ambiguity}
  For any $x \in \X$, denote by $\tau\vert_x$ the conditional probability of $\tau$ given $x$, and define the set $S_x$ as \[ S_x = \bigcap_{S \in \supp(\tau\vert_x)} S. \] The weak distribution $\tau$ is said {\em non-ambiguous} if, for every $x\in\X$, $S_x$ is a singleton.
  Moreover, we say that $\tau$ is {\em strictly non-ambiguous} if it is non-ambiguous and there exists ${\eta\in (0,1)}$ such that, for all ${x \in \X}$ and ${z \notin S_x}$
  \[\PP_{S \sim \tau\vert_x}(z \in S) \leq 1 - \eta.\]
\end{definition}
This concept is similar to the one by~\citet{Cour2011}, but more subtle because this quantity only depends on~$\tau$, and makes no assumption on the original distribution~$\rho$ describing the fully supervised process that we can not access.
In this sense, it is also more general.

When $\tau$ is non-ambiguous, we can write $S_x = \brace{y_x}$ for any $x$, where $y_x$ is the only element of $S_x$.
In this case it is natural to identify $\rho^\star$ as the one satisfying $\rho^\star\vert_x = \delta_{y_x}$.
Actually, such a $\rho^\star$ is characterized without $S_x$ as the only deterministic distribution that is eligible for $\tau$.
Because deterministic distributions are characterized as minimizing the minimum risk~\eqref{il:eq:risk}, we introduce the following {\em minimum variability principle} to disambiguate between all eligible $\rho$'s, and identify $\rho^\star$,
\begin{equation}\label{il:eq:infimum-disambiguation}
  \rho^\star \in \argmin_{\rho\vdash \tau}{\cal E}(\rho),\qquad {\cal E}(\rho) = \inf_{f:\X\to\Y} {\cal R}(f; \rho).
\end{equation}
The quantity ${\cal E}$ can be identified as a variance, since if $f_\rho$ is the minimizer of ${\cal R}(f; \rho)$, $f_\rho(x)$ can be seen as the mean of $\rho\vert_x$ and $\ell$ the natural distance in $\Y$.
Indeed, when $\ell = \ell_2$ is the mean square loss, this is exactly the case.
The principle above recovers exactly $\rho^\star\vert_x = \delta_{y_x}$, when $\tau$ is non-ambiguous, as stated by Theorem \ref{il:thm:ambiguity}, proven in Appendix \ref{il:proof:ambiguity}.
\begin{proposition}[Non-ambiguity determinism]\label{il:thm:ambiguity}
  When $\tau$ is non-ambiguous, the solution $\rho^\star$ \eqref{il:eq:infimum-disambiguation} exists and satisfies that, for any $x\in\X$, $\rho^\star\vert_x = \delta_{y_x}$, where $y_x$ is the only element of $S_x$.
\end{proposition}
Proposition \ref{il:thm:ambiguity} provides a justification for the usage of the minimum variability principle.
Indeed, under non-ambiguity assumption, following this principle will allow us to build an algorithm that recovers the original fully supervised distribution.
Therefore, given samples $(x_i, S_i)$, it is of interest to test if $\tau$ is non-ambiguous.
Such tests should leverage other regularity hypotheses on $\tau$, which we will not address in this work.

Now, we characterize the minimum variability principle in terms of a variational optimization problem that we can tackle in Section \ref{il:sec:algorithm} via empirical risk minimization.

\subsection{Variational formulation via the infimum loss}
Given a partial labeling problem $(\ell, \tau)$, define the solutions based on the minimum variability principle as the functions minimizing the recovered risk
\begin{equation}\label{il:eq:fstar-of-rhostar}
  f^* \in \argmin_{f:\X \to \Y} {\cal R}(f; \rho^\star).
\end{equation}
for $\rho^\star$ a distribution solving~\eqref{il:eq:infimum-disambiguation}.
As shown in Theorem \ref{il:thm:infimum-loss} below, proven in Appendix \ref{il:proof:infimum-loss}, the proposed disambiguation paradigm naturally leads to a variational framework involving the {\em infimum loss}.
\begin{theorem}[Infimum loss (\emph{IL})]\label{il:thm:infimum-loss}
  The functions $f^*$ defined in~\eqref{il:eq:fstar-of-rhostar} are characterized as \[ f^* \in \argmin_{f:\X \to \Y} {\cal R}_S(f), \] where the risk ${\cal R}_S$ is defined as
  \begin{equation}\label{il:eq:set-risk}
    {\cal R}_S(f) = \E_{(X,S)\sim\tau}\bracket{L(f(X), S)},
  \end{equation}
  and $L$ is the {\em infimum loss}
  \begin{equation}\label{il:eq:infimum-loss}
    L(z, S) = \inf_{y\in S} \ell(z, y).
  \end{equation}
\end{theorem}
The infimum loss, also known as the ambiguous loss \citep{Luo2010,Cour2011}, or as the optimistic superset loss \citep{Hullermeier2014}, captures the idea that, when given a set~$S$, this set contains the good label $y$ but also a lot of bad ones, that should not be taken into account when retrieving~$f$.
In other terms, $f$ should only match the best guess in $S$.
Indeed, if $\ell$ is seen as a distance, $L$ is its natural extension to sets.

\subsection{Recovery of the fully supervised solutions}
In this subsection, we investigate the setting where an original fully supervised learning problem $\rho_0$ has been weakened due to incomplete labeling, leading to a weak distribution~$\tau$.
The goal here is to understand under which conditions on $\tau$ and $\ell$ it is possible to recover the original fully supervised solution based on the infimum loss framework.
Denote $f_0$ the function minimizing ${\cal R}(f;\rho_0)$.
The theorem below, proven in Appendix \ref{il:proof:non-ambiguity}, shows that under non-ambiguity and deterministic conditions, it is possible to fully recover the function $f_0$ also from $\tau$.
\begin{theorem}[Supervision recovery]\label{il:thm:non-ambiguity}
  For an instance $(\ell, \rho_0, \tau)$ of the weakened supervised problem, if we denote by $f_0$ the minimizer of~\eqref{il:eq:risk}, we have the under the conditions that (1)~$\tau$ is not ambiguous (2)~for all $x\in\X$, $S_x = \brace{f_0(x)}$; the infimum loss recovers the original fully supervised solution, {\em i.e.} the $f^*$ defined in~\eqref{il:eq:fstar-of-rhostar} verifies $f^* = f_0$.

  Furthermore, when $\rho_0$ is deterministic and $\tau$ not ambiguous, the $\rho^\star$ defined in~\eqref{il:eq:infimum-disambiguation} verifies $\rho^\star = \rho_0$.
\end{theorem}
At a comprehensive level, this theorem states that under non-ambiguity of the partial labeling process, if the labels are a deterministic function of the inputs, the infimum loss framework makes it possible to recover the solution of the original fully supervised problem while only accessing weak labels.
In the next subsection, we will investigate which is the relation between the two problems when dealing with an estimator $f$ of $f^*$.
\subsection{Comparison inequality}\label{il:sec:calibration}
In the following, we want to characterize the error performed by ${\cal R}(f;\rho^\star)$ with respect to the error performed by ${\cal R}_S(f)$.
This will be useful since, in the next section, we will provide an estimator for $f^*$ based on structured prediction, that minimizes the risk ${\cal R}_S$.
First, we introduce a measure of discrepancy for the loss function.
\begin{definition}[Discrepancy of the loss $\ell$]
  Given a loss function $\ell$, the {\em discrepancy degree} $\nu$ of $\ell$ is defined as
  \[ \nu = \log\sup_{y, z'\neq z} \frac{\ell(z, y)}{\ell(z, z')}. \]
  $\Y$ will be said discrete for $\ell$ when $\nu < +\infty$, which is always the case when $\Y$ is finite.
\end{definition}
Now we are ready to state the comparison inequality that generalizes a result on classification with the $0-1$ loss from \citet{Cour2011} to arbitrary losses and output spaces.
\begin{proposition}[Comparison inequality]\label{il:thm:calibration}
  When $\Y$ is discrete and $\tau$ is strictly non-ambiguous for a given $\eta \in (0,1)$, then the following holds
  \begin{equation}\label{il:eq:calibration-bound}
    {\cal R}(f; \rho^\star) - {\cal R}(f^*;\rho^\star) \leq C({\cal R}_S(f) - {\cal R}_S(f^*)),
  \end{equation}
  for any measurable function $f \in \Y^\X$, where $C$ does not depend on $\tau, f$, and is defined as follows and always finite
  \[ C = \eta^{-1} e^\nu.\]
\end{proposition}
When $\rho_0$ is deterministic, since we know from Theorem~\ref{il:thm:non-ambiguity} that $\rho^\star=\rho_0$, this theorem allows bounding the error made on the original fully supervised problem with the error measured with the infimum loss on the weakly supervised one.

Note that the constant presented above is the product of two independent terms, the first measuring the ambiguity of the weak distribution $\tau$, and the second measuring a form of discrepancy for the loss.
In the appendix, we provide a more refined bound for $C$, that is $C = C(\ell, \tau)$, that shows a more elaborated interaction between $\ell$ and $\tau$.
This may be interesting in situations where it is possible to control the labeling process and may suggest strategies to active partial labeling, with the goal of minimizing the costs of labeling while preserving the properties presented in this section and reducing the impact of the constant $C$ in the learning process.
An example is provided in the Appendix \ref{il:discussion:refinement-C}.

\section{Consistent algorithm for partial labeling}\label{il:sec:algorithm}
In this section, we provide an algorithmic approach based on structured prediction to solve the weak supervised learning problem expressed in terms of infimum loss from Theorem~\ref{il:thm:infimum-loss}.
From this viewpoint, we could consider different structured prediction frameworks as structured SVM \citep{Tsochantaridis2005}, conditional random fields \citep{Lafferty2001} or surrogate mean estimation \citep{Ciliberto2016}.
For example, \citet{Luo2010} used a margin maximization formulation in a structured SVM fashion, \citet{Hullermeier2015} went for nearest neighbors, and \citet{Cour2011} design a surrogate method specific to the 0-1 loss, for which they show consistency based on~\citet{Bartlett2006}.

In the following, we will use the structured prediction method of~\citet{Ciliberto2016,Nowak2019}, which allows us to derive an explicit estimator, easy to train and with strong theoretical properties, in particular, consistency and finite sample bounds for the generalization error.
The estimator is based on the pointwise characterization of $f^*$ as
\[
  f^*(x) \in \argmin_{z\in\Y} \E_{S\sim \tau\vert_x}\bracket{\inf_{y\in S}\ell(z, y)},
\]
and weights $\alpha_i(x)$ that are trained on the dataset such that $\hat{\tau}_{\vert x} = \sum_{i=1}^n \alpha_i(x) \delta_{S_i}$ is a good approximation of $\tau\vert_x$.
Plugging this approximation in the precedent equation leads to our estimator, that is defined explicitly as follows
\begin{equation}\label{il:eq:algorithm}
  f_n(x) \in \argmin_{z\in\Y} \inf_{y_i \in S_i} \sum_{i=1}^n \alpha_i(x) \ell(z, y_i).
\end{equation}
Among possible choices for $\alpha$, we will consider the following kernel ridge regression estimator to be learned at training time
\[
  \alpha(x) = (K + n\lambda)^{-1}v(x),
\]
with $\lambda > 0$ a regularizer parameter and $K = (k(x_i, x_j))_{i,j} \in \R^{n\times n}, v(x) = (k(x, x_i))_{i} \in \R^n$ where $k\in \X\times\X \to \R$ is a positive-definite kernel \citep{Scholkopf2001} that defines a similarity function between input points ({\em e.g.}, if $\X = \R^d$ for some $d \in \N$ a commonly used kernel is the Gaussian kernel $k(x,x') = e^{-\|x-x'\|^2}$).
Other choices can be done to learn $\alpha$, beyond kernel methods, a particularly appealing one is harmonic functions, incorporating a prior on low density separation to boost learning \citep{Zhu2003,Zhou2003,Bengio2006}.
Here we use the kernel estimator since it allows deriving strong theoretical results, based on kernel conditional mean estimation \citep{Muandet2017}.

\subsection{Theoretical guarantees}
In this following, we want to prove that $f_n$ converges to $f^*$ as $n$ goes to infinity, and we want to quantify it with finite sample bounds.
The intuition behind this result is that as the number of data points tends toward infinity, $\hat{\tau}$ concentrates toward $\tau$, making our algorithm in~\eqref{il:eq:algorithm} converging to a minimizer of~\eqref{il:eq:set-risk} as explained more in detail in Appendix \ref{il:proof:consistency}.
\begin{theorem}[Consistency]\label{il:thm:consistency}
  Let $\Y$ be finite and $\tau$ be a non-ambiguous probability.
  Let $k$ be a bounded continuous universal kernel, {\em e.g.} the Gaussian kernel \citep[see][for details]{Micchelli2006}, and $f_n$ the estimator in~\eqref{il:eq:algorithm} trained on $n \in \N$ examples and with $\lambda = n^{-1/2}$.
  Then, holds with probability~$1$
  \[ \lim_{n \to \infty} {\cal R}(f_n; \rho^\star) = {\cal R}(f^*; \rho^\star).\]
\end{theorem}
In the next theorem, instead we want to quantify how fast $f_n$ converges to $f^*$ depending on the number of examples.
To obtain this result, we need a finer characterization of the infimum loss $L$ as:
\[
  L(z, S) = \scap{\psi(z)}{\phi(S)},
\]
where ${\cal H}$ is a Hilbert space and $\psi: \Y \to {\cal H}, \phi: 2^{\Y} \to {\cal H}$ are suitable maps.
Such a decomposition always exists in finite case (as
for the infimum loss over $\Y$ finite) and many explicit examples for losses of interest are presented by~\citet{Nowak2019}.
We now introduce the conditional expectation of $\phi(S)$ given $x$, defined as
\[
  \myfunction{g}{\X}{\cal H}{x}{\E_{\tau}\bracket{\phi(S)\midvert X = x}.}
\]
The idea behind the proof is that the distance between $f_n$ and $f$ is bounded by the distance of $g_n$ an estimator of $g$ that is implicitly computed via $\alpha$.
If $g$ has some form of regularity, {\em e.g.} $g \in {\cal G}$, with ${\cal G}$ the space of functions representable by the chosen kernel \citep[see][]{Scholkopf2001}, then it is possible to derive explicit rates, as stated in the following theorem.
\begin{theorem}[Convergence rates]\label{il:thm:learning-rates}
  In the setting of Theorem~\ref{il:thm:consistency}, if $\tau$ is $\eta$-strictly non-ambiguous for $\eta\in(0, 1)$, and if $g \in {\cal G}$, then there exists a $\tilde{C}$, such that, for any $\delta \in (0, 1)$ and $n\in \N$, holds with probability at least $1 - \delta$,
  \begin{equation}\label{il:eq:rates}
    {\cal R}(f_n; \rho^\star) - {\cal R}(f^*; \rho^\star) \leq
    \tilde{C} \log\paren{\frac{8}{\delta}}^2 n^{-1/4}.
  \end{equation}
\end{theorem}
Those last two theorem are proven in Appendix \ref{il:proof:consistency} and combines the consistency and learning results for kernel ridge regression \citep{Caponnetto2007,Smale2007}, with a comparison inequality of~\citet{Ciliberto2016} which relates the excess risk of the structured prediction problem with the one of the surrogate loss ${\cal R}_S$, together with our Theorem \ref{il:thm:calibration}, which relates the error ${\cal R}$ to ${\cal R}_S$.

Those results make our algorithm the first algorithm for partial labeling that, to our knowledge, is applicable to a generic loss $\ell$ and has strong theoretical guarantees as consistency and learning rates.
In the next section we will compare with the state of the art and other variational principles.

\section{Previous works and baselines}
\label{il:sec:inconsistency}
Partial labeling was first approached through discriminative models, proposing to learn $\paren{Y \midvert X}$ among a family of parametrized distributions by maximizing the log likelihood based on expectation-maximization scheme~\citep{Jin2002}, eventually integrating knowledge on the partial labeling process~\citep{Grandvalet2002,Papandreou2015}.
In the meanwhile, some applications of clustering methods have involved special instances of partial labeling, like segmentation approached with spectral method~\citep{Weiss1999}, semi-supervision approached with max-margin~\citep{Xu2004}.
Also, initially geared toward clustering,~\citet{Bach2007} considered the infimum principle on the mean square loss, and this was generalized to weakly supervised problems~\citep{Joulin2010}.
The infimum loss as an objective to minimize when learning from partial labels was introduced by~\citet{Cour2011} for the classification instance and used by~\citet{Luo2010,Hullermeier2014} in generic cases.
Compared to those last two, we provide a framework that derives the use of infimum loss from first principles and from which we derive an explicit and easy to train algorithm with strong statistical guarantees, which were missing in previous work.
In the rest of the section, we will compare the infimum loss with other variational principles that have been considered in the literature, in particular the supremum loss~\citep{Guillaume2017} and the average loss~\citep{Denoeux2013}.

\paragraph{Average loss (\emph{AC}).}
A simple loss to deal with uncertainty is to average over all potential candidates, assuming $S$ discrete,
\[
  L_{\textit{ac}}(z, S) = \frac{1}{\module{S}} \sum_{y\in S} \ell(z, y).
\]
It is equivalent to a fully supervised distribution $\rho_{\textit{ac}}$ by sampling $Y$ uniformly at random among $S$
\[ \rho_{\textit{ac}}(y) = \int_{{\cal S}} \frac{1}{\module{S}} \ind{y\in S} \diff\tau(S).\]
This directly follows from the definition of $L_{\textit{ac}}$ and of the risk ${\cal R}(z; \rho_{\textit{ac}})$.
However, as soon as the loss $\ell$ has discrepancy, {\em i.e.} $\nu > 0$, the average loss will implicitly advantage some labels, which can lead to inconsistency, even in the deterministic not ambiguous setting of Theorem \ref{il:thm:calibration} (see Appendix \ref{il:app:other-losses} for more details).

\paragraph{Supremum loss (\emph{SP}).}
Another loss that has been considered is the supremum loss~\citep{Wald1945,Madry2018}, bounding from above the fully supervised risk in~\eqref{il:eq:risk}.
It is widely used in the context of robust risk minimization and reads
\[
  R_{\textit{sp}}(f) = \sup_{\rho\vdash\tau} \E_{(X,Y)\sim\rho}\bracket{\ell(f(x), S)}.
\]
Similarly to the infimum loss in Theorem~\ref{il:thm:infimum-loss}, this risk can be written from the loss function
\[
  L_{\textit{sp}}(z, S) = \sup_{y\in S} \ell(z, y).
\]
Yet, this adversarial approach is not consistent for partial labeling, even in the deterministic non-ambiguous setting of Theorem \ref{il:thm:calibration}, since it finds the solution that best agrees with {\em all} the elements in $S$ and not only the true one (see Appendix \ref{il:app:other-losses} for more details).

\subsection{Instance showcasing superiority of our method}
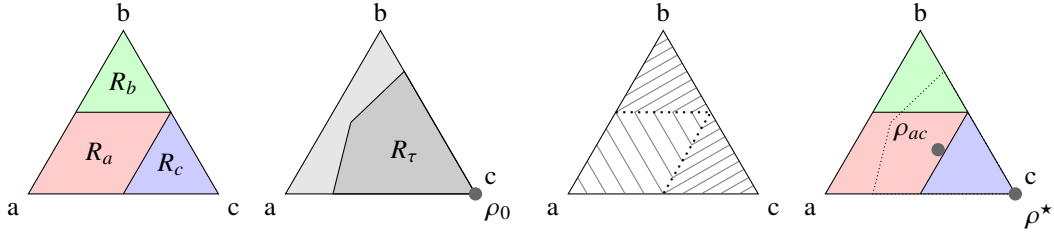
\begin{figure}[t]
  \centering
  \begin{tikzpicture}[scale=2.5]
  \coordinate(a) at (0, 0);
  \coordinate(b) at ({1/2}, {sin(60)});
  \coordinate(c) at (1, 0);
  \coordinate(ha) at ({3/4}, {sin(60)/2});
  \coordinate(hb) at ({1/2}, 0);
  \coordinate(hc) at ({1/4}, {sin(60)/2});

  \coordinate (mc) at (1,-.25);
  \coordinate (mca) at (0,-.25);
  \fill[fill=white] (a) -- (c) -- (mc) -- (mca) -- cycle;

  \fill[fill=red!20] (a) -- (hb) -- (ha) -- (hc) -- cycle;
  \fill[fill=green!20] (b) -- (ha) -- (hc) -- cycle;
  \fill[fill=blue!20] (c) -- (ha) -- (hb) -- cycle;
  \draw (a) node[anchor=north east]{a} -- (b) node[anchor=south]{b} --
        (c) node[anchor=north west]{c} -- cycle;
  \draw (hb) -- (ha) -- (hc);
  \node at ({3/8}, {sin(60)/4}) {$R_a$};
  \node at ({1/2}, {3*sin(60)/4 - 1/16}) {$R_b$};
  \node at ({3/4}, {sin(60)/4 - 1/16}) {$R_c$};
\end{tikzpicture}
\begin{tikzpicture}[scale=2.5]
  \coordinate(a) at (0, 0);
  \coordinate(b) at ({cos(60)}, {sin(60)});
  \coordinate(c) at (1, 0);
  \coordinate(r) at ({1/4}, 0);
  \coordinate(s) at ({7/32 + 1/8}, {7*sin(60)/16});
  \coordinate(t) at ({3/8 + 1/4}, {3*sin(60)/4});

  \coordinate (mc) at (1,-.25);
  \coordinate (mca) at (0,-.25);
  \fill[fill=white] (a) -- (c) -- (mc) -- (mca) -- cycle;

  \fill[fill=black!20] (c) -- (r) -- (s) -- (t) -- cycle;
  \fill[fill=black!10] (a) -- (r) -- (s) -- (t) -- (b) -- cycle;
  \draw (a) node[anchor=north east]{a} -- (b) node[anchor=south]{b} --
        (c) node[anchor=south west]{c} -- cycle;
  \draw (c) -- (r) -- (s) -- (t) -- cycle;
  \node at ({5/8}, {sin(60)/3 - 1/16}) {$R_{\tau}$};
  \fill[fill=black!60] (1, 0) circle (.1em) node[anchor=north west] {$\rho_0$};
\end{tikzpicture}
\begin{tikzpicture}[scale=2.5]  
  \coordinate (a) at (0,0) ;
  \coordinate (b) at ({1/2},{sin(60)}) ;
  \coordinate (c) at (1,0);
  \coordinate (mb) at (.25, {sin(60)});
  \coordinate (mba) at ({-.25*cos(30)},{.25*sin(30)});
  \coordinate (mc) at (1,-.25);
  \coordinate (mca) at (0,-.25);
  \coordinate(ha) at ({3/4}, {sin(60)/2});
  \coordinate(hb) at ({1/2}, 0);
  \coordinate(hc) at ({1/4}, {sin(60)/2});

  \foreach \x in {0,.05,...,.25}
  \draw[gray, rotate=30] ({sin(60)}, \x) -- ({sin(60) - (tan(60)*\x)}, \x) --
  ({sin(60) - (tan(60)*\x)}, -\x) -- ({sin(60)}, {-\x}); 
  \foreach \x in {0.25,.3,...,.5}
  \draw[gray, rotate=30] ({sin(60)}, \x) -- ({sin(60)-tan(60)*(.5-\x)}, \x);
  \foreach \x in {0.25,.3,...,.5}
  \draw[gray, rotate=30] ({sin(60)-tan(60)*(.5-\x)}, -\x) -- ({sin(60)}, {-\x}); 
  \foreach \x in {.25,.3,...,.5}
  \draw[gray, rotate=30] ({sin(60) - (tan(60)*\x)}, .25) -- ({sin(60) - (tan(60)*\x)}, -.25); 

  \fill[fill=white] (a) -- (c) -- (mc) -- (mca) -- cycle;
  \fill[fill=white] (a) -- (b) -- (mb) -- (mba) -- cycle;
  \draw (a) node[anchor=north east]{a} -- (b) node[anchor=south]{b} --
  (c) node[anchor=north west]{c} -- cycle;
  \draw[dotted, thick] (hc) -- (ha) -- (hb);
\end{tikzpicture}
\begin{tikzpicture}[scale=2.5]
  \coordinate(a) at (0, 0);
  \coordinate(b) at ({1/2}, {sin(60)});
  \coordinate(c) at (1, 0);
  \coordinate(ha) at ({3/4}, {sin(60)/2});
  \coordinate(hb) at ({1/2}, 0);
  \coordinate(hc) at ({1/4}, {sin(60)/2});
  \coordinate(r) at ({1/4}, 0);
  \coordinate(s) at ({7/32 + 1/8}, {7*sin(60)/16});
  \coordinate(t) at ({3/8 + 1/4}, {3*sin(60)/4});

  \coordinate (mc) at (1,-.25);
  \coordinate (mca) at (0,-.25);
  \fill[fill=white] (a) -- (c) -- (mc) -- (mca) -- cycle;

  \fill[fill=red!20] (a) -- (hb) -- (ha) -- (hc) -- cycle;
  \fill[fill=green!20] (b) -- (ha) -- (hc) -- cycle;
  \fill[fill=blue!20] (c) -- (ha) -- (hb) -- cycle;
  \draw (a) node[anchor=north east]{a} -- (b) node[anchor=south]{b} --
        (c) node[anchor=south west]{c} -- cycle;
  \draw (hb) -- (ha) -- (hc);
  \draw[densely dotted] (c) -- (r) -- (s) -- (t) -- cycle;
  \fill[fill=black!60] (1, 0) circle (.1em) node[anchor=north west] {$\rho^\star$};
  \fill[fill=black!60] ({13/96+11/24}, {13*sin(60)/48}) circle (.1em)
                       node[anchor=south east] {$\rho_{\textit{ac}}$};
\end{tikzpicture}
  \vspace*{-.3cm}
  \caption{Simplex $\Prob{\Y}$.
    (Left) Decision frontiers.
    (Middle left) Full and weak distributions.
    (Middle right) Level curves of the piecewise linear objective ${\cal E}$~\eqref{il:eq:infimum-disambiguation}, to optimize when disambiguating $\tau$ into $\rho^\star$.
    (Right) Disambiguation of \emph{AC} and \emph{IL}.}
  \label{il:fig:inconsistency}
\end{figure}

In the rest of this section, we consider a pointwise example to showcase the underlying dynamics of the different methods.
It is illustrated in Figure~\ref{il:fig:inconsistency}.
Consider $\Y = \brace{a, b, c}$ and a proper symmetric loss function such that $\ell(a, b) = \ell(a, c) = 1$, $\ell(b, c) = 2$.
The simplex $\Prob{\Y}$ is naturally split into decision regions, for $e\in\Y$,
\[
  R_e = \brace{\rho\in\Prob{\Y} \midvert e\in\argmin_{z\in\Y}\E_{\rho}[\ell(z, Y)]}.
\]
Both {\em IL} and {\em AC} solutions can be understood geometrically by looking at where $\rho^\star$ and $\rho_{\textit{ac}}$ fall in the partition of the simplex $(R_e)_{e\in\Y}$.
Consider a fully supervised problem with distribution $\delta_c$, and a weakening $\tau$ of $\rho$ defined by $\tau(\brace{a, b, c}) = \frac{5}{8}$ and $\tau(\brace{c}) = \tau(\brace{a,c}) = \tau(\brace{b,c}) = \frac{1}{8}$.
This distribution can be represented on the simplex in terms of the region $R_\tau = \brace{\rho\in\Prob{\Y}\midvert \rho\vdash \tau}$.
Finding $\rho^\star$ correspond to minimizing the piecewise linear function $ {\cal E}(\rho)$~\eqref{il:eq:infimum-disambiguation} inside $R_\tau$.
On this example, it is minimized for $\rho^\star = \delta_c$, which we know from
Theorem~\ref{il:thm:calibration}.
Now note that if we use the average loss, it disambiguates $\rho$ as
\[
  \rho_{\textit{ac}}(c) = \frac{11}{24} =
  \frac{1}{3}\frac{5}{8} + \frac{1}{8} + 2\cdot\frac{1}{2}\frac{1}{8}, \quad
  \rho_{\textit{ac}}(b) = \rho_{\textit{ac}}(a) = \frac{13}{48}.
\]
This distribution falls in the decision region of $a$, which is inconsistent with the real label $y=c$.
For the supremum loss, one can show, based on ${\cal R}_{\textit{sp}}(a) = \ell(a, c) = 1$, ${\cal R}_{\textit{sp}}(b) = \ell(b, c) = 2$ and ${\cal R}_{\textit{sp}}(c) = 3/2$, that the supremum loss is minimized for $z = a$, which is also inconsistent.
Instead, by using the infimum loss, we have $f^* = f_0 = c$, and moreover that $\rho^\star = \rho_0$ that is the optimal one.
\subsection{Algorithmic considerations for AC, SP}
The averaging candidates principle, approached with the framework of quadratic surrogates \citep{Ciliberto2016}, leads to the following algorithm
\begin{align*}
  f_{\textit{ac}}(x) & \in \argmin_{z\in\Y}
  \sum_{i=1}^n \alpha_{i}(x)\frac{1}{\module{S_i}}\sum_{y\in S_i} \ell(z, y)
  \\ &= \argmin_{z\in\Y} \sum_{y\in\Y}
  \paren{\sum_{i=1}^n \mathbf{1}_{y\in S_i}\frac{\alpha_i(x)}{ \module{S_i}}} \ell(z,y).
\end{align*}
This estimator is computationally attractive because the inference complexity is the same as the inference complexity of the original problem when approached with the same structured prediction estimator.
Therefore, one can directly reuse algorithms developed to solve the original inference problem~\citep{Nowak2019}.
Finally, with a similar approach to the one in Section \ref{il:sec:algorithm}, we can derive the following algorithm for the supremum loss
\[
  f_{\textit{sp}}(x) \in \argmin_{z\in\Y}
  \sup_{y_i \in S_i}\sum_{i=1}^n \alpha_i(x) \ell(z, y_i).
\]
In the next section, we will use the average candidates as baseline to compare with the algorithm proposed in this paper, as the supremum loss consistently performs worth, as it is not fitted for partial labeling.

\section{Applications and experiments}\label{il:sec:application}
In this section, we will apply~\eqref{il:eq:algorithm} to some synthetic and real datasets from different prediction problems and compared with the average estimator presented in the section above, used as a baseline.
Code is available online.\footnote{\url{https://github.com/VivienCabannes/partial_labelling}}

\subsection{Classification}\label{il:sec:classification}
Classification consists in recognizing the most relevant item among $m$ items.
The output space is isomorphic to the set of indices $\Y=\bbracket{1, m}$, and the usual loss function is the 0-1 loss
\[
  \ell(z, y) = \ind{y\neq z}.
\]
It has already been widely studied with several approaches that are calibrated in non-ambiguous deterministic settings, notably by~\citet{Cour2011}.
The infimum loss reads $L(z, S) = \ind{z\notin S}$, and its risk in~\eqref{il:eq:set-risk} is minimized for
\[
  f(x) \in \argmax_{z\in \Y} \PP\paren{z\in S \midvert X=x}.
\]
Based on data $(x_i, S_i)_{i\leq n}$, our estimator~\eqref{il:eq:algorithm} reads
\[
  f_n(x) = \argmax_{z\in\Y} \sum_{i;z\in S_i} \alpha_i(x).
\]
For this instance, the supremum loss is really conservative, only learning from set that are singletons $L_{\textit{sp}}(z, S) = \ind{S\neq\brace{z}}$, while the average loss is similar to the infimum one, adding an evidence weight depending on the size of $S$, $L_{\textit{ac}}(z, S) \simeq \ind{z\notin S} / \module{S}$.
\begin{figure}[t]
  \centering
  \includegraphics[width=.45\textwidth]{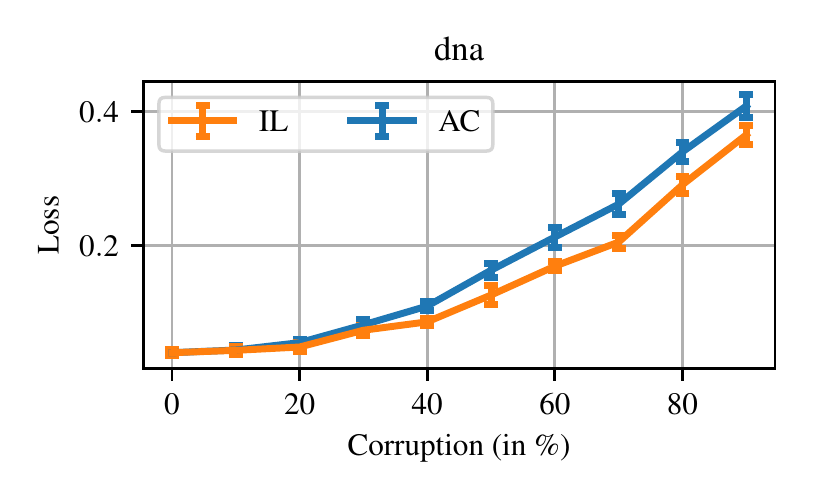}
  \includegraphics[width=.45\textwidth]{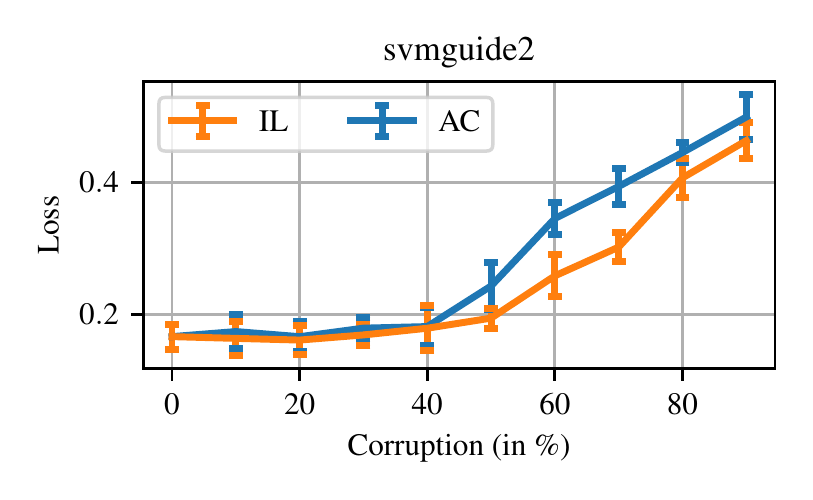}
  \vspace*{-.3cm}
  \caption{Classification.
    Testing risks (from~\eqref{il:eq:risk}) achieved by {\em AC} and {\em IL} on the ``DNA'' and ``svmguide2'' datasets from {\em LIBSVM} as a function of corruption parameter $c$, when the corruption is as follows: for $y$ being the most present labels of the dataset, and $z'\neq z$, $\PP\paren{z'\in S\midvert Y=z} = c\cdot \ind{z=y}$.
    Plotted intervals show the standard deviation on eight-fold cross-validation.
    Experiments were done with the Gaussian kernel.
    See all experimental details in~\eqref{il:app:experiments}.}
  \label{il:fig:libsvm}
\end{figure}
\paragraph{Real data experiment.} To compare {\em IL} and {\em AC}, we used {\em LIBSVM} datasets~\citep{Chang2011} on which we corrupted labels to simulate partial labeling.
When the corruption is uniform, the two methods perform the same.
Yet, when labels are unbalanced, such as in the ``DNA'' and ``svmguide2'' datasets, and we only corrupt the most frequent label $y\in\Y$, the infimum loss performs better as shown in Figure~\ref{il:fig:libsvm}.

\subsection{Ranking}\label{il:sec:ranking}
Ranking consists in ordering $m$ items based on an input~$x$ that is often the conjunction of a user $u$ and a query $q$, ($x=(u,q)$).
An ordering can be thought of as a permutation, that is, $\Y=\Sfrak_m$.
While designing a loss for ranking is intrinsically linked to a voting system \citep{Arrow1950}, making it a fundamentally hard problem; \citet{Kemeny1959} suggested approaching it through pairwise disagreement, which is current machine learning standard~\citep{Duchi2010}, leading to the Kendall embedding
\[
  \phi(y) = \paren{\sign\paren{y_i - y_j}}_{i<j \leq m},
\]
and the Kendall loss~\citep{Kendall1938}, with $C = m(m-1)/2$
\[
  \ell(y, z) = C - \phi(y)^T\phi(z).
\]
Supervision often comes as partial order on items, {\em e.g.},
\[
  S = \brace{y\in\Sfrak_m \midvert y_i > y_j > y_k, y_l > y_m}.
\]
It corresponds to fixing some coordinates in the Kendall embedding.
In this setting, \emph{AC} and \emph{SP} are not consistent, as one can recreate a similar situation to the one in Section~\ref{il:sec:inconsistency}, considering $m=3$, $a = (1,2,3)$, $b=(2,1,3)$ and $c=(1,3,2)$ (permutations being represented with $(\sigma^{-1}(i))_{i\leq m}$), and supervision being most often $S = (1>3) = \brace{a,b,c}$ and sometimes $S = (1>3>2) = \brace{c}$.

\paragraph{Minimum feedback arc set.}
Dealing with Kendall's loss requires solving problem of the form,
\[
  \argmin_{y \in S} \scap{c}{\phi(y)},
\]
for $c\in\R^{m^2}$, and constraints due to partial ordering encoded in $S\subset\Y$.
This problem is an instance of the constrained minimum feedback arc set problem.
We provide a simple heuristic to solve it in Appendix~\ref{il:app:fas}, which consists of approaching it as an integer linear program.
Such heuristics are analyzed and refined for analysis purposes by~\citet{Ailon2005,vanZuylen2007}.

\paragraph{Algorithm specification.}
At inference, the infimum loss requires solving:
\begin{equation}\tag{\ref{il:eq:algorithm}}
  f_n(x) = \argmax_{z\in\Y} \sup_{(y_i) \in S_i} \sum_{i=1}^n \alpha_i(x) \scap{\phi(z)}{\phi(y_i)}.
\end{equation}
It can be approached with alternate minimization, initializing $\phi(y_i) \in \hull(\phi(S_i))$, by putting $0$ on unseen observed pairwise comparisons, then, iteratively, solving a minimum feedback arc set problem in $z$, then solving several minimum feedback arc set problems with the same objective, but different constraints in $(y_i)$.
This is done efficiently using warm start on the dual simplex algorithm.
\begin{figure}[t]
  \centering
  \includegraphics{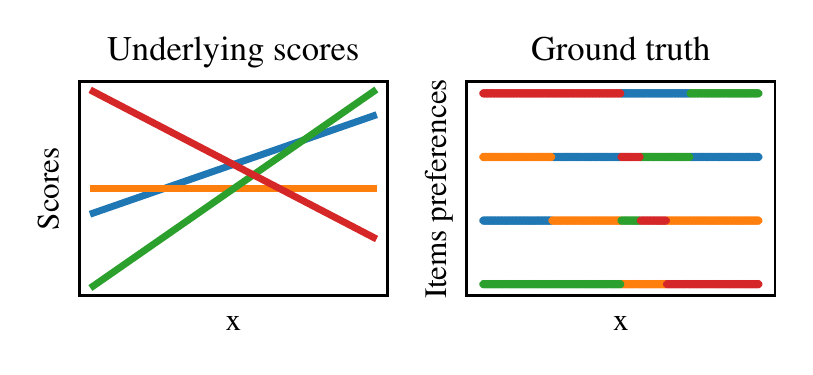}
  \vspace*{-.3cm}
  \caption{Ranking, experimental setting.
    Colors represent four different items to rank.
    Each item is associated with a utility function of $x$ shown on the left figure.
    From those scores, an ordering $y$ of the items is retrieved as represented on the right.}
  \label{il:fig:rk:setting}
\end{figure}
\begin{figure}[t]
  \centering
  \includegraphics{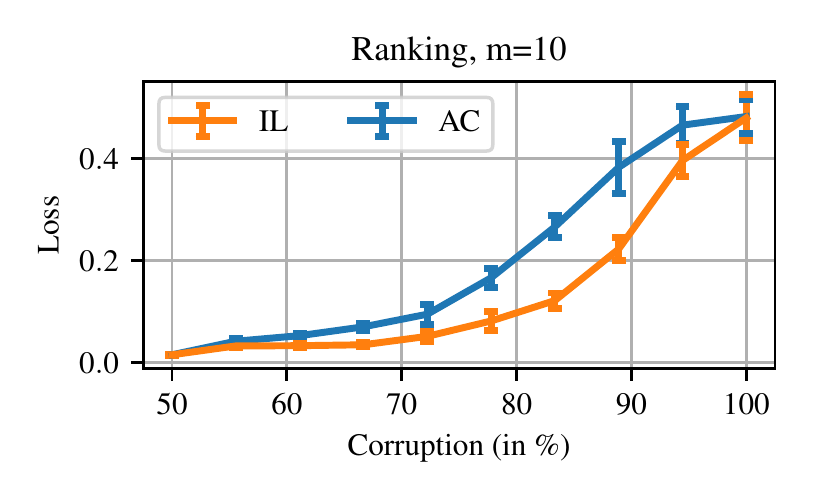}
  \vspace* {-.3cm}
  \caption{Ranking, results.
    Testing risks~(from~\eqref{il:eq:risk}) achieved by {\em AC} and {\em IL} as a function of corruption parameter $c$.
    When $c=1$, both risks are similar at $0.5$.
    The simulation setting is the same as in Figure~\ref{il:fig:libsvm}.
    The error bars are defined as for Figure~\ref{il:fig:libsvm}, after cross-validation over eight folds.
      {\em IL} clearly outperforms {\em AC}.}
  \label{il:fig:rk:corruption}
\end{figure}

\begin{figure*}[t]
  \centering
  \includegraphics[width=.45\textwidth]{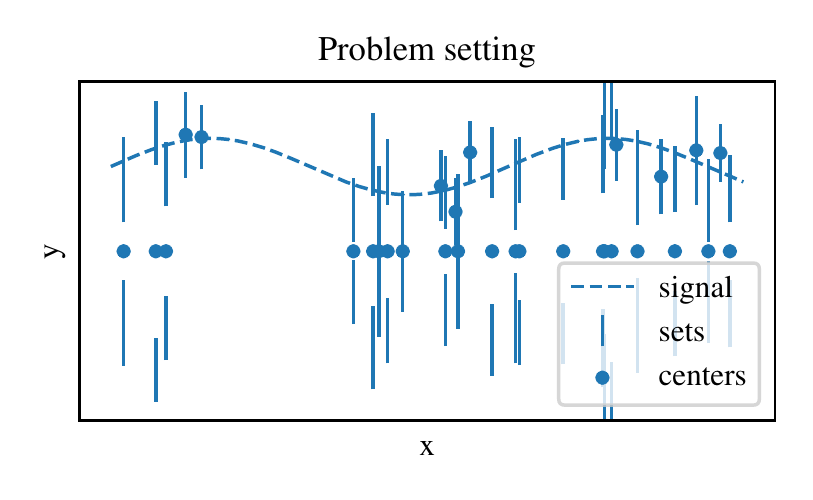}
  \includegraphics[width=.45\textwidth]{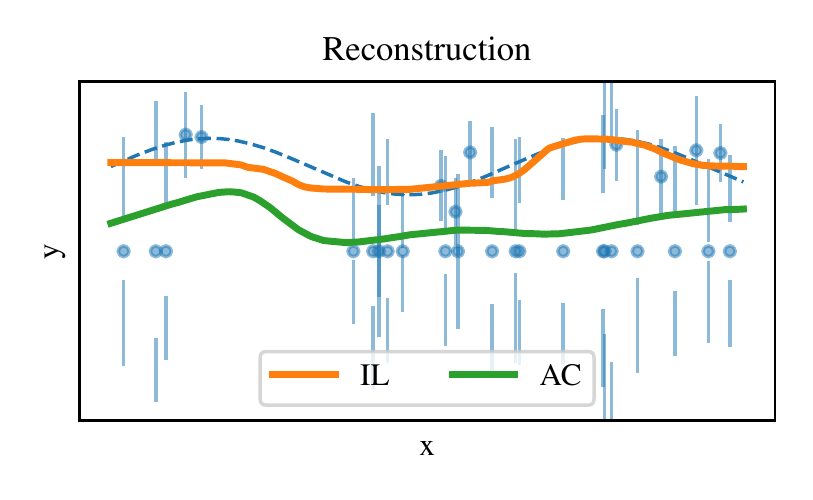}
  \vspace* {-.3cm}
  \caption{Partial regression on $\R$.
    In this setting we aim at recovering a signal $y(x)$ given upper and lower bounds on its amplitude, and in thirty percent of case, information on its phase, or equivalently in $\R$, its sign.
      {\em IL} clearly outperforms the baseline.
    Indeed, {\em AC} is a particularly ill-fitted method on such a problem, since it regresses on the barycenter of the resulting sets.}
  \label{il:fig:interval-regression}
\end{figure*}

\paragraph{Synthetic experiments.}
Let us consider $\X = [0,1]$ embodying some input features.
Let $\{1,\dots,m\}$, $m \in \N$ be abstract items to order, each item being linked to a utility function $v_i \in \R^\X$, that characterizes the value of $i$ for $x$ as $v_i(x)$.
Labels $y(x)\in\Y$ are retrieved by sorting $(v_i(x))_{i\leq m}$.
To simulate a problem instance, we set $v_i$ as $v_i(x) = a_i\cdot x + b_i$, where $a_i$ and $b_i$ follow a standard normal distribution.
Such a setting is illustrated in Figure~\ref{il:fig:rk:setting}.

After sampling $x$ uniformly on $[0, 1]$ and retrieving the ordering $y$ based on scores, we simulate partial labeling by randomly losing pairwise comparisons.
The comparisons are formally defined as coordinates of the Kendall's embedding $(\phi(y)_{jk})_{jk\leq m}$.
To create non-symmetric perturbations we corrupt more often items whose scores differ a lot.
In other words, we suppose that the partial labeling focuses on pairs that are hard to discriminate.
The corruption is set upon a parameter $c\in[0,1]$.
In fact, for $m=10$, until $c=0.5$, our corruption is fruitless since it can most often be inverted based on transitivity constraints in ordering, while the problem becomes non-trivial with $c \geq 0.5$.
In the latter setting, {\em IL} clearly outperforms {\em AC} on Figure~\ref{il:fig:rk:corruption}.

\subsection{Partial regression}\label{il:sec:partial-regression} Partial regression is an example of non-discrete partial labeling problem, where $\Y=\R^m$ and the usual loss is the Euclidean distance
\[
  \ell(y, z) = \norm{y - z}^2.
\]
This partial labeling problem consists of regression where observations are sets $S\subset \R^m$ that contain the true output $y$ instead of $y$.
Among others, it arises for example in economical models, where bounds are preferred over approximation when acquiring training labels~\citep{Tobin1958}.
As an example, we will illustrate how partial regression could appear for some phase problems arising with physical measurements.
Suppose a physicist wants to measure the law between a vector quantity $Y$ and some input parameters~$X$.
Suppose that, while she can record the input parameters $x$, her sensors do not exactly measure $y$ but render an interval in which the amplitude $\norm{y}$ lays and only occasionally render its phase $y / \norm{y}$, in a fashion that leads to a set of candidates $S$ for $y$.
The geometry over $\ell^2$ makes it a perfect example to showcase superiority of the infimum loss as illustrated in Figure~\ref{il:fig:interval-regression}.

In this figure, we consider ${\cal Y} = \mathbb{R}$ and suppose that $Y$ is a deterministic function of $X$ as shown by the dotted blue line signal.
If, for a given $x_i$, measurements only provides that $\module{y_i} \in [1, 2]$ without the sign of $y_i$, a situation where the phase is lost, this corresponds to the set $S_i = [-2, -1] \cup [1, 2]$, explaining the shape of observed sets that are symmetric around the origin.
Whenever the acquired data has no phase, which happens seventy percent of the time in our simulation, AC will target the set centers, explaining the green curve.
On the other hand, IL is aiming at passing by each set, which explains the orange curve, crossing all blue bars.

\section{Conclusions}
In this paper, we deal with the problem of weakly supervised learning, beyond standard regression and classification, focusing on the more general case of arbitrary loss functions and structured prediction.
We provide a principled framework to solve the problem of learning with partial labeling, from which a natural variational approach based on the infimum loss is derived.
We prove that under some identifiability assumptions on the labeling process the framework is able to recover the solution of the original supervised learning problem.
The resulting algorithm is easy to train and with strong theoretical guarantees.
In particular, we prove that it is consistent, and we provide generalization error rates.
Finally, the algorithm is tested on simulated and real datasets, showing that when the acquisition process of the labels is more adversarial in nature, the proposed algorithm performs consistently better than baselines.
This paper focuses on the problem of partial labeling, however the resulting mathematical framework is quite flexible in nature, and it is interesting to explore the possibility to extend it to tackle also other weakly supervised problems, as imprecise labels from non-experts~\citep{Dawid1979}, more general constraints over the set $(y_i)_{i\leq n}$ \citep{Quadrianto2009} or semi-supervision \citep{Chapelle2006}.

\begin{subappendices}
  \chapter*{Appendix}
  \addcontentsline{toc}{chapter}{Appendix}
  \section{Proofs}
\label{il:app:proof}

In the paper, we have implicitly considered $\X, \Y$ separable and completely metrizable topological spaces, {\em i.e.} Polish spaces, which allows considering probabilities.
Moreover, we assumed that $\Y$ is compact, to have minimizers well-defined.
The observation space was considered to be the set of closed subsets of $\Y$ endowed with the Hausdorff distance, ${\cal S} = {\operatorname{Cl}(\Y), d_H}$.
As such, ${\cal S}$ is also a Polish metric space, inheriting this property from $\Y$ \citep{Beer1993}.
In the following, we will show that the closeness of sets is important in order to switch from the minimum variability principle to the infimum loss.

In terms of notations, we use the simplex notation $\Prob{\cal A}$ to denote the space of Borel probability measures over the space ${\cal A}$.
In particular, $\Prob{\X\times\Y}$, $\Prob{\X\times{\cal S}}$ and $\Prob{\X\times\Y\times{\cal S}}$ are endowed with the weak-* topology and are Polish, inheriting the properties from original spaces \citep{Aliprantis2006}.
The fact that such spaces are Polish allows defining the conditional probabilities given $x \in \X$.
We will denote this conditional probability $\rho\vert_x$ when, for example, $\rho\in\Prob{\X\times\Y}$.
Finally, we will denote by $\rho_\X$ the marginal distributions of $\rho$ over $\X$.

Before diving into proofs, we would like to point out that many of our results are pointwise results.
At an intuitive level, we only leverage the structure of the loss on the output space and aggregate those results over $\X$.

\begin{remark}[Going pointwise]
  The learning frameworks in~\eqref{il:eq:risk}, \eqref{il:eq:infimum-disambiguation} and \eqref{il:eq:set-risk} are {\em pointwise separable} as their solutions can be written as aggregation of pointwise solutions \citep{Devroye1996}.
  More exactly, the partial labeling risk (and similarly the fully supervised one) can be expressed as
  \[
    {\cal R}_S(f) = \E_X\bracket{{\cal R}_{S,X}(f(X))},
  \]
  where the conditional risk reads,
  \[
    {\cal R}_{S, x}(z) = \E_{S\sim\tau\vert_x}\bracket{L(z, S)},
  \]
  with $\tau\vert_x$ the conditional distribution of $\paren{S\midvert X=x}$.
  Thus, minimizing ${\cal R}_S$ globally for $f\in\Y^\X$ is equivalent to minimizing locally ${\cal R}_{S,x}$ for $f(x)$ for almost all $x$.
  Similarly, for~\eqref{il:eq:infimum-disambiguation},
  \[
    {\cal E}(\rho) = \inf_{f:\X\to\Y}\E_{\rho}\bracket{\ell(f(X), Y)}
    = \E_X\bracket{\inf_{z\in\Y} \E_{Y\sim\rho\vert_x}\bracket{\ell(z, Y) \midvert X=x}}.
  \]
  Therefore, studies on risk can be done pointwise on instances $(\ell, \rho\vert_x, \tau\vert_x)$, before integrating along $\X$.
  Actually, Theorems \ref{il:thm:ambiguity}, \ref{il:thm:infimum-loss}, \ref{il:thm:non-ambiguity} and \ref{il:thm:calibration} are pointwise results.
\end{remark}

\subsection{Proof of Theorem~\ref{il:thm:ambiguity}}
\label{il:proof:ambiguity}

Here we want to prove that when $\tau$ is non-ambiguous, then it is possible to define an optimal $\rho^\star$ that is deterministic on $\Y$, and that this $\rho^\star$ is characterized by solving~\eqref{il:eq:infimum-disambiguation}.
\begin{lemma}\label{il:lem:ambiguity}
  When $\tau$ is non-ambiguous, and there is one, and only one, deterministic distribution eligible for $\tau$.
  More exactly, if we write, for any $x\in\X$ in the support of $\tau_\X$, based on Definition~\ref{il:def:non-ambiguity}, $S_x = \brace{y_x}$, then this deterministic distribution is characterized as $\rho\vert_x=\delta_{y_x}$ almost everywhere.
\end{lemma}
\begin{proof}
  Let us consider a probability measure $\tau\in\Prob{\X\times{\cal S}}$.
  We begin by working on the concept of eligibility.
  Consider $\rho\in\Prob{\X\times\Y}$ eligible for $\tau$ and a suitable $\pi$ as defined in Definition~\ref{il:def:eligibility}.
  First of all, the condition that, for $y\in S$, $\PP_\pi\paren{S\midvert Y=y} = 0$, can be stated formally in terms of measure as
  \[
    \pi(\brace{(x, y, S)\in\X\times\Y\times{\cal S}\midvert y\notin S}) = 0,
  \]
  from which we deduced that, for $y\in\Y$ and $x\in\X$,
  \begin{align*}
    \rho\vert_x(y) & = \pi\vert_x(\brace{y}\times{\cal S})
    = \pi\vert_x(\brace{y} \times \brace{S \in{\cal S}\midvert y\in S})
    \\&\leq \pi\vert_x(\Y \times \brace{S \in{\cal S}\midvert y\in S})
    = \tau\vert_x(\brace{S\in{\cal S}\midvert y\in S}).
  \end{align*}
  It follows that when $\rho$ is deterministic, if we write $\rho\vert_x = \delta_{y_x}$, then we have
  \(
  \tau\vert_x(\brace{S\in{\cal S}\midvert y_x\in S}) = 1,
  \)
  which means that $y_x$ is in all sets that are in the support of $\tau\vert_x$, or that, using notations of Definition~\ref{il:def:non-ambiguity}, $y_x \in S_x$.
  So far, we have proved that if there exists a deterministic distribution, $\rho\vert_x=\delta_{y_x}$, that is eligible for $\tau\vert_x$, we have $y_x\in S_x$.
  Reciprocally, one can do the reverse derivations, to show that if $\rho\vert_x = \delta_{y_x}$, with $y_x \in S_x$, for all $x\in\X$, then $\rho$ is eligible for $\tau$
  When $\tau$ is non-ambiguous, $S_x$ is a singleton and therefore, there could be only one deterministic eligible distribution for $\tau$, that is characterized in the lemma.
\end{proof}
Now we use the characterization of deterministic distribution through the minimization of the risk~\eqref{il:eq:risk}.
\begin{lemma}[Deterministic characterization]\label{il:lem:deterministic}
  When $\Y$ is compact and $\ell$ proper, deterministic distribution are exactly characterized by minimum variability~\eqref{il:eq:infimum-disambiguation} as
  \[
    {\cal E}(\rho) = \inf_{f: \X\to\Y}\E_\rho\bracket{\ell(f(X), Y)} = 0.
  \]
\end{lemma}
\begin{proof}
  Let's consider $\rho\in\Prob{\X\times\Y}$, because $\Y$ is compact and $\ell$ continuous, we can consider $f_\rho$ a minimizer of ${\cal R}(f;\rho)$.
  Let's now suppose that ${\cal R}(f_\rho; \rho) = 0$, since $\ell$ is non-negative, it means that almost everywhere
  \[
    \E_{Y\sim\rho\vert_x}\bracket{\ell(f_\rho(x), Y)} = 0.
  \]
  Suppose that $\rho\vert_x$ is not deterministic, then there is at least two points $y$ and $z$ in $\Y$ in its support, then, because $\ell$ is proper, we come to the absurd conclusion that
  \[
    \E_{Y\sim\rho\vert_x}\bracket{\ell(f_\rho(x), Y)}
    \geq \rho\vert_x(y) \ell(f_\rho(x), y) + \rho\vert_x(z) \ell(f_\rho(x), z) > 0.
  \]
  So ${\cal R}(f_\rho; \rho) = 0$ implies that $\rho$ is deterministic.
  Reciprocally, when $\rho$ is deterministic it is easy to show that the risk is minimized at zero.
\end{proof}

\subsection{Proof of Theorem~\ref{il:thm:infimum-loss}}
\label{il:proof:infimum-loss}

At a comprehensive level, Theorem~\ref{il:thm:infimum-loss} is composed of two parts:
\begin{itemize}
  \item A double minimum switch, to take the minimum over $\rho$ before the minimum over $f$, and for which we need some compactness assumption to consider the joint minimum.
  \item A minimum-expectation switch, to take the minimum over $\rho\vdash\tau$ as a minimum $y\in S$ before the expectation to compute the risk, and for which we need some measure properties.
\end{itemize}
We begin with the minimum-expectation switch.
To proceed with derivations, we need first to reformulate the concept of eligibility in Definition~\ref{il:def:eligibility} in terms of measures.
\begin{lemma}[Measure eligibility]\label{il:lem:eligibility}
  Given a probability $\tau$ over $\X \times {\cal S}$, the space of probabilities over $\X\times\Y$ satisfying $\rho \vdash \tau$ is characterized by all probability measures of the form
  \[
    \rho(C) = \int_{\X\times\Y\times{\cal S}} \ind{C}(x, y)
    \diff\pi\vert_{x,S}(y) \diff\tau(x, S),
  \]
  for any $C$ a closed subset of $\X\times\Y$, and where $\pi$ is a probability measure over $\X\times\Y\times{\cal S}$ that satisfies $\pi_{\X\times{\cal S}} = \tau$ and $\pi\vert_{x,S}(S) = 1$ for any $(x, S)$ in the support of $\tau$.
\end{lemma}
\begin{proof}
  For any $\rho$ that is eligible for $\tau$ there exists a suitable $\pi$ on $\X \times \Y \times {\cal S}$ as specified by Definition~\ref{il:def:eligibility}.
  Actually, the set of $\pi$ leading to an eligible $\rho:=\pi_{\X\times\Y}$ is characterized by satisfying $\pi_{\X\times{\cal S}} = \tau$ and
  \[
    \pi(\{(x,y,S) \in \X\times\Y\times{\cal S} ~|~ y \notin S\}) = 0.
  \]
  This last property can be reformulated with the complementary space as
  \[
    \pi(\{(x,y,S) \in \X\times\Y\times{\cal S} ~|~ y \in S\}) = 1,
  \]
  which equivalently reads, that for any $(x, S)$ in the support of $\tau$, we have
  \[
    \pi\vert_{x, S}(S) = \pi\vert_{x, S}(\brace{y\in\Y\midvert y\in S}) = 1.
  \]
  Finally, using the conditional decomposition we have that, for $C$ a closed subset of $\X\times\Y$
  \[
    \rho(C) = \pi_{\X\times\Y}(C)
    = \int_{\X\times\Y\times{\cal S}} \ind{C}(x, y)\diff\pi(x, y, S)
    = \int_{\X\times\Y\times{\cal S}} \ind{C}(x, y)\diff\pi\vert_{x, S}(y)
    \diff\pi_{\X\times{\cal S}}(x, S),
  \]
  which ends the proof since $\tau = \pi_{\X\times{\cal S}}$.
\end{proof}
We are now ready to state the minimum-expectation switch.
\begin{lemma}[Minimum-Expectation switch]\label{il:lem:eligibility-infimum}
  For a probability measure $\tau\in\Prob{\X\times{\cal S}}$, and measurable functions $\ell \in \R^{\Y\times\Y}$ and $f \in \Y^\X$, the infimum of eligible expectations of $\ell$ is the expectation of the infimum of $f$ over $S$ where $S$ is distributed according to $\tau$.
  Formally
  \[
    \inf_{\rho\vdash\tau}\E_{(X, Y)\sim\rho}\bracket{\ell(f(X), Y)} =
    \E_{(X,S)\sim\tau}\bracket{\inf_{y\in S}\ell(f(X), y)}.
  \]
\end{lemma}
\begin{proof}
  Before all, note that $(x, S) \to \inf_{y\in S} \ell(f(x), y)$ inherit measurability from $f$ allowing to consider such an expectation \citep[see Theorem 18.19 of][and references therein for details]{Aliprantis2006}.
  Moreover, let us use Lemma~\ref{il:lem:eligibility} to reformulate the right hand side problem as
  \[
    \inf_{\rho\vdash\tau}\E_{(X, Y)\sim\rho}\bracket{\ell(f(X), Y)} =
    \inf_{\pi \in {\cal M}} \int_{\X\times\Y\times{\cal S}} \ell(f(x),
    y)\diff\pi_{x, S}(y)\diff\tau(x, S).
  \]
  Where we denote by ${\cal M} \subset \Prob{\X\times\Y\times{\cal S}}$ the space of probability measures $\pi$ that satisfy the assumption of Lemma~\ref{il:lem:eligibility}.
  We will now prove the equality by showing that both quantities bound the other one.

  \paragraph{($\geq$).} To proceed with the first bound, notice that for
  $x\in\X$ and $S\in{\cal S}$, when $\pi\vert_{x, S}\in\Prob{\Y}$ only charge $S$, {\em i.e.} if $\pi \in {\cal M}$, then
  \[
    \int_\Y \ell(f(x), y) \diff\pi_{x, S}(y) \geq \inf_{y\in S} \ell(f(x), y).
  \]
  The first bound is then obtained by taking the expectation over $\tau$ of this pointwise property.

  \paragraph{($\leq$).} For the second bound, we consider the function $Y \in \Y^{\X\times{\cal S}}$ define as
  \[
    Y(x, S) = \argmin_{y\in S} \ell(f(x), y).
  \]
  Such a function is well-defined since $S$ is compact due to the fact that $\Y$ is compact and ${\cal S}$ is the set of closed set.
  However, in more general cases, one can consider a sequence that minimizes $\ell(f(x), y)$ rather than the argmin to show the same as what we are going to show.
  Now, if we define $\pi^{(f)}$ with $\pi^{(f)}_{\X\times{\cal S}}:= \tau$ and $\pi^{(f)}\vert_{x, S} := \delta_{Y(x,S)}$, because $Y(x, S)$ is in $S$, we have that $\pi^{(f)}$ is in ${\cal M}$, so, for $x \in \X$ and $S\in{\cal S}$
  \[
    \inf_{\pi\in{\cal M}} \int_{\Y} \ell(f(x), y)\diff\pi_{x, S}(y) \leq
    \int_{\Y} \ell(f(x), y)\diff\pi^{(f)}_{x, S}(y) = \ell(f(x), Y(x, S)) =
    \inf_{y\in S} \ell(f(x), y).
  \]
  We end the proof by integrating this over $\tau$.
\end{proof}
Now, we will move on to the minimum switch.
First, we make sure that the infimum loss minimizer is well-defined.
\begin{lemma}[Infimum loss minimizer]
  When $\Y$ is compact and the observed set are closed, there exists a measurable function $f_S \in \Y^\X$ that minimize the infimum loss risk
  \[
    {\cal R}_S(f_S) = \inf_{f:\X\to\Y} {\cal R}_S(f), \qquad\text{where}\qquad
    {\cal R}_S (f) = \int \min_{y \in S} \ell(f(x), y) \diff\tau(x, S).
  \]
  The infimum on the right hand side is a minimum because $S$ is a closed subset of $\Y$ compact, and therefore, is compact.
\end{lemma}
\begin{proof}
  First note that $d(y,y') = \sup_{z \in \Y} \module{\ell(z,y) - \ell(z,y')}$ is a metric on $\Y$ when $\ell$ is a proper loss: the triangular inequality holds trivially; when $y = y'$ then $d(y,y') = 0$; when $y\neq y'$, by properness we have $\ell(y,y) = 0$ and $d(y, y') \geq \ell(y,y') > 0$.
  Moreover, note that $L(z,S) = \min_{y \in S} \ell(z,y)$ is continuous and $1$-Lipschitz with respect to the topology induced by the Hausdorff distance $d_H$ based on $d$, indeed given two sets $S,S' \in {\cal S}$
  \begin{align*}
    \module{L(z,S) - L(z,S')} & \leq \max\brace{
      \max_{y \in S}\min_{y' \in S'} \module{\ell(z,y)-\ell(z,y')},~
    \max_{y' \in S'}\min_{y \in S} \module{\ell(z,y)-\ell(z,y')}} \\
                              & \leq \max\brace{
      \max_{y \in S}\min_{y' \in S'} d(y,y'),~
      \max_{y' \in S'}\min_{y \in S} d(y,y')} = d_H(S,S').
  \end{align*}
  The result of existence of a measurable $f_S$ minimizing ${\cal R}_S(f) = \int L(f(x),S) d\tau(x, S)$ follows by the compactness of $\Y$, the continuity of $L(z,S)$ in the first variable with respect to the topology induced by $d$, in the second with respect to the topology induced by $d_H$ and measurability of $\tau\vert_x$ in $x$, via Berge maximum theorem \citep[see Theorem 18.19 of][and references therein]{Aliprantis2006}.
\end{proof}
We can state the minimum switch now.
\begin{lemma}[Minimum switch]\label{il:lem:switch}
  When $\Y$ is compact, and observed sets are closed, solving the partial labeling through the minimum variability principle
  \[
    f^* \in \argmin_{f\in\Y^\X} \E_{\rho^\star}\bracket{\ell(f(X), Y)}, \qquad\text{with}\qquad
    \rho^\star \in \argmin_{\rho\vdash\tau} \inf_{f\in\Y^\X} \E_{\rho}\bracket{\ell(f(X), Y)}.
  \]
  can be done jointly in $f$ and $\rho$, and rewritten as
  \[
    f^*\in\argmin_{f\in\Y^\X}\inf_{\rho\vdash\tau}\E_{\rho}\bracket{\ell(f(X), Y)}.
  \]
\end{lemma}
\begin{proof}
  When $(\rho^\star, f^*)$ is a minimizer of the top problem, it also minimizes the joint problem $(\rho, f) \to {\cal R}(f; \rho)$, and we can switch the infimum order.
  The hard part is to show that when $f_S$ minimizes the bottom risk, the infimum over $\rho$ is indeed a minimum.
  Indeed, we know from Lemma~\ref{il:lem:eligibility-infimum} that $f_S$ is characterized as a minimizer of the infimum risk ${\cal R}_S$, those are well-defined as shown in precedent lemma.
  To $f_S$, we can associate $\rho_S := \pi^{(f)}$ as defined in the proof of Lemma~\ref{il:lem:eligibility-infimum}, which is due to the closeness of sets in ${\cal S}$ and the compactness of $\Y$.
  Indeed, $(f_S, \rho_S)$ minimize jointly the objective ${\cal R}(f, \rho)$, so we have that
  \[
    \rho_S \in \argmin_{\rho\vdash\tau}\inf_{f:\X\to\Y} {\cal R}(f;
    \rho),\qquad\text{and}\qquad
    f_S \in \argmin_{f:\X\to\Y}{\cal R}(f; \rho_S).
  \]
  From which we deduced that $\rho_S$ can be written as a $\rho^\star$ and $f_S$ as an $f^*$.
\end{proof}
\begin{remark}[A counter example when sets are not closed.]
  The minimum switch relies on compactness assumption, which can be violated when the observed sets in ${\cal S}$ are not closed.
  Let us consider the case where $\Y = \R$, $\ell=\ell_2$ is the mean square loss.
  Consider the pointwise weak supervision
  \[
    \tau = \frac{1}{2} \delta_{\Q} + \frac{1}{2}\delta_{\sqrt{2}\Q},
  \]
  In this case, we have $\rho^\star = \delta_0$.
  Yet, for any $z$, we do have ${\cal R}_{S, x}(z) = 0$ for any $z \in \R$.
  For example, if $z=\sqrt{2}$, one can consider
  \[
    \rho_n = \frac{1}{2}\delta_{\sqrt{2}} + \frac{1}{2}\delta_{\frac{\floor{10^n\sqrt{2}}}{10^n}},
  \]
  to show that $z\in\argmin_{z\in\Y}\inf_{\rho\vdash\tau} {\cal R}(z, \rho)$.
  As one can see this is counter example is based on the fact that $\brace{\rho\midvert \rho\vdash\tau}$ is not complete, so that there exists infimum of ${\cal R}_{x}(z, \rho)$ that are not minimum such as ${\cal R}_x(\sqrt{2}, \delta_{\sqrt{2}})$.
\end{remark}
\subsection{Proof of Theorem~\ref{il:thm:non-ambiguity}}
\label{il:proof:non-ambiguity}

If $\tau$ is not ambiguous, then, almost surely for $x\in\X$, if $y_x$ is the only element in $S_x$ of Definition~\ref{il:def:non-ambiguity}, we know that $\rho^\star\vert_x = \delta_{y_x}$, and consequently we derive $f^*(x)=y_x$, so for it to be consistent with $f_0$, we need that $f_0(x)=y_x$.

Moreover, because $\tau$ is a weakening of $\rho_0$, $\rho_0$ is eligible for $\tau$.
When $\rho_0$ is deterministic, we know from considerations in the proof of Lemma~\ref{il:lem:ambiguity}, that it is $\rho^\star$, the only deterministic distribution eligible for $\tau$.
Thus, in fact, the condition $S_x = \brace{f_0(x)}$ is implied by $\rho_0$ deterministic.

\subsection{Proof of Theorem~\ref{il:thm:calibration}}
\label{il:proof:calibration}

When $\tau$ is not ambiguous, we know from Theorem~\ref{il:thm:ambiguity}, that $\rho^\star$ is deterministic.
Let us write $\rho^\star\vert_x = \delta_{y_x}$, we have $f^*(x) = y_x$, and ${\cal R}_x(f^*) = 0$, moreover, because $y_x$ is in every $S$ in the support of $\tau\vert_S$, then ${\cal R}_{S, x}(f^*) = 0$.
Similarly to the bound given by~\citet{Cour2011} for the 0-1 loss, we have
\begin{align*}
  {\cal R}_{S, x}(z) & = \E_{S\sim\tau\vert_x}[\inf_{z'\in S} \ell(z, z')]
  = \sum_{S; z\notin S} \inf_{z'\notin S} \ell(z, z')\PP_{S\sim\tau\vert_x}(S)
  \\&\geq \inf_{z'\neq z} \ell(z, z')\PP_{S\sim\tau\vert_x}(z\notin S)
  \geq \inf_{z'\neq z} \ell(z, z')\eta,
\end{align*}
while ${\cal R}_x(z) = \ell(z, y)$, so we deduce locally
\begin{align*}
  {\cal R}_{x}(z; \rho^\star\vert_x) - {\cal R}_{x}(f^*(x); \rho^\star\vert_x)
   & \leq \frac{\ell(z, y)}{\inf_{z'\neq z} \ell(z, z')} \eta^{-1} \paren{{\cal R}_{S, x}(z) - {\cal R}_{S, x}(f^*(x))}
  \\&\leq e^\nu\eta^{-1}\paren{{\cal R}_{S, x}(z) - {\cal R}_{S, x}(f^*(x))}.
\end{align*}
Integrating over $x$ this last equation gives us the bound in Theorem~\ref{il:thm:calibration}.

\subsection{Refined bound analysis of Theorem~\ref{il:thm:calibration}}
\label{il:discussion:refinement-C}
The constant $C$ that appears in Theorem~\ref{il:thm:calibration} is the result of controlling separately the corruption process and the discrepancy of the loss.
Indeed, they can be controlled together, leading to a better constant.
To relates the two risk ${\cal R}$ and ${\cal R}_S$, we will consider the pointwise setting $\tau\in\Prob{2^\Y}$ and $\rho_0\in\Prob{\Y}$ that satisfies $\rho_0\vdash\tau$, we will also consider a prediction $z\in\Y$.
\begin{proposition}[Bound refinement]\label{il:prop:refinement}
  When $\Y$ is discrete and $\tau$ not ambiguous, the best $C$ that verifies~\eqref{il:eq:calibration-bound} in the pointwise setting $\tau\in\Prob{2^\Y}$ is the maximum of $\lambda^{-1}$, for $\lambda \in [0, 1]$ such that there exists a point $z\neq y$ and signed measured $\sigma$ that verify ${\cal R}(z;\sigma) = 0$ and such that $\sigma + \lambda \delta_y + (1-\lambda)\delta_z$ is a probability measure that is eligible for $\tau$.
\end{proposition}
\begin{proof}
  First, let's extend our study to the space ${\cal M}_\Y$ of signed measure over $\Y$.
  We extend the risk definition in~\eqref{il:eq:risk} to any signed measure $\mu\in {\cal M}_\Y$, with
  \[
    {\cal R}_x(z; \mu) = \int_{\Y} \ell(z, y) \diff\mu(y).
  \]
  Note that the risk is a linear function of the distribution $\mu$.
  Two spaces are going to be of particular interest: the one of measures of mass one ${\cal M}_{\Y, 1}$, and the one of measures of mass null ${\cal M}_{\Y, 0}$, where
  \[
    {\cal M}_{\Y, p} = \brace{\mu\in{\cal M} \midvert \mu(\Y) = p}.
  \]
  Let's now relates for a $\rho_0$, $\tau$ and $z$, the risk ${\cal R}_x(z; \rho_0)$ and ${\cal R}_{S,x}(z)$.
  To do so, we introduce the space of signed measures of null mass, that could be said orthogonal to $(\ell(z, y))_{y\in\Y}$, formally
  \[
    D_z = \brace{\mu\in{\cal M}_{\Y, 0} \midvert {\cal R}_x(z; \mu) = 0}.
  \]
  There are two alternatives: (1) either ${\cal R}_x(z; \rho_0) = 0$, and so ${\cal R}_{S, x}(z) = 0$ too, and we have related the two risk; (2) either ${\cal R}_x(z, \rho_0) \neq 0$, and the space ${\cal M}_{\Y, 1}$ can be decomposed as
  \[
    {\cal M}_{\Y, 1} = D_z + \brace{\lambda \rho_0 + (1-\lambda)\delta_z \midvert
      \lambda \in \R}.
  \]
  To prove it take $\mu \in {\cal M}_{\Y, 1}$, and use linearity of the risk after writing
  \[
    \mu = \lambda \rho_0 + (1-\lambda)\delta_z + \paren{\mu - (\lambda \rho_0 + (1-\lambda)\delta_z)},
    \qquad\text{with}\qquad \lambda = \frac{{\cal R}_{x}(z, \mu)}{{\cal R}_{x}(z, \rho_0)}.
  \]
  For such a $\mu$, using the linearity of the risk, and the properness of the loss, if we denote by $d_z$ the part in $D_z$ of the last decomposition, we have
  \[
    {\cal R}_x(z; \mu) = \lambda {\cal R}_x(z; \rho_0) + (1-\lambda){\cal R}_x(z;
    \delta_z) + {\cal R}_x(z; d_z) = \lambda {\cal R}_x(z; \rho_0)
  \]
  If we denote by $R_\tau = \brace{\rho\in\Prob{\Y}\midvert \rho\vdash \tau}$, we can conclude that
  \[
    \frac{{\cal R}_{S, x}(z)}{{\cal R}_x(z; \rho_0)} = \inf\brace{\lambda\midvert
      (\lambda \rho_0 + (1-\lambda)\delta_z) \in R_\tau + D_z}.
  \]
  Finally, when $\tau$ is not ambiguous, we know that $\rho^\star$ is deterministic, and if $\rho_0$ is deterministic then $\rho_0 = \rho^\star$.
  In this case, there exists a $y$ such that $\rho_0 = \delta_y$, and we can suppose this $y$ is different from $z$ otherwise ${\cal R}_x(z; \rho_0) = 0$.
  In this case, we also have ${\cal R}_x(z^*) = {\cal R}_{S, x}(z^*) = 0$ with $z^* = y$, and thus the excess of risk to relates in~\eqref{il:eq:calibration-bound} is indeed the relation between the two risks.

  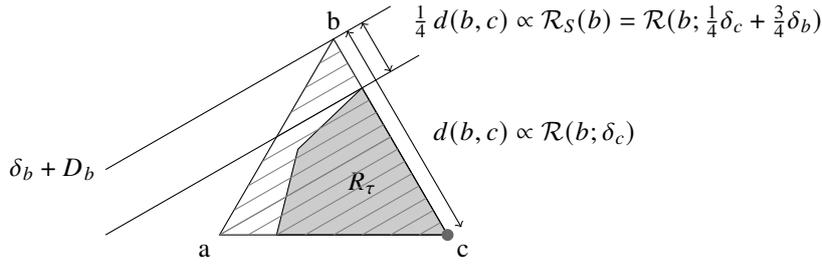
\begin{figure}[t]
    \centering
    \begin{tikzpicture}[scale=3]
  \coordinate(a) at (0, 0);
  \coordinate(b) at ({cos(60)}, {sin(60)});
  \coordinate(c) at (1, 0);
  \coordinate(r) at ({1/4}, 0);
  \coordinate(s) at ({7/32 + 1/8}, {7*sin(60)/16});
  \coordinate(t) at ({3/8 + 1/4}, {3*sin(60)/4});

  \fill[fill=black!20] (c) -- (r) -- (s) -- (t) -- cycle;
  \draw (a) node[anchor=north east]{a} -- (b) node[anchor=south]{b} --
        (c) node[anchor=north west]{c} -- cycle;
  \draw (c) -- (r) -- (s) -- (t) -- cycle;
  \node at ({5/8}, {sin(60)/3 - 1/16}) {$R_{\tau}$};
  \fill[fill=black!60] (1, 0) circle (.075em);

  \foreach \x in {0,.125,...,1} \draw[gray]
           ({\x / 2}, {\x*sin(60)}) -- ({(3 - \x) / 4)}, {(1 + \x) * sin(60)/2}); 
  \foreach \x in {0,.125,...,1} \draw[gray]
           ({\x, 0}) -- ({(3 + \x) / 4)}, {(1 - \x) * sin(60)/2});
  \draw (-.5,0) -- ({7/8}, {11*sin(60)/12});
  \draw (-.5,{sin(60)/3}) node[anchor=east] {$\delta_b + D_b$} -- ({3/4}, {7*sin(60)/6});
  \draw[<->] ({3/4}, {5*sin(60)/6}) -- ({11/16}, {23*sin(60)/24}) node[anchor=south west]
             {$\quad \frac{1}{4}\, d(b, c) \propto {\cal R}_S(b) =
              {\cal R}(b; \frac{1}{4}\delta_c + \frac{3}{4}\delta_b)$} -- ({5/8}, {13*sin(60)/12});    
  \draw[<->] ({1/2 + 3 * .04 / 2}, {(1 + .04) * sin(60)})  --
             ({7/8}, {2*(1 + 2*.04 - 7/8)*sin(60)}) node[anchor=south west]
             {$\, d(b, c) \propto {\cal R}(b; \delta_c)$} -- ({1 + 3 * .04 / 2}, {.04*sin(60)});    
\end{tikzpicture}
    \vspace*{-.3cm}
    \caption{Geometrical understanding of Proposition~\ref{il:prop:refinement}, showing the link between the infimum and the fully supervised risk.
    The drawing is set in the affine span of the simplex ${\cal M}_{\Y, 1}$, where we identify $a$ with $\delta_a$.
    The underlying instance $(\ell, \tau)$ is taken from Section~\ref{il:sec:inconsistency}, and can be linked to the setting of Proposition~\ref{il:prop:refinement} with $z=b$, $y=c$.
    Are represented in the simplex the level curves of the function $\rho\rightarrow {\cal R}(z;\rho)$.
    Based on this drawing, one can recover ${\cal R}_S(b) = {\cal R}(b)/4$, which is better than the bound given in Theorem~\ref{il:thm:calibration}.
    }
    \label{il:fig:simplex-calibration}
  \end{figure}
\end{proof}

\begin{remark}[Proposition \ref{il:prop:refinement} as a variant of Thales theorem]
  Proposition \ref{il:prop:refinement} can be seen as a variant of the Thales theorem.
  Indeed, with the geometrical embedding $\pi$ of the simplex in $\R^\Y$, $\pi(\rho) = (\rho(y))_{y\in\Y}$, one can have, with $d$ the Euclidean distance
  \[
    \frac{{\cal R}_{S, x}(z)}{{\cal R}_x(z; \rho_0)} = \frac{d(\pi(\delta_z +
      D_z), \pi(R_\tau))}{d(\pi(\delta_z + D_z), \pi(\rho_0))}.
  \]
  And conclude by using the following variant of Thales theorem, that can be derived from Figure~\ref{il:fig:proof-thales}:
  For $x, y, z \in \R^d$, and $S \subset \R^d$, with $d$ the Euclidean distance, if $y \in S$, $d(z+x^\perp, S) = \gamma d(z+x^\perp, y) $, where
  \[
    \gamma = \min\brace{\module{\lambda}\midvert \lambda\in \R,
      (\lambda y + (1- \lambda) z + x^\perp) \cap S \neq \emptyset}.
  \]
  Moreover, notice that if $S$ is contained in the half-space that contains $y$ regarding the cut with the hyperplane $z + x^\perp$, $\lambda$ can be restricted to be in $[0,1]$.

  \begin{figure}[t]
    \centering
    \begin{tikzpicture}[scale=1]
  \coordinate(xz1) at (0,3);
  \coordinate(xz2) at (7,3);
  \coordinate(xs1) at (0,2);
  \coordinate(xs2) at (7,2);
  \coordinate(xy1) at (0,0);
  \coordinate(xy2) at (7,0);
  \coordinate(s1) at (2,0);
  \coordinate(s2) at (4,0);
  \coordinate(s3) at (4,1);
  \coordinate(s4) at (3,2);
  \coordinate(s5) at (1,.5);
  \draw (xz1) node[anchor=east] {$z + x^\perp$} -- (xz2);
  \draw (xs1) node[anchor=east] {\footnotesize{$\lambda y + (1-\lambda)z + x^\perp$}} -- (xs2);
  \draw (xy1) node[anchor=east] {$y + x^\perp$} -- (xy2);
  \filldraw[fill=black!20] (s1) -- (s2) -- (s3) -- (s4) -- (s5) -- cycle;

  \fill[fill=black] (2,0) circle (.2em) node[anchor=north west] {$y$};
  \fill[fill=black] (1,3) circle (.2em) node[anchor=north west] {$z$};

  \coordinate(d1) at (4.5,2);
  \coordinate(d2) at (4.5,2.5);
  \coordinate(d3) at (4.5,3);
  \draw[<->, dashed] (d1) -- (d2) node [anchor=west] {\footnotesize{$\,\,\,\lambda d(z+x^\perp, y)$}} -- (d3);
  \coordinate(s) at (3, 1);
  \draw (s) node [anchor=north] {$S$};
\end{tikzpicture}
    \vspace*{-.3cm}
    \caption{A variant of Thales theorem.}
    \label{il:fig:proof-thales}
  \end{figure}
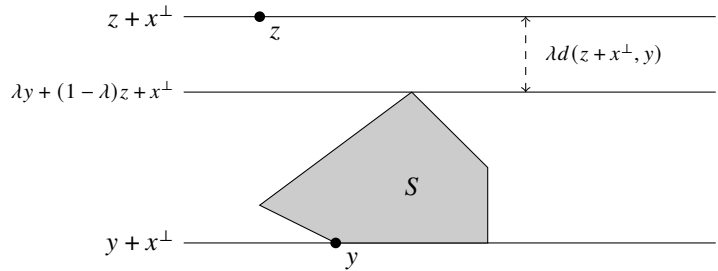
\end{remark}

\begin{remark}[Active labeling]
  When annotating data, as a partial labeler, you could ask yourself how to optimize your labeling.
  For example, suppose that you want to poll a population to retrieve preferences among a set of presidential candidates.
  Suppose that for a given polled person, you can only ask her to compare between four candidates.
  Which candidates would you ask her to compare?
  According to the questions you are asking, you will end up with different sets of potential weak distribution $\tau$.
  If aware of the problem $\ell$ that your dataset is intended to tackle, and aware of a constant $C = C(\ell, \tau)$ that verifies~\eqref{il:eq:calibration-bound}, you might want to design your questions in order to maximize on average over potential $\tau$, the quantity $C(\ell,\tau)$.
  An example where $\tau$ is not well-designed according to $\ell$ is given in Figure~\ref{il:fig:example-calibration}.

  \begin{figure}[t]
    \centering
    \begin{tikzpicture}[scale=3]
  \coordinate(a) at (0, 0);
  \coordinate(b) at ({cos(60)}, {sin(60)});
  \coordinate(c) at (1, 0);
  \coordinate(r) at ({1/2}, 0);
  \coordinate(s) at ({1/4}, {sin(60)/2});
  \coordinate(t) at ({3/4}, {sin(60)/2});
  
  \fill[fill=black!20] (c) -- (r) -- (s) -- (t) -- cycle;
  \draw (a) node[anchor=east]{a} -- (b) node[anchor=south east]{b} -- (c) node[anchor=north west]{c} -- cycle;
  \draw (c) -- (r) -- (s) -- (t) -- cycle;
  \node at ({5/8}, {sin(60)/3 - 1/16}) {$R_{\tau}$};
  \fill[fill=black!60] (1, 0) circle (.075em);

  \foreach \x in {0,.075,...,1} \draw[gray] ({\x}, 0) -- ({(1 + \x) / 2)}, {(1 - \x) * sin(60)});
  \foreach \x in {.25} \draw ({-\x / 2}, {- \x * sin(60)}) node[anchor=east] {$\ell_b^\perp$} -- ({(1 + \x) / 2}, {(1 + \x) * sin(60)});
\end{tikzpicture}
    \vspace*{-.3cm}
    \caption{Example of a bad link between $\tau$ and $\ell$.
      Same representation as Figure~\ref{il:fig:simplex-calibration} with a different instance where $\tau = \frac{1}{2} \delta_{\brace{a, c}} + \frac{1}{2}\delta_{\brace{b, c}}$ and $\ell(b, a) = 0$, $\ell(b, c) = 1$.
      In this example $C_\ell(\tau)=+\infty$, and the infimum loss is $0$ on $\Y$ and therefore not consistent.
      Given the loss structure, partial labeling acquisition should focus on specifying sets that do not intersect $\brace{a, b}$.
      Note that this instance violates the proper loss assumption, explaining its inconsistency.
    }
    \label{il:fig:example-calibration}
  \end{figure}
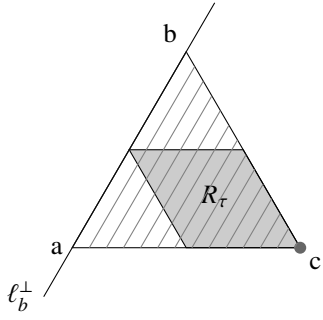
\end{remark}

\subsection{Proof of Theorems~\ref{il:thm:consistency} and \ref{il:thm:learning-rates}}
\label{il:proof:consistency}
First, note that, since ${\cal R}_S(f)$ is characterized by ${\cal R}_S(f) = \mathbb{E}_{(x,S)\sim\tau} \min_{u \in S} \ell(f(x), u)$, then the problem
\[
  f^* = \argmin_{f:\X \to \Y} {\cal R}_S(f) = \argmin_{f:\X \to \Y}
  \E_{(x,S)\sim\tau}\bracket{\min_{y \in S} \ell(f(x), y)}.
\]
can be considered as an instance of structured prediction with loss $L(z,S) = \min_{y \in S} \ell(f(x), y)$.
The framework for structured prediction presented in \citet{Ciliberto2016}, and extended in \citet{Ciliberto2020}, provides consistency and learning rates in terms of the excess risk ${\cal R}_S(f_n) - {\cal R}_S(f^*)$ when $f^*$ is estimated via $f_n$ defined as in~\eqref{il:eq:algorithm} and when the structured loss $L$ admits the decomposition
\[
  L(z,S) = \langle \psi(z), \phi(S) \rangle_{\cal H},
\]
for a separable Hilbert space ${\cal H}$ and two maps $\psi: \Y \to {\cal H}$ and $\phi: {\cal S} \to {\cal H}$.
Note that since $\Y$ is finite $L$ always admits the decomposition, indeed the cardinality of $\Y$ is finite, {\em i.e.}, $|\Y| < \infty$ and $|{\cal S}| = 2^{|\Y|}$.
Choose an ordering for the elements in $\Y$ and in ${\cal S}$ and denote them respectively $o_\Y:\N \to \Y$ and $o_{\cal S}:\N \to {\cal S}$.
Let $n_\Y: \Y \to \N$ be the inverse of $o_\Y$, {\em i.e.} $o_\Y(n_\Y(y)) = y$ and $n_\Y(o_\Y(i)) = i$ for $y \in \Y$ and $i \in {1,\dots,|\Y|}$, define analogously $n_{\cal S}$.
Now let ${\cal H} = \R^{|\Y|}$ and define the matrix $B \in \R^{|\Y| \times 2^{|\Y|}}$ with element $B_{i,j} = L(o_\Y(i), o_{\cal S}(j))$ for $i=1,\dots,|\Y|$ and $j=1,\dots,2^{|\Y|}$, then define
\[
  \psi(z) = e^{|\Y|}_{n_\Y(z)}, \quad \phi(S) = B e^{2^{|\Y|}}_{n_{\cal S}(S)},
\]
where $e^k_i$ is the $i$-th element of the canonical basis of $\R^k$.
We have that
\[
  \langle \psi(z), \phi(S) \rangle_{\cal H} = \langle e^{|\Y|}_{n_\Y(z)}, B
  e^{2^{|\Y|}}_{n_{\cal S}(S)} \rangle_{R^{|\Y|}} = B_{n_\Y(z), n_{\cal S}(S)} =
  L(i_\Y(n_\Y(z)), i_{\cal S}(n_{\cal S}(S))) = L(z,S),
\]
for any $z \in \Y, S \in {\cal S}$.
So we can apply Theorem 4 and 5 of \cite{Ciliberto2016} \citep[see also their extended forms in Theorem 4 and 5 of][]{Ciliberto2020}.
The last step is to connect the excess risk on ${\cal R}_S$ with the excess risk on ${\cal R}(f, \rho^\star)$, which is done by our comparison inequality in Theorem~\ref{il:thm:calibration}.
\begin{remark}[Illustrating the consistency in a discrete setting]
  Suppose that $\tau_{\vert x}$ has been approximate, as a signed measure $\hat{\tau}_{\vert x} = \sum_{i=1}^n \alpha_i(x)\delta_{S_i}$.
  After renormalization, one can represent it as a region $R_{\hat\tau_{\vert x}}$ in the affine span of $\Prob{\Y}$.
  Retaking the settings of Section~\ref{il:sec:inconsistency}, suppose that
  \[
    \hat{\tau}(\brace{a,b}) = \frac{1}{2},\qquad\hat{\tau}(\brace{c}) = \frac{1}{2},\qquad
    \hat{\tau}(\brace{a,c}) = \frac{1}{4},\qquad \hat{\tau}(\brace{a,b,c}) = -\frac{1}{4}.
  \]
  This corresponds to the region $R_{\hat\tau}$ represented in Figure~\ref{il:fig:consistency}.
  It leads to a disambiguation $\hat{\rho}$ that minimizes ${\cal E}$~\eqref{il:eq:infimum-disambiguation}, inside this space as
  \[
    \hat{\rho}(a) = \frac{1}{2}, \qquad \hat{\rho}(b) = -\frac{1}{4}, \qquad \hat{\rho}(c) = \frac{3}{4},
  \]
  and to the right prediction $\hat{z}=c$, since $\hat{\rho}$ felt in the decision region $R_c$.
  As the number of data augments, $R_{\hat\tau}$ converges toward $R_\tau$, so does $\hat{\rho}$ toward $\rho^\star$ and the risk ${\cal R}(\hat{f})$ toward its minimum.

  \begin{figure}[t]
    \centering
    \begin{tikzpicture}[scale=3]
  \coordinate(a) at (0, 0);
  \coordinate(b) at ({cos(60)}, {sin(60)});
  \coordinate(c) at (1, 0);
  \coordinate(r) at ({1/4}, 0);
  \coordinate(s) at ({7/32 + 1/8}, {7*sin(60)/16});
  \coordinate(t) at ({3/8 + 1/4}, {3*sin(60)/4});

  \coordinate(ra) at ({1/4},0);
  \coordinate(rb) at ({3/8},{-sin(60)/4});
  \coordinate(rc) at ({5/8},{-sin(60)/4});
  \coordinate(rd) at (1,{sin(60)/2});
  \coordinate(re) at ({1/2},{sin(60)/2});
  
  \fill[fill=black!10] (c) -- (r) -- (s) -- (t) -- cycle;
  \fill[fill=black!30] (ra) -- (rb) -- (rc) -- (rd) -- (re) -- cycle;
  \draw (a) node[anchor=north east]{a} -- (b) node[anchor=south]{b} -- (c) node[anchor=south west]{c} -- cycle;
  \draw (c) -- (r) -- (s) -- (t) -- cycle;
  \draw (ra) -- (rb) -- (rc) -- (rd) -- (re) -- cycle;
  \node at ({5/8}, {sin(60)/2 + 1/16}) {$R_{\tau}$};
  \node at ({9/16}, {-sin(60)/4 + 1/8}) {$R_{\hat{\tau}}$};
  \fill[fill=red!100] (1, 0) circle (.1em) node[anchor=north west] {$\rho^\star$};
  \fill[fill=red!100] ({5/8}, {-sin(60)/4}) circle (.1em) node[anchor=north west] {$\hat\rho$};

  \foreach \x in {0,.05,...,.6}
  \draw[gray, very thin, rotate=30] (1.25, \x) -- ({sin(60) - (tan(60)*\x)}, \x) -- ({sin(60) - (tan(60)*\x)}, -\x) -- ({1.25}, {-\x}); 

  \draw[dashed, rotate=30] (1.25, 0) -- ({sin(60)}, 0) -- ({sin(60) - (tan(60)*.55)}, .55); 
  \draw[dashed, rotate=30] ({sin(60)}, 0) --  ({sin(60) - (tan(60)*.55)}, -.55); 
\end{tikzpicture}
    \vspace*{-.3cm}
    \caption{Understanding convergence of the algorithm~\eqref{il:eq:algorithm}.
      Our method is approximating $\tau$ as a signed measured $\hat\tau$, which leads to $R_{\hat{\tau}}$ in dark gray compared to the ground truth $R_{\tau}$ in light gray.
      The disambiguation of $\hat\rho$ and $\rho^\star$ is done on those two domains with the same objective ${\cal E}$~\eqref{il:eq:infimum-disambiguation}, which level curves are represented with light lines.}
    \label{il:fig:consistency}
  \end{figure}
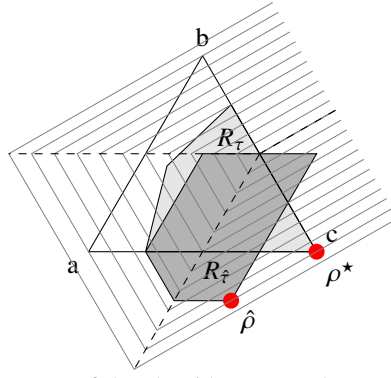
\end{remark}

\subsection{Understanding of the average and the supremum loss}
\label{il:app:other-losses}

For the average loss, if there is discrepancy in the loss $\nu > 0$, then there exists $a, b, c$ such that $\ell(b, c) = (1+\epsilon) \ell(a, b)$, for some $\epsilon > 0$.
In this case, one can recreate the example of Section~\ref{il:sec:inconsistency} by considering $\rho_0 = \rho^\star = \delta_c$ and
\[
  \tau = \lambda \delta_{\brace{c}} + (1 - \lambda) \delta_{\brace{a, b, c}},
  \qquad\text{with}\qquad
  \lambda = \frac{1}{2}\frac{\epsilon}{3\ell(a, b) + \epsilon},
\]
to show the inconsistency of the average loss.
Similarly, supposing, without loss of generality that $\ell(a, c) \in \bracket{\ell(a, b), \ell(b, c)}$, the case where $\rho_0 = \rho^\star = \delta_b$ and
\[
  \tau = \lambda \delta_{\brace{b}} + (1 - \lambda) \delta_{\brace{a, b, c}},
  \qquad \text{with}\qquad
  \lambda = \frac{1}{2} \min\paren{\frac{\epsilon}{1+\epsilon},
    \frac{1+\epsilon - x}{2+\epsilon - x}},\qquad x=\frac{\ell(a,c)}{\ell(a, b)},
\]
will fail the supremum loss, which will recover $z^* = a$, instead of $z^* = b$.
  \section{Experiments}
\label{il:app:experiments}

\subsection{Classification}\label{il:app:classification}
Let us consider the classification setting of Section~\ref{il:sec:classification}.
The infimum loss reads $L(z, S) = \ind{z\notin S}$.
Given a weak distribution $\tau$, the infimum loss is therefore solving for
\[
  f(x) \in \argmin_{z\in\Y} \E_{S\sim\tau\vert_x}\bracket{L(z, S)}
  = \argmin_{z\in\Y} \E_{S\sim\tau\vert_x}\bracket{\ind{z\notin S}}
  = \argmin_{z\in\Y} \PP_{S\sim\tau\vert_x}(z\notin S)
  = \argmax_{z\in\Y} \PP_{S\sim\tau\vert_x}(z\in S).
\]
Given data, $(z_i, S_i)$ our estimator consists in approximating the conditional distributions $\tau\vert_x$ as
\[
  \hat\tau\vert_x = \sum_{i=1}^n \alpha_i(x) \delta_{S_i},
\]
from which we deduce the inference formula, that we could also be derived from~\eqref{il:eq:algorithm},
\[
  \hat{f}(x) \in \argmax_{z\in\Y} \sum_{i=1}^n \alpha_i(x) \ind{z\in S_i}
  = \argmax_{z\in\Y} \sum_{i; z\in S_i} \alpha_i(x).
\]

\subsubsection{Complexity analysis}
The complexity of our algorithm~\eqref{il:eq:algorithm} can be split in two parts:
\begin{itemize}
  \item a training part, where given $(x_i, S_i)$ we precompute quantities that will be useful at inference.
  \item an inference part, where given a new $x$, we compute the corresponding prediction $\hat{f}(x)$.
\end{itemize}
In the following, we will review the time and space complexity of both parts.
We give this complexity in terms of $n$ the number of data and $m$ the number of items in $\Y$.
Results are summed up in Table~\ref{il:tab:cl:complexity}.
\begin{table}[t]
  \caption{Complexity of our algorithm for classification.}
  \label{il:tab:cl:complexity}
  \vskip 0.3cm
  \centering
  {\small\sc
    \begin{tabular}{lcc}
      \toprule
      Complexity & Time                 & Space              \\
      \midrule
      Training   & ${\cal O}(n^2(n+m))$ & ${\cal O}(n(n+m))$ \\
      Inference  & ${\cal O}(nm)$       & ${\cal O}(n+m)$    \\
      \bottomrule
    \end{tabular}
  }
\end{table}
\paragraph{Training.}
Let us suppose that computing $L(y, S) = \ind{y\notin S}$ can be done in a constant cost that does not depend on $m$.
We first compute the following matrices in ${\cal O}(nm)$ and ${\cal O}(n^2)$ in time and space.
\[
  L = (L(y, S_i))_{i\leq n, y\in\Y} \in \R^{n\times m},\qquad
  K_\lambda = (k(x_i, x_j) + n\lambda\delta_{i=j})_{ij} \in \R^{n\times n}.
\]
We then solve the following, based on the {\tt \_gesv} routine of Lapack, in ${\cal O}(n^3 + n^2m)$ in time and ${\cal O}(n(n+m))$ in space \citep[see][for details]{Golub1996}
\[
  \beta = K_\lambda^{-1}L \in \R^{n\times m}.
\]

\paragraph{Inference.}
At inference, we first compute in ${\cal O}(n)$ in both time and space
\[
  v(x) = (k(x, x_i))_{i\leq n} \in \R^n.
\]
Then we do the following multiplication in ${\cal O}(nm)$ in time and ${\cal O}(m)$ in space,
\[
  {\cal R}_{S, x} = v(x)^T \beta \in \R^m.
\]
Finally, we take the minimum of ${\cal R}_{S, x}(z)$ over $z$ in ${\cal O}(m)$ in time and ${\cal O}(1)$ in space.

\subsubsection{Baselines}
The average loss is really similar to the infimum loss, it reads
\[
  L_{\textit{ac}}(z, S) = \frac{1}{\module{S}}\sum_{y\in S} \ell(z, y) = 1 -
  \frac{\ind{z\in S}}{\module{S}} \simeq \frac{1}{\module{S}}\cdot\ind{z\notin S} =
  \frac{1}{\module{S}} L(z, S).
\]
Following similar derivations to the one for the infimum loss, given a distribution $\tau$, one can show that the average loss is solving for
\[
  f_{\textit{ac}}(x) \in \argmax_{z\in\Y} \sum_{S; z\in S}
  \frac{1}{\module{S}}\tau\vert_x(S),
\]
which is consistent when $\tau$ is not ambiguous.
The difference with the infimum loss is due to the term in $\module{S}$.
It can be understood as an evidence weight, giving less importance to big sets that do not allow discriminating efficiently between candidates.
Given data $(x_i, S_i)$, it leads to the estimator
\[
  \hat{f}_{\textit{ac}}(x) \in \argmin_{z\in\Y} \sum_{i; z\in S_i} \frac{\alpha_i(x)}{\module{S_i}}.
\]
The supremum loss is really conservative since
\[
  L_{\textit{sp}}(z, S) = \sup_{y\in S} \ell(y, z) = \sup_{y\in S} \ind{y\neq z}
  = \ind{S\neq\brace{z}}.
\]
It is solving for
\[
  f(x) \in \argmax_{z\in\Y} \tau\vert_x(\brace{z}),
\]
which empirically correspond to discarding all the set with more than one element
\[
  \hat{f}_{\textit{sp}}(x) \in \argmin_{z\in\Y} \sum_{i; S_i = \brace{z}} \alpha_i(x).
\]
Note that $\tau$ could be not ambiguous while charging no singleton, in this case, the supremum loss is not informative, as its risk is the same for any prediction.

\subsubsection{Corruptions on the \emph{LIBSVM} datasets}\label{il:sec:libsvm}

\begin{figure}[t]
  \centering
  \includegraphics[width=.45\textwidth]{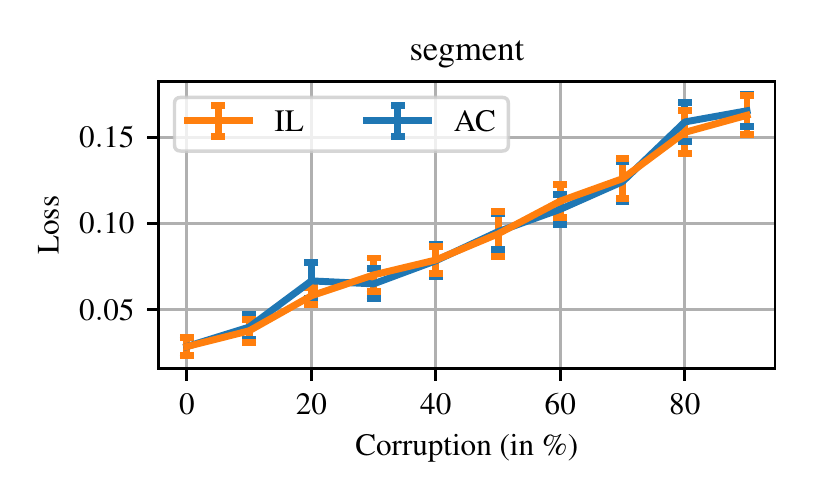}
  \includegraphics[width=.45\textwidth]{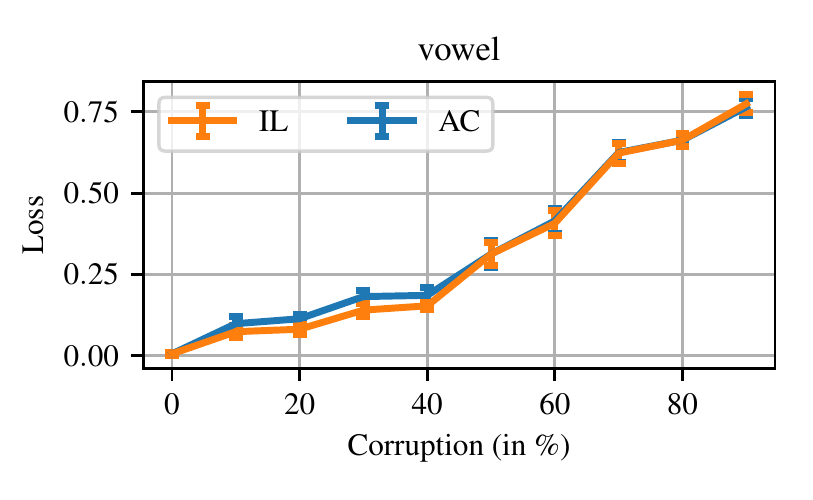}
  \vspace*{-.3cm}
  \caption{Classification.
    Testing risks \eqref{il:eq:risk} achieved by {\em AC} and {\em IL} on the ``segment'' and ``vowel'' datasets from {\em LIBSVM} as a function of corruption parameter $c$, when the corruption is uniform, as described in Section~\ref{il:sec:libsvm}.}
  \label{il:fig:cl:uniform}
\end{figure}

To illustrate the dynamic of our method versus the average baseline, we used {\em LIBSVM} datasets~\citep{Chang2011}, that we corrupted by artificially adding false class candidates to transform fully supervised pairs $(x, y)$ into weakly supervised ones $(x, S)$.
We experiment with two types of corruption processes.
\begin{itemize}
  \item A uniform one, reading, with the $\mu$ of Definition~\ref{il:def:eligibility}, for $z\neq y$,
        \[
          \PP_{(Y, S)\sim\mu\vert_{\Y\times 2^\Y}}\paren{z\in S\midvert Y=y} = c.
        \]
        with $c$ a corruption parameter that we vary between zero and one.
        In this case, the average loss and the infimum one works the same as shown on Figure~\ref{il:fig:cl:uniform}.
  \item A skewed one, where we only corrupt the pair $(x, y)$ when $y$ is the most present class in the dataset.
        More exactly, if $y$ is the most present class in the dataset, for $z\in\Y$, and $z'\neq z$, our corruption process reads
        \[
          \PP_{(Y, S)\sim\mu\vert_{\Y\times 2^\Y}}\paren{z'\in S\midvert Y=z} = c\cdot\ind{z=y}.
        \]
        In unbalanced dataset, such as the ``DNA'' and ``svmguide2'' datasets, where the most present class represents more than fifty percent of the labels as shown Table~\ref{il:tab:cl:libsvm}, this allows fooling the average loss as shown Figure~\ref{il:fig:libsvm}.
        Indeed, this corruption was designed to fool the average loss since we knew of the evidence weight $\frac{1}{\module{S}}$ appearing in its solution.
\end{itemize}
\begin{table}[t]
  \caption{{\em LIBSVM} datasets characteristics, showing the number of data, of classes, of input features, and the proportion of the most present class when labels are unbalanced.}
  \label{il:tab:cl:libsvm}
  \vskip 0.3cm
  \centering
  {\small\sc
    \begin{tabular}{lccccc}
      \toprule
      Dataset   & Data ($n$) & Classes ($m$) & Features ($d$) & Balanced     & Most present \\
      \midrule
      DNA       & 2000       & 3             & 180            & $\times$     & 52.6\%       \\
      Svmguide2 & 391        & 3             & 20             & $\times$     & 56.5\%       \\
      Segment   & 2310       & 7             & 19             & $\checkmark$ & -            \\
      Vowel     & 528        & 11            & 10             & $\checkmark$ & -            \\
      \bottomrule
    \end{tabular}
  }
\end{table}

\subsubsection{Reproducibility specifications}
All experiments were run with {\em Python}, based on the {\em NumPy} library.
Randomness was controlled by instantiating the random seed of {\em NumPy} to $0$ before doing any computations.
Results of Figures~\ref{il:fig:libsvm} and \ref{il:fig:cl:uniform} were computed by using eight folds, and trying out several hyperparameters, before keeping the set of hyperparameters that hold the lowest mean error over the eight folds.
Because we used a Gaussian kernel, there were two hyperparameters, the Gaussian kernel parameter $\sigma$, and the regularization parameter $\lambda$.
We search for the best hyperparameters based on the heuristic
\[
  \sigma = c_\sigma d,\qquad \lambda = c_\lambda n^{-1/2},
\]
where $d$ is the dimension of the input $\X$ (or the number of features), and where the Gaussian kernel reads
\[
  k(x, x') = \exp\paren{-\frac{\norm{x-x'}^2}{2\sigma^2}}.
\]
We tried $c_\sigma \in \brace{10, 5, 1, .5, .1, .01}$ and $c_\lambda \in\brace{10^i\midvert i\in\bbracket{3, -3}}$.

\subsection{Ranking}
Consider the ranking setting of Section~\ref{il:sec:ranking}, where $\Y = \Sfrak_m$, $\phi$ is the Kendall's embedding and the loss is equivalent to $\ell(z, y) = - \phi(y)^T\phi(z)$.

\subsubsection{Complexity analysis}
Given data $(x_i, S_i)$, our algorithm is solving at inference for
\[
  f(x) \in \argmin_{z\in\Y}\inf_{y_i \in S_i} - \sum_{i=1}^n \alpha_i(x)
  \phi(z)^T\phi(y_i)
  = \argmax_{z\in\Y}\sup_{y_i \in S_i} \sum_{i=1}^n \alpha_i(x) \phi(z)^T\phi(y_i)
\]
We solved it through alternate minimization, by iteratively solving in $z$ for
\[
  \phi(z)^{(t+1)} = \argmax_{\xi \in \phi(\Y)} \scap{\xi}{ \sum_{i=1}^n \alpha_i(x) \phi(y_i)^{(t)}},
\]
and solving for each $y_i$ for
\[
  \phi(y_i)^{(t+1)} = \argmax_{\xi\in\phi(S_i)} \alpha_i(x)\scap{\xi}{\phi(z)}.
\]
We initialize the problem with the coordinates of $\phi(y_i)$ put to 0 when not specified by the constraint $y_i \in S_i$.\footnote{%
  Coordinates of the Kendall's embedding correspond to pairwise comparison between two items $j$ and $k$, so we put to 0 the coordinates for which we can not infer preferences from $S$
  between items $j$ and $k$.
}
Those two problems are minimum feedback arc set problems, that are \textit{NP}-hard in $m$, meaning that one has to check for all potential solutions, and there is $m!$ of them, which is the cardinal of $\Sfrak_m$.
We suggest solving them using an integer linear programming (ILP) formulation that we relax into linear programming as explained in Appendix~\ref{il:app:fas}.
All the problems in $y_i$ share the same objective, up to a change in sign, but different constraint $\xi\in\phi(S_i)$, such a setting is particularly suited for warm start on the dual simplex algorithm to solve efficiently one after the other the linear programs associated to each $y_i$.

To give numbers, at training time, we compute the inverse $K_\lambda^{-1}$ in ${\cal O}(n^3)$ in time and ${\cal O}(n^2)$ in space, and at inference we compute $\alpha(x) K_\lambda^{-1}v(x)$ in ${\cal O}(n^2)$ in time and ${\cal O}(n)$ in space, before solving iteratively $n$ \textit{NP}-hard problem in $m$ of complexity $n \textit{NP}(m)$, that cost $nm^2$ in space to represent using {\em Cplex}~\citep{Cplex}, if we allow our self $e$ iterations, the inference complexity is ${\cal O}(n^2 + e\,n\,\textit{NP}(m))$ in time and ${\cal O}(nm^2)$ in space.

\subsubsection{Baselines}
The supremum loss is really similar to the infimum loss, only changing an infimum by a supremum.
However, algorithmically, this change leads to solving for a local saddle point rather than solving for a local minimum.
While the latter are always defined, there might be instances where no saddle point exists.
In this case, the supremum optimization might stall without getting to any stable solution, and the user might consider stopping the optimization after a certain number of iteration and outputting the current state as a solution.
The average loss, despite its simple formulation, does not lead to an easy implementation either.
Indeed, when given a set $S$, the average loss is implicitly computing the center of this set $c(S)$, and replacing $L_{\textit{ac}}(z, S)$ by $\ell(z, c(S))$, more exactly
\[
  L_{\textit{ac}}(z, S) \simeq - \frac{1}{\module{S}} \sum_{y\in S} \phi(z)^T\phi(y)
  = -\phi(y)^T\paren{\frac{1}{\module{S}}\sum_{y\in S}\phi(y)}.
\]
To compute the center $\paren{\frac{1}{\module{S}}\sum_{y\in S}\phi(y)}$, we sample $c_k \sim {\cal N}(0, I_{m^2})$, solve the resulting minimum feedback arc set problem, with the constraint $y\in S$, and end up with solutions $\phi(y_k)$.
After removing duplicates, we estimate the average with the empirical one.
Note that this work is done at training, leading the average loss to have a quite good inference complexity in ${\cal O}(nm + \textit{NP}(m))$ in time.

\subsubsection{Synthetic example: ordering lines}
In the following, we explain our synthetic example of Section~\ref{il:sec:ranking}.
It corresponds of choosing $\X = [0, 1]$, choose $m$ a number of items, simulate $a, b \sim {\cal N}(0, I_m)$, compute scores $v_i(x) = ax + b$, and order items according to their scores as shown on Figure~\ref{il:fig:rk:setting}.
For Figure~\ref{il:fig:rk:corruption}, we chose $m=10$, as this is the biggest $m$ for which can rely on our minimum feedback arc set heuristic to recover the real minimum feedback arc set solution and therefore not to play a role in what our algorithm will output.
The corruption process was defined as losing coordinates in the Kendall's embedding, more exactly given a point $x\in\X$, we have a score $(v_i(x))_{i\leq m}$ and an ordering $y\in\Y$.
To create a skewed corruption, we first compute the normalized distance between scores as
\[
  d_{ij} = \frac{\module{v_i - v_j}}{\max_{k,l} \module{v_k - v_l}} \in [0, 1]
\]
and remove the pairwise comparison for which $d_{ij} > c$, where $c$ is a
corruption parameter between 0 and 1, formally
\[
  S = \brace{z \in \Y\midvert \forall\, (j,k) \in I,\ \phi(z)_{jk} = \phi(y)_{jk}},\qquad\text{where}\qquad
  I = \brace{(j,k) \midvert d_{(j, k)} < c },
\]
Because of the transitivity constraint, when $c$ is small the comparison that we lost can be found back using transitivity between comparisons.
\begin{figure}[t]
  \centering
  \includegraphics[width=\linewidth]{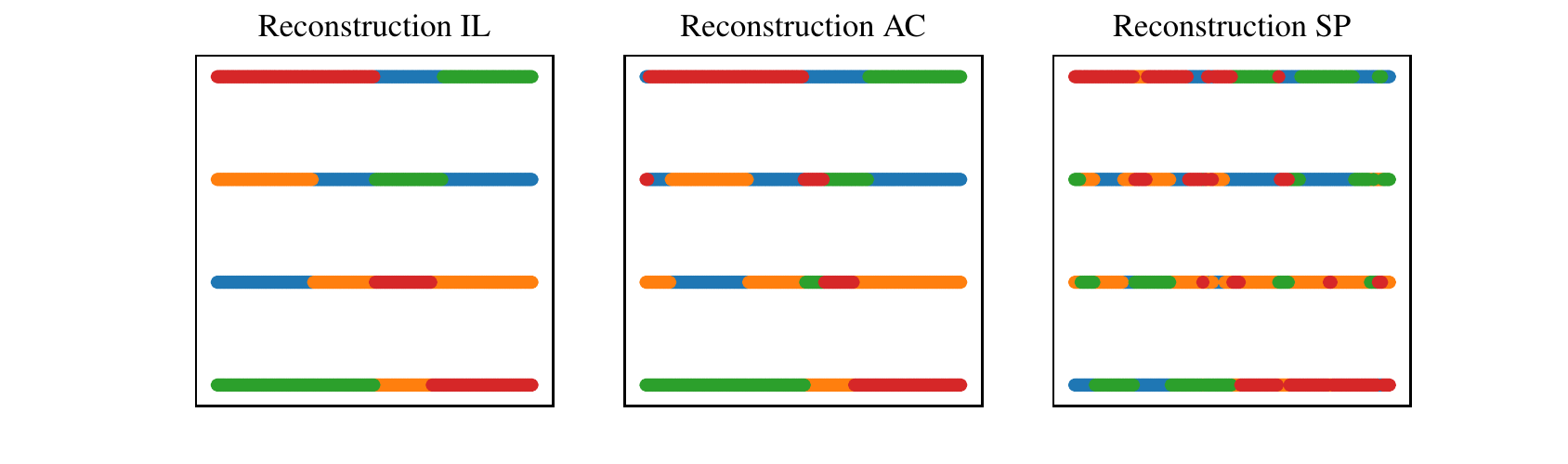}
  \vspace*{-.3cm}
  \caption{Reconstruction of the problem of Figure~\ref{il:fig:rk:setting}, given $n=50$ random points $(x_i, y_i)_{i\leq n}$, after losing at random fifty percent of the coordinates $(\phi(y_i))_{i\leq n}$, leading to sets $(S_i)_{i\leq n}$ of potential candidates.
  Hyperparameter were chosen as $\sigma = 1$ for the Gaussian kernel and $\lambda = 10^{-3}n^{-1/2}$ for the regularization parameter.
  The percentage of error in the reconstructed Kendall's embedding is 3\% for {\em IL}, 4\% for {\em AC} and 13\% for {\em SP}.
  As for classification, with such a random corruption process, {\em AC} and {\em IL} show similar behaviors.}
  \label{il:fig:ranking_reconstruction}
\end{figure}

\subsubsection{Reproducibility specification}
To get Figure~\ref{il:fig:rk:corruption}, we generate eight problems that correspond to ordering $m=10$ lines, seen as eight folds.
We only cross validated results with the same heuristics as in Appendix~\ref{il:app:classification}, yet, because computations were expensive we only tried $c_\sigma \in \brace{1, .5}$, and $c_\lambda \in \brace{10^3, 1, 10^{-3}}$.
Again, randomness was controlled by instantiating random seeds to 0.
Solving the linear program behind our minimum feedback arc set was done using {\em Cplex} \citep{Cplex}, which is the fastest linear program solver we are aware of.

\subsection{Multilabel}
Multilabel is another application of partial labeling that we did not mention in our experiment section in the core paper.
This omission was motivated by the fact that, under natural weak supervision, the three losses (infimum, average and supremum) are basically the same.
However, we will provide, now, an explanation of this problem and our algorithm to solve it.

Multilabel prediction consists in finding which are the relevant tags (possibly more than one) among $m$ potential tags.
In this case, one can represent $\Y = \brace{-1, 1}^m$, with $y_i = 1$ (resp.~$y_i = -1$), meaning that tag $i$ is relevant (resp.~not relevant).
The classical loss is the Hamming loss, which is the decoupled sum of errors for each label:
\[
  \ell(y, z) = \sum_{i=1}^m \ind{y_i\neq z_i}.
\]
Natural weak supervision consists in mentioning only a few relevant or irrelevant tags.
This is the setting of~\citet{Yu2014}.
This leads to sets $S$ that are built from a set $P$ of relevant items, and a set $N$ of irrelevant items.
\[
  S = \brace{y\in \Y \midvert \forall\,i\in P, y_i = 1, \forall\,i\in N, y_i = -1}.
\]
In this case, the infimum loss reads,
\[
  L(z, S) = \sum_{i\in P} \ind{z_i = -1} + \sum_{i\in N} \ind{z_i = 1}.
\]
For such supervision, the infimum, the average and the supremum loss are intrinsically the same, they only differs by constants, due to the fact that for each unseen labels, the infimum loss pays $0$, the average loss $1/2$ and the supremum loss $1$.

When considering data $(x_i, S_i)_{i\leq n}$, where $(S_i)$ is built from $(N_i, P_i)$, our algorithm in~\eqref{il:eq:algorithm} reads $\hat{f}(x) = (\sign(\hat{f}_j(x)))_{j\leq m}$, based on the scores
\[
  \hat{f}_j(x) = \sum_{i;j\in P_i} \alpha_i(x) - \sum_{i;j\in N_i} \alpha_i(x).
\]

\subsubsection{Tackling positive bias.}
In the precedent development, we implicitly assumed that the ratio between positive and negative labels given by the weak supervision reflects the one of the full distribution.
An assumption that is often violated in practice.
It is common that partial labeling only mentions a subset of the relevant tags ({\em i.e.}, $N = \emptyset$).
This case is ill-conditioned as always outputting all tags ($y = \textbf{1}$) will minimize the infimum loss.
To solve this problem, we can constrain the prediction space to the top-$k$ space $\Y_{k} = \brace{y\in\Y\midvert \sum_{i=1}^m \ind{y_j = 1} = k}$, which will lead to taking the top-$k$ over the score $(\hat{f}_j)_{j\leq m}$.
We can also break the loss symmetry and add a penalization with $\epsilon >0$,
\[
  \ell_\epsilon(z, y) = \ell(z, y) + \epsilon \sum_{i=1}^m \ind{z_i=1}.
\]
In this case, the inference algorithm will threshold scores at $\epsilon$ rather than $0$.

\subsubsection{Complexity analysis}
The complexity analysis is similar to the one for classification.
At training, we compute $L = (\ind{j\in P_i} - \ind{j \in N_i})$, and we solve for $\beta = K_\lambda^{-1}L$ in $\R^{n\times m}$.
At testing, we compute $v(x)$ and $\beta^T v(x)$ in $\R^m$, before thresholding it or taking the top-$k$ in either ${\cal O}(m)$ or ${\cal O}(m\log(m))$.
As such, complexity reads similarly as for the classification case.
Yet notice that, for multilabeling, the dimension of $\Y$ is not $m$ but $2^m$, meaning we do not scale with $\#\Y$ but with the intrinsic dimension.
\begin{table}[t]
  \caption{Complexity of our algorithm for multilabels.}
  \label{il:tab:ml:complexity}
  \vskip 0.3cm
  \centering
  {\small\sc
    \begin{tabular}{lcc}
      \toprule
      Complexity        & Time                      & Space              \\
      \midrule
      Training          & ${\cal O}(n^2(n+m))$      & ${\cal O}(n(n+m))$ \\
      Inference         & ${\cal O}(nm)$            & ${\cal O}(n+m)$    \\
      Inference top-$k$ & ${\cal O}(nm + m\log(m))$ & ${\cal O}(n+m)$    \\
      \bottomrule
    \end{tabular}
  }
\end{table}

\subsubsection{Corruptions on the \emph{MULAN} datasets}\label{il:sec:mulan}

\begin{figure}[t]
  \centering
  \includegraphics{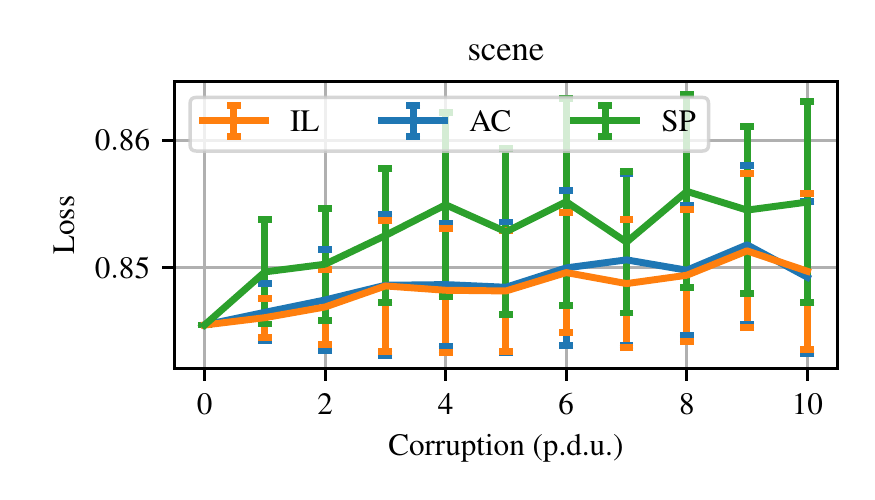}
  \vspace*{-.3cm}
  \caption{Multilabeling.
    Testing risks (from~\eqref{il:eq:risk}) achieved by {\em AC} and {\em IL} on the ``scene'' dataset from MULAN as a function of corruption parameter c, shown in procedure defined unit, when the supervision is given as Hamming balls, as described in Section~\ref{il:sec:mulan}.}
  \label{il:fig:mulan}
\end{figure}

When sets are given by few positive and negative tags, all losses are the same.
Yet, under other types of supervision, such as when the sets come as Hamming balls, defined by
\[
  B(z, r) = \brace{y\in \Y \midvert \ell(z, y) \leq r},
\]
the methods will not behave the same.
We experiment on MULAN datasets provided by~\citet{Tsoumakas2011}.
Because supervision with Hamming balls does not lead to efficient implementation, we went for extensive grid search for the best solution, which reduces our ability to consider large $m$.
Among MULAN datasets, we went for the ``scene'' one, with $m=6$ tags, and $n=2407$ data.
When given a pair $(x, y)$, we add corruption on $y$, by first sampling a radius parameter $r\sim\uniform([0, c*(m+1)])$, with $c$ a corruption parameter.
We then sample, with replacement, $\floor{r}$ coordinates to modify to pass from $y$ to a center $c$.
We then consider the supervision $S = B(c, r)$.
For such randomness, somehow uniform corruption, the infimum loss works slightly better than the average loss that both outperform the supremum loss as shown on Figure~\ref{il:fig:mulan}.

\subsubsection{Reproducibility specification}
To get Figure~\ref{il:fig:mulan}, we follow the same cross-validation scheme as for classification and ranking.
More exactly, we cross-validated over eight folds with the same heuristics for $\sigma$, the Gaussian kernel parameter, and $\lambda$, the regularization one, with $c_\sigma \in \brace{10, 5, 1, .5, .1, .01}$, and $c_\lambda \in \brace{10^i\midvert i\in\bbracket{-3, 3}}$.

\subsection{Partial regression}
Partial regression is the regression instance of partial labeling.
When supervision comes as intervals, it is known as interval regression, and known as censored regression when sets come as half-lines.
Note that for censored regression, neither the average, nor the supremum loss can be properly defined.

\subsubsection{Baselines}
Given a bounded set $S$, learning with the average loss corresponds to considering the center of this set, since, for $z\in\Y$, with $\lambda$ the Lebesgue measure
\begin{align*}
  L_{\textit{ac}}(z, S) & = \frac{1}{\lambda(S)}\int_{S} \norm{z - y}^2 \lambda(\diff y)
  = \norm{z}^2 - 2\scap{z}{\frac{1}{\lambda(S)}\int_{S} y \lambda(\diff y)} +
  \frac{1}{\lambda(S)}\int_{S} \norm{y}^2 \lambda(\diff y)
  \\&= \norm{z - \frac{1}{\lambda(S)}\int_{S} y \lambda(\diff y)}^2 + \frac{1}{\lambda(S)}\int_{S} \norm{y}^2 \lambda(\diff y) - \norm{\frac{1}{\lambda(S)}\int_{S} y \lambda(\diff y) }^2
  = \norm{z - c(S)}^2 + C_S,
\end{align*}
where $c(S) = \frac{1}{\lambda(S)}\int_{S} y \lambda(\diff y)$ is the center of $S$.
As such, the average loss is always convex.
As the supremum of convex function, the supremum loss is also convex.

\subsubsection{Reproducibility specification}
To compute Figure~\ref{il:fig:interval-regression}, for both {\em AC} and {\em IL}, we consider $\sigma$, the Gaussian kernel parameter, and $\lambda$, the regularization parameter, achieving the best risk when measured with the fully supervised distribution~\eqref{il:eq:risk}.
We tried over $\sigma \in \brace{1, .5, .1, .05, .01}$ and $\lambda \in \brace{10^3, 1, 10^{-3}}$.
Randomness was controlled by instantiating random seeds.

\subsection{Beyond}
Beyond the examples showcased previously, advances in dealing with weak supervision could be beneficial for several problems.
Supervision on \emph{image segmentation} problems usually comes as partial pixel annotation.
This problem is often tackled through conditional random fields~\citep{Verbeek2007}, making it a perfect mix between partial labeling and structured prediction.
\emph{Action retrieval} on instructional video, where partial supervision is retrieved from the audio track is another interesting application~\citep{Alayrac2018}.
  \section{Minimum feedback arc set}
\label{il:app:fas}

\subsection{Formulation}
Consider a directed weighted graph with vertices $\bbracket{1, m}$ and edges $\brace{i\rightarrow j}$ with weights $(w_{ij})_{i,j\leq m} \in \R_+^{m^2}$.
The goal is to find a directed acyclic graph $G = (V, E)$ that maximizes the weights on remaining edges
\[
  \argmax_{E} \sum_{i\rightarrow j \in E} w_{ij}.
\]
This directed acyclic graph can be seen as a preference graph, item $j$ being preferred over item $i$.
Since $w_{ij}$ are non-negative, the underlying ordering in $G$ is necessarily total, and therefore can be written based on a score function, that can be embedded in the permutation of $\bbracket{1, m}$, $\sigma \in \Sfrak_m$, with $\sigma(j) > \sigma(i)$ meaning that $j$ is preferred over $i$.
Thus the problem reads equivalently
\begin{align*}
  \argmax_{\sigma\in\Sfrak_m}\sum_{i,j \leq m} w_{ij} \ind{\sigma(j) > \sigma(i)}
   & = \argmax_{\sigma\in\Sfrak_m}\sum_{i<j \leq m} c_{ij} \ind{\sigma(j) > \sigma(i)}
  = \argmax_{\sigma\in\Sfrak_m}\sum_{i<j \leq m} c_{ij} \sign\paren{\sigma(j) - \sigma(i)}
  \\&= \argmin_{\sigma\in\Sfrak_m}\sum_{i<j \leq m} c_{ij} \sign\paren{\sigma(i) - \sigma(j)}
  = \argmin_{\sigma\in\Sfrak_m}\sum_{i<j \leq m} c_{ij} \ind{\sigma(i) > \sigma(j)}
\end{align*}
with $c_{ij} = w_{ij} - w_{ji}$.
This last formulation is the one usually encountered for ranking algorithms in machine learning~\citep{Duchi2010}.

We are going to study in depth this problem under the formulation
\begin{equation}\label{il:eq:fas}
  \argmin_{\sigma \in\Sfrak_m} \sum_{i<j\leq m} c_{ij} \sign\paren{\sigma(i) - \sigma(j)}
\end{equation}

\subsection{Integer linear programming}

\begin{definition}[Kendall's embedding]\label{il:def:ken}
  For $\sigma \in \Sfrak_m$, define Kendall's embedding, with $m_e =m(m-1) / 2$,
  \[
    \phi(\sigma) = \sign\paren{\sigma(i) - \sigma(j)}_{i<j \leq m} \in \brace{-1, 1}^{m_e}.
  \]
  We associate it to the Kendall's polytope of order $m$, $\hull\paren{\phi(\Sfrak_m)}$.
\end{definition}
The Kendall's embedding, Definition~\ref{il:def:ken}, cast the minimum feedback arcset problem~\eqref{il:eq:fas} as a linear program
\[
  \minimize{\scap{c}{x}}{x\in \hull\paren{\phi(\Sfrak_m)}.}
\]
Since the objective is linear, the solution is known to lie on a vertex of the constraint polytope, which is the set of Kendall's embeddings of permutations.
Yet, how to describe Kendall's polytope?

\begin{definition}[Transitivity polytope]\label{il:def:tr}
  The transitivity polytope of order $m$ is defined in $\R^{m_e}$ as
  \[
    {\cal M} = \brace{x\in\R^{m_e} \midvert \forall\,i<k<j; -1 \leq x_{ij} + x_{jk} - x_{ik} \leq 1 }
  \]
  This polytope encodes the transitivity constraints of Kendall's embeddings, Definition~\ref{il:def:ken}.
\end{definition}

The transitivity polytope, Definition~\ref{il:def:hull}, will be used to approximate Kendall's polytope based on the following property.

\begin{proposition}[Relaxed polytope]\label{il:prop:rel}
  The intersection between the transitivity polytope and the vertex of the hypercube is exactly the set of Kendall's embeddings of permutations.
  Mathematically
  \[
    \phi(\Sfrak_m) = {\cal M} \cap \brace{-1, 1}^{m_e}.
  \]
\end{proposition}
\begin{proof}
  First of all it is easy to show that $\phi(\Sfrak_m) \subset \brace{-1, 1}^{m_e}$, and that, $\phi(\Sfrak_m)\subset{\cal M}$.

  Let's now consider $x \in {\cal M} \cap\brace{-1,1}^{m_e}$.
  Let's associate to $x$ the symmetric embedding
  \[
    \tilde{x}_{ij} = \left\{
    \begin{array}{ccc}
      x_{ij}  & \text{if} & i < j \\
      0       & \text{if} & i = j \\
      -x_{ji} & \text{if} & j < i
    \end{array}
    \right.
  \]
  Let's consider the permutation $\sigma$ resulting from the ordering of $\sum_{k}\tilde{x}_{ik}$
  \[
    \sigma^{-1}(1) = \argmin_{i \in\bbracket{1,m}} \sum_{k = 1}^m \tilde{x}_{ik}\qquad\text{and}\qquad
    \sigma^{-1}(i) = \argmin_{i \in\bbracket{1,m}\setminus \sigma^{-1}\paren{\bbracket{1, i-1}} } \sum_{k=1}^m \tilde{x}_{ik}.
  \]
  Let's now show that $\phi(\sigma) = x$, or equivalently that $\tilde{\phi}(\sigma) = (\sign(\sigma(i) - \sigma(j)))_{i,j\leq m} = \tilde{x}$.
  First, one can show that $\tilde{x}$ verify the transitivity constraints
  \[
    \forall\, i, j, k \leq m, \qquad -1 \leq \tilde{x}_{ij} + \tilde{x}_{jk} - \tilde{x}_{ik} \leq 1.
  \]
  This can be proven for any ordering of $i, j, k$ based on the fact that $x \in {\cal M}$.
  For example, if $i < k < j$, we have
  \[
    \bracket{-1, 1} \ni x_{ik} + x_{kj} - x_{ij} = \tilde{x}_{ik} - \tilde{x}_{jk} - \tilde{x}_{ij}.
  \]
  which leads to
  \[
    \tilde{x}_{ij} + \tilde{x}_{jk} - \tilde{x}_{ik} \in -\bracket{-1, 1} = \bracket{-1, 1}.
  \]
  Now suppose, without loss of generality, that $\tilde{x}_{ij} = 1$ (if $\tilde{x}_{ij} = -1$, just consider $\tilde{x}_{ji} = 1$).
  The transitivity constraints tell us that $\tilde{x}_{ik} \geq \tilde{x}_{jk}$ for all $k$, therefore
  \[
    \sum_{k\not\in\brace{i,j}} \tilde{x}_{ik} \geq \sum_{k\not\in\brace{i,j}} \tilde{x}_{jk},
    \qquad\Rightarrow\qquad
    \sum_{k=1}^{m} \tilde{x}_{ik} > \sum_{k=1}^{m} \tilde{x}_{jk}.
    \qquad\Rightarrow\qquad
    \sigma(i) > \sigma(j).
  \]
  This shows that $\tilde{\phi(\sigma)}_{ij} = 1 = \tilde{x}_{ij}$.
  Thus, we have shown that $x \in \phi(\Sfrak_m)$, which concludes the proof.
\end{proof}

\begin{figure}[t]
  \centering
  \includegraphics[width=.45\textwidth]{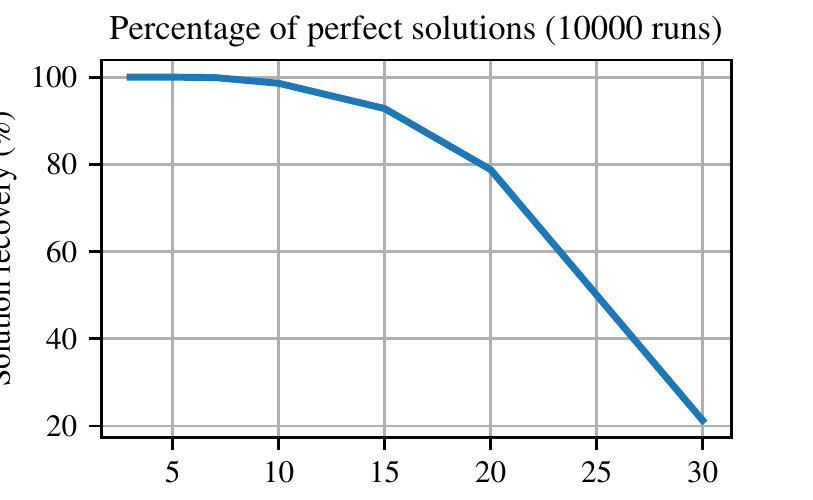}
  \includegraphics[width=.45\textwidth]{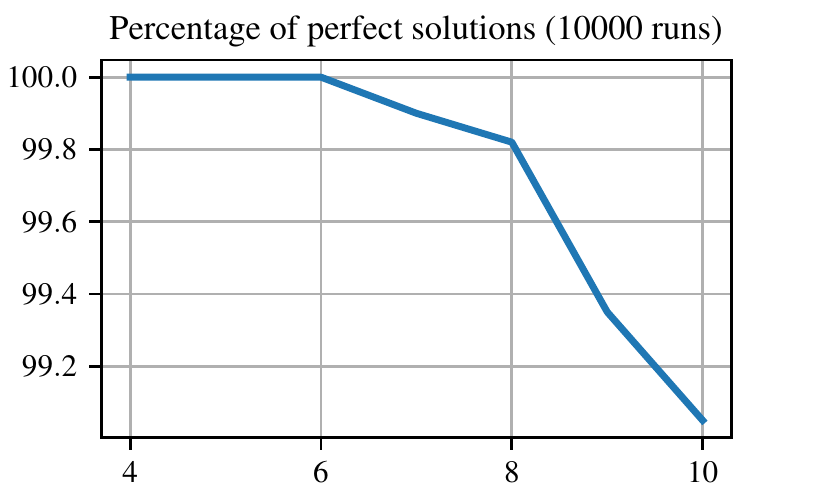}
  \vspace*{-.3cm}
  \caption{Evaluating the percentage of exact solutions of the ILP relaxation as $m$ grows large.
    Evaluation is done by choosing an objective $c\sim{\cal N}(0, I_{m_e})$, solving the ILP relaxation, Definition~\ref{il:def:hull}, and evaluating if the solution is in $\brace{-1, 1}^{m_e}$.
    The experience is repeated several times to estimate how often, on average, the original solution of~\eqref{il:eq:fas} is returned by the ILP.
  }
  \label{il:fig:ilp_rel}
\end{figure}

\begin{definition}[ILP relaxation]\label{il:def:hull}
  Based on Proposition~\ref{il:prop:rel}, we define the canonical polytope ${\cal C} = {\cal M} \cap \bracket{-1, 1}^{m_e}$, and relax the problem~\eqref{il:eq:fas} into
  \[
    \minimize{\scap{c}{x}}{x\in{\cal C}}
  \]
  As soon as the solution $x$ is in $\brace{-1, 1}^{m_e}$, Proposition~\ref{il:prop:rel} tells us that $x$ recover the exact minimum feedback arc set solution~\eqref{il:eq:fas}.
\end{definition}

\iftrue
  In small dimensions, the canonical polytope ${\cal C}$ is the same as the Kendall's one, and the ILP relaxation gives the right solution.
  Yet, as shown in Figure~\ref{il:fig:ilp_rel}, as soon as $m > 5$, there exists vertex in ${\cal C}$ that does not correspond to a permutation embedding.
  For small dimensions, proving that ${\cal C}$ is exactly the Kendall's polytope is done with a simple drawing for $m=3$, using unimodularity of the transitivity constraint matrix is enough for $m=4$~\citep{Hoffman2010}.
  The case $m=5$ is also provable, based on several tricks that we will not discuss here.
\else
  \begin{proposition}
    The ILP relaxation, Definition~\ref{il:def:hull}, always recovers the solution of~\eqref{il:eq:fas} if and only if $m \leq 5$.
  \end{proposition}
  \begin{proof}
    For $m > 5$, computation done to draw Figure~\ref{il:fig:ilp_rel} gives counter-examples, and small dimension counter-examples can be cast in bigger ones as $m$ grows.
    For $m=3$, one can draw the Kendall's polytope and the canonical one, they are the same.
    For $m=4$, let us use basic knowledge on integer linear program and unimodularity~\citep{Hoffman2010}.
    With $x = (x_{12}, x_{13}, x_{14}, x_{23}, x_{24}, x_{34})$, the transitivity constraints are built from
    \[
      A =
      \paren{\begin{array}{cccccc}
          1 & -1 & 0  & 1 & 0  & 0 \\
          1 & 0  & -1 & 0 & 1  & 0 \\
          0 & 1  & -1 & 0 & 0  & 1 \\
          0 & 0  & 0  & 1 & -1 & 1 \\
        \end{array}}
    \]
    More precisely, the entire constraints read $Ax \leq 1$, $-Ax\leq -1$, $x\leq 1$, $-x\leq 1$.

    The matrix $A$, and the one that build the transitivity constraints for $m=5$, is totally unimodular, meaning that any invertible sub-square matrix of $A$ is invertible in $\mathbb{Z}$, or equivalently, any sub-square matrix has determinant -1, 0 or 1.
    It is easy to show that $\tilde{A} = [A, -A, I, -I]^T$ is also totally unimodular.

    Consider a vertex $x$ in ${\cal C}$, it can be defined from $m_e$ linearly independent constraint hyperplanes, it solves an equation of the type $\tilde{A}_{\vert I} x = b$, with $b\in\brace{-1, 1}^{m_e}$.
    $\tilde{A}_{\vert I}$ being an invertible sub-square matrix of $\tilde{A}$, it is invertible in $\mathbb{Z}$, leading to $x = \tilde{A}_{\vert I}^{-1}b$ being integer.
    Therefore, $x\in\brace{-1, 0, 1}^{m_e}$.
    To show that $x\in\brace{-1, 1}^{m_e}$, let's consider $x' = (x + 1)/2$, it corresponds to the embedding $(\ind{\sigma(i) > \sigma(j)})_{ij}$.
    Solving the original problem with $x'$ or with $x$ is the same.
    The constraints for $x'$ reads $Ax' \leq 1$, $-Ax'\leq 0$, $x' \leq 1$, $-x'\leq 0$.
    Therefore, for the same reason as $x$, $x'$ is integer, more exactly $x'\in\brace{0, 1}^{m_e}$, from which $x$ is deduced to be in $\brace{-1,1}^{m_e}$.
    By showing that all vertices of ${\cal C}$ are in $\brace{-1, 1}^{m_e}$, using Proposition~\ref{il:prop:rel} we have shown that the ILP relaxation does give the exact solution for $m = 4$.

    For $m=5$, one can not use unimodularity as the transitivity constraint matrix shows sub-matrices of determinant $\pm 2$, for example
    \[
      \tilde{A} = \begin{blockarray}{cccccc}
        x_{45} & x_{34} & x_{25} & x_{13} & x_{12} \\
        \begin{block}{(ccccc)c}
          1 & 1 & 0 & 0 & 0 & x_{34} + x_{45} - x_{35} \\
          1 & 0 & -1 & 0 & 0 & x_{24} + x_{45} - x_{25} \\
          0 & 1 & 0 & 1 & 0 & x_{13} + x_{34} - x_{14} \\
          0 & 0 & 1 & 0 & 1 & x_{12} + x_{25} - x_{15} \\
          0 & 0 & 0 & -1 & 1 & x_{12} + x_{23} - x_{13} \\
        \end{block}
      \end{blockarray}
    \]
    This matrix is not invertible in $\mathbb{Z}$.
    The constraints and coefficients corresponding are written around it.

    Let's first prove that ${\cal C} \in \frac{1}{2}\mathbb{Z}$.
    Mainly we used that all sub-matrix of dimension $7\times 7$ are not invertible, the one of dimension $6\times 6$ are of determinant -1, 1, 0, and the one of $5\times 5$ are the only ones with determinant sometimes being $\pm 2$.
    This shows that the determinant of the matrix defining the constraint for a vertex is at most 2, by developing the determinant with respect to constraint of the type $x_{ij} = \pm 1$.
    Which shows that ${\cal C} \in \frac{1}{2}\mathbb{Z}$.
    With the same embedding trick, with $x' = (x+1)/2$, we show that $x\in\mathcal{Z}$.
    We know that the only potential trouble is when the matrix defining the constraints has a sub-factor of determinant $\pm 2$, and all the other constraint are of the type $x_{ij} = \pm 1$, meaning that there is at least 5 coordinates among the 10 non-zero.
    If we put those into the 5 transitivity constraints, if we put one of them in a coefficient that appear twice, then necessarily there will be one transitivity constraint that get two of its three coefficient fixed by the simple constraints, this constraint can so be replaced by a fixed one on the last coefficient, meaning that the vertex could also be defined by 4 transitivity constraint and 6 simple constraint, in this case it will be invertible.
    Meaning that we should distribute the simple constraint among the coefficients that appears only once.
    This links all the remaining coefficients between them in a really simple way, either they are all zeros, or all integers.
    If they are all zeros, one can check that they are not on the border by showing for example that $\max_{y\in{\cal C}}\scap{x}{y}$ is not only achieve for $y=x$, indeed any completing of the partial order given by the simple constraint will work
  \end{proof}
\fi

\begin{remark}[Low noise consistency]
  Remark that the low-noise setting considered by~\citet{Duchi2010} correspond to having $\sign(c) = -\phi(y)$ for a $y\in\Y$, in this case our algorithm is consistent and does recover the best solution $z=y$.
\end{remark}

\subsection{Sorting heuristics}
When formatting and solving the integer linear program takes too much time, one can go for a simple sorting heuristic, mainly based on a heuristic to compare items two by two and using quick sorting.
A review of some heuristic with guarantees is provided by~\citet{Ailon2005},
Similar study when in presence of constraints on the resulting total order can be found in~\citet{vanZuylen2007}.

\end{subappendices}
\chapter{Disambiguation Framework}

The following is a reproduction of \cite{Cabannes2021}.

Machine learning approached through supervised learning requires expensive annotation of data.  This motivates weakly supervised learning, where data are annotated with incomplete yet discriminative information.
In this paper, we focus on partial labeling, an instance of weak supervision where, from a given input, we are given a set of potential targets.
We review disambiguation principles to recover full supervision from weak supervision, and propose an empirical disambiguation algorithm. We prove exponential convergence rates of our algorithm under classical learnability assumptions, and we illustrate the usefulness of our method on practical examples.
\section{Introduction}

In many applications of machine learning, such as recommender systems, where an input $x$ characterizing a user should be matched with a target $y$ representing an ordering of a large number $m$ of items, accessing fully supervised data $(x, y)$ is not an option.
Instead, one should expect weak information on the target $y$, which could be a list of previously taken (if items are online courses), watched (if items are plays), {\em etc.}, items by a user characterized by the feature vector $x$.
This motivates {\em weakly supervised learning}, aiming at learning a mapping from inputs to targets in such a setting where tools from supervised learning can not be applied off-the-shelves.

Recent applications of weakly supervised learning showcase impressive results in solving complex tasks such as action retrieval on instructional videos \citep{Miech2019}, image semantic segmentation \citep{Papandreou2015}, salient object detection \citep{Wang2017}, 3D pose estimation \citep{Dabral2018}, text-to-speech synthesis \citep{Jia2018}, to name a few.
However, those applications of weakly supervised learning are usually based on clever heuristics, and theoretical foundations of learning from weakly supervised data are scarce, especially when compared to statistical learning literature on supervised learning \citep{Vapnik1995,Boucheron2005,Steinwart2008}.
We aim to provide a step in this direction.

In this paper, we focus on partial labeling, a popular instance of weak supervision, approached with a structured prediction point of view \cite{Ciliberto2020}.
We detail this setup in Section \ref{df:sec:setup}.
Our contributions are organized as follows.
\begin{itemize}
  \item In Section \ref{df:sec:algo}, we introduce a disambiguation algorithm to retrieve fully supervised samples from weakly supervised ones, before applying off-the-shelf supervised learning algorithms to the completed dataset.
  \item In Section \ref{df:sec:stat}, we prove exponential convergence rates of our algorithm, in terms of the fully supervised excess of risk, given classical learnability assumptions.
  \item In Section \ref{df:sec:optim}, we explain why disambiguation algorithms are intrinsically non-convex, and provide guidelines based on well-grounded heuristics to implement our algorithm.
\end{itemize}
We end this paper with a review of literature in Section \ref{df:sec:litterature}, before showcasing the usefulness of our method on practical examples in Section \ref{df:sec:example}, and opening on perspectives in Section \ref{df:sec:opening}.

\section{Disambiguation of partial labeling}
\label{df:sec:setup}

In this section, we review the supervised learning setup, introduce the partial labeling problem along with a principle to tackle this instance of weak supervision.

Algorithms can be formalized as mapping an input $x$ to a desired output $y$, respectively belonging to an input space~$\X$ and an output space $\Y$.
Machine learning consists in automating the design of the mapping $f:\X\to\Y$, based on a joint distribution $\mu \in \prob{\X\times\Y}$ over input/output pairings $(x, y)$ and a loss function $\ell:\Y\times\Y\to\R$, measuring the error cost of outputting $f(x)$ when one should have output $y$.
The optimal mapping is defined as satisfying
\begin{equation}
  \label{df:eq:sol_fs}
  f^* \in \argmin_{f:\X\to \Y} \E_{(X, Y) \sim \mu} \bracket{\ell(f(X), Y)}.
\end{equation}
In \emph{supervised learning}, it is assumed that one does not have access to the full distribution $\mu$, but only to independent samples $(X_i, Y_i)_{i\leq n} \sim \mu^{\otimes n}$.
In practice, accessing such samples means building a dataset of examples.
While input data $(x_i)$ are usually easily accessible, getting output pairings $(y_i)$ generally requires careful annotation, which is both time-consuming and expensive.
For example, in image classification, $(x_i)$ can be collected by scrapping images over the Internet.
Subsequently, a ``data labeler'' might be asked to recognize a rare feline $y_i$ on an image $x_i$.
While getting $y_i$ will be hard in this setting, recognizing that it is a feline and describing elements of color and shape is easy, and already helps to determine what outputs $f(x_i)$ are acceptable.
A second example is given when pooling a known population $(x_i)$ to get estimation of their political orientation $(y_i)$, one might get information from recent election of percentage of voters across the political landscape, leading to global constraints that $(y_i)$ should verify.
A supervision that gives information on $(y_i)_{i\leq n}$ without giving its precise value is called \emph{weak supervision}.

\emph{Partial labeling}, also known as ``superset learning'', is an instance of weak supervision, in which, for an input $x$, we do not access the precise label $y$ but only a set $s$ of potential labels, $y\in s\subset \Y$.
For example, on a caracal image $x$, one might not get the label ``caracal'' $y$, but the set $s$ ``feline'', containing all the labels $y$ corresponding to felines.
It is modelled through a distribution $\nu \in \prob{\X\times 2^\Y}$ over $\X\times 2^\Y$ generating samples $(X, S)$, which should be compatible with the fully supervised distribution $\mu \in \prob{\X\times\Y}$ as formalized by the following definition.

\begin{definition}[Compatibility, \citet{Cabannes2020}]
  \label{df:def:compatibility}
  A fully supervised distribution $\mu \in \prob{\X\times\Y}$ is \emph{compatible} with a weakly supervised distribution $\nu \in \prob{\X\times\Y}$, denoted by $\mu\vdash\nu$ if there exists an underlying distribution $\pi \in \prob{\X\times\Y\times 2^\Y}$, such that $\mu$, and $\nu$, are the respective marginal distributions of $\pi$ over $\X\times \Y$ and $\X\times 2^\Y$, and such that $y\in s$ for any tuple $(x, y, s)$ in the support of $\pi$ (or equivalently $\pi\vert_{s} \in \prob{s}$, with $\pi\vert_{s}$ denoting the conditional distribution of $\pi$ given $s$).
\end{definition}

This definition means that a weakly supervised sample $(X, S) \sim \nu$ can be thought as proceeding from a fully supervised sample $(X, Y) \sim \mu$ after losing information on $Y$ according to the sampling of $S\sim \pi\vert_{X, Y}$.
The goal of partial labeling is still to learn $f^*$ from \eqref{df:eq:sol_fs}, yet without accessing a fully supervised distribution $\mu \in \prob{\X\times\Y}$ but only the weakly supervised distribution $\nu \in \prob{\X\times 2^\Y}$.
As such, this is an ill-posed problem, since $\nu$ does not discriminate between all $\mu$ compatible with it.
Following {\em lex parsimoniae}, \citet{Cabannes2020} have suggested looking for $\mu$ such that the labels are the most deterministic function of the inputs, which they measure with a loss-based ``variance'', leading to the disambiguation
\begin{equation}
  \label{df:eq:principle}
  \mu^* \in \argmin_{\mu\vdash\nu} \inf_{f:\X\to\Y} \E_{(X, Y)\sim\mu}\bracket{\ell(f(X), Y)},
\end{equation}
and to the definition of the optimal mapping $f^*:\X\to\Y$
\begin{equation}
  \label{df:eq:solution}
  f^* \in \argmin_{f:\X\to\Y} \E_{(X, Y)\sim\mu^*}\bracket{\ell(f(X), Y)}.
\end{equation}
This principle is motivated by Theorem 1 of \citet{Cabannes2020} showing that $f^*$ in \eqref{df:eq:solution} is characterized by $f^* \in \argmin_{f:\X\to\Y} \E_{(X, S)\sim\nu}\bracket{\inf_{y\in S} \ell(f(X), y)}$, matching a prior formulation based on infimum loss \citep{Cour2011,Luo2010,Hullermeier2014}.
In practice, it means that if $(S\vert X=x)$ has probability 50\% to be the set ``feline'' and 50\% the set ``orange with black stripes'', $(Y\vert X=x)$ should be considered as 100\% ``tiger'', rather than 20\% ``cat'', 30\% ``lion'' and 50\% ``orange car with black stripes'', which could also explain $(S\vert X=x)$.
Similarly to supervised learning, partial labeling consists in retrieving $f^*$ without accessing $\nu$ but only samples $(X_i, S_i)_{i\leq n} \sim \nu^{\otimes n}$.

\begin{remark}[Measure of determinism]
  \label{df:rmk:determinism}
  \eqref{df:eq:principle} is not the only variational way to push toward distribution where labels are deterministic function of the inputs.
  For example, one could minimize entropy \citep[{\em e.g.},][]{Berthelot2019,Lienen2021}.
  However, a loss-based principle is appreciable since the loss usually encodes structures of the output space \citep{Ciliberto2020}, which will allow sample and computational complexity of consequent algorithms to scale with an intrinsic dimension of the space rather than the real one, {\em e.g.}, $m$ rather than $m!$ when $\Y = \Sfrak_m$ and $\ell$ is a suitable ranking loss \citep[see Section~\ref{df:sec:ranking} or ][]{Nowak2019}.
\end{remark}

\section{Learning algorithm}
\label{df:sec:algo}

In this section, given weakly supervised samples, we present a disambiguation algorithm to retrieve fully supervised samples based on an empirical expression of \eqref{df:eq:principle}, before learning a mapping from $\X$ to $\Y$ based on those fully supervised samples, according to \eqref{df:eq:solution}.

Given a partially labeled dataset ${\cal D}_n = (x_i, s_i)_{i\leq n}$, sampled accordingly to $\nu^{\otimes n}$, we retrieve fully supervised samples, based on the following empirical version of \eqref{df:eq:principle}, with $C_n = \prod_{i\leq n} s_i \subset \Y^n$
\begin{equation}
  \label{df:eq:disambiguation}
  (\hat{y}_i)_{i\leq n} \in \argmin_{(y_i)_{i\leq n} \in C_n} \inf_{(z_i)_{i\leq n} \in \Y^n} \sum_{i, j=1}^n \alpha_j(x_i) \ell(z_i, y_j),
\end{equation}
where $(\alpha_i(x))_{i\leq n}$ is a set of weights measuring how much one should base its prediction for $x$ on the observations made at $x_i$.
This formulation is motivated by the Bayes approximate rule proposed by \citet{Stone1977}, which can be seen as the approximation of $\mu$ by $n^{-1}\sum_{i,j=1}^n \alpha_j(x_i)\delta_{x_i} \otimes \delta_{y_j}$ in \eqref{df:eq:principle}.

Once fully supervised samples $(x_i, \hat{y_i})$ have been recollected, one can learn $f_n:\X\to\Y$, approximating $f^*$, with classical supervised learning techniques.
In this work, we will consider the structured prediction estimator introduced by \citet{Ciliberto2016}, defined as
\begin{equation}
  \label{df:eq:estimate}
  f_n(x) \in \argmin_{z\in \Y} \sum_{i=1}^n \alpha_i(x) \ell(z, \hat{y}_i).
\end{equation}

\paragraph{Weighting scheme $\alpha$.}
For the weighting scheme $\alpha$, several choices are appealing.
Laplacian diffusion is one of them as it incorporates a prior on low density separation to boost learning \citep{Zhu2003,Zhou2003,Bengio2006,Hein2007}.
Kernel ridge regression is another due to its theoretical guarantees \citep{Ciliberto2020}.
In the theoretical analysis, we will use nearest neighbors.
Assuming $\X$ is endowed with a distance $d$, and assuming, for readability’ sake, that ties to define nearest neighbors do not happen, it is defined as
\[
  \alpha_i(x) = \left\{
  \begin{array}{cc}
    k^{-1} & \text{if}\quad \sum_{j=1}^n \ind{d(x, x_j) \leq d(x, x_i)} \leq k \\
    0      & \text{otherwise},
  \end{array}\right.
\]
where $k$ is a parameter fixing the number of neighbors.
Our analysis, leading to Theorem \ref{df:thm:convergence}, also holds for other local averaging methods such as partitioning or Nadaraya-Watson estimators.

\section{Consistency result}
\label{df:sec:stat}

In this section, we assume $\Y$ finite, and prove the convergence of $f_n$ toward $f^*$ as $n$, the number of samples, grows to infinity.
To derive such a consistency result, we introduce a surrogate problem that we relate to the risk through a calibration inequality.
We later assume that weights are given by nearest neighbors and review classical assumptions, that we work to derive exponential convergence rates.

In the following, we are interested in bounding the expected generalization error, defined as
\begin{equation}
  \label{df:eq:excess}
  {\cal E}(f_n) = \E_{{\cal D}_n} {\cal R}(f_n) - {\cal R}(f^*),
\end{equation}
where
\(
{\cal R}(f) = \E_{(X, Y)\sim\mu^*}\bracket{\ell(f(X), Y)},
\)
by a quantity that goes to zero, when $n$ goes to infinity.
This implies, under boundedness of $\ell$, convergence in probability (the randomness being inherited from ${\cal D}_n$) of ${\cal R}(f_n)$ toward $\inf_{f:\X\to\Y} {\cal R}(f)$, which is referred as \emph{consistency} of the learning algorithm.\footnote{If $\E|X| < +\infty$ and $\E[\abs{X_n - X}]\to 0$, $X_n \to X$ in probability.}
We first introduce a few objects.

\paragraph{Disambiguation ground truth \texorpdfstring{$(y_i^*)$}{}.}
Introduce $\pi^* \in \prob{\X\times\Y\times 2^\Y}$ expressing the compatibility of $\mu^*$ and $\nu$ as in Definition \ref{df:def:compatibility}.
Given samples $(x_i, s_i)_{i\leq n}$ forming a dataset ${\cal D}_n$, we enrich this dataset by sampling $y_i^* \sim \pi^*\vert_{x_i, s_i}$, which build an underlying dataset $(x_i, y_i^*, s_i)$ sampled accordingly $(\pi^*)^{\otimes n}$.
Given ${\cal D}_n$, while {\em a priori}, $y_i^*$ are random variables, sampled accordingly to $\pi^*\vert_{x_i, s_i}$, because of the definition of $\mu^*$ \eqref{df:eq:principle}, under basic definition assumptions, they are actually deterministic, defined as $y_i^* = \argmin_{y\in s_i} \ell(f^*(x_i), y)$.
As such, they should be seen as ground truth for $\hat{y}_i$.

\paragraph{Surrogate estimates.}
The approximate Bayes rule was successfully analyzed recently through the prism of plug-in estimators by \citet{Ciliberto2020}.
While we will not cast our algorithm as a plug-in estimator, we will leverage this surrogate approach, introducing two mappings $\phi$ and $\psi$ from~$\Y$ to a Hilbert space ${\cal H}$ such that
\begin{equation}
  \label{df:eq:loss}
  \forall\, z, y \in \Y, \qquad \ell(z, y) = \scap{\psi(z)}{\phi(y)},
\end{equation}
Such mappings always exist when $\Y$ is finite, and have been used to encode ``problem structure'' defined by the loss $\ell$ \citep{Nowak2019}.
We introduce three surrogate quantities that will play a major role in the following analysis, they map $\X$ to ${\cal H}$ as
\begin{gather}
  \label{df:eq:surrogate}
  g^*(x) = \E_{\mu^*}\bracket{\phi(Y)\midvert X=x},\qquad
  g_n(x) = \sum_{i=1}^n \alpha_i(x) \phi(\hat{y}_i),\nonumber\\
  g_n^*(x) = \sum_{i=1}^n \alpha_i(x) \phi(y^*_i).
\end{gather}
It is known that $f^*$ and $f_n$ are retrieved from $g^*$ and $g_n$, through the decoding, retrieving $f:\X\to\Y$ from $g:\X\to{\cal H}$ as
\begin{equation}
  \label{df:eq:decoding}
  f(x) = \argmin_{z\in\Y} \scap{\psi(z)}{g(x)},
\end{equation}
which explains the wording of {\em plug-in} estimator \citep{Ciliberto2020}.
We now introduce a {\em calibration inequality}, that relates the error between $f_n$ and $f^*$ with surrogate error quantities.

\begin{lemma}[Calibration inequality]\label{df:lem:cal}
  When $\Y$ is finite, and the labels are a deterministic function of the input, {\em i.e.}, when $\mu^*\vert_x$ is a Dirac for all $x\in\supp\nu_\X$, for any weighting scheme such that $\sum_{i=1}^n \abs{\alpha_i(x)} \leq 1$ for all $x\in\supp\nu_\X$,
  \begin{align}
    \label{df:eq:cal}
     & {\cal R}(f_n) - {\cal R}(f^*) \leq 4c_\psi \norm{g_n^* - g_n}_{L^1} \hspace*{2cm}\nonumber \\
     & \hspace*{1cm} \qquad+
    8c_\psi c_\phi \Pbb_X\paren{\norm{g_n^*(X) - g^*(X)} > \delta},
  \end{align}
  with $c_\psi = \sup_{z\in\Y} \norm{\psi(z)}$, $c_\phi = \sup_{z\in\Y}\norm{\phi(y)}$, and $\delta$ a parameter that depend on the geometry of $\ell$ and its decomposition through $\phi$.
\end{lemma}

This lemma, proven in Appendix \ref{df:proof:cal}, separates a part reading in $\norm{g_n - g_n^*}$, due to the {\em disambiguation error} between $(\hat{y}_i)$ and $(y_i^*)$ together with the {\em stability} of the learning algorithm when substituting $(\hat{y}_i)$ for $(y_i^*)$, and a part in $\norm{g_n^* - g^*}$ due to the {\em consistency} of the fully supervised learning algorithm.
The expression of the first part relates to Theorem 7 in \citet{Ciliberto2020} while the second part relates to Theorem 6 in \citet{Cabannes2021b}.

\subsection{Classical learnability assumptions}
In the following, we suppose that the weights $\alpha$ are given by nearest neighbors, that $\X$ is a compact metric space endowed with a distance $d$, that $\Y$ is finite and that $\ell$ is proper in the sense that it strictly positive except on the diagonal of $\Y\times\Y$ diagonal where it is zero.
We now review classical assumptions to prove consistency.
First, assume that $\nu_\X$ is regular in the following sense.

\begin{assumption}[$\nu_\X$ well-behaved]\label{df:ass:mass}
  Assume that $\nu_\X$ is such that there exists $h_1, c_\mu, q > 0$ satisfying, with ${\cal B}$ designing balls in $\X$,
  \[
    \forall\, x\in\supp\nu_\X,\ \forall\, r < h_1,
    \qquad \nu_\X({\cal B}(x, r)) > c_\mu r^q.
  \]
\end{assumption}

Assumption \ref{df:ass:mass} is useful to make sure that neighbors in ${\cal D}_n$ are closed with respect to the distance $d$, it is usually derived by assuming that $\X$ is a subset of $\R^q$; that $\nu_\X$ has a density $p$ against the Lebesgue measure $\lambda$ with {\em minimal mass} $p_{\min}$ in the sense that for any $x\in\supp\nu_\X$, $p(x) > p_{\min}$; and that $\supp\nu_\X$ has regular boundary in the sense that $\lambda({\cal B}(x, r) \cap \supp\nu_\X) \geq c\lambda({\cal B}(x, r)$ for any $x\in\supp\nu_\X$ and $r < h$ \citep[{\em e.g.},][]{Audibert2007}.

We now switch to a classical assumption in partial labeling, allowing for population disambiguation.

\begin{assumption}[Non ambiguity, \citet{Cour2011}]
  \label{df:ass:non-ambiguity}
  Assume the existence of $\eta \in [0, 1)$, such that for any $x\in\supp\nu_\X$, there exists $y_x \in \Y$, such that $\Pbb_{\nu}\paren{y_x \in S \midvert X=x} = 1$, and
  \[
    \forall\, z\neq y_x,\quad
    \Pbb_{\nu}\paren{z \in S \midvert X=x} \leq \eta.
  \]
\end{assumption}

Assumption \ref{df:ass:non-ambiguity} states that when given the full distribution $\nu$, there is one, and only one, label that is coherent with every observable sets for a given input.
It is a classical assumption in literature about the learnability of the partial labeling problem \citep[{\em e.g.},][]{Liu2014}.
When $\ell$ is proper, this implies that $\mu^*\vert_x = \delta_{y_x}$, and $f^*(x) = y_x$.

Finally, we assume that $g^*$ is regular.
As we are considering local averaging method, we will use Lipschitz-continuity, which is classical in such a setting.\footnote{Its generalization through H\"older-continuity would work too.}

\begin{assumption}[Regularity of $g^*$]\label{df:ass:lipschitz}
  Assume that there exists $c_g > 0$, such that for any $x, x'\in\X$, we have
  \[
    \norm{g^*(x) - g^*(x')}_{\cal H} \leq c_g d(x, x').
  \]
\end{assumption}

It should be noted that regularity of $g^*$, Assumption \ref{df:ass:lipschitz}, together with determinism of $\mu^*\vert_x$ inherited from Assumption~\ref{df:ass:non-ambiguity} implies that classes $\X_y = \brace{x\midvert f^*(x) = y}$ are separated in $\X$, in the sense that there exists $h_2 > 0$, such that for any $y, y' \in \Y$ and $(x, x') \in \X_y \times \X_{y'}$, $d(x, x') > h_2$, which is a classical assumption to derive consistency of semi-supervised learning algorithm \citep[{\em e.g.},][]{Rigollet2007}.
We detailed those implications in Appendix~\ref{df:proof:ass}.

\subsection{Exponential convergence rates}
We are now ready to state our convergence result.
We introduce $h = \min(h_1, h_2)$ and $p = c_\mu h^q$, so that for any $x \in \supp\nu_\X$, $\nu_\X({\cal B}(x, h)) > p$.

\begin{theorem}[Exponential convergence rates]
  \label{df:thm:convergence}
  When the weights $\alpha$ are given by nearest neighbors, under Assumptions \ref{df:ass:mass}, \ref{df:ass:non-ambiguity} and \ref{df:ass:lipschitz}, the excess of risk in~\eqref{df:eq:excess} is bounded by
  \begin{align}
    \label{df:eq:convergence}
    {\cal E}(f_n) & \leq 8c_\psi c_\phi (n+1)\exp\paren{-{\frac{np}{16}}} \hspace*{2cm} \nonumber \\
                  & \hspace*{1cm} \qquad + 8c_\psi c_\phi m \exp\paren{-k\abs{\log(\eta)}},
  \end{align}
  as soon as $k < np / 4$, with $m = \card{\Y}$.
  By taking $k_n = k_0 n$, for $k_0 < p/4$, this implies exponential convergence rates ${\cal E}(f_n) = O(n\exp(-n))$.
\end{theorem}

\begin{proof}[Sketch for Theorem \ref{df:thm:convergence}]
  In essence, based on Lemma \ref{df:lem:cal}, Theorem \ref{df:thm:convergence} can be understood as two folds.
  \begin{itemize}
    \item A fully supervised error between $g_n^*$ and $g^*$.
          This error can be controlled in $\exp(-np)$ as the non-ambiguity assumption implies a hard Tsybakov margin condition, a setting in which {\em the fully supervised estimate $g_n^*$ is known to converge to the population solution $g^*$ with such rates} \citep{Cabannes2021b}.
    \item A weakly disambiguation error, that is exponential too, since, based on Assumption \ref{df:ass:non-ambiguity}, disambiguating between $z \in \Y$ and $y_x$ from $k$ sets $S$ sampled accordingly to $\nu\vert_x$ can be done in $\eta^k$, and disambiguating between all $z\neq y_x$ and $y_x$ in $m\eta^k = m \exp(-k\abs{\log(\eta)})$.
  \end{itemize}
  Appendix~\ref{df:proof:convergence} provides details.
\end{proof}

Theorem \ref{df:thm:convergence} states that under a non-ambiguity assumption and a regularity assumption implying no-density separation, one can expect exponential convergence rates of $f_n$ learned with weakly supervised data to $f^*$ the solution of the fully supervised learning problem, measured with excess of fully supervised risk.
Because of exponential convergence rates, we could expect polynomial convergence rates for a broader class of problems that are approximated by problems satisfying assumptions of Theorem \ref{df:thm:convergence}.
{\em The derived rates in $n\exp(-n)$ should be compared with rates in $n^{-1/2}$ and $n^{-1/4}$}, respectively derived, {\em under the same assumptions}, by \citet{Cour2011,Cabannes2020}.

\subsection{Discussion on assumptions}
While we have retaken classical assumptions from literature, those assumptions are quite strong, which allows us, by understanding their strength, to derive exponential convergence rates.
Assumptions \ref{df:ass:mass} and \ref{df:ass:lipschitz} are classical in the nearest neighbor literature with full supervision.
If we were using (reproducing) kernel methods to define the weighting scheme $\alpha$, those assumptions would be mainly replaced with ``$g^*$ belonging to the RKHS''.
Assumption \ref{df:ass:non-ambiguity} is the strongest assumption in our view, that we will now discuss.

\paragraph{How to check it in practice ?}
First, for Assumption \ref{df:ass:non-ambiguity} to hold, the labels have to be a deterministic function of the inputs.
In other words, a zero error is achievable.
Finally, Assumption \ref{df:ass:non-ambiguity} is related to dataset collection.
If dealing with images, weak supervision could take the form of some information on shape, color, or texture, etc., Assumption \ref{df:ass:non-ambiguity} supposes that the weak information potentially given on a specific image~$x$ allows retrieving the unique label $y$ of the image ({\em e.g.}, a “pig” could be recognized from its shape and its color).
This is a reasonable assumption, if, for a given $x$, we ask at random a data labeler to provide us information on shape, color, or texture, etc.
However, it will not be the case, if for some reasons ({\em e.g.} the dataset is built from several weakly annotated datasets), in some regions of the input space, we only get shape information, and in other regions, we only get color information.
In particular, it is not verified for semi-supervised learning when the support of the unlabeled data distribution is not the same as the support of the labeled input data distribution.

\paragraph{How to relax it and what results to expect?}
Previous works used Assumption \ref{df:ass:non-ambiguity} to derive a calibration inequality between the infimum loss to the original loss \citep[{\em e.g.}, see Proposition 2 by][]{Cabannes2020}.
In contrast, we relate the surrogate and original problem through a refined calibration inequality \eqref{df:eq:cal}.
This technical progress allows us to derive exponential convergence rates similarly to the work of \citet{Cabannes2021b}.
Importantly, in comparison with previous work, our calibration inequality Lemma \ref{df:lem:cal} can easily be extended without the determinism assumption provided by Assumption \ref{df:ass:non-ambiguity}.
Essentially, in our work, Assumption \ref{df:ass:non-ambiguity} is used to simplify the study of $(\hat{y}_i)_{i\leq n}$ given by the disambiguation algorithm \eqref{df:eq:disambiguation}, and therefore the study of the disambiguation error in~\eqref{df:eq:cal}.
The study of $(\hat{y}_i)_{i\leq n}$ without Assumption \ref{df:ass:non-ambiguity} would require other tools than the one presented in this paper.
It could be studied in the realm of graphical model and message passing algorithm, or with Wasserstein distance and topological considerations on measures.
With much milder forms of Assumption \ref{df:ass:non-ambiguity}, we expect the rates to degrade smoothly with respect to a parameter defining the hardness of the problem, similarly to the works of \citet{Audibert2007,Cabannes2021b}.

\section{Optimization considerations}
\label{df:sec:optim}

In this section, we focus on implementations to solve~\eqref{df:eq:disambiguation}.
We explain why disambiguation objectives, such as~\eqref{df:eq:principle} are intrinsically non-convex and express a heuristic strategy to solve~\eqref{df:eq:disambiguation} besides non-convexity in classical well-behaved instances of partial labeling.
Note that we do not study implementations to solve~\eqref{df:eq:estimate} as this study has already been done by \citet{Nowak2019}.
We end this section by considering a practical example to make derivations more concrete.

\subsection{Non-convexity of disambiguation objectives}

For readability, suppose that $\X$ is a singleton, justifying to remove the dependency on the input in the following.
Consider $\nu \in \prob{2^\Y}$ a distribution modelling weak supervision.
While the domain $\brace{\mu\in\prob{\Y} \midvert \mu \vdash \nu}$ is convex, a disambiguation objective ${\cal E}:\prob{\Y} \to \R$ defining $\mu^* \in \argmin_{\mu\vdash\nu} {\cal E}(\mu)$, similarly to~\eqref{df:eq:principle}, that is minimized for deterministic distributions, which correspond to $\mu$ a Dirac, {\em i.e.}, minimized on vertices of its definition domain $\prob{\Y}$, can not be convex.
In other terms, any disambiguation objective that pushes toward distributions where targets are deterministic function of the input, as mentioned in Remark \ref{df:rmk:determinism}, can not be convex.

Indeed, smooth disambiguation objectives such as entropy and our piecewise linear loss-based principle~\eqref{df:eq:principle}, reading pointwise
\(
{\cal E}(\mu) = \inf_{z\in\Y} \E_{Y\sim\mu}[\ell(z, Y)],
\)
are concave.
Similarly, its quadratic variant
\(
{\cal E}'(\mu) = \E_{Y, Y'\sim\mu}[\ell(Y, Y')],
\)
is concave as soon as $(\ell(y, y'))_{y, y'\in\Y}$ is semi-definite negative.
We illustrate those considerations on a concrete example with graphical illustration in Appendix \ref{df:app:example}.
We should see how this translates on generic implementations to solve the empirical objective~\eqref{df:eq:disambiguation}.

\subsection{Generic implementation for~\eqref{df:eq:disambiguation}}

Depending on $\ell$ and on the type of observed set $(s_i)$, \eqref{df:eq:disambiguation}~might be easy to solve.
In the following, however, we will introduce optimization considerations to solve it in a generic structured prediction fashion.
To do so, we recall the decomposition of $\ell$~\eqref{df:eq:loss} and rewrite~\eqref{df:eq:disambiguation} as
\begin{equation*}
  (\hat{y}_i)_{i\leq n} \in \argmin_{y_i \in C_n} \inf_{(z_i) \in \Y^n} \sum_{i,j=1}^n \alpha_j(x_i) \psi(z_i)^\top \phi(y_j).
\end{equation*}
Since, given $(y_j)$, the objective is linear in $\psi(z_j)$, the constraint $\psi(z_j) \in \psi(\Y)$ can be relaxed with $\zeta_i \in \hull\psi(\Y)$.\footnote{The minimization pushes toward extreme points of the definition domain.}
Similarly, with respect to $\phi(y_j)$, this objective is the infimum of linear functions, therefore is concave, and the constraint $\phi(y_j) \in \phi(s_j)$, could be relaxed with $\xi_i \in \hull\phi(s_j)$.
Hence, with ${\cal H}_0 = \hull\psi(\Y)$ and $\Gamma_n = \prod_{j\leq n} \hull\phi(s_j)$, the optimization is cast as
\begin{equation}
  \label{df:eq:algo}
  (\hat{\xi}_i)_{i\leq n} \in \argmin_{(\xi_i) \in \Gamma_n} \inf_{(\zeta_i) \in{\cal H}_0^n} \sum_{i,j=1}^n \alpha_j(x_i) \zeta_i^\top \xi_j.
\end{equation}
Because of concavity, $(\hat{\xi}_i)$ will be an extreme point of $\Gamma_n$, that could be decoded into $\hat{y}_i = \phi^{-1}(\hat{\xi}_i)$.
However, it should be noted that if only interested in $f_n$ and not in the disambiguation $(\hat{y}_i)$, this decoding can be avoided, since~\eqref{df:eq:estimate} can be rewritten as $f_n(x) \in \argmin_{z\in\Y} \psi(z)^\top \sum_{i=1}^n \alpha_i(x) \hat{\xi}_i$.

\subsection{Alternative minimization with good initialization}

To solve~\eqref{df:eq:algo}, we suggest using an alternative minimization scheme.
The output of such a scheme is highly dependent to the variable initialization.
In the following, we introduce well-behaved problem, where $(\xi_i)_{i\leq n}$ can be initialized smartly, leading to an efficient implementation to solve~\eqref{df:eq:algo}.

\begin{definition}[Well-behaved partial labeling problem]
  \label{df:def:init}
  A partial labeling problem $(\ell, \nu)$ is said to be well-behaved if for any $s \in \supp\nu_{2^\Y}$, there exists a signed measure $\mu_s$ on $\Y$ such that the function from $\Y$ to $\R$ defined as $z\to \int_{\Y} \ell(z, y) \diff\mu_s(y)$ is minimized for, and only for, $z \in s$.
\end{definition}

We provide a real-world example of a well-behaved problem in Section \ref{df:sec:ranking} as well as a synthetic example with graphical illustration in Appendix \ref{df:app:example}.
On those problems, we suggest solving~\eqref{df:eq:algo} by considering the initialization $\xi^{(0)}_i = \E_{Y\sim\mu_{s_i}}[\phi(Y)]$, and performing alternative minimization of~\eqref{df:eq:algo}, until attaining $\xi^{(\infty)}$ as the limit of the alternative minimization scheme (which exists since each step decreases the value of the objective in~\eqref{df:eq:algo} and there is a finite number of candidates for $(\xi_i)$).
It corresponds to a disambiguation guess $\tilde{y}_i = \phi^{-1}(\xi_i^{(\infty)})$.
Then we suggest learning $\hat{f}_n$ from $(x_i, \tilde{y}_i)$ based on~\eqref{df:eq:estimate}, and existing algorithmic tools for this problem \citep{Nowak2019}.
To assert the well-foundedness of this heuristic, we refer to the following proposition, proven in Appendix \ref{df:proof:init}.

\begin{proposition}
  \label{df:prop:init}
  Under the non-ambiguity hypothesis, Assumption \ref{df:ass:non-ambiguity}, the solution of~\eqref{df:eq:solution} is characterized by
  \(
  f^* \in \argmin_{f:\X\to\Y} \E_{(X, S) \sim \nu}\bracket{\E_{Y\sim\mu_S}[\ell(f(X), Y)]}.
  \)
  Moreover, if the surrogate function $g_n^{\circ}:\X\to{\cal H}$ defined as $g_n^{\circ}(x) = \sum_{i=1}^n\alpha_i(x)\xi_{s_i}$, with $\xi_s = \E_{Y\sim\mu_s}[\phi(Y)]$, converges toward $g^\circ(x) = \E_{S\sim\nu\vert_x}[\xi_S]$ in $L^1$, $f_n^{\circ}$ defined through the decoding~\eqref{df:eq:decoding} converges in risk toward $f^*$.
\end{proposition}

Given that our algorithm scheme is initialized for $\xi_i^{(0)} = \xi_{s_i}$ and $\zeta_i^{(0)} = f_n^{\circ}(x_i)$ and stopped once having attained $\xi_i^{(\infty)}$ and $\zeta_i^{(\infty)} = \hat{f}_n(x_i)$, $\hat{f}_n$ is arguably better than $f_n^{\circ}$, which given consistency result exposed in Proposition \ref{df:prop:init}, is already good enough.

\begin{remark}[IQP implementation for~\eqref{df:eq:disambiguation}]
  Other heuristics to solve~\eqref{df:eq:disambiguation} are conceivable.
  For example, considering $z_i = y_i$ in this equation, we remark that the resulting problem is isomorphic to an integer quadratic program (IQP).
  Similarly to integer linear programming, this problem can be approached with relaxation of the ``integer constraint'' to get a real-valued solution, before ``thresholding'' it to recover an integer solution.
  This heuristic can be seen as a generalization of the Diffrac algorithm \citep{Bach2007,Joulin2010}.
  We present it in details in Appendix~\ref{df:app:diffrac}.
\end{remark}

\begin{remark}[Link with EM, \citep{Dempster1977}]
  Arguably, our alternative minimization scheme, optimizing respectively the targets $\xi_i = \phi(y_i)$ and the function estimates $\zeta_i = \psi(f_n(x_i))$ can be seen as the non-parametric version of the Expectation-Maximization algorithm, popular for parametric model \citep{Dempster1977}.
\end{remark}

\subsection{Application: ranking with partial ordering}
\label{df:sec:ranking}

Ranking is a problem consisting, for an input $x$ in an input space $\X$, to learn a total ordering $y$, belonging to $\Y = \Sfrak_m$, modelling preference over $m$ items.
It is usually approached with the Kendall loss $\ell(y, z) = - \phi(y)^\top \phi(z)$, with $\phi(y) = (\sign\paren{y(i) - y(j)})_{i,j\leq m} \in \brace{-1,1}^{m^2}$ \citep{Kendall1938}.
Full supervision corresponds, for a given $x$, to be given a total ordering of the $m$ items.
This is usually not an option, but one could expect to be given partial ordering that $y$ should follow \citep{Cao2007,Hullermeier2008,Korba2018}.
Formally, this equates to the observation of some, but not all, coordinates $\phi(y)_i$ of the vector $\phi(y)$ for some $i\in I \subset \bbracket{1, m}^2$.

In this setting, $s\subset \Y$ is a set of total orderings that match the given partial ordering.
It can be represented by a vector $\xi_s \in {\cal H}$, that satisfies the partial ordering observation, $(\xi_s)_I = \phi(y)_I$, and that is agnostic on unobserved coordinates, $(\xi_s)_{^c I} = 0$.
This vector satisfies that $z\to\psi(z)^\top\xi_s$ is minimized for, and only for, $z\in s$.
Hence, it constitutes a good initialization for the alternative minimization scheme detailed above.
We provide details in Appendix \ref{df:proof:ranking}, where we also show that $\xi_s$ can be formally translated in a $\mu_s$ to match the Definition \ref{df:def:init}, proving that ranking with partial labeling is a well-behaved problem.

Many real world problems can be formalized as a ranking problem with partial ordering observations.
For example, $x$ could be a social network user, and the $m$ items could be posts of her connection that the network would like to order on her feed accordingly to her preferences.
One might be told that the user $x$ prefer posts from her close rather than from her distant connections, which translates formally as the constraint that for any $i$ corresponding to a post of a close connection and $j$ corresponding to a post of a distant connection, we have $\phi(y)_{ij} = 1$.
Nonetheless, designing non-parametric structured prediction models that scale well when the intrinsic dimension $m$ of the space $\Y$ is very large (such as the number of post on a social network) remains an open problem, that this paper does not tackle.

\section{Related work}
\label{df:sec:litterature}

Weakly supervised learning has been approached through parametric and non-parametric methods.
Parametric models are usually optimized through maximum likelihood \citep{Heitjan1991,Jin2002}.
\citet{Hullermeier2014} show that this approach, as formalized by \citet{Denoeux2013}, equates to disambiguating sets by averaging candidates, which was shown inconsistent by \citet{Cabannes2020} when data are {\em not missing at random}.
Among non-parametric models, \citet{Xu2004,Bach2007} developed an algorithm for clustering, that has been cast for weakly supervised learning problem \citep{Joulin2010,Alayrac2016}, leading to a disambiguation algorithm similar than ours, yet without consistency results.
More recently, half-way between theory and practice, \citet{Gong2018} derived an algorithm geared toward classification, based on a disambiguation objective, incorporating several heuristics, such as class separation, and Laplacian diffusion.
Those heuristics could be incorporated formally in our model.

The infimum loss principle has been considered by several authors, among them \citet{Cour2011,Luo2010,Hullermeier2014}.
It was recently analyzed through the prism of structured prediction by \citet{Cabannes2020}, leading to a consistent non-parametric algorithm that will constitute the baseline of our experimental comparison.
This principle is interesting as it does not assume knowledge on the corruption process $(S\vert Y)$ contrarily to the work of \citet{CidSueiro2014} or \citet{vanRooyen2018}.

The non-ambiguity assumption has been introduced by \citet{Cour2011} and is a classical assumption of learning with partial labeling \citep{Liu2014}.
Assumptions of Lipschitzness and minimal mass are classical assumptions to prove convergence of local averaging method \citep{Audibert2007,Biau2015}.
Those assumptions imply class separation in $\X$, which has been leverage in semi-supervised learning, justifying Laplacian regularization \citep{Rigollet2007,Zhu2003}.

Note that those assumptions might not hold on raw representation of the data, but with appropriate metrics, which could be learned through unsupervised \cite{Duda2000} or self-supervised learning \cite{Doersch2017}.
As such, the practitioner might consider weights $\alpha$ given by similarity metrics derived through such techniques, before computing the disambiguation~\eqref{df:eq:disambiguation} and learning $f_n$ from the recollected fully supervised dataset with deep learning.

\section{Experiments}
\label{df:sec:example}

In this section, we review a baseline, and experiments that showcase the usefulness of our algorithm -- which corresponds to ~\eqref{df:eq:disambiguation}~and~\eqref{df:eq:estimate}.

\paragraph{Baseline.}
We consider as a baseline the work of \citet{Cabannes2020}, which is a consistent structured prediction approach to partial labeling through the infimum loss.
It is arguably the state-of-the-art of partial labeling approached through structured prediction.
It follows the same loss-based variance disambiguation principle, yet in an implicit fashion, leading to the inference algorithm, $f_n:\X\to\Y$,
\begin{equation}
  \label{df:eq:baseline}
  f_n(x) \in \argmin_{z\in\Y} \inf_{(y_i) \in C_n} \sum_{i=1}^n \alpha_i(x) \ell(z, y_i).
\end{equation}
Statistically, exponential convergence rates similar to Theorem \ref{df:thm:convergence} could be derived.
Yet, as we will see, our algorithm outperforms this state-of-the-art baseline.

\begin{figure*}[t]
  \centering
  \includegraphics[width=.45\linewidth]{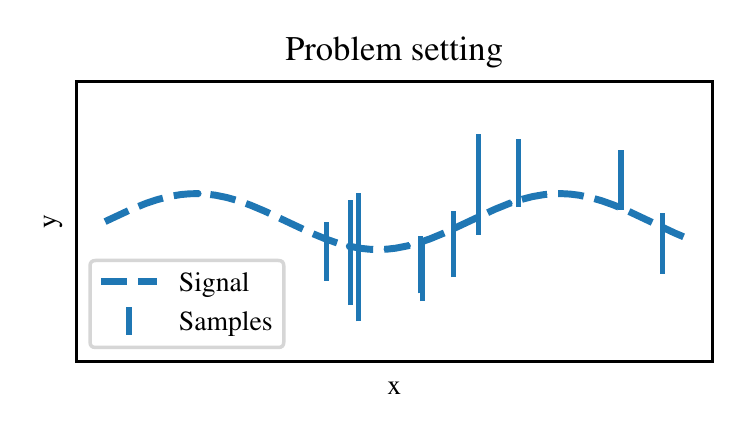}
  \includegraphics[width=.45\linewidth]{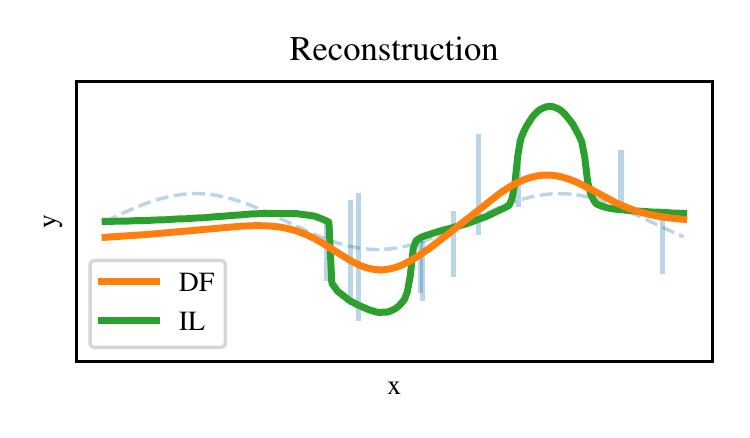}
  \caption{
    Interval regression.
    See Appendix \ref{df:app:experiments} for the exact reproducible experimental setup (Left) Setup.
    The goal is to learn $f^*:\X\to\R$ represented by the dashed line, given samples $(x_i, s_i)$, where $(s_i)$ are intervals represented by the blue segments.
    (Right) We compare the infimum loss (IL) baseline~\eqref{df:eq:baseline} shown in green, with our disambiguation framework (DF), \eqref{df:eq:disambiguation} and \eqref{df:eq:estimate}, shown in orange; with weights $\alpha$ given by kernel ridge regression.
    (DF) retrieves $\hat{y_i}$ before learning a smooth $f_n$ based on $(x_i, \hat{y}_i)$, while (IL) implicitly retrieves $\hat{y}_i(x)$ differently for each input, leading to irregularity of the consequent estimator of $f^*$.
  }
  \label{df:fig:ir}
\end{figure*}

\paragraph{Disambiguation coherence - Interval regression.}
The baseline~\eqref{df:eq:baseline} implicitly requires disambiguating $(\hat{y}_i(x))$ differently for every $x\in\X$.
This is counterintuitive since $(y_i^*)$ does not depend on $x$.
It means that $(\hat{y}_i)$ could be equal to some $(\hat{y}_i^{(0)})$ on a subset $\X_0$ of $\X$, and to another $(\hat{y}_i^{(1)})$ on a disjoint subset $\X_1 \subset \X$, leading to irregularity of $f_n$ between $\X_0$ and $\X_1$.
We illustrate this graphically on Figure \ref{df:fig:ir}.
This figure showcases an interval regression problem, which corresponds to the regression setup ($\Y = \R$, $\ell(y, z) = \abs{y - z}^2$) of partial labeling, where one does not observe $y\in\R$ but an interval $s\subset \R$ containing $y$.
Among others this problem appears in physics \citep{Sheppard1897} and economy \citep{Tobin1958}.

\paragraph{Computation attractiveness - Ranking.}
Computationally, the baseline requires to solve a disambiguation problem, recovering $(\hat{y_i}(x)) \in C_n$ for every $x\in\X$ for which we want to infer $f_n(x)$.
This is much more costly, than doing the disambiguation of $(\hat{y_i}) \in C_n$ once, and solving the supervised learning inference problem~\eqref{df:eq:estimate}, for every $x\in\X$ for which we want to infer $f_n(x)$.
To illustrate the computation attractiveness of our algorithm, consider the case of ranking, defined in Section \ref{df:sec:ranking}.
Fully supervised inference scheme~\eqref{df:eq:estimate} corresponds to solving a NP-hard problem, equivalent to the minimum feedback arcset problem \citep{Duchi2010}.
While disambiguation approaches with alternative minimization implied by~\eqref{df:eq:disambiguation} and~\eqref{df:eq:baseline} require to solve this NP-hard problem for each minimization step.
In other terms, the baseline ask to solve multiple NP-hard problem every time one wants to infer $f_n$ given by~\eqref{df:eq:baseline} on an input $x\in\X$.
Meanwhile, our disambiguation approach asks to solve multiple NP-hard problem upfront to solve~\eqref{df:eq:disambiguation}, yet only require to solve one NP-hard problem to infer $f_n$ given by~\eqref{df:eq:estimate} on an input $x\in\X$.

\begin{figure*}[t]
  \centering
  \includegraphics[width=.45\linewidth]{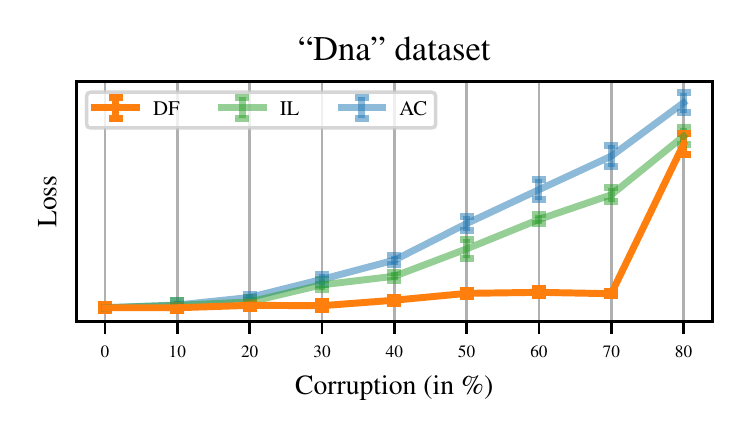}
  \includegraphics[width=.45\linewidth]{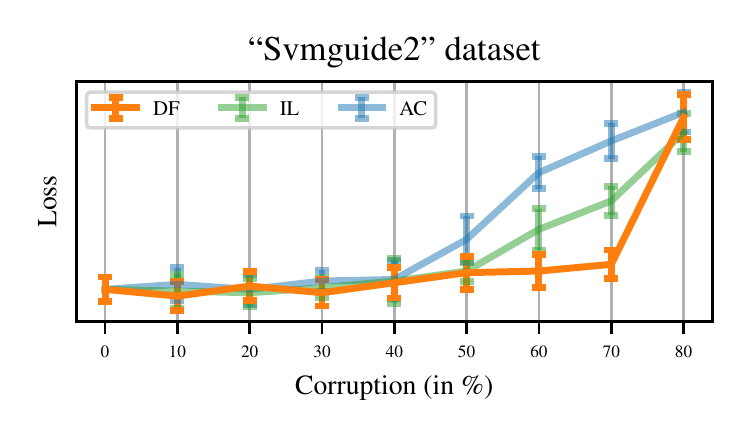}
  \caption{Testing errors as function of the supervision corruption on real dataset corresponding to classification with partial labels.
    We split fully supervised LIBSVM datasets into training and testing dataset.
    We corrupt training data in order to get partial labels.
    Corruption is managed through a parameter, represented by the $x$-axis, that relates to the ambiguity degree $\eta$ of Assumption \ref{df:ass:non-ambiguity}.
    For each method (our algorithm (DF), the baseline (IL), and the baseline of the baseline (AC, consisting of averaging candidates $y_i$ in sets $S_i$)), we consider weights $\alpha$ given by kernel ridge regression with Gaussian kernel, for which we optimized hyperparameters with cross-validation on the training set.
    We then learn an estimate $f_n$ that we evaluate on the testing set, represented by the $y$-axis, on which we have full supervision.
    The figure shows the superiority of our method, that achieves error similar to baseline when full supervision ($x=0$) or no supervision ($x=100\%$) is given, but performs better when only in presence of partial supervision.
    See Appendix \ref{df:app:experiments} for reproducibility specifications, where we also provide Figure \ref{df:fig:rk} showcasing similar empirical results in the case of ranking with partial ordering.
  }
  \label{df:fig:cl}
\end{figure*}

\paragraph{Better empirical results - Classification.}
Finally, we compare our algorithm, our baseline~\eqref{df:eq:baseline} and the baseline considered by \citet{Cabannes2020} on real datasets from the LIBSVM dataset \citep{Chang2011}.
Those datasets $(x_i, y_i)$ correspond to fully supervised classification problem.
In this setup, $\Y = \bbracket{1,m}$ for $m$ a number of classes, and $\ell(y, z) = \ind{y\neq z}$.
We ``corrupt'' labels in order to create a synthetic weak supervision datasets $(x_i, s_i)$.
We consider skewed corruption, in the sense that $(s_i)$ is generated by a probability such that $\sum_{z\in\Y} \Pbb_{S_i}(z\in S_i \vert y_i)$ depends on the value of $y_i$.
This corruption is parametrized by a parameter that related with the ambiguity parameter $\eta$ of Assumption \ref{df:ass:non-ambiguity}.
Results on Figure \ref{df:fig:cl} show that, in addition to having a lower computation cost, our algorithm performs better in practice than the state-of-the-art baseline.\footnote{All the code is available online - \url{https://github.com/VivienCabannes/partial_labelling/}.}

\begin{figure*}[ht]
  \centering
  \includegraphics[width=.3\linewidth]{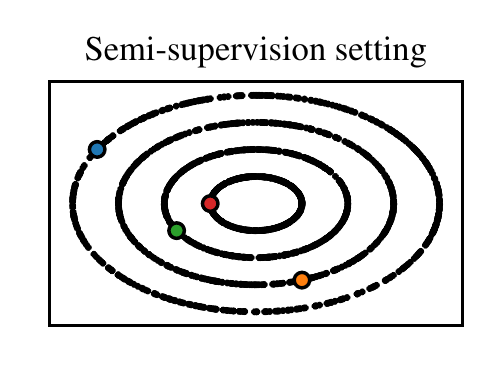}
  \includegraphics[width=.3\linewidth]{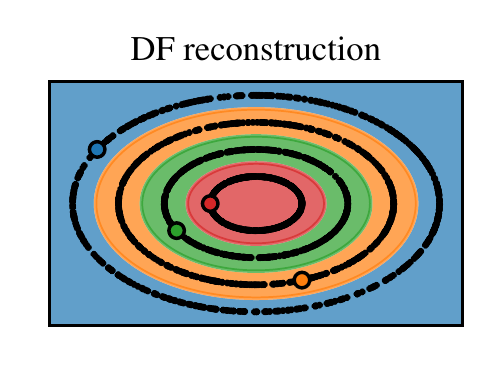}
  \includegraphics[width=.3\linewidth]{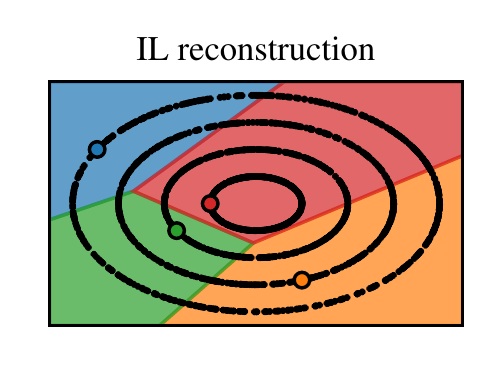}
  \caption{Semi-supervised learning, ``concentric circle'' instance with four classes (red, green, blue, yellow).
    Reproducibility details provided in Appendix \ref{df:app:experiments}.
    (Left) We represent points $x_i \in \X \subset \R^2$, there is many unlabeled points (represented by black dots and corresponding to $S_i = \Y$), and one labeled point for each class (represented in color, corresponding to $S_i = \brace{y_i}$).
    (Middle) Reconstruction $f_n:\X \to \Y$ given by our algorithm~\eqref{df:eq:disambiguation} and \eqref{df:eq:estimate}.
    Our algorithm succeeds to comprehend the concentric circle structure of the input distribution and clusters classes accordingly.
    (Right) Reconstruction $f_n:\X\to\Y$ given by the baseline~\eqref{df:eq:baseline}.
    The baseline performs as if only the four supervised data points where given.}
  \label{df:fig:ss}
\end{figure*}

\paragraph{Beyond~\eqref{df:eq:principle} - Semi-supervised learning.}
The main limitation of~\eqref{df:eq:principle} is that it is a pointwise principle that decorrelates inputs, in the sense that the optimization of $\mu^*\vert_x$, for $x\in\X$, only depends on $\nu\vert_x$ and not on what is happening on $\X\setminus \brace{x}$.
As such, this principle failed to tackle semi-supervised learning, where $\nu\vert_x$ is equal to $\mu\vert_x$ (in the sense that $\pi\vert_{x, y} =\delta_{\brace{y}}$) for $x\in\X_l$ and is equal to $\delta_\Y$ for $x \in \X_u:=\X\setminus\X_l$.
In such a setting, for $x\in\X_u$, $\mu^*\vert_x$ can be set to any $\delta_y$ for $y\in\Y$.
Interestingly, in practice, while the baseline suffer the same limitation, for our algorithm, {\em weighting schemes have a regularization effect}, that contrasts with those considerations.
We illustrate it on Figure \ref{df:fig:ss}.

\section{Conclusion}
\label{df:sec:opening}

In this work, we have introduced a structured prediction algorithm~\eqref{df:eq:disambiguation} and \eqref{df:eq:estimate}, to tackle partial labeling.
We have derived exponential convergence rates for the nearest neighbors instance of this algorithm under classical learnability assumptions.
We provided optimization considerations to implement this algorithm in practice, and have successfully compared it with the state-of-the-art.
Several open problems offer prospective follow-up of this works:

\begin{itemize}
  \item \emph{Semi-supervised learning and beyond.}
        While we only proved convergence in situation where $\mu^*$ of~\eqref{df:eq:principle} is uniquely defined, therefore excluding semi-supervised learning, Figure \ref{df:fig:ss} suggests that our algorithm \eqref{df:eq:disambiguation} could be analyzed in a broader setting than the one considered in this paper.
        Among others, we conjecture that the non-ambiguity assumption could be replaced by a cluster assumption \citep{Rigollet2007} together with a non-ambiguity assumption cluster-wise in Theorem~\ref{df:thm:convergence}.
  \item \emph{Hard-coded weak supervision.}
        Variational principles~\eqref{df:eq:principle} and \eqref{df:eq:solution} could be extended beyond partial labeling to any type of hard-coded weak supervision, which is when weak supervision can be cast as a set of hard constraint that $\mu$ should satisfy, formally written as a set of fully supervised distributions compatible with weak information.
        Hard-coded weak supervision includes label proportion \citep{Quadrianto2009,Dulac2019}, but excludes supervision of the type ``80\% of the experts say this nose is broken, and 20\% say it is not''.
        Providing a unifying framework for those problems would make an important step in the theoretical foundation of weakly supervised learning.
  \item \emph{Missing input data.}
        While weak supervision assumes that only $y$ is partially known, in many applications of machine learning, $x$ is also only partially known, especially when the feature vector $x$ is built from various source of information, leading to missing data.
        While we only considered a principle to fill missing output information, similar principles could be formalized to fill missing input information.
\end{itemize}

\begin{subappendices}
  \chapter*{Appendix}
  \addcontentsline{toc}{chapter}{Appendix}
  \section{Proofs}

\paragraph{Mathematical assumptions.}
To make formal what should be seen as implicit assumptions heretofore, we consider $\X$ and $\Y$ Polish spaces, $\Y$ compact, $\ell:\Y\times\Y\to\R$ continuous, ${\cal H}$ a separable Hilbert space, $\phi$ measurable, and $\psi$ continuous.
We also assume that for $\nu_x$-almost every $x\in\X$, and any $\mu\vdash\nu$, that the pushforward measure $\phi_*\mu\vert_x$ has a second moment.
This is the sufficient setup in order to be able to formally define objects and solutions considered all along the paper.

\paragraph{Notations.}
Beside standard notations, we use $\card{\Y}$ to design the cardinality of $\Y$, and $2^\Y$ to design the set of subsets of $\Y$.
Regarding measures, we use $\mu_\X$ and $\mu\vert_x$ respectively the marginal over $\X$ and the conditional accordingly to $x$ of $\mu\in\prob{\X\times\Y}$.
We denote by $\mu^{\otimes n}$ the distribution of the random variable $(Z_1, \cdots, Z_n)$, where the $Z_i$ are sampled independently according to $\mu$.
For $A$ a Polish space, we consider $\prob{A}$ the set of Borel probability measures on this space.
For $\phi:\Y\to{\cal H}$ and $S\subset \Y$, we denote by $\phi(S)$ the set $\brace{\phi(y)\midvert y\in S}$.
For a family of sets $(S_i)$, we denote by $\prod S_i$ the Cartesian product $S_1\times S_2\times\cdots$, also defined as the set of points $(y_i)$ such that $y_i \in S_i$ for all index $i$, and by $\Y^n$ the Cartesian product $\prod_{i\leq n}\Y$.
Finally, for $E$ a subset of a vector space $E'$, $\hull E$ denotes the convex hull of $E$ and $\Span(E)$ its span.

\paragraph{Abuse of notations.}
For readability’ sake, we have abused notations.
For a signed measure $\mu$, we denote by $\E_{\mu}[X]$ the integral $\int x\diff\mu(x)$, extending this notation usually reserved to probability measure.
More importantly, when considering $2^\Y$, we should actually restrict ourselves to the subspace ${\cal S} \subset 2^\Y$ of closed subsets of $\Y$, as ${\cal S}$ is a Polish space (metrizable by the Hausdorff distance) while $2^\Y$ is not always.
However, when $\Y$ is finite, those two spaces are equals, $2^\Y = {\cal S}$.

\subsection{Proof of Lemma \ref{df:lem:cal}}
\label{df:proof:cal}

From Lemma 3 in \citet{Cabannes2021b}, we pull the calibration inequality
\[
  {\cal R}(f_n) - {\cal R}(f^*) \leq 2c_\psi
  \E\bracket{\ind{\norm{g_n(X) - g^*(X)} > d(g^*(X), F)}\norm{g_n(X) - g^*(X)}}.
\]
Where $F$ is defined as the set of points $\xi \in \hull\phi(\Y)$ leading to two decodings
\[
  F = \brace{\xi \in \hull\phi(\Y)\midvert
    \card{\argmin_{z\in\Y}\scap{\psi(z)}{\xi}} > 1},
\]
and $d$ is defined as the extension of the norm distance to sets, for $\xi \in {\cal H}$
\[
  d(\xi, F) = \inf_{\xi'\in F} \norm{\xi - \xi'}_{\cal H}.
\]
Using that $\norm{g_n(X) - g^*(X)} \leq \norm{g_n(X) - g_n^*(X)} + \norm{g_n^*(X) - g^*(X)}$ and that, if $a \leq b + c$,
\[
  \ind{a > \delta} a \leq \ind{b + c > \delta} b + c \leq \ind{2\sup(b, c) >
    \delta} 2 \sup{b, c} = 2 \sup_{e\in{b, c}} \ind{e > \delta} e
  \leq 2 \ind{b>\delta} b + 2\ind{c > \delta} c.
\]
We get the refined inequality
\begin{align*}
   & {\cal R}(f_n) - {\cal R}(g^*) \\&\quad\leq 4 c_\psi
  \E\bracket{\ind{2\norm{g_n(X) - g_n^*(X)} > d(g^*(X), F)}\norm{g_n(X) -
      g_n^*(X)}
    + \ind{2\norm{g_n^*(X) - g^*(X)} > d(g^*(X), F)}\norm{g_n^*(X) - g^*(X)}}.
\end{align*}
The first term is bounded by
\[
  \E\bracket{\ind{2\norm{g_n(X) - g_n^*(X)} > d(g^*(X), F)}\norm{g_n(X) -
      g_n^*(X)}} \leq \norm{g_n - g_n^*}_{L^1}.
\]
While for the second term, we proceed with
\begin{align*}
   & \E\bracket{\ind{2\norm{g_n^*(X) - g^*(X)} > d(g^*(X), F)}\norm{g_n^*(X)
      -g^*(X)}}
  \\&\qquad\leq \norm{g_n^* - g^*}_{L^\infty} \Pbb_X\paren{2\norm{g_n^*(X) - g^*(X)} >
    \inf_{x\in\supp\nu_\X} d(g^*(X), F)}.
\end{align*}
When weights are positive and sum to one, both $g_n^*(X)$ and $g^*(X)$ are averaging of $\phi(y)$ for $y\in\Y$, therefore
\[
  \norm{g_n^* - g^*}_{L^\infty} \leq 2 c_\phi.
\]
The same is true when $\sum_{i\leq n}\abs{\alpha_i(x)} \leq 1$.
Finally, when the labels are a deterministic function of the input, $g^*(X) = \phi(f^*(X))$, and $d(g^*(X), F) \leq \sup_{y\in\Y} d(\phi(y), F)$.
Defining $\delta := \sup_{y\in\Y} d(\phi(y), F) /2$, and adding everything together leads to Lemma~\ref{df:lem:cal}.

\subsection{Implication of Assumptions \ref{df:ass:non-ambiguity} and
  \ref{df:ass:lipschitz}}
\label{df:proof:ass}

Assume that Assumption \ref{df:ass:non-ambiguity} holds, consider $x\in\supp\nu_\X$, let us show that $f^*(x) = y_x$ and $\mu^*\vert_x = \delta_{y_x}$.
First of all, notice that \(\bigcap_{S;S\in\supp\nu\vert_x} = \brace{y_x} \); that $\delta_{y_x} \vdash \nu\vert_x$, as it corresponds to $\pi\vert_{x, S} = \delta_{y_x} \in \prob{S}$, for all $S$ in the support of $\nu\vert_{x}$; and that, because $\ell$ is well-behaved,
\[
  \inf_{z\in\Y} \ell(z, y_x) = \ell(y_x, y_x) = 0.
\]
This infimum is only achieved for $z = y_x$, hence if we prove that $\mu^*\vert_x = \delta_{y_x}$, we directly have that $f^*(x) = y_x$.
Finally, suppose that $\mu\vert_x\vdash\nu\vert_x$ charges $y \neq y_x$.
Because $y$ does not belong to all sets charged by $\nu\vert_x$, $\mu\vert_x$ should charge another $y'\in\Y$, and therefore
\[
  \inf_{z\in\Y} \E_{Y\sim\mu\vert_x}[\ell(z, y)] \geq
  \inf_{z\in\Y}\mu\vert_x(y) \ell(z, y) +
  \mu\vert_x(y') \ell(z, y') > 0.
\]
Which shows that $\mu^*\vert_x = \delta_{y_x}$.
We deduce that $g^*(x) = y_x$.

Now suppose that Assumption \ref{df:ass:lipschitz} holds too, and consider two $x, x' \in\supp\nu_\X$ belonging to two different classes $f(x) = y$ and $f(x') = y'$.
We have that $g^*(x) = \phi(y)$ and $g^*(x') =
  \phi(y')$, therefore,
\[
  d(x, x') \geq c^{-1} \norm{\phi(y) - \phi(y')}_{\cal H}.
\]
Define $h_2 = \inf_{y\neq y'} c^{-1} \norm{\phi(y) - \phi(y')}_{\cal H}$.
Let us now show that $h_2 > 0$.
When $\Y$ is finite, this infimum is a minimum, therefore, $h_2 = 0$, only if there exists a $y \neq y'$, such that $\phi(y) = \phi(y')$, which would implies that $\ell(\cdot, y) = \ell(\cdot,y')$ and therefore $\ell(y, y') = \ell(y, y)$ which is impossible when $\ell$ is proper.

\subsection{Proof of Theorem \ref{df:thm:convergence}}
\label{df:proof:convergence}

Reusing Lemma \ref{df:lem:cal}, we have
\[
  {\cal E}(f_n) \leq 4c_\psi \E_{{\cal D}_n, X}\bracket{\norm{g_n^*(X) - g_n(X)}_{\cal H}} +
  8c_\psi c_\phi \E_{{\cal D}_n, X}\bracket{\ind{\norm{g_n^*(X) - g^*(X)} > \delta}}.
\]
We will first prove that
\[
  \E_{{\cal D}_n}\bracket{\ind{\norm{g_n^*(X) - g^*(X)} > \delta}}
  \leq \exp\paren{-\frac{np}{8}}
\]
as long as $k < np / 2$.
The error between $g^*$ and $g_n$ relates to classical supervised learning of $g^*$ from samples $(X_i, Y_i) \sim \mu^*$.
We invite the reader who would like more insights on this fully supervised part of the proof to refer to the several monographs written on local averaging methods and, in particular, nearest neighbors, such as \citet{Biau2015}.
Because of class separation, we know that if $k$ points fall at distance at most $h$ of $x \in \supp\nu_\X$, $g_n^*(x) = k^{-1}\sum_{i; X_i\in{\cal N}(x)} \phi(y_i) = \phi(y_x) = g^*(x)$, where ${\cal N}(x)$ designs the $k$-nearest neighbors of $x$ in $(X_i)$.
Because the probability of falling at distance $h$ of $x$ for each $X_i$ is lower bounded by~$p$, we have that
\[
  \Pbb_{{\cal D}_n}(g_n^*(x) \neq g^*(x)) \leq \Pbb(\text{Bernoulli}(n, p) < k).
\]
This can be upper bound by $\exp(- np / 8)$ as soon as $k < np/2$, based on Chernoff multiplicative bound \citep[see][for a reference]{Biau2015}, meaning
\[
  \E_{{\cal D}_n, X} \bracket{\ind{\norm{g_n^*(X) - g^*(X)} \geq \delta}}
  \leq \exp(- np / 8).
\]

For the disambiguation part in $\norm{g_n - g_n^*}_{L^1}$, we distinguish two types of datasets, the ones where for any input $X_i$ its $k$-neighbors at are distance at least $h$, ensuring that disambiguation can be done by clusters, and datasets that does not verify this property.
Consider the event
\[
  \mathbb{D} = \brace{(X_i)_{i\leq n} \midvert \sup_{i} d(X_i, X_{(k)}(X_i)) < h}
\]
where $X_{(k)}(x)$ design the $k$-th nearest neighbor of $x$ in $(X_i)_{i\leq n}$.
We proceed with
\[
  \E_{{\cal D}_n, X}\bracket{\norm{g_n^*(X) - g_n(X)}_{\cal H}}
  \leq \sup_{X\in\X} \norm{g_n^* - g_n}_{\infty} \Pbb_{{\cal
      D}_n}((X_i) \notin\mathbb{D})
  + \E_{{\cal D}_n, X}\bracket{\norm{g_n^*(X) - g_n(X)}_{\cal H}\midvert (X_i) \in \mathbb{D}},
\]
Which is based on $E[Z] = \Pbb(Z\in A)\E[Z\vert A] + \Pbb(Z\notin A)\E[Z\vert ^cA]$.
For the term corresponding to bad datasets, we can bound the disambiguation error with the maximum error.
Similarly to the derivation for Lemma \ref{df:lem:cal}, because $g_n^*(x)$ and $g_n^*(X)$, are averaging of $\phi(y)$, we have that
\[
  \sup_{x\in\supp\nu_\X} \norm{g_n(x) - g_n^*(x)} \leq 2c_\phi.
\]
Indeed, we allow ourselves to pay the worst error on those datasets as their probability is really small, which can be proved based on the following derivation.
\begin{align*}
  \Pbb_{{\cal D}_n}((X_i)_{i\leq n} \notin\mathbb{D})
   & = \Pbb_{(X_i)}( \sup_{i} d(X_i, X_{(k)}(X_i)) \geq h)
  = \Pbb_{(X_i)}\paren{\cup_{i\leq n}\brace{d(X_i, X_{(k)}(X_i)) \geq h}}
  \\&\leq \sum_{i=1}^n \Pbb_{(X_i)}\paren{d(X_i, X_{(k)}(X_i)) \geq h}
  = n \Pbb_{X, {\cal D}_{n-1}}\paren{d(X, X_{(k)}(X)) \geq h}.
\end{align*}
This last probability has already been work out when dealing with the fully supervised part, and was bounded as
\[
  \Pbb_{X, {\cal D}_{n-1}}\paren{d(X, X_{(k)}(X)) \geq h} \leq
  \exp\paren{-(n-1)p / 8}.
\]
as long as $k < (n-1)p / 2$.
Finally, we have
\[
  \sup_{X\in\X} \norm{g_n^* - g_n}_{\infty}
  \Pbb_{{\cal D}_n}((X_i)_{i\leq n} \notin\mathbb{D})
  \leq 2c_\phi n \exp\paren{-(n-1)p / 8}.
\]

For the expectation term, corresponding to datasets ${\cal D}_n \in \mathbb{D}$ that cluster data accordingly to classes, we have to make sure that $\hat{y}_i = y_i^*$ is the only acceptable solution of \eqref{df:eq:disambiguation}, which is true as soon as the intersection of $S_j$, for $x_j$ the neighbors of $x_i$, only contained $y_i^*$.
To work out the disambiguation algorithm, notice that
\begin{align*}
  \norm{g_n - g_n^*}_{L^1}
   & =\int_\X \norm{\sum_{i=1}^n \alpha_i(x) \phi(\hat{y}_i) - \phi(y^*_i)} \diff\nu_\X(x)
  \leq \int_\X k^{-1}\sum_{i=1}^n \ind{X_i \in{\cal N}(x)} \norm{\phi(\hat{y}_i) - \phi(y^*_i)} \diff\nu_\X(x)
  \\ &= k^{-1}\sum_{i=1}^n \Pbb_X\paren{X_i \in{\cal N}(X)} \norm{\phi(\hat{y}_i) - \phi(y^*_i)}
  \leq 2c_\phi k^{-1}\sum_{i=1}^n \Pbb_X\paren{X_i \in{\cal N}(X)} \ind{\phi(\hat{y}_i) \neq \phi(y^*_i)}.
\end{align*}
Finally we have, after proper conditioning, considering the variability in $S_i$ while fixing $X_i$ first,
\begin{align*}
   & \E_{{\cal D}_n, X}\bracket{\norm{g_n^*(X) - g_n(X)}_{\cal H}\midvert (X_i) \in \mathbb{D}}
  \\&\qquad= 2c_\phi k^{-1}\E_{(X_i)}\bracket{\sum_{i=1}^n \Pbb_X\paren{X_i \in{\cal
        N}(X)} \E_{(S_i)}\bracket{\ind{\phi(\hat{y}_i) \neq
        \phi(y^*_i)} \midvert (X_i)} \midvert (X_i) \in \mathbb{D}}
  \\&\qquad= 2c_\phi k^{-1}\E_{(X_i), X}\bracket{\sum_{i=1}^n \ind{X_i \in{\cal
        N}(X)} \Pbb_{(S_i)}\paren{\phi(\hat{y}_i) \neq
      \phi(y^*_i) \midvert (X_i)} \midvert (X_i) \in \mathbb{D}}.
\end{align*}
We design $\mathbb{D}$ so that the $k$-th nearest neighbor of any input $X_i$ is at distance at most $h$ of $X_i$, meaning the because of class separation, $y_{x_i} \in S_j$ for any $X_j \in {\cal N}(X_i)$.
This mean that outputting $(\hat{y}_i) = (y^*_i)$ and $z_j = y_j$, will lead to an optimal error in \eqref{df:eq:disambiguation}.
Now suppose that there is another solution for \eqref{df:eq:disambiguation} such that $\hat{y}_i \neq y_i^*$, it should also achieve an optimal error, therefore it should verify $z_j = \hat{y}_j$ for all $j$ as well as $\hat{y}_j = \hat{y}_i$ for all $j$ such that $X_j$ is one of the $k$ nearest neighbors of $X_i$.
This implies that $\hat{y}_i \in \cap_{j;X_j \in {\cal N}(X_i)} S_j$, which happen with probability
\[
  \Pbb_{(S_j)_{j;X_j\in{\cal N}(X_i)}}(\exists z \neq y_i, z \in \cap_j S_j)
  \leq m \Pbb_{S_j}(z \in S_j)^k
  \leq m \eta^k = m\exp(-k\abs{\log(\eta)}).
\]
With $m = \card{\Y}$ the number of elements in $\Y$.
We deduce that
\[
  \Pbb_{(S_i)}\paren{\phi(\hat{y}_i) \neq \phi(y^*_i) \midvert (X_i)}
  \leq m \exp(-k\abs{\log(\eta)}).
\]
And because $\sum_{i=1}^n \ind{X_i \in {\cal N}(X)} = k$, we conclude that
\[
  \E_{{\cal D}_n, X}\bracket{\norm{g_n^*(X) - g_n(X)}_{\cal H}\midvert (X_i)
    \in \mathbb{D}} \leq 2c_\phi m \exp(-k\abs{\log(\eta)}).
\]
Finally, adding everything together we get
\[
  {\cal E}(f_n) \leq 8 c_\phi c_\psi \exp\paren{-\frac{np}{8}}
  + 8 c_\phi c_\psi n \exp\paren{-\frac{(n-1)p}{8}}
  + 8 c_\phi c_\psi m \exp\paren{-k\abs{\log(\eta)}}.
\]
as long as $k < (n-1) p /2$, which implies Theorem \ref{df:thm:convergence} as long as $n \geq 2$.

\begin{remark}[Other approaches]
  While we have proceeded with analysis based on local averaging methods, other paths could be explored to prove convergence results of the algorithm provided \eqref{df:eq:disambiguation} and \eqref{df:eq:estimate}.
  For example, one could prove Wasserstein convergence of $\sum_{i=1}^n\delta_{(x_i, \hat{y}_i)}$ toward $\sum_{i=1}^n \delta_{(x_i, \hat{y}^*_i)}$, together with some continuity of the learning algorithm as a function of those distributions.\footnote{The Wasserstein metric is useful to think in terms of distributions, which is natural when considering partial supervision that can be cast as a set of admissible fully supervised distributions.
    This approach has been successfully followed by \citet{Perchet2015} to deal with partial monitoring in games.}
  This analysis could be understood as tripartite:
  \begin{itemize}
    \item A disambiguation error, comparing $\hat{y}_i$ to $y_i^*$.
    \item A stability / robustness measure of the algorithm to learn $f_n$ from data
          when substituting $y_i^*$ by $\hat{y}_i$.
    \item A consistency result regarding $f_n^*$ learned with $(x_i, y_i^*)$.
  \end{itemize}
  Our analysis followed a similar path, yet with the first two parts tackled jointly.
\end{remark}

\subsection{Proof of Proposition \ref{df:prop:init}}
\label{df:proof:init}

Under the non-ambiguity hypothesis (Assumption \ref{df:ass:non-ambiguity}), the solution of \eqref{df:eq:solution} is characterized pointwise by $f^*(x) = y_x$ for all $x\in\supp\nu_\X$.
Similarly, under Assumption \ref{df:ass:non-ambiguity}, we have the characterization $f^*(x) \in \cap_{S\in\supp\nu\vert_x} S$.
With the notation of Definition \ref{df:def:init}, since $f^*(x)$ minimizes $z\to\E_{Y\sim\mu_S}[\ell(z, Y)]$ for all $S\in\supp\nu\vert_x$, it also minimizes $z\to\E_{S\sim\nu\vert_x}\E_{Y\sim\mu_S} [\ell(z, Y)]$.

For the second part of the proposition, we use the structured prediction framework of \citet{Ciliberto2020}.
Define the signed measure $\mu^\circ$ defined as $\mu^\circ_\X := \nu_\X$ and $\mu^\circ\vert_x := \E_{S\sim\nu\vert_x}\E_{Y\sim\mu_S}[\delta_Y]$, and $f^\circ:\X\to\Y$ the solution $f^\circ \in \argmin_{f:\X\to\Y}\E_{(X, Y)\sim\mu^\circ}[\ell(f(X), Y)] = \argmin_{f:\X\to\Y} \E_{(X, Y)\sim\nu}\bracket{\E_{Y\sim\mu_S}[\ell(f(X), Y)]}$.
The first part of the proposition tells us that $f^\circ = f^*$ under Assumption \ref{df:ass:non-ambiguity}.
The framework of \citet{Ciliberto2020}, tells us that $f^\circ$ is obtained after decoding \eqref{df:eq:decoding} of $g^\circ:\X\to{\cal H}$, and that if $g_n^\circ$ converges to $g^\circ$ with the $L^1$ norm, $f_n^\circ$ converges to $f^\circ$ in terms of the $\mu^\circ$-risk.
Under Assumption \ref{df:ass:non-ambiguity} and mild hypothesis on $\mu^\circ$, it is possible to prove that convergence in terms of the $\mu^\circ$-risk implies convergence in terms of the $\mu$-risk (for example through calibration inequality similar to Proposition 2 of \citet{Cabannes2020}).

\subsection{Ranking with partial ordering is a well-behaved problem}
\label{df:proof:ranking}

Here, we discuss building directly $\xi_S$ to initialize our alternative minimization scheme or considering $\mu_S$ given by the definition of well-behaved problem (Definition \ref{df:def:init}).
Since the existence of $\mu_S$ implying $\xi_S$ defined as $\E_{Y\sim\mu_S}[\phi(Y)]$, we will only study when $\xi_S$ can be cast as a $\mu_S$.

In ranking, we have that $\psi = -\phi$, which corresponds to ``correlation losses''.
In this setting, we have that $\Span(\phi(\Y)) = \Span(\psi(\Y))$.
More generally, looking at a ``minimal'' representation of $\ell$, one can always assume the equality of those spans, as what happens on the orthogonal of the intersection of those spans, does not modify the scalar product $\phi(y)^\top\psi(z)$.
Similarly, $\xi_S$ can be restricted to $\Span(\psi(\Y))$, and therefore $\Span(\phi(\Y))$, which exactly the image by $\mu\to\E_{Y\sim\mu}[\phi(Y)]$ of the set of signed measures, showing the existence of a $\mu_S$ matching Definition \ref{df:def:init}.
  \section{IQP implementation}
\label{df:app:diffrac}

In this section, we introduce an IQP implementation to solve for \eqref{df:eq:disambiguation}.
We first mention that our alternative minimization scheme is not restricted to well-behaved problems, before motivating the introduction of the IQP algorithm in two different ways, and finally describing its implementation.

\subsection{Initialization of alternative minimization for non-well-behaved problem}
Before describing the IQP implementation to solve \eqref{df:eq:algo}, we would like to stress that, even for non-well-behaved partial labeling problems, it is possible to search for smart ways to initialize variables of the alternative minimization scheme.
For example, one could look at $z_i^{(0)} \in \cap_{j;x_j\in{\cal N}_{k_i}} S_j$, where ${\cal N}_k$ designs the $k$ nearest neighbors of $x_i$ in $(x_j)_{j\leq n}$, and $k_i$ is chosen such that this intersection is a singleton.

\subsection{Link with Diffrac and empirical risk minimization}
Our IQP algorithm is similar to an existing disambiguation algorithm known as the Diffrac algorithm \citep{Bach2007,Joulin2010}.\footnote{The Diffrac algorithm was first introduced for clustering, which is a classical approach to unsupervised learning.
  In practice, it consists of changing the constraint set $C_n = \prod S_i$ by a set of the type $C_n = \argmax_{(y_i) \in \Y^n} \sum_{i,j=1}^n \ind{y_i\neq y_j}$ in \eqref{df:eq:disambiguation} and \eqref{df:eq:iqp}, meaning that $(y_i)$ should be disambiguated into different classes.}
This algorithm was derived by implicitly following empirical risk minimization of \eqref{df:eq:principle}.
This approach leads to algorithms written as
\[
  (y_i) \in \argmin_{(y_i) \in C_n} \inf_{f\in{\cal F}} \sum_{i=1}^n \ell(f(x_i), y_i)
  + \lambda \Omega(f),
\]
for ${\cal F}$ a space of functions, and $\Omega:{\cal F} \to \R_+$ a measure of complexity.
Under some conditions, it is possible to simplify the dependency in $f$ \citep[{\em e.g.},][]{Xu2004,Bach2007}.
For example, if $\ell(y, z)$ can be written as $\norm{\phi(y) - \phi(z)}^2$ for a mapping $\phi:\X\to\Y$, {\em e.g.} the Kendall loss detailed in Section \ref{df:sec:ranking},\footnote{Since $\norm{\phi(y)}$ is constant.} and the search of $\phi(f):\X\to\phi(\Y)$ is relaxed as a $g:\X\to{\cal H}$.
With $\Omega$ and ${\cal F}$ linked with kernel regression on the surrogate functional space $\X\to{\cal H}$, it is possible to solve the minimization with respect to $g$ as $g(x_i) = \sum_{j=1}^n \alpha_j(x_i)\phi(y_i)$, with $\alpha$ given by kernel ridge regression \citep{Ciliberto2016}, and to obtain a disambiguation algorithm written as
\[
  \argmin_{y_i \in S_i} \sum_{i=1}^n\big\|\sum_{j=1}^n \alpha_j(x_i) \phi(y_j) - \phi(y_i)\big\|^2.
\]
This IQP is a special case of the one we will detail. As such, our IQP is a generalization of the Diffrac algorithm, and this paper provides, to our knowledge, {\em the first consistency result for Diffrac}.

\subsection{Link with another determinism measure}
While we have considered the measure of determinism given by \eqref{df:eq:principle}, we could have considered its quadratic variant
\[
  \mu^\star \in \argmin_{\mu\vdash\nu} \inf_{f:\X\to\Y} \E_{X\sim\nu_\X}
  \bracket{\E_{Y, Y'\sim\mu\vert_x}\bracket{\ell(Y, Y')}}.
\]
This corresponds to the right drawing of Figure \ref{df:fig:objective}.
We could arguably translate it experimentally as
\begin{equation}
  \label{df:eq:iqp}
  (\hat{y}_i) \in \argmin_{(y_i) \in C_n} \sum_{i,j=1}^n \alpha_i(x_j)
  \ell(y_i, y_j),
\end{equation}
and still derive Theorem \ref{df:thm:convergence} when substituting \eqref{df:eq:disambiguation} by \eqref{df:eq:iqp}.
When the loss is a correlation loss $\ell(y, z) = -\phi(y)^\top\phi(z)$.
This leads to the quadratic problem
\[
  (\hat{y}_i) \in \argmin_{(y_i) \in C_n} -\sum_{i,j=1}^n \alpha_i(x_j)
  \phi(y_i)^\top \phi(y_j).
\]

\subsection{IQP implementation}
In order to make our implementation possible for any symmetric loss $\ell:\Y\times\Y\to\R$, on a finite space $\Y$, we introduce the following decomposition.

\begin{proposition}[Quadratic decomposition]
  When $\Y$ is finite, any proper symmetric loss $\ell$ admits a decomposition with two mappings $\phi:\Y\to\R^m$, $\psi:\Y\to\R^m$, for an $m\in\N$ and a $c\in\R$, reading
  \begin{equation}
    \label{df:eq:qloss}
    \forall\, y, z\in \Y, \quad\ell(y, z) = \psi(y)^\top\psi(z) - \phi(y)^\top
    \phi(z)
    \qquad \text{with}\qquad \norm{\phi(y)} = \norm{\psi(y)} = c
  \end{equation}
\end{proposition}
\small
\begin{proof}
  Consider $\Y = {y_1, \cdot, y_m}$ and $L = (\ell(y_i, y_j))_{i,j\leq m} \in \R^{m\times m}$.
  $L$ is a symmetric matrix, diagonalizable as
  \(
  L = \sum_{i=1}^m \lambda_i u_i\otimes u_i,
  \)
  with $(u_i)$ an orthonormal basis of $\R^m$, and $\lambda_i \in \R$ its eigenvalues.
  We have, with $(e_i)$ the Cartesian basis of $\R^m$,
  \[
    \ell(y_j, y_k) = L_{jk} = \scap{e_j}{Le_k} = \sum_{i=1}^m (\lambda_i)_+
    \scap{e_j}{u_i} \scap{e_k}{u_i}
    - \sum_{i=1}^m (\lambda_i)_- \scap{e_j}{u_i} \scap{e_k}{u_i}.
  \]
  We build the decomposition
  \[
    \tilde{\psi}(y_k) = \paren{\sqrt{(\lambda_i)_+} \scap{e_k}{u_i}}_{i\leq m},
    \qquad\text{and}\qquad
    \tilde{\phi}(y_k) = \paren{\sqrt{(\lambda_i)_-} \scap{e_k}{u_i}}_{i\leq m}.
  \]
  It satisfies
  \(
  \ell(y_j, y_k) = \tilde\psi(y)^\top\tilde\psi(z) - \tilde\phi(y)^\top\tilde\phi(z).
  \)
  We only need to show that we can consider $\phi$ of constant norm.
  For this, first consider $C = \max_i\abs{\lambda_i}$, we have
  \(
  \norm{\tilde{\psi}(y_k)}^2 = \sum_{i=1}^m (\lambda_i)_+ \scap{u_i}{e_k}^2
  \leq C \sum_{i=1}^m \scap{u_i}{e_k}^2 = C\norm{e_k}^2 = C
  \)
  The last equalities being due to the fact that $(u_i)$ is orthonormal.
  Now, introduce the correction vector $\xi:\Y\to\R^m$,
  \(
  \xi(y_i) = \sqrt{C - \norm{\tilde\psi(y)}^2} e_i.
  \)
  And consider $\phi = \binom{\tilde\phi}{\xi}$, $\psi = \binom{\tilde\psi}{\xi}$.
  By construction, $\psi$ is of constant norm being equal to $C$ and that $\ell(y, z) = \psi(y)^T\psi(z) - \phi(y)^T\phi(z)$.
  Finally, because $\ell(y, z) = 0$, we also have $\phi$ of constant norm.
\end{proof}
\normalsize

Using the decomposition \eqref{df:eq:qloss}, \eqref{df:eq:iqp} reads, with
$\textbf{y} = (y_i)$
\[
  {\bf \hat y} \in \argmin_{{\bf y}\in C_n}\sum_{i=1}^n \alpha_i(x_j) \psi(y_i) \psi(y_j) - \sum_{i=1}^n \alpha_i(x_j) \phi(y_i) \phi(y_j).
\]
By defining the matrix $A = (\alpha_i(x_j))_{ij\leq n} \in \R^{n\times n}$, $\Psi({\bf y}) = (\psi(y_i))_{i\leq n} \in \R^{n\times m}$ and $\Phi({\bf y}) = (\phi(y_i))_{i\leq n} \in \R^{n\times m}$, we cast it as
\[
  {\bf \hat y} \in \argmin_{{\bf y}\in C_n}\trace\paren{A\Psi({\bf y})\Psi({\bf y})^\top} - \trace\paren{A\Phi({\bf y})\Phi({\bf y})^\top}.
\]

\paragraph{Objective convexification.} As $\alpha_i(x_j)$ is a measure of similarity between $x_i$ and $x_j$, $A$ is usually symmetric positive definite, making this objective convex in $\Psi$ and concave in $\Phi$.
However, recalling \eqref{df:eq:qloss}, we have $\trace\Phi\Phi^\top = \trace{\Psi\Psi^\top} = n c$, therefore considering the spectral norm of $A$, we convexify the objective as
\[
  {\bf \hat y} \in \argmin_{{\bf y}\in C_n}\trace\paren{(\norm{A}_*I +
    A)\Psi({\bf y})\Psi({\bf y})^\top} + \trace\paren{(\norm{A}_*I - A)\Phi({\bf y})\Phi({\bf y})^\top}.
\]
Considering
\[
  B = \paren{\begin{array}{cc} \norm{A}_*I + A & 0 \\ 0 & \norm{A}_*I - A \end{array}}
  \qquad\text{and}\quad
  \Xi({\bf y}) = \binom{\Psi({\bf y})}{\Phi({\bf y})},
\]
simplifies this objective as
\[
  {\bf \hat y} \in \argmin_{{\bf y}\in C_n}\trace\paren{B\Xi({\bf y})\Xi({\bf y})^\top}.
\]
When parametrized by $\xi = \Xi({\bf y})$, this is an optimization problem with a convex quadratic objective and ``integer-like'' constraint $\xi \in \Xi(C_n)$, identifying to an integer quadratic program (IQP).

\paragraph{Relaxation.} IQP are known to be NP-hard, several tools exist in literature and optimization libraries implementing them.
The most classical approach consists in relaxing the integer constraint $\xi \in \Xi(C_n)$ into the convex constraint $\xi \in \hull(\Xi(C_n))$, solving the resulting convex quadratic program, and projecting back the solution toward an extreme of the convex set.
Arguably, our alternative minimization approach is a better grounded heuristic to solve our specific disambiguation problem.
  \section{Example with graphical illustrations}
\label{df:app:example}

To ease the understanding of the disambiguation principle \eqref{df:eq:principle}, we provide a toy example with a graphical illustration, Figure~\ref{df:fig:objective}.
Since \eqref{df:eq:principle} decorrelates inputs, we will consider $\X$ to be a singleton, in order to remove the dependency to $\X$.
In the following, we consider $\Y = \brace{a, b, c}$, with the loss given by
\[
  L = (\ell(y, z))_{y, z\in \Y} =
  \paren{\begin{array}{ccc}
      0 & 1 & 1 \\
      1 & 0 & 2 \\
      1 & 2 & 0
    \end{array}}.
\]
This problem can be represented on a triangle through the embedding of probability measures reading $\xi:\prob{\Y}\to \R^3; \mu\to\mu(a)e_1 + \mu(b)e_2 + \mu(c)e_3$, and onto the triangle $\brace{z\in\R_+^3 \midvert z^\top 1 = 1}$. Note that $\xi$ can be extended from any signed measure of total mass normalized to one onto the plane $\brace{z\in\R^3 \midvert z^\top 1 = 1}$, as well as the drawings Figure \ref{df:fig:objective} can be extended onto the affine span of the represented triangles.
The objective \eqref{df:eq:principle} reads pointwise as $\prob{\Y}\to\R; \mu \to \min_{i\leq 3} e_i^\top L \xi(\mu)$, while its quadratic version reads $\prob{\Y}\to\Y; \mu\to \xi(\mu)^\top L \xi(\mu)$.
Note that while $L$ is not definite negative, one can check that the restriction of $\R^3 \to \R; z \to z^\top Lz$ to the definition domain $\brace{z\in\R^3\midvert z^\top 1 = 1}$ is concave, as suggested by the right drawing of Figure \ref{df:fig:objective}.

\begin{figure*}[t]
  \centering
  \begin{tikzpicture}[scale=2.5]
  \coordinate(a) at (0, 0);
  \coordinate(b) at ({1/2}, {sin(60)});
  \coordinate(c) at (1, 0);
  \coordinate(ha) at ({3/4}, {sin(60)/2});
  \coordinate(hb) at ({1/2}, 0);
  \coordinate(hc) at ({1/4}, {sin(60)/2});

  \coordinate (mc) at (1,-.25);
  \coordinate (mca) at (0,-.25);
  \fill[fill=white] (a) -- (c) -- (mc) -- (mca) -- cycle;

  \fill[fill=red!20] (a) -- (hb) -- (ha) -- (hc) -- cycle;
  \fill[fill=green!20] (b) -- (ha) -- (hc) -- cycle;
  \fill[fill=blue!20] (c) -- (ha) -- (hb) -- cycle;
  \draw (a) node[anchor=north east]{a} -- (b) node[anchor=south]{b} --
  (c) node[anchor=north west]{c} -- cycle;
  \draw (hb) -- (ha) -- (hc);
  \node at ({3/8}, {sin(60)/4}) {$R_a$};
  \node at ({1/2}, {3*sin(60)/4 - 1/16}) {$R_b$};
  \node at ({3/4}, {sin(60)/4 - 1/16}) {$R_c$};
\end{tikzpicture}
\begin{tikzpicture}[scale=2.5]
  \coordinate(a) at (0, 0);
  \coordinate(b) at ({cos(60)}, {sin(60)});
  \coordinate(c) at (1, 0);
  \coordinate(r) at ({1/4}, 0);
  \coordinate(s) at ({7/32 + 1/8}, {7*sin(60)/16});
  \coordinate(t) at ({3/8 + 1/4}, {3*sin(60)/4});

  \coordinate (mc) at (1,-.25);
  \coordinate (mca) at (0,-.25);
  \fill[fill=white] (a) -- (c) -- (mc) -- (mca) -- cycle;

  \fill[fill=black!20] (c) -- (r) -- (s) -- (t) -- cycle;
  \fill[fill=black!10] (a) -- (r) -- (s) -- (t) -- (b) -- cycle;
  \draw (a) node[anchor=north east]{a} -- (b) node[anchor=south]{b} --
  (c) node[anchor=north west]{c} -- cycle;
  \draw (c) -- (r) -- (s) -- (t) -- cycle;
  \node at ({5/8}, {sin(60)/3 - 1/16}) {$R_{\nu}$};
\end{tikzpicture}
\begin{tikzpicture}[scale=2.5]  
  \coordinate (a) at (0,0) ;
  \coordinate (b) at ({1/2},{sin(60)}) ;
  \coordinate (c) at (1,0);
  \coordinate (mb) at (.25, {sin(60)});
  \coordinate (mba) at ({-.25*cos(30)},{.25*sin(30)});
  \coordinate (mc) at (1,-.25);
  \coordinate (mca) at (0,-.25);
  \coordinate(ha) at ({3/4}, {sin(60)/2});
  \coordinate(hb) at ({1/2}, 0);
  \coordinate(hc) at ({1/4}, {sin(60)/2});

  \foreach \x in {0,.05,...,.25}
  \draw[gray, rotate=30] ({sin(60)}, \x) -- ({sin(60) - (tan(60)*\x)}, \x) -- ({sin(60) - (tan(60)*\x)}, -\x) -- ({sin(60)}, {-\x}); 
  \foreach \x in {0.25,.3,...,.5}
  \draw[gray, rotate=30] ({sin(60)}, \x) -- ({sin(60)-tan(60)*(.5-\x)}, \x);
  \foreach \x in {0.25,.3,...,.5}
  \draw[gray, rotate=30] ({sin(60)-tan(60)*(.5-\x)}, -\x) -- ({sin(60)}, {-\x}); 
  \foreach \x in {.25,.3,...,.5}
  \draw[gray, rotate=30] ({sin(60) - (tan(60)*\x)}, .25) -- ({sin(60) - (tan(60)*\x)}, -.25); 

  \fill[fill=white] (a) -- (c) -- (mc) -- (mca) -- cycle;
  \fill[fill=white] (a) -- (b) -- (mb) -- (mba) -- cycle;
  \draw (a) node[anchor=north east]{a} -- (b) node[anchor=south]{b} -- (c) node[anchor=north west]{c} -- cycle;
  \draw[dotted, thick] (hc) -- (ha) -- (hb);
\end{tikzpicture}
\begin{tikzpicture}[scale=2.5]
  \coordinate (a) at (0,0) ;
  \coordinate (b) at ({1/2},{sin(60)}) ;
  \coordinate (c) at (1,0) ;
  \coordinate (mb) at (.25, {sin(60)});
  \coordinate (mba) at ({-.25*cos(30)},{.25*sin(30)});
  \coordinate (mc) at (1,-.25);
  \coordinate (mca) at (0,-.25);
  \coordinate(ha) at ({3/4}, {sin(60)/2});
  \coordinate(hb) at ({1/2}, 0);
  \coordinate(hc) at ({1/4}, {sin(60)/2});

  \foreach \x in {0,.05,...,.5}
  \draw[gray,rotate=30] ({sin(60)}, 0) + (0, {\x}) arc (90:270:{sqrt(3)*\x} and {\x});

  \fill[fill=white] (a) -- (c) -- (mc) -- (mca) -- cycle;
  \fill[fill=white] (a) -- (b) -- (mb) -- (mba) -- cycle;
  \draw (a) node[anchor=north east]{a} -- (b) node[anchor=south]{b} -- (c) node[anchor=north west]{c} -- cycle;
  \draw[dotted, thick] (hc) -- (ha) -- (hb);
\end{tikzpicture}
  \caption{Exposition of a pointwise problem in the simplex $\prob\Y$, with $\Y = \brace{a, b, c}$ and a proper symmetric loss defined by $\ell(a, b) = \ell(a, c) = \ell(b, c) / 2$.
    (Left) Representation of the decision regions $R_z = \brace{\mu\in\prob{\Y}\midvert z\in \argmin_{z'\in\Y} \E_{Y\sim\mu}[\ell(z, y)]}$ for $z\in\Y$.
    (Middle Left) Representation of $R_\nu = \brace{\mu\in\prob{\Y} \midvert \mu\vdash\nu}$ for $\nu = (5\delta_{\brace{a, b, c}} + \delta_{\brace{c}} + \delta_{\brace{a, c}} + \delta_{\brace{b, c}}) / 8$.
    (Middle Right) Level curves of the piecewise function $\prob{\Y}\to\R; \mu\to\min_{z\in\Y}\E_{Y\sim\mu}[\ell(z, Y)]$ corresponding to \eqref{df:eq:principle}.
    (Right) Level curves of the quadratic function $\prob{\Y}\to\R; \mu\to \E_{Y, Y'\sim\mu}[\ell(Y, Y')]$. Our disambiguation \eqref{df:eq:principle} corresponds to minimizing the concave function represented in the middle right drawing on the convex domain represented in the middle left drawing.}
  \label{df:fig:objective}
\end{figure*}

It should be noted that $(\ell, \nu)$ being a well-behaved partial labeling problem can be understood graphically, as having the intersection of the decision regions $\cap_{z\in S} R_z$ non-empty for any set $S$ in the support of $\nu$.
As such, it is easy to see that our toy problem is well-behaved for any distribution $\nu$.
Formally, to match Definition \ref{df:def:init}, we can define $\mu_{\brace{e}} = \delta_e$ for $e\in\brace{a, b, c}$ and
\[
  \mu_{\brace{a,b}} = .5\delta_a + .5\delta_b,\quad
  \mu_{\brace{a,c}} = .5\delta_b + .5\delta_c,\quad
  \mu_{\brace{b, c}} = \delta_b + \delta_c - \delta_a,\quad
  \mu_{\brace{a, b, c}} = .5\delta_b + .5\delta_c.
\]
Graphically $\xi(\mu_{\brace{a,b}})$ can be chosen as any points on the horizontal dashed line on the middle right drawing of Figure \ref{df:fig:objective} (similarly for $\xi\mu_{\brace{a, c}}$), while $\xi(\mu_{\brace{a, b, c}})$ has to be chosen has the intersection $.5e_2 + .5e_3$, and while $\xi(\mu_{\brace{b,c}})$ has to be chosen outside the simplex on the half-line leaving $.5e_2 + .5e_3$ supported by the perpendicular bisector of $[e_2, e_3]$ and not containing $e_1$.

  \section{Experiments}
\label{df:app:experiments}
While our results are much more theoretical than experimental, out of principle, as well as for reproducibility, comparison and usage sake, we detail our experiments.

\subsection{Interval regression - Figure \ref{df:fig:ir}}
Figure \ref{df:fig:ir} corresponds to the regression setup consisting of learning $f^*:[0, 1]\to\R; x\to\sin(\omega x)$, with $\omega=10\approx 3\pi$.
The dataset represented on Figure \ref{df:fig:ir} is collected in the following way.
We sample $(x_i)_{i\leq n}$ with $n = 10$, uniformly at random on $\X=[0, 1]$, after fixing a random seed for reproducibility.
We collect $y_i = f(x_i)$.
We create $(s_i)$ by sampling $u_i$ uniformly on $[0,1]$, defining $r_i = r - \gamma \log(u_i)$, with $r=1$ and $\gamma = 3^{-1}$, sampling $c_i$ uniformly at random on $[0, r_i]$, and defining $s_i = y_i + \sign(y_i)\cdot c_i + [-r_i, r_i]$.
The corruption is skewed on purpose to showcase disambiguation instability of the baseline \eqref{df:eq:baseline} compared to our method.
We solve \eqref{df:eq:disambiguation} with alternative minimization, initialized by taking $y_i^{(0)}$ at the center of $s_i$, and stopping the minimization scheme when $\sum_{i\leq n}\vert y_i^{(t+1)} - y_i^{(t)}\vert < \epsilon$ for $\epsilon$ a stopping criterion fixed to $10^{-6}$.
For $x\in \X$, the inference \eqref{df:eq:estimate} and \eqref{df:eq:baseline} is done through grid search, considering, for $f_n(x)$, 1000 guesses dividing uniformly $[-6, 6] \subset \Y = \R$.
We consider weights $\alpha$ given by kernel ridge regression with Gaussian kernel, defined as
\[
  \alpha(x) = (K+n\lambda I)^{-1}K_x \in \R^n, \quad
  K = (k(x_i, x_j))_{i,j\leq n} \in \R^{n\times n},\quad
  K_x = (k(x_i, x))_{i\leq n} \in \R^n,
\]
with $k(x, x') = \exp\paren{-\frac{\norm{x-x'}^2}{2\sigma^2}}$, and $\lambda$ a regularization parameter, and $\sigma$ a standard deviation parameter.
In our simulation, we fix $\sigma = .1$ based on simple considerations on the data, while we consider $\lambda \in [10^{-1}, 10^{-3}, 10^{-6}]$.
The evaluation of the mean square error between $f_n$ and $f^*$, which is equivalent to evaluating the risk with the regression loss $\ell(y, z) = \norm{y - z}^2$, is done by considering 200 points dividing uniformly $\X = [0,1]$ and evaluating $f_n$ and $f^*$ on it.
The best hyperparameter $\lambda$ is chosen by minimizing this error.
It leads to $\lambda = 10^{-1}$ for the baseline \eqref{df:eq:baseline}, and $\lambda = 10^{-6}$ for our algorithm \eqref{df:eq:disambiguation} and \eqref{df:eq:estimate}.
This difference in $\lambda$ is normal since both methods are not estimating the same surrogate quantities.
The fact that $\lambda$ is smaller for our algorithm is natural as our disambiguation objective \eqref{df:eq:disambiguation} already has a regularization effect on the solution.\footnote{%
  Moreover, the analysis in \citet{Cabannes2020} suggests that the baseline is estimating a surrogate function in $\X\to 2^\R$, while our method is estimating a function in $\X\to\R$, which is a much smaller function space, hence needing less regularization.
  However, those reflections are based on upper bounds, that might be suboptimal, which could invalidate those considerations.}
Note that we used the same weights $\alpha$ for \eqref{df:eq:disambiguation} and \eqref{df:eq:estimate}, which is suboptimal, but fair to the baseline, as, consequently, both methods have the same number of hyperparameters.

\subsection{Classification - Figure \ref{df:fig:cl}}
Figure \ref{df:fig:cl} corresponds to classification problems, based on real datasets from the LIBSVM datasets repository.
At the time of writing, the datasets are available at \url{https://www.csie.ntu.edu.tw/~cjlin/libsvmtools/datasets/multiclass.html}.
We present results on the ``DNA'' and ``Svmguide2'' datasets, that both have 3 classes ($m=3$), and respectively have 4000 samples with 180 features ($n=4000$,$d=180$) and 391 samples with 20 features ($n = 391$, $d=20$).

In terms of {\em complexity}, when $\Y = \bbracket{1, m} = \brace{1, 2, \cdots, m}$, and weights based on kernel ridge regression with Gaussian kernel as described in the last paragraph the complexity of performing inference for \eqref{df:eq:estimate} and \eqref{df:eq:baseline} can be done in $O(nm)$ in time and $O(n+m)$ in space, where $n$ is the number of training samples \citep{Nowak2019,Cabannes2020}.
The disambiguation \eqref{df:eq:disambiguation} performed with alternative minimization is done in $O(cn^2m)$ in time and in $O(n(n+m))$ in space, with $c$ the number of steps in the alternative minimization scheme.
In practice, $c$ is really small, which can be understood since we are minimizing a concave function and each step leads to a guess on the border of the constraint domain.

Based on the dataset $(x_i, y_i)$, we create $(s_i)$ by sampling it accordingly to $\gamma \delta_{\brace{y_i}} + 1-\gamma \delta_{\brace{y, y_i}}$, with $y$ the most present labels in the dataset (indeed we choose the two datasets because they were not too big and presenting unequal labels proportion), and $\gamma\in[0,1]$ the corruption parameter represented in percentage on the $x$-axis of Figure \ref{df:fig:cl}.
This skewed corruption allows distinguishing methods and invalidates the simple approach consisting of averaging candidate (AC) in set to recover $y_i$ from $s_i$, which works well when data are {\em missing at random} \citep{Heitjan1991}.
We separate $(x_i, s_i)$ in 8 folds, consider $\sigma \in d\cdot[1, .1, .01]$, where $d$ is the dimension of $\X$, and $\lambda \in n^{-1/2}\cdot [1, 10^{-3}, 10^{-6}]$, where $n$ is the number of data. We tested the different hyperparameter setup and reported the best error for each corruption parameter on Figure \ref{df:fig:cl}.
Those errors are measured with the 0-1 loss, computed as averaged over the 8 folds, {\em i.e.} cross-validated, with standard deviation represented as error bars on the figure.
The best hyperparameter generally corresponds to $\sigma = .1$ and $\lambda = 10^{-3}$ when the corruption is small and $\sigma = 1$, $\lambda = 10^{-3}$ when the corruption is big.
Differences between cross-validated error and testing error were small, and we presented the first one out of simplicity.

In terms of {\em energy cost}, the experiments were run on a personal laptop that has two processors, each of them running 2.3 billion instructions per second.
During experiments, all the data were stored on the random access memory of 8 GB.
Experiments were run on Python, extensively relying on the NumPy library \citep{Harris2020}.
The heaviest computation is Figure \ref{df:fig:cl}.
Its total runtime, cross-validation included, was around 70 seconds.
This paper is the result of experimentation, we evaluate the total cost of our experimentation to be three orders of magnitude higher than the cost of reproducing the final computations presented on Figure \ref{df:fig:ir}, \ref{df:fig:cl} and \ref{df:fig:ss}.
The total computational energy cost is negligible.

\subsection{Semi-supervised learning - Figure \ref{df:fig:ss}}

\begin{figure*}[ht]
  \centering
  \includegraphics{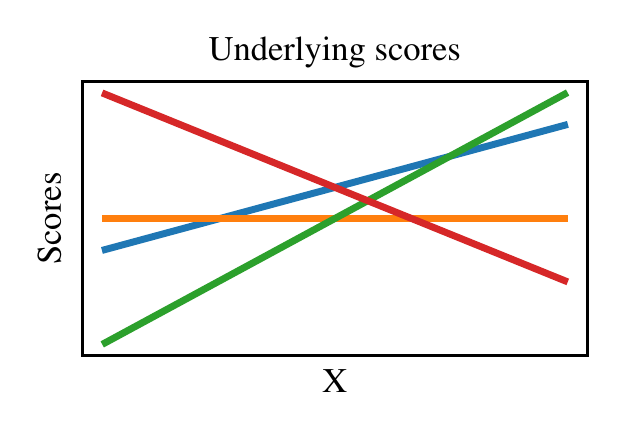}
  \includegraphics{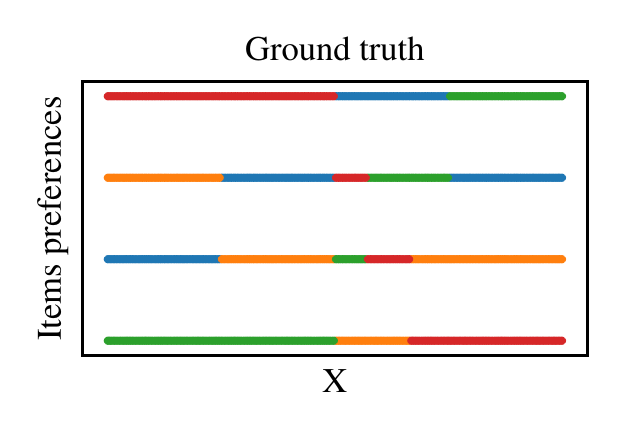}
  \caption{
    Ranking setting.
    We consider $\X$ an interval of $\R$, and $\Y = \Sfrak_m$ with $m=4$ on the figure.
    (Right) To create a ranking dataset, we sample randomly $m$ lines in $\R^2$, embedding a value, or equivalently a score, associated to each item as a function of the input $x$.
    (Left) By ordering those lines, we create preferences between items as a function of $x$. On the figure, when $x$ is small, the ``red'' item is preferred over the ``orange'' item, itself preferred over the ``blue'' item, itself preferred over the ``green'' item. While when $x$ is big, ``green'' is preferred over ``blue'', preferred over ``orange'', preferred over ``red''.
    We create a partial labeling dataset by sampling $(x_i) \in \X^n$, and providing only partial ordering that the $(y_i)$ follow.
    For example, for a small $x$, we might only give the partial information that ``red'' is preferred over ``blue''.
  }
  \label{df:fig:rk_set}
\end{figure*}

\begin{figure*}[ht!]
  \centering
  \includegraphics{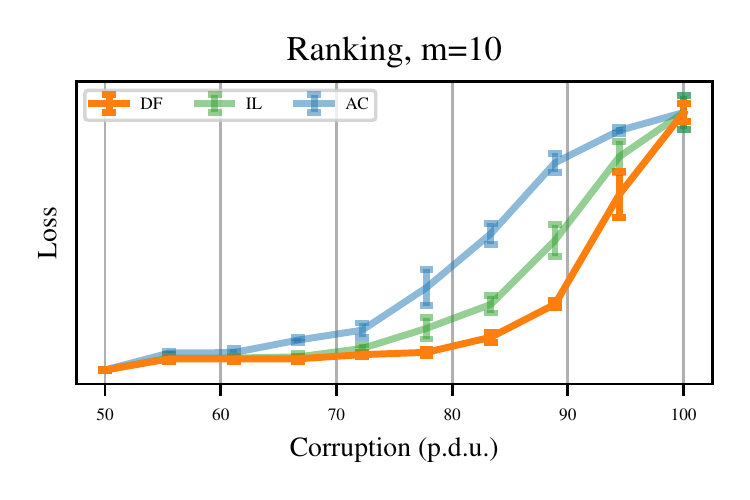}
  \caption{
    Performance of our algorithm for ranking with partial ordering.
    This figure is similar to Figure \ref{df:fig:cl}, but is based on the ranking problem illustrated on Figure \ref{df:fig:rk_set}.
    For this figure, we consider $m = 10$, as it is arguably the limit where the LP relaxation provided by \citet{Cabannes2020} of the NP-hard minimum feedback arcset problem still performs well.
    The corruption parameter corresponds to the proportion of coordinates lost in the Kendall embedding when creating $s_i$ from $y_i$.
    Because the Kendall embedding satisfies transitivity constraints, a corruption smaller than 50\% is almost ineffective to remove any information.
    In this figure, we observe a similar behavior for ranking to the one observed for classification on Figure \ref{df:fig:cl}, suggesting that those empirical findings are not spurious.
  }
  \label{df:fig:rk}
\end{figure*}

On Figure \ref{df:fig:ss}, we review a semi-supervised classification problem with $\Y = \bbracket{1, 4}$, $\X = [-4.5, 4.5]^2$, $\mu_\X$ only charging $\brace{x=(x_1, x_2)\in\R^2\midvert x_1^2 + x_2^2 \in \N^*}$ and the solution $f^*:\X \to \Y$ being defined almost everywhere as $f^*(x) = x_1^2 + x_2^2$.
We collect a dataset $(x_i, s_i)$, by sampling 2000 points $\theta_i$ uniformly at random on $[0, 1]$, as well as $r_i$ uniformly at random in $\bbracket{1, 4} = \brace{1, 2, 3, 4}$, before building $x_i = r_i\cdot (\cos(2\pi \theta_i), \sin(2\pi\theta_i)) \in \X$, and $s_i = \Y$.
We add four labeled points to this dataset $x_{2001} = (-2\sqrt{3}, 2)$ with $s_{2001} = \brace{4}$, $x_{2001} = (1, -2\sqrt{2})$ with $s_{2002} = \brace{3}$, $x_{2001} = (-\sqrt{3}, -1)$ with $s_{2003} = \brace{2}$ and $x_{2001} = (-1, 0)$ with $s_{2004} = \brace{1}$.
We designed the weights $\alpha$ in \eqref{df:eq:disambiguation} with $k$-nearest neighbors, with $k=20$, and solve this equation with a variant of alternative minimization, leading to the optimal solution $\tilde{y}_i = y_i^*$.
In order to be able to compute the baseline \eqref{df:eq:baseline}, we design weights $\alpha$ for the inference task based on Nadaraya-Watson estimators with Gaussian kernel, defined as $\alpha_i(x) = \exp\paren{\norm{x-x_i}^{2} / h}$, with $h = .08$. We solve the inference task on a grid of $\X$ composed of 2500 points, and artificially recreate the observations to make them neat and reduce the resulting PDF size.
Note that it is possible to design weights $\alpha$ that capture the cluster structure of the data, which, in this case, will lead to a nice behavior of the baseline as well as our algorithm.
Arguably, this experiment showcases a regularization property of our algorithm \eqref{df:eq:disambiguation}.

\subsection{Ranking with partial ordering}

To conclude this experiment section, we look at ranking with partial ordering.
We refer to Section \ref{df:sec:ranking} for a clear description of this instance of partial labeling.
We provide to the reader eager to use our method, an implementation of our algorithm, available online at \url{https://github.com/VivienCabannes/partial_labelling/}.
It is based on LP relaxation of the NP-hard minimum feedback arcset problem.
This relaxation was proven exact when $m \leq 6$ by \citet{Cabannes2020}.
The LP implementation relies on CPLEX \citep{Cplex}.
As complementary experiments, we will not provide much reproducibility details, those details would be really similar to the previous paragraphs, and the curious reader could run our code instead.
We present our ranking setup on Figure \ref{df:fig:rk_set} and our results on Figure \ref{df:fig:rk}.
\end{subappendices}

\chapter{Laplacian Regularization}

The following is a reproduction of \cite{Cabannes2021c}.

As annotations of data can be scarce in large-scale practical problems, leveraging unlabeled examples is one of the most important aspects of machine learning.
This is the aim of semi-supervised learning.
To benefit from the access to unlabeled data, it is natural to diffuse knowledge of labeled data to unlabeled one.
This induces the use of Laplacian regularization.
Yet, current implementations of Laplacian regularization suffer from several drawbacks, notably the well-known curse of dimensionality.
In this paper, we provide a statistical analysis to overcome those issues, and unveil a large body of spectral filtering methods that exhibit desirable behaviors.
They are implemented through (reproducing) kernel methods, for which we provide realistic computational guidelines in order to make our method usable with large amounts of data.

\section{Introduction}

In the last decade, machine learning has been able to tackle amazingly complex tasks, which was mainly allowed by computational power to train large learning models on large annotated datasets.
For instance, ImageNet is made of tens of millions of images, which have all been manually annotated by humans \citep{ImageNet}.
The greediness in data annotation of such a current learning paradigm is a major limitation.
In particular, when annotation of data demands in-depth expertise, relying on techniques that require zillions of labeled data is not viable.
This motivates several research streams to overcome the need for annotations, such as self-supervised learning for images or natural language processing \citep{Devlin2019}.
Aiming for generality, semi-supervised learning is the most classical one, assuming access to a vast amount of input data, but among which only a scarce percentage is labeled.
To leverage the presence of unlabeled data, most semi-supervised techniques assume a form of low-density separation hypothesis, as detailed in the recent review of \citet{Engelen2020}, and illustrated by state-of the-art models \citep{Berthelot2019,Verma2019}.
This hypothesis assumes that the function to learn from the data varies smoothly in highly populated regions of the input space, but might vary more strongly in sparsely populated areas, or that the decision frontiers between classes lie in regions with low-density.
In such a setting, it is natural to enforce constraints on the variations of the function to learn.
While semi-supervised learning is an important learning framework, it has not provided as many exciting realizations as one could have expected.
This might be related to the fact that it is classically approached through graph-based Laplacian, a technique that does not scale well with the dimension of the input space~\citep{Bengio2006}.

\paragraph{Paper organization.} In Section \ref{lap:sec:laplacian}, we motivate Laplacian regularization, and recall drawbacks of naive implementations.
These limitations are overcome in Section \ref{lap:sec:kernel} where we expose a theoretically principled path to derive well-behaved algorithms.
More precisely, we unveil a vast class of estimates based on spectral filtering.
We turn to implementation in Section~\ref{lap:sec:implementation} where we provide realistic guidelines to ensure scalability of the proposed algorithms.
Statistical properties of our estimators are stated in Section \ref{lap:sec:statistics}.

\paragraph{Contributions.} They are two folds.
({\em i}) Statistically, we explain that Laplacian regularization can be properly leveraged based on functional space considerations, and that those considerations can be turned into concrete implementations thanks to kernel methods.
As a result, we provide consistent estimators that exhibit fast convergence rates under a low density separation hypothesis, and that, in particular, do not suffer from the curse of dimensionality.
({\em ii}) Computationally, we avoid dealing with large matrices of derivatives by providing a low-rank approximation that allows dealing with $n^{\gamma}\log(n)\times n^{\gamma}\log(n)$ matrices, with a parameter $\gamma \in (0, 1]$ depending on the regularity of the problem, instead of $n(d+1)\times n(d+1)$ matrices, thus cutting down to ${\cal O}(\log(n)^2n^{1+2\gamma}d)$ the potential ${\cal O}(n^3d^3)$ training cost.

\paragraph{Related work.} Interplay between graph theory and machine learning were proven successful in the 2000s \citep{Smola2003}.
The seminal paper of \citet{Zhu2003} introduced graph-Laplacian as a transductive method in the context of semi-supervised learning.
A smoothing variant was proposed by \citep{Zhou2003}, which is coherent with the fact that enforcing constraints on labeled points leads to spikes \citep{Alaoui2016}.
Interestingly, graph Laplacian do converge to diffusion operators linked with the weighted Laplace Beltrami operator \citep{Hein2007,GarciaTrillos2019}.
However, these local diffusion methods are known to suffer from the curse of dimensionality \citep{Bengio2006}.
That is, local averaging methods are intuitive learning methods that have been used for more than half a century \citep{Fix1951}.
Yet, those methods do not scale well with the dimension of the input space \citep{Yang1999}.
This is related to the fact that to cover $[0,1]^d$, we need $\epsilon^{-d}$ balls of radius $\epsilon$.
Interestingly, if the function to learn is $m$ times differentiable with smooth partial derivatives, it is possible to leverage more information from function evaluations and overcome the curse of dimensionality when $m \gtrsim d$.
This property is related to covering numbers (a.k.a. capacity) of Sobolev spaces \citep{Kolmogorov1959}, and is leveraged by (reproducing) kernel methods \citep{Steinwart2008,Caponnetto2007}.
The crux of this paper is to apply this fact to Laplacian regularization techniques.
Note that derivatives with reproducing kernel methods in machine learning have already been considered in different settings by \citep{Zhou2008,Rosasco2013,Eriksson2018}.

\section{Laplacian regularization}
\label{lap:sec:laplacian}

In this section, we introduce the notations and concepts related to the semi-supervised learning regression problem, noting that most of our results extend to any convex loss beyond least-squares.
We motivate and describe Laplacian regularization that will allow us to leverage the low-density separation hypothesis.
We explain statistical drawbacks usually linked with Laplacian regularization, and discuss how to circumvent them.

In the following, we denote by $\X = \R^d$ the input space, $\Y=\R$ the output space, and by $\rho\in\prob{\X\times\Y}$ the joint distribution on $\X\times\Y$.
For simplicity, we assume that $\rho$ has compact support.
In the following, we denote by $\rho_\X$ the marginal of $\rho$ over $\X$, and by $\rho\vert_x$ the conditional distribution of $Y$ given $X=x$.
As usual, for $p \in \N^*$, $L^p(\R^d)$ is the space of functions $f$ such that $f^p$ is integrable.
Moreover, we define usual Sobolev spaces: for $s \in \N$, $W^{s,p}(\R^d)$ stands for the space of functions whose weak derivatives of order $s$-th are in $L^p(\R^d)$.
When $p=2$, they have a Hilbertian structure, and we denote, $H^{s}(\R^d) = W^{s,2}(\R^d)$ these Hilbertian spaces.
Ideally, we would like to retrieve the mapping $g^*:\X\to\Y$ defined as
\begin{equation}
  \label{lap:eq:least_square}
  g^* = \argmin_{g\in L^2(\rho_\X)}\E_{(X, Y)\sim \rho}\bracket{\norm{g(X) - Y}^2}
  = \argmin_{g\in L^2(\rho_\X)}\norm{g - g_\rho}^2_{L^2(\rho_\X)} = g_\rho,
\end{equation}
where $g_\rho:\X\to\Y$ is defined as $g_\rho(x) = \E\bracket{Y\midvert X=x}$.
In semi-supervised learning, we assume that we do not have access to $\rho$, but we have access to $n$ independent samples $(X_i)_{i\leq n} \sim\rho_\X^{\otimes n}$, among which we have $n_\ell$ labels $Y_i \sim \rho\vert_{X_i}$ for $i \leq n_\ell$, with $n_\ell$ potentially much smaller than $n$.
In other terms, we have $n_\ell$ supervised pairs $(X_i, Y_i)_{i\leq n_\ell}$, and $n-n_\ell$ unsupervised samples $(X_i)_{n_\ell < i \leq n}$.
While we restrict ourselves to real-valued regression for simplicity, our exposition indeed applies generically to partially supervised learning.
In particular, it can be used off-the-shelve to complement the approaches of \citep{Cabannes2020,Cabannes2021} as we detailed in Appendix \ref{lap:sec:extension}.

\subsection{Diffusion operator \texorpdfstring{${\cal L}$}{}}
\begin{figure*}[t]
  \centering
  \includegraphics{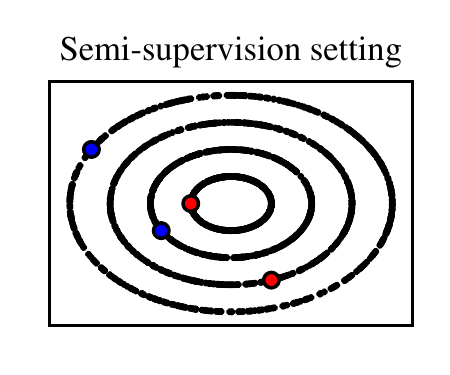}
  \includegraphics{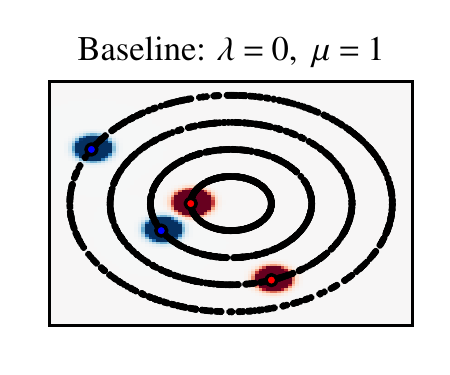}
  \includegraphics{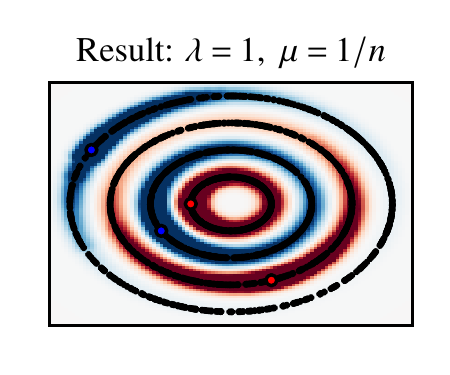}
  \vskip -0.2in
  \caption{
    Motivating example.
    (Left) We suppose given $n = 2000$ points in $\X=\R^2$, represented as black dots, spanning $4$ concentric circles.
    Among those points are $n_\ell = 4$ labeled points, with labels being either $1$ represented in red, and $-1$ represented in blue.
    In this setting, it is natural to assume that $g^*$ should be constant on each circle, which can be encoded as $\norm{\nabla g^*} = 0$ on $\supp\rho_\X$.
    (Middle) Kernel ridge regression estimate based on the labeled points with Gaussian kernel of bandwidth $\sigma = .2r$, $r$ being the radius of the innermost circle.
    (Right) Laplacian regularization reconstruction.
    The reconstruction is based on approximate empirical risk minimization with $p=n$, which ensures a computational complexity of $O(p^2 nd)$, instead of $O(n^3d^3)$ needed to recover the exact empirical risk minimizer \eqref{lap:eq:estimate}.
  }
  \label{lap:fig:intro}
  \vskip -0.2in
\end{figure*}

In order to leverage unlabeled data, we will assume that $g^*$ varies smoothly on highly populated regions of $\X$, and might vary highly on low density regions.
For example, this is the case when data are clustered in well separated regions of space, and labels are constant on clusters.
This is captured by the fact that the Dirichlet energy
\begin{equation}
  \label{lap:eq:diffusion_operator}
  \int_{\X} \norm{\nabla g^*(x)}^2 \rho_\X(\diff x)
  = \E_{X\sim\rho_\X} \bracket{\norm{\nabla g^*(X)}^2}
  =: \norm{{\cal L}^{1/2} g}_{L^2(\rho_\X)}^2,
\end{equation}
is assumed to be small.
Because the quadratic functional \eqref{lap:eq:diffusion_operator} will play a crucial role in our exposition, we define ${\cal L}$ as the self-adjoint operator on $L^2(\rho_\X)$, extending the operator on $H^1(\rho_\X)$ representing this functional.
Under mild assumptions on $\rho_\X$, ${\cal L}^{-1}$ can be shown to be a compact operator, which we will assume in the following.
In essence, we will assume that if we have a lot of unlabeled data and $\norm{{\cal L}^{1/2} g}$ can be well approximated for any function $g$, then we do not need a lot of labeled data to estimate correctly $g^*$.
To illustrate this, at one extreme, if we know that $\norm{{\cal L}^{1/2} g^*} = 0$, then $g^*$ is known to be constant on each connected component of $\rho_\X$ so that, along with the knowledge of $\rho_\X$, only a few labeled points would be sufficient to recover perfectly $g^*$.
We illustrate those considerations on Figure \ref{lap:fig:intro}.

\subsection{Drawbacks of naive Laplacian regularization}
Following the motivations presented previously, it is natural to consider the regularized objective and solution defined, for $\lambda > 0$, as
\begin{equation}
  \label{lap:eq:laplacian_tikhonov}
  \begin{split}
    g_\lambda
    &= \argmin_{g\in H^1(\rho_\X)} \E_{(X,Y)\sim\rho}\bracket{\norm{g(X) - Y}^2} + \lambda
    \E_{X\sim\rho_\X}\bracket{\norm{\nabla g(X)}^2_{\R^d}}
    \\&= \argmin_{g\in H^1(\rho_\X)} \norm{g - g_\rho}^2_{L^2(\rho_\X)} + \lambda
    \norm{{\cal L}^{1/2} g}^2_{L^2(\rho_\X)} = (I+\lambda{\cal L})^{-1} g_\rho.
  \end{split}
\end{equation}
This regularization has nice properties.
In particular, for small $\lambda$, it can be seen as a first order approximation of the heat equation solution $e^{-\lambda {\cal L}}g_\rho$, which represents the temperature profile at time $t=\lambda$, instantiated with the initial profile $g_\rho$, and with $\rho_\X$ modeling the thermal conductivity.
It also has interpretations in terms of random walk and Langevin diffusion \citep{PillaudVivien2020,Klus2020}.
In a word, $g_\lambda$ is the diffusion of $g_\rho$ with respect to the density $\rho_\X$, which relates to the idea of diffusing labeled data with respect to the intrinsic geometry of the data, which is the idea captured by \citep{Zhu2003}.

However, from a learning perspective, \eqref{lap:eq:laplacian_tikhonov}~is linked with the prior that $g^*$ belongs to $H^1(\rho_\X)$, a prior that is not strong enough to overcome the curse of dimensionality as we saw in the related work section.
Moreover, assuming we have enough unsupervised data to suppose known $\rho_\X$, and therefore $\cal L$, \eqref{lap:eq:laplacian_tikhonov}~leads to the naive empirical estimate
\(
g_{\text{(naive)}} \in \argmin_{g:\X\to\R} \sum_{i=1}^{n_\ell} \norm{g(X_i) - Y_i}^2 + n_\ell\lambda \norm{{\cal L}^{1/2}g}^2.
\)
While the definition of $g_{\text{(naive)}}$ could seem like a great idea, in fact, such an estimate $g_{\text{(naive)}}$ is known to be mostly constant and spiking to interpolate the data $(X_i, Y_i)$ as soon as $d > 2$ \citep{Nadler2009}.
This is to be related with the capacity of the space associated with the pseudo-norm $\norm{{\cal L}^{1/2}g}$ in $L^2$.
This capacity, related to $H^1$, is too large for the Laplacian regularization term to constraint $g_{\text{(naive)}}$ in a meaningful way.
In other terms, we need to regularize with stronger penalties.

\subsection{Stronger regularization}
\label{lap:sec:kernel_free_reg}

In this subsection, we discuss techniques to overcome the issues encountered with $g_{\text{(naive)}}$.
Those techniques are based on functional space constraints or on spectral filtering techniques.

\paragraph{Functional spaces.} A solution to overcome the capacity issue of $H^1$ in $L^2$ is to constrain the estimate of $g^*$ to belong to a smaller functional space.
In the realm of graph Laplacian, \citep{Alaoui2016} proposed to solve this problem by considering the $r$-Laplacian regularization reading
\(
\Omega_{r} = \int_\X \norm{\nabla g(X)}^r \rho(\diff x),
\)
with $r > d$.
In essence, this restricts $g$ to live in $W^{1,r}(\rho_\X)$ for $r > d$, and allows avoiding spikes associated with $g_{\text{(naive)}}$.
However, considering high power of the gradient is likely to introduce instability (think that $d$ is the potentially colossal dimension of the input space), and from a learning perspective, the capacity of $W^{1,r}$, which compares to the one of $H^2$, is still too big.
In this paper, we will rather keep the diffusion operator ${\cal L}$, and add a second penalty to reduce the space in which we look for the solution.
With ${\cal G}$ a Hilbert space of functions, we could look for, with $\mu > 0$ a second regularization parameter
\begin{equation}
  \label{lap:eq:tikhonov}
  g_{\lambda, \mu} = \argmin_{g:{\cal G}\cap H^1(\rho_\X)} \norm{g - g_\rho}^2_{L^2(\rho_\X)}
  + \lambda \norm{{\cal L}^{1/2} g}^2_{L^2(\rho_\X)} + \lambda\mu\norm{g}_{\cal G}^2.
\end{equation}
This formulation restricts $g_{\lambda, \mu}$ to belong both to ${H}^1(\rho_\X)$ (thanks to the term in $\lambda$) and ${\cal G}$ (thanks to the term in $\mu$).
In particular the resulting space $H^1(\rho_\X) \cap {\cal G}$ to which $g_{\lambda, \mu}$ belongs, has a smaller capacity in $L^2$ than the one of ${\cal G}$ in $L^2$.
In practice, we do not have access to $\rho$ and $\rho_\X$ but to $(X_i, Y_i)_{i\leq n_\ell}$ and $(X_i)_{i\leq n}$, and we might consider the empirical estimator defined through empirical risk minimization
\begin{equation}
  \label{lap:eq:estimate}
  g_{n_\ell,n} = \argmin_{g\in{\cal G}} n_\ell^{-1}\sum_{i=1}^{n_\ell} \norm{g(X_i) - Y_i}^2
  + \lambda n^{-1} \sum_{i=1}^n\norm{\nabla g(X_i)}^2 + \lambda\mu\norm{g}^2_{\cal G}.
\end{equation}
For example, we could consider ${\cal G}$ to be the Sobolev space ${H}^m(\diff x)$.
Note the difference between ${\cal G}$ linked with $\diff x$, the Lebesgue measure, that is known, and ${\cal L}$ linked with $\rho_\X$, the marginal of $\rho$ over $\X$, that is not known.
In this setting, the regularization $\|{\cal L}^{1/2}g\|^2 + \mu\|g\|_{\cal G}^2$ reads $\int_\X \norm{D g(x)}^2 \rho_\X(\diff x) + \mu \int_\X \sum_{\alpha=0}^m \norm{D^\alpha g(x)}^2 \diff x$.
Because of the size of ${H}^m$ in $L^2$, this allows for efficient approximation of $g_{\lambda, \mu}$ based on empirical risk minimization.
In particular, if $n = +\infty$, we expect the minimizer \eqref{lap:eq:estimate} to converge toward $g_{\lambda, \mu}$ at rates in $L^2$ scaling similarly to $n_\ell^{-m/d}$ in $n_\ell$.
To complete the picture, depending on a prior on $g_\rho$, $g_{\lambda, \mu}$ might exhibit good convergence properties toward $g_\rho$ as $\lambda$ and $\mu$ go to zero.
This contrasts with the problem encountered with $g_{\text{(naive)}}$.
Those considerations are exactly what reproducing kernel Hilbert space will provide, additionally with a computationally friendly framework to perform the estimation.
Note that quantities similar to $g_{\lambda,\mu}$ were considered in \citep{Zhou2008,Rosasco2013}.

\paragraph*{Spectral filtering.}
Without looking for higher power-norm, \citep{Nadler2009} proposed to overcome the capacity issue by considering approximation of the operator ${\cal L}$ based on the graph-based technique provided by \citep{Belkin2003,Coifman2006} and to reduce the search of $g_{n_\ell}$ on the space spanned by the first few eigenvectors of the Laplacian.
In particular, on Figure \ref{lap:fig:intro}, $g^*$ could be searched in the null space of ${\cal L}$, that is, among functions that are constant on each connected component of $\supp\rho_\X$.
This technique exhibits two parts, the ``unsupervised'' estimation of ${\cal L}$ that will depend on the total number of data $n$, and the ``supervised'' search for $g_\rho$ on the first few eigenvectors of ${\cal L}$ that will depend on the number of labels~$n_\ell$.
While, at first sight, this technique seems to be completely different from Tikhonov regularization~\eqref{lap:eq:tikhonov}, it can be cast, along with gradient descent, into the same \emph{spectral filtering} framework \citep{Lin2020}.
This point of view enables the use of a wide range of techniques offered by spectral manipulations on the diffusion operator ${\cal L}$.

This paper is motivated by the fact that current well-grounded semi-supervised learning techniques are implemented based on graph-based Laplacian, which is a local averaging method that does not leverage smartly functional capacity.
In particular, as recalled earlier, graph-based Laplacian is known to suffer from the curse of dimensionality, in the sense that the convergence of the empirical estimator $\widehat{\cal L}$ toward the ${\cal L}$ exhibits a rate of convergence of order ${\cal O}(n^{-1/d})$ with $d$ the dimension of the input space ${\cal X}$ \citep{Hein2007}.
In this work, we will bypass this curse of dimensionality by looking for $g$ in a smooth universal reproducing kernel Hilbert space, which will lead to efficient empirical estimates.

\section{Spectral filtering with kernel Laplacian}
\label{lap:sec:kernel}

In this section, we approach Laplacian regularization from a functional analysis perspective.
We first introduce kernel methods and derivatives in reproducing kernel Hilbert space (RKHS).
We then translate the considerations provided in Section \ref{lap:sec:kernel_free_reg} in the realm of kernel methods.

\subsection{Kernel methods and derivatives evaluation maps}

In this subsection, we introduce kernel methods (see \citep{Aronszajn1950,Scholkopf2001,Steinwart2008} for more details).
Consider $({\cal H}, \scap{\cdot}{\cdot}_{\cal H})$ a reproducing kernel Hilbert space, that is a Hilbert space of functions from $\X$ to $\R$ such that the evaluation functionals $L_x:{\cal H}\to\R;g\to g(x)$ are continuous linear forms for any $x\in\X$.
Such forms can be represented by $k_x \in {\cal H}$ such that, for any $g\in{\cal H}$, $L_x(g) = \scap{k_x}{g}_{\cal H}$.
A reproducing kernel Hilbert space can alternatively be defined from a symmetric positive semi-definite kernel $k:\X\to\X\to\R$, that is a function such that for any $n\in\N$ and $(x_i)_{i\leq n} \in \X^n$ the matrix $(k(x_i,x_j))_{i,j}$ is symmetric positive semi-definite, by building $(k_x)_{x\in\X}$ such that $k(x, x') = \scap{k_x}{k_{x'}}_{\cal H}$.
From a learning perspective, it is useful to use the evaluation maps to rewrite ${\cal H} = \brace{g_\theta:x\to\scap{k_x}{\theta}_{\cal H}\midvert \theta\in{\cal H}}$.
As such, kernel methods can be seen as ``linear models'' with features $k_x$, allowing to parametrize large spaces of functions \citep{Micchelli2006}.
In the following, we will differentiate $\theta$ seen as an element of ${\cal H}$ and $g_\theta$ seen as its embedding in $L^2$.
To make this distinction formal, we define the embedding $S:({\cal H}, \scap{\cdot}{\cdot}_{\cal H})\hookrightarrow (L^2(\rho_\X), \scap{\cdot}{\cdot}_{L2}); \theta\to g_\theta$, as well as its adjoint $S^\star:L^2(\rho_\X)\to{\cal H}$.

Given a linear parametric model of functions $g_\theta(x) = \scap{\theta}{k_x}_{\cal H}$, it is possible to compute derivatives of $g_\theta$ based on derivatives of the feature vector -- think of ${\cal H} = \R^p$ and of $k_x = \phi(x)$ as a feature vector with $\phi:\R^d \to \R^p$.
For $\alpha \in \N^d$, with $\abs\alpha = \sum_{i\leq d} \alpha_i$, we have the following equality of partial derivatives, when $k$ is $2\abs{\alpha}$ times differentiable,
\[
  D^\alpha g_\theta(x) = \scap{\theta}{D^\alpha k_x},\qquad\text{where}\qquad
  D^\alpha = \frac{\partial^{\abs\alpha}}{(\partial x_1)^{\alpha_1} (\partial x_2)^{\alpha_2}\cdots (\partial x_d)^{\alpha_d}}.
\]
Here and $D^\alpha k_x$ has to be understood as the partial derivative of the mapping of $x\in\X$ to $k_x\in{\cal H}$, which can be shown to belong to ${\cal H}$ \citep{Zhou2008}.
In the following, we assume that $k$ is twice differentiable with continuous derivatives, and will make an extensive use of derivatives of the form $\partial_i k_x = \partial k_x / \partial x_i$ for $i \leq d$ and $x\in\X$.
Note that, as well as we can describe completely the Hilbertian geometry of the space $\Span\brace{k_x \midvert x\in\X}$ through $k(x, x') = \scap{k_x}{k_x'}$, for $x, x'\in\X$, we can describe the Hilbertian geometry of $\Span\brace{k_x \midvert x\in\X} + \Span\brace{\partial_i k_x\midvert x\in\X}$, through
\[
  \partial_{1,i} k(x, x') = \scap{\partial_i k_x}{k_{x'}}_{\cal H},
  \qquad\text{and}\qquad
  \partial_{1,i}\partial_{2,j} k(x, x') = \scap{\partial_i k_x}{\partial_j
    k_{x'}}_{\cal H},
\]
where $\partial_{1,i}$ denotes the partial derivative with respect to the $i$-th coordinates of the first variable.
This echoes the so-called ``representer theorems''.

\begin{example}[Gaussian kernel]
  \label{lap:ex:rbf}
  A classical kernel is the Gaussian kernel, also known as radial basis function, defined for $\sigma > 0$ as the following $k$, and satisfying, for $i\neq j$, the following equalities,
  \begin{gather*}
    k(x, x') = \exp\paren{-\frac{\norm{x-x'}^2}{2\sigma^2}},\qquad
    \partial_{1,i}\partial_{2,j} k(x, y) = - \frac{(x_i - y_i)(x_j -
      y_j)}{\sigma^4} k(x, y),
    \\
    \partial_{1,i} k(x, y) = -\frac{(x_i - y_i)}{\sigma^2} k(x, y),
    \qquad
    \partial_{1,i}\partial_{2,i} k(x, y) = \paren{\frac{1}{\sigma^2} -
      \frac{(x_i - y_i)^2}{\sigma^4}} k(x, y),
  \end{gather*}
  where $x_i$ designs the $i$-th coordinates of the vector $x\in\X=\R^d$.
\end{example}

\subsection{Tikhonov, spectral filtering and dimensionality reduction}
Given the kernel $k$, its associated RKHS ${\cal H}$ and $S$ the embedding of ${\cal H}$ in $L^2$, we rewrite~\eqref{lap:eq:tikhonov} under its ``parametrized'' version
\begin{equation}
  \tag{\ref{lap:eq:tikhonov}}
  g_{\lambda, \mu} = S \, \argmin_{\theta\in{\cal H}} \brace{\norm{S\theta - g_\rho}^2_{L^2(\rho_\X)}
  + \lambda \norm{{\cal L}^{1/2} S\theta}^2_{L^2(\rho_\X)} + \lambda\mu\norm{\theta}_{\cal H}^2 }.
\end{equation}
Do not hesitate to refer to Table \ref{lap:tab:notations} to keep track of notations.
In the following, we will use that
\(
\norm{{\cal L}^{1/2} S\theta}^2_{L^2(\rho_\X)} + \mu\norm{\theta}_{\cal
  H}^2
= \norm{(S^\star {\cal L} S + \mu I)^{1/2} \theta}_{\cal H}^2.
\)
This equality explains why we consider $\mu\lambda$ instead of $\mu$ in the last term.
In the RKHS setting, the study of~\eqref{lap:eq:tikhonov} unveils the three operators $\Sigma$, $L$, and $I$ on ${\cal H}$, (indeed
\(
g_{\lambda, \mu} = S \, \argmin_{\theta\in{\cal H}} \brace{
  \theta^\star(\Sigma + \lambda L + \lambda\mu)\theta - 2\theta^\star S^\star g_\rho }
\))
where $I$ is the identity, and, as we detail in Appendix \ref{lap:app:operators},
\begin{equation}
  \label{lap:eq:block_op}
  \Sigma = S^\star S = \E_{X\sim\rho_\X}\bracket{k_X \otimes k_X},
  \qquad\text{and}\qquad
  L = S^\star {\cal L}S = \E_{X\sim\rho_\X}
  \bracket{\sum_{i=1}^d \partial_j k_X \otimes \partial_j k_X}.
\end{equation}
Regularization and spectral filtering have been well-studied in the inverse-problem literature.
In particular, the regularization~\eqref{lap:eq:tikhonov} is known to be linked with the generalized singular value decomposition of $[\Sigma; L+\mu I]$ (see, \emph{e.g.}, \cite{Edelman2020}), which is linked to the generalized eigenvalue decomposition of $(\Sigma, L+\mu I)$ \citep{Golub2013}.
We derive the following characterization of~\eqref{lap:eq:tikhonov}, whose proof is reported in Appendix \ref{lap:app:algebra}.

\begin{proposition}
  \label{lap:thm:decomposition}
  Let $(\lambda_{i,\mu})_{i\in\N} \in \R^\N, (\theta_{i,\mu})_{i\in\N} \in {\cal H}^\N$ be the generalized eigenvalue decomposition of the pair $(\Sigma, L + \mu I)$, that is $(\theta_{i,\mu})$ generating ${\cal H}$ and such that for any $i, j \in \N$, $\Sigma \theta_{i,\mu} = \lambda_{i,\mu} (L + \mu I) \theta_{i,\mu}$, and $\scap{\theta_{i,\mu}}{(L+\mu I) \theta_{j,\mu}} = \ind{i=j}$.
  \eqref{lap:eq:tikhonov} can be rewritten as
  \begin{equation}
    \label{lap:eq:filtering}
    g_{\lambda, \mu} =
    \paren{\sum_{i\in\N} \psi(\lambda_{i,\mu}) S\theta_{i,\mu} \otimes S\theta_{i, \mu}} g_\rho =
    \sum_{i\in\N} \psi(\lambda_{i,\mu}) \scap{S^\star g_\rho}{\theta_{i,\mu}} S\theta_{i,\mu},
  \end{equation}
  with $\psi:\R_+ \to \R; x\to (x+\lambda)^{-1}$.
  \eqref{lap:eq:filtering}~should be seen as a specific instance of spectral filtering based on a filter function $\psi:\R_+\to\R$.
\end{proposition}

\begin{figure*}[t]
  \centering
  \includegraphics{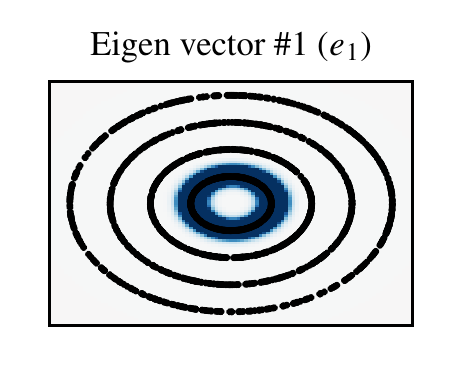}
  \includegraphics{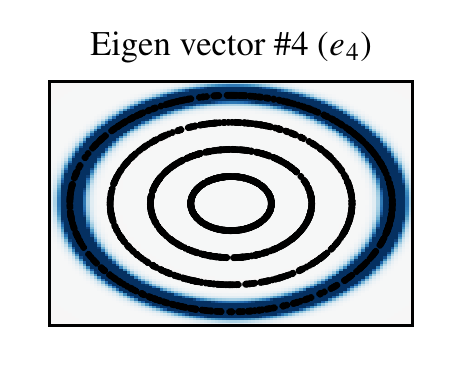}
  \includegraphics{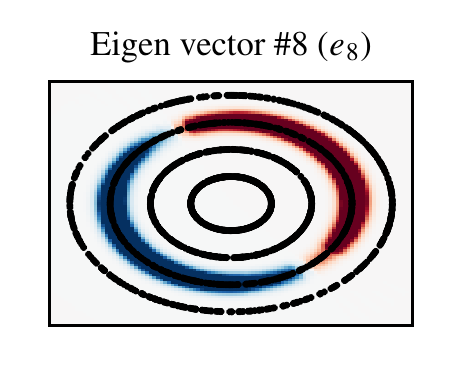}
  \vskip -0.2in
  \caption{
    Few of the first generalized eigenvectors of $(\hat\Sigma; \hat{L} + \mu I)$ (with $\mu = 1/n$).
    The first four eigenvectors correspond to constant functions on each circle, as shown with $e_1$ and $e_4$.
    The few eigenvectors after correspond to second harmonics localized on a single circle as shown with $e_8$.
  }
  \label{lap:fig:unsupervised}
  \vskip -0.2in
\end{figure*}

Interestingly, the generalized eigenvalue decomposition of the pair $(\Sigma, L+\mu I)$ was already considered by \citet{PillaudVivien2020} to estimate the first eigenvalue of the Laplacian.
Moreover, \citet{PillaudVivien2020b} suggests leveraging this decomposition for dimensionality reduction based on the first eigenvectors of the Laplacian.
As well as \eqref{lap:eq:tikhonov}~contrasts with graph-based semi-supervised learning techniques, this dimensionality reduction technique contrasts with methods based on graph Laplacian provided by \citep{Belkin2003,Coifman2006}.
Remarkably, the semi-supervised learning algorithm that consists in using the unsupervised data to perform dimensionality reduction based on the Laplacian eigenvalue decomposition, before solving a small linear regression problem on the small resulting space, can be seen as a specific instance of spectral filtering, based on regularization by thresholding/cutting-off eigenvalue, which corresponds to $\psi:x\to x^{-1}\ind{x>\lambda}$ for a given threshold $\lambda > 0$ in~\eqref{lap:eq:filtering}.

\section{Implementation}
\label{lap:sec:implementation}

In this section, we discuss how to practically implement estimates for~\eqref{lap:eq:filtering} based on empirical data $(X_i, Y_i)_{i\leq n_\ell}$ and $(X_i)_{n_\ell < i \leq n}$.
We first review how we can approximate the integral operators of~\eqref{lap:eq:block_op} based on data.
We then discuss how to implement our methods practically on a computer.
We end this section by considering approximations that allow cutting down high computational costs associated with kernel methods involving derivatives.

\begin{algorithm}[H]
  \caption{Empirical estimates based on spectral filtering.}
  \KwData{$(X_i, Y_i)_{i\leq n_\ell}$, $(X_i)_{n_\ell < i\leq n}$, a kernel $k$, a filter $\psi$, a regularizer $\mu$}
  \KwResult{$\hat{g}_p$ through $c \in \R^p$ defining $\hat{g}_p(x) = \sum_{i=1}^n c_i k(x, X_i) = k_x^\star T_ac$}

  Compute $S_nT_a = (k(X_i, X_j))_{i\leq n, j\leq p} \in \R^{n\times p}$ in ${\cal O}(pn)$\\
  Compute $Z_nT_a = (\partial_{1,j} k(X_l, X_i))_{(j\leq d, l\leq n), i\leq p} \in \R^{nd\times p}$ in ${\cal O}(pnd)$\\
  Build $T_a^\star \hat\Sigma T_a = n^{-1}(S_nT_a)^\top(S_nT_a)$ in ${\cal O}(p^2n)$\\
  Build $T_a^\star \hat{L} T_a = n^{-1}(Z_nT_a)^\top(Z_nT_a)$ in ${\cal
        O}(p^2nd)$\footnote{
    Building this matrix can be avoided by using the generalized singular value decomposition rather than the generalized eigenvector decomposition.
    Implemented with Lapack, such a procedure will also require $O(p^2 nd)$ floating point operations, but with a smaller constant in the big $O$ \citep{Golub2013}.}\\
  Build $T_a^\star T_a = (k(X_i, X_j))_{i,j\leq p} \in \R^{p\times p}$ in ${\cal O}(1)$ as a partial copy of $S_nT_a$\\
  Get $(\lambda_{i, \mu}, u_{i,\mu})_{i\leq n}$ the generalized eigenelements of $(T_a^\star \hat\Sigma T_a, T_a^\star (\hat{L} + \mu I) T_a)$ in~${\cal O}(p^3)$\\
  Get $b = T_a^\star \hat\theta = (n_\ell^{-1}\sum_{i=1}^{n_\ell} Y_i k(X_i, X_j))_{j\leq p} \in \R^p$ in ${\cal O}(pn_\ell)$\\
  Return $c = \sum_{i=1}^n \psi(\lambda_i) u_i u_i^\top b \in \R^p$ in ${\cal O}(p^3)$.
  \label{lap:alg:imp}
\end{algorithm}

\subsection{Integral operators' approximation}
The classical empirical risk minimization in~\eqref{lap:eq:estimate} can be understood as the plugging of the approximate distributions $\hat\rho = n_\ell^{-1}\sum_{i=1}^{n_\ell} \delta_{X_i} \otimes \delta_{Y_i}$ and $\hat\rho_\X = n^{-1} \sum_{i=1}^n \delta_{X_i}$ instead of $\rho$ and $\rho_\X$ in~\eqref{lap:eq:tikhonov}.
It can also be understood as the same replacement when dealing with integral operators, leading to the three following important quantities to rewrite~\eqref{lap:eq:filtering},
\begin{equation}
  \label{lap:eq:approximate_operator}
  \hat\Sigma := n^{-1}\sum_{i=1}^{n} k_{X_i}\otimes k_{X_i},\
  \hat{L} := n^{-1}\sum_{i=1}^n \sum_{j=1}^d \partial_j k_{X_i}\otimes \partial_j k_{X_i},\
  \hat\theta := \widehat{S^\star g_\rho} := n_\ell^{-1}\sum_{i=1}^{n_\ell} Y_i k_{X_i}.
\end{equation}
It should be noted that while considering $n$ in the definition of $\hat\Sigma$ is natural from the spectral filtering perspective, to make it formally equivalent with the empirical risk minimization \eqref{lap:eq:estimate}, it should be replaced by $n_\ell$.
\eqref{lap:eq:approximate_operator}~allows rewriting~\eqref{lap:eq:filtering} without relying on the knowledge of $\rho$, by considering $(\hat\lambda_{i,\mu}, \hat\theta_{i,\mu})$ the generalized eigenvalue decomposition of $(\hat\Sigma, \hat L)$ and considering
\begin{equation}
  \hat g = \sum_{i\in\N} \psi(\hat\lambda_{i,\mu}) \scap{\widehat{S^\star g_\rho}}{\hat\theta_{i,\mu}} S\hat\theta_{i,\mu},
\end{equation}
We present the first eigenvectors (after plunging them in $L^2$ through $S$) of the generalized eigenvalue decomposition of $(\hat\Sigma, \hat{L} + \mu I)$ on Figure \ref{lap:fig:unsupervised}.
The first eigenvectors recover the null space of ${\cal L}$.
This explains clearly the behavior on the right of Figure \ref{lap:fig:intro}.

\subsection{Matrix representation and approximation of operators}
Currently, we are dealing with operators ($\hat\Sigma, \hat{L}$) and vectors (\emph{e.g.}, $\hat\theta$) in the Hilbert space ${\cal H}$.
It is natural to wonder how to represent this on a computer.
The answer is the object of representer theorems (see Theorem 1 of \citep{Zhou2008}), and consists in noticing that all the objects introduced are actually defined in, or operate on, ${\cal H}_n + {\cal H}_{n,\partial} \subset {\cal H}$, with ${\cal H}_n = \Span\brace{k_{X_i} \midvert i\leq n}$ and ${\cal H}_{n,\partial} = \Span\brace{\partial_j k_{X_i} \midvert i\leq n, j\leq d}$.
This subspace of ${\cal H}$ is of dimension at most $n(d+1)$ and if $T:\R^p \to {\cal H}_n + {\cal H}_{n,\partial}$ (with $p \leq n(d+1)$) parametrizes ${\cal H}_n + {\cal H}_{n,\partial}$, our problem can be cast in $\R^{p}$ by considering the $p\times p$ matrices $T^\star \hat\Sigma T$ and $T^\star (\hat{L} + \mu I)T$ instead of the operators $\hat\Sigma$ and $\hat{L} + \mu I$.
The canonical representation consists in taking $p = n(d+1)$ and considering for $c \in \R^{n(d+1)}$, the mapping $T_c c = \sum_{i=1}^n c_{i0} k_{X_i} + \sum_{j=1}^d c_{ij}\partial_j k_{X_i}$ \citep{Zhou2008,Rosasco2013}.

This exact implementation implies dealing and finding the generalized eigenvalue decomposition of $p\times p$ matrices with $p = n(d+1)$, which leads to computational costs in ${\cal O}(n^3d^3)$, which can be prohibitive.
Two solutions are known to cut down prohibitive computational costs of kernel methods.
Both methods consist in looking for a space that can be parametrized by $\R^p$ for a small $p$ and that approximates well the space ${\cal H}_n + {\cal H}_{n,\partial} \subset {\cal H}$.
The first solution is provided by random features \citep{Rahimi2007}.
It consists in approximating ${\cal H}$ with a space of small dimension $p\in\N$, linked with an explicit representation $\phi:\X\to\R^p$ that approximate $k(x, x') \simeq k_\phi(x, x') = \scap{\phi(x)}{\phi(x')}_{\R^p}$.
In theory, it substitutes the kernel $k$ by $k_\phi$.
In practice, all computations can be done with the explicit feature $\phi$.

\paragraph{Approximate solution.}
The second solution, which we are going to use in this work, consists in approximating ${\cal H}_n + {\cal H}_{n,\partial}$ by ${\cal H}_p = \Span\brace{k_{X_i}}_{i\leq p}$ for $p \leq n$.
This method echoes the celebrated Nystr\"om method \citep{Williams2000}, as well as the Rayleigh–Ritz method for Sturm–Liouville problems.
In essence, \citep{Rudi2015} shows that, when considering subsampling based on leverage score, $p=n^{\gamma}\log(n)$, with $\gamma \in (0,1]$ linked to the ``size'' of the RKHS and the regularity of the solution, is a good enough approximation, in the sense that it only downgrades the sample complexity by a constant factor.
In theory, we know that the space ${\cal H}_p$ will converge to ${\cal H} = \text{Closure}\Span\brace{k_x}_{x\in\supp\rho_\X}$ as $p$ goes to infinity.
In practice, it means considering the approximation mapping $T_a: \R^p \to {\cal H}; c \to \sum_{i=1}^p c_i k_{X_i}$, and dealing with the $p\times p$ matrices $T_a^\star \Sigma T_a$ and $T_a^\star L T_a$.
It should be noted that the computation of $T_a^\star L T_a$ requires to multiply a $p\times nd$ matrix by its transpose.
Overall, training this method can be done with ${\cal O}(p^2 n d)$ basic operations, and inference with this method can be done in ${\cal O}(p)$.
The saving cost of this approximate method is huge: without compromising the precision of our estimator, we went from $O(n^3 d^3)$ run time complexities to $O(\log(n)^2n^{1+2\gamma} d)$ computations, with $\gamma$ possibly very small.
Similarly, the memory cost went from $O(n^2 d^2)$ down to $O(nd + n^{2\gamma})$.\footnote{%
  Our code is available online at \url{https://github.com/VivienCabannes/partial_labelling}.}

\section{Statistical analysis}
\label{lap:sec:statistics}

In this section, we are interested in quantifying the risk of the learned mapping $\hat{g}$.
We study it through the generalization bound, which consists in obtaining a bound on the averaged excess risk~$\E_{\textrm{data}}\|\hat{g} - g_\rho\|_{L^2}^2$.
In particular, we want to answer the following points.
\begin{enumerate}
  \item How, and under which assumptions, Laplacian regularization boosts learning?
  \item How the excess of risk relates to the number of labeled and unlabeled data?
\end{enumerate}
In terms of priors, we want to leverage a low-density separation hypothesis.
In particular, we can suppose that when diffusing $g_\rho$ with $e^{-t{\cal L}}$ we stay close to $g_\rho$, or that $g_\rho$ is supported on a finite dimensional space of functions on which $\norm{{\cal L}^{1/2}g}$ (which measures the variation of $g$) is small.
Both those assumptions can be made formal by assuming the $g_\rho$ is supported by the first eigenvectors of the diffusion operator ${\cal L}$.
\begin{assumption}[Source condition]
  \label{lap:ass:source}
  $g_\rho$ is supported on a finite dimensional space that is left stable by the diffusion operator ${\cal L}$.
  In other terms, if $(e_i) \in (L^2)^\N$ are the eigenvectors of ${\cal L}$, there exists $r\in\N$, such that $g_\rho \in \Span\brace{e_i}_{i\leq r}$.
\end{assumption}
We will also assume that the diffusion operator ${\cal L}$ can be well approximated by the RKHS associated with $k$.
In practice, under mild assumptions, {\em c.f.} Appendix \ref{lap:app:operators}, the eigenvectors of the Laplacian are known to be regular, in particular to belong to $H^m$ for $m\in\N$ bigger than $d$.
As such, many classical kernels would verify the following assumption.
\begin{assumption}[Approximation condition]
  \label{lap:ass:approximation}
  The eigenvectors $(e_i)$ of ${\cal L}$ belong to the RKHS ${\cal H}$.
\end{assumption}
We add one technical assumptions regarding the eigenvalue decay of the operator $\Sigma$ compared to the operator $L$, with $\preceq$ denoting the L\"owner order ({\em i.e.}, for $A$ and $B$ symmetric, $A \preceq B$ if $B-A$ is positive semi-definite).
\begin{assumption}[Eigenvalue decay]
  \label{lap:ass:decay}
  There exists $a \in [0, 1]$ and $c > 0$ such that $L \preceq c \Sigma^\alpha$.
\end{assumption}
Note that, in our setting, $L$ is compact and bounded and Assumption \ref{lap:ass:decay} is always satisfied with $a = 0$.
For translation-invariant kernels, such as Gaussian or Laplace kernels, based on considerations linking eigenvalue decay of operators with functional space capacities \citep{Steinwart2008}, under mild assumptions, we can take $a > 1 - 2 / d$.
We discuss all assumptions in more detail in Appendix~\ref{lap:app:operators}.

To study the consistency of our algorithms, we can reuse the extensive literature on kernel ridge regression \citep{Caponnetto2007,Lin2020}.

This literature body provides an extensive picture on convergence rates relying on various filters and assumptions of capacity, a.k.a. effective dimension, and source conditions.
Our setting is slightly different and showcases two specificities:
({\em i}) the eigenelements $(\lambda_{i, \mu}, \theta_{i,\mu})$ are dependent of $\mu$;
({\em ii}) the low-rank approximation in Algorithm \ref{lap:alg:imp} is specific to settings with derivatives.
We end our exposition with the following convergence result, proven in Appendix \ref{lap:app:consistency}.
Note that the dependency of $p$ in $n$ can be improved based on subsampling techniques that leverage expressiveness of the different $(k_{X_i})$ \citep{Rudi2015}.
Moreover, universal consistency results could also be provided when the RKHS is dense in $H^1$, as well as convergence rates for other filters and laxer assumptions which we discuss in Appendix \ref{lap:app:consistency} (in particular, the source condition can be relaxed by considering the biggest $q\in(0, 1]$ such that $g\in\ima{\cal L}^q$).

  \begin{figure}[t]
    \centering
    \includegraphics{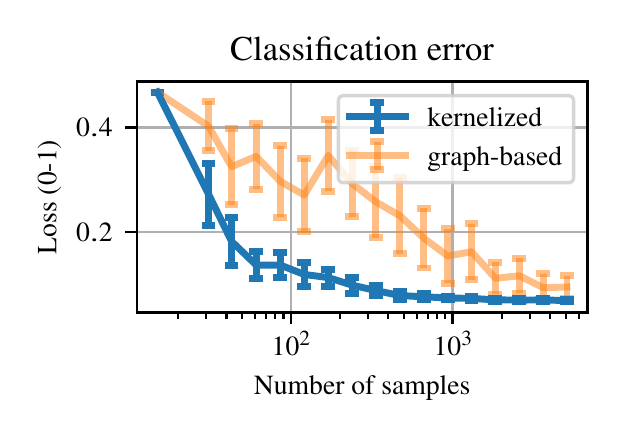}
    \includegraphics{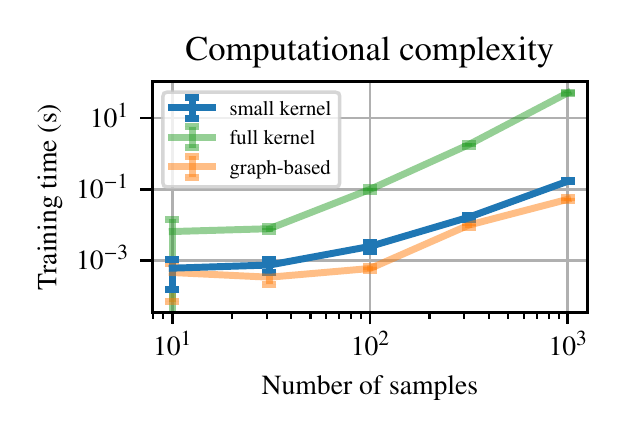}
    \vskip -0.1in
    \caption{(Left) Comparison between our kernelized Laplacian method
      (Tikhonov regularization version with $\lambda = 1$, $\mu_n = n^{-1}$, $p=50$) and graph-based Laplacian based on the same Gaussian kernel with bandwidth $\sigma_n = n^{-\frac{1}{d+4}}\log(n)$ as suggested by graph-based theoretical results \citep{Hein2007}.
      We report classification error as a function of the number of samples $n$.
      The error is averaged over 50 trials, with error bars representing standard deviations.
      We fixed the ratio $n_\ell / n$ to one tenth, and generated the data according to two Gaussian in dimension $d=10$ with unit variance and whose centers are at distance $\delta = 3$ of each other (similar to the setting of \citep{Castelli1995,Lelarge2019}).
      Our method discovers the structure of the data much faster than graph-based Laplacian (to get a 20\% error we need $40$ points, while graph-based needs $700$).
      (Right) Time to perform training with graph-based Laplacian in orange, with Algorithm \ref{lap:alg:imp} in blue (with the specification of the left figure), and with the naive representation in $\R^{n(d+1)}$ of the empirical minimizer \eqref{lap:eq:estimate} in green.
      When dealing with $1000$ points, our algorithm, as well as graph-based Laplacian, can be computed in about one tenth of a second on a 2 GHz processor, while the naive kernel implementation requires 10 seconds.
      We show in Appendix \ref{lap:sec:experiment} that this cut in costs is not associated with a loss in performance.
    }
    \label{lap:fig:exp}
    \vskip -0.15in
  \end{figure}

  \begin{theorem}[Convergence rates]
    \label{lap:thm:consistency}
    Under Assumptions \ref{lap:ass:source}, \ref{lap:ass:approximation} and \ref{lap:ass:decay}, for $n_\ell, n \in \N$, when considering the spectral filtering Algorithm \ref{lap:alg:imp} with $\psi_{\lambda}: x \to (x + \lambda)^{-1}$, there exists a constant $C$ independent of $n$, $n_\ell$, $\lambda$, $\mu$ and $p$ such that the estimate $\hat{g}_p$ defined in Algorithm \ref{lap:alg:imp} verifies
    \begin{equation}
      \E_{{\cal D}_n}\bracket{\norm{\hat{g}_p - g_\rho}_{L^2}^2} \leq
      C \Big(\lambda^2 + \lambda\mu + \frac{\sigma^2_\ell n_\ell^{-1} + n_\ell^{-2} + n^{-1}}{\lambda\mu}
      + \frac{\log(p)}{p}
      + \lambda \frac{\log(p)^a}{p^a}\Big),
    \end{equation}
    with $\sigma_\ell^2$ is a variance parameter that relates to the variance of the variable $Y(I + \lambda {\cal L})^{-1}\delta_X$, inheriting its randomness from $(X, Y) \sim \rho$.
    In particular, when the ratio $r = n_\ell / n$ is fixed, with the regularization scheme $\lambda_n = \lambda_0 n^{-1/4}$, $\mu_n = \mu_0 n^{-1/4}$, for any $\lambda_0 > 0$ and $\mu_0 > 0$, and the subsampling scheme $p_n = p_0 n^s \log(n)$ for any $p_0 > 0$ and with $s = \max(\sfrac{1}{2}, \sfrac{1}{4a})$, there exists a constant $C'$ independent of $n$ and $n_\ell$ such that the excess of risk verifies
    \begin{equation}
      \E_{{\cal D}_n}\bracket{\norm{\hat{g}_p - g_\rho}_{L^2}^2} \leq
      C' (n^{-1/2} + \sigma_\ell^2 n_\ell^{-1/2}).
    \end{equation}
  \end{theorem}

  Theorem \ref{lap:thm:consistency} answers the two questions asked at the beginning of this section.
  In particular, it characterizes the dependency of the need for labeled data to a variance parameter linked with the diffusion of observations $(X_i, Y_i)$ based on the density $\rho_\X$ through the operator ${\cal L}$.
  Intuitively, if the index set $I \subset \{1, 2, \cdots, n\}$ of data $(X_i)_{i\in I}$ we labeled does not change the profile of the diffusion solution $\hat{g}$, then we do not need that much labeled data -- as this is the case on Figure \ref{lap:fig:intro}.

  Finally, Theorem~\ref{lap:thm:consistency} is remarkable in that it exhibits no dependency to the dimension of $\X$ in the power of $n$ and $n_\ell$.
  This contrasts with graph-based Laplacian methods that do not scale well with the input space dimensionality \citep{Bengio2006,Hein2007}.
  Indeed, Figure \ref{lap:fig:exp} shows the superiority of our method over graph-based Laplacian in dimension $d=10$, with a mixture of Gaussian.
We provide details as well as additional experiments in Appendix \ref{lap:sec:experiment}.

\section{Conclusion}

Diffusing information or enforcing regularity through penalties on derivatives are natural ideas to tackle many machine learning problems.
Those ideas can be captured informally with graph-based techniques and finite element differences, or captured more formally with the diffusion operator we introduced in this work.
This formalization allowed us to shed light on Laplacian regularization techniques based on statistical learning considerations.
In order to make our method usable in practice, we provided strong computational guidelines to cut down prohibitive costs associated with a naive implementation of our methods.
In particular, we were able to develop computationally efficient semi-supervised techniques that do not suffer from the curse of dimensionality.

This work paves the way to many extensions beyond semi-supervised learning.
For example, in Appendix \ref{lap:sec:extension}, we describe its usefulness to the partial supervised learning problem, where minimizing the Dirichlet energy provide a learning principle, in order to bypass the restrictive non-ambiguity assumption usually made in this setup \citep{Cour2011,Liu2014,Cabannes2020,Cabannes2021}.
Moreover, in the context of active learning, retaking the strategy of \citet{Karzand2020}, this energy provides a computationally-effective, theoretically-grounded, data-dependent score to select the next point to query.
As such, follow-ups would be of interest to see how this introductory theoretical paper makes its way into the world of concrete applications.

\begin{subappendices}
  \chapter*{Appendix}
  \addcontentsline{toc}{chapter}{Appendix}

  \section{Extension to partially supervised learning}
\label{lap:sec:extension}

In this section, we first show how our work can be extended to generic semi-supervised learning problems, beyond real-valued regression.
This first extension is based on the least-square surrogate introduced by \citet{Ciliberto2020} for structured prediction problems.
We later show how our work can be extended to generic partially-supervised learning.
This second extension is based on the work of \citet{Cabannes2020}.

\subsection{Structured prediction and least-square surrogate}
\label{lap:sec:structured_prediction}
Until now, we have considered the least-square problem with $Y \in \R$.
Indeed, our work can be extended easily to a wide class of learning problems.
Consider $\Y$ an output space, $\ell:\Y\times\Y\to\R$ a loss function, and keep $\X\subset \R^d$ and $\rho\in\prob{\X\times\Y}$.
Suppose that we want to retrieve
\begin{equation}
  \label{lap:eq:structured_prediction}
  f^* = \argmin_{f:\X\to\Y} {\cal R}(f),\qquad\text{with}\qquad
  {\cal R}(f) = \E_{(X, Y)\sim \rho} \bracket{\ell(f(X), Y)}.
\end{equation}
\citet{Ciliberto2020} showed that
as soon as $\ell$ can be decomposed through two mappings $\phi:\Y\to{\cal H}_\Y$ and $\psi:\Y\to{\cal H}_\Y$ with ${\cal H}_\Y$ a Hilbert space as $\ell(y, z) = \scap{\phi(y)}{\psi(z)}_{{\cal H}_\Y}$, it is possible to leverage the least-square regression by considering the surrogate problem
\begin{equation}
  \label{lap:eq:surrogate}
  g^* \in \argmin_{g:\X\to{\cal H}_\Y}
  \E_{(X, Y)\sim\rho}\bracket{\norm{g(X) - \phi(Y)}^2_{{\cal H}_\Y}}.
\end{equation}
This surrogate problem relates to the original one through the decoding $d$ that relates a surrogate estimate $g:\X\to{\cal H}_\Y$ to an estimate of the original problem $f:\X\to\Y$ as $f = d(g)$ defined through, for $x\in\supp\rho_\X$,
\begin{equation}
  \label{lap:eq:decoding}
  f(x) = \argmin_{z\in\Y} \scap{\psi(z)}{g(x)}_{{\cal H}_\Y}.
\end{equation}
In the real-valued regression case, presented previously, our estimates for $g_n$ can all be written as $g_n(x) = \sum_{i=1}^{n_\ell} \beta_i(x) Y_i$, where $\beta_i(x)$ is a function of the $(X_i)_{i\leq n}$, involving the kernel $k$ and its derivatives.
Those estimates can be cast to vector-valued regression by considering coordinates-wise regression,\footnote{%
To parametrize functions $g$ from $\X$ to ${\cal H}_\Y$, we can parametrize independently each coordinates $\scap{g}{e_i}_{{\cal H}_\Y}$, for $(e_i)$ a basis of ${\cal H}_\Y$, by the space ${\cal G}$ -- note that it is possible to generalize real-valued kernel to parametrize coordinates in a joint fashion \citep{Caponnetto2007}.
The coordinate-wise parametrization corresponds to the tensorization ${\cal H}' = {\cal H}_\Y \otimes {\cal H}$ and to the parametric space ${\cal G}' = \brace{x\to \Theta k_x \midvert \Theta \in {\cal H}'}$ of functions from $\X$ to ${\cal H}_\Y$.
${\cal G}'$ naturally inherits the Hilbertian structure of ${\cal H}'$, itself inherited from the structure of ${\cal H}$ and ${\cal H}_\Y$.
} which leads to $g_n(x) = \sum_{i=1}^{n_\ell}\beta_i(x) \phi(Y_i)$, and to the original estimates, for any $x\in\supp\rho_\X$,
\begin{equation}
  \label{lap:eq:loss_trick}
  f_n(x) \in \argmin_{z\in{\cal Z}} \sum_{i=1}^{n_\ell} \beta_i(x) \ell(z, Y_i).
\end{equation}
The behavior of $f_n$ being independent of the decomposition $(\phi, \psi)$ of $\ell$ was referred to as the loss trick.
In particular, \citet{Ciliberto2020} showed that convergence rates derived between $\norm{g_n - g^*}_{L^2}$ does not change if we consider $g:\X\to\R$ or $g:\X\to{\cal H}_\Y$ and that those rates can be cast directly as convergence rates between ${\cal R}(f_n)$ and ${\cal R}(f^*)$ with $f_n = d(g_n)$ defined by~\eqref{lap:eq:decoding}.
Moreover, when $\Y$ is a discrete output space, it is possible to get much better generalization bound on ${\cal R}(f_n) - {\cal R}^*$by introducing geometrical considerations regarding $g^*$ and decision frontier between classes \citep{Cabannes2021b}.

\begin{example}[Binary classification]
  \label{lap:ex:binary}
  This framework aims at generalizing well known surrogate considerations in the case of the binary classification.
  Binary classification corresponds to $\Y = \brace{-1, 1}$, $\ell$ the $0-1$ loss.
  In this setting, ${\cal H}_\Y = \R$, $\phi: \Y\to\R; y \to y$, and $\psi = -\phi$.
  This definition verifies $\ell(y, z) = .5 - .5 \phi(y)^\top \phi(z) \simeq \phi(y)^\top \psi(z)$.
  This corresponds to the usual least-square surrogate, which is \( {\cal R}_S(g) = \E[\norm{g(X) - Y}^2]\), \(g(x) = \E\bracket{Y\midvert X=x}\) and $f = \sign g$.
\end{example}

\paragraph{Beyond least-squares.}
Considering a least-square surrogate assumes that retrieving $g^*$ \eqref{lap:eq:surrogate} is the way to solve the original problem \eqref{lap:eq:structured_prediction} and that the low-density separation hypothesis can be expressed as Assumption \ref{lap:ass:source} being verified by $g^*$.
We would like to point out that the low-density separation could be expressed under a much weaker form, which is that there exists $g$ such that $f^* = d(g)$ \eqref{lap:eq:decoding} and $g$ verifies Assumption \ref{lap:ass:source}.
In particular, the cluster assumption \citep{Rigollet2007} could be understood as assuming that $g = \phi(f^*)$, the trivial embedding of $f^*$ in ${\cal H}_\Y$, is constant on clusters, which means that $g$ belongs to the kernel of the Laplacian operator ${\cal L}$.
Yet, $g^*:x\to\E[\phi(Y)\vert X=x]$, which depends on the labeling noise, could be really non-smooth, even under the cluster assumption.
Those considerations are related to an open problem in machine learning, which is that we do not know what is the best statistical way (and the best surrogate problem) to solve the fully supervised binary classification problem \citep[see {\em e.g.}][]{Zhang2020}.
However, many points introduced in the work could be retaken with other surrogate, could it be SVM (which leads to $g^* = \phi(f^*)$, with $g^*$ minimizing the Hinge loss), softmax regression (used in deep learning) or others.

\subsection{Partially supervised learning}
Partial supervision is a popular instance of weak supervision, which generalizes semi-supervised learning.
It has been known under the name of partial labeling \citep{Cour2011}, superset learning \citep{Liu2014}, as well as learning with partial label \citep{Grandvalet2002}, with partial annotation \citep{Lou2012}, with candidate labeling set \citep{Luo2010} or with multiple label \citep{Jin2002}.
It encompasses many problems such as ``classification with partial labels'' \citep{Nguyen2008,Cour2011}, ``multilabeling with missing labels'' \citep{Yu2014}, ``ranking with partial ordering'' \citep{Hullermeier2008}, ``regression with censored data'' \citep{Tobin1958}, ``segmentation with pixel annotation'' \citep{Verbeek2007,Papandreou2015}, as well as instances of ``action retrieval'', especially on instructional videos \citep{Alayrac2016,Miech2019}.
It consists, for a given input $x$, in not observing its label $y\in\Y$, but observing a set of potential labels $s\in 2^\Y$ that contains the labels ($y\in s$).
Typically, if $\Y$ is the space $\Sfrak_m$ of orderings between $m$ items (\emph{e.g.} movies on a streaming website), for a given input $x$ (\emph{e.g.} some feature vectors characterizing a user) $s$ might be specified by a partial ordering that the true label $y$ should satisfy (\emph{e.g.} the user prefers romantic movies over action movies).

In this setting, it is natural to create consensus between the different sets giving information on $(y\vert x)$, which has been formalized mathematically by the infimum loss $(z,s)\in\Y\times 2^\Y \to \inf_{y\in s} \ell(z, y)\in\R$ for $\ell:\Y\times\Y\to\R$ a specified loss on the underlying fully supervised learning problem.
This leads, for $\tau \in \prob{\X\times 2^\Y}$ encoding the distribution generating samples $(X, S)$, to the formulation
\(
f^* \in {\cal F} = \argmin_{f:\X\to\Y} \E_{(X,
  S)\sim\tau}\bracket{\inf_{Y\in S}\ell(f(X), Y)}.
\)
To study this problem, a non-ambiguity assumption is usually made \citep{Cour2011,Luo2010,Liu2014,Cabannes2020,Cabannes2021}.
This is a very strong assumption to ensure that ${\cal F}$ is, in essence, a singleton.
Highly adequate to this setting, the Laplacian regularization allows relaxing this assumption, assuming that ${\cal F}$ can be big, but that we can discriminate between function in ${\cal F}$ by looking for the smoothest one in the sense defined by the Laplacian penalty.
Moreover, the loss trick \eqref{lap:eq:loss_trick} allows endowing, in an off-the-shelf fashion, the recent work of \citet{Cabannes2020,Cabannes2021} on the partial supervised learning problem with our considerations on Laplacian regularization.
  \section{Experiments}
\label{lap:sec:experiment}

\subsection{Low-rank approximation}

\begin{figure*}[t]
  \centering
  \includegraphics{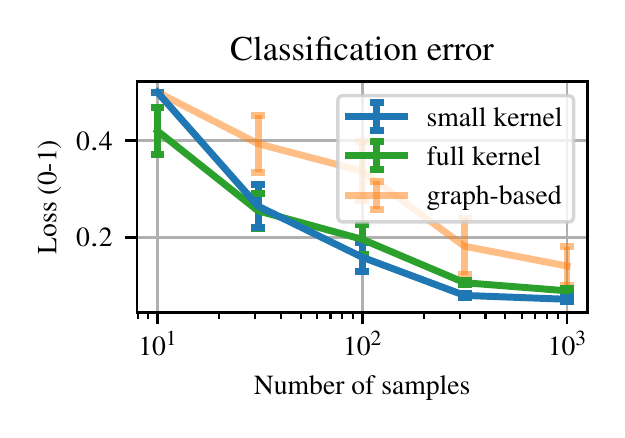}
  \includegraphics{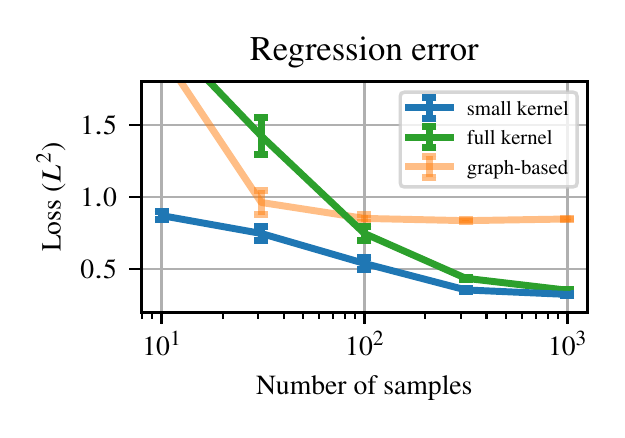}
  \vskip -0.1in
  \caption{
    Cut in computation costs are not associated with a loss in performance.
    The estimate $\hat{g}_p$ Algorithm \ref{lap:alg:imp} (in blue), based on low-rank approximation that cut computation cuts performs as well as the exact computation of $\hat{g}$~\eqref{lap:eq:estimate}.
    (Left) Classification error in the setting of Figure \ref{lap:fig:exp}.
    (Right) Regression error in the same setting.
    The fact that the error of the graph-based method stalls around one, is due to the amplitude of the estimate being very small, which is coherent with behaviors described in \citep{Nadler2009}.
  }
  \label{lap:fig:cut_loss}
\end{figure*}

Cutting computation cost thanks to low-rank approximation, as we did by going from the naive exact empirical risk minimizer $\hat{g}$~\eqref{lap:eq:estimate} to the smart implementation $\hat{g}_p$ Algorithm \ref{lap:alg:imp}, is associated with a trade-off between computational versus statistical performance.
This trade-off can be studied theoretically thanks to Theorem \ref{lap:thm:consistency}, which shows that under mild assumptions, considering $p = n^{1/2}\log(n)$ does not lead to any loss in performance, in the sense that the convergence rates in $n$, the number of samples, are only changed by a constant factor.
We show on Figure \ref{lap:fig:cut_loss} that in the setting of Figure \ref{lap:fig:exp}, our low-rank approximation is not associated with a loss in performance.
Actually low-rank approximation can even be beneficial as it tends to lower the risk for overfitting as discussed by \citet{Rudi2015}.

\subsection{Comparison with graph-based Laplacian}

One the main goal of this paper is to make people drop graph-based Laplacian methods and adopt our ``kernelized'' technique.
As such, we would like to discuss in more detail our comparison with graph-based Laplacian.
In particular, we will discuss how and why we choose the hyperparameters and the setting of Figure \ref{lap:fig:exp}.

\begin{figure*}[t]
  \centering
  \includegraphics{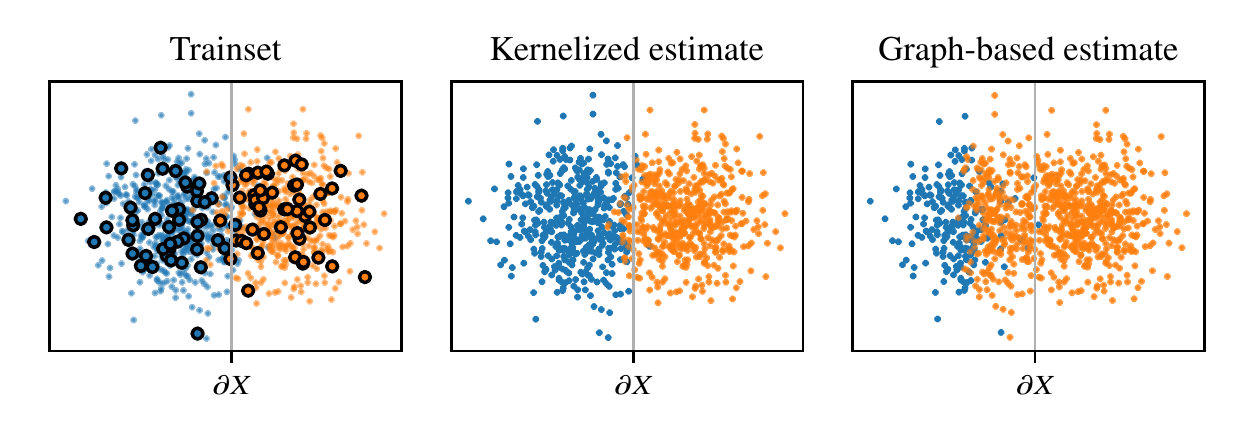}
  \vskip -0.1in
  \caption{
    Setting of Figure \ref{lap:fig:exp} with $n = 1000$.
    (Left) Training set.
    We represent a cut of $\X \subset \R^d$ according to the two first coordinates $\brace{(x_1, x_2) \midvert (x_1, x_2, \cdots, x_d)\in \X}$.
    We have two Gaussian distributions with unit variance, one centered at $x = (0, 0, \cdots 0)$ and the other one centered at $x = (3, 0, \cdots, 0)$.
    One of the Gaussian distributions is associated with the blue class, the other one with the orange class.
    We consider $n = 1000$ unlabeled points, represented by small points, colored according to their classes, and $n_\ell = 100$ labeled points, represented in color with black edges.
    (Middle) Reconstruction with our kernelized Laplacian methods.
    Our method uncovers correctly the structure of the problem, and allows making a quite optimal reconstruction.
    The optimal decision frontier is illustrated by the gray line $\partial X$.
    (Right) Reconstruction with graph-Laplacian.
    The graph-Laplacian diffuses information too far away from what it should, leading to many incorrect guesses.
  }
  \label{lap:fig:exp_setting}
\end{figure*}

The setting of Figure \ref{lap:fig:exp} is the one of Figure \ref{lap:fig:exp_setting}, we considered two Gaussian with unit variance and whose centers are at distance $\delta = 3$ of each other.
We chose Gaussian distributions as it is a well-understood setting.
We chose $\delta = 3$ so that there is a mild overlap between the two distributions.
For the bandwidth parameter, we considered $\sigma_n = Cn^{-\frac{1}{d+4}}\log(n)$ as this is known to be the optimal bandwidth for graph Laplacian \citep{Hein2007}.
We chose $C = 1$ as this leads to $\sigma_n$ of the order of $\delta$.
We chose $\lambda = 1$ to enforce Laplacian regularization and $\mu_n = 1/n$, as this is a classical regularization parameter in RKHS.
Furthermore, we did not cross-validate parameters in order to be fair with graph-Laplacian that do not have as many parameters as our kernel method.
We compute the error in a transductive setting, retaking the exact problem and algorithm of \citet{Zhu2003}.
We choose $d = 10$, as we know that this is a good dimension parameter in order to illustrate the curse of dimensionality phenomenon without needing too much data.\footnote{%
  Note that our consistency result Theorem \ref{lap:thm:consistency} describes a convergence regime that applies to a vast class of problems.
  Such a regime usually takes place after a certain number of data (depending on the value of the constant $C$).
  Before entering this regime, describing the error of our algorithm would require more precise analysis specific to each problem instance, eventually involving tools from random matrix theory.
}

\subsection{Usefulness of Laplacian regularization}

\begin{figure*}[t]
  \centering
  \includegraphics{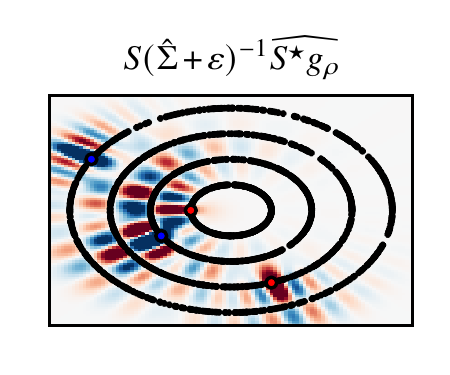}
  \includegraphics{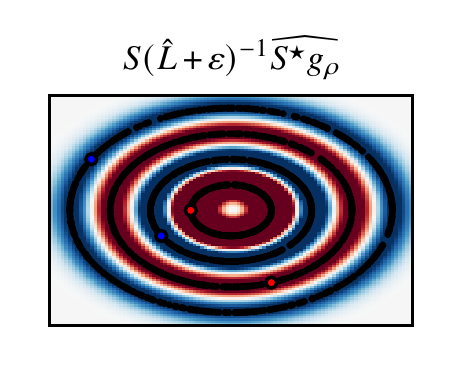}
  \vskip -0.2in
  \caption{
    Usefulness of Laplacian regularization.
    We illustrate the reconstruction based on our spectral filtering techniques based on the sole use of the covariance matrix $\Sigma$ on the left, and on the sole use of the Laplacian matrix $L$ on the right.
    We see that the covariance matrix does not capture the geometry of the problem, which contrasts with the use of Laplacian regularization.
  }
  \label{lap:fig:use}
  \vskip -0.1in
\end{figure*}

It is natural to ask about the relevance of Laplacian regularization.
To give convergence results, we have used Assumptions \ref{lap:ass:source} and \ref{lap:ass:approximation}, which imply that $g^*$ belongs to the RKHS ${\cal H}$, and we got convergence rates in $n_l^{1/2}$, which is not better than the rates we could get with pure kernel ridge regression.
In particular, our algorithm can be split between an unsupervised part that learn the penalty $\norm{{\cal L}^{1/2}g}_{L^2(\rho_\X)}^2$ and a supervised part, that solve the problem of estimating $g_\lambda$ from few labels $(X_i, Y_i)$ given the penalty associated to ${\cal L}$.
But the same method can be used for pure kernel ridge regression: unsupervised data could be leveraged to learn the covariance matrix $\Sigma$ \eqref{lap:eq:block_op}, and supervised data could be used to get $\widehat{S^\star g_\rho}$ to converge toward $S^\star g_\rho$.
The same analysis would yield the same type of convergence rates.
Yet the parameter $\sigma_\ell$ appearing in Theorem \ref{lap:thm:consistency} would not be linked with the variance of $Y(I + \lambda{\cal L} + \lambda\mu K^{-1})^{-1}\delta_X$ but with the variance of $Y(I + \mu K^{-1})^{-1}\delta_X$.
This is a key fact, the geometry of the covariance operator $\Sigma$ is not supposed to be that relevant to the problem, while the one of $L$ is.
We illustrate this fact on Figure \ref{lap:fig:use}.

  \section{Central operators}
\label{lap:app:operators}

The paper makes an intensive use of operators.
This section aims at providing details and intuitions on those operators, in order to help the reader.
In particular, we discuss Assumptions \ref{lap:ass:source} and \ref{lap:ass:approximation}, and we prove the equality in~\eqref{lap:eq:block_op}.

\begin{table}[ht]
  \caption{Notations}
  \label{lap:tab:notations}
  \vskip 0.1in
  \centering
  \begin{tabular}{cc}
    \toprule
    Symbol                         & Description                                                               \\
    \midrule
    $(X_i)_{i\leq n}$              & $n$ samples of input data                                                 \\
    $(Y_i)_{i\leq n_l}$            & $n_l$ labels                                                              \\
    $\rho$                         & Distribution of $(X, Y)$                                                  \\
    $g_\rho$                       & Function to learn \eqref{lap:eq:least_square}                             \\
    $\lambda, \mu$                 & Regularization parameters                                                 \\
    $g_{\lambda}, g_{\lambda,\mu}$ & Biased estimates (\ref{lap:eq:laplacian_tikhonov}, \ref{lap:eq:tikhonov}) \\
    $\hat{g}$                      & Empirical estimate \eqref{lap:eq:estimate}                                \\
    $\hat{g}_p$                    & Empirical estimate with low-rank approximation (Algo. \ref{lap:alg:imp})  \\
    ${\cal H}$                     & Reproducing kernel Hilbert space                                          \\
    $k$                            & Reproducing kernel                                                        \\
    $S$                            & Embedding of ${\cal H}$ in $L^2$                                          \\
    $S^\star$                      & Adjoint of $S$, operating from $L^2$ to ${\cal H}$                        \\
    $\Sigma = S^\star S$           & Covariance operator on ${\cal H}$                                         \\
    $K = SS^\star$                 & Equivalent of $\Sigma$ on $L^2$                                           \\
    ${\cal L}$                     & Diffusion operator (a.k.a. Laplacian)                                      \\
    $L = S^\star{\cal L}S$         & Restriction of the diffusion operator to ${\cal H}$                       \\
    $g$                            & Generic element in $L^2$                                                  \\
    $\theta$                       & Generic element in ${\cal H}$                                             \\
    $\lambda_i$                    & Generic eigenvalue                                                        \\
    $e_i$                          & Generic eigenvector in $L^2$                                              \\
    \bottomrule
  \end{tabular}
\end{table}

\subsection{The diffusion operator}
In this subsection, we extend on the diffusion operator, and recall its basic properties.

The diffusion operator is a well-known operator in the realm of partial differential equations.
Let us assume that $\rho_\X$ admits a smooth density $\rho_\X(dx) = p(x) dx$, say $p \in {\cal C}^2(\R^d)$ that cancels outside a domain $\Omega \subset \R^d$.
Then the diffusion operator ${\cal L}$ can be explicitly written, for $g$ twice differentiable, as
\[
  {\cal L}g(x) = - \Delta g(x) + \frac{1}{p(x)} \scap{\nabla p(x)}{\nabla g(x)}.
\]
This follows from the fact that for $f$ once and $g$ twice differentiable, using Stokes theorem,
\begin{align*}
  \scap{f}{{\cal L} g}_{L^2(\rho_\X)}
   & = \scap{\nabla f}{\nabla g}_{L^2(\rho_\X)}
  = \scap{\nabla f}{ p \nabla g}_{L^2(\diff x)}
  \\&= \int_\X \Div(f p \nabla g ) \diff x -
  \scap{f}{\Div(p\nabla g)}_{L^2(\diff x)}
  = -\scap{f}{\Div(p \nabla g)}_{L^2(\diff x)}
  \\&= -\scap{f}{p^{-1}\Div (p \nabla g)}_{L^2(\rho_\X)}
  = -\scap{f}{\Div \nabla g + p^{-1} (\nabla p). \nabla g}_{L^2(\rho_\X)}.
\end{align*}
Note that when the distribution is uniform on $\Omega$, the diffusion operator is exactly the opposite of the usual Laplacian operator $\Delta$.
As for the Laplacian case, it can be shown that under mild assumption on $p$, whose smoothness properties directly translates to the smoothness properties of the boundary of $\Omega$, that the diffusion operator ${\cal L}$ has a compact resolvent (that is, for $\lambda \notin \spec({\cal L})$,
$({\cal L} + \lambda I)^{-1}$ is compact).
This is a standard result implied by a standard version of the famous Rellich-Kondrachov compactness embedding theorem: $H^2(\Omega)$ is compactly injected in $L^2(\Omega)$ whenever $\Omega$ is a bounded open with $C^2$-boundaries.

In such a setting, we can consider the eigenvalue decomposition of ${\cal L}^{-1}$, that is, $(\lambda_i, e_i) \in (\R_+ \times L^2)^\N$, with $(e_i)_{i\in\N}$ an orthonormal basis of $L^2$ and $(e_i)_{i\leq \dim\ker{\cal L}}$ generating the null space of ${\cal L}$, with the convention $\lambda_i = M$ for $i \leq \dim\ker {\cal L}$, with $M$ an abstraction representing $+\infty$, and $(\lambda_i)$ decreasing toward zero afterwards.
This decomposition reads
\begin{equation}
  \label{lap:eq:L}
  {\cal L}^{-1} = \sum_{i \in \N} \lambda_i e_i \otimes e_i.
\end{equation}
Note that the fact that all the $(\lambda_i)$ are positive, is due to the fact that ${\cal L}^{-1}$ is the inverse of a positive self-adjoint operator.
As a consequence, the diffusion operator has discrete spectrum, and can be written as
\begin{equation}
  \label{lap:eq:evd}
  {\cal L} = \sum_{i\in\N} \lambda_i^{-1} e_i \otimes e_i.
\end{equation}
In such a setting, the kernel-free Tikhonov regularization~\eqref{lap:eq:laplacian_tikhonov} reads
\begin{equation}
  \label{lap:eq:decomposition_zero}
  g_\lambda = \sum_{i\in\N} \psi(\lambda_i)\scap{g_\rho}{\lambda_i^{1/2} e_i}_{L^2}
  \lambda_i^{1/2} e_i,
\end{equation}
with $\psi:x \to (x+\lambda)^{-1}$, and the convention $M\psi(M) = 1$.

\subsection{Regularity of the eigenvectors of the diffusion operator}
In this subsection, we extend on the regularity assumed in Assumption \ref{lap:ass:approximation}.

Introducing the kernel $k$ and its associated RKHS ${\cal H}$ is useful when the eigenvectors of ${\cal L}$ can be well approximated by functions in ${\cal H}$.
In applications, people tend to go for kernels that are translation-invariant, which implied that the RKHS ${\cal H}$ is made of smooth functions, could it be analytical functions (for the Gaussian kernel) or functions in $H^m$ (for Sobolev kernels).
As a consequence, we should investigate the regularity of those eigenvectors.
Indeed, if $\rho$ derives from a Gibbs potential, that is $\rho(\diff x) = e^{-V(x)} \diff x$, the eigenvectors of ${\cal L}$ can be shown to inherit from the smoothness of the potential $V$ \citep{PillaudVivien2020b}.
For example if $V$ belongs to $H^m$, and $H^m \subset {\cal H}$, we expect $(e_i)$ to belong to ${\cal H}$, thus verifying Assumption \ref{lap:ass:approximation}.

\paragraph{Counter-example and beyond Assumption \ref{lap:ass:approximation}.}
Note that if $\rho$ has several connected components of non-empty interiors, the null space of ${\cal L}$ is made of functions that are constants on each connected component of $\supp\rho_\X$.
Those functions are not analytic.
In such a setting, the Gaussian kernel is not sufficient for Assumption \ref{lap:ass:approximation} to hold, and one should favor kernel associated with richer functional space such as the Laplace kernel or the neural tangent kernel \citep{Chen2021}.
However, as illustrated by Figure \ref{lap:fig:unsupervised}, $e_i$ not belonging to ${\cal H}$ does not mean that $e_i$ can not be well approximated by ${\cal H}$.
Indeed, it is well known that the approximation power of ${\cal H}$ for $e_i$ can be measure in the biggest power $p$ such that $e_i \in \ima K^p$ \citep{Caponnetto2007}, where $K = SS^*$.
Assumption \ref{lap:ass:approximation} corresponds to $p=1/2$, but it should be seen as a specific instance of more generic approximation conditions.

\paragraph{Handling constants in RKHS.}
Finally, note also that many RKHS do not contain constant functions, and therefore might not contain the constant function $e_0$ (although we are only looking for equality in the support of $\rho_\X$), however this specific point with $e_0$ can easily be circumvented either by assuming that $g_\rho$ has zero mean, either by centering the covariance matrices $\Sigma$ and $\hat\Sigma$ \citep{PillaudVivien2020b}.
This relates with the usual technique for SVM consisting in adding an unpenalized bias \citep{Steinwart2008}.

\subsection{Low-density separation}
In this subsection, we discuss how Assumption \ref{lap:ass:source} relates to the idea of low-density separation.

\paragraph{Low-variation intuition.}
The low-density separation supposes that the variations of $g^*$ take place in region with low-density, so that $\norm{{\cal L}^{1/2}g^*} / \norm{g^*}$ is small.
As such, using Courant-Fischer principle, Assumption \ref{lap:ass:source} can be reformulated as $g^*$ belonging to the space
\[
  \Span\brace{e_i}_{i\leq r} =
  \argmin_{\substack{{\cal F}\subset L^2;\\ \dim{\cal F} = r}}
  \max_{g\in{\cal F}} \frac{\norm{{\cal L}^{-1/2}g}_{L^2}^2}{\norm{g}_{L^2}^2}.
\]
In other terms, Assumption \ref{lap:ass:source} can be restated as $g^*$ belonging to a finite dimensional space that minimizes a measure of variation given by the Dirichlet energy.

To tell the story differently, suppose that we are in a classification setting, {\em i.e.} $Y\in\brace{-1, 1}$, and that the $\supp\rho_\X$ is connected.
Then we know that the null space of ${\cal L}$ is made of constant functions.
Then the first eigenvector $e_2$ of ${\cal L}$ is a function that is orthogonal to constants.
Hence, $e_2$ is a function that changes its sign and that is ``balanced'' in the sense that $\E[e_2] = 0$ -- {\em i.e.} if $e_2(x) = \E_{\mu}[Y\,\vert\, X=x]$ for some measure $\mu$, we have $\E_\mu[Y] = 0$, meaning that classes are ``balanced''.
Moreover, in order to minimize $\norm{{\cal L}^{1/2}e_2}$, the variations of $e_2$ should take place in low-density regions of $\X$.

\paragraph{Diffusion intuition.}
Finally, as ${\cal L}$ is a diffusion operator, we also have an interpretation of Assumption \ref{lap:ass:source} in terms of diffusion.
Consider $(\lambda_i, e_i)$ the eigenelements of~\eqref{lap:eq:evd}.
The diffusion of $g_\rho$ according the density $\rho_\X$ can be written as, for $t \in \R$,
\[
  g_t = e^{-t {\cal L}} g_\rho = \sum_{i\in\N} e^{-t\lambda_i^{-1}} \scap{g_\rho}{e_i} e_i.
\]
This diffusion will cut off the high frequencies of $g_\rho$ that corresponds to $\scap{g_\rho}{e_i}$ for big $i$, and big $\lambda_i^{-1}$.
Indeed, the difference between the diffusion and the original $g_\rho$ can be measured as
\[
  \norm{g_t - g_\rho}_{L^2}^2 = \sum_{i\in\N} (e^{-t\lambda_i^{-1}} - 1)^2\scap{g_\rho}{e_i}^2
  = \sum_{i\in\N} t^2\lambda_i^{-2} \scap{g_\rho}{e_i}^2 + o(t^2\lambda_i^{-2}).
\]
Hence, assuming that $g_\rho$ is supported on a few of the first eigenvectors of ${\cal L}$, can be rephrased as saying that the diffusion of $g_\rho$ does not modify it too much.

\paragraph{The variance $\sigma_\ell$.}
Theorem \ref{lap:thm:consistency} shows that the need for labels depends on the variance parameter $\sigma_{\ell}^2$.
It is natural to wonder how this parameter relates to the low-density hypothesis.
As we discussed, this parameter is linked to the variance of $Z = Y(I + \lambda {\cal L})^{-1}\delta_X$.
We can separate the variability of this variable due to $X$ and the variability due to $Y$
\[
  Z = Z_X + Z_Y, \qquad\text{with}\qquad
  Z_X = (I + \lambda {\cal L})^{-1} g_\rho(X)\delta_X,\quad
  Z_Y = (I + \lambda {\cal L})^{-1} (Y - \E[Y\,\vert\,X])\delta_X.
\]
As such we see that this variance depends on the structure of the density $\rho_\X$ with the variance of $(I + \lambda{\cal L})^{-1}\delta_X$, and the labeling noise with the variance of $(Y\,\vert\, X)$.
The low-density separation does not tell us anything about the level of noise in $Y$ or the diffusion structure linked with $\rho_\X$, but additional hypotheses could be made to characterize those.

\subsection{Kernel operators}

In this subsection, we define formally the operators $S$ and $\Sigma$.

We now turn toward operators linked with the Hilbert space ${\cal H}$.
Recall that for $k:\X\to\X\to\R$ a kernel, ${\cal H}$ is defined the closure of the span of the elements $k_x$ under the scalar product $\scap{k_x}{k_{x'}} = k(x, x')$.
In particular, $\norm{k_x}^2_{\cal H} = k(x, x)$.
${\cal H}$ parametrizes a vast class of functions in $\R^\X$ through the mapping
\[
  \myfunction{S}{\cal H}{\R^\X}{\theta}{(\scap{k_x}{\theta})_{x\in\X}.}
\]
Under mild assumptions, $S$ maps ${\cal H}$ to a space of functions belonging to $L^2$.

\begin{proposition}
  When $x\to k(x, x)$ belongs to $L^1(\rho_\X)$, $S$ is a continuous mapping from ${\cal H}$ to $L^2(\rho_\X)$.
  This is particularly the case when $\rho_\X$ has compact support and $k$ is continuous.
\end{proposition}
\begin{proof}
  Consider $\theta\in{\cal H}$, we have
  \begin{align*}
    \norm{S\theta}_{L^2}^2 & = \int_\X \scap{k_x}{\theta}^2\rho_\X(\diff x)
    \leq \int_\X \scap{k_x}{\theta}^2_{\cal H}\rho_\X(\diff x)
    \leq \int_\X \norm{k_x}^2_{\cal H} \norm{\theta}^2_{\cal H}\rho_\X(\diff x)
    \\&= \norm{\theta}_{\cal H}^2 \int_\X k(x, x)\rho_\X(\diff x)
    = \norm{\theta}^2_{\cal H}\norm{x\to k(x, x)}_{L^1}.
  \end{align*}
  Moreover, when $\rho_\X$ has compact support and $k$ is continuous, $k$ is bounded on the support of $\rho_\X$ therefore $x\to k(x, x)$ belongs to $L^1$.
\end{proof}

As a continuous operator from the Hilbert space ${\cal H}$ to the Hilbert space $L^2$, $S$ is naturally associated with many linear structures: in particular its adjoint $S^\star$, but also the self-adjoint operators $K = SS^\star$ and $\Sigma = S^\star S$.

\begin{proposition}
  The adjoint of $S$ is defined as
  \[
    \myfunction{S^\star}{L^2}{\cal H}{g}{\int_\X g(x)k_x \rho_\X(\diff x)
      = \E_{X\sim\rho_\X}[g(X)k_X].}
  \]
  To $S$ is associated the kernel self-adjoint operator on $L^2$
  \[
    \myfunction{K := SS^\star}{L^2}{L^2}{g}{(x\to\int_\X k(x, x') g(x') \rho_\X(\diff x')),}
  \]
  as well as the (non-centered) covariance on ${\cal H}$,
  \(
  \Sigma := S^\star S = \E_{X\sim\rho_\X}[k_X\otimes k_X].
  \)
\end{proposition}
\begin{proof}
  We shall prove the equality defining those operators.
  Consider $\theta \in {\cal H}$ and $g \in L^2$, we have
  \[
    \scap{S^\star g}{\theta}_{\cal H}
    = \scap{g}{S\theta}_{L^2}
    = \E_{X\sim\rho_\X}[g(X) \scap{k_X}{\theta}_{\cal H}]
    = \scap{\E_{X\sim\rho_\X}[g(X)k_X]}{\theta}_{\cal H}.
  \]
  We also have, for $x\in \X$,
  \[
    (SS^\star g)(x) = \scap{k_x}{\E_{X\sim\rho_\X}[g(X)k_X]}_{\cal H}
    = \E_{X\sim\rho_\X}[g(X)\scap{k_x}{k_X}_{\cal H}]
    = \E_{X\sim\rho_\X}[g(X)k(X, x)].
  \]
  Finally, we have
  \[
    S^\star S \theta = \E_{X\sim\rho_\X}[S\theta(X)k_X]
    = \E_{X\sim\rho_\X}[\scap{\theta}{k_X}_{\cal H}k_X]
    = \E_{X\sim\rho_\X}[k_X\otimes k_X]\theta.
  \]
  This provides the last of all the equalities stated above.
\end{proof}

\paragraph{The functional space \texorpdfstring{${\cal H}$}{}.}
In the main text, we have written everything in terms of $\theta$, highlighting the parametric nature of kernel methods.
This made it easier to dissociate the norm on functions derived from ${\cal H}$ and the one derived from $L^2$ or $H^1$.
In literature, people tend to keep everything in terms of functions $g_\theta = S\theta$ without even mentioning the dependency in $\theta$.
Such a setting consists in considering directly the RKHS ${\cal H}$ whose scalar product is defined for $g, g'\in (\ker K)^\perp$ by $\scap{g}{g'}_{\cal H} = \scap{g}{K^{-1}g'}_{L^2}$.

\subsection{Derivative operators}
In this subsection, we extend on derivatives in RKHS, and we formally define the operator $L$.

As well as evaluation maps can be represented in ${\cal H}$, under mild assumptions, derivative evaluation maps can benefit from such a property.
Indeed, for $g_\theta = S\theta$, $x \in \X$ and $u \in {\cal B}_\X(0, 1)$ a unit vector, we have
\[
  \partial_u g_\theta(x) = \lim_{t\to 0} \frac{g_\theta(x+tu) - g_\theta(x)}{t}
  = \lim_{t\to 0} \frac{\scap{\theta}{k_{x+tu}}_{\cal H} - \scap{g}{k_x}_{\cal H}}{t}
  = \lim_{t\to 0} \scap{\theta}{\frac{k_{x+tu} - k_x}{t}}_{\cal H}
\]
As a linear combination of elements in ${\cal H}$, the difference quotient evaluation map $t^{-1}(k_{x+tu} - k_x)$ belongs to ${\cal H}$ and has a norm
\[
  \norm{\frac{k_{x+tu} - k_x}{t}}^2_{\cal H}
  = \frac{k(x+tu, x+tu) - 2k(x+tu, x) + k(x, x)}{t^2}.
\]
In order for the limit when $t$ goes to zero to belong to ${\cal H}$, we see the importance of $k$ to be twice differentiable.
This limit $\partial_u k_x$, whose existence is proven formally by \citet{Zhou2008}, provides a derivative evaluation map in the sense that
\[
  \partial_u g_\theta(x) = \scap{\theta}{\partial_u k_x}_{\cal H}.
\]
From this equality, we derive that $\partial_{1i}k(x, x') = \scap{k_{x'}}{\partial_i k_x}$, and recursively that $\scap{\partial_i k_x}{\partial_j k_{x'}} = \partial_{1i}\partial_{2j} k(x, x')$.

Similarly to the operator $S$, we can introduce the operators $Z_i$ for $i\in\bracket{1, d}$, defined as
\[
  \myfunction{Z_i}{\cal H}{\R^\X}{\theta}{(\scap{\partial_ik_x}{\theta}_{\cal H})_{i\leq d}}.
\]
Once again, under mild assumptions, $\ima Z_i$ inherit from a Hilbertian structure.
\begin{proposition}
  When $x\to \partial_{1i} \partial_{2i} k(x, x)$ belongs to $L^1(\rho_\X)$, $Z_i$ is a continuous mapping from ${\cal H}$ to $L^2(\rho_\X)$.
  This is particularly the case when $\rho_\X$ has compact support and $k$ is twice differentiable with continuous derivatives.
\end{proposition}
\begin{proof}
  Consider $\theta\in{\cal H}$, similarly to before, we have
  \begin{align*}
    \norm{Z\theta}_{L^2}^2 & = \int_\X \scap{\partial_i k_x}{\theta}^2\rho_\X(\diff x)
    \leq \norm{\theta}^2_{\cal H} \int_\X \norm{\partial_i k_x}^2_{\cal H}\rho_\X(\diff x)
    = \norm{\theta}^2_{\cal H}\norm{x\to \partial_{1,i}\partial_{2,i} k(x, x)}_{L^1}.
  \end{align*}
  Moreover, when $\rho_\X$ has compact support and $\partial_{1,i}\partial_{2,i}k$ is continuous, $\partial_{1,i}\partial_{2,i}k$ is bounded on the support of $\rho_\X$ therefore $x\to\partial_{1,i}\partial_{2,i}k$ belongs to $L^1$.
\end{proof}
Among the linear operators that can be built from $Z_i$, in the theoretical part of this paper, we are mainly interested in $Z_i^\star Z_i$.
In the empirical part however, we might be interested in $Z_i Z_j^\star$ as well as $Z_iS^\star$ as it appears in Algorithm \ref{lap:alg:imp} (where the notation $Z_n$ there has to be understood as the empirical version of $Z = [Z_1; \cdots; Z_d]$).

\begin{proposition}
  The Dirichlet energy on ${\cal H}$ can be represented through the operator
  \[
    S^\star {\cal L} S = \sum_{i=1}^d Z_i^\star Z_i
    = \sum_{i=1}^d \E_{X\sim\rho_\X}[\partial_i k_X\otimes \partial_i k_X].
  \]
\end{proposition}
\begin{proof}
  Let $\theta \in {\cal H}$ and $g_\theta = S\theta$, we have
  \begin{align*}
    \scap{g_\theta}{{\cal L}g_\theta}_{L^2}
     & = \scap{\theta}{S^\star{\cal L}S\theta}_{\cal H}
    = \E_{X\sim\rho_\X}\bracket{\norm{\nabla g_\theta(X)}^2}
    = \sum_{i=1}^d\E_{X\sim\rho_\X}\bracket{(\partial_i g_\theta(X))^2}
    \\&= \sum_{i=1}^d\E_{X\sim\rho_\X}\bracket{\scap{\partial_i k_X}{\theta}^2_{\cal H}}
    = \sum_{i=1}^d\norm{Z_i\theta}^2_{L^2}
    = \scap{\theta}{\sum_{i=1}^d Z_i^\star Z_i \theta}_{\cal H}
    \\&= \sum_{i=1}^d\E_{X\sim\rho_\X}\bracket{\scap{\theta}{(\partial_i k_X \otimes \partial_i k_X)\theta}_{\cal H}}
    = \scap{\theta}{\sum_{i=1}^d\E_{X\sim\rho_\X}\bracket{\partial_i k_X \otimes \partial_i k_X}\theta}_{\cal H}.
  \end{align*}
  Since the three operators are self-adjoint, and they all represent the same quadratic form, they are equals.
\end{proof}

\subsection{Relation between \texorpdfstring{$\Sigma$}{} and \texorpdfstring{$L$}{}}

In this subsection, we discuss the relation between $\Sigma$ and $L$ and show that we can expect the existence of $a \in (1 - 2 / d, 1]$ and $c > 0$ such that $L \preceq c \Sigma^a$.

\paragraph{Informal capacity considerations.}
We want to compare $\Sigma$ and $L$, as $L \preceq c\Sigma^a$ with the biggest $a$ possible.
This depends on how fast the eigenvalues are decreasing, which is linked to the entropy numbers of those two compact operators.
Those entropy numbers are linked with the capacity of the functional spaces $\brace{g\in L^2\midvert \norm{K^{-1/2}g}_{L^2} < \infty}$ and $\brace{g\in L^2\midvert \norm{K^{-1/2}{\cal L}^{-1/2}g}_{L^2} < \infty}$.
The first space is the reproducing kernel Hilbert space linked with $k$, the second space is, roughly speaking, a space of functions whose integral belongs to the first space.
As such, if the first space is ${\cal H}^m$, the second is ${\cal H}^{m-1}$, and we can consider $a = (m - 1) / m$.
Because we are considering kernels, we have $m > d / 2$ (this to make sure that the evaluation functionals $L_X : f\to f(x)$ are continuous), so that $a > 1 - 2 / d$.
Without trying to make those ``algebraic'' considerations more formal, we will give an example on the torus.

\paragraph{Translation-invariant kernel and Fourier transform.}
Consider $L^2([0, 1]^d, \diff x)$ the space of periodic functions in dimension $d$, square integrable against the Lebesgue measure on $[0, 1]^d$.
For simplicity, we will suppose that $\rho_\X$ is the Lebesgue measure on $[0, 1]^d$.
Consider a translation invariant kernel
\[
  k(x, y) = q(x - y) \qquad \text{for } q:\R^d \to \R \text{ that is one periodic}.
\]
In this setting, the operator $K$, operating on $L^2$, is the convolution by $q$, that is
\[
  \myfunction{K}{L^2}{L^2}{g}{q*g}, \qquad\text{hence}\qquad
  \widehat{Kg} = \hat{q}\hat{g}.
\]
Where we have used the fact that convolutions can be represented by a product in the Fourier domain.
Note that, from B\"ochner theorem, we know that $k$ being positive definite implies that the Fourier transform of $q$ exists and is not negative.
Let us define the Fourier coefficient and the inverse Fourier transform as
\[
  \forall\,\omega\in\Z^d, \quad
  \hat{g}(\omega) = \int_{[0,1]^d} g(x)e^{-2i\pi \omega^\top x} \diff x,
  \quad\text{and}\quad
  \forall\,x\in[0, 1]^d, \quad
  g(x) = \sum_{\omega \in \Z^d} e^{2i\pi \omega^\top x} \hat{g}(\omega).
\]
$K$ being a convolution operator, it is diagonalizable with eigenelements $(\hat{q}(\omega), x\to e^{2i\pi \omega^\top x})_{\omega\in\Z^d}$.
From this, we can make explicit many of our abstract operators.
First of all, using Perceval's theorem,
\[
  \norm{g}_{\cal H}^2 = \scap{g}{K^{-1}g}_{L^2}
  = \sum_{\omega\in\Z^d} \frac{\abs{\hat{g}(\omega)}^2}{\hat{q}(\omega)}.
\]
Hence, we can parametrize ${\cal H}$ with $(\theta_\omega)_{\omega \in \Z^d} \in \C^{\Z^d}$ and the $\ell^2$-metric, where $\theta_\omega = \hat{g}(\omega) / \sqrt{\hat{q}(\omega)}$ and
\[
  (S\theta)(x) = g_\theta(x) = \sum_{\omega\in\Z^d} e^{2i\pi\omega^\top x}\sqrt{\hat{q}(\omega)}\theta_\omega.
\]
Note that this is not the usual parametrization of ${\cal H}$ by elements $\theta\in{\cal H}$ as $(\C^{\Z^d}, \ell^2)$ is not a space of functions.
However, such a parametrization of ${\cal H}$ does not change any of the precedent algebraic considerations on the operators $S$, $\Sigma$, $K$, and $L$.

\paragraph{Diffusion operator and Fourier transform.}
As well as convolution operators are well represented in the Fourier domain, derivation operators are.
Indeed, when $g$ is regular, we have
\[
  \norm{{\cal L}^{1/2} g}^2_{L^2} =
  \norm{\nabla g}^2_{L^2} = \sum_{j=1}^d \norm{\partial_j g}^2_{L^2}
  = \sum_{j=1}^d \sum_{\omega \in\Z^d} \omega_j^2 \abs{\hat{g}(\omega)}^2.
\]
As a consequence, using the expression of $S\theta$, we have
\[
  \Sigma\theta = \sum_{\omega\in\Z^d} \hat{q}(\omega) \theta_\omega,
  \qquad\text{while}\qquad
  L\theta = \sum_{\omega\in\Z^d} \norm{\omega}_2^2 \hat{q}(\omega) \theta_\omega,
  \quad\text{where}\quad \norm{\omega}_2^2 = \sum_{j=1}^d \omega_i^2.
\]
In this setting, the eigenelements of $\Sigma$ are $(\hat{q}(\omega), \delta_\omega)_{\omega\in\Z^d}$ while the one of $L$ are $(\norm{\omega}^2_2\hat{q}(\omega), \delta_\omega)_{\omega\in\Z^d}$.

\paragraph{Eigenvalue decay comparison.}
Hence, having $L \preceq c\Sigma^a$ is equivalent to having $\norm{\omega}_2^2\hat{q}(\omega) \leq c\hat{q}(\omega)^a$.
Now suppose that the decay of $\hat{q}$ is governed by
\[
  c_1 (1 + \sigma^{-1}\norm{\omega}_2^2)^{-m} \leq \hat{q}(\omega)
  \leq c_2 (1 + \sigma^{-1}\norm{\omega}_2^2)^{-m},
\]
for two constants $c_1, c_2 > 0$.
In particular, this is verified for Mat{\'e}rn kernels, corresponding to the fractional Sobolev space $H^m$, and for the Laplace kernel with $m = (d+1) / 2$, which reads $k(x, y) = \exp(-\sigma^{-1}\norm{x-y})$.
The Gaussian kernel could be seen as $m = +\infty$ as it has exponential decay.
With such a decay we have, assuming without restrictions that we are in one dimension
\begin{align*}
  \omega^2\hat{q}(\omega) \leq c_2\omega^2 (1 + \sigma^{-1}\omega^2)^{-m}
  \leq c_2\sigma(1 + \sigma^{-1}\omega^2)^{-(m-1)}
  \leq c_1^{\frac{m}{m-1}}c_2\sigma \hat{q}(\omega)^{\frac{m-1}{m}}.
\end{align*}
In other terms, we can consider $c = c_1^{\frac{m}{m-1}}c_2 \sigma$ and $a = (m-1)/m$.
Assuming that $q$ is square-integrable, so is $\hat{q}$, which implies that $2m > d$.
As a consequence, we do have $a > 1 - 2 / d$.
Note that this reasoning could be extended to the case where $\rho_\X$ has a density against the Lebesgue measure, that is bounded above and below away from zero.
  \section{Spectral decomposition}
\label{lap:app:algebra}

In this section, we recall facts on spectral regularization, before proving Proposition \ref{lap:thm:decomposition} and extending it to the case $\mu = 0$.

\subsection{Generalized singular value with matrices}

\paragraph{Generalized singular value decomposition.}
Let $A \in \R^{m1\times n}$ and $B \in \R^{m_2\times n}$ be two matrices.
There exists $U \in \R^{m_1 \times m_1}$, $V\in \R^{m_2 \times m_2}$ two orthogonal matrices, $c \in \R^{m_1 \times r}$ and $s\in \R^{m_2 \times r}$ two $1$-diagonal matrices such that $c^\top c + s^\top s = I_r$, and $H\in \R^{n\times r}$ a non-singular matrix such that
\[
  A = UcH^{-1}, \qquad B = VsH^{-1}.
\]
To be more precise $c$ is such that only entries $c_{ii} = \cos (\theta_i)$ for $i < \min(r, m_2)$ are non-zeros and $s$ such that only entries $s_{m_1-i, r-i} = \sin(\theta_{r-i})$ for $i < \min(r, m_1)$ are non-zeros, with $\theta_i \in [-\pi/2, \pi/2]$ some angle.
Here, $c$ stands for cosine, $s$ for sinus and $r$ for rank.

\paragraph{Link with generalized eigenvalue problem.}
As well as the singular value of $A$ is linked with the eigenvalue of $A^\top A$, the generalized singular value decomposition of $[A; B]$ is linked with the generalized eigenvalue problem linked with $(A^\top A, B^\top B)$.
Indeed, we have
\[
  A^\top A = H^{-\top} c^\top c H^{-1}, \qquad B^\top B = H^{-\top} s^\top s H^{-1}.
\]
In particular, with $(e_i)$ the canonical basis of $\R^r$, and $h_i$ the $i$-th column of $H$, we get
\[
  H^\top A^\top A h_i = \cos(\theta_i)^2 e_i = \tan(\theta_i)^{-2}
  \sin(\theta_i)^2 e_i
  = \tan(\theta_i)^{-2} H^\top B^\top B h_i.
\]
From which we deduce that, since $\ima A \cup \ima B \subset \ima H^\top$,
\[
  A^\top A h_i = \tan(\theta_i)^{-2} B^\top B h_i,\qquad
  h_j^\top B^\top B h_i = \sin(\theta_i)^2 \ind{i=j}.
\]
So if we denote by $f_i = \abs{\sin(\theta_i)}^{-1} h_i$ and $\lambda_i = \tan(\theta_i)^{-2}$, assuming $\lambda_i \neq 0$ for all $i \leq r$ (which corresponds to $\ker B \subset \ker A$), $(\lambda_i)_{i\leq r}, (f_i)_{i\leq r}$ provide the generalized eigenvalue decomposition of $(A^\top A, B^\top B)$ in the sense that
\[
  A^\top A f_i = \lambda_i B^\top B f_i,\qquad
  f_j^\top B^\top B f_i = \ind{i=j},\qquad
  f_j^\top A^\top A f_i = \lambda_i \ind{i=j}.
\]

\subsection{Tikhonov regularization}
Define the Tikhonov regularization
\[
  x_\lambda = \argmin_{x\in \R^n} \norm{Ax - b}^2 + \lambda \norm{Bx}^2.
\]
When this problem is well-defined, the solution is defined as
\[
  x_\lambda = (A^\top A + \lambda B^\top B)^\dagger A^\top b.
\]
With the generalized singular value decomposition of $A$ and $B$, we have
\[
  A^\top A + \lambda B^\top B = H^{-\top} \gamma_\lambda H^{-1},
  \quad\text{with}\quad \gamma_\lambda = c^\top c + \lambda s^\top s.
\]
Using the fact that $A^\top b = H^{-\top} c^\top U^\top b$, we get
\[
  x_\lambda = H \gamma_\lambda^{-1} c^\top U^\top b
  = \paren{\sum_{i=1}^r \frac{\cos(\theta_i)}{\cos(\theta_i)^2 + \lambda \sin(\theta_i)^2} h_i \otimes u_i} b.
\]
Now, we would like to replace $c_{ii}$, $s_{ii}$, $h_i$ and $u_i$ with quantities that depend on $\lambda_i$, $f_i$ and $A$.
To do so recall that $AH = Uc$, therefore $\cos(\theta_i) u_i = Ah_i$, and recall that $h_i = \sin(\theta_i) f_i$ and $\lambda_i = \cos(\theta_i)^2 / \sin(\theta_i)^2$.
Inputting this equality in the last expression of $x_\lambda$ we get
\[
  x_\lambda = \paren{\sum_{i=1}^r \frac{\sin(\theta_i)^2}{\cos(\theta_i)^2 +
      \lambda \sin(\theta_i)^2} f_i \otimes Af_i} b
  = \paren{\sum_{i=1}^r \frac{1}{\lambda_i +
      \lambda} f_i \otimes Af_i} b.
\]
Finally,
\[
  b_\lambda = A x_\lambda
  = \sum_{i=1}^r \psi(\lambda_i) \scap{Af_i}{b} Af_i,
  \qquad\text{where}\qquad
  \psi(x) = \frac{1}{x+\lambda}.
\]

\subsection{Extension to operators}
To end the proof of Proposition \ref{lap:thm:decomposition}, we should prove that we can apply the generalized eigenvalue decomposition to operators.
We will only prove that it is possible for $(\Sigma, L+\mu)$ based on simple considerations.
\begin{proposition}
  When $k$ is continuous and $\supp\rho_\X$ is bounded, $\Sigma$ is a compact operator.
\end{proposition}
\begin{proof}
  We have $\Sigma = \E[k_X\otimes k_X]$ and $\norm{k_x} = k(x, x)$.
  Since $k$ is continuous and $\supp\rho_\X$ is compact, for $x\in\supp\rho_\X$, $k(x, x)$ is bounded.
  Hence, $\Sigma$ is a trace class and compact operator.
\end{proof}
\begin{proposition}
  When $k$ is twice differentiable with continuous derivative, and $\supp\rho_\X$ is compact, $L$ is a compact operator.
  As a consequence, $L$ has a compact spectrum, and has a pseudo-inverse that we will denote, with a slight abuse of notation, by $L^{-1}$.
\end{proposition}
\begin{proof}
  The proof is similar to the one showing that $\Sigma$ is compact, based on the fact that $L = \sum_{i=1}^d \E[\partial_i k_X \otimes \partial_i k_X]$, and $\norm{\partial_i k_X}^2 = \partial_{1i}\partial_{2i} k(x, x)$.
\end{proof}
\begin{proposition}
  When $\Sigma$ is compact, for all $\mu > 0$, $(L+\mu)^{-1/2}\Sigma (L+\mu)^{-1/2}$ is a compact operator.
\end{proposition}
\begin{proof}
  The proof is straightforward
  \[
    \trace((L+\mu)^{-1/2} \Sigma (L+\mu)^{-1/2}) =
    \trace(\Sigma (L+\mu)^{-1}) \leq
    \norm{(L+\mu)^{-1}}_{\op} \trace(\Sigma)
    \leq \mu^{-1} \trace(\Sigma) < +\infty.
  \]
  Therefore, the operator is trace class, hence compact.
\end{proof}
\begin{proposition}
  For any $\mu > 0$, the generalized eigenvalue decomposition of $(\Sigma, L+\mu)$ as defined in Proposition \ref{lap:thm:decomposition} exists.
\end{proposition}
\begin{proof}
  Using the spectral theorem, since $(L+\mu)^{-1/2}\Sigma (L+\mu)^{-1/2}$ is positive self-adjoint compact operator, there exists $(\xi_i)$ a basis of ${\cal H}$ and $(\lambda_i) \in \R_+$ a decreasing sequence (note that $\ker (L+\mu) = \ker \Sigma = \brace{0}$), such that
  \[
    (L+\mu)^{-1/2}\Sigma (L+\mu)^{-1/2} = \sum_{i\in\N} \lambda_i \xi_i \otimes \xi_i.
  \]
  Taking $\theta_i = (L + \mu)^{-1/2} \xi_i$, we get $\Sigma \theta_i = \lambda_i L\theta_i$.
  Because $(\xi_i)$ generates ${\cal H}$, and $(L+\mu)^{-1/2}$ is bijective (since $L$ is compact, $(L+\mu)^{-1}$ is coercive), $((L+\mu)^{-1/2}\xi_i)$ generates ${\cal H}$.
\end{proof}

Proposition \ref{lap:thm:decomposition} follows from prior discussion on Tikhonov regularization extended to infinite summations.

\subsection{The case \texorpdfstring{$\mu = 0$}{}}

When $\mu = 0$, \eqref{lap:eq:filtering} should be seen as the rewriting of~\eqref{lap:eq:decomposition_zero} based on the RKHS ${\cal G} = \ima S$.
This can only be done when the eigenvectors of ${\cal L}$ appearing in~\eqref{lap:eq:evd} belongs to ${\cal G} = \ima S = \ima K^{1/2}$, which is exactly what Assumption \ref{lap:ass:approximation} provides.
In such a setting, we can find $(\theta_i) \in {\cal H}^\N$ to write $\lambda_i^{1/2} e_i = S\theta_i$ as soon as $\lambda_i \neq 0$ (write $M e_i = S\theta_i$ for $M$ an abstraction representing $+\infty$ when $\lambda_i = 0$, handling the potential fact that $\ker B \not\subset \ker A$), we get $\theta_iS^*S\theta_j = \lambda_i\ind{i=j}$, and $L \theta_i = \lambda_i^{-1}\Sigma\theta_i$, and we can extend Proposition \ref{lap:thm:decomposition} to the case $\mu = 0$, with
\begin{equation}
  \label{lap:eq:filtering_zero}
  g_\lambda = \sum_{i\in\N} \psi(\lambda_i) \scap{S^\star g_\rho}{\theta_i} S\theta_i,
\end{equation}
where we handle the null space of ${\cal L}$ with the equality $M\psi(M) = 1$, verified by $M$ our abstraction representing $+\infty$, so that $\psi(M) \scap{S^\star g_\rho}{\theta_i}S\theta_i = \scap{g_\rho}{e_i} e_i$ as soon as $\lambda_i = 0$.

\paragraph{Beyond Assumption \ref{lap:ass:approximation}.}
Assumption \ref{lap:ass:approximation} could be made generic by considering the biggest $(p_i)\in\R_+^\N$ such that $K^{-p_i} e_i$ belongs to $L^2$, and rewriting~\eqref{lap:eq:filtering_zero} under the form $g_\lambda = \sum_{i\in\N} \psi(\lambda_i) \scap{(S_0 K^{p_i})^\star g_\rho}{\theta_i} S_0 K^{p_i} \theta_i$, with $\theta_i = \lambda_i^{-1/2} S_0^{-1} K^{-p_i}\theta_i$ and $S_0 = K^{-1/2}S$ the isomorphism between ${\cal H}$ and $L^2$ (assuming that $S$ is dense in $L^2$).
Such an assumption would describe all situations from no assumption ($p_i = 0$ for all $i$), Assumption \ref{lap:ass:approximation} ($p_i = 1/2$ for all $i$) to even more optimistic assumptions ($p_i \geq 1$ for all $i$).

  \section{Consistency analysis}
\label{lap:app:consistency}

This section is devoted to the proof of Theorem \ref{lap:thm:consistency}.
The proof is based on~\eqref{lap:eq:filtering} and \eqref{lap:eq:filtering_zero}, and splits the error of $\norm{g_\rho - \hat{g}_p}_{L^2}^2$ into several components linked with how well we approximate $S^\star g_\rho$, and how well we approximate the eigenvalue decomposition $(\lambda_i, \theta_{i})$ of $(\Sigma, L)$.

\subsection{Sketch and understanding of the proof}
In this subsection, we explain how the proofs work for consistency theorems such as Theorem \ref{lap:thm:consistency}.

Let us define the mapping $G:{\cal H}\times {\cal C}\to L^2$ with $\cal C$ the set of pairs of self-adjoint operators on ${\cal H}$ that admit a generalized eigenvalue decomposition, as
\begin{equation}
  G(\theta, (A, B)) = \sum_{i\in\N} \psi(\lambda_i) \scap{\theta}{\theta_i} S\theta_i
  \qquad\text{with}\qquad (\lambda_i, \theta_i) \in \operatorname{GEVD}(A, B).
\end{equation}
$G(\theta, (A, B)) \in L^2$ corresponds to writing $\theta\in{\cal H}$ in the basis associated with the generalized eigenvalue decomposition (GEVD) of $(A, B)$.

\begin{proposition}
  Under Assumptions \ref{lap:ass:source} and \ref{lap:ass:approximation}, and with $\psi$ defined in Theorem \ref{lap:thm:consistency}
  \[
    g_\lambda = G(S^\star g_\rho, (\Sigma, L)), \qquad\text{and}\qquad
    \hat{g}_p = G(\widehat{S^\star g_\rho}, (P\hat\Sigma P, P\hat LP + \mu P)),
  \]
  with $P$ the projection matrix from ${\cal H}$ to $\Span\brace{k_{X_i}}_{i\leq p}$.
\end{proposition}
\begin{proof}
  This is a direct application of Assumptions~\ref{lap:ass:source}, \ref{lap:ass:approximation}, \eqref{lap:eq:filtering_zero} and Algorithm~\ref{lap:alg:imp}.
\end{proof}

The main point of the proof is to relate $g_\rho$ to $\hat{g}_p$.
To do so, we will use several functions in $L^2$ generated by $G$.
We detail our steps in Table \ref{lap:tab:step}.
Table \ref{lap:tab:step} gives a first answer to the two questions asked in the opening of Section \ref{lap:sec:statistics}.
The number of unlabeled data controls the convergence of the operators $(P\hat\Sigma P, P\hat LP+\mu P)$ toward $(\Sigma, L + \mu)$.
The number of labeled data controls the convergence of the vector $\widehat{S^\star g_\rho}$ toward $S^\star g_\rho$.
Priors on the structure of the problem, such as source and approximation conditions, control the convergence of the bias estimate $g_{\lambda, \mu}$ toward $g_\rho$.
Furthermore, a more precise study reveals that the concentration of operators are related to efficient dimension \citep{Caponnetto2007} and are accelerated by capacity assumptions on the functional space whose norm is $\norm{g} = \norm{({\cal L} + K^{-1})^{1/2}g}_{L^2}$, and that the concentration of the vector $\widehat{S^\star g_\rho}$ is accelerated by assumptions on moments of the variable $Y(I + \lambda {\cal L})^{-1}\delta_X$ (inheriting randomness from $(X, Y) \sim \rho$).
\begin{table}[h]
  \caption{Steps in the consistency analysis}
  \label{lap:tab:step}
  \vskip 0.1in
  \centering
  \begin{tabular}{cccc}
    \toprule
    Estimate           & Vector                     & Property of convergence                    & Basis                             \\
    \midrule
                       &                            &                                            &                                   \\
    $\hat{g}_p$        & $\widehat{S^\star g_\rho}$ &                                            & $(P\hat\Sigma P, P\hat LP+\mu P)$ \\
                       &                            &                                            &                                   \\
                       &                            & Low-rank approx. \citep{Rudi2015}          & $\downarrow$                      \\
                       &                            &                                            &                                   \\
    $\hat{g}$          & $\widehat{S^\star g_\rho}$ &                                            & $(\hat\Sigma , \hat L+\mu )$      \\
                       &                            &                                            &                                   \\
                       &                            & Operator concentration \citep{Minsker2017} & $\downarrow$                      \\
                       &                            &                                            &                                   \\
    $g_{n_\ell}$       & $\widehat{S^\star g_\rho}$ &                                            & $(\Sigma, L +\mu)$                \\
                       &                            &                                            &                                   \\
                       &                            & Vector concentration \citep{Yurinskii1970} & $\downarrow$                      \\
                       &                            &                                            &                                   \\
    $g_{\lambda, \mu}$ & $S^\star g_\rho$           &                                            & $(\Sigma, L +\mu)$                \\
                       &                            &                                            &                                   \\
                       &                            & Source condition \citep{Lin2020}           & $\downarrow$                      \\
                       &                            &                                            &                                   \\
    $g_{\lambda}$      & $S^\star g_\rho$           &                                            & $(\Sigma,L)$                      \\
                       &                            &                                            &                                   \\
    $\downarrow$       &                            & Source condition \citep{Caponnetto2007}    &                                   \\
                       &                            &                                            &                                   \\
    $g_\rho$           &                            &                                            &                                   \\
    \bottomrule
  \end{tabular}
\end{table}

\paragraph{Control of biases.}
We begin our study in a downward fashion regarding Table \ref{lap:tab:step}.
Indeed, for Tikhonov regularization~\eqref{lap:eq:laplacian_tikhonov}, we can show that for $q \in [0, 1]$,
\[
  \norm{g_\lambda - g_\rho}_{L^2}
  \leq \lambda^q \norm{{\cal L}^q g_\rho}_{L^2}.
\]
Meaning that if we have the source condition $g_\rho \in \ima {\cal L}^q$ (which is a condition on how fast $(\scap{g_\rho}{e_i})_{i\in\N}$ decreases compared to $(\lambda_i)_{i\in\N}$ for $(\lambda_i, e_i)$ the eigenvalue decomposition of ${\cal L}^{-1}$), the rates of convergence of this term when $n$ goes to infinity is controlled by the regularization scheme $\lambda_n^q$.

Similarly to the kernel-free bias above, for $(q_i) \in (0, 1)^\N$, we can have
\[
  \norm{g_{\lambda, \mu} - g_\lambda}_{L^2}^2
  \leq 2 \sum_{i\in\N} \lambda^{2q_i} \mu^{2q_i} \paren{\frac{\lambda_i}{\lambda+\lambda_i}}^2\abs{\scap{e_i}{g_\rho}}^2 \norm{K^{-q_i}e_i}_{L^2}^2.
\]
This shows explicitly the usefulness of controlling at the same time how $g_\rho$ is supported on the eigenspaces of ${\cal L}$ and how the eigenvectors are well approximated by the RKHS ${\cal H}$, which can be read in the value of $(q_i)$ such that all $e_i \in \ima K^{q_i}$.

\paragraph{Vector concentration.}
Let us now switch to concentration of $\widehat{S^\star g_\rho} = n_\ell^{-1}\sum_{i=1}^{n_\ell} Y_i k_{X_i}$ toward $S^\star g_\rho = \E_{(X, Y)\sim \rho}\bracket{Yk_X}$, it will allow controlling $\norm{g_{n_\ell} - g_{\lambda, \mu}}_{L^2}^2$ with the notations appearing in Table~\ref{lap:tab:step}.
Note how this convergence should be measured in terms of the reconstruction error
\[
  \norm{\sum_{i\in\N} \psi(\lambda_{i,\mu})
    \scap{S^\star g_\rho - \widehat{S^\star g_\rho}}{\theta_{i,\mu}}
    S\theta_{i,\mu}}_{L^2}.
\]
This error might behave in a much better fashion than the $L^2$ error between $S S^\star g_\rho$ and $S\widehat{S^\star g_\rho}$.
In particular, on Figure \ref{lap:fig:intro}, we can consider $\psi(\lambda_{i,\mu})= 0$ for $i > 4$, and we might have $\scap{Yk_{X}}{\theta_{i,\mu}} = Y\ind{x\in C_i}$, for $i \leq 4$ and $(X, Y) \in \X\times\Y$, where $C_i$ is the $i$-th innermost circle.
In this setting, when all four $\paren{Y\midvert X\in C_i}$ are deterministic, we only need one labeled point per circle to clear the reconstruction error.
Based on concentration results in Hilbert space, when $\abs{Y}$ is bounded by a constant $c_\Y$, and $x\to k(x, x)$ by a constant $\kappa^2$, we have, with ${\cal D}_{n_\ell}\sim\rho^{\otimes {n_\ell}}$ the dataset generating the labeled data
\[
  \E_{{\cal D}_{n_\ell}}\bracket{\norm{g_{n_\ell} - g_{\lambda, \mu}}^2_{L^2}}
  \leq 2 \sigma_\ell^2 (\mu\lambda n_\ell)^{-1} + \frac{4}{9} c_\Y^2 \kappa^2 (\mu\lambda n_\ell^2)^{-1}.
\]
where $\sigma_\ell^2 \leq c_\Y^2 \trace(\Sigma)$ is a variance parameter to relate to the variance of $Y(I + \lambda {\cal L})^{-1} \delta_X$ (where the randomness is inherited from $(X, Y)\sim\rho$).
The fact that the need for labeled data depends on the variance of $(X, Y)$ after being diffused through ${\cal L}$ is coherent with the results obtained by \citet{Lelarge2019} in the specific case of a mixture of two Gaussian.

\paragraph{Basis concentration.}
We are left with the comparison of $g_{n_\ell}$, which is the filtering of $\widehat{S^\star g_\rho}$ with the operators $(\Sigma, L + \mu)$, and $\hat{g}_p$, which is the filtering of the same vector with the operators $(P\hat\Sigma P, P \hat{L} P + \mu P)$.
As the number of samples grows toward infinity, we know that $(P\hat\Sigma P, P \hat{L} P + \mu P)$ will converge in operator norm toward $(\Sigma, L + \mu)$.
Yet, how to leverage this property to quantify the convergence of $\hat{g}_p$ toward $g_{n_\ell}$?
Let us write $(\lambda_i, \theta_i) = \operatorname{GEVD}(\Sigma, L + \mu)$, and $(\lambda_i', \theta_i') = \operatorname{GEVD}(P\hat\Sigma P, P \hat{L} P + \mu P)$, we have
\[
  \norm{\hat{g}_p - g_{n_\ell}}_{L^2} = \norm{\sum_{i\in\N}
    \psi(\lambda_i)\scap{\theta_i}{\hat\theta_\rho} S\theta_i -
    \psi(\lambda_i')\scap{\theta_i'}{\hat\theta_\rho} S\theta_i'}
  \qquad\text{with}\qquad \hat\theta_\rho = \widehat{S^\star g_\rho}.
\]
The generic study of this quantity requires controlling eigenspaces one by one.
Note that we expect convergence of eigenspaces to depend on gaps between eigenvalues.
However, when considering Tikhonov regularization, this quantity can be written under a simpler form.
In particular, the concentration of operators is controlled, up to few leftovers, through the quantity
$\norm{(\Sigma + \lambda L + \mu\lambda)^{-1/2}((\Sigma - \hat\Sigma) +
    \lambda(L - \hat{L}))(\Sigma + \lambda L + \mu\lambda)^{-1/2}}_{\op}$
where $\norm{\cdot}_{\op}$ designs the operator norm.
In this setting, the low-rank approximation is controlled through $\norm{(\Sigma + \lambda L)^{1/2}(I - P)}_{\op}$, and when $L \preceq c\Sigma^\alpha$, this term can be controlled by $\norm{\Sigma^{1/2} (I - P)}_{\op} + \lambda^{1/2}\norm{\Sigma^{1/2}(I-P)}^\alpha_{\op}$ which can be controlled based on the work of \citet{Rudi2015}.

\subsection{Risk decomposition}
In this subsection, we decompose the risk appearing in Theorem \ref{lap:thm:consistency}.

\subsubsection{Control of biases}
We begin by splitting the error $\norm{g_\rho - \hat{g}_p}_{L^2}$ between a bias term due to the regularization parameters and a variance term due to the data.
With the notation of Table \ref{lap:tab:step},
\begin{equation}
  \norm{g_\rho - \hat{g}_p}_{L^2} \leq \norm{g_\rho - g_\lambda}_{L^2} +
  \norm{g_\lambda - g_{\lambda, \mu}}_{L^2} + \norm{g_{\lambda, \mu} - \hat{g}_p}_{L^2}.
\end{equation}
We will control the first two terms here, and the last term in the following subsections.

\begin{proposition}[Bias in $\lambda$]
  Under Assumption \ref{lap:ass:source}
  \begin{equation}
    \norm{g_\lambda - g_\rho}_{L^2} \leq \lambda \norm{{\cal L}g_\rho}_{L^2}.
  \end{equation}
\end{proposition}
\begin{proof}
  Based on the definition of $g_\lambda = (I + \lambda {\cal L})^{-1}g_\rho$, we have
  \[
    g_\lambda - g_\rho = ((I + \lambda {\cal L})^{-1} - I)g_\rho
    = -\lambda {\cal L} g_\rho.
  \]
  Because $g_\rho$ is supported on the first eigenvectors of the Laplacian (Assumption \ref{lap:ass:source}), we have $g_\rho = \sum_{i=1}^r \scap{g_\rho}{e_i} e_i$, with $e_i$ the eigenvector of ${\cal L}$ appearing in~\eqref{lap:eq:evd}, and
  \[
    \norm{{\cal L}g_\rho}_{L^2}^2 = \norm{\sum_{i=1}^r \lambda_i^{-1}
      \scap{g_\rho}{e_i} e_i}_{L^2}^2
    = \sum_{i=1}^r \lambda_i^{-2} \scap{g_\rho}{e_i}^2 \leq
    \lambda_r^{-2} \norm{g_\rho}_{L^2}^2 < +\infty.
  \]
  This ends the proof of this proposition.
\end{proof}

\begin{proposition}[Bias in $\mu$]
  Under Assumptions \ref{lap:ass:source} and \ref{lap:ass:approximation}, we have
  \begin{equation}
    \norm{g_{\lambda, \mu} - g_\lambda}_{L^2}^2
    \leq \lambda\mu c_a^2\norm{g_\rho}^2_{L^2},\qquad\text{with}\qquad
    c_a^2 = \sum_{i=1}^r \norm{K^{-1/2}e_i}_{L^2}^2 = \sum_{i=1}^r \norm{e_i}_{\cal H}^2.
  \end{equation}
\end{proposition}
\begin{proof}
  Before diving into the proof, recall that the RKHS norm penalization can be written as $\norm{g}_{\cal H} = \norm{K^{-1/2}g}_{L^2}$.
  Using the fact that $A^{-1} - B^{-1} = A^{-1} (B - A) B^{-1}$, we have
  \begin{align*}
    g_{\lambda, \mu} - g_\lambda & =
    ((I + \lambda {\cal L} + \mu\lambda K^{-1})^{-1} - (I + \lambda {\cal L})^{-1}) g_\rho
    \\&= - (I + \lambda {\cal L} + \mu\lambda K^{-1})^{-1} \lambda\mu K^{-1}(I + \lambda {\cal L})^{-1} g_\rho
    \\&= - (\lambda\mu)^{1/2} (I + \lambda {\cal L} + \mu\lambda K^{-1})^{-1/2} (I + \lambda {\cal L} + \mu\lambda K^{-1})^{-1/2} \\&
    \qquad\cdots\times (\lambda\mu K^{-1})^{1/2} K^{-1/2}(I + \lambda {\cal L})^{-1} g_\rho.
  \end{align*}
  As a consequence,
  \begin{align*}
    \norm{g_{\lambda, \mu} - g_\lambda}^2_{L^2}
     & \leq \lambda \mu \norm{K^{-1/2}(I + \lambda {\cal L})^{-1} g_\rho}^2_{L^2},
  \end{align*}
  where we used the fact that $I + \lambda {\cal L} + \mu\lambda K^{-1} \succeq I$, so that $\norm{(I + \lambda {\cal L} + \mu\lambda K^{-1})^{-1/2}}_{\op} \leq 1$ (with $\norm{\cdot}_{\op}$ the operator norm), and that
  \begin{align*}
    \norm{(I + \lambda {\cal L} + \mu\lambda K^{-1})^{-1/2} (\lambda\mu K^{-1})^{1/2}}_{\op}^2
     & =
    \lambda\mu\norm{K^{-1/2}(I + \lambda {\cal L} + \mu\lambda K^{-1})^{-1} K^{-1/2}}_{\op}
    \\&=
    \lambda\mu\norm{(K + \lambda K^{1/2}{\cal L}K^{1/2} + \mu\lambda)^{-1}}_{\op}
    \leq 1.
  \end{align*}
  We continue the proof with
  \begin{align*}
    \norm{K^{-1/2}(I + \lambda {\cal L})^{-1} g_\rho}
     & = \norm{\sum_{i=1}^r \frac{\lambda_i}{\lambda + \lambda_i} \scap{g_\rho}{e_i} K^{-1/2}e_i}
    \leq \sum_{i=1}^r \frac{\lambda_i}{\lambda + \lambda_i} \abs{\scap{g_\rho}{e_i}} \norm{K^{-1/2}e_i}
    \\&\leq \sum_{i=1}^r \abs{\scap{g_\rho}{e_i}} \norm{K^{-1/2}e_i}_{L^2}
    \leq \norm{g_\rho}_{L^2} \paren{\sum_{i\leq r} \norm{K^{-1/2}e_i}^2_{L^2}}^{1/2}.
  \end{align*}
  Putting all the pieces together ends the proof.
\end{proof}

\subsubsection{Vector concentration}
We are left with the study of the variance $\norm{\hat{g}_p - g_{\lambda,\mu}}$.
To ease derivations, we denote $C = \Sigma + \lambda L$, $\hat{C} = \hat\Sigma + \lambda \hat L$, $\theta_\rho = S^\star g_\rho$, $\hat\theta_\rho = \widehat{S^\star g_\rho}$ and $P$ the projection from ${\cal H}$ to $\Span\brace{k_{X_i}}_{i\leq p}$.
We have, for Tikhonov regularization
\begin{align*}
  \norm{\hat{g}_p - g_{\lambda, \mu}}_{L^2}
   & = \norm{S \paren{P(P\hat{C}P + \lambda\mu)^{-1} P \hat\theta_\rho - (C + \lambda\mu)^{-1}\theta_\rho}}_{L^2}                \\
   & = \norm{\Sigma^{1/2} \paren{P(P\hat{C}P + \lambda\mu)^{-1} P \hat\theta_\rho - (C + \lambda\mu)^{-1}\theta_\rho}}_{\cal H}.
\end{align*}
We begin by isolating the dependency to labeled data
\begin{equation}
  \begin{split}
    \norm{\hat{g}_p - g_{\lambda,\mu}}_{L^2}
    &\leq \norm{\Sigma^{1/2} P(P\hat{C}P + \lambda\mu)^{-1} (\hat\theta_\rho - \theta_\rho)}_{\cal H}
    \\&\qquad\cdots+ \norm{\Sigma^{1/2} \paren{P(P\hat{C}P + \lambda\mu)^{-1} \theta_\rho - (C+\lambda\mu)^{-1}\theta_\rho}}_{\cal H}.
  \end{split}
\end{equation}
We will control the first term here, and the second term in the following subsection.

\begin{lemma}[Vector term]
  When $\norm{(C + \lambda\mu)^{-1/2} (\hat{C} - C) (C+\lambda\mu)^{-1/2}}_{\op} \leq 1/2$, we have
  \begin{equation}
    \norm{\Sigma^{1/2} P(P\hat{C}P + \lambda\mu)^{-1} P (\hat\theta_\rho - \theta_\rho)}_{\cal H}
    \leq 2 \norm{(C + \lambda\mu)^{-1/2}(\hat\theta_\rho - \theta_\rho)}_{\cal H}.
  \end{equation}
\end{lemma}
\begin{proof}
  We begin with the splitting
  \begin{align*}
    \norm{\Sigma^{1/2} P(P\hat{C}P + \lambda\mu)^{-1} P (\hat\theta_\rho - \theta_\rho)}_{\cal H}
     & \leq \norm{\Sigma^{1/2} P(P\hat{C}P + \lambda\mu)^{-1} P (C + \lambda\mu)^{1/2}}_{\op}
    \\&\qquad\cdots \times\norm{(C + \lambda\mu)^{-1/2}(\hat\theta_\rho - \theta_\rho)}_{\cal H}.
  \end{align*}
  The first term will concentrate toward a matrix smaller than identity, while the second term concentrates toward zero.
  We can make those considerations more formal.
  Following basic properties with the L\"owner order on operators, we have
  \begin{align*}
     & (C + \lambda\mu)^{-1/2} (C - \hat{C}) (C+\lambda\mu)^{-1/2} \preceq t
    \\\Rightarrow\qquad
     & \hat C \succeq (1 - t) C - t\lambda\mu
    \\\Rightarrow\qquad
     & P\hat CP \succeq (1 - t) PCP - t\lambda\mu P \succeq (1 - t) PCP - t\lambda\mu
    \\\Rightarrow\qquad
     & P\hat CP + \lambda\mu \succeq (1 - t) (PCP + \lambda\mu)
    \\\Rightarrow\qquad
     & (P\hat CP + \lambda\mu)^{-1} \preceq (1 - t)^{-1} (PCP + \lambda\mu)^{-1}
    \\\Rightarrow\qquad
     & (C+\lambda\mu)^{1/2}P(P\hat CP + \lambda\mu)^{-1}P(C+\lambda \mu)^{1/2}
    \\&\qquad\preceq (1 - t)^{-1} (C+\lambda\mu)^{1/2}P(PCP + \lambda\mu)^{-1}P(C+\lambda\mu)^{1/2}
    \preceq (1-t)^{-1},
  \end{align*}
  where we have used the fact that the last operator is a projection.
  As a consequence, for any $t \in (0, 1)$, we have
  \begin{align*}
     & \norm{(C + \lambda\mu)^{-1/2} (\hat{C} - C) (C+\lambda\mu)^{-1/2}}_{\op} \leq t
    \\\qquad\Rightarrow\qquad&
    \norm{(C + \lambda \mu)^{1/2} P(P\hat{C}P + \lambda\mu)^{-1}(C+\lambda\mu)^{1/2}}_{\op}
    \leq (1-t)^{-1}.
    \\\qquad\Rightarrow\qquad&
    \norm{\Sigma^{1/2} P(P\hat{C}P + \lambda\mu)^{-1}P (C+\lambda\mu)^{1/2}}_{\op}
    \leq (1-t)^{-1}.
  \end{align*}
  Where the last implication follows from the fact that $C + \lambda \mu = \Sigma + \lambda L + \lambda\mu \succeq \Sigma$.
\end{proof}

\subsubsection{Basis concentration}
We are left with the study of the basis concentration with the number of unlabeled data.

\begin{lemma}[Basis term]
  When $\norm{(C + \lambda\mu)^{-1/2} (\hat{C} - C) (C+\lambda\mu)^{-1/2}}_{\op} \leq 1/2$, we have
  \begin{equation}
    \begin{split}
      &\norm{\Sigma^{1/2}(P(P\hat{C}P + \lambda\mu)^{-1} - (C + \lambda\mu)^{-1})\theta_\rho}_{\cal H}
      \\&\qquad
      \leq 3 \norm{C^{1/2} (I - P)}_{\op} \norm{g_\lambda}_{\cal H}
      + 2 \norm{(C + \lambda\mu)^{-1/2} (\hat C - C) (C+\lambda\mu)^{-1}\theta_\rho}_{\cal H}.
    \end{split}
  \end{equation}
  Notice that Assumptions \ref{lap:ass:source} and \ref{lap:ass:approximation} imply $\norm{g_\lambda}_{\cal H} \leq c_a\norm{g_\rho}_{L^2} < +\infty$.
\end{lemma}
\begin{proof}
  First of all, using that $A^{-1} - B^{-1} = A^{-1}(B - A)B^{-1}$, notice that
  \begin{align*}
     & \norm{\Sigma^{1/2}(P(P\hat{C}P + \lambda\mu)^{-1} - (C + \lambda\mu)^{-1})\theta_\rho}_{\cal H}
    \\&
    = \norm{\Sigma^{1/2}P(P\hat{C}P + \lambda\mu)^{-1}P(C - \hat{C}P )(C + \lambda\mu)^{-1}\theta_\rho
      - \Sigma^{1/2}(I-P)(C + \lambda\mu)^{-1}\theta_\rho}_{\cal H}
    \\&
    \leq \norm{\Sigma^{1/2}P(P\hat{C}P + \lambda\mu)^{-1}P(\hat{C}P - C)(C + \lambda\mu)^{-1}\theta_\rho}_{\cal H}
    + \norm{\Sigma^{1/2}(I-P)(C + \lambda\mu)^{-1}\theta_\rho}_{\cal H}
    \\&
    \leq \norm{\Sigma^{1/2}P(P\hat{C}P + \lambda\mu)^{-1}P (C+\lambda\mu)^{1/2}}_{\op}
    \norm{(C+\lambda\mu)^{-1/2}P(\hat{C}P - C)(C+\lambda\mu)^{-1}\theta_\rho}_{\cal H}
    \\&
    \qquad\cdots+ \norm{\Sigma^{1/2}(I-P)}_{\op}\norm{(C + \lambda\mu)^{-1}\theta_\rho}_{\cal H}.
  \end{align*}
  Because $\Sigma \preceq \Sigma + \lambda L = C$, we have $\norm{\Sigma^{1/2}(I-P)}_{\op} \leq \norm{C^{1/2} (I-P)}_{\op}$, and we also have
  \[
    \norm{(C+\lambda\mu)^{-1}\theta_\rho}_{\cal H} \leq
    \norm{C^{-1}\theta_\rho}_{\cal H}
    = \norm{K^{-1/2} S C^{-1}\theta_\rho}_{\cal H}
    = \norm{K^{-1/2}g_\lambda}_{L^2} = \norm{g_\lambda}_{\cal H}.
  \]
  Recall, that, for any $t \in (0, 1)$, we have already shown that
  \begin{align*}
     & \norm{(C + \lambda\mu)^{-1/2} (\hat{C} - C) (C+\lambda\mu)^{-1/2}}_{\op} \leq t
    \\\qquad\Rightarrow\qquad&
    \norm{\Sigma^{1/2} P(P\hat{C}P + \lambda\mu)^{-1}P (C+\lambda\mu)^{1/2}}_{\op}
    \leq (1-t)^{-1}.
  \end{align*}
  We are left with one last term to work on
  \begin{align*}
    \norm{(C+\lambda\mu)^{-1/2}P(\hat{C}P - C)(C+\lambda\mu)^{-1}\theta_\rho}_{\cal H}
     & \leq \norm{(C+\lambda\mu)^{-1/2}P(\hat{C} - C)P(C+\lambda\mu)^{-1}\theta_\rho}_{\cal H}
    \\&\cdots+ \norm{(C+\lambda\mu)^{-1/2}C (I-P)(C+\lambda\mu)^{-1}\theta_\rho}_{\cal H}.
  \end{align*}
  We control the first term with the fact for $A, B, C$ three self-adjoint operators and $x$ a vector we have
  \[
    \norm{APBPCx} = \norm{APBPC x\otimes x C PBPA}_{\op}^{1/2},
  \]
  and that
  \begin{align*}
     & PCx\otimes xCP \preceq Cx\otimes xC
    \\\qquad\Rightarrow\qquad&
    PBPC x\otimes xCPBP \preceq BPCx\otimes xCPB \preceq BCx\otimes xCB
    \\\qquad\Rightarrow\qquad&
    APBPC x\otimes xCPBPA \preceq ABCx\otimes xCBA,
  \end{align*}
  so that
  \[
    \norm{(C+\lambda\mu)^{-1/2}P(\hat{C} -
      C)P(C+\lambda\mu)^{-1}\theta_\rho}_{\cal H}
    \leq
    \norm{(C+\lambda\mu)^{-1/2}(\hat{C} - C)(C+\lambda\mu)^{-1}\theta_\rho}_{\cal H}.
  \]
  We control the second term with
  \begin{align*}
     & \norm{(C+\lambda\mu)^{-1/2}C (I-P)(C+\lambda\mu)^{-1}\theta_\rho}
    \\&\qquad\leq \norm{(C+\lambda\mu)^{-1/2}C^{1/2}}\norm{C^{1/2}(I-P)}\norm{(C+\lambda\mu)^{-1}\theta_\rho}.
  \end{align*}
  Using that
  \(
  (C+\lambda\mu)^{-1/2}C^{1/2}\preceq I,
  \)
  we can add up everything to get the lemma.

  For the part concerning $\norm{g_\lambda}_{\cal H}$, notice that
  \begin{align*}
    \norm{g_\lambda}_{\cal H} & = \norm{K^{-1/2}g_\lambda}_{L^2}
    = \norm{ \sum_{i=1}^r \frac{\lambda_i}{\lambda_i +
        \lambda}\scap{g_\rho}{e_i} K^{-1/2} e_i}_{L^2}
    \leq \sum_{i=1}^r \abs{g_\rho}{e_i} \norm{K^{-1/2}e_i}
    \\&\leq \norm{g_\rho}_{L^2} \paren{\sum_{i=1}^d \norm{K^{-1/2}e_i}^2_{L^2}}^{1/2}
    = c_a \norm{g_\rho}_{L^2},
  \end{align*}
  with $c_a$ defined as before.
\end{proof}

\subsubsection{Conclusion}
Based on the last subsections, we have proved the following proposition.

\begin{proposition}[Risk decomposition]
  When $\norm{(C+\lambda\mu)^{-1/2}(\hat{C} - C)(C+\lambda\mu)^{-1/2}} \leq 1/2$, Under the assumptions \ref{lap:ass:source} and \ref{lap:ass:approximation},
  \begin{equation}
    \label{lap:eq:risk_dec}
    \begin{split}
      &\norm{\hat{g}_p - g_\rho}_{L^2}^2 \leq
      4\lambda^2 \norm{{\cal L}g_\rho}_{L^2}^2
      + 4\lambda\mu c_a^2 \norm{g_\rho}_{L^2}^2
      + 8 \norm{(C+\lambda\mu)^{-1/2}(\hat\theta_\rho - \theta_\rho)}_{\cal H}^2
      \\&\qquad\cdots+ 12c_a^2\norm{C^{1/2}(I - P)}_{\op}^2\norm{g_\rho}_{L^2}^2
      + 8\norm{(C+\lambda\mu)^{-1/2} (\hat{C} - C)(C+\lambda\mu)^{-1}\theta_\rho}_{\cal H}^2.
    \end{split}
  \end{equation}
\end{proposition}

We are left with the quantification of the different convergences when the number of labeled and unlabeled data grows toward infinity.
We will quantify those convergences based on concentration inequalities.

\subsection{Probabilistic inequalities}
In this subsection, we bound each term appearing in~\eqref{lap:eq:risk_dec} based on concentration inequalities.

\subsubsection{Vector concentration}
The concentration of $\hat\theta_\rho = \widehat{S^\star g_\rho}$ toward $\theta_\rho = S^\star g_\rho$ is controlled through Bernstein inequality.

\begin{theorem}[Concentration in Hilbert space \citep{Yurinskii1970}]
  \label{lap:thm:bernstein-vector}
  Let denote by ${\cal A}$ a Hilbert space and by $(\xi_{i})$ a sequence of independent random vectors in ${\cal A}$ such that $\E[\xi_{i}] = 0$, that are bounded by a constant $M$, with finite variance $\sigma^2 = \E[\sum_{i=1}^{n}\norm{\xi_{i}}^2]$.
  For any $t>0$,
  \[
    \Pbb(
    \norm{\sum_{i=1}^{n} \xi_{i}} \geq t) \leq 2\exp\paren{-\frac{t^2}{2\sigma^2 +
        2tM / 3}}.
  \]
\end{theorem}

\begin{proposition}[Vector concentration]
  When $\abs{Y}$ is bounded by a constant $c_\Y$, and $x\to k(x, x)$ by a constant $\kappa^2$, we have, with ${\cal D}_{n_\ell}\sim\rho^{\otimes {n_\ell}}$ the dataset generating the labeled data
  \begin{equation}
    \Pbb_{{\cal D}_{n_\ell}}\paren{\norm{(C+\lambda\mu)^{-1/2}(\hat\theta_\rho - \theta_\rho)}_{\cal H} \geq t}
    \leq
    2\exp\paren{-\frac{n_\ell t^2}{2\sigma_\ell^2(\mu\lambda)^{-1} + 2t c_\Y (\mu\lambda)^{-1/2}\kappa / 3}},
  \end{equation}
  where $\sigma_\ell^2 \leq c_\Y^2 \trace(\Sigma)$ is a variance parameter to relate with the variance of $Y(I + \lambda {\cal L})^{-1} \delta_X$ (where the randomness is inherited from $(X, Y)\sim\rho$).
\end{proposition}
\begin{proof}
  Recall that
  \[
    (C+\lambda\mu)^{-1/2}(\hat\theta_\rho - \theta_\rho) = (\Sigma + \lambda L +\lambda\mu)^{-1/2} (n_\ell^{-1}\sum_{i=1}^{n_\ell}Y_i k_{X_i} - \E_{\rho}[Yk_X])
  \]
  We want to apply Bernstein inequality to the vector $\xi_{i} = (\Sigma + \lambda L + \mu \lambda)^{-1/2}Y_ik_{X_i}$, after centering it.
  Let us denote by $c_\Y$ a bound on $\abs{Y}$, $c_\Y \in \R$ exists since we have supposed $\rho$ of compact support.
  We have
  \begin{align*}
    \sigma^2 & = \E[\sum_{i=1}^{n_\ell}\norm{\xi_{i} - \E[\xi_i]}^2]
    = n_\ell \E[\norm{\xi - \E[\xi_i]}^2]
    \leq n_\ell \E[\norm{\xi}^2]
    \\&= n_\ell \E_{(X, Y)\sim\rho}\bracket{ Y^2 \scap{k_X}{(\Sigma + \lambda L + \mu \lambda)^{-1} k_{X}}}
    \\&\leq n_\ell c_\Y^2 \E_{X\sim\rho_\X}\bracket{\scap{k_X}{(\Sigma + \lambda L + \mu \lambda)^{-1} k_{X}}}
    \\&= n_\ell c_\Y^2 \trace\paren{(\Sigma + \lambda L + \mu \lambda)^{-1} \Sigma}
    \\&\leq n_\ell c_\Y^2 \trace(\Sigma) \norm{(\Sigma + \lambda L + \mu \lambda)^{-1}}_{\op}
    \leq n_\ell c_\Y^2 \trace(\Sigma) (\mu \lambda)^{-1}.
  \end{align*}
  Note that we have proceed with a generic upper bound, but we expect this variance, which is related to the variance of $Y(I + \lambda {\cal L} + \lambda \mu K^{-1})^{-1} \delta_X$ to be potentially much smaller -- if we remove the term in $P$ the vector concentration is the concentration of the vector
  $S(S^*S + \lambda S^\star {\cal L}S +
    \lambda\mu)^{-1}Yk_X \simeq
    K^{1/2}(K + \lambda K^{1/2} {\cal L}K^{1/2} +
    \lambda\mu)^{-1}K^{1/2}S^{-\star}Yk_X
    = (I + \lambda L + \lambda\mu K^{-1})^{-1}
    YS^{-\star}k_X \simeq (I + \lambda L + \lambda\mu K^{-1})^{-1}Y\delta_X$.
  To capture this fact, we will write $\sigma^2 \leq n_\ell \sigma_\ell^2(\mu\lambda)^{-1}$, with $\sigma_\ell = c_\Y \trace(\Sigma)^{1/2}$ in our analysis, but potentially much smaller under refined hypothesis and in practice.
  Similarly to the bound on the variance, we have
  \[
    \norm{\xi - \E[\xi]} \leq \norm{\xi}
    = \norm{(\Sigma + \lambda L + \mu\lambda)^{-1/2} Y_i k_{X_i}}
    \leq (\mu\lambda)^{-1/2} c_\Y \kappa,
  \]
  with $\kappa$ an upper bound on $k(x, x)^{1/2}$ for $x\in\supp\rho_\X$.
  As a consequence, applying Bernstein concentration inequality, we get, for any $t>0$,
  \[
    \Pbb_{{\cal D}_{n_\ell}}(\norm{n_\ell^{-1}\sum_{i=1}^{n_\ell} \xi_{i}- E[\xi_i]} \geq t) \leq
    2\exp\paren{-\frac{n_\ell t^2}{2\sigma_\ell^2(\mu\lambda)^{-1} + 2t c_\Y (\mu\lambda)^{-1/2}\kappa / 3}}.
  \]
  This ends the proof.
\end{proof}

\subsubsection{Operator concentration}
The convergence of $\hat{C}$ toward $C$ is controlled with Bernstein inequality for self-adjoint operators.

\begin{theorem}[Bernstein inequality for self-adjoint \citep{Minsker2017}]
  Let ${\cal A}$ be a separable Hilbert space, and $(\xi_i)$ a sequence of independent random self-adjoint operators on ${\cal A}$.
  Assume that $(\xi_i)$ are bounded by $M \in \R$, in the sense that almost everywhere $\norm{\xi}_{\op} < M$, and have a finite variance $\sigma^2 = \norm{\sum_{i=1}^n\E[\xi_i^2]}_{\op}$.
  For any $t > 0$,
  \begin{align*}
     & \Pbb\paren{\norm{\sum_{i=1}^{n}(\xi_i - \E[\xi_i])}_{\op} > t}
    \leq 2 \paren{1 + 6\ \frac{\sigma^2 + M t / 3}{t^2}} \frac{\trace\paren{\sum_{i=1}^n\E[\xi_i^2]}}{\norm{\sum_{i=1}^n\E[\xi_i^2]}_{\op}}
    \exp\paren{-\frac{t^2}{2 \sigma^2 + 2 t M / 3}}.
  \end{align*}
\end{theorem}

\begin{proposition}[Operator concentration]
  When $x\to k(x, x)$ is bounded by $\kappa^2$, and $x \to \partial_{1,j}\partial_{2,j} k(x, x)$ is bounded by $\kappa_j^2$, we have
  \begin{equation}
    \begin{split}
      &\Pbb_{{\cal D}_{n}}\paren{\norm{(C + \mu\lambda)^{-\frac{1}{2}}(C - \hat C)(C + \mu\lambda)^{-\frac{1}{2}}}_{\op} > 1/2}
      \leq \paren{2 + 56\ \frac{\kappa^2 + \lambda\sum_{i=1}^d \kappa_j^2}{\lambda\mu n}}
      \\&\qquad\cdots \times(1 + \lambda\mu \norm{C}_{\op}^{-1})
      \frac{\kappa^2 + \lambda\sum_{j=1}^d \kappa_j^2}{\lambda\mu}
      \exp\paren{-\frac{\lambda\mu n}{10\paren{\kappa^2 + \lambda\sum_{j=1}^d \kappa^2_j}}}.
    \end{split}
  \end{equation}
\end{proposition}
\begin{proof}
  We want to apply the precedent concentration inequality to
  \[
    \xi_i = (\Sigma + \lambda L +\lambda\mu)^{-1/2}(k_{X_i}\otimes k_{X_i} +
    \lambda\sum_{j=1}^d \partial_j k_{X_i}\otimes \partial_j k_{X_i})(\Sigma + \lambda L +\lambda\mu)^{-1/2},
  \]
  since we have, based on the fact that $C = \Sigma + \lambda L$ and that $\Sigma = \E[k_X\otimes k_X]$ and $L = \E[\sum_{j=1}^n\partial_j k_X \otimes \partial_j k_X]$,
  \[
    \norm{(C+\lambda\mu)^{-1/2}(\hat C - C)(C+\lambda\mu)^{-1/2}}_{\op}
    = n^{-1}\norm{\sum_{i=1}^n \xi_i - \E[\xi_i]}_{\op}.
  \]
  We bound $\xi$ with
  \begin{align*}
    \norm{\xi}_{\op} & = \norm{(C + \mu\lambda)^{-\frac{1}{2}}
      \paren{k_{X} \otimes k_{X} + \lambda\sum_{j=1}^d \partial_j k_X \otimes \partial_j k_X } (C + \mu\lambda)^{-\frac{1}{2}}}_{\op}
    \\&\leq \trace\paren{(C + \mu\lambda)^{-\frac{1}{2}} \paren{k_X \otimes k_X + \lambda\sum_{j=1}^d \partial_j k_X \otimes \partial_j k_X } (C + \mu\lambda)^{-\frac{1}{2}}}
    \\&= \trace\paren{(C + \mu\lambda)^{-\frac{1}{2}} k_X \otimes k_X (C + \mu\lambda)^{-\frac{1}{2}}}
    \\&\qquad\cdots + \lambda \sum_{j=1}^d \trace\paren{(C + \mu\lambda)^{-\frac{1}{2}} \partial_j k_X \otimes \partial_j k_X (C + \mu\lambda)^{-\frac{1}{2}}}
    \\&= \norm{(C + \mu\lambda)^{-\frac{1}{2}} k_{X}}_{\cal H}^2
    + \lambda\sum_{j=1}^d \norm{(C + \mu\lambda)^{-\frac{1}{2}} \partial_j k_X}_{\cal H}^2
    \leq (\lambda\mu)^{-1} \paren{\kappa^2 + \lambda\sum_{j=1}^d \kappa^2_j}.
  \end{align*}
  With $\kappa^2$ an upper bound on the kernel $k$ and $\kappa_j^2$ an upper bound on $\partial_{1,j}\partial_{2,j} k$.
  For the variance we have, using L\"owner order,
  \begin{align*}
    \E[\xi^2]
     & \preceq \sup_{X\in\X} \norm{\xi(X)}_{\op} \E[\xi]
    \preceq (\lambda\mu)^{-1}\paren{\kappa^2 + \lambda\sum_{j=1}^d \kappa^2_j} \E[\xi]
    \\&= (\lambda\mu)^{-1}\paren{\kappa^2 + \lambda\sum_{j=1}^d \kappa^2_j} (C + \lambda)^{-1}C
    \preceq (\lambda\mu)^{-1}\paren{\kappa^2 + \lambda\sum_{j=1}^d \kappa^2_j}.
  \end{align*}
  Therefore, we get for any $t>0$,
  \begin{align*}
     & \Pbb_{{\cal D}_{n}}\paren{\norm{(C + \mu\lambda)^{-\frac{1}{2}}(C - \hat C)(C + \mu\lambda)^{-\frac{1}{2}}}_{\op} > t}
    \\&\qquad\leq 2\paren{1 + 6\ \frac{(\kappa^2 + \lambda \sum_{i=1}^d \kappa_j^2)(1 + t / 3)}{\lambda\mu n t^2}}\, \frac{\norm{C}_{\op} + \lambda\mu}{\norm{C}_{\op}}
    \trace\paren{(C + \lambda)^{-1}C}
    \\&\qquad\cdots\times\exp\paren{-\frac{nt^2}{2 (\lambda\mu)^{-1}\paren{\kappa^2 + \lambda\sum_{j=1}^d \kappa^2_j} (1 + t/3)}}.
  \end{align*}
  Remark that
  \[
    \trace\paren{(C + \mu\lambda)^{-1}C} \leq \norm{(C +
      \mu\lambda)^{-1}}_{\op}\trace(C) \leq
    (\lambda\mu)^{-1}(\kappa^2 + \lambda \sum_{j=1}^d \kappa_j^2).
  \]
  Taking $t = 1/2$ ends the lemma.
\end{proof}

\subsubsection{Basis concentration}
Similarly, we could control $\norm{(C+\lambda\mu)^{-1/2}(\hat C - C)(C+\lambda\mu)^{-1}\theta_\rho}_{\cal H}$ by using concentration of self-adjoint, yet this will lead to laxer bounds, than using concentration on vectors.

\begin{proposition}[Basis concentration]
  When $x\to k(x, x)$ is bounded by $\kappa^2$, $x \to \partial_{1,j}\partial_{2,j} k(x, x)$ is bounded by $\kappa_j^2$, with Assumptions \ref{lap:ass:source} and \ref{lap:ass:approximation}, we have
  \begin{equation}
    \Pbb_{{\cal D}_{n}}\paren{\norm{(C + \mu\lambda)^{-\frac{1}{2}}(C - \hat C)(C + \mu\lambda)^{-1}\theta_\rho}_{\cal H} > t}
    \leq 2\exp\paren{
      - \frac{\mu\lambda nt^2}{2 c_1(c_1 + \lambda^{1/2}\mu^{1/2} t / 3)}
    },
  \end{equation}
  with $c_1 = (\kappa^2 + \lambda\sum_{i=1}^d \kappa_j^2) c_a \norm{g_\rho}_{L^2}$.
\end{proposition}
\begin{proof}
  We want to apply Bernstein concentration inequality to the vectors
  \[
    \xi_i = (C+\mu\lambda)^{-1/2} \paren{k_{X_i}\otimes k_{X_i} + \lambda \sum_{j=1}^d
      \partial_j k_{X_i} \otimes \partial_j k_{X_i}} (C+\lambda\mu)^{-1}\theta_{\rho},
  \]
  since
  \[
    \norm{(C + \mu\lambda)^{-\frac{1}{2}}(C - \hat C)(C + \mu\lambda)^{-1}\theta_\rho}_{\cal H}
    = n^{-1} \norm{\sum_{i=1}^n \xi_i - \E[\xi_i]}_{\cal H}.
  \]
  We bound $\xi$, reusing prior derivations, with
  \begin{align*}
    \norm{\xi_i}_{\cal H}
     & = \norm{(C+\mu\lambda)^{-1/2} \paren{k_{X_i}\otimes k_{X_i} + \lambda \sum_{j=1}^d
        \partial_j k_{X_i} \otimes \partial_j k_{X_i}} (C+\lambda\mu)^{-1}\theta_{\rho}}_{\cal H}
    \\&\leq \norm{(C+\mu\lambda)^{-1/2}}_{\op} \norm{\paren{k_{X_i}\otimes k_{X_i} + \lambda \sum_{j=1}^d
        \partial_j k_{X_i} \otimes \partial_j k_{X_i}}}_{\op} \norm{(C+\mu\lambda)^{-1}\theta_\rho}_{\cal H}.
    \\&\leq (\mu\lambda)^{-1/2} (\kappa^2 + \lambda\sum_{i=1}^d \kappa_j^2) c_a \norm{g_\rho}_{L^2}.
  \end{align*}
  For the variance, we have, similarly to prior derivations,
  \begin{align*}
    \E[\norm{\xi}^2]
     & \leq
    \sup_{X\in\X}\norm{k_{X}\otimes k_{X} + \lambda \sum_{j=1}^d
      \partial_j k_{X} \otimes \partial_j k_{X}}_{\op}\norm{(C+\lambda\mu)^{-1}\theta_{\rho}}^2
    \\&\qquad\qquad\cdots \times \E\bracket{ \norm{(C+\mu\lambda)^{-1} \paren{k_{X}\otimes k_{X} + \lambda \sum_{j=1}^d
          \partial_j k_{X_i} \otimes \partial_j k_{X}}_{\op}}}
    \\&\leq
    \paren{\kappa^2 + \lambda\sum_{i=1}^d \kappa_i^2} c_a^2 \norm{g_\rho}^2_{L^2}
    \\&\qquad\qquad\cdots\E\bracket{\norm{(C+\mu\lambda)^{-1} k_{X}\otimes k_{X}}_{\op} + \lambda \sum_{j=1}^d
      \norm{(C+\mu\lambda)^{-1}\partial_j k_{X_i} \otimes \partial_j k_{X}}_{\op}}
    \\&=
    \paren{\kappa^2 + \lambda\sum_{i=1}^d \kappa_i^2} c_a^2 \norm{g_\rho}^2_{L^2}
    \trace\paren{(C+\mu\lambda)^{-1} C }
    \\&\leq (\lambda\mu)^{-1}\paren{\kappa^2 + \lambda\sum_{i=1}^d \kappa_i^2}^2 c_a^2 \norm{g_\rho}^2_{L^2}.
  \end{align*}
  As a consequence, using Bernstein inequality,
  \[
    \Pbb\paren{n^{-1}\norm{\sum_{i=1}^n \xi_{i} - \E[\xi_i]} > t}
    \leq 2\exp\paren{
      - \frac{\mu\lambda nt^2}{2 c_1(c_1 + \lambda^{1/2}\mu^{1/2} t / 3)},
    }
  \]
  with $c_1 = (\kappa^2 + \lambda\sum_{i=1}^d \kappa_j^2) c_a \norm{g_\rho}_{L^2}$.
  Note that we have bounded naively the variable $\xi$ and its variance, but we have shown how appears $\sup_{X\in\X} \norm{(C+\lambda\mu)^{-1}k_X} + \lambda\sum_{i=1}^d \norm{(C+\lambda\mu)^{-1}\partial_i k_X}$ and $\trace((C+\lambda\mu)^{-1}C)$, which under interpolation and capacity assumptions could be controlled in a better fashion.
\end{proof}

\subsubsection{Low-rank approximation}
We now switch to Nystr\"om approximation.

\begin{proposition}[Low-rank approximation]
  When $x\to k(x, x)$ is bounded by $\kappa^2$, for any $p \in \N$ and $t > 0$, we have
  \begin{equation*}
    \Pbb_{{\cal D}_p}\paren{\norm{(I-P)\Sigma^{1/2}}^2 > t}
    \leq \paren{2 + \frac{116\kappa^2}{t p}} (2 + t \norm{\Sigma}^{-1}_{\op}) \frac{\kappa^2}{t}
    \exp\paren{- \frac{p t }{10\kappa^2}},
  \end{equation*}
\end{proposition}
\begin{proof}
  Reusing Proposition 3 of \citet{Rudi2015}, for any $\gamma > 0$, we have, with $P$ the projection on $\Span\brace{k_{X_i}}_{i\leq p}$ and $\hat\Sigma = p^{-1}\sum_{i=1}^p k_{X_i} \otimes k_{X_i}$,
  \[
    \norm{(I-P)\Sigma^{1/2}}^2
    \leq \gamma\norm{(\hat\Sigma + \gamma)^{-1/2}\Sigma^{1/2}}^2_{\op}
    \leq \gamma \norm{\Sigma^{1/2} (\hat\Sigma + \gamma)^{-1} \Sigma^{1/2}}_{\op}.
  \]
  As a consequence, skipping derivations that can be retaken from our precedent proofs,
  \begin{align*}
     & \Pbb_{{\cal D}_p}\paren{\norm{(I-P)\Sigma^{1/2}}^2 > t}
    \leq \inf_{\gamma > 0} \Pbb_{{\cal D}_p}
    \paren{ \gamma \norm{\Sigma^{1/2} (\hat\Sigma + \gamma)^{-1} \Sigma^{1/2}}_{\op} > t}
    \\&\qquad\leq \inf_{\gamma > 0} \Pbb_{{\cal D}_p}
    \paren{\norm{(\Sigma + \gamma)^{-1/2}(\hat\Sigma - \Sigma)
        (\Sigma + \gamma)^{-1/2}}_{\op} > (1 - \gamma t^{-1})}
    \\&\qquad\leq \inf_{\gamma > 0}
    \paren{2 + 56\ \frac{\kappa^2}{\gamma p}} (1 + \gamma \norm{\Sigma}^{-1}_{\op}) frac{\kappa^2}{\gamma}
    \exp\paren{- \frac{p\gamma u^2}{2\kappa^2 (1 + u/ 3)}},
  \end{align*}
  with $u = (1 - \gamma t^{-1})$.
  Taking $\gamma = t/2$, this term is simplified as
  \[
    \Pbb_{{\cal D}_p}\paren{\norm{(I-P)\Sigma^{1/2}}^2 > t}
    \leq \paren{2 + 116\ \frac{\kappa^2}{t p}} (2 + t \norm{\Sigma}^{-1}_{\op}) \frac{\kappa^2}{t}
    \exp\paren{- \frac{p t }{10\kappa^2}},
  \]
  which is the object of this proposition.
\end{proof}

\begin{lemma}
  When $L \leq c_d \Sigma^a$, we have
  \begin{equation}
    \norm{(I-P)C^{1/2}}_{\op}^2 \leq \norm{(I-P)\Sigma^{1/2}}^2_{\op}
    + c_d\lambda\norm{(I-P)\Sigma^{1/2}}^{2a}_{\op}.
  \end{equation}
\end{lemma}
\begin{proof}
  This follows from the fact that
  \begin{align*}
    \norm{C^{1/2}(I - P)}^2_{\op} & = \norm{(I-P) C (I-P)}_{\op}
    = \norm{(I-P) (\Sigma + \lambda L)(I-P)}_{\op}
    \\&\leq \norm{(I-P)\Sigma(I-P)}_{\op}
    + \lambda \norm{(I-P)L(I-P)}_{\op}
    \\&\leq \norm{(I-P)\Sigma(I-P)}_{\op}
    + \lambda c_d \norm{(I-P)\Sigma^a(I-P)}_{\op}
    \\&= \norm{(I-P)\Sigma^{1/2}}_{\op}^2
    + \lambda c_d \norm{(I-P)\Sigma^{a/2}}_{\op}^2
    \\&= \norm{(I-P)\Sigma^{1/2}}_{\op}^2
    + \lambda c_d \norm{(I-P)^{a}\Sigma^{a/2}}_{\op}^2
    \\&\leq \norm{(I-P)\Sigma^{1/2}}_{\op}^2
    + \lambda c_d \norm{(I-P)\Sigma^{1/2}}_{\op}^{2a},
  \end{align*}
  where we used the fact that $(I - P)^a = (I-P)$ and that $\norm{A^s B^s} \leq \norm{AB}^s$ for $s \in [0, 1]$ and $A, B$ positive self-adjoint.
\end{proof}

\subsection{Averaged excess of risk - ending the proof}

Based on the precedent excess of risk decomposition, and precedent concentration inequalities, we have all the elements to derive convergence rates of our algorithm.
We will enunciate this convergence in terms of the averaged excess of risk of $\E_{{\cal D}_n}\bracket{\norm{\hat{g}_p - g_\rho}_{L^2}^2}$.

\begin{lemma}
  Under Assumptions \ref{lap:ass:source} and \ref{lap:ass:approximation},
  \begin{equation}
    \begin{split}
      &\E_{{\cal D}_n}\bracket{\norm{\hat{g}_p - g_\rho}_{L^2}^2}
      \leq 4c_\Y^2\Pbb\paren{\norm{(C+\lambda\mu)^{-1/2}(\hat{C} - C)(C+\lambda\mu)^{-1/2}} \leq 1/2}
      \\&\qquad\cdots + 4\lambda^2 \norm{{\cal L}g_\rho}_{L^2}^2
      + 4\lambda\mu c_a^2 \norm{g_\rho}_{L^2}^2
      \\&\qquad\cdots + 8 \E_{{\cal D}_n}\bracket{ \norm{(C+\lambda\mu)^{-1/2}(\hat\theta_\rho - \theta_\rho)}_{\cal H}^2}
      + 12c_a^2\norm{g_\rho}_{L^2}^2 \E_{{\cal D}_n}\bracket{ \norm{C^{1/2}(I - P)}_{\op}^2}
      \\&\qquad\cdots+ 8\E_{{\cal D}_n}\bracket{\norm{(C+\lambda\mu)^{-1/2} (\hat{C} - C)(C+\lambda\mu)^{-1}\theta_\rho}_{\cal H}^2}.
    \end{split}
  \end{equation}
\end{lemma}
\begin{proof}
  We proceed using the fact that $\E[X] = \E[X\,\vert\, ^cA]\Pbb(^cA) + \E[X\,\vert\, A]\Pbb(A) \leq \sup X \Pbb(^cA) + \E[X\, \vert\, A]\Pbb(A)$, with $A = \brace{{\cal D}_n \midvert \norm{(C+\lambda\mu)^{-1/2}(\hat{C} - C)(C+\lambda\mu)^{-1/2}} \leq 1/2}$,
  \[
    \E_{{\cal D}_n}\bracket{\norm{\hat{g}_p - g_\rho}_{L^2}^2}
    \leq \sup_{{\cal D}_n} \norm{\hat{g}_p - g_\rho}^2\Pbb\paren{^c A}
    + \E_{{\cal D}_n}\bracket{\norm{\hat{g}_p - g_\rho}^2 \midvert A}\Pbb(A).
  \]
  When $Y$ is bounded by $c_\Y$, because $g_\rho$ is a convex combination of $Y$, we know that $\norm{g_\rho}_{L^2} \leq c_\Y$, as a consequence, we can clip $\hat{g}_p$ to $[-c_\Y, c_\Y]$, which will only improve the estimation of $g_\rho$, as a consequence, we can consider the clipping estimate for which we have
  \(
  \sup_{{\cal D}_n} \norm{\hat{g}_p - g_\rho}_{L^2}^2 \leq 4 c_\Y^2.
  \)
  Regarding the second part, we have already decomposed the risk under the event $A = \brace{{\cal D}_n \midvert \norm{(C+\lambda\mu)^{-1/2}(\hat{C} - C)(C+\lambda\mu)^{-1/2}} \leq 1/2}$.
  As a consequence, we have
  \begin{align*}
     & \E_{{\cal D}_n}\bracket{\norm{\hat{g}_p - g_\rho}_{L^2}^2}
    \leq 4c_\Y^2\Pbb(^cA)
    + 4\lambda^2 \norm{{\cal L}g_\rho}_{L^2}^2 \Pbb(A)
    + 4\lambda\mu c_a^2 \norm{g_\rho}_{L^2}^2 \Pbb(A)
    \\&\cdots + 8 \E_{{\cal D}_n}\bracket{ \norm{(C+\lambda\mu)^{-1/2}(\hat\theta_\rho - \theta_\rho)}_{\cal H}^2 \midvert A}\Pbb(A)
    \\&\cdots+ 12c_a^2\norm{g_\rho}_{L^2}^2 \E_{{\cal D}_n}\bracket{ \norm{C^{1/2}(I - P)}_{\op}^2 \midvert A}\Pbb(A)
    \\&\cdots+ 8\E_{{\cal D}_n}\bracket{\norm{(C+\lambda\mu)^{-1/2} (\hat{C} - C)(C+\lambda\mu)^{-1}\theta_\rho}_{\cal H}^2 \midvert A}\Pbb(A).
  \end{align*}
  To control the conditional expectation, we use that, when $X$ is positive
  \[
    \E\bracket{X\midvert A} P(A) = \E[X] - \E\bracket{X \midvert ^cA}\Pbb(^cA) \leq \E[X].
  \]
  This ends the proof.
\end{proof}

Based on deviation inequalities, we can control expectations based on the equality, for $X$ positive, $\E[X] = \int_0^{+\infty} \Pbb(X > t) \diff t$.

\begin{lemma}
  In the setting of the paper,
  \begin{equation}
    \E_{{\cal D}_n}\bracket{ \norm{(C+\lambda\mu)^{-1/2}(\hat\theta_\rho - \theta_\rho)}_{\cal H}^2}
    \leq 8 \sigma_\ell^2 (n_\ell \mu\lambda)^{-1} +
    8c_\Y^2\kappa^2(n_\ell^{2} \mu\lambda)^{-1}.
  \end{equation}
\end{lemma}
\begin{proof}
  First, recall that
  \begin{align*}
    \Pbb\paren{ \norm{(C+\lambda\mu)^{-1/2}(\hat\theta_\rho - \theta_\rho)}_{\cal H} > t}
     & \leq 2\exp\paren{-\frac{n_\ell t^2}{2\sigma_\ell^2 (\mu\lambda)^{-1} + 2 t c_\Y(\lambda\mu)^{-1/2} \kappa / 3}}
    \\&
    \leq 2\exp\paren{-\frac{n_\ell t^2}{2\max\paren{2\sigma_\ell^2 (\mu\lambda)^{-1}, 2 t c_\Y(\lambda\mu)^{-1/2} \kappa / 3}}}
    \\&
    \leq 2\exp\paren{-\frac{n_\ell \mu\lambda t^2}{4\sigma_\ell^2}}
    + 2\exp\paren{-\frac{3n_\ell \mu^{1/2}\lambda^{1/2} t}{4 c_\Y \kappa}}.
  \end{align*}
  As a consequence
  \begin{align*}
     & \E\bracket{ \norm{(C+\lambda\mu)^{-1/2}(\hat\theta_\rho - \theta_\rho)}_{\cal H}^2}
    =
    \int_0^{+\infty}\Pbb\paren{ \norm{(C+\lambda\mu)^{-1/2}(\hat\theta_\rho - \theta_\rho)}_{\cal H}^2 > t}\diff t
    \\&\qquad\leq 2 \int \exp\paren{-\frac{n_\ell \mu\lambda t}{4\sigma_\ell^2}}\diff t
    + 2\int \exp\paren{-\frac{3n_\ell \mu^{1/2}\lambda^{1/2} t^{1/2}}{4 c_\Y \kappa}}\diff t.
    \\&\qquad = 8 \sigma_\ell^2 (n_\ell \mu\lambda)^{-1}
    + \frac{64c_\Y^2\kappa^2}{9}(n_\ell^{2} \mu\lambda)^{-1}.
  \end{align*}
  This is the result stated in the lemma.
\end{proof}

\begin{lemma}
  In the setting of the paper,
  \begin{equation}
    \begin{split}
      \E_{{\cal D}_n}\bracket{\norm{(C+\lambda\mu)^{-1/2} (\hat{C} - C)(C+\lambda\mu)^{-1}\theta_\rho}_{\cal H}^2}
      &\leq 8(\kappa^2 + \lambda\partial \kappa^2)^2 c_a^2 \norm{g_\rho}_{L^2}^2
      \\&\qquad\cdots\times\paren{(\mu\lambda n)^{-1} + (\mu \lambda n^2)^{-1}},
    \end{split}
  \end{equation}
  with $\partial \kappa^2 = \sum_{i=1}^d \kappa_i^2$.
\end{lemma}
\begin{proof}
  Let us denote by $A$ the quantity $\norm{(C+\lambda\mu)^{-1/2} (\hat{C} -
      C)(C+\lambda\mu)^{-1}\theta_\rho}_{\cal H}$, and
  $\partial \kappa^2 = \sum_{i=1}^d \kappa_j^2$.
  Recall that
  \begin{align*}
    \Pbb\paren{A > t}
     & \leq 2\exp\paren{-\frac{\mu\lambda n t^2}{2c_1(c_1 + \lambda^{1/2}\mu^{1/2} t/3)}}
    \\&\leq 2\exp\paren{-\frac{\mu\lambda n t^2}{4c_1^2}}
    + 2\exp\paren{-\frac{3(\mu\lambda)^{1/2} n t}{4c_1}}.
  \end{align*}
  We conclude the proof similarly to the precedent lemma.
\end{proof}

\begin{lemma}
  Under Assumption \ref{lap:ass:decay},
  \begin{equation}
    \begin{split}
      \E_{{\cal D}_n}\bracket{\norm{C^{1/2}(I - P)}_{\op}^2}
      &\leq
      \paren{\frac{10\kappa^2\log(p)}{p} + \frac{10^a\kappa^{2a} c_d \lambda \log(p)^a}{p^a}}
      \\&\cdots \times \paren{1 + \frac{2\kappa^2}{\norm{\Sigma}_{\op}\log(p)} \paren{1 +
          \frac{6}{\log(p)}} \paren{\frac{1}{p} + \frac{1}{5\log(p)}}}.
    \end{split}
  \end{equation}
\end{lemma}
\begin{proof}
  Once again, this result comes from integration of the tail bound obtained on $\norm{C^{1/2}(I-P)}_{\op}^2$ through the one we have on $\norm{\Sigma^{1/2}(I - P)}_{\op}^2$ and the fact that $\norm{C^{1/2}(I - P)}_{\op}^2 \leq \norm{\Sigma^{1/2}(I - P)}^2_{\op} + c_d\lambda \norm{\Sigma^{1/2}(I - P)}^{2a}_{\op}$.
  For any $a, b > 0$, we have
  \begin{align*}
     & \E_{{\cal D}_n}\bracket{\norm{\Sigma^{1/2}(I - P)}_{\op}^2}
    = \int_0^\infty \Pbb_{{\cal D}_n}\paren{\norm{\Sigma^{1/2}(I - P)}_{\op}^2 > t} \diff t
    \\&\qquad\leq \int_0^\infty \min\brace{1, 2\kappa^2 \norm{\Sigma}^{-1}_{\op} \paren{1 + \frac{58\kappa^2}{t p}} \paren{1 + \frac{2\kappa^2}{t}}
      \exp\paren{- \frac{p t }{10\kappa^2}}} \diff t
    \\&\qquad =
    \frac{10\kappa^2 a}{p} \int_0^\infty \min\brace{1, 2\kappa^2 \norm{\Sigma}^{-1}_{\op} \paren{1 +
        \frac{58}{10 au}} \paren{1 + \frac{p}{5 au}}
      \exp\paren{- au }} \diff u
    \\&\qquad \leq
    \frac{10\kappa^2 a}{p} \paren{b + \int_b^\infty 2\kappa^2 \norm{\Sigma}^{-1}_{\op} \paren{1 +
        \frac{6}{au}} \paren{1 + \frac{p}{5 au}}
      \exp\paren{- au } \diff u}
    \\&\qquad \leq
    \frac{10\kappa^2}{p} \paren{ab + 2\kappa^2 \norm{\Sigma}^{-1}_{\op} \paren{1 +
        \frac{6}{ab}} \paren{1 + \frac{p}{5 ab}}
      \exp\paren{- ab }}.
  \end{align*}
  This last quantity is optimized for $ab = \log(p)$, which leads to the first part of the lemma.
  Similarly,
  \begin{align*}
     & \E_{{\cal D}_n}\bracket{\norm{\Sigma^{1/2}(I - P)}_{\op}^{2a}}
    = \int_0^\infty \Pbb_{{\cal D}_n}\paren{\norm{\Sigma^{1/2}(I - P)}_{\op}^{2a} > t} \diff t
    \\&\qquad = \int_0^\infty \Pbb_{{\cal D}_n}\paren{\norm{\Sigma^{1/2}(I - P)}_{\op}^{2} > t^{1/a}} \diff t
    \\&\qquad\leq \int_0^\infty \min\brace{1, 2\kappa^2 \norm{\Sigma}^{-1}_{\op} \paren{1 + \frac{58\kappa^2}{t^{1/a} p}} \paren{1 + \frac{2\kappa^2}{t^{1/a}}}
      \exp\paren{- \frac{p t^{1/a} }{10\kappa^2}}} \diff t
    \\&\qquad =
    \frac{10^a\kappa^{2a} a c^a}{p^a} \int_0^\infty \min\brace{u^{a-1}, 2\kappa^2 \norm{\Sigma}^{-1}_{\op} \paren{1 +
        \frac{58}{10 cu}} \paren{1 + \frac{p}{5 cu}}
      \frac{1}{u^{1-a}}\exp\paren{- cu }} \diff u
    \\&\qquad \leq
    \frac{10^a\kappa^{2a} a c^a}{p^a} \paren{\frac{b^a}{a} + \int_b^\infty 2\kappa^2 \norm{\Sigma}^{-1}_{\op} \paren{1 +
        \frac{6}{cu}} \paren{1 + \frac{p}{5 cu}}
      \frac{1}{u^{1-a}}\exp\paren{- cu } \diff u}
    \\&\qquad \leq
    \frac{10^a\kappa^{2a}}{p^a} \paren{(cb)^a + 2\kappa^2 \norm{\Sigma}^{-1}_{\op} \paren{1 +
        \frac{6}{cb}} \paren{1 + \frac{p}{5 cb}}
      \frac{1}{(cb)^{1-a}}\exp\paren{- cb }}.
  \end{align*}
  Once again this is optimized for $cb = \log(p)$.
\end{proof}

\begin{remark}[Leverage scores]
  Out of simplicity, we only present a low rank approximation with random subsampling.
  Yet, we can improve the result by considering subsampling based on leverage scores.
  If we consider the Gaussian kernel, $Sk_x\in L^2$ can be thought of as a function that is a little bump around $x\in\X$.
  In essence, subsampling based on leverage scores, consists in representing the solution on a subsampled sequence $(k_{X_i})_{i\in I}$ where the $X_i$ are far from one another so that the bump functions $(Sk_{X_i})$ can approximate a maximum of functions.
  \citep{Rudi2015} shows that with leverage scores, we can take $p = (\mu\lambda)^{\gamma} \log(n)$, with $\gamma$ linked with the capacity of the RKHS linked with the kernel $k$.
\end{remark}

If we add all derivations, we have derived the following theorem.

\begin{theorem}
  Under Assumptions \ref{lap:ass:source}, \ref{lap:ass:approximation} and \ref{lap:ass:decay},
  \begin{equation}
    \label{lap:eq:pre_final}
    \begin{split}
      &\E_{{\cal D}_n}\bracket{\norm{\hat{g} - g_\rho}_{L^2}^2}
      \\&\qquad\leq 8c_\Y^2 \paren{1 + 28\ \frac{\kappa^2 + \lambda \partial\kappa^2}{\lambda\mu n}} (1 + \lambda\mu \norm{C}_{\op}^{-1})
      \frac{\kappa^2 + \lambda\partial\kappa^2}{\lambda\mu}
      \exp\paren{-\frac{\lambda\mu n}{10\paren{\kappa^2 + \lambda\partial\kappa^2}}}
      \\&\qquad\cdots + 4\lambda^2 \norm{{\cal L}g_\rho}_{L^2}^2
      + 4\lambda\mu c_a^2 \norm{g_\rho}_{L^2}^2
      + 64 \sigma_\ell^2 (n_\ell \mu\lambda)^{-1} +
      57 c_\Y^2\kappa^2(n_\ell^{2} \mu\lambda)^{-1}
      \\&\qquad\cdots + 64(\kappa^2 + \lambda\partial \kappa^2)^2 c_a^2 \norm{g_\rho}_{L^2}^2 (\mu\lambda n)^{-1}
      + 57 (\kappa^2 + \lambda\partial \kappa^2)^2 c_a^2 \norm{g_\rho}_{L^2}^2 (\mu\lambda n^2)^{-1}
      \\&\qquad\cdots +
      12 c_a^2 \norm{g_\rho}^2_{L^2}\paren{\frac{10\kappa^2\log(p)}{p} + \frac{10^a\kappa^{2a} c_d \lambda \log(p)^a}{p^a}}
      \\&\qquad\qquad\qquad\cdots \times
      \paren{1 + \frac{2\kappa^2}{\norm{\Sigma}_{\op}\log(p)} \paren{1 +
          \frac{6}{\log(p)}} \paren{\frac{1}{p} + \frac{1}{5\log(p)}}}.
    \end{split}
  \end{equation}
  where $c_\Y$ is an upper bound on $Y$, $\kappa^2$ is an upper bound on $x \to k(x, x)$, $\partial\kappa^2 = \sum_{i=1}^d \kappa_i^2$ with $\kappa_i^2$ a bound on $x \to \partial_{1i}\partial_{2i} \partial k_{x_i}$, $c_d$ and $a$ the constants appearing in Assumption \ref{lap:ass:decay}, $c_a$ a constant such that $\norm{g}_{\cal H} \leq c_a \norm{g}_{L^2}$ and $\sigma_\ell^2 \leq c_\Y^2 \kappa^2$ a variance parameter linked with the variance of $Y(I + \lambda{\cal L})^{-1}\delta_X$.
\end{theorem}

Theorem \ref{lap:thm:consistency} is a corollary of this theorem.

\end{subappendices}

\part{Active Labeling}
\label{part:collection}
\chapter{Streaming Stochastic Gradients}
\label{chap:sgd}

The following is a reproduction of \cite{Cabannes2022}.

The workhorse of machine learning is stochastic gradient descent.
To access stochastic gradients, it is common to consider iteratively input/output pairs of a training dataset.
Interestingly, it appears that one does not need full supervision to access stochastic gradients, which is the main motivation of this paper.
After formalizing the ``active labeling'' problem, which generalizes active learning based on partial supervision, we provide a streaming technique that provably minimizes the ratio of generalization error over the number of samples.
We illustrate our technique in depth for robust regression.
\section{Introduction}

A large amount of the current hype around artificial intelligence was fueled by the recent successes of supervised learning.
Supervised learning consists in designing an algorithm that maps inputs to outputs by learning from a set of input/output examples.
When accessing many samples, and given enough computation power, this framework is able to tackle complex tasks.
Interestingly, many of the difficulties arising in practice do not emerge from choosing the right statistical model to solve the supervised learning problem, but from the problem of collecting and cleaning enough data \cite[see Chapters 1 and 2 of][for example]{Geron2017}.
Those difficulties are not disjoint from the current trends toward data privacy regulations \citep{GDPR}.
This fact motivates this work, where we focus on how to efficiently collect information to carry out the learning process.

In this paper, we formalize the ``active labeling'' problem for weak supervision, where the goal is to learn a target function by acquiring the most informative dataset given a restricted budget for annotation.
We focus explicitly on weak supervision that comes as a set of label candidates for each input, aiming to partially supervise input data in the most efficient way to guide a learning algorithm.
We also restrict our study to the streaming variant where, for each input, only a single partial information can be collected about its corresponding output.
The crux of this work is to leverage the fact that full supervision is not needed to acquire unbiased stochastic gradients, and perform stochastic gradient descent.

The following summarizes our contributions.
\begin{enumerate}
  \item First, we introduce the ``active labeling'' problem, which is a relevant theoretical framework that encompasses many useful problems encountered by practitioners trying to annotate their data in the most efficient fashion, as well as its streaming variation, in order to deal with privacy preserving issues. This is the focus of Section~\ref{sgd:sec:fram}.
  \item Then, in Section~\ref{sgd:sec:sgd}, we give a high-level framework to access unbiased stochastic gradients with weak information only. This provides a simple solution to the streaming ``active labeling'' problem.
  \item Finally, we detail this framework for a robust regression task in Section~\ref{sgd:sec:median}, and provide an algorithm whose optimality is proved in Section~\ref{sgd:sec:stat}.
\end{enumerate}
As a proof of concept, we provide numerical simulations in Section~\ref{sgd:sec:exp}.
We conclude with a high-level discussion around our methods in Section~\ref{sgd:sec:discussion}.

\paragraph{Related work.}
Active query of information is relevant to many settings.
The most straightforward applications are searching games, such as Bar Kokhba or twenty questions \citep{Walsorth1882}.
We refer to \citet{Pelc2002} for an in-depth survey of such games, especially when liars introduce uncertainty, and their relations with coding on noisy channels.
But applications are much more diverse, {\em e.g.} for numerical simulation \citep{Chevalier2014}, database search \citep{Qarabaqi2014}, or shape recognition \citep{Geman1993}, to name a few.

In terms of motivations, many streams of research can be related to this problem, such as experimental design \citep{Chernoff1959}, statistical queries \citep{Kearns1998,Fotakis2021}, crowdsourcing \citep{Doan2011}, or aggregation methods in weak supervision \citep{Ratner2020}.
More precisely, ``active labeling''\footnote{Note that the wording ``active labeling'' has been more or less used as synonymous of ``active learning'' \cite[\emph{e.g.,}][]{Wang2014}. In contrast, we use ``active labeling'' to design ``active weakly supervised learning''.} consists in having several inputs and querying partial information on the labels. It is close to active learning \citep{Settles2010,Dasgupta2011,Hanneke2014}, where there are several inputs, but exact outputs are queried; and to active ranking \citep{Valiant1975,Ailon2011,Braverman2019}, where partial information is queried, but there is only one input.
The streaming variant introduces privacy preserving constraints, a problem that is usually tackled through the notion of differential privacy \citep{Dwork2006}.

In terms of formalization, we build on the partial supervision formalization of \cite{Cabannes2020}, which casts weak supervision as sets of label candidates and generalizes semi-supervised learning \citep{Chapelle2006}.
Finally, our sequential setting with a unique final reward is similar to combinatorial bandits in a pure-exploration setting \citep{Garivier2016,Fiez2019}.

\section{The ``active labeling'' problem}
\label{sgd:sec:fram}

Supervised learning is traditionally modeled in the following manner.
Consider $\X$ an input space, $\Y$ an output space, $\ell:\Y\times\Y\to\R$ a loss function, and $\rho\in\prob{\X\times\Y}$ a joint probability distribution.
The goal is to recover the function
\begin{equation}
  \label{sgd:eq:obj}
  f^*\in\argmin_{f:\X\to\Y} {\cal R}(f) := \E_{(X, Y)\sim\rho}[\ell(f(X), Y)],
\end{equation}
yet, without accessing $\rho$, but a dataset of independent samples distributed according to $\rho$, ${\cal D}_n = (X_i, Y_i)_{i\leq n}\sim\rho^{\otimes n}$.
In practice, accessing data comes at a cost, and it is valuable to understand the cheapest way to collect a dataset allowing to discriminate $f^*$.

We shall suppose that the input data $(X_i)_{i\leq n}$ are easy to collect, yet that labeling those inputs to get outputs $(Y_i)_{i\leq n}$ demands a high amount of work.
For example, it is relatively easy to scrap the web or medical databases to access radiography images, but labeling them by asking radiologists to recognize tumors on zillions of radiographs will be both time-consuming and expensive.
As a consequence, \emph{we assume the $(X_i)_{i\leq n}$ given but the $(Y_i)_{i\leq n}$ unknown}.
As getting information on the labels comes at a cost ({\em e.g.}, paying a pool of label workers, or spending your own time), given a budget constraint, what information should we query on the labels?

To quantify this problem, we will assume that {\em we can sequentially and adaptively query $T$ information of the type $\ind{Y_{i_t} \in S_t}$, for any index $i_t\in\brace{1, \cdots, n}$ and any set of labels $S_t \subset \Y$ (belonging to a specified set of subsets of $\Y$).}
Here, $t \in \brace{1, \cdots, T}$ indexes the query sequence, and $T\in\N$ is a fixed budget.
The goal is to optimize the design of the sequence $(i_t, S_t)$ in order to get the best estimate of $f^*$ in terms of risk minimization~\eqref{sgd:eq:obj}.
In the following, we give some examples to make this setting more concrete.

\begin{figure}[ht]
  \centering
  \includegraphics[width=.3\textwidth]{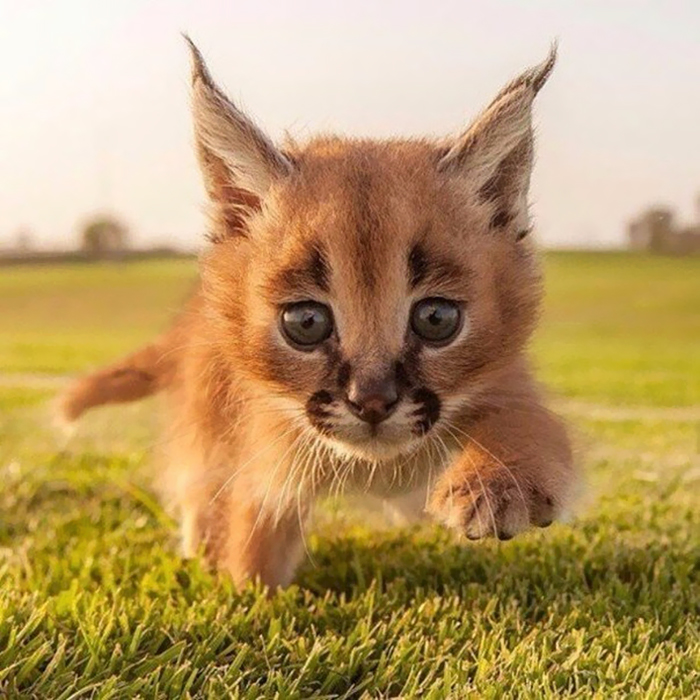}
  \caption{Recognizing fine-grained classes is difficult, but recognizing attributes is easy.}
  \label{sgd:fig:caracal}
\end{figure}

\begin{example}[Classification with attributes]
  Suppose that a labeler is asked to provide fine-grained classes on images \citep{Krause2016,Zheng2019}, such as the label ``caracal'' in Figure~\ref{sgd:fig:caracal}.
  This would be difficult for many people. Yet, it is relatively easy to recognize that the image depicts a ``feline'' with ``tufted-ears'' and ``sandy color''.
  As such, a labeler can give the weak information that $Y$ belongs to the set ``feline'', $S_1 = \brace{\text{``cat'', ``lion'', ``tiger''}, \dots}$, and the set ``tufted ears'', $S_2 = \brace{\text{``Great horned owl'', ``Aruacana chicken''}, \dots}$.
  This is enough to recognize that $Y \in S_1 \cap S_2 = \brace{\text{``caracal''}}$.
  The question $\ind{Y\in S_1}$, corresponds to asking if the image depicts a feline.
  Literature on hierarchical classification and autonomic taxonomy construction provides interesting ideas for this problem \citep[{\em e.g.},][]{Cesa-Bianchi2006,Gangaputra2006}.
\end{example}

\begin{example}[Ranking with partial ordering]
  Consider a problem where for a given input $x$, characterizing a user, we are asked to deduce their preferences over $m$ items.
  Collecting such a label requires knowing the exact ordering of the $m$ items induced by a user. This might be hard to ask for.
  Instead, one can easily ask the user which items they prefer in a collection of a few items. The user's answer will give weak information about the labels, which can be modeled as knowing $\ind{Y_i\in S} = 1$, for $S$ the set of total orderings that satisfy this partial ordering.
  We refer the curious reader to active ranking and dueling bandits for additional contents \citep{Jamieson2011,Bengs2021}.
\end{example}

\begin{example}[Pricing a product]
  Suppose that we want to sell a product to a consumer characterized by some features $x$, this consumer is ready to pay a price $y\in \R$ for this product.
  We price it $f(x)\in\R$, and we observe $\ind{f(x) < y}$, that is if the consumer is willing to buy this product at this price tag or not \citep{Cesa-Bianchi2019,Liu2021}.
  Although, in this setting, the goal is often to minimize the regret, which contrasts with our pure exploration setting.
\end{example}

As a counter-example, our assumptions are not set to deal with missing data, {\em i.e.} if some coordinates of some input feature vectors $X_i$ are missing \citep{Rubin1976}.
Typically, this happens when input data comes from different sources ({\em e.g.}, when trying to predict economic growth from country information that is self-reported).

\paragraph{Streaming variation.}
The special case of the active labeling problem we shall consider consists in its variant without resampling.
This corresponds to the online setting where one can only ask one question by sample, formally $i_t=t$.
This setting is particularly appealing for privacy concerns, in settings where the labels $(Y_i)$ contain sensitive information that should not be revealed totally.
For example, some people might be more comfortable giving a range over a salary rather than the exact value; or in the context of polling, one might not call back a previous respondent characterized by some features $X_i$ to ask them again about their preferences captured by $Y_i$.
Similarly, the streaming setting is relevant for web marketing, where inputs model new users visiting a website, queries model sets of advertisements chosen by an advertising company, and one observes potential clicks.

\section{Weak information as stochastic gradients}
\label{sgd:sec:sgd}

In this section, we discuss how stochastic gradients can be accessed through weak information.

Suppose that we model $f = f_\theta$ for some Hilbert space $\Theta\ni\theta$.
With some abuse of notations, let us denote $\ell(x, y, \theta) := \ell(f_\theta(x), y)$.
We aim to minimize
\(
{\cal R}(\theta) = \E_{(X, Y)}\bracket{\ell(X, Y, \theta)}.
\)
Assume that ${\cal R}$ is differentiable (or sub-differentiable) and denote its gradients by $\nabla_\theta {\cal R}$.

\begin{definition}[Stochastic gradient]
  A stochastic gradient of ${\cal R}$ is any random function $G:\Theta \to \Theta$ such that
  \(
  \E[G(\theta)] = \nabla_\theta{\cal R}(\theta).
  \)
  Given some step size function $\gamma:\N\to\R^*$, a stochastic gradient descent (SGD) is a procedure, $(\theta_t) \in \Theta^\N$, initialized with some $\theta_0$ and updated as
  \(
  \theta_{t+1} = \theta_t - \gamma(t) G(\theta_t),
  \)
  where the realization of $G(\theta_t)$ given $\theta_t$ is independent of the previous realizations of $G(\theta_{s})$ given $\theta_s$.
\end{definition}

In supervised learning, SGD is usually performed with the stochastic gradients $\nabla_\theta \ell(X, Y, \theta)$.
More generally, stochastic gradients are given by
\begin{equation}
  \label{sgd:eq:sgd_def}
  G(\theta) = \ind{\nabla_\theta \ell(X, Y, \theta) \in T}\cdot \tau(T),
\end{equation}
for $\tau:{\cal T}\to\Theta$ with ${\cal T} \subset 2^\Theta$ a set of subsets of $\Theta$, and $T$ a random variable on ${\cal T}$, such that
\begin{equation}
  \label{sgd:eq:condition}
  \forall\, \theta \in\Theta,\quad \E_T[\ind{\theta\in T} \cdot \tau(T)] = \theta.
\end{equation}
Stated otherwise, if you have a way to image a vector $\theta$ from partial measurements $\ind{\theta\in T}$ such that you can reconstruct this vector in a linear fashion \eqref{sgd:eq:condition}, then it provides you a generic strategy to get an unbiased stochastic estimate of this vector from a partial measurement \eqref{sgd:eq:sgd_def}.

For $\psi:\Y\to\Theta$ a function from $\Y$ to $\Theta$ ({\em e.g.}, $\psi = \nabla_\theta(X, \cdot, \theta)$), a question $\ind{\psi(Y) \in T}$ translates into a question $\ind{Y\in S}$ for some set $S = \psi^{-1}(T) \subset \Y$, meaning that the stochastic gradient \eqref{sgd:eq:sgd_def} can be evaluated from a single query.
As a proof of concept, we derive a generic implementation for $T$ and $\tau$ in Appendix \ref{sgd:app:generic}.
This provides a generic SGD scheme to learn functions from weak queries when there are no constraints on the sets to query.

\begin{remark}[Cutting plane methods]
  While we provide here a descent method, one could also develop cutting-plane/ellipsoid methods to localize $\theta^*$ according to weak information, which corresponds to the techniques developed for pricing by \cite{Cohen2020} and related literature.
\end{remark}

\section{Median regression}
\label{sgd:sec:median}

In this section, we focus on efficiently acquiring weak information providing stochastic gradients for regression problems.
In particular, we motivate and detail our methods for the absolute deviation loss.

Motivated by seminal works on censored data \citep{Tobin1958}, we shall suppose that {\em we query half-spaces}.
For an output $y\in\Y=\R^m$, and any hyper-plane $z + u^\perp \subset \R^m$ for $z \in \R^m$, $u \in \mathbb{S}^{m-1}$, we can ask a labeler to tell us which half-space $y$ belongs to.
Formally, {\em we access the quantity $\sign(\scap{y - z}{u})$ for a given unit cost}.

\paragraph{Least-squares.}
For regression problems, it is common to look at the mean square loss
\[
  \ell(X, Y, \theta) = \norm{f_\theta(X) - Y}^2,\qquad
  \nabla_\theta\ell(X, Y, \theta) = 2(f_\theta(X) - Y)^\top Df_\theta(X),
\]
where $Df_\theta(x)\in\Y\otimes\Theta$ denotes the Jacobian of $\theta \to f_\theta(x)$.
In rich parametric models, it is preferable to ask questions on $Y\in\Y$ rather than on gradients in $\Theta$ which is a potentially much bigger space.
If we assume that $Y$ and $f_\theta(X)$ are bounded in $\ell^2$-norm by $M \in \R_+$, we can adapt~\eqref{sgd:eq:sgd_def} and \eqref{sgd:eq:condition} through the fact that for any $z \in \Y$, such that $\norm{z} \leq 2M$, as proven in Appendix \ref{sgd:app:generic},
\[
  \E_{U, V}\bracket{\ind{\scap{z}{U} \geq V}\cdot U}
  = c_1\cdot z,\quad\text{where}\quad
  c_1 = \E_{U, V}\bracket{\ind{\scap{e_1}{U} \geq V} \cdot\scap{e_1}{U}}
  = \frac{\pi^{3/2}}{2M (m^2 + 4m + 3)},
\]
for $U$ uniform on the sphere $\mathbb{S}^{m-1}$ and $V$ uniform on $[0, 2M]$.
Applied to $z = f_\theta(X) - Y$, it designs an SGD procedure by querying information of the type
\(
\ind{\scap{Y}{U} < \scap{f_\theta(X)}{U} - V}.
\)

\paragraph{A case for median regression.}
Motivated by robustness purposes, we will rather expand on median regression.
In general, we would like to learn a function that, given an input, replicates the output of I/O samples generated by the joint probability $\rho$.
In many instances, $X$ does not characterize all the sources of variations of $Y$, {\em i.e.} input features are not rich enough to characterize a unique output, leading to randomness in the conditional distributions $(Y \vert X)$.
When many targets can be linked to a vector $x\in\X$, how to define a consensual $f(x)$?
For analytical reasons, statisticians tend to use the least-squares error which corresponds to asking for $f(x)$ to be the mean of the distribution $\paren{Y\vert X=x}$.
Yet, means are known to be too sensitive to rare but large outputs \citep[see \emph{e.g.},][]{Huber1981}, and cannot be defined as good and robust consensus in a world of heavy-tailed distributions.
This contrasts with the median, which, as a consequence, is often much more valuable to summarize a range of values.
For instance, median income is preferred over mean income as a population indicator \citep[see \emph{e.g.},][]{USCensus}.

\paragraph{Median regression.} The geometric median is variationally defined through the absolute deviation loss, leading to
\begin{equation}
  \label{sgd:eq:median}
  \ell(X, Y, \theta) = \norm{f_\theta(X) - Y},\qquad
  \nabla_\theta \ell(X, Y, \theta) =
  \paren{\frac{f_\theta(X) - Y}{\norm{f_\theta(X) - Y}}}^\top
  Df_\theta(X).
\end{equation}
Similarly to the least-squares case, we can access weakly supervised stochastic gradients through the fact that for $z \in \mathbb{S}^{m-1}$, as shown in Appendix \ref{sgd:app:generic},
\begin{equation}
  \label{sgd:eq:median_sgd}
  \E_{U}\bracket{\sign\paren{\scap{z}{U}}\cdot U}
  = c_2\cdot z,\quad\text{where}\quad
  c_2 = \E_{U}\bracket{\sign\paren{\scap{e_1}{U}} \cdot\scap{e_1}{U}}
  = \frac{\sqrt{\pi}\Gamma(\frac{m-1}{2})}{m \Gamma(\frac{m}{2})},
\end{equation}
where $U$ is uniformly drawn on the sphere $\mathbb{S}^{m-1}$, and $\Gamma$ is the gamma function.
This suggests Algorithm~\ref{sgd:alg:sgd}.

\begin{algorithm}[H]
  \caption{Median regression with SGD.}
  \KwData{A model $f_\theta$ for $\theta \in \Theta$, some data $(X_i)_{i\leq n}$, a labeling budget $T$, a step size rule $\gamma:\N\to\R_+$}
  \KwResult{A learned parameter $\hat{\theta}$ and the predictive function
    $\hat{f} = f_{\hat{\theta}}$.}

  Initialize $\theta_0$.\\
  \For{$t\gets 1$ \KwTo $T$}{
    Sample $U_t$ uniformly on $\mathbb{S}^{m-1}$.\\
    Query $\epsilon = \sign(\scap{Y_t-z}{U_t})$ for $z =
      f_{\theta_{t-1}}(X_t)$.\\
    Update the parameter
    $\theta_{t} = \theta_{t-1} + \gamma(t) \epsilon\cdot U_t^\top(Df_{\theta_{t-1}}(X_t))$.
  }
  Output $\hat{\theta} = \theta_T$, or some average, {\em e.g.}, $\hat{\theta} = T^{-1}\sum_{t=1}^T \theta_t$.
  \label{sgd:alg:sgd}
\end{algorithm}

\section{Statistical analysis}
\label{sgd:sec:stat}

In this section, we quantify the performance of Algorithm~\ref{sgd:alg:sgd} by proving optimal rates of convergence when the median regression problem is approached with (reproducing) kernels.
For simplicity, we will assume that $f^*$ can be parametrized by a linear model (potentially of infinite dimension).

\begin{assumption}
  \label{sgd:ass:source}
  Assume that the solution $f^*:\X\to\R^m$ of the median regression problem~\eqref{sgd:eq:obj} and \eqref{sgd:eq:median} can be parametrized by some separable Hilbert space ${\cal H}$, and a bounded feature map $\phi:\X\to{\cal H}$, such that, for any $i \in [m]$, there exists some $\theta_i^* \in \cal H$ such that
  \(
  \scap{f^*(\cdot)}{e_i}_{\Y} = \scap{\theta_i^*}{\phi(\cdot)}_{\cal H},
  \)
  where $(e_i)$ is the canonical basis of $\R^m$.
  Written into matrix form, there exists $\theta^* \in \Y\otimes{\cal H}$, such that
  \(
  f^*(\cdot) = \theta^* \phi(\cdot).
  \)
\end{assumption}

The curious reader can easily relax this assumption in the realm of reproducing kernel Hilbert spaces following the work of \citet{PillaudVivien2018}.
Under the linear model of Assumption~\ref{sgd:ass:source}, Algorithm~\ref{sgd:alg:sgd} is specified with
\(
u^\top Df_\theta(x) = u\otimes \phi(x).
\)
Note that rather than working with $\Theta = \Y \otimes {\cal H}$ which is potentially infinite-dimensional, empirical estimates can be represented in the finite-dimensional space $\Y \otimes \Span\brace{\phi(X_i)}_{i\leq n}$, and well approximated by small-dimensional spaces to ensure efficient computations \citep{Williams2000,Meanti2020}.

One of the key points of SGD is that gradient descent is so gradual that one can use noisy or stochastic gradients without loosing statistical guarantees while speeding up computations.
This is especially true when minimizing convex functions that are nor strongly-convex, {\em i.e.}, bounded below by a quadratic, nor smooth, {\em i.e.}, with Lipschitz-continuous gradient \citep[see, \emph{e.g.},][]{Bubeck2015}.
In particular, the following theorem, proven in Appendix~\ref{sgd:proof:sgd}, states that Algorithm~\ref{sgd:alg:sgd} minimizes the population risk at a speed at least proportional to $O(T^{-1/2})$.

\begin{theorem}[Convergence rates]
  \label{sgd:thm:sgd}
  Under Assumption~\ref{sgd:ass:source}, and under the knowledge of $\kappa$ and $M$ two real values such that $\E[\norm{\phi(X)}^2] \leq \kappa^2$ and $\norm{\theta^*} \leq M$, with a budget $T\in\N$, a constant step size $\gamma = \frac{M}{\kappa\sqrt{T}}$ and the average estimate $\hat{\theta} = \frac{1}{T}\sum_{t=0}^{T-1} \theta_t$, Algorithm~\ref{sgd:alg:sgd} leads to an estimate $f$ that suffers from an excess of risk
  \begin{equation}
    \label{sgd:eq:thm}
    \E\bracket{{\cal R}\paren{f_{\hat{\theta}}}} - {\cal R}(f^*)
    \leq \frac{2\kappa M}{c_2 \sqrt{T}}
    \leq \kappa M m^{3/2} T^{-1/2},
  \end{equation}
  where the expectation is taken with respect to the randomness of $\hat{\theta}$ that depends on the dataset $(X_i, Y_i)$ as well as the questions $(i_t, S_t)_{t\leq T}$.
\end{theorem}

While we give here a result for a fixed step size, one could retake the extensive literature on SGD to prove similar results for decaying step sizes that do not require to know the labeling budget in advance ({\em e.g.} setting $\gamma(t) \propto t^{-1/2}$ at the expense of an extra term in $\log(T)$ in front of the rates), as well as different averaging strategies \citep[see \emph{e.g.},][]{Bach2023}.
In practice, one might not know {\em a priori} the parameter $M$ but could nonetheless find the right scaling for $\gamma$ based on cross-validation.

The rate in $O(T^{-1/2})$ applies more broadly to all the strategies described in Section~\ref{sgd:sec:sgd} as long as the loss $\ell$ and the parametric model $f_\theta$ ensure that ${\cal R}(\theta)$ is convex and Lipschitz-continuous.
Although the constants appearing in front of rates depend on the complexity to reconstruct the full gradient $\nabla_\theta \ell(f_\theta(X_i, Y_i))$ from the reconstruction scheme \eqref{sgd:eq:condition}.
Those constants correspond to the second moment of the stochastic gradient.
For example, for the least-squares technique described earlier one would have to replace $c_2$ by $c_1$ in~\eqref{sgd:eq:thm}.

Theorem~\ref{sgd:thm:minmax_opt}, proven in Appendix~\ref{sgd:proof:lower}, states that any algorithm that accesses a fully supervised learning dataset of size $T$ cannot beat the rates in $O(T^{-1/2})$, hence any algorithm that collects weaker information on $(Y_i)_{i\leq T}$ cannot display better rates than the ones verified by Algorithm~\ref{sgd:alg:sgd}. This proves minimax optimality of our algorithm up to constants.

\begin{theorem}[Minimax optimality]
  \label{sgd:thm:minmax_opt}
  Under Assumption~\ref{sgd:ass:source} and the knowledge of an upper bound on $\norm{\theta^*}\leq M$, assuming that $\phi$ is bounded by $\kappa$, there exists a universal constant $c_3$ such that for any algorithm~${\cal A}$ that takes as input ${\cal D}_T = (X_i, Y_i)_{i\leq T} \sim\rho^{\otimes T}$ for any $T\in\N$ and output a parameter $\theta$,
  \begin{equation}
    \sup_{\rho\in {\cal M}_{M}} \E_{{\cal D}_T\sim\rho^{\otimes T}}\bracket{{\cal R}(f_{{\cal A}({\cal D}_T; \rho)})} - {\cal R}(f_\rho; \rho) \geq c_3 M\kappa T^{-1/2}.
  \end{equation}
  The supremum over $\rho\in{\cal M}_{M}$ has to be understood as the supremum over all distributions $\rho\in\prob{\X\times\Y}$ such that the problem defined through the risk ${\cal R}(f; \rho) := \E_{\rho}[\ell(f(X), Y)]$ is minimized for $f_\rho$ that verifies Assumption~\ref{sgd:ass:source} with $\norm{\theta^*}$ bounded by a constant $M$.
\end{theorem}

The same theorem applies for least-squares with a different universal constant.
It should be noted that minimax lower bounds are in essence quantifying worst cases of a given class of problems.
In particular, to prove Theorem~\ref{sgd:thm:minmax_opt}, we consider distributions that lead to hard problems; more specifically, we assumed the variance of the conditional distribution $\paren{Y\midvert X}$ to be high.
The practitioner should keep in mind that it is possible to add additional structure on the solution, leverage active learning or semi-supervised strategy such as uncertainty sampling \citep{Nguyen2021}, or Laplacian regularization \citep{Zhu2003,Cabannes2021c}, and reduce the optimal rates of convergence.

To conclude this section, let us remark that most of our derivations could easily be refined for practitioners facing a slightly different cost model for annotation.
In particular, they might prefer to perform batches of annotations before updating $\theta$ rather than modifying the question strategy after each input annotation.
This would be similar to mini-batching in gradient descent.
Indeed, the dependency of our result on the annotation cost model and on Assumption~\ref{sgd:ass:source} should not be seen as a limitation but rather as a proof of concept.

\section{Numerical analysis}
\label{sgd:sec:exp}

In this section, we illustrate the differences between our active method versus a classical passive method, for regression and classification problems.
Extensive details are provided in Appendix \ref{sgd:app:experiments}.
Our code is available online at \url{anonymized-url}.

\begin{figure}[ht]
  \centering
  \includegraphics{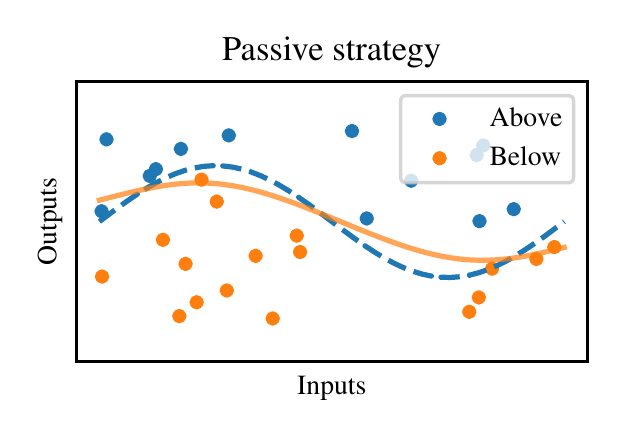}
  \includegraphics{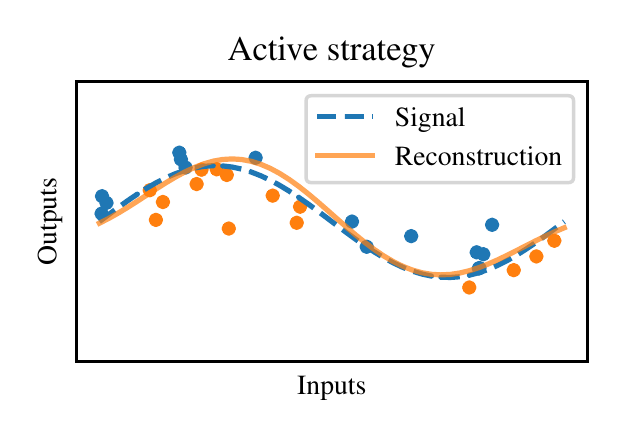}
  \caption{
    {\em Visual comparison of active and passive strategies.}
    Estimation in orange of the original signal $f^*$ in dashed blue based on median regression in a noiseless setting.
    Any orange point $(x, u)\in\R^2$ corresponds to an observation made that $u$ is below $f^*(x)$, while a blue point corresponds to $u$ above $f^*(x)$.
    The passive strategy corresponds to acquiring information based on $\paren{U\midvert x}$ following a normal distribution, while the active strategy corresponds to $\paren{u\midvert x} = f_{\theta}(x)$.
    The active strategy reconstructs the signal much better given the budget of $T=30$ observations.
  }
  \label{sgd:fig:exp_1}
\end{figure}

Let us begin with the regression problem that consists in estimating the function $f^*$ that maps $x\in[0,1]$ to $\sin(2\pi x) \in \R$.
Such a regular function, which belongs to any H\"older or Sobolev classes of functions, can be estimated with the Gaussian kernel, which would ensure Assumption \ref{sgd:ass:source}, and that corresponds to a feature map $\phi$ such that $k(x, x'):=\scap{\phi(x)}{\phi(x')} = \exp(-\abs{x-x'}/(2\sigma^2))$ for any bandwidth parameter $\sigma > 0$.\footnote{A noteworthy computational aspect of linear models, often refer as the ``kernel trick'', is that the features map $\phi$ does not need to be explicit, the knowledge of $k:\X\times\X\to\R$ being sufficient to compute all quantities of interest \citep{Scholkopf2001}. This ``trick'' can be applied to our algorithms.}
On Figure \ref{sgd:fig:exp_1}, we focus on estimating $f^*$ given data $(X_i)_{i\in [T]}$ that are uniform on $[0, 1]$ in the noiseless setting where $Y_i = f^*(X_i)$, based on the minimization of the absolute deviation loss.
The passive baseline consists in randomly choosing a threshold $U_i\sim{\cal N}(0, 1)$ and acquiring the observations $(\ind{Y_i > U_i})_{i\in [T]}$ that can be cast as the observation of the half-space $S_i = \brace{y\in\Y\midvert \ind{y > U_i} = \ind{Y_i > U_i}} =: s(Y_i, U_i)$.
In this noiseless setting, a good baseline to learn $f^*$ from the data $(X_i, S_i)$ is provided by the infimum loss characterization \citep[see][]{Cabannes2020}
\[
  f^* = \argmin_{f:\X\to\Y} \E_{(X, S)}[\inf_{y\in S} \ell(f(X), y)],
\]
where the distribution over $X$ corresponds to the marginal of $\rho$ over $\X$, and the distribution over $\paren{S\midvert X=x}$ is the pushforward of $U\sim{\cal N}(0, 1)$ under $s(f^*(x), \cdot)$.
The left plot on Figure \ref{sgd:fig:exp_1} corresponds to an instance of SGD on such an objective based on the data $(X_i, S_i)$, while the right plot corresponds to Algorithm \ref{sgd:alg:sgd}.
We take the same hyperparameters for both plots, a bandwidth $\sigma=0.2$ and an SGD step size $\gamma =0.3$.
We refer the curious reader to Figure \ref{sgd:fig:exp_1_app} in Appendix \ref{sgd:app:experiments} for plots illustrating the streaming history, and to Figure \ref{sgd:fig:exp_libsvm} for ``real-world'' experiments.

To illustrate the versatility of our method, we approach a classification problem through the median surrogate technique presented in Proposition \ref{sgd:prop:sur}.
To do so, we consider the classification problem with $m\in\N$ classes, $\X = [0,1]$ and the conditional distribution $\paren{Y\midvert X}$ linearly interpolating between Dirac in $y_1$, $y_2$ and $y_3$ respectively for $x=0$, $x=1/2$ and $x=1$ and the uniform distribution for $x=1/4$ and $x=3/4$; and $X$ uniform on $\X \setminus ([1/4 - \epsilon, 1/4+\epsilon] \cup [3/4-\epsilon, 3/4+\epsilon])$.

\begin{figure}[ht]
  \centering
  \includegraphics{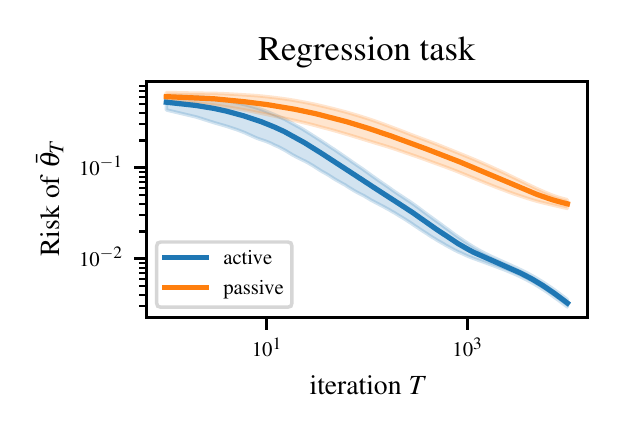}
  \includegraphics{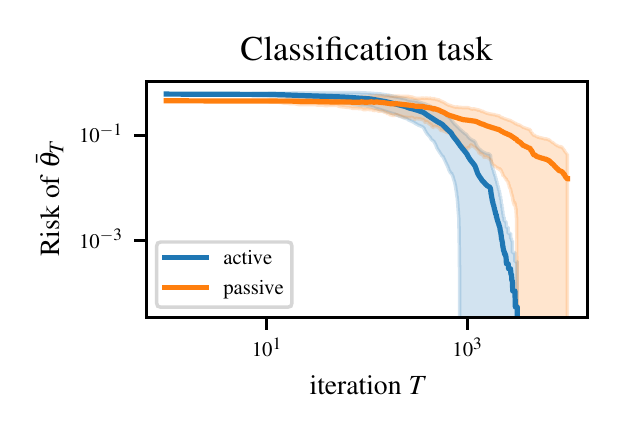}
  \caption{
    {\em Comparison of generalization errors of passive and active strategies} as a function of the annotation budget $T$.
    This error is computed by averaging over 100 trials.
    In solid is represented the average error, while the height of the dark area represents one standard deviation on each side.
    In order to consider the streaming setting where $T$ is not known in advance, we consider the decreasing step size $\gamma(t) = \gamma_0 /\sqrt{t}$; and to smooth out the stochasticity due to random gradients, we consider the average estimate $\bar\theta_t = (\theta_1 + \cdots + \theta_t) / t$.
    The left figure corresponds to the noiseless regression setting of Figure \ref{sgd:fig:exp_1}, with $\gamma_0 = 1$. We observe the convergence behavior in $O(T^{-1/2})$ of our active strategy.
    The right setting corresponds to the classification problem setting described in the main text with $m=100$, $\epsilon = 1/20$, and approached with the median surrogate.
    We observe the exponential convergence phenomenon described by \cite{Cabannes2021b}; its kicks in earlier for the active strategy.
    The two plots are displayed with logarithmic scales on both axes.
  }
  \label{sgd:fig:exp_2}
\end{figure}

\section{Discussion}
\label{sgd:sec:discussion}

\subsection{Discrete output problems}

Learning problems with discrete output spaces are not as well understood as regression problems.
This is a consequence of the complexity of dealing with combinatorial structures in contrast with continuous metric spaces.
In particular, gradients are not defined for discrete output models.
The current state-of-the-art framework to deal with discrete output problems is to introduce a continuous surrogate problem whose solution can be decoded as a solution on the original problem \citep{Bartlett2006}.
For example, one could solve a classification task with a median regression surrogate problem, which is the object of the next proposition, proven in Appendix~\ref{sgd:proof:sur}.

\begin{proposition}[Consistency of median surrogate]
  \label{sgd:prop:sur}
  The classification setting where $\Y$ is a finite space, and $\ell:\Y\times\Y\to\R$ is the zero-one loss $\ell(y, z) = \ind{y\neq z}$ can be solved as a regression task through the simplex embedding of $\Y$ in $\R^\Y$ with the orthonormal basis $(e_y)_{y\in\Y}$.
  More precisely, if $g^*:\X\to\R^\Y$ is the minimizer of the median surrogate risk ${\cal R}_S(g) = \E\bracket{\norm{g(X) - e_Y}}$, then $f^*:\X\to\Y$ defined as $f^*(x) = \argmax_{y\in\Y} g^*_y(x)$ minimizes the original risk ${\cal R}(f) = \E\bracket{\ell(f(X), Y)}$.
\end{proposition}

More generally, any discrete output problem can be solved by reusing the consistent least-squares surrogate of \cite{Ciliberto2020}.
Algorithm \ref{sgd:alg:sgd} can be adapted to the least-squares problem based on specifications at the beginning of Section \ref{sgd:sec:median}.
This allows using our method in an off-the-shelve fashion for all discrete output problems.
In this setting, Theorem \ref{sgd:thm:sgd} can be refined under margin conditions where our approach would exhibit exponential convergence rates as illustrated on Figure \ref{sgd:fig:exp_2}.
As a side note, while we are not aware of any generic theory encompassing the absolute-deviation surrogate of Proposition~\ref{sgd:prop:sur}, we showcase its superiority over least-squares on at least two types of problems on Figures \ref{sgd:fig:med_ls} and \ref{sgd:fig:med_simplex} in Appendix \ref{sgd:proof:sur}.

\subsection{Supervised learning baseline with resampling}
When resampling is allowed a simple baseline for the active labeling problem is provided by supervised learning.
In regression problems with the query of any half-space, a method that consists in annotating each $(Y_i)_{i\leq n(T, \epsilon)}$ up to precision $\epsilon$, before using any supervised learning method to learn $f$ from $(X_i, Y_i)_{i\leq n(T, \epsilon)}$ could acquire $n(T, \epsilon) \simeq T / m \log_2(\epsilon^{-1})$ data points with a dichotomic search along all directions, assuming $Y_i$ bounded or sub-Gaussian.
In terms of minimax rates, such a procedure cannot perform better than in $n(T,\epsilon)^{-1/2} + \epsilon$, the first term being due to the statistical limit in Theorem~\ref{sgd:thm:minmax_opt}, the second due to the incertitude $\epsilon$ on each $Y_i$ that transfers to the same level of incertitude on $f$.
Optimizing with respect to $\epsilon$ yields a bound in $O(T^{-{1/2}}\log(T)^{1/2})$.
Therefore, this not-so-naive baseline is only suboptimal by a factor $\log(T)^{1/2}$.
In the meanwhile, Algorithm \ref{sgd:alg:sgd} can be rewritten with resampling, as well as Theorem \ref{sgd:thm:sgd}, which we prove in Appendix \ref{sgd:proof:resampling}.
Hence, our technique will still achieve minimax optimality for the problem ``with resampling''.
In other terms, by deciding to acquire more imprecise information, our algorithm reduces annotation cost for a given level of generalization error (or equivalently reduces generalization error for a given annotation budget) by a factor $\log(T)^{1/2}$ when compared to this baseline.

The picture is slightly different for discrete-output problems.
If one can ask any question $s\in 2^\Y$ then with a dichotomic search, one can retrieve any label with $\log_2(m)$ questions.
Hence, to theoretically beat the fully supervised baseline with the SGD method described in Section~\ref{sgd:sec:sgd}, one would have to derive a gradient strategy \eqref{sgd:eq:sgd_def} with a small enough second moment ({\em e.g.}, for convex losses that are non-smooth nor strongly convex, the increase in the second moment compared to the usual stochastic gradients should be no greater than $\log_2(m)^{1/2}$).
How to best refine our technique to better take into account the discrete structure of the output space is an open question.
Introducing bias that does not modify convergence properties while reducing variance eventually thanks to importance sampling is a potential way to approach this problem.
A simpler idea would be to remember information of the type $Y_i \in s$ to restrict the questions asked in order to locate $f_{\theta_t}(X_i) - Y_i$ when performing stochastic gradient descent with resampling.
Combinatorial bandits might also provide helpful insights on the matter.
Ultimately, we would like to build an understanding of the whole distribution $\paren{Y\midvert X}$ and not only of $f^*(X)$ as we explore labels in order to refine this exploration.

\subsection{Min-max approaches}
Min-max approaches have been popularized for searching games and active learning, where one searches for the question that minimizes the size of the space where a potential guess could lie under the worst possible answer to that question.
A particularly well illustrative example is the solution of the Mastermind game proposed by \cite{Knuth1977}.
While our work leverages plain SGD, one could build on the vector field point-of-view of gradient descent \citep[see, \emph{e.g.},][]{Bubeck2015} to tackle min-max convex concave problems with similar guarantees.
In particular, we could design weakly supervised losses $L(f(x), s; \ind{y\in s})$ and min-max games where a prediction player aims at minimizing such a loss with respect to the prediction $f$, while the query player aims at maximizing it with respect to the question $s$, that is querying information that best elicit mistakes made by the prediction player.
For example, the dual norm characterization of the norm leads to the following min-max approach to the median regression
\[
  \argmin_{f:\X\to\Y}{\cal R}(f) = \argmin_{f:\X\to\Y}\max_{U\in(\mathbb{S}^{m-1})^{\X\times\Y}} \E_{(X, Y)\sim\rho}\bracket{\scap{U(x, y)}{f(x) - y}}.
\]
Such min-max formulations would be of interest if they lead to improvement of computational and statistical efficiencies, similarly to the work of \cite{Babichev2019}.
For classification problems, the following proposition introduces such a game and suggests its suitability.
Its proof can be found in Appendix \ref{sgd:proof:minmax}.

\begin{proposition}
  \label{sgd:prop:minmax}
  Consider the classification problem of learning $f^*:\X\to\Y$ where $\Y$ is of finite cardinality, with the 0-1 loss $\ell(z, y) = \ind{z\neq y}$, minimizing the risk \eqref{sgd:eq:obj} under a distribution $\rho$ on $\X\times\Y$.
  Introduce the surrogate score functions $g:\X\to\prob{\Y}; x\to v$ where $v = (v_y)_{y\in \Y}$ is a family of non-negative weights that sum to one, as well as the surrogate loss function $L:\prob{\Y}\times{\cal S}\times\brace{-1, 1}\to \R; (v, S, \epsilon) = \epsilon(1 - 2\sum_{y\in S} v_y)$, and the min-max game
  \begin{equation}
    \label{sgd:eq:minmax}
    \min_{g:\X\to\prob{\Y}} \max_{\mu:\X\to\prob{\cal S}} \E_{(X,Y)\sim\rho} \E_{S\sim \mu(x)}\bracket{L(g(x), S; \ind{Y\in S}-\ind{Y\notin S})}.
  \end{equation}
  When ${\cal S}$ contains the singletons and with the low-noise condition that $\Pbb\paren{Y\neq f^*(x)\midvert X=x} < 1/2$ almost everywhere, then $f^*$ can be learned through the relation $f^*(x) = \argmin_{y\in\Y} g^*(x)_y$ for the unique minimizer $g^*$ of \eqref{sgd:eq:minmax}.
  Moreover, the minimization of the empirical version of this objective with the stochastic gradient updates for saddle point problems provides a natural ``active labeling'' scheme to find this $g^*$.
\end{proposition}

\section{Conclusion}
We have introduced the ``active labeling'' problem, which corresponds to ``active partially supervised learning''.
We provided a solution to this problem based on stochastic gradient descent.
Although our method can be used for any discrete output problem, we detailed how it works for median regression, where we show that it optimizes the generalization error for a given annotation budget.
In a near future, we would like to focus on better exploiting the discrete structure of classification problems, eventually with resampling strategies.

Understanding more precisely the key issues in applications concerned with privacy, and studying how weak gradients might provide a good trade-off between learning efficiently and revealing too much information also provide interesting follow-ups.
Finally, regarding dataset annotation, exploring different paradigms of weakly supervised learning would lead to different active weakly supervised learning frameworks.
While this work is based on partial labeling, similar formalization could be made based on other weak supervision models, such as aggregation \citep[\emph{e.g.},][]{Ratner2020}, or group statistics \citep{Dietterich1997}.
In particular, annotating a huge dataset is often done by bagging inputs according to predicted labels and correcting errors that can be spotted on those bags of inputs \citep{ImageNet}.
We left for future work the study of variants of the ``active labeling'' problem that model those settings.

\begin{subappendices}
  \chapter*{Appendix}
  \addcontentsline{toc}{chapter}{Appendix}
	\section{Proofs of the statistical analysis}
In the following proofs, we assume $\X$ to be Polish and $\Y = \R^m$, so to define the joint probability $\rho\in\prob{\X\times\Y}$.
Moreover, we assume that $\E[\norm{Y}] < +\infty$ in order to define the risk of median regression.
We consider ${\cal H}$ to be a Hilbert space that is separable ({\em i.e.} only the origin is in all the neighborhood of the origin), and $\phi$ to be a measurable mapping from $\X$ to ${\cal H}$.

In terms of notations, we denote $\brace{1, 2, \cdots, n}$ by $[n]$ for any $n\in\N^*$, and by $(x_i)_{i\leq n}$ the family $(x_1, \cdots, x_n)$ for any sequence $(x_i)$.
The unit sphere in $\R^m$ is denoted by $\mathbb{S}^{m-1}$.
The symbol $\otimes$ denotes tensors, and is extended to product measures in the notation $\rho^{\otimes n} = \rho\times\rho\times\cdots\times\rho$.
We have used the isometry between trace-class linear mappings from ${\cal H}$ to $\Y$ and the tensor space $\Y\otimes{\cal H}$, which generalizes the matrix representation of linear map between two finite-dimensional vector spaces.
This space inherits from the Hilbertian structure of ${\cal H}$ and $\Y$ and we denote by $\norm{\cdot}$ the Hilbertian norm that generalizes the Frobenius norm on linear maps between Euclidean spaces.

\subsection{Upper bound for stochastic gradient descent}
\label{sgd:proof:sgd}

This subsection is devoted to the proof of Theorem \ref{sgd:thm:sgd}.
For simplicity, we will work with the rescaled step size $\gamma_t := c_2 \gamma(t)$ rather than the step size described in the main text $\gamma(t)$.

Convergence of stochastic gradient descent for non-smooth problems is a known result. For completeness, we reproduce and adapt a usual proof to our setting.
For $t\in\N$, let us introduce the random functions
\[
	{\cal R}_t(\theta) = c_2^{-1}\abs{\scap{\theta\phi(X_t) - Y_t}{U_t}},
	\qquad\text{where}\qquad
	c_2 = \E_U[\abs{\scap{e_1}{U}}] = \E_U[\sign(\scap{e_1}{U}) \scap{e_1}{U}]
\]
for $(X_t, Y_t) \sim \rho$, $U_t$ uniform on the sphere $\mathbb{S}^{m-1} \subset \Y$.
Those random functions all average to ${\cal R}(\theta) = \E_{\rho} \E_U[c_2^{-1}\abs{\scap{\theta\phi(X) - Y}{U}}] = \E_\rho[\norm{\theta\phi(X) - Y}]$.
After a random initialization $\theta_0 \in\Theta$, the stochastic gradient update rule can be written for any $t \in \N$ as
\[
	\theta_{t+1} = \theta_t - \gamma_t \nabla {\cal R}_t(\theta_t),
\]
where $\nabla {\cal R}_t$ denotes any sub-gradients of ${\cal R}_t$.
We can compute
\[
	\nabla {\cal R}_t(\theta_t)
	= c_2^{-1} \nabla \abs{\scap{\theta\phi(X_t) - Y_t}{U_t}}
	= c_2^{-1} \sign\paren{\scap{\theta\phi(X_t) - Y_t}{U_t}} U_t\otimes \phi(X_t).
\]
This corresponds to the gradient written in Algorithm \ref{sgd:alg:sgd}.

Let us now express the recurrence relation on $\norm{\theta_{t+1} - \theta^*}$.
We have
\begin{align*}
	\norm{\theta_{t+1} - \theta^*}^2
	 & = \norm{\theta_t - \gamma_t \nabla {\cal R}_t(\theta_t) - \theta^*}^2
	\\&= \norm{\theta_t - \theta^*}^2 + \gamma_t^2\norm{\nabla {\cal R}_t(\theta_t)}^2 - 2\gamma_t\scap{\nabla {\cal R}_t(\theta_t)}{\theta_t - \theta^*}.
\end{align*}
Because ${\cal R}_t$ is convex, it is above its tangents
\[
	{\cal R}_t(\theta^*) \geq {\cal R}_t(\theta_t) + \scap{\nabla{\cal R}_t(\theta_t)}{\theta^* - \theta_t}.
\]
Hence,
\[
	\norm{\theta_{t+1} - \theta^*}^2
	\leq \norm{\theta_t - \theta^*}^2 + \gamma_t^2\norm{\nabla {\cal R}_t(\theta_t)}^2 + 2\gamma_t ({\cal R}_t(\theta^*) - {\cal R}_t(\theta_t)).
\]
This allows bounding the excess of risk as
\[
	2({\cal R}_t(\theta_t) - {\cal R}_t(\theta^*))
	\leq \frac{1}{\gamma_t}(\norm{\theta_t - \theta^*}^2 - \norm{\theta_{t+1} - \theta^*}^2) + \gamma_t c_2^{-2}\norm{\phi(X_t)}^2.
\]
where we used the fact that $\norm{\nabla {\cal R}_t} = c_2^{-1}\norm{\phi(X_t)}$.
Let us multiply this inequality by $\eta_t > 0$ and sum from $t=0$ to $t=T-1$, we get
\begin{align*}
	 & 2(\sum_{t=0}^{T-1} \eta_t {\cal R}_t(\theta_t) - \sum_{t=0}^{T-1} \eta_t {\cal R}_t(\theta^*))
	\leq \sum_{t=0}^{T-1} \frac{\eta_t}{\gamma_t}(\norm{\theta_t - \theta^*}^2 - \norm{\theta_{t+1} - \theta^*}^2) + \sum_{t=0}^{T-1} \eta_t\gamma_t c_2^{-2}\norm{\phi(X_t)}^2
	\\&\qquad= \frac{\eta_0}{\gamma_0} \norm{\theta_0-\theta^*}^2 - \frac{\eta_{T-1}}{\gamma_{T-1}} \norm{\theta_T - \theta^*}^2 + \sum_{t=1}^{T-1} \paren{\frac{\eta_t}{\gamma_t} - \frac{\eta_{t-1}}{\gamma_{t-1}}}\norm{\theta_t - \theta^*}^2 + \sum_{t=0}^{T-1} \eta_t\gamma_t c_2^{-2}\norm{\phi(X_t)}^2.
\end{align*}
From here, there is several options to obtain a convergence result, either one assume $\norm{\theta_t - \theta^*}$ bounded and take $\eta_t\gamma_{t-1} \geq \eta_{t-1}\gamma_t$; or one take $\eta_t = \gamma_t$ but at the price of paying an extra $\log(T)$ factor in the bound; or one take $\gamma_t$ and $\eta_t$ independent of $t$.
Since we suppose the annotation budget given, we will choose $\gamma_t$ and $\eta_t$ independent of $t$, only depending on $T$.
\begin{align*}
	 & 2(\sum_{t=0}^{T-1} \eta {\cal R}_t(\theta_t) - \sum_{t=0}^{T-1} \eta {\cal R}_t(\theta^*))
	\leq \frac{\eta}{\gamma} \norm{\theta_0-\theta^*}^2 + \sum_{t=0}^{T-1} \eta\gamma c_2^{-2}\norm{\phi(X_t)}^2.
\end{align*}

Let now take the expectation with respect to all the random variables, for the risk
\begin{align*}
	\E_{(X_s, Y_s, U_s)_{s\leq t}}[{\cal R}_t(\theta_t)]
	 & = \E_{(X_s, Y_s, U_s)_{s\leq t}}\bracket{\E_{(X_t, Y_t)}\bracket{\E_{U_t}\bracket{{\cal R}_t(\theta_t)\midvert \theta_t}\midvert \theta_t}}
	\\&= \E_{(X_s, Y_s, U_s)_{s\leq t}}\bracket{{\cal R}(\theta_t)}
	= \E[{\cal R}(\theta_t)].
\end{align*}
For the variance, $\E[\norm{\phi(X_s)}^2] = \E[\norm{\phi(X)}^2] = \kappa^2$.

Let us fix $T$ and consider $\eta_t = 1/ T$, by Jensen we can bound the following averaging
\begin{align*}
	2\paren{{\cal R}\paren{\sum_{t=0}^{T-1} \eta_t \theta_t} - {\cal R}(\theta^*)}
	 & \leq 2\paren{\sum_{t=0}^{T-1} \eta_t {\cal R}\paren{\theta_t} - {\cal R}(\theta^*)}
	= 2\E\bracket{\sum_{t=0}^{T-1} \eta_t ({\cal R}_t\paren{\theta_t} - {\cal R}_t(\theta^*))}
	\\&\qquad\leq \frac{1}{T\gamma}\norm{\theta_0-\theta^*}^2 + \gamma c_2^{-2}\kappa^2.
\end{align*}
Initializing $\theta_0$ to zero, we can optimize the resulting quantity to get the desired result.

\subsection{Upper bound for resampling strategy}
\label{sgd:proof:resampling}

For resampling strategies, the proof is built on classical statistical learning theory considerations.
Let us decompose the risk between estimation and optimization errors.
Recall the expression of the risk ${\cal R}$, the function taking as inputs measurable functions from $\X$ to $\Y$ and outputting a real number
\[
	{\cal R}(f) = \E_{\rho}[\norm{f(X) - Y}].
\]
Let us denote by ${\cal F}$ the class of functions from $\X$ to $\Y$ we are going to work with.
Let $f_n$ be our estimate of $f^*$ which maps almost every $x\in\X$ to the geometric median of $\paren{Y\midvert X}$.
Denote by ${\cal R}_{{\cal D}_n}^*$ the best value that can be achieved by our class of functions to minimize the empirical average absolute deviation
\[
	{\cal R}^*_{{\cal D}_n} = \inf_{f\in{\cal F}} {\cal R}_{{\cal D}_n}(f).
\]
Assumption \ref{sgd:ass:source} states that we have a well-specified model ${\cal F}$ to estimate the median, {\em i.e.} $f^* \in {\cal F}$.
Hence, the excess of risk can be decomposed as an estimation and an optimization error, without approximation error (it is not difficult to add an approximation error, but it will make the derivations longer and the convergence rates harder to parse for the reader).
Using the fact that ${\cal R}_{{\cal D}_n}(f^*) \geq {\cal R}_{{\cal D}_n}^*$ by definition of the infimum, we have
\begin{equation}
	\label{sgd:eq:decomposition}
	{\cal R}(f_n) - {\cal R}(f^*)
	\leq \underbrace{{\cal R}(f_n) - {\cal R}_{{\cal D}_n}(f_n)
	+ {\cal R}_{{\cal D}_n}(f^*) - {\cal R}(f^*)}_{\text{estimation error}}
	+ \underbrace{{\cal R}_{{\cal D}_n}(f_n) - {\cal R}_{{\cal D}_n}^*}_{\text{optimization error}}.
\end{equation}

\paragraph{Estimation error.}
Let us begin by controlling the estimation error.
We have two terms in it.
${\cal R}_{{\cal D}_n}(f^*) - {\cal R}(f^*)$ can be controlled with a concentration inequality on the empirical average of $\norm{f^*(X) - Y}$ around its population mean.
Assuming sub-Gaussian moments of $Y$, it can be done with Bernstein inequality.

${\cal R}_{{\cal D}_n}(f_n) - {\cal R}(f_n)$ is harder to control as $f_n$ depends on ${\cal D}_n$, so we can not use the same technique.
The classical technique consists in going for the brutal uniform majoration,
\begin{equation}
	\label{sgd:eq:sup_rade}
	{\cal R} (f_n) - {\cal R}_{{\cal D}_n}(f_n) \leq
	\sup_{f\in{\cal F}} \paren{{\cal R} (f) - {\cal R}_{{\cal D}_n}(f)},
\end{equation}
where ${\cal F}$ denotes the set of functions that $f_n$ could be in concordance with our algorithm.
While this bound could seem highly suboptimal, when the class of functions is well-behaved, we can indeed control the deviation ${\cal R}(f) - {\cal R}_{{\cal D}_n}(f)$ uniformly over this class without losing much (indeed for any class of functions, it is possible to build some really adversarial distribution $\rho$ so that this supremum behaves similarly to the concentration we are looking for \citep{Vapnik1995,Anthony1999}).
This is particularly the case for our model linked with Assumption \ref{sgd:ass:source}.
Expectations of supremum processes have been extensively studied, allowing to get satisfying upper bounds (note that when the $\norm{f(X) - Y}$ is bounded, deviation of the quantity of interest around its expectation can be controlled through McDiarmid inequality).
In the statistical learning literature, it is usual to proceed with Rademacher complexity.

\begin{lemma}[Uniform control of functions deviation with Rademacher complexity]
	The expectation of the excess of risk can be bounded as
	\begin{equation}
		\label{sgd:eq:rademacher}
		\frac{1}{2}\E_{{\cal D}_n} \bracket{\sup_{f\in{\cal F}} \paren{{\cal R} (f) - {\cal R}_{{\cal D}_n}(f)}}
		\leq \mathfrak{R}_n({\cal F}, \ell, \rho) :=
		\frac{1}{n} \E_{{\cal D}_n, (\sigma_i)}\bracket{ \sup_{f\in{\cal F}}\sigma_i \ell(f(X_i), Y_i)},
	\end{equation}
	where $(\sigma_i)_{i\leq n}$ is defined as a family of Bernoulli independent variables taking value one or minus one with equal probability, and $\mathfrak{R}_n({\cal F}, \ell, \rho)$ is called Rademacher complexity.
\end{lemma}

\begin{proof}
	This results from the reduction to larger supremum and a symmetrization trick,
	\begin{align*}
		\E_{{\cal D}_n} \bracket{\sup_{f\in{\cal F}} \paren{{\cal R} (f) - {\cal R}_{{\cal D}_n}(f)}}
		 & = \E_{{\cal D}_n} \bracket{\sup_{f\in{\cal F}} \paren{\E_{{\cal D}_n'}{\cal R}_{{\cal D}_n'} (f) - {\cal R}_{{\cal D}_n}(f)}}
		\\&\leq \E_{{\cal D}_n} \E_{{\cal D}_n'}\bracket{\sup_{f\in{\cal F}} \paren{{\cal R}_{{\cal D}_n'} (f) - {\cal R}_{{\cal D}_n}(f)}}
		\\&= \E_{(X_i, Y_i), (X_i', Y_i')}\bracket{\sup_{f\in{\cal F}} \paren{\frac{1}{n} \sum_{i=1}^n \ell(f(X_i'), Y_i') - \ell(f(X_i), Y_i)}}
		\\&= \E_{(X_i, Y_i), (X_i', Y_i'), (\sigma_i)}\bracket{\sup_{f\in{\cal F}} \paren{\frac{1}{n} \sum_{i=1}^n \sigma_i\paren{\ell(f(X_i'), Y_i') - \ell(f(X_i), Y_i)}}}
		\\&\leq 2 \E_{(X_i, Y_i), (\sigma_i)}\bracket{\sup_{f\in{\cal F}} \paren{\frac{1}{n} \sum_{i=1}^n \sigma_i\paren{\ell(f(X_i), Y_i)}}},
	\end{align*}
	which ends the proof.
\end{proof}

In our case, we want to compute the Rademacher complexity for $\ell$ given by the norm of $\Y$, and ${\cal F} = \brace{x\to \theta \phi(x)\midvert \theta \in \Y\otimes{\cal H}, \norm{\theta} < M}$, for $M > 0$ a parameter to specify in order to make sure that $\norm{\theta^*} < M$, where the norm has to be understood as the $\ell^2$-product norm on $\Y \otimes{\cal H} \simeq {\cal H}^m$.
Working with linear models and Lipschitz losses is a well-known setting, allowing to derive directly the following bound.

\begin{lemma}[Rademacher complexity of linear models with Lipschitz losses]
	The complexity of the linear class of vector-valued function ${\cal F} = \brace{x\to \theta \phi(x)\midvert \theta \in \Y\otimes{\cal H}, \norm{\theta} < M}$ is bounded as
	\begin{equation}
		\E_{(\sigma_i)}\bracket{\sup_{f\in{\cal F}}\paren{\frac{1}{n} \sum_{i=1}^n \sigma_i \norm{f(x_i) - y_i}}}
		\leq M \kappa n^{-1/2}.
	\end{equation}
\end{lemma}
\begin{proof}
	This proposition is usually split in two.
	First using the fact that the composition of a space of functions with a Lipschitz function does not increase the entropy of the subsequent space \citep{Vitushkin1954}.
	Then bounding the Rademacher complexity of linear models.
	We refer to \cite{Maurer2016} for a self-contained proof of this result (stated in its Section 4.3).
\end{proof}

Adding all the pieces together we have proven the following proposition, using the fact that the previous bound also applies to $\sup_{f\in{\cal F}} {\cal R}_{{\cal D}_n}(f) - {\cal R}(f)$ by symmetry, hence it can be used for the deviation of ${\cal R}_{{\cal D}_n}(f^*) - {\cal R}(f^*)$.

\begin{proposition}[Control of the estimation error]
	Under Assumption \ref{sgd:ass:source}, with the model of computation ${\cal F} = \brace{x\in\X\to \theta\phi(x) \in \Y \midvert \norm{\theta} \leq M}$, the generalization error of $f_n$ is controlled by a term in $n^{-1/2}$ plus an optimization error on the empirical risk minimization
	\begin{equation}
		\E_{{\cal D}_n}\bracket{{\cal R}(f_n) - {\cal R}(f^*)}
		\leq \frac{4 M\kappa}{n^{1/2}} + \E_{{\cal D}_n}\bracket{{\cal R}_{{\cal D}_n}(f_n) - {\cal R}_{{\cal D}_n}^*},
	\end{equation}
	as long as $f^* \in {\cal F}$.
\end{proposition}

Note that this result can be refined using regularized risk \citep{Sridharan2008}, which would be useful under richer (stronger or weaker) source assumptions \citep[\emph{e.g.},][]{Caponnetto2007}.
Such a refinement would allow switching from a constraint $\norm{\theta} < M$ to define ${\cal F}$ to a regularization parameter $\lambda\norm{\theta}^2$ added in the risk without restrictions on $\norm{\theta}$, which would be better aligned with the current practice of machine learning.
Under Assumption \ref{sgd:ass:source}, this will not fundamentally change the result.
The estimation error can be controlled with the derivation in Appendix \ref{sgd:proof:sgd}, where stochastic gradients correspond to random sampling of a coefficient $i_t\leq n$ plus the choice of a random $U_t$.
For the option without resampling, there exists an acceleration scheme specific to different losses in order to benefit from the strong convexity \citep[\emph{e.g.},][]{Bach2013}.

\subsection{Lower bound}
\label{sgd:proof:lower}

In this section, we prove Theorem \ref{sgd:thm:minmax_opt}.
Let us consider any algorithm ${\cal A}:\cup_{n\in\N}(\X\times\Y)^n \to \Theta$ that matches a dataset ${\cal D}_n$ to an estimate $\theta_{{\cal D}_n}\in\Theta$.
Let us consider jointly a distribution $\rho$ and a parameter $\theta$ such that Assumption \ref{sgd:ass:source} holds, that is $f_\rho := \argmin_{f:\X\to\Y} \E_\rho[\ell(f(X), Y)] = f_\theta$.
We are interested in characterizing for each algorithm the worst excess of risk it can achieve with respect to an adversarial distribution.
The best worst performance that can be achieved by algorithms matching datasets to parameter can be written as
\begin{equation}
	{\cal E} = \inf_{\cal A}\sup_{\theta\in\Theta, \rho\in\prob{\X\times\Y}; f_{\rho} = f_{\theta}} \E_{{\cal D}_n \sim \rho^{\otimes n}}\bracket{\E_{(X, Y)\sim\rho}\bracket{\ell(f_{{\cal A}({\cal D}_n)}(X), Y) - \ell(f_{\theta}(X), Y)}}.
\end{equation}
This provides a lower bound to upper bounds such as~\eqref{sgd:eq:thm} that can be derived for any algorithm.
There are many ways to get lower bounds on this quantity.
Ultimately, we want to quantify the best certainty one can have on an estimate $\theta$ based on some observations $(X_i, Y_i)_{i\leq n}$.
In particular, the algorithms ${\cal A}$ can be seen as rules to discriminate a model $\theta$ from observations ${\cal D}_n$ made under $\rho_\theta$, and where the error is measured through the excess of risk ${\cal R}(f_{\hat\theta}, \rho_\theta) - {\cal R}(f_\theta; \rho_\theta)$ where ${\cal R}(f; \rho) = \E_\rho[\ell(f(X), Y)]$ and $\rho_\theta$ is a distribution parametrized by $\theta$ such that $f_\theta = f_\rho$.

Let us first characterize the measure of error.
Surprisingly, when in presence of Gaussian noise or uniform noise, the excess of risk behaves like a quadratic metric between parameters.

\begin{lemma}[Quadratic behavior of the median regression excess of risk with Gaussian noise]
	Consider the random variable $Y \sim {\cal N}(\mu, \sigma^2 I_m)$, denote by $\hat{\mu}$ an estimate of $\mu$, the excess of risk can be developed as
	\begin{equation}
		\E_{{\cal N}(\mu, \sigma^2 I_m)}[\norm{\hat{\mu} - Y} - \norm{\mu - Y}]
		= \frac{c_4\norm{\hat{\mu} - \mu}^2}{\sigma}
		+ o\paren{\frac{\norm{\hat{\mu} - \mu}^3}{\sigma^2}},
	\end{equation}
	where $c_4 = \Gamma(\frac{m+1}{2}) / (2\sqrt{2}\Gamma(\frac{m+2}{2})) \geq (m+2)^{-1/2} / 2$.
\end{lemma}
\begin{proof}
	With this specific noise model, one can do the following derivations.
	\[
		\E_{{\cal N}(\mu, \sigma^2 I_m)}[\norm{\hat{\mu} - Y}]
		= \E_{{\cal N}(0, I_m)} [\norm{\hat{\mu} - \mu - \sigma Y}]
		= \sigma \E_{{\cal N}(0, I_m)} \bracket{\norm{\frac{\hat{\mu} - \mu}{\sigma} - Y}}.
	\]
	We recognize the mean of a non-central $\chi$-distribution of parameter $k=m$ and $\lambda = \norm{\frac{\hat{\mu} - \mu}{\sigma}}$.
	It can be expressed through the generalized Laguerre functions, which allows us to get the following Taylor expansion
	\begin{align*}
		\E_{{\cal N}(\mu, \sigma^2 I_m)}[\norm{\hat{\mu} - Y}]
		 & = \frac{\sqrt{\pi}\sigma}{\sqrt{2}} L_{\frac{1}{2}}^{(\frac{m-2}{2})}\paren{-\frac{\norm{\hat{\mu} - \mu}^2}{2\sigma^2}}
		\\&= \frac{\sqrt{\pi} \sigma}{\sqrt{2}} \paren{L_{\frac{1}{2}}^{(\frac{m-2}{2})}(0) + \frac{\norm{\hat{\mu} - \mu}^2}{2\sigma^2} L_{-\frac{1}{2}}^{(\frac{m}{2})}(0)}
		+ o\paren{\frac{\norm{\hat{\mu} - \mu}^3}{\sigma^2}}.
	\end{align*}
	Hence, the following expression of the excess of risk,
	\begin{align*}
		\E_{{\cal N}(\mu, \sigma^2 I_m)}[\norm{\hat{\mu} - Y} - \norm{\mu - Y}]
		 & = \frac{\sqrt{\pi}\norm{\hat{\mu} - \mu}^2}{2\sqrt{2}\sigma} L_{-\frac{1}{2}}^{(\frac{m}{2})}(0)
		+ o\paren{\frac{\norm{\hat{\mu} - \mu}^3}{\sigma^2}}
		\\&= \frac{\Gamma(\frac{m+1}{2})\norm{\hat{\mu} - \mu}^2}{2\sqrt{2}\Gamma(\frac{m+2}{2})\sigma}
		+ o\paren{\frac{\norm{\hat{\mu} - \mu}^3}{\sigma^2}}.
	\end{align*}
	Note that in dimension one, the calculation can be done explicitly by computing integrals with the error function.
	\begin{align*}
		\E_{{\cal N}(\mu, \sigma^2)}[\norm{\hat{\mu} - Y}]
		 & = \sigma \E_{{\cal N}(0, 1)} \bracket{Y - \frac{\hat{\mu} - \mu}{\sigma} + 2\ind{Y < \frac{\hat{\mu} - \mu}{\sigma}} \paren{\frac{\hat{\mu} - \mu}{\sigma} - Y}}
		\\&= \mu - \hat{\mu} + 2(\hat{\mu} - \mu) \E_{{\cal N}(0, 1)}\bracket{\ind{Y < \frac{\hat{\mu} - \mu}{\sigma}}}
		- 2\sigma \E_{{\cal N}(0, 1)}\bracket{Y\ind{Y < \frac{\hat{\mu} - \mu}{\sigma}}}
		\\&= \mu - \hat{\mu} + 2(\hat{\mu} - \mu) \paren{\frac{1}{2} + \frac{1}{2}\operatorname{erf}\paren{\frac{\hat{\mu} - \mu}{\sqrt{2}\sigma}}}
		- \frac{\sqrt{2}\sigma}{\sqrt{\pi}} \int_{-\infty}^{\frac{\hat{\mu} - \mu}{\sigma}} y e^{-\frac{y^2}{2}}\diff y
		\\&= (\hat{\mu} - \mu) \operatorname{erf}\paren{\frac{\hat{\mu} - \mu}{\sqrt{2}\sigma}}
		- \frac{\sqrt{2}\sigma}{\sqrt{\pi}} e^{-\frac{(\hat{\mu} - \mu)^2}{2\sigma^2}},
	\end{align*}
	where we used the error function, which is the symmetric function defined for $x\in\R_+$ as
	\[
		\operatorname{erf}(x) = \frac{2}{\sqrt{\pi}} \int_0^x e^{-t^2} \diff t
		= \frac{2}{\sqrt{2 \pi}} \int_0^{\sqrt{2}x} e^{-\frac{u^2}{2}} \diff u
		= 2 \E_{{\cal N}(0, 1)}[\ind{0 \leq Y \leq \sqrt{2} x}].
	\]
	Developing those two functions in the Taylor series leads to the same quadratic behavior.
\end{proof}

Let us now add a context variable.

\begin{lemma}[Reduction to least-squares]
	For $\Y=\R^m$, there exists a $\sigma_m > 0$, such that if $\phi$ is bounded by $\kappa$, and $f^*$ belongs to the class of functions ${\cal F} = \brace{x\to \theta\phi(x)\midvert \theta \in \Y\otimes{\cal H}, \norm{\theta}\leq M}$, and the conditional distribution are distributed as $\paren{Y\midvert X} \sim {\cal N}(f^*(x), \sigma^2 I_m)$, with $\sigma > 2M\kappa\sigma_m$,
	\begin{equation}
		\forall\, f\in{\cal F}, \qquad{\cal R}(f) - {\cal R}(f^*)
		\geq \frac{c_4\norm{f - f^*}_{L^2(\rho_\X)}^2}{2\sigma}.
	\end{equation}
\end{lemma}
\begin{proof}
	According to the precedent lemma, there exists $\sigma_m$ such that $\norm{\hat{\mu} - \mu}\sigma^{-1} \leq \sigma_m^{-1}$ leads to\footnote{This best value for $\sigma_m$ can be derived by studying the Laguerre function, which we will not do in this paper.}
	\[
		\E_{{\cal N}(\mu, \sigma^2 I_m)}[\norm{\hat{\mu} - Y} - \norm{\mu - Y}]
		\geq \frac{c_4\norm{\hat{\mu} - \mu}^2}{2\sigma}.
	\]
	Let $f$ and $f^*\in{\cal F}$ be parametrized by $\theta$ and $\theta^*$.
	For a given $x$, setting $\hat\mu = f_\theta(x) = \theta\phi(x)$ and $\mu = f_{\theta^*}(x)$, we get that, using the operator norm,
	\[
		\norm{\hat{\mu} - \mu} = \norm{(\theta - \theta^*)\phi(x)}
		\leq \norm{\theta - \theta^*}_{\op}\norm{\phi(x)}
		\leq \norm{\theta - \theta^*}\norm{\phi(x)}
		\leq 2M\kappa.
	\]
	Hence, as soon as $2M\kappa \leq \sigma\sigma_m^{-1}$, we have that for almost all $x\in\X$
	\[
		\E_Y\bracket{\norm{f(X) - Y} - \norm{f^*(X) - Y}\midvert X=x}
		\geq \frac{c_4\norm{f(X) - f^*(X)}^2}{2\sigma}.
	\]
	The result follows from integration over $\X$.
\end{proof}

We now have a characterization of the excess of risk that will allow us to reuse lower bounds for least-squares regression.
We will follow the exposition of \cite{Bach2023} that we reproduce and comment here for completeness.
It is based on the generalized Fano's method \citep{Ibragimov1977,Birge1983}.

Learnability over a class of functions depends on the size of this class of functions.
For least-squares regression with a Hilbert class of functions, the right notion of size is given by the Kolmogorov entropy.
Let us call $\epsilon$-packing of ${\cal F}$ with a metric $d$ any family $(f_i)_{i\leq N}\in {\cal F}^N$ such that $d(f_i, f_j) > \epsilon$.
The logarithm of the maximum cardinality of an $\epsilon$-packing defines the $\epsilon$-capacity of the class of functions ${\cal F}$.
We refer the interested reader to Theorem~6 in \cite{Kolmogorov1959} to make a link between the notions of capacity and entropy of a space.
To be perfectly rigorous, the least-squares error in not a norm on the space of $L^2$ functions, but we will call it a {\em quasi-distance} as it verifies symmetry, positive definiteness and the inequality $d(x, y) \leq K(d(x, z) + d(z, y))$ for $K \geq 1$.
Let us define an $\epsilon$-packing with respect to a quasi-distance similarly as before.

The $\epsilon$-capacity of a space ${\cal F}$ gives a lower bound on the number of information to transmit in order to recover a function in ${\cal F}$ up to precision $\epsilon$.
We will leverage this fact in order to show our lower bound.
Let us first reduce the problem to a statistical test.

\begin{lemma}[Reduction to statistical testing]
	Let us consider a class of functions ${\cal F}$ and an $\epsilon$-packing $(f_i)_{i\leq N}$ of ${\cal F}$ with respect to a quasi-distance $d(\cdot, \cdot)$ verifying the triangular inequality up to a multiplicative factor $K$.
	Then the minimax optimality of an algorithm ${\cal A}$ that takes as input the dataset ${\cal D}_n = (X_i, Y_i)_{i \leq n}$ and output a function in ${\cal F}$ can be related to the minimax optimality of an algorithm ${\cal C}$ that takes an input the dataset ${\cal D}_n$ and output an index $j\in[m]$ through
	\begin{equation}
		\inf_{\cal A} \sup_{\rho} \E_{{\cal D}_n\sim\rho^{\otimes n}}\bracket{d\paren{f_{{\cal A}({\cal D}_n)}, f_\rho}}
		\geq \frac{\epsilon}{2K}
		\inf_{\cal C} \sup_{i\in[N]} \Pbb_{{\cal D}_n\sim(\rho_i)^{\otimes n}}\paren{{{\cal C}({\cal D}_n)} \neq i},
	\end{equation}
	where the supremum over $\rho$ has to be understood as taken over all measures whose marginals can be written ${\cal N}(f^*(x), \sigma)$ for $\sigma$ bigger than a threshold $\sigma_m$ and $f^*\in{\cal F}$, and the supremum over $\rho_i$ taken over the same type of measures with $f^* \in (f_i)_{i\leq N}$.
\end{lemma}
\begin{proof}
	Consider an algorithm ${\cal A}$ that takes as input a dataset ${\cal D}_n = (X_j, Y_j)_{j\leq n}$ and output a function $f\in{\cal F}$.
	We would like to see ${\cal A}$ as deriving from a classification rule and relate the classification and regression errors.
	The natural classification rule associated with the algorithm ${\cal A}$ can be defined through $\pi$ the projection from ${\cal F}$ to $[N]$ that minimizes $d\paren{f, f_{\pi(f)}}$.
	The classification error and regression error made by $\pi\circ{\cal A}$ can be related thanks to the $\epsilon$-packing property.
	For any index $j\in[N]$
	\[
		d\paren{f_{\pi\circ{\cal A}({\cal D}_n)}, f_j} \geq \epsilon \ind{\pi\circ{\cal A}({\cal D}_n) \neq j}.
	\]
	The error made by $f_{\cal A}({\cal D}_n)$ relates to the one made by $f_{\pi\circ{\cal A}({\cal D}_n)}$ thanks to the modified triangular inequality, using the definition of the projection
	\[
		d\paren{f_{\pi\circ{\cal A}({\cal D}_n)}, f_j}
		\leq K \paren{d\paren{f_{\pi\circ{\cal A}({\cal D}_n)}, f_{{\cal A}({\cal D}_n)}}
			+ d\paren{f_{{\cal A}({\cal D}_n)}, f_j}}
		\leq 2 Kd\paren{f_{{\cal A}({\cal D}_n)}, f_j}.
	\]
	Finally,
	\[
		d\paren{f_{{\cal A}({\cal D}_n)}, f_{j}} \geq \frac{\epsilon}{2K} \ind{\pi\circ{\cal A}({\cal D}_n) \neq j}.
	\]
	Assuming that the data were generated by a $\rho_i$ and taking the expectation, the supremum over $\rho_i$ and the infimum over ${\cal A}$ leads to
	\[
		\inf_{\cal A}\sup_{\rho_i} \E_{{\cal D}_n\sim\rho_i^{\otimes n}}\bracket{d\paren{f_{{\cal A}({\cal D}_n)}, f_i}}
		\geq \frac{\epsilon}{2K}\inf_{{\cal C}=\pi\circ{\cal A}}\sup_{(\rho_i)} \Pbb_{{\cal D}_n\sim\rho_i^{\otimes n}}\paren{{\cal C}({\cal D}_n) \neq i}.
	\]
	Because $\pi\circ{\cal A}$ are part of classification rules (indeed it parametrizes all the classification rules, simply consider ${\cal A}$ that matches a dataset to one of the functions $(f_i)_{i\leq N}$), and because the distributions $\rho_i$ are part of the distributions $\rho$ defined in the lemma, this last equation implies the stated result.
\end{proof}

One of the harshest inequalities in the last proof is due to the usage of the $\epsilon$-packing condition without considering error made by $d\paren{f_{\pi\circ{\cal A}({\cal D}_n)}, f_j}$ that might be much worse than $\epsilon$.
We will later add a condition on the $\epsilon$-packings to ensure that the $(f_i)$ are not too far from each other.
This will not be a major problem when considering small balls in big dimension spaces.

\subsubsection{Results from statistical testing}
In this section, we expand on lower bounds for statistical testing.
We refer the curious reader to \cite{Cover1991}.
We begin by relaxing the supremum by an average
\begin{align}
	\inf_{\cal C} \sup_{i\in[N]} \Pbb_{{\cal D}_n\sim(\rho_i)^{\otimes n}}\paren{{{\cal C}({\cal D}_n)} \neq i}
	 & = \inf_{\cal C} \sup_{p\in\prob{N}} \sum_{i=1}^N p_i \Pbb_{{\cal D}_n\sim(\rho_i)^{\otimes n}}\paren{{{\cal C}({\cal D}_n)} \neq i}
	\\&\geq \inf_{\cal C} \frac{1}{N}\sum_{i=1}^N \Pbb_{{\cal D}_n\sim(\rho_i)^{\otimes n}}\paren{{{\cal C}({\cal D}_n)} \neq i}.
\end{align}
The last quantity can be seen as the best measure of error that can be achieved by a decoder ${\cal C}$ of a signal $i\in[N]$ based on noisy observations ${\cal D}_n$ of the signal.
A lower bound on such a similar quantity is the object of Fano's inequality \citep{Fano1968}.

\begin{lemma}[Fano's inequality]
	Let $(X, Y)$ be a couple of random variables in $\X\times\Y$ with $\X$, $\Y$ finite, and $\hat{X}:\Y\to\X$ be a classification rule.
	Then, the error $e = e(X, Y) = \ind{X \neq \hat{X}(Y)}$ verifies
	\[
		H\paren{X\midvert Y} \leq H(e) + \Pbb(e) \log(\card{\X} - 1) \leq \log(2) + \Pbb(e)\log(\card\X).
	\]
	Where for $(X, Y)\in\prob{\X\times\Y}$, $H(X)$ and $H\paren{X\midvert Y}$ denotes the entropy and conditional entropy, defined as, with the convention $0\log 0 = 0$,
	\begin{align*}
		 & H(X) = -\sum_{x\in\X} \Pbb(X=x) \log(\Pbb(X=x)),
		\\ &H\paren{X\midvert Y} = -\sum_{x\in\X,y\in\Y} \Pbb(X=x, Y=y) \log(\Pbb\paren{X=x\midvert Y=y}).
	\end{align*}
\end{lemma}

\begin{proof}
	This lemma is actually the result of two properties.
	The first part of the proof is due to some manipulation of the entropy, consisting in showing that
	\begin{equation}
		\label{sgd:eq:proc_1}
		H\paren{X\midvert \hat{X}(Y)} \leq H(e) + \Pbb(e)\log(\card\X - 1).
	\end{equation}
	Let us first recall the following additive property of entropy
	\begin{align*}
		H\paren{X, Y\midvert Z} & = -\sum_{x\in\X, y\in\Y, z\in{\cal Z}} \Pbb(X=x, Y=y, Z=z)\log(\Pbb\paren{X=x,Y=y\midvert Z=z})
		\\&= -\sum_{x\in\X, y\in\Y, z\in{\cal Z}} \Pbb(X=x, Y=y, Z=z)\log(\Pbb\paren{Y=y\midvert X=x, Z=z})
		\\&\qquad\qquad\qquad\qquad - \sum_{x\in\X, y\in\Y, z\in{\cal Z}} \Pbb(X=x, Y=y, Z=z)\log(\Pbb\paren{X=x\midvert Z=z})
		\\&= H\paren{Y\midvert X, Z} + H\paren{X\midvert Z}.
	\end{align*}
	Using this chain rule, we get
	\begin{align*}
		H\paren{e, X\midvert \hat{X}} & = H\paren{e\midvert X, \hat{X}} + H\paren{X\midvert\hat{X}}
		\\&= H\paren{X\midvert e, \hat{X}} + H\paren{e\midvert\hat{X}}
	\end{align*}
	Because $e$ is a function of $\hat{X}$ and $X$ one can check that $H\paren{e\midvert X, \hat{X}} = 0$,
	\begin{align*}
		H\paren{e\midvert X,\hat{X}}
		 & = -\sum_{e, X, \hat{X}} \Pbb(X, \hat{X}) \Pbb\paren{e\midvert X,\hat{X}} \log(\Pbb\paren{e\midvert X,\hat{X}})
		\\&= -\sum_{e, X, \hat{X}} \Pbb(X, \hat{X}) \ind{e=\ind{X\neq\hat{X}}}\log(\ind{e=\ind{X\neq\hat{X}}})
		= -\sum_{e, X,\hat{X}} \Pbb(X, \hat{X}) \cdot 0 = 0.
	\end{align*}
	Using Jensen inequality for the logarithm, we get
	\begin{align*}
		H\paren{X\midvert e, \hat{X}}
		 & = -\sum_{X, e, \hat{X}} \Pbb(X, e, \hat{X}) \log(\Pbb\paren{X\midvert e, \hat{X}})
		\\&= -\sum_{x, x'} \Pbb(X=x, e=0, \hat{X}=x') \log(\Pbb\paren{X=x\midvert e=0, \hat{X}=x'})
		\\&\qquad\qquad\qquad\qquad- \Pbb(X=x, e=1, \hat{X}=x') \log(\Pbb\paren{X=x\midvert e=1, \hat{X}=x'})
		\\&= -\sum_{x, x'} \Pbb\paren{X = x, \hat{X}=x'}\ind{x=x'} \log(\ind{x=x'})
		\\&\qquad\qquad\qquad\qquad- \Pbb(e=1)\ind{x\neq x'}\Pbb(X=x, \hat{X}=x') \log(\Pbb\paren{X=x\midvert \hat{X}=x'})
		\\&= \Pbb(e=1)\sum_{x'} \Pbb(\hat{X}=x') \sum_{x\neq x'} \Pbb\paren{X=x\midvert \hat{X}=x'} \log\paren{\frac{1}{\Pbb\paren{X=x\midvert \hat{X}=x'}}}
		\\&\leq \Pbb(e=1)\sum_{x'} \Pbb(\hat{X}=x') \log\paren{\sum_{x\neq x'} \Pbb\paren{X=x\midvert \hat{X}=x'} \frac{1}{\Pbb\paren{X=x\midvert \hat{X}=x'}}}
		\\&= \Pbb(e=1)\log(\card{\X} - 1).
	\end{align*}
	Using that conditioning reduces the entropy, which follows again from Jensen inequality,
	\begin{align*}
		H(X) - H\paren{X\midvert Y}
		 & = \sum_{x, y} \Pbb(X=x,Y=y) \log\paren{\frac{\Pbb\paren{X=x\midvert Y=y}}{\Pbb\paren{X=x}}}
		\\&= -\sum_{x, y} \Pbb(X=x,Y=y) \log\paren{\frac{\Pbb\paren{X=x}\Pbb\paren{Y=y}}{\Pbb\paren{X=x, Y=y}}}
		\\&\geq -\log\paren{\sum_{x, y} \Pbb(X=x,Y=y) \frac{\Pbb\paren{X=x}\Pbb\paren{Y=y}}{\Pbb\paren{X=x, Y=y}}} = 0,
	\end{align*}
	we get
	\[
		H\paren{e \midvert \hat{X}} \leq H(e) \leq \log(2).
	\]
	Hence, we have proven that
	\[
		H\paren{X\midvert \hat{X}} \leq \Pbb(e=1)\log(\card\X - 1) + H(e).
	\]

	The rest of the proof follows from the so-called data processing inequality, that is
	\begin{equation}
		H\paren{X\midvert \hat{X}(Y)} \geq H\paren{X\midvert Y}.
	\end{equation}
	We will not derive it here, since it will not be used in the following.
\end{proof}

In our case, a slight modification of the proof of Fano's inequality leads to the following Proposition.
\begin{lemma}[Generalized Fano's method]
	For any family of distributions $(\rho_i)_{i\leq N}$ on $\X\times\Y$ with $N\in\N^*$, any classification rule ${\cal C}:{\cal D}_n \to [N]$ cannot beat the following average lower bound
	\begin{equation}
		\inf_{\cal C} \frac{1}{N}\sum_{i=1}^N \Pbb_{{\cal D}_n \sim \rho_i^{\otimes n}}({\cal C}({\cal D}_n) \neq i) \log(N - 1) \geq \log(N) - \log(2) - \frac{n}{N^2} \sum_{i, j \in [N]} K\paren{\rho_i\midvertvert \rho_j},
	\end{equation}
	where $K\paren{p\midvertvert q}$ is the Kullback-Leibler divergence defined for any measure $p$ absolutely continuous with respect to a measure $q$ as
	\[
		K\paren{p\midvertvert q} = \E_{X\sim q}\bracket{-\log\paren{\frac{\diff p(X)}{\diff q(X)}}}.
	\]
\end{lemma}
\begin{proof}
	Let us consider the joint variable $(X, Y)$ where $X$ is a uniform variable on $[N]$ and $\paren{Y\midvert X}$ is distributed according to $\rho_X^{\otimes n}$.
	For any classification rule $\hat{X}:{\cal D}_n\to[N]$, using~\eqref{sgd:eq:proc_1} we get
	\[
		\frac{1}{N}\sum_{i=1}^N \Pbb_{{\cal D}_n\sim\rho_i^{\otimes n}}\paren{\hat{X}({\cal D}_n) \neq i} = \Pbb(\hat{X} \neq X)\log(N-1) \geq H\paren{X\midvert \hat{X}} - \log(2).
	\]

	We should work on $H\paren{X\midvert \hat{X}\midvert X}$ with similar ideas to the data processing inequality.
	First of all, using the chain rule for entropy
	\[
		H\paren{X\midvert \hat{X}} = H(X, \hat{X}) - H(\hat{X}) = H(X) + (H(X, \hat{X}) - H(X) - H(\hat{X})) = \log(N) - I(X,\hat{X}),
	\]
	where $I$ is the mutual information defined as, for $X$ and $Z$ discrete
	\begin{align*}
		I(X, Z) & = H(X) + H(Z) - H(X, Z)
		= \sum_{x, z} \Pbb\paren{X=x, Z=z} \log\paren{\frac{\Pbb(X=x, Z=z)}{\Pbb(X=x)\Pbb(Z=z)}}
		\\&= \sum_x \Pbb(X=x) \sum_z \Pbb\paren{Z=z\midvert X=x} \log\paren{\frac{\Pbb\paren{Z=z\midvert X=x)}}{\Pbb(Z=z)}}.
	\end{align*}
	Similarly, one can define the mutual information for continuous variables.
	In particular, we are interested in the case where $X$ is discrete and $Y$ is continuous, denote by $\mu_\Y$ the marginal of $(X, Y)$ over $Y$ and by $\mu\vert_x$ the conditional $\paren{Y\midvert X=x}$.
	\[
		I(X, Y) = \sum_{x}\Pbb(X=x)\int_{y} \mu\vert_x(\diff y) \log\paren{\frac{\mu\vert_x(\diff y)}{\mu(\diff y)}}.
	\]

	Let us show the following version of the data processing inequality
	\begin{equation}
		I(X, \hat{X}(Y)) \leq I(X, Y).
	\end{equation}
	To do so, we will use the conditional independence of $X$ and $\hat{X}$ given $Y$, which leads to
	\begin{align*}
		\Pbb\paren{X=x\midvert \hat{X}=x'}
		 & = \int \Pbb\paren{X=x\midvert Y=\diff y}\Pbb\paren{Y=\diff y\midvert \hat{X}=z}
		\\&= \int \frac{\Pbb(X=x)\mu\vert_x(\diff y)}{\mu(\diff y)}\Pbb\paren{Y=\diff y\midvert \hat{X}=z}.
	\end{align*}
	Hence, using Jensen inequality,
	\begin{align*}
		I(X, \hat{X})
		 & = H(X) - H\paren{X\midvert \hat{X}}
		\\&= H(X) + \sum_{z} \Pbb(\hat{X}=z) \sum_{x} \Pbb(X=x) \log(\Pbb\paren{X=x\midvert \hat{X}=z})
		\\&= H(X) + \sum_{z} \Pbb(\hat{X}=z) \sum_{x} \Pbb(X=x) \log\paren{\int \frac{\Pbb(X=x)\mu\vert_x(\diff y)}{\mu(\diff y)}\Pbb\paren{Y=\diff y\midvert \hat{X}=z}}
		\\&\leq H(X) + \sum_{z} \Pbb(\hat{X}=z) \sum_{x} \Pbb(X=x)\int \Pbb\paren{Y=\diff y\midvert \hat{X}=z} \log\paren{\frac{\Pbb(X=x)\mu\vert_x(\diff y)}{\mu(\diff y)}}
		\\&= H(X) + \sum_{x} \Pbb(X=x) \int \mu(\diff y) \log\paren{\frac{\Pbb(X=x)\mu\vert_x(\diff y)}{\mu(\diff y)}}
		\\&=\sum_{x} \Pbb(X=x) \paren{\int \mu(\diff y) \log\paren{\frac{\Pbb(X=x)\mu\vert_x(\diff y)}{\mu(\diff y)}} - \log(P(X=x)}
		\\&=\sum_{x} \Pbb(X=x) \int \mu(\diff y) \log\paren{\frac{\mu\vert_x(\diff y)}{\mu(\diff y)}}
		\\&= I(X, Y).
	\end{align*}

	We continue by computing the value of $I(X, Y)$, by definition and using Jensen inequality, we get
	\begin{align*}
		I(X, Y) & = \frac{1}{N}\sum_{i\in[N]} \int_{{\cal D}_n\sim\rho_i^{\otimes n}} \rho_i^{\otimes n}(\diff {\cal D}_n) \log\paren{\frac{\rho_i^{\otimes n}(\diff {\cal D}_n)}{\frac{1}{N}\sum_{j\in[N]} \rho_j^{\otimes n}(\diff {\cal D}_n)}}
		\\&\leq \frac{1}{N}\sum_{i\in[N]} \int_{{\cal D}_n\sim\rho_i^{\otimes n}} \rho_i^{\otimes n}(\diff {\cal D}_n) \frac{1}{N}\sum_{j\in[N]}\log\paren{\frac{\rho_i^{\otimes n}(\diff {\cal D}_n)}{\rho_j^{\otimes n}(\diff {\cal D}_n)}}
		= \frac{1}{N^2}\sum_{i,j \in[N]} K\paren{\rho_i^{\otimes n}\midvertvert\rho_j^{\otimes n}}.
	\end{align*}
	We conclude from the fact that for $p$ and $q$ two distributions on a space ${\cal Z}$, we have
	\begin{align*}
		K\paren{p^{\otimes n}\midvertvert q^{\otimes n}}
		 & = \int_{{\cal Z}^n} -\log\paren{\frac{\diff p^{\otimes n}(z_1,\cdots, z_n)}{\diff q^{\otimes n}(z_1,\cdots, z_n)}} q^{\otimes n}(\diff z_1,\cdots, \diff z_n)
		\\&= \int_{{\cal Z}^n} -\log\paren{\frac{\prod_{i\leq n}\diff p(z_i)}{\prod_{i\leq n}\diff q(z_i)}} q^{\otimes n}(\diff z_1,\cdots, \diff z_n)
		\\&= \sum_{i\leq n}\int_{{\cal Z}^n} -\log\paren{\frac{\diff p(z_i)}{\diff q(z_i)}} q^{\otimes n}(\diff z_1,\cdots, \diff z_n)
		\\&= \sum_{i\leq n}\int_{\cal Z} -\log\paren{\frac{\diff p(z_i)}{\diff q(z_i)}} q(\diff z_i)
		= n K\paren{p\midvertvert q}.
	\end{align*}
	This explains the result.
\end{proof}

Let us assemble all the results proven thus far.
In order to reduce our excess risk to a quadratic metric, we have assumed that the conditional distribution $\rho_i\vert_x$ to be Gaussian noise.
In order to integrate this constraint into the precedent derivations, we leverage the following lemma.

\begin{lemma}[Kullback-Leibler divergence with Gaussian noise]
	If $\rho_i$ and $\rho_j$ are two different distributions on $\X\times\Y$ such that there marginal over $\X$ are equal and the conditional distributions $\paren{Y\midvert X=x}$ are respectively equal to ${\cal N}(f_i(x), \sigma I_m)$ and ${\cal N}(f_j(x), \sigma I_m)$, then
	\[
		K\paren{\rho_i \midvertvert \rho_j} = \frac{1}{2\sigma^2} \norm{f_i - f_j}^2_{L^2(\rho_\X)}.
	\]
\end{lemma}
\begin{proof}
	We proceed with
	\begin{align*}
		K\paren{\rho_i \midvertvert \rho_j}
		 & = \int_\X \E_{Y\sim{\cal N}(f_j(x), \sigma I_m)}\bracket{\frac{\norm{Y-f_i(x)}^2 - \norm{Y-f_j(x)}^2}{2\sigma^2}}\rho_j(\diff x)
		\\&= \int_\X \E_{Y\sim{\cal N}\paren{\frac{f_j(x) - f_i(x)}{\sqrt{2}\sigma}, I_m}}\bracket{\norm{Y}^2} - \E_{Y\sim{\cal N}(0, I_m)}\bracket{\norm{Y}^2}\rho_j(\diff x)
		\\&= \int_\X \paren{m + \frac{\norm{f_j(x) - f_i(x)}^2}{2\sigma^2} - m} \rho_j(\diff x)
		= \frac{\norm{f_j - f_i}^2_{L^2(\rho_\X)}}{2\sigma^2},
	\end{align*}
	where we have used the fact that the mean of a non-central $\chi$-square variable of parameter $(m, \mu^2)$ is $m+\mu^2$.
	One could also develop the first two squared norms and use the fact that for any vector $u\in\R^m$, $\E[\scap{Y-f_i(x)}{u}] = 0$ to get the result.
\end{proof}

Combining the different results leads to the following proposition.
\begin{lemma}
	\label{sgd:prop:step_1}
	Under Assumption \ref{sgd:ass:source} with ${\cal F} = \brace{x\in \X\to \theta\phi(x) \in \Y\midvert \norm{\theta} \leq M}$ and $\phi$ bounded by $\kappa$, for any family $(f_i)_{i\leq N_\epsilon} \in {\cal F}^N$ and any $\sigma > 2M\kappa \sigma_m$
	\begin{align*}
		 & \inf_{\cal A}\sup_\rho \E_{{\cal D}_n\sim\rho^{\otimes n}}[{\cal R}(f_{{\cal A}({\cal D}_n)}; \rho)] - {\cal R}^*(\rho)
		\\&\qquad\qquad\qquad\geq \frac{\min_{i,j\in[N]} \norm{f_i - f_j}_{L^2(\rho_\X)}^2}{16(m+2)^{1/2}\sigma} \paren{1 - \frac{\log(2)}{\log(N)} - \frac{n\max_{i,j\in[N]} \norm{f_i - f_j}^2_{L^2(\rho_\X)}}{2\sigma^2 \log(N)}},
	\end{align*}
	for any algorithm ${\cal A}$ that maps a dataset ${\cal D}_n \in (\X\times\Y)^n$ to a parameter $\theta \in \Theta$.
\end{lemma}

\subsubsection{Covering number for linear model}

We are left with finding a good packing of the space induced by Assumption \ref{sgd:ass:source}.
To do so, we shall recall some property of reproducing kernel methods.

\begin{lemma}[Linear models are ellipsoids]
	\label{sgd:lem:lin_ell}
	For ${\cal H}$ a separable Hilbert space and $\phi:\X\to{\cal H}$ bounded, the class of functions ${\cal F} = \brace{x\in \X\to \theta\phi(x) \in \Y\midvert \norm{\theta} \leq M}$ can be characterized by
	\begin{equation}
		{\cal F} = \brace{f:\X\to\Y\midvert \norm{K^{-1/2}f}_{L^2(\rho_\X)} \leq M},
	\end{equation}
	where $\rho_\X$ is any distribution on $\X$ and $K$ is the operator on $L^2(\rho_\X)$ that map $f$ to
	\[
		Kf(x') = \int_{x\in\X} \scap{\phi(x)}{\phi(x')} f(x) \rho_\X(\diff x),
	\]
	whose image is assumed to be dense in $L^2$.
\end{lemma}
\begin{proof}
	This follows for isometry between elements in ${\cal H}$ and elements in $L^2$.
	More precisely, let us define
	\[\myfunction{S}{\Y\otimes{\cal H}}{L^2(\X, \Y, \rho_\X)}{\theta}{x\to \theta\phi(x).}\]
	The adjoint of $S$ is characterized by
	\[\myfunction{S^*}{L^2(\X, \Y, \rho_\X)}{\Y\otimes{\cal H}}{f}{\E[f(x)\otimes\phi(X)],}\]
	which follows from the fact that for $\theta \in \Y\otimes{\cal H}$, $f\in L^2$ we have
	\begin{align*}
		\scap{\theta}{S^*f}_{\Y\otimes{\cal H}} = \scap{S\theta}{f}_{L^2} & = \sum_{i=1}^m \int_{\X} f_i(x) \scap{\theta_i}{\phi(x)}_{\cal H} \rho_\X(\diff x)
		\\&= \sum_{i=1}^m \scap{\theta_i}{\E[f_i(X) \phi(X)]}_{\cal H}
		= \scap{\theta}{\E[f(X)\otimes\phi(X)]}_{\Y\otimes{\cal H}}.
	\end{align*}
	When $SS^*$ is compact and dense in $L^2$, we have
	\[
		\norm{\theta}_{\Y\otimes{\cal H}} = \norm{(SS^*)^{-1/2} S\theta}_{L^2(\rho_\X)}.
	\]
	The compactness allows considering spectral decomposition hence fractional powers.
	We continue by observing that $SS^* = K$, which follows from
	\begin{align*}
		(SS^*f)(x') = (S\E[f(X)\otimes\phi(X)])(x')
		= \E[f(X)\otimes\phi(X)] \phi(x')
		= \E[\scap{\phi(X)}{\phi(x')} f(X)].
	\end{align*}
	The compactness of $K$ derives from the fact that
	\[
		\norm{Kf(x')}^2 = \norm{\E[\scap{\phi(X)}{\phi(x')} f(X)]}^2
		\leq \E[\norm{\scap{\phi(X)}{\phi(x')} f(X)}^2]
		\leq \kappa^2 \norm{f}_{L^2}^2.
	\]
	Hence, $\norm{K}_{\text{op}} \leq \kappa^2$.
	Indeed, it is not hard to prove that the trace of $K$ is bounded by $m\kappa^2$, hence $K$ is not only compact but trace-class.
\end{proof}

It should be noted that the condition on $K$ being dense in $L^2(\rho_\X)$ is not restrictive, as indeed all the problem is only seen through the lens of $\phi$ and $\rho_\X$: one can replace $\X$ by $\supp\rho_\X$ and $L^2(\rho_\X)$ by the closure of the range of $K$ in $L^2(\rho_\X)$ without modifying nor the analysis, nor the original problem.

We should study packing in the ellipsoid ${\cal F} = \brace{f\in L^2\midvert \|K^{-1/2}f\|_{L^2(\rho_\X)} \leq M}$.
It is useful to split the ellipsoid between a projection on a finite dimensional space that is isomorphic to the Euclidean space $\R^k$ and on a residual space $R$ where the energies $(\norm{f\vert_R}_{L^2(\rho_\X)})_{f\in{\cal F}}^2$ are uniformly small.
We begin with the following packing lemma, sometimes referred to as Gilbert-Varshamov bound \citep{Gilbert1952,Varshamov1957} which corresponds to a more generic result in coding theory.

\begin{lemma}[$\ell_2^2$-packing of the hypercube]
	For any $k \in \N^*$, there exists a $k/4$-packing of the hypercube $\brace{0, 1}^k$, with respect to Hamming distance, of cardinality $N = \exp(k/8)$.
\end{lemma}
\begin{proof}
	Let us consider $\epsilon > 0$, and a maximal $\epsilon$-packing $(x_i)_{i\leq N}$ of the hypercube with respect to the distance $d(x, y) = \sum_{i\in[k]} \ind{x_i\neq y_i} = \norm{x-y}_1 = \norm{x-y}_2^2$.
	By maximality, we have $\brace{0,1}^k \subset \cup_{i\in[N]}B_d(x_i,\epsilon)$, hence
	\[
		2^k \leq N \card{\brace{x\in\brace{0,1}^k\midvert \norm{x}_1 \leq \epsilon}}.
	\]
	This inequality can be rewritten with $Z$ a binomial variable of parameter $(k, \sfrac{1}{2})$ as
	\(
	1 \leq N \Pbb(Z \leq \epsilon).
	\)
	Using Hoeffding inequality \citep{Hoeffding1963}, when $\epsilon = k/4$ we get
	\[
		N^{-1} \leq \Pbb(Z \leq k/4) = \Pbb(Z - \E[Z] \leq k/4)
		\leq \exp\paren{-\frac{2k^2}{4^2 k}} = \exp\paren{-k/8}.
	\]
	This is the desired result.
\end{proof}

\begin{lemma}[Packing of infinite-dimensional ellipsoids]
	\label{sgd:lem:pack_ell}
	Let ${\cal F}$ be the function in $L^2(\rho_\X)$ such that $\norm{K^{-1/2}f}_{L^2(\rho_\X)}\leq M$ for $K$ a compact operator and $M$ any positive number.
	For any $k\in\N^*$, it is possible to find a family of $N \geq \exp(k/8)$ elements $(f_i)_{i\in[N]}$ in ${\cal F}$ such that for any $i\neq j$,
	\begin{equation}
		\frac{kM^2}{\sum_{i\leq k} \lambda_i^{-1}} \leq \norm{f_i - f_j}_{L^2(\rho_\X)}^2 \leq \frac{4k M^2}{\sum_{i\leq k}\lambda_i^{-1}},
	\end{equation}
	where $(\lambda_i)_{i\in\N}$ are the ordered (with repetition) eigenvalues of $K$.
\end{lemma}
\begin{proof}
	Let us denote by $(\lambda_i)_{i\in \N}$ the eigenvalues of $K$ and $(u_i)_{i\in\N}$ in $L^2$ the associated eigenvectors.
	Consider $(a_s)_{s\in[N]}$ a $k$-packing of the hypercube $\brace{-1,1}^k$ for $N \geq \exp(k/8)$ with respect to the $\ell^2_2$ quasi-distance and define for any $a\in\brace{a_s}$
	\[
		f_a = \frac{M}{c} \sum_{s=1}^k a_i u_i,
	\]
	with $c^2 = \sum_{i=1}^k \lambda_i^{-1}$.
	We verify that
	\begin{align*}
		 & \norm{K^{-1/2} f_a}_{L^2}^2 = \frac{M^2}{c^2} \sum_{i=1}^k \lambda_i^{-1} = M^2.
		\\&\norm{f_a - f_b}_{L^2}^2 = \frac{M^2}{c^2} \sum_{i=1}^k \abs{a_i - b_i}^2
		= \frac{M^2}{c^2} \norm{a_i - b_i}_2^2
		\in \frac{M^2}{c^2}\cdot [k, 4k].
	\end{align*}
	This is the object of the lemma.
\end{proof}

So far, we have proven the following lower bound.
\begin{lemma}
	\label{sgd:prop:step_2}
	Under Assumption \ref{sgd:ass:source} with ${\cal F} = \brace{x\in \X\to \theta\phi(x) \in \Y\midvert \norm{\theta} \leq M}$ and $\phi$ bounded by $\kappa$, for any family $(f_i)_{i\leq N_\epsilon} \in {\cal F}^N$ and any $\sigma > 2M\kappa \sigma_m$ and $km > 10$,
	\[
		\inf_{\cal A}\sup_\rho \E_{{\cal D}_n\sim\rho^{\otimes n}}[{\cal R}(f_{{\cal A}({\cal D}_n)}; \rho)] - {\cal R}^*(\rho)
		\geq \frac{1}{128}\min\brace{\frac{M^2}{\sigma m^{1/2}\sum_{i\leq k} (k\lambda_i)^{-1}}, \frac{\sigma km^{1/2}}{32 n}},
	\]
	for any algorithm ${\cal A}$ that maps a dataset ${\cal D}_n \in (\X\times\Y)^n$ to a parameter $\theta \in \Theta$, and where $(\lambda_i)$ are the ordered eigenvalue of the operator $K$ on $L^2(\X,\R,\rho_\X)$ that maps any function $f$ to the function $Kf$ defines for $x'\in\X$ as
	\[
		(Kf)(x') = \int_{x\in\X} \scap{\phi(x)}{\phi(x')} f(x)\rho_\X(\diff x).
	\]
	In particular, when $\lambda_i = \kappa^2 i^{-a} / \zeta(\alpha)$, where $\zeta$ denotes the Riemann zeta function, we get the following bounds.
	If we optimize with respect to $\sigma$, there exists $n_\alpha \in \N$ such that for any $n > n_\alpha$.
	\begin{equation}
		\inf_{\cal A}\sup_\rho \E_{{\cal D}_n\sim\rho^{\otimes n}}[{\cal R}(f_{{\cal A}({\cal D}_n)}; \rho)] - {\cal R}^*(\rho)
		\geq \frac{M\kappa}{725\zeta(\alpha)^{1/2} n^{1/2}}.
	\end{equation}
	If we fix $\sigma = \beta M\kappa$ with $\beta \geq 2$, and we optimize with respect to $k$, there exists a constant $c_\beta$ and an integer $n_0$ such that for $n > n_0$ we have
	\begin{equation}
		\inf_{\cal A}\sup_\rho \E_{{\cal D}_n\sim\rho^{\otimes n}}[{\cal R}(f_{{\cal A}({\cal D}_n)}; \rho)] - {\cal R}^*(\rho)
		\geq \frac{M\kappa c_\beta}{\zeta(\alpha)^{\frac{1}{1+\alpha}} n^{\frac{\alpha}{\alpha+1}}}.
	\end{equation}
\end{lemma}
\begin{proof}
	Reusing Lemma \ref{sgd:prop:step_1}, with the same notations, we have the lower bound in
	\[
		\frac{\min\norm{f_i - f_j}^2}{16\sigma(m+2)^{1/2}}\paren{1 - \frac{\log(2)}{\log(N)} - \frac{n\max\norm{f_i - f_j}^2}{2\sigma^2\log(N)}}.
	\]
	Let $K$ and $K_\Y$ be the self-adjoint operators on $L^2(\X, \R, \rho_\X)$ and $L^2(\X, \Y, \rho_\X)$ respectively, both defined through the formula
	\[
		(Kf)(x') = \int_{x\in\X} \scap{\phi(x)}{\phi(x')} f(x)\rho_\X(\diff x).
	\]
	When $K$ is compact, it admits an eigenvalue decomposition $K = \sum_{i\in\N} \lambda_i u_i\otimes u_i$ where the equality as to be understood as the convergence of operator with respect to the operator norm based on the $L^2$-topology.
	It follows from the product structure of $L^2(\X, \Y, \rho_\X) \simeq L^2(\X,\R,\rho_\X)^m$ that $K_\Y = \sum_{i\in\R, j\in[m]} \sum_{i\in\N, j\in[m]} \lambda_i (e_i\otimes y_j) \otimes (e_i\otimes u_j)$ with $(e_j)$ the canonical basis of $\Y = \R^m$.
	As a consequence, if $(\lambda_i)_{i\in\N}$ are the ordered eigenvalues of $K$ then $(\lambda_{\floor{i / m}})$ are the ordered eigenvalues of $K_\Y$.
	Hence, with Lemmas \ref{sgd:lem:lin_ell} and \ref{sgd:lem:pack_ell}, it is possible to find $N=\exp(km/8)$ functions in ${\cal F}$ such that
	\[
		\frac{km M^2}{m\sum_{i\leq k} \lambda_i^{-1}} \leq \norm{f_i - f_j}_{L^2(\rho_\X)}^2 \leq \frac{4km M^2}{m\sum_{i\leq k}\lambda_i^{-1}}.
	\]
	If we multiply those functions by $\eta\in[0,1]$ we get a lower bound in
	\[
		\frac{\eta^2 M^2}{16\sigma(m+2)^{1/2}\sum_{i\leq k}(k\lambda_i)^{-1}}\paren{1 - \frac{8\log(2)}{km} - \frac{16 M^2 n\eta^2}{\sigma^2 km \sum_{i\leq k} (k\lambda_i)^{-1}}}.
	\]
	Making sure that the last two terms are smaller than one fourth and one half respectively we get the following conditions on $k$ and $\eta$, with $\Lambda_k = \sum_{i\leq k} (k\lambda_i)^{-1}$,
	\[
		km \geq 32 \log(2),\qquad
		32M^2 n \eta^2 \leq \sigma^2 km\Lambda_k.
	\]
	Using the fact that $\eta < 1$, the lower bound becomes
	\[
		\frac{M^2}{128\sigma m^{1/2}\Lambda_k} \min\brace{1, \frac{\sigma^2 km \Lambda_k}{32M^2 n}}
		=\frac{1}{128}\min\brace{\frac{M^2}{\sigma m^{1/2}\Lambda_k}, \frac{\sigma km^{1/2}}{32 n}},
	\]
	as long as $km > 10$.
	When $\lambda_i^{-1} = i^\alpha \zeta(\alpha) / \kappa^2$, since $\Lambda_k \leq \lambda_k^{-1}$, we simplify the last expression as
	\[
		\frac{1}{128}\min\brace{\frac{M^2 \kappa^2}{\sigma m^{1/2}k^\alpha \zeta(\alpha)}, \frac{\sigma km^{1/2}}{32 n}}.
	\]
	Optimizing with respect to $\sigma$ leads to
	\[
		\sigma^2 = \frac{32nM^2\kappa^2}{mk^{1+\alpha} \zeta(\alpha)} \geq 4M^2\kappa^2 \sigma_m.
	\]
	This gives
	\[
		n_{\alpha, m} = m\zeta(\alpha) \sigma_m^2 / 8.
	\]
	The dependency of $n_\alpha$ to $m$ can be removed since any problem with $\Y=\R^m$ can be cast as a problem in $\R^{m+1}$ by adding a spurious coordinate.
	Taking $k=1$ and $m=10$ leads to the result stated in the lemma.
	When $n < n_\alpha$, one can artificially multiply the bound by $n_\alpha^{1/2}$, since an optimal algorithm can not do better with fewer data.
	After checking that one can take $\sigma_1 \geq 1$, this leads to a bound in
	\[
		\frac{M\kappa}{2048 n^{1/2}}.
	\]
	Optimizing with respect to $k$ leads to $k^{\alpha + 1} = 32 M^2 \kappa^2 n / (\sigma^2 m \zeta(\alpha))$ and a bound in
	\[
		\frac{(\sigma m^{1/2})^{\frac{\alpha-1}{\alpha+1}} (M\kappa)^{\frac{2}{\alpha+1}}}{128(32 n)^{\frac{\alpha}{\alpha+1}} \zeta(\alpha)^{\frac{1}{\alpha+1}}}.
	\]
	The condition $k > \min\brace{10m^{-1}, 1}$ and $\sigma \geq 2M\kappa\sigma_m$ translates into the condition
	\[
		4M^2\kappa^2\sigma_m^2 \leq \sigma^2 \leq \frac{32 M^2\kappa^2 n}{m\zeta(\alpha)} \min\brace{1, \frac{m^{1+\alpha}}{10^{1+\alpha}}}.
	\]
	We deduce that $\sigma_m = O(m^{-1/2})$, otherwise we would not respect the upper bound derived with Rademacher complexity (or have made a mistake somewhere).
	Once again we can remove the dependency to $m$. Considering $\sigma =\beta M\kappa$ leads to the result stated in the lemma.
\end{proof}

\subsubsection{Controlling eigenvalues decay}

Based on Lemma \ref{sgd:prop:step_2}, in order to prove Theorem \ref{sgd:thm:minmax_opt}, we only need to show that there exists a mapping $\phi$, an input space $\X$ and a distribution $\rho_\X$ such that the integral operator $K$ introduced in the lemma verifies the assumption on its eigenvalues.
Notice that we show in the proof of Lemma \ref{sgd:prop:step_2} that the universal constant $c_3$ can be taken as $c_3 = 2^{-11}$.

To proceed, let us consider any infinite dimensional Hilbert space ${\cal H}$ with a basis $(e_i)_{i\in\N}$, $\X=\N$ and $\phi:\N\to{\cal H}; i\to \kappa e_i$.
For $a:\N\to\R$ we have
\[
	(Ka)(i) = \sum_{j\in\N} \scap{\phi(i)}{\phi(j)} a(j) \rho(j) = \kappa^2 a(i)\rho(i).
\]
Hence, the eigenvalues of $K$ are $(\kappa^2\rho_\X(i))_{i\leq n}$.
It suffices to consider $\rho_\X(i) = i^{-\alpha} / \zeta(\alpha)$ to conclude.

The eigenvalue decay in $O(i^{-\alpha})$ can also be witnessed in many regression problems.
One way to build those cases is to turn a sequence of non-negative real values into a one-periodic function $h$ from $\R^d$ to $\R$ thanks to the inverse Fourier transform.
Using \cite{Bochner1933}, one can construct a map $\phi$ such that the convolution operator linked with $h$ corresponds to the operator $K$.
When $\rho$ is uniform on $[0,1]^d$, diagonalizing this convolution operator with the Fourier functions and using the property in Lemma \ref{sgd:lem:lin_ell} shows that the class of functions ${\cal F}$ are akin to Sobolev spaces.
Similar behavior can be proven when $\X=\R^d$ and $\rho_\X$ is absolutely continuous with respect to the Lebesgue measure and has bounded density \citep{Widom1963}.
We refer the curious reader to \cite{Scholkopf2001} or \cite{Bach2023} for details.

\section{Unbiased weakly supervised stochastic gradients}
\label{sgd:app:generic}

In this section, we provide a generic scheme to acquire unbiased weakly supervised stochastic gradients, as well as specifications of the formula given in the main text for least-squares and median regression.

\subsection{Generic implementation}

Suppose that $\Theta$ is finite dimensional, or that it can be approximated by a finite dimensional space without too much approximation error.
For example, in the realm of scalar-valued kernel methods, it is usual to consider either the random finite dimensional space $\Span\brace{\phi(x_i)}_{i\leq n}$ for $(x_i)$ the data points, or the finite dimension space linked to the first eigenspaces of the operator $\E[\phi(X)\otimes \phi(X)]$.
In the context of neural networks, the parameter space is always finite-dimensional.

Suppose also that, given $\theta$, we know an upper bound $M_\theta$ on the amplitude of $\nabla_\theta \ell(f_\theta(x), y)$, or that we know how to handle clipped gradients at amplitude $M_\theta$ for SGD.
Then, similarly to the least-squares method proposed in the main text, we can access weakly supervised gradient through the formula
\begin{equation*}
	\nabla_\theta\ell(f_\theta(x), y) = \frac{2M_\theta(\card\Theta^2 + 4\card\Theta + 3)}{\pi^{3/2}} \E_{U \sim \uniform{B_\Theta},V\sim\uniform{[0, M_\theta]}}[\ind{y\in(z\to\scap{U}{\nabla_\theta \ell(f_\theta(x), z)})^{-1}([V, \infty))} U],
\end{equation*}
where $B_\Theta$ is the unit ball of $\Theta$.

This scheme is really generic, and we do not advocate for it in practice as one may hope to leverage specific structure of the loss function and the parametric model in a more efficient way.
This formula is rather a proof of concept to illustrate that our technique can be applied generically, and is not specific to least-squares or median regression.

\subsection{Specific implementations}

Let us prove the two formulas to get stochastic gradients for both least-squares and median regression.
We begin with median regression.
Consider $z\in\mathbb{S}^{m-1}$, and let us denote
\[
	x = \E_U[\sign(\scap{z}{U}) U].
\]
The direction $x/\norm{x}\in\mathbb{S}^{m-1}$ is characterized by the argmax over the sphere of the linear form
\[
	y\to \scap{\E_U[\sign(\scap{z}{U})U]}{y}
	= \E_U[\sign(\scap{z}{U})\scap{U}{y}].
\]
This linear form has a unique maximizer on $\mathbb{S}^{m-1}$ and by invariance by symmetry over the axis $z$, this maximizer is aligned with $z$, hence $x = c_x\cdot z$.
We compute the amplitude with the formula, because $z$ is a unit vector
\[
	c_x = \scap{x}{z} = \E_U[\sign(\scap{z}{U})\scap{U}{z}].
\]
By invariance by rotation of both the uniform distribution and the scalar product, $c_x$ is actually a constant, it is equal to its value $c_2 = c_{e_1}$.

The same type of reasoning applies for the least-squares case.
Consider $z \in\R^m$, and denote
\[
	x = \E_{U, V}\bracket{\ind{\scap{z}{U} \geq V}\cdot U}.
\]
For the same reasons as before $x = c_x\cdot u$ for $u = z / \norm{z}$, and $c_x$ verifies
\begin{align*}
	c_x
	 & = \scap{x}{u}
	= \E_{U,V}[\ind{\scap{z}{U} \geq V}\scap{U}{u}]
	= \E_U[\E_{V}[\ind{\scap{z}{U} \geq V}]\scap{U}{u}]
	\\&= \E_{U}[\ind{\scap{z}{U} > 0} \frac{\scap{z}{U}}{M}\scap{U}{u}]
	= \frac{\norm{z}}{M}\E_{U}[\ind{\scap{u}{U}>0} \scap{U}{u}^2].
\end{align*}
Hence,
\[
	x = \frac{1}{M}\E_{U}[\ind{\scap{u}{U}>0} \scap{U}{u}^2] \cdot z = c_1 \cdot z.
\]
This explains the formula for least-squares.

\begin{lemma}[Constant for the uniform strategy]
	Under the uniform distribution on the sphere
	\begin{equation}
		c_2 = \E_{u\sim\mathbb{S}^{m-1}} \bracket{\abs{\scap{u}{e_1}}}
		= \frac{\sqrt{\pi}\,\Gamma(\frac{m-1}{2})}{m\,\Gamma(\frac{m}{2})} \geq \frac{\sqrt{2 \pi}}{m^{3/2}}.
	\end{equation}
\end{lemma}
\begin{proof}
	Let us compute $c_2 = \E_{u\sim\mathbb{S}^{m-1}} \bracket{\abs{\scap{u}{e_1}}}$.
	This constant can be written explicitly as
	\[
		c_2 = \frac{\int_{x\in\mathbb{S}^{m-1}} \abs{x_1} \diff x}{\int_{x\in\mathbb{S}^{m-1}} \diff x}.
	\]
	Remark that for any function $f:\R\to\R$, we have
	\[
		\int_{\mathbb{S}^{m-1}} f(x_1) \diff x
		= \int_{x_1 \in [-1, 1]} f(x_1)\diff x_1 \int_{\tilde{x}\in \sqrt{1-x_1^2}\cdot\mathbb{S}^{m-2}} \diff \tilde{x}
		= \int_{x_1\in[-1,1]} f(x_1) (1 - x_1^2)^{\frac{m-2}{2}} \diff x_1 \int_{\tilde{x}\in\mathbb{S}^{m-2}} \diff \tilde{x}.
	\]
	By denoting $S_{m}$ the surface of the $m$-sphere, the last integral is nothing but $S_{m-2}$.
	By setting $f(x)=1$, we can retrieve by recurrence the expression of $S_m$.
	In our case, $f(x)=\abs{x}$, so we compute, with $u = 1-x^2$
	\[
		\int_{x_1\in[-1,1]} \abs{x_1} (1 - x_1^2)^\frac{m-2}{2} \diff x_1
		=2\int_{x_1\in[0,1]} x_1 (1 - x_1^2)^\frac{m-2}{2} \diff x_1
		= \int_{u=0}^1 u^\frac{m-2}{2} \diff u = \frac{1}{m}.
	\]
	This leads to
	\[
		c_2 = \frac{S_{m-2}}{m S_{m-1}} = \frac{\sqrt{\pi}\,\Gamma(\frac{m-1}{2})}{m\,\Gamma(\frac{m}{2})}.
	\]
	The ratio $S_{m-2} / S_{m-1}$ can be expressed with the integral corresponding to $f=1$, but it is common knowledge that $S_{m-1} = \sfrac{2\pi^{m/2}}{\Gamma(m/2)}$.
\end{proof}

\begin{lemma}[Constant for least-squares]
	Under the uniform distributions on $[0, M]$ and the sphere
	\begin{equation}
		c_1 = \E_{y\sim[0, M]}\E_{u\sim\mathbb{S}^{m-1}} \bracket{\ind{\scap{u}{e_1} > v} \scap{u}{e_1}} = \frac{\pi^{3/2}}{M (m^2 + 4m + 3)}.
	\end{equation}
\end{lemma}
\begin{proof}
	Similarly to the previous case, this constant can be written explicitly as
	\[
		c_1 = \frac{1}{2}\frac{\int_{y\in[0,M]}\int_{x\in\mathbb{S}^{m-1}} \abs{x_1} \ind{\abs{x_1} > y} \diff y\diff x}{M\int_{x\in\mathbb{S}^{m-1}} \diff x}
		= \frac{\int_{x\in\mathbb{S}^{m-1}} x_1^2 \diff x}{2M\int_{x\in\mathbb{S}^{m-1}} \diff x}.
	\]
	We continue as before with
	\[
		\int_{x_1\in[-1,1]} \abs{x_1}^2 (1 - x_1^2)^\frac{m-2}{2} \diff x_1
		= 2\int_{x\in[0,1]} x^2 (1 - x^2)^\frac{m-2}{2} \diff x
		= \frac{2\pi \Gamma(\frac{m}{2})}{4\Gamma(\frac{m+3}{2})}.
	\]
	This leads to
	\[
		c_1 = \frac{\pi \Gamma(\frac{m}{2})}{4M\Gamma(\frac{m+3}{2})}\cdot\frac{\sqrt{\pi}\,\Gamma(\frac{m-1}{2})}{\Gamma(\frac{m}{2})}
		= \frac{\pi^{3/2} \Gamma(\frac{m-1}{2})}{4M\Gamma(\frac{m+3}{2})}
		= \frac{\pi^{3/2}}{M (m^2 + 4m + 3)}.
	\]
	This is the result stated in the lemma.
\end{proof}
	\section{Median surrogate}
\label{sgd:proof:sur}

Let us begin this section by proving Proposition \ref{sgd:prop:sur}.
This result is actually the integration over $x\in\X$ of a pointwise result, so let us fix $x\in\X$.
Consider a probability distribution $p\in\prob{\Y}$ over $\Y$, and its median $\Theta^* \subset \R^\Y$ defined as the minimizer of ${\cal R}_S(\theta) = \E_p[\norm{\theta - e_Y}]$.
We will to prove that $\cup_{\theta\in\Theta^*}\argmax_{y\in\Y} \theta_y = \argmax_{y\in\Y} p(y)$.

Let us begin by the inclusion $\argmax_{y\in\Y} p(y) \subset \cup_{\theta\in\Theta^*}\argmax_{y\in\Y} \theta_y$.
To do so, consider $\theta\in\R^\Y$ and $\sigma \in \Sfrak_\Y$ the transposition of two elements $y$ and $z$ in $\Y$.
Denote by $\theta_\sigma \in \R^\Y$, the vector such that $(\theta_\sigma)_{y'} = \theta_{\sigma(y')}$ for any $y'\in\Y$.
We have
\begin{align*}
	{\cal R}_S(\theta) - {\cal R}_S(\theta_\sigma)
	 & = \sum_{y'\in \Y} p(y') \paren{\norm{\theta - e_{y'}} - \norm{\theta_\sigma - e_{y'}}}
	\\&= \sum_{y'\in \Y} p(y') \paren{\sqrt{\sum_{z'\in\Y}\theta_{z'}^2 + (1 - \theta_{y'})^2 - \theta_{y'}^2} -
		\sqrt{\sum_{z'\in\Y}\theta_{\sigma(z')}^2 + (1 - \theta_{\sigma(y')})^2 - \theta_{\sigma(y')}^2}}
	\\&= (p(y) - p(z)) \paren{\sqrt{\sum_{z'\in\Y}\theta_{z'}^2 + 1 - 2\theta_{y}} -
		\sqrt{\sum_{z'\in\Y}\theta_{z'}^2 + 1 - 2\theta_{z}}}.
\end{align*}
Because, for any $a \in \R_+$, the function $x \to \sqrt{a - 2x}$ is increasing, if $p(y) > p(z)$, then to minimize ${\cal R}$, we should make sure that $\theta_y \geq \theta_z$i.
As a consequence, because of symmetry, the modes of $p$ do correspond to argmax of $(\theta^*_y){y\in\Y}$ for some $\theta^*\in\Theta^*$.

Let us now prove the second inclusion.
To do so, suppose that $p(1) > p(2)$, and let us show that $\theta^*_1 > \theta^*_2$.
Let us parametrize $\theta_1 = a + \epsilon$ and $\theta_2 = a - \epsilon$ for a given $a$, and show that $\epsilon = 0$ is not optimal in order to minimize the risk ${\cal R}_S$ seen as a function of $\epsilon$.
To do so, we can use the Taylor expansion of $\sqrt{1 + x} = 1 + x / 2$.
Hence, with $A = \sum_{y > 2} (\theta^*_y)^2$, retaking the last derivations
\begin{align*}
	{\cal R}_S(\epsilon)
	 & = p(1)\sqrt{(a+\epsilon)^2 + (a-\epsilon)^2 + A + 1 - 2(a + \epsilon)}
	\\&\qquad+ p(2)\sqrt{(a+\epsilon)^2 + (a-\epsilon)^2 + A + 1 - 2(a - \epsilon)}
	\\&\qquad+ \sum_{y>2} p(y)\sqrt{(a+\epsilon)^2 + (a-\epsilon)^2 + A + 1 - 2\theta^*_y}
	\\&= p(1)\sqrt{2a^2+ 2\epsilon^2 + A + 1 - 2a - 2\epsilon}
	\\&\qquad+ p(2)\sqrt{2a^2+2\epsilon^2 + A + 1 - 2a + 2\epsilon} + c + o(\epsilon)
	\\&= \tilde{c} + \frac{\epsilon}{\sqrt{2a^2 + A + 1 - 2a}}(p(2) - p(1)) + o(\epsilon).
\end{align*}
This shows that taking $\theta_1^* = \theta_2^*$, that is $\epsilon=0$, is not optimal, hence we have the second inclusion, which ends the proof.
Note that we have proven a much stronger result, we have shown that $(\theta_y)$ and $p(y)$ are order in the exact same fashion (with respect to the strict comparison $p(y)>p(z)\Rightarrow \theta_y^* > \theta_z^*$ for any $\theta^*\in\Theta^*$).

\subsection{Discussion around the median surrogate.}
The median surrogate have some nice properties for a surrogate method, in particular it does not fully characterize the distribution $p(y)$ in the sense that there is no one-to-one mapping from $p$ to $\theta^*$.
For example, when $\Y = \brace{1, 2, 3}$ if
\(
p(y=e_1), p(y=e_2), p(y=e_3) \propto (1, 1, 2\cos(\pi / 6)),
\)
then the geometric median correspond to $\theta^* = e_3$.
This differs from smooth surrogates, such as logistic regression or least-squares, that implicitly learn the full distribution $p$, which should be seen as a waste of resources.
Non-smooth surrogates tend to exhibit faster rates of convergence (in terms of decrease of the original risk as a function of the number of samples) than smooth surrogates when rates are derived through calibration inequalities \citep{Nowak2021}.
It would be nice to derive generic calibration inequality for the median surrogate for multiclass, and see how to derive a median surrogate for more structured problems such as ranking problems.

\begin{figure}[ht]
	\centering
	\includegraphics{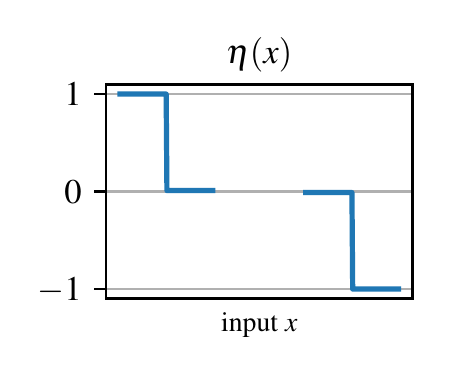}
	\includegraphics{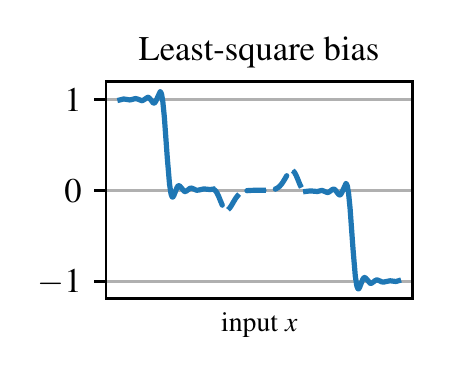}
	\includegraphics{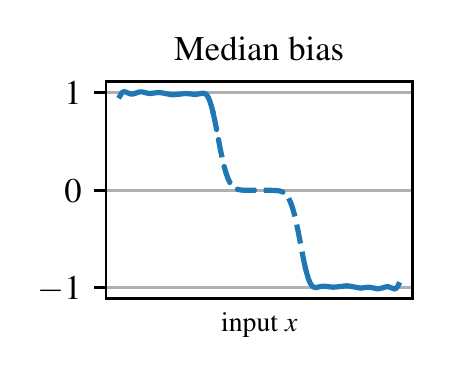}
	\includegraphics{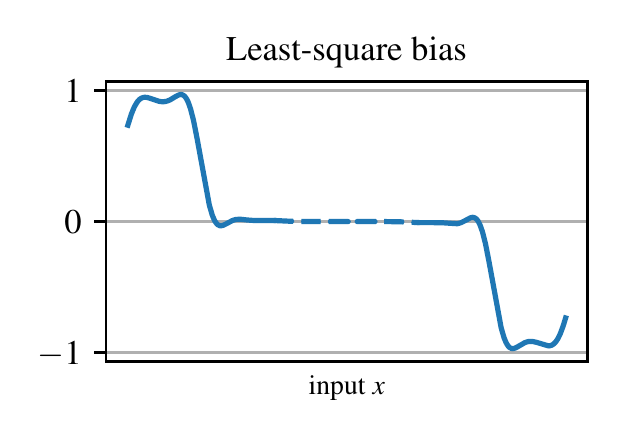}
	\includegraphics{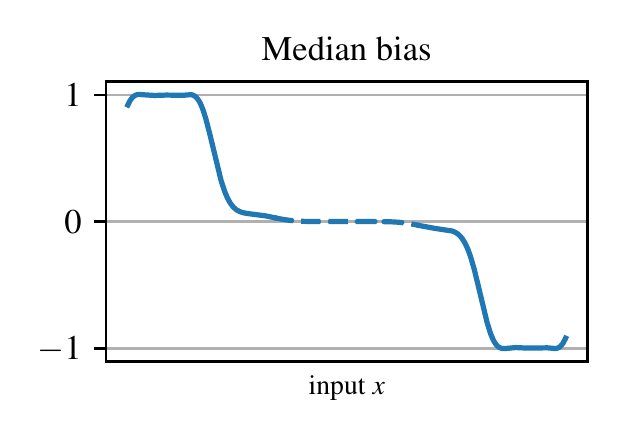}
	\caption{
	{\em Comparison of least-squares and absolute deviation with noise irregularity} for a classification problem specified by $\X = [0, 3]$, $\Y = \brace{-1, 1}$ with $X$ uniform on $[0, 1]\cup[2, 3]$ and $\eta(x) = \E\brace{Y\midvert X=x}$ specified on the left figure.
	The optimal classifier, with respect to the zero-one loss, $f^*(x) = \sign \eta$ takes value one on $[0, 1]$ and value minus one on $[2, 3]$.
	The regularized solution are defined as $\argmin_g \E[\norm{\scap{\phi(X)}{\theta} - Y}^p] + \lambda \norm{\theta}$ with $p=2$ for least-squares (middle), and $p=1$ for the median (right).
	They can be translated into classifiers with the decoding $f = \sign g$.
	In this figure, we choose $\phi$ implicitly through the Gaussian kernel $k(x, x') = \scap{\phi(x)}{\phi(x')} = \exp(-\norm{x-x'}^2 / 2\sigma^2)$ with $\sigma = .1$ which explains the frequency of the observed oscillations, and choose $\lambda = 10^{-6}$ (top) and $\lambda = 10^{-2}$ (bottom).
	On the one hand, because the least-squares surrogate is trying to estimate $\eta$ it suffers from its lack of regularity, leading to Gibbs phenomena that restricts it to be a perfect classifier.
	On the other hand, the absolute deviation is trying to approach the function $f^*$ itself, and does not suffer from its lack of regularity.
	In this setting, if we approach the original classification problem by minimization of the surrogate empirical risks, and denote by $g_n$ this minimizer and $f_n =\sign g_n$ its decoding, $f_n$ obtained through median regression will converge exponentially fast toward $f^*$, while $f_n$ obtained through least-squares will never converge to the solution $f^*$.
	}
	\label{sgd:fig:med_ls}
\end{figure}

\begin{figure}[ht]
	\centering
	\includegraphics{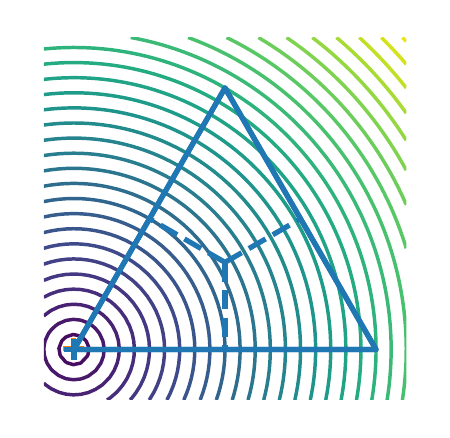}
	\includegraphics{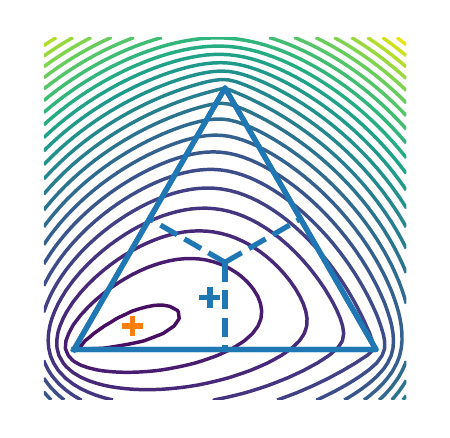}
	\includegraphics{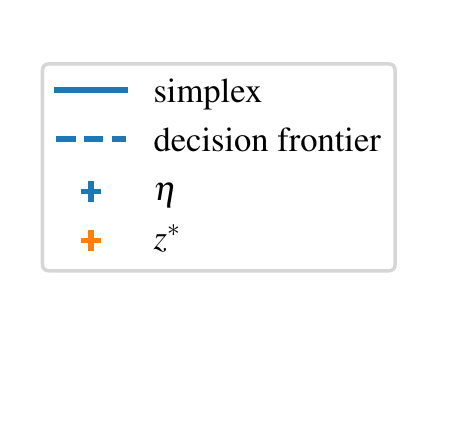}
	\caption{
	{\em Comparison of least-squares and median surrogate without context.}
	Consider a context-free classification problem that consists in estimating the mode of a distribution $p\in\prob{\Y}$, or equivalently the minimizer of the 0-1 loss.
	Such a problem can be visualized on the simplex $\prob{m}$ where $\Y = \brace{y_1, \cdots, y_m} \simeq \brace{1, \cdots, m}$ is mapped to the canonical basis $\brace{e_i}_{i\in[m]} \in \R^m$.
	The figure illustrates the case $m=3$.
	The least-squares and median surrogate methods can be understood as working in this simplex, estimating a quantity $z \in \prob{\Y}$, before performing the decoding $y(z) = \argmax_y \scap{z}{e_y}$.
	Such a decoding partitions the simplex in regions whose frontiers are represented in dashed blue on the figure.
	The distribution $p$ is characterized on the simplex by $\eta = \E_{Y\sim p}[e_Y] = \argmin \E_{Y\sim p} [\norm{z - e_Y}^2]$.
	This quantity $\eta$ is exactly the quantity estimated by the least-squares surrogate.
	The median surrogate searches the minimizer $z^*$ of the quantity ${\cal E}(z) = \E_{Y\sim p}[\norm{z - e_Y}]$, whose level lines are represented in solid on the figure.
	One of the main advantage of the median surrogate compared to the least-squares one is that $z^*$ is always farther away from the boundary frontier than $\eta$, meaning that for a similar estimation error on this quantity, the error on the decoding, which corresponds to an estimate of the mode of $p$, will be much smaller for the median surrogate.
	The left figure represents the case $p = (1, 0, 0)$, the right figure the case $p =(.45, .35, .2)$.}
	\label{sgd:fig:med_simplex}
\end{figure}

\begin{figure}[ht]
	\centering
	\includegraphics{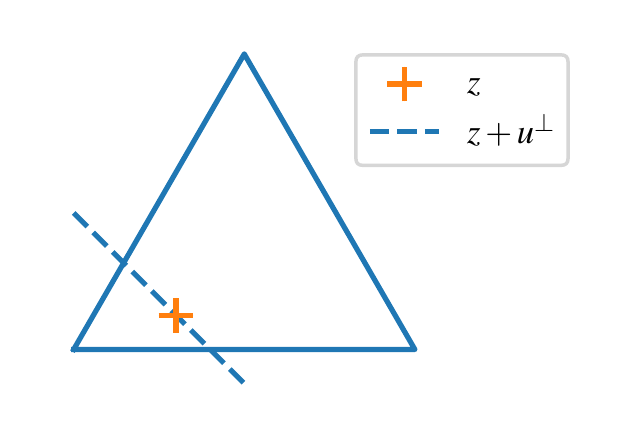}
	\caption{
		{\em Query strategy based on regression surrogate.}
		Retaking the simplex representation of Figure \ref{sgd:fig:med_simplex}, the query strategy for classification approached with least-squares surrogate or median surrogate consists in looking at the current surrogate estimate $z$ in the simplex $\prob{\Y}$, taking a random direction $u\in\R^\Y$ and querying $\sign(\scap{e_Y - z}{u})$.
		We see that with three elements, when $Y$ is deterministic, the optimal query strategy consists in considering $s =\brace{y}$, while surrogate strategies, such as least-squares and median regression, that learn $z^* = e_y$, would only make such a query only two third of the time (which is the ratio of the solid angle of $[e_2, e_3]$ from $e_1$ divided by $\pi$).
		This shows that those surrogate strategies do not fully leverage the specific structure of the output.
	}
	\label{sgd:fig:med_query}
\end{figure}

\section{Classification with a min-max game}
\label{sgd:proof:minmax}

In this section, we prove and extend on Proposition \ref{sgd:prop:minmax}.
First of all, let us consider the average loss, for $(v_y)\in\R^\Y$ summing to one
\[
	\bar L(v, s) = 1 - \sum_{y\in s} v_y = \sum_{y\notin s} v_y.
\]
Consider now this loss conditioned on the observation $\ind{y\in s}$, we have plenty of characterizations of $L$,
\begin{align*}
	L(v, s; \ind{y\in s} - \ind{y\notin s})
	 & = \ind{y\in s} \bar L(v, s) + \ind{y\notin s}\bar L(v, \Y\setminus s)
	= \ind{y\in s}\sum_{y\notin s} v_y + \ind{y\notin s}\sum_{y\in S} v_y
	\\&= \ind{y\in s} + (\ind{y\notin s} - \ind{y\in s})\sum_{y\in s} v_y
	= \ind{y\notin s} + (\ind{y\in s} - \ind{y\notin s})\sum_{y\notin s} v_y
	\\&= \frac{1}{2} - \frac{1}{2} (\ind{y\in s} - \ind{y\notin s})\paren{\sum_{y\in s} v_y - \sum_{y\notin s} v_y}
	= \frac{1}{2} + \frac{1}{2} (\ind{y\in s} - \ind{y\notin s})\paren{1 - 2\sum_{y\in s} v_y}.
\end{align*}
Minimizing this loss or the loss $2L-1$ as defined in Proposition \ref{sgd:prop:minmax} is equivalent.

\subsection{Consistency}
Let us consider the loss as defined in this proposition, we have the characterization
\[
	L(v, s; \ind{y\in s} - \ind{y\notin s}) = (\ind{y\in s} - \ind{y\notin s})\paren{\sum_{y\in s} v_y - \sum_{y\notin s} v_y}.
\]
Let us rewrite~\eqref{sgd:eq:minmax} based on this previous characterization of the loss, we have
\[
	\E_Y[L(v, s, \ind{Y\in s} - \ind{Y\notin s})]
	= -(\Pbb_Y(Y\in s) - \Pbb_Y(Y\notin s))\paren{\sum_{y\in s} v_y - \sum_{y\notin s} v_y}.
\]
Hence, without any context variable, the min-max game \eqref{sgd:eq:minmax} can be rewritten as
\begin{equation}
	\min_{v\in\prob{\Y}} \max_{\mu\in\prob{\cal S}}
	- \sum_{s\in{\cal S}} \mu_s (\Pbb_Y(Y\in s) - \Pbb_Y(Y\notin s))\paren{\sum_{y\in s} v_y - \sum_{y\notin s} v_y}.
\end{equation}
We will analyze this problem through the lens of a mix-actions zero-sum game.
We know from \cite{VonNeumann1944} that a solution to this min-max problem exists, and that one can switch the min-max to a max-min without modifying the value of the solution.
Let us denote by $(v^*,\mu^*)$ the argument of a solution.
To minimize the value of this game, the player $v$ should play such that
\[
	\sign(\sum_{y\in s} v_y^* - \sum_{y\notin s} v_y^*)
	= \sign (\Pbb(Y\in s) - \Pbb(Y\notin s))
	= \sign (\sum_{y\in s} \Pbb(Y=y) - \sum_{y\notin s} \Pbb(Y=y)),
\]
which allows this player to ensure a negative value to the game.
Stated otherwise
\begin{equation}
	\label{sgd:eq:loss_imp}
	\forall\, s\in{\cal S}, \qquad
	\Pbb(Y\in s) > \frac{1}{2}\quad\Rightarrow\quad\sum_{y\in s} v_y^* \geq \frac{1}{2}.
\end{equation}
As a consequence, if there exists any set such that $\Pbb(Y\in s) = 1/2$, the best strategy of player $\mu$ is to play only those sets to ensure the value zero, and any $v$ that satisfies~\eqref{sgd:eq:loss_imp} is optimal.
It should be noted that~\eqref{sgd:eq:loss_imp} does not generally imply that $(v_y)_{y\in\Y}$ has the same ordering as $(\Pbb(Y=y))_{y\in\Y}$.

When $\brace{y^*}\in{\cal S}$ and $\Pbb(Y=y^*) > 1/2$, if $v = \delta_{y^*}$, the prediction player is able to ensure a value of $\max_{s\in{\cal S}}-\abs{2\Pbb(Y\in s) - 1}$, which is maximized by the query player with $s = \brace{y^*} \cup s'$ for any $s'$ such that $\Pbb(Y\in s') = 0$. Other strategies for $v$ will only increase this value, hence $v^* = \delta_{y^*}$ which implies the first part of Proposition \ref{sgd:prop:minmax}.

\paragraph{A counter example.}
While we hope that the solution $(v^*, \mu^*)$ does characterize the original solution $y^*$, it should be noted that $v^*$ alone does not characterize $y^*$.
Indeed, it is even possible to have $v^*$ uniquely defined without having $y^* = \argmax_{y\in\Y} v^*_y$.
For example, consider the case where $\Y = \brace{1, 2, 3}$ and $(\Pbb(Y=i))_{i\in[3]} = (.4, .3, .3)$.
By symmetry, the player $\mu$ only has to play on ${\cal S} = \brace{\brace{1}, \brace{2}, \brace{3}}$, which leads to the min-max game
\[
	\min_{v} \max_{\mu}
	\paren{
		\begin{array}{c}
			\mu_{\brace 1} \\
			\mu_{\brace 2} \\
			\mu_{\brace 3} \\
		\end{array}
	}^\top
	\paren{
		\begin{array}{ccc}
			.2  & -.2 & -.2 \\
			-.4 & .4  & -.4 \\
			-.4 & -.4 & .4  \\
		\end{array}
	}
	\paren{
		\begin{array}{c}
			v_1 \\
			v_2 \\
			v_3 \\
		\end{array}
	}.
\]
The value of this game is $-.1$ and is achieved for $\mu^* = (.5, .25, .25)$, $v^* = (.25, .375, .375)$.

\subsection{Optimization procedure}

Let us rewrite the problem through the objective
\[
	{\cal E}(g, \mu) = \E_{(X, y)\sim \rho}\E_{S\sim\mu(x)}[L(g(X), S, \ind{Y\in S} - \ind{Y\notin S})].
\]
We want to solve the min-max problem $\min_g\max_\mu{\cal E}(g, \mu)$.
This problem can be solved efficiently based on the vector field point of view of gradient descent \citep{Bubeck2015} if:
\begin{itemize}
	\item we can parametrize the function $g:\X\to\prob\Y$ such that ${\cal E}$ is convex with respect to the parametrization of $g$;
	\item we can access unbiased stochastic gradients of ${\cal E}$ with respect to $g$ that have a small second moment;
	\item we can parametrize the function $\mu:\X\to\prob{\cal S}$ such that ${\cal E}$ is concave with respect to the parametrization of $\mu$;
	\item we can access unbiased stochastic gradients of ${\cal E}$ with respect to $\mu$ that have a small second moment.
\end{itemize}
The first two points are no problems, $g$ can be parametrized with softmax regression, and since $L$ is linear with respect to the scores, it will keep the problem convex.
Moreover, to access a stochastic gradient of ${\cal E}$, one can sample $X_i\sim\rho_\X$ and $S_i\sim\mu(X_i)$ before querying $\ind{Y_i\in S_i}$ and computing the gradient of $L(g(X_i), S_i, \ind{Y_i\in S_i} - \ind{Y_i\notin S_i})$ with respect to the parametrization of $g$.

The third point is slightly harder to tackle.
Since ${\cal E}$ is linear with respect to $\mu$, one way to proceed is to find a linear parametrization of $\mu$.
In particular, one can take a family $(g_i)_{i\in[N]}$ of linearly independent functions from $\X$ to $\prob{\cal S}$ and search for $g$ under the form $\sum_{i\in[N]} c_i g_i$ for $(c_i)$ positive summing to one.
To build such a family, one can eventually use ``atom functions'' and simple operations such as symmetry with respect to $\Y$ and ${\cal S}$, rescaling, translation, rotations with respect to $\X$.
For example if $\X$ is a Banach space, one could define atom functions as, for $y_i\in\Y$
\[
	g_i: x\to \frac{\norm{x}}{1+\norm{x}} \frac{1}{\card{\cal S}} \sum_{s\in{\cal S}} e_s + \frac{1}{1+\norm{x}} e_{\brace{y_i}}.
\]
Those functions could be rescaled and translated as $g_{\sigma, \tau, i}(x) = g_i(\sigma (x-\tau))$, in order to specify a family $(g_{\sigma, \tau, i})$ from few values for $\tau$ and $\sigma$.

The last point is the most difficult one.
Without context variables, and with no-parametrization for $\mu$, a naive unbiased gradient strategy for $\mu$ consists in asking random questions to update the full knowledge of $(\Pbb(Y\in s))_{s\in{\cal S}}$.
But such a strategy will be much worse than our median surrogate technique with queries $\ind{Y\in\brace{y}}$ for $y$ sampled uniformly at random in $\Y$.
Eventually, one should go for a biased gradient strategy, while making sure to update $\mu$ coherently to avoid getting stalled on bad estimates as a result of biases.
	\section{Experimental details}
\label{sgd:app:experiments}

Our experiments are done in {\em Python}. We leverage the {\em C} implementation of high-level array instructions by \cite{Harris2020}, as well as the visualization library of \cite{Hunter2007}.
Randomness in experiments is controlled by choosing explicitly the seed of a pseudo-random number generator.

\begin{figure}[ht!]
	\centering
	\includegraphics{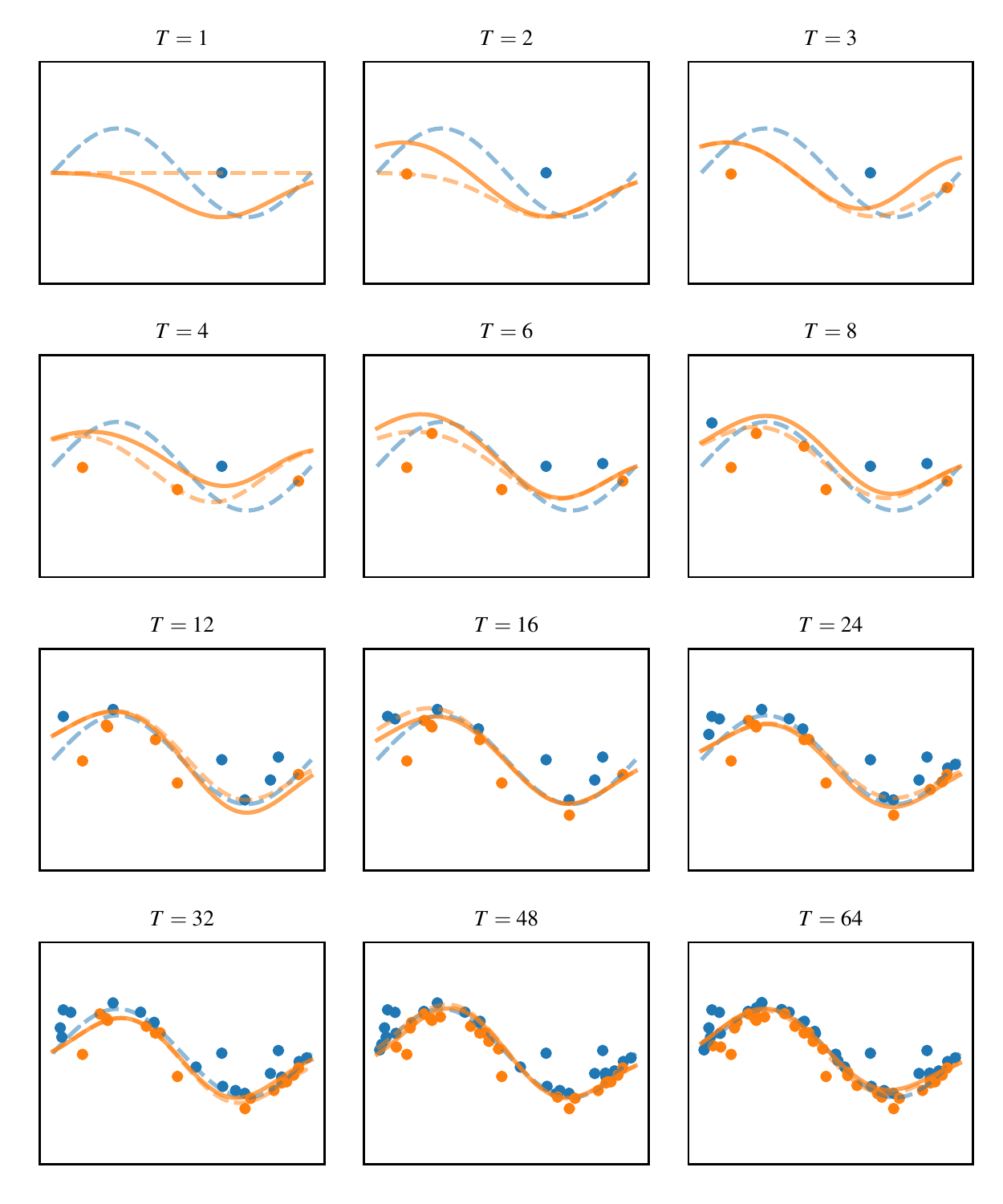}
	\caption{
		{\em Streaming history of the active strategy} to reconstruct the signal in dashed blue in the same setting as Figure \ref{sgd:fig:exp_1}.
		At any time $t$, a point $X_t$ is given to us, our current estimate of $\theta_t$ plotted in dashed orange gives us $z = f_{\theta_t}(X_t)$, and we query $\sign(Y_t - z)$.
		Based on the answer to this query, we update $\theta_t$ to $\theta_{t+1}$ leading to the new estimate of the signal in solid orange.
		In this figure, we see that it might be useful for the practitioners in a streaming setting to reduce the bandwidth of $\phi$ as they advance in time.}
	\label{sgd:fig:exp_1_app}
\end{figure}

\subsection{Comparison with fully supervised SGD}

\begin{figure}[t]
	\centering
	\includegraphics{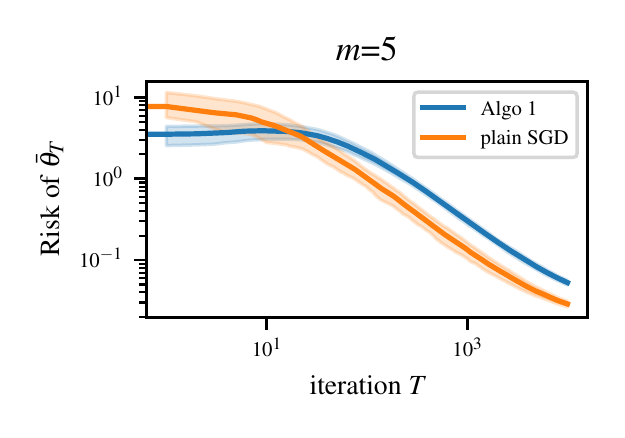}
	\includegraphics{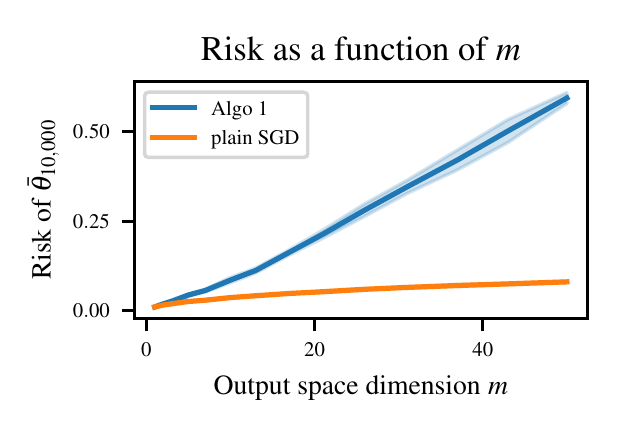}
	\caption{
	{\em Comparison of generalization errors of weakly and fully supervised SGD} as a function of the annotation budget $T$ and output space dimension $m$.
	The setting is similar to Figure \ref{sgd:fig:exp_2}.
	We observe a transitory regime before convergence rates follows the behavior described by Theorem \ref{sgd:thm:sgd}.
	The right side plots the error of both procedures after 10,000 iterations as a function of the output space dimension $m$ between 1 and 50.
	The number of iteration ensures that, for all values of $m\in [50]$, the reported error is well characterized by our theory, in other terms that we have entered the regime described by Theorem \ref{sgd:thm:sgd}.}
	\label{sgd:fig:plain_sgd}
\end{figure}

In this section, we investigate the difference between weakly and fully supervised SGD.
According to Theorem \ref{sgd:thm:sgd}, we only lost a constant factor of order $m^{3/2}$ in our rates compared to fully supervised (or plain) SGD.
This behavior can be checked by adding the plain SGD curve on Figure \ref{sgd:fig:exp_2}. 
On the left side of Figure \ref{sgd:fig:plain_sgd}, we do observe that the risk of both Algorithm \ref{sgd:alg:sgd} and plain SGD decrease with same exponent with respect to number of iteration but with a different constant in front of the rates: that is we observe the same slopes on the logarithm scaled plot, but different intercepts.
Going one step further to check the tightness of our bound, one can plot the intercept, or the error achieved by both Algorithm \ref{sgd:alg:sgd} and plain SGD as a function of the output space dimension $m$. 
The right side of Figure \ref{sgd:fig:plain_sgd} shows evidence that this error grows as $m^\epsilon$ for some $\epsilon \in [1, 3/2]$, which is coherent with our upper bound.
Similarly to Figure \ref{sgd:fig:exp_2}, this figure was computed after cross validation to find the best scaling of the step sizes for each dimension $m$.

\subsection{Passive strategies for classification}
A simple passive strategy for classification based on median surrogate consists in using the active strategy with coordinates sampling, that is $u$ being uniform on $\brace{e_y}_{y\in\Y}$, where $(e_y)_{y\in\Y}$ is the canonical basis of $\R^\Y$ used to define the simplex $\prob{\Y}$ as the convex hull of this basis.
Querying $\ind{\scap{g_\theta(x) - e_y}{e_y} > 0}$ is formally equivalent to the query of $\ind{Y = y}$ when $g_\theta(x) \in \prob{\Y}$.
This is the baseline we plot on Figure \ref{sgd:fig:exp_2}.

\begin{figure}[ht]
	\centering
	\includegraphics{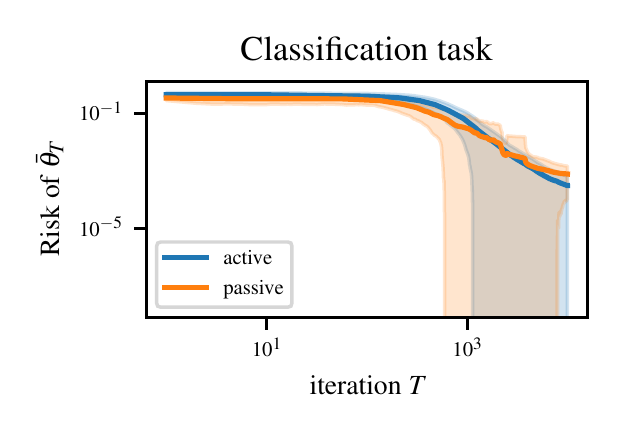}
	\caption{
		{\em Comparison with the infimum loss with better conditioned passive supervision} in a similar setting to Figure \ref{sgd:fig:exp_2} yet with $m=10$, $\epsilon=0$, that is $X$ uniform on $\X$, and $\gamma_0 = 7.5$ for the active strategy and $\gamma_0 = 15$ for the passive strategy.
		We see no major differences between the active strategy based on the median surrogate and the passive strategy based on the median surrogate with the infimum loss.
		Note that the standard deviation is sometimes bigger than the average of the excess of risk, explaining the dive of the dark area on this logarithmic-scaled plot.
	}
	\label{sgd:fig:exp_2_app}
\end{figure}

A more advanced passive baseline is provided by the infimum loss \citep{Cour2011,Cabannes2020}.
It consists in solving
\[
	\argmin_{f:\X\to\Y}{\cal R}_I(f):=\E_{(X, Y)\sim\rho}\E_{S}\bracket{L(f(X), S, \ind{Y\in S})},
\]
where $S$ is a random subset of $\Y$ and $L$ is defined from the original loss $\ell:\Y\times\Y\to\R$ as, for $z\in\Y$, $s\subset \Y$ and $y\in\Y$,
\[
	L(z, s, \ind{y\in s}) = \left\{\begin{array}{cl} \inf_{y'\in s} \ell(z, y') & \text{if } y\in s \\ \inf_{y'\notin s} \ell(z, y') & \text{otherwise.}\end{array}\right.
\]
Random subsets $S$ could be generated by making sure that the variable $(y \in S)_{y\in\Y}$ are independent balanced Bernoulli variables; and by removing the trivial sets $S = \emptyset$ and $S = \Y$ from the subsequent distribution.
In order to optimize this risk in practice, one can use a parametric model and a surrogate differentiable loss together with stochastic gradient descent on the empirical risk.
For classification with the 0-1 loss, we can reuse the surrogate introduced in Proposition \ref{sgd:prop:sur} and minimize, assuming that we always observed $\ind{Y_i\in S_i} = 1$ for simplicity,
\[
	\hat{\cal R}_{I, S}(\theta) = \sum_{i=1}^n \inf_{y \in S_i} \norm{g_\theta(X_i) - e_y}.
\]
Stochastic gradients are then given by, assuming ties have no probability to happen,
\[
	\nabla_\theta \inf_{y \in S_t} \norm{g_\theta(X_t) - e_y}
	= \paren{\frac{g_\theta(X_t) - e_{y^*}}{\norm{g_\theta(X_t) - e_{y^*}}}}^\top D g_\theta(X_t)
	\quad\text{with}\quad y^* := \argmax_{y\in S_t} \scap{g_\theta(X_t)}{e_y}.
\]
This gives a good passive baseline to compare our active strategy with.
In our experiments with the Gaussian kernel, see Figure \ref{sgd:fig:exp_2_app} for an example, we witness that this baseline is highly competitive.
Although we find that it is slightly harder to properly tune the step size for SGD, and that the need to compute an argmax for each gradient slows-down the computations.

\subsection{Real-world classification datasets}

\begin{figure}[ht]
	\centering
	\includegraphics{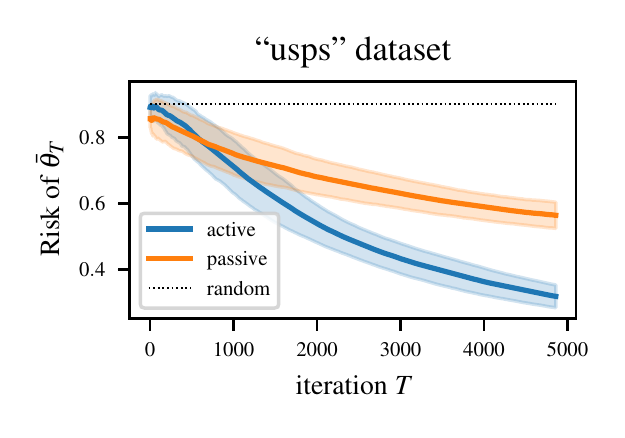}
	\includegraphics{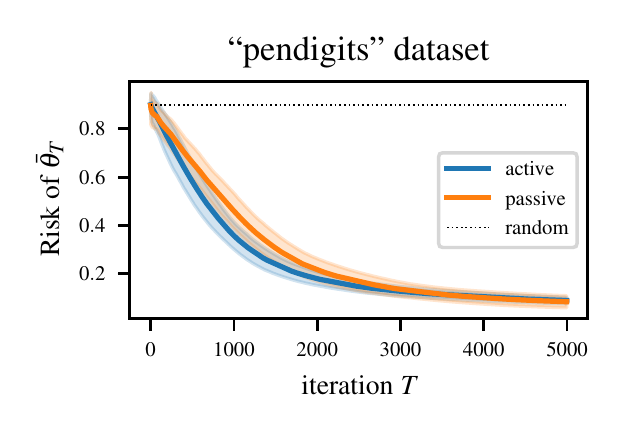}
	\caption{
		{\em Testing errors on two LIBSVM datasets} with a similar setting to Figure \ref{sgd:fig:exp_2_app}.
		Those empirical errors are reported after averaging over 100 different splits of the datasets.
		The step size parameter was optimized visually, which led to $\gamma_0=15$ for the active strategy on ``USPS'', $\gamma_0=60$ for the passive one, $\gamma_0=7.5$ for the active strategy on ``pen digits'', $\gamma_0=30$ for the passive one.
		The dotted line represents ${\cal R} = 1 - m^{-1}$ which is the performance of a random model.
	}
	\label{sgd:fig:exp_libsvm}
\end{figure}

In Figure \ref{sgd:fig:exp_libsvm}, we compare the ``well-conditioned'' passive baseline with our active strategy on the real-world problems of LIBSVM \citep{Chang2011}.
We choose the ``USPS'' and ``pen digits'' datasets as they contain $m=10$ classes each with $n=7291$ and $n=7494$ samples respectively, with $d=50$ and $d=16$ features each.
We have chosen those datasets as they present enough classes that leads to many different sets $S$ to query, and they are made of the right number of samples to do some experiments on a laptop without the need for ``advanced'' computational techniques such as caching or low-rank approximation \citep{Meanti2020}.
On Figure \ref{sgd:fig:exp_libsvm}, we use the same linear model as for Figure \ref{sgd:fig:exp_2}, that is a Gaussian kernel.
We choose the bandwidth to be $\sigma = d/5$, and we normalize the features beforehand to make sure that they are all centered with unit variance.
We report error by taking two thirds of the samples for training and one third for testing, and averaging over one hundred different ways of splitting the datasets.
We observe that the active strategy leads to important gains on the ``USPS'' dataset, yet is not that useful for the ``pen digits'' dataset.
We have not dug in to understand those two different behaviors.

\subsection{Real-world regression dataset \& Nystr\"om method}

In this section, we provide two experiments on real-world datasets.

In order to deal with big regression datasets, it is useful to approximate the parameter space $\Y\otimes{\cal H}$ in Assumption \ref{sgd:ass:source} with a small dimensional space.
To do so, let us remark that given samples $(X_i)_{i\leq n} \in \X^n$ for $n\in\N$, we know that our estimate $f_{\theta_n}$ can be represented as
\[
	f_{\theta_n}(\cdot) = \sum_{i\leq n}\sum_{j\leq m} a_{ij} \scap{\phi(x_i)}{\phi(\cdot)} e_j,
\]
for some $(a_{ij}) \in \R^{p\times m}$ and where $(e_j)_{j\leq m}$ is the canonical basis of $\Y = \R^m$.
For large datasets, that is when $n$ is large, it is smart to approximate this representation through the parameterization
\[
	f_{a}(x) = \sum_{i\leq p}\sum_{j\leq m}a_{ij} k(x, x_i) e_j,
\]
where $p\leq n$ is the rank of our approximation, and $k$ is the kernel defined as $k(x, x') = \scap{\phi(x)}{\phi(x')}$.
Stated with words, we only use a small number $p$, instead of $n$, of vectors $\phi(x_i)$ to parameterize $f$.
This allows to only keep a matrix of size $p\times m$ in memory instead of $n\times m$, while not fundamentally changing the statistical guarantee of the method \citep{Rudi2015}.
In this setting, the stochastic gradients are specified from the fact that 
\[
	u^\top D_a f_a(x) = (u_j k(x, x_i))_{i,j} \in \R^{p\times m}.
\]
In other terms, in order to update the parameter $a$ with respect to the observation made at $(x, u)$, we check how much each coordinate of $a$ determines the value of $u^\top f_a(x)$.

In the following, we experiment with two real-world datasets.
In order to learn the relation between inputs and outputs, we use a Gaussian kernel after normalizing input features so that each of them has zero mean and unit variance.
To keep computational cost, we sample $p$ random (Nystr\"om) representers among the training inputs which are used to parameterize functions.
To avoid overfitting, we add a small regularization to the empirical objective. It reads $\lambda \norm{\theta}_{{\cal H}}^2$ with our notations and corresponds to the Hilbertian norm inherited from the reproducing kernel $k$ of the function $f_{\theta}$ \citep{Scholkopf2001}.

\begin{figure}[h]
	\centering
	\includegraphics{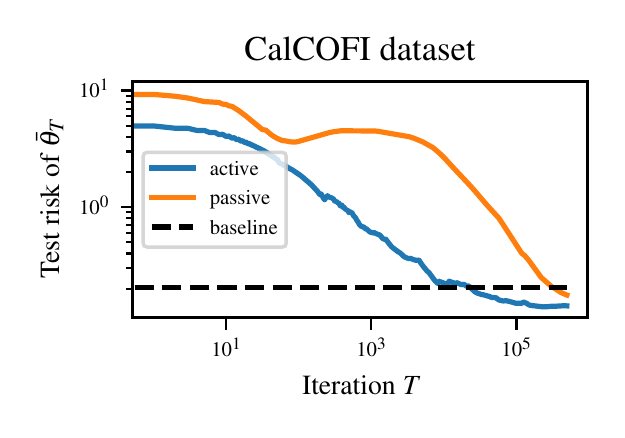}
	\includegraphics{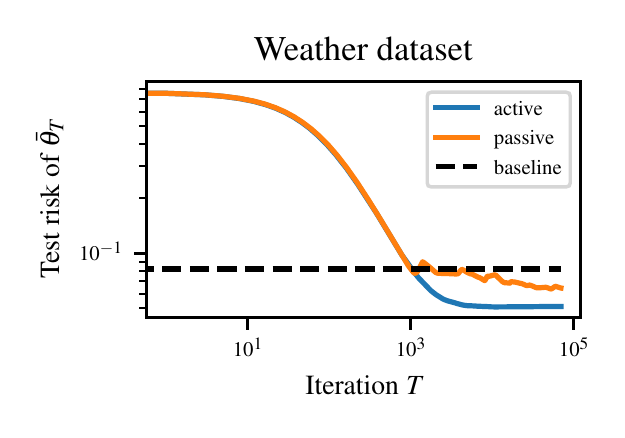}
	\caption{
	{\em Testing error on two real-world regression datasets.}
	On both datasets, a single pass was made through the data in a chronological fashion, and errors were computed from the 26,453 most recent data samples for the ``Weather'' dataset, and from a random sample of 10,000 samples among the 155,140 most recent samples for the ``CalCOFI'' dataset.}
	\label{sgd:fig:real_world}
\end{figure}

Our first experiment is based on the data collected by the California Cooperative Oceanic Fisheries Investigation between March 1949 and November 2016.
It consists of more than 800,000 seawater samples including measurements of nutriments (set aside in our experiments) together with pressure, temperature, salinity, water density, dynamic height (providing five input parameters), as well as dissolved oxygen, and oxygen saturation (the two outputs we would like to predict).
We assume that we can measure if any weighted sum of oxygen concentration and saturation is above a threshold by letting some population of bacteria evolves in the water sample and checking if it survives after a day.
If the measurements are done on the day of the sample collection, this setting exactly fits in the streaming active labeling framework.
After cleaning the dataset for missing values, the dataset contains 655,140 samples.
The ``CalCOFI'' dataset results are reported on the left of Figure \ref{sgd:fig:real_world}, parameters were chosen as $p=100$, $\sigma = 10$, $\lambda = 10^{-6}$ and $\gamma_0 = 1$.
For the passive strategy, random queries were chosen to follow a normal distribution with the same mean as the targets and one third of their standard deviation ({\em i.e.} we ask if the apparent temperature is lower than the usual one plus or minus a perturbation).
The plotted baseline corresponds to linear regression performed over the entire dataset.
It takes about 10,000 samples for our active strategy to be competitive with this baseline, and 200,000 samples for the passive one.

The second experiment makes use of data collected through the Dark Sky API (which is now part of Apple WeatherKit). 
It is made of 96,454 weather summaries between 2006 and 2016 in the city of Szeged, Hungary.
Our task consists in computing the apparent temperature from real temperature, humidity, wind speed, wind bearing, visibility and pressure.
The apparent temperature is an index that searches to quantify the subjective feeling of heat that humans perceive, it is expressed on the same scale as real temperature.
One way to measure it would be to ask some humans if the outside is hotter or colder than a controlled room with a specific temperature and neutral meteorological conditions.
Once again, this exactly fits into our streaming active labeling setting.
The ``Weather'' dataset results are reported on the right of Figure \ref{sgd:fig:real_world}.
The baseline consists in predicting the apparent temperature as the real temperature.
We observe a transitory regime where the first 1,000 samples seem to be used to calibrate the weights $\alpha$.
During this regime, our estimate is too bad for the active strategy to make smarter queries than the ``random'' ones that have been calibrated on temperature statistics.
The main difference in the learning dynamic between the active and passive strategies is observed on the remaining 69,000 training samples.
The parameters were the same as the ``CalCOFI'' dataset but for $\gamma_0 = 10^{-2}$.

\end{subappendices}

\chapter*{Conclusion}
\pagestyle{plain}
\addcontentsline{toc}{part}{Conclusion}
In this thesis, we have approached weakly supervised learning through partial supervision.
We have leveraged the algorithm of \cite{Ciliberto2016} to provide consistent estimators once we had formulated the problem through the infimum loss, which was the focus of \cite{Cabannes2020}.
By characterizing implicit disambiguation due to the infimum loss, we have revisited and generalized the approach of \cite{Bach2007} to any type of discrete output problems in \cite{Cabannes2021}.
Such a disambiguation strategy has the advantage of working with a smaller surrogate space (working on functions from $\X$ to $\R^\Y$ rather than from $\X$ to $\R^{2^\Y}$), hence reducing the variance of our estimates, and making approximation assumptions more comprehensible.
Regarding optimization, our min-min formulations are hard to tackle in a generic fashion; while one can reuse the relaxation in \cite{Bach2007}, our implementations were based on alternate minimization with good starting points; readers and practitioners eager to come up with different strategies might refer to literature on bilevel optimization for non-convex problems.
In \cite{Cabannes2021c}, advocating for the leverage of unlabeled data, we have brought up the approach of \cite{Zhu2003} to the realm of kernel methods;
and to ensure computational efficiency, we have adapted the low-rank approximation of \cite{Rudi2015} and its proof to our specific setting with derivatives.
We hope to see future works pushing forwards the usage of kernel methods with derivatives, could it be for sampling through Langevin dynamics \citep[in the spirit of][]{PillaudVivien2020} or for penalties inducing sparsity \citep[in the spirit of][]{Rosasco2013,Follain2022}.
Providing efficient and user-friendly open-source implementations would arguably foster the diffusion of kernels with derivatives in the research community.

In order to help the practitioner collecting data, we finally set ourselves before the data collection process, and introduced the ``active labeling'' problem.
Following the observation that one does not need to access full information in order to build stochastic gradients, we provided a first solution to this problem based on stochastic gradient descent in \cite{Cabannes2022} that is naturally adapted to streaming settings.
In a near future, we would like to focus on exploiting the discrete-output structure of classification problems more subtly than surrogate methods, notably with ideas steaming from the ``bandit'' literature.
In particular, we will explore how entropic coding could help to tackle the active labeling problem.
Beside this project, the ``active'' setup opens up new possibilities, {\em e.g.}, to detach ourselves from min-min formulations and try to come up with well-grounded min-max formulations that are easier to optimize, or, on a completely different level, to deal with privacy constraints.

While working on the proof of \cite{Cabannes2021}, we discovered that usual rates of convergence on discrete output learning problems are often suboptimal thanks to the work of \cite{Audibert2007}.
In substance, if you have $L^\infty$-exponential concentration inequality, and margin conditions, then you can have up to exponential convergence rates.
We studied the question in more detail in \cite{Cabannes2021b} and \cite{Cabannes2022b}.
Those derivations can still be pushed further, could it be to better understand the interplay between estimation error (a.k.a. variance) and approximation error (a.k.a. bias) without decoupling them, or by adapting the proof structure with concentration on other quantities than suprema and corresponding relative hardness conditions.
Regarding statistical learning, a project of interest to pursue, yet which require other tools than the ones presented in the thesis, is to approach learning theory without thinking in terms of functions classes, but by thinking directly with measures, especially at the light of the disambiguation principles that we expressed directly in the space of measures.
Interestingly, such an approach might help to exchange (if not unify) ideas between works on weakly supervised learning ({\em i.e.} missing output data) and works on missing (input) data.

Finally, while this thesis was more of theoretical nature, in order to maximize my ``tangible impact'' as a researcher, I would like to look at issues that are closer to ``production'' and focus more on real-world experiments.
I hope that my future postdoc position at FAIR New York will provide me with such opportunities, for example by exposing myself to ongoing experimental works with self-supervised learning.

\cleardoublepage
\phantomsection
\addcontentsline{toc}{chapter}{Bibliography}
\bibliography{main}

\begin{thebibliography}{312}
\bibitem[Whittaker(1915)]{Whittaker1915}
Edmund Whittaker.
\newblock On the functions which are represented by the expansions of the
  interpolation theory.
\newblock \emph{Proceedings of the Royal Society of Edinburgh}, 35:\penalty0
  181--194, 1915.
\end{thebibliography}

\end{document}